\RequirePackage{fix-cm}
\documentclass[
11pt, 
english, 
singlespacing, 
headsepline, 
]{MastersDoctoralThesis} 
\usepackage{babel}
\usepackage{multirow} 

\usepackage[utf8]{inputenc} 
\usepackage[T1]{fontenc} 
\usepackage{mathpazo} 

\usepackage[backend=bibtex,style=authoryear,natbib=true,backref=true]{biblatex} 

\usepackage{textpos}
\usepackage{tikz}
\usetikzlibrary{quotes,arrows.meta,positioning,decorations.pathreplacing,calc,3d,arrows}

\usepackage{tabularx}
\usepackage{mathtools}

\usepackage{xcolor}
\usepackage{transparent}
\usepackage[autostyle=true]{csquotes} 

\usepackage{listings}
\usepackage{color}
\usepackage{lscape} 
\usepackage{xparse}
\usepackage{xstring}

\definecolor{forestgreen(web)}{rgb}{0.13, 0.55, 0.13}

\newcommand{\mycolorbox}[2]{%
    \StrBefore{#1}{,}[\rval]%
    \StrBehind{#1}{,}[\tempa]%
    \StrBefore{\tempa}{,}[\gval]%
    \StrBehind{\tempa}{,}[\tempb]%
    \StrBefore{\tempb}{,}[\bval]%
    \StrBehind{\tempb}{,}[\aval]%
    \tikz[baseline=(X.base)] \node[fill={rgb,255:red,\rval;green,\gval;blue,\bval}, fill opacity=\aval, text opacity=1, rectangle, rounded corners=2pt] (X) {#2};%
}

\usepackage{verbatim}
\usepackage{tabularray}

\usepackage[toc]{appendix}
\let\cite\parencite
\usepackage{lmodern}
\PassOptionsToPackage{hyphens}{url}
\usepackage[colorlinks=true,urlcolor=blue,linkcolor=red,citecolor=blue]{hyperref}

\usepackage[normalem]{ulem}
\useunder{\uline}{\ul}{}

\usepackage{textcomp}
\usepackage{amssymb}
\usepackage{amsmath}
\usepackage{titlesec}
\titleformat{\paragraph}
{\normalfont\small\bfseries}{\theparagraph}{1em}{}
\titlespacing*{\paragraph}
{0pt}{3.25ex plus 1ex minus .2ex}{1.5ex plus .2ex}

\usepackage{algcompatible}
\usepackage{float}
\usepackage{caption}
\usepackage{subcaption}
\usepackage{varwidth}
\usepackage{dsfont}
\usepackage{algorithmicx}
\usepackage[hang,flushmargin]{footmisc}
\usepackage{indentfirst}
\usepackage{adjustbox}
\usepackage{rotating}
\usepackage[graphicx]{realboxes}
\usepackage{titlesec}
\usepackage{algorithm}

\usepackage{thmtools}

\definecolor{myblue}{HTML}{55caff}
\definecolor{mygreen}{HTML}{55d555}
\definecolor{myorange}{HTML}{fab771}
\definecolor{greenF2}{HTML}{00c41e}
\definecolor{redF2}{HTML}{FF5959}
\definecolor{blueF2}{HTML}{7b84f7}
\definecolor{orangeF2}{HTML}{ffa000}
\definecolor{magentaF2}{HTML}{FF7FFF}
\definecolor{yellowF2}{HTML}{ffdd37}
\definecolor{oilF2}{HTML}{d2dd37}
\definecolor{purpleF2}{HTML}{E0B0FF}
\definecolor{cyanF2}{HTML}{00d4f7}
\definecolor{green2F2}{HTML}{00e884}

\titleformat{\paragraph}[runin]{\normalfont\normalsize\bfseries}{\theparagraph}{1em}{}[]


\makeatletter
\def\BState{\State\hskip-\ALG@thistlm}
\makeatother

\addto\captionsfrench{\renewcommand{\abstractname}{Abstract}}
\addto\captionsfrench{\renewcommand{\acknowledgementname}{Remerciements}}

\thesistitle{Deep Learning For Time Series Analysis With Application On Human Motion}
\supervisor{
    Pr. Michel \textsc{Hassenforder}\\
    Pr. Pierre-Alain \textsc{Muller}\\
    Pr. Germain \textsc{Forestier}\\
    Dr. Jonathan \textsc{Weber}
} 
\advisor{
} 
\examiner{} 
\degree{Doctor of Philosophy} 
\author{Ali El Hadi \textsc{ISMAIL FAWAZ}} 
\addresses{} 

\subject{} 
\university{Université De Haute-Alsace} 
\department{{Institut de Recherche en Informatique, Mathématiques, Automatique et Signal}} 
\group{MSD} 
\faculty{Faculté des Sciences et Techniques} 

\AtBeginDocument{
\hypersetup{pdftitle=\ttitle} 
\hypersetup{pdfauthor=\authorname} 
\hypersetup{pdfkeywords=\keywordnames} 
}

\newcommand{\pr}{$\mathcal{P}_{\mathcal{R}}$}
\newcommand{\pg}{$\mathcal{P}_{\mathcal{G}}$}
\newtheorem{theorem}{Theorem}

\begin{document}

\newcounter{definition}[section]
\newenvironment{definition}[1][]{\refstepcounter{definition}\par\medskip
    \noindent \textbf{Definition~\thedefinition. #1} \rmfamily}{\medskip}

\sloppy 
\frontmatter 

\pagestyle{plain} 


\begin{titlepage}
\begin{textblock}{}(0.03,0.29)
\begin{figure}
    \centering
    \includegraphics[width=3.5\linewidth]{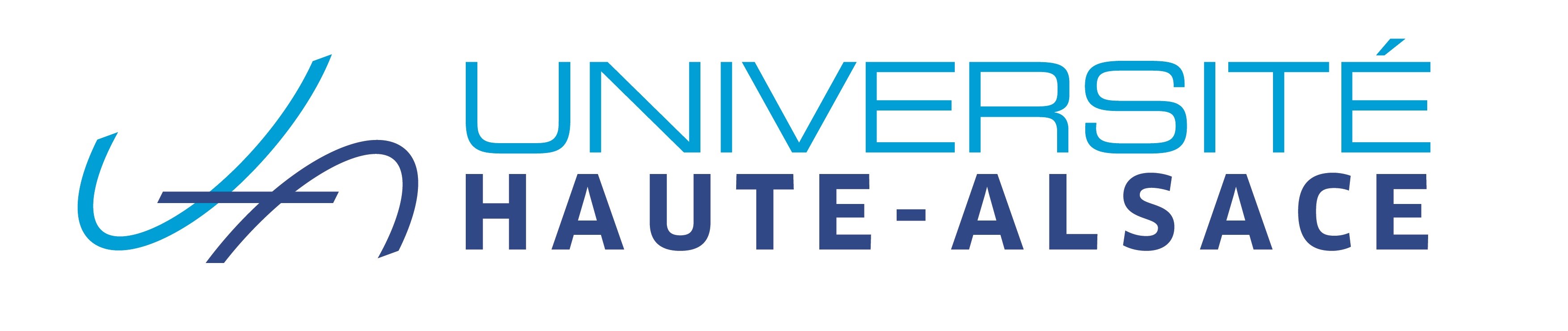}
\end{figure}
\end{textblock}
\begin{textblock}{}(8.5,0.29)
\begin{figure}
    \centering
    \includegraphics[width=1.5\linewidth]{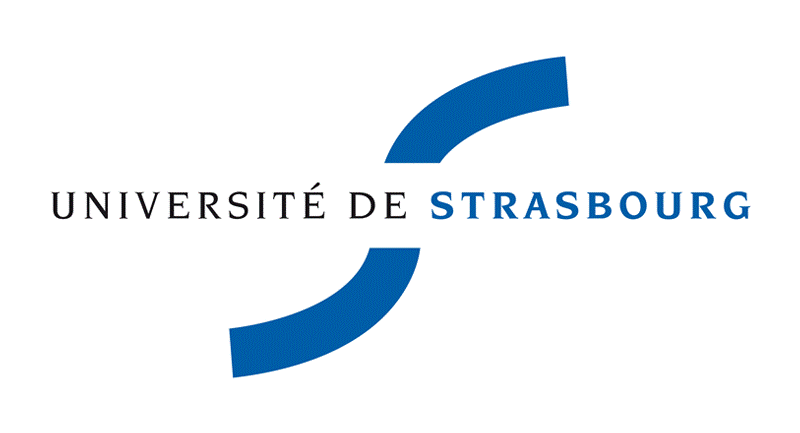}
\end{figure}
\end{textblock}
\begin{textblock}{2.0}(0.03,0.9)
Année 2024
\end{textblock}

\begin{textblock}{5.0}(5.5,0.9)
N$^\circ$  d’ordre : (attribué par le SCD)
\end{textblock}

\begin{center}

\vspace*{.09\textheight}
{\scshape\LARGE \univname}\\ 
{\scshape\small Université De Strasbourg}\vspace{0.5cm} \\
\textsc{\LARGE \textbf{Thèse}}\\[0.5cm] 
{\small Pour l'obtention du grade de \\ 
\textbf{Docteur de l'Université de Haute-Alsace}\vspace{0.4cm} \\
\textbf{École Doctorale: Mathématiques, Sciences de l'Information et de l'Ingénieur (ED 269)}
Discipline : Informatique}\vspace{0.5cm} \\

{\small Présentée et soutenue publiquement \vspace{0.2cm} \\
par \vspace{0.2cm} \\ 
\LARGE{\authorname} \vspace{0.4cm} \\ 
\small The 8th of January 2024\\ \vspace{1.4cm}
}

\HRule \\[0.4cm] 
{\huge \bfseries \ttitle\par}\vspace{0.4cm} 
\HRule \\[1.5cm] 


{\normalsize Jury} \vspace{-0.2cm}

\begin{flushleft}
{\normalsize
\begin{table}[h!]
    \begin{tabular}{@{}ll}
	Rapporteur & Prof. Pierre-François MARTEAU, Université de Bretagne Sud\\
	Rapporteur & Prof. Laurent OUDRE, Ecole Normale Supérieure Paris-Saclay\\
	Examinatrice & Dr. HDR Latifa OUKHELLOU, Université Gustave Eiffel \\
	Examinatrice & Dr. Georgiana IFRIM, University College Dublin\\
	Directeur de thèse & Prof. Germain FORESTIER, Université de Haute-Alsace\\
	co-Directeur de thèse & Prof. Jonathan WEBER, Université de Haute-Alsace \\
	co-Directeur de thèse & Dr. Maxime DEVANNE, Université de Haute-Alsace\\
	Co-Encadrant de thèse & Dr. HDR Stefano BERRETTI, University of Florence \\
    \end{tabular}
\end{table}
}
\end{flushleft}

%
%
%
%
\end{center}
\end{titlepage}

\newpage

\begin{extraAbstract}
\addchaptertocentry{\resumename} 
Les séries temporelles, définies par des points de données espacés 
régulièrement dans le temps, sont essentielles dans divers domaines, 
notamment la médecine, les télécommunications et l'énergie.
L'analyse des séries temporelles consiste à extraire des informations, 
des modèles et des tendances pour répondre à diverses tâches.
La classification peut identifier des individus ayant des mouvements 
normaux ou anormaux en fonction de leurs séquences de mouvements basées sur le squelette.
Les tâches de régression incluent la prédiction de la progression de 
la récupération d'un patient en fonction de ses mouvements.
Le clustering peut analyser des données boursières pour détecter des 
comportements similaires entre actions.
Le prototypage, utilisé dans les exercices de rééducation, augmente 
la quantité de données pour les chercheurs.

Étant donné les dépendances temporelles, développer des algorithmes efficaces 
pour l'analyse de séries temporelles nécessite de considérer ces relations.
Ces dernières années, de nombreuses approches ont émergé pour résoudre diverses 
tâches d'analyse de séries temporelles.
Elles incluent des méthodes basées sur la distance pour la classification, 
l'extraction de caractéristiques temporelles pour la régression et l'analyse 
de formes pour le clustering.
Récemment, l'apprentissage profond, grâce à son succès dans des domaines comme 
le traitement du langage naturel et la classification d'images, s'est montré 
efficace pour l'analyse de séries temporelles.

Cette thèse aborde plusieurs tâches d'analyse de séries temporelles en 
utilisant des techniques d'apprentissage profond.
Nous contribuons à la classification en améliorant les modèles profonds 
avec une ingénierie avancée des caractéristiques, en proposant des modèles 
de base comme points de départ et en développant une architecture novatrice, 
le plus petit réseau atteignant des performances de pointe.
Nous traitons également le manque de données annotées avec des approches 
d'apprentissage auto-supervisé.

Nos contributions sont mises en valeur à travers des applications réelles, 
notamment dans l'analyse de séquences de mouvements humains pour la reconnaissance 
d'actions et la rééducation.
Nous avons développé un modèle génératif pour les données de mouvements humains, 
utile pour des applications telles que la production cinématographique et le 
développement de jeux.
De plus, dans le domaine du prototypage de séries temporelles, nous proposons
une méthode pour générer des échantillons synthétiques via une analyse basée 
sur les formes, augmentant ainsi les données disponibles pour entraîner les 
modèles d'apprentissage profond pour les tâches de régression lorsque la 
collecte de données est coûteuse.

Enfin, nous entreprenons une évaluation approfondie des modèles discriminatifs 
et génératifs, en discutant les limitations des méthodologies d'évaluation actuelles.
Nous plaidons pour l'établissement d'un processus d'évaluation unifié et 
équitable, robuste et résistant à la manipulation.
À travers des expériences approfondies sur des données publiques, nous visons 
à faire progresser le domaine de l'analyse de séries temporelles en apportant 
de nouveaux méthodologies, démontrant leur impact significatif 
sur des applications pratiques.

\end{extraAbstract}

\begin{abstract}
\addchaptertocentry{\abstractname} 
Time series data, defined by equally spaced data points over time, is fundamental across various domains, 
notably in medical, telecommunication and energy etc.
The analysis and mining of time series data 
involve extracting meaningful information, patterns, and trends to address various tasks.
Classification can include identifying individuals with normal or abnormal movement based on their skeleton-based motion sequences.
Clustering applications might involve analyzing stock market data to find similar behavior patterns among stocks.
Prototyping is used in physical therapy exercises to expand the available data for researchers.
Regression tasks include predicting the progress of a patient's recovery based on their movement data.
Each of these applications highlights the diverse and critical uses of time series analysis.

Given the inherent temporal dependencies, developing effective time series analysis algorithms requires careful consideration of these temporal relationships.
Over the past decade, numerous approaches have emerged to tackle different time series analysis tasks. 
These include for example distance-based methods for classification, temporal feature extraction for regression and shape 
analysis for clustering. Recently, deep learning has gained significant attention due to its sucess in other 
fields like natural language processing and image classification, and proved to be effective for time series analysis.

This thesis aims to address multiple time series analysis tasks using deep learning techniques. We contribute 
to the field of classification by enhancing deep models with advanced feature engineering, proposing foundation 
models to be used as starting points, and developing a novel architecture that remains the smallest 
network achieving state-of-the-art performance. Additionally, we tackle the issue of limited labeled data with 
self-supervised learning approaches.

Our contributions are showcased through real-world applications, particularly in analyzing human motion sequences 
for action recognition and physical rehabilitation. We developed a generative model for human motion data, valuable
for applications such as cinematic production and game development. Additionally, in time series prototyping,
we propose a method to generate synthetic 
samples via shape-based analysis, expanding the data available for training deep learning models for regression 
tasks when data collection is costly.

Finally, we undertake a thorough evaluation of both discriminative and generative models, shedding light on 
the limitations of current evaluation methodologies. We advocate for the establishment of a unified and 
fair evaluation process that is robust and resistant to manipulation. Through extensive experiments on public 
data, we aim to advance the field of time series analysis by providing novel insights and methodologies, 
demonstrating their significant impact on practical applications.

\end{abstract}


\begin{acknowledgements}
\addchaptertocentry{\acknowledgementname}

This work would not have been possible without the invaluable contributions and 
support of many individuals, whom I would like to sincerely thank here.

First and foremost, I would like to express my deepest gratitude to my 
supervisors, Dr. Maxime Devanne, Dr. Stefano Berretti, Prof. Jonathan Weber, 
and Prof. Germain Forestier. Their supervision, guidance, and encouragement 
over the past three years have shaped not only this thesis but also my growth 
as a researcher. The experience and knowledge I have gained through their 
mentorship have been truly enriching.

I would like to express my gratitude  to Prof. Pierre-François Marteau and Prof. Laurent Oudre 
for accepting to review my manuscript, and to Dr. Latifa Oukhellou and 
Dr. Georgiana Ifrim for serving as members of the jury. Your 
time and expertise in evaluating this work are greatly appreciated.

I would like to thank Dr. Stefano Berretti for hosting me at 
the MICC Institute at the University of Florence for an academic research 
visit. This collaboration was an invaluable opportunity, and I deeply 
appreciate the hospitality and support.

I would like to thank Prof. Geoffrey I. Webb for hosting me at 
Monash University for a two-week academic research visit. The discussions and 
collaboration we shared were instrumental to my growth, and I am grateful for 
the opportunity to gain new perspectives and insights.

I would like to thank Dr. Patrick Schäfer for hosting me at Humboldt-University 
in Berlin and for inviting me to give a seminar on my PhD work.

I would like to thank Dr. François Petitjean and Dr. Hassan Ismail Fawaz 
for the many discussions we had on time series analysis.

I would like to thank Prof. Anthony Bagnall for nominating me as a core 
developer in the aeon toolkit, which allowed me to advance 
my research and promote my work.

I would also like to express my gratitude to the Mésocentre of Strasbourg for 
providing essential computational resources, including access to the GPU cluster, 
which was crucial for conducting my experiments.

I am thankful to the Agence Nationale de la Recherche for funding my PhD 
through the DELEGATION ANR project (grant NR-21-CE23-0014), without which this 
research would not have been possible.

Last but certainly not least, I would like to thank all of my colleagues 
at the Université de Haute-Alsace. Your contributions, both professionally 
and personally, have been invaluable throughout this journey.

Finally, I want to express my deepest gratitude to my parents, brothers, 
and friends for their unwavering support over the past three years and in 
all the years leading up to this point. Your encouragement and belief in me 
have been my constant source of strength.

\end{acknowledgements}


\tableofcontents 

\listoffigures 

\listoftables 

\listoftheorems[ignoreall,show={theorem,defn}]


\begin{abbreviations}{ll}
\textbf{ECG} & \textit{Electrocardiogram}\\
\textbf{EEG} & \textit{Electroencephalogram}\\
\textbf{TSC} & \textit{Time Series Classification}\\
\textbf{TSCL} & \textit{Time Series Clustering}\\
\textbf{TSP} & \textit{Time Series Prototyping}\\
\textbf{TSER} & \textit{Time Series Extrinsic Regression}\\
\textbf{UTS} & \textit{Univariate Time Series}\\
\textbf{MTS} & \textit{Multivariate Time Series}\\
\textbf{NN} & \textit{Nearest Neighbor}\\
\textbf{DTW} & \textit{Dynamic Time Warping}\\
\textbf{ED} & \textit{Euclidean Distance}\\
\textbf{ShapeDTW} & \textit{Shape Dynamic Time Warping}\\
\textbf{SoftDTW} & \textit{Soft Dynamic Time Warping}\\
\textbf{EE} & \textit{Elastic Ensemble}\\
\textbf{PF} & \textit{Proximity Forest}\\
\textbf{RF} & \textit{Random Forest}\\
\textbf{ADTW} & \textit{Amerced Dynamic Time Warping}\\
\textbf{hctsa} & \textit{highly comparative time-series analysis}\\
\textbf{Catch22} & \textit{Canonical Time Series Characteristics}\\
\textbf{TSFresh} & \textit{Time Series Feature Extraction based on Scalable Hypothesis Tests}\\
\textbf{RotF} & \textit{Rotation Forest}\\
\textbf{CNN} & \textit{Convolutional Neural Network}\\
\textbf{ROCKET} & \textit{RandOm Convolutional KErnel Transform}\\
\textbf{PPV} & \textit{Proportion of Positive Values}\\
\textbf{MPV} & \textit{Mean of Positive Values}\\
\textbf{MIPV} & \textit{Mean of Indices of Positive Values}\\
\textbf{LSPV} & \textit{Longest Stretch of Positive Values}\\
\textbf{HYDRA} & \textit{HYbrid Dictionary-Rocket Architecture}\\
\textbf{STC} & \textit{Shapelet Transform Classifier}\\
\textbf{RDST} & \textit{Random Dilated Shapelet Transform}\\
\textbf{SAX} & \textit{Symbolic Aggregate approXimation}\\
\textbf{PAA} & \textit{Piecewise Aggregate Approximation}\\
\textbf{VSM} & \textit{Vector Space Model}\\
\textbf{SFA} & \textit{Symbolic Fourier Approximation}\\
\textbf{MCB} & \textit{Multiple Coefficient Binning}\\
\textbf{BOSS} & \textit{Bag-of-SFA-Symbols}\\
\textbf{WEASEL} & \textit{Word Extraction for Time Series Classification}\\
\textbf{TSF} & \textit{Time Series Forest}\\
\textbf{CIF} & \textit{Canonical Interval Forest}\\
\textbf{DrCIF} & \textit{Diverse Representation Canonical Interval Forest}\\
\textbf{TS-CHIEF} & \textit{Time Series Combination of Heterogeneous and Integrated Embedding Forest}\\
\textbf{RISE} & \textit{Random Interval Spectral Ensemble}\\
\textbf{COTE} & \textit{Collective Of Transformation-based Ensemble}\\
\textbf{CAWPE} & \textit{Cross-validation Accuracy Weighted Probabilistic Ensemble}\\
\textbf{HIVE-COTE} & \textit{HIerarchical VotE Collective Of Transformation Ensemble}\\
\textbf{HC} & \textit{HIVE-COTE}\\
\textbf{RNN} & \textit{Recurrent Neural Network}\\
\textbf{FC} & \textit{Fully Connected}\\
\textbf{SGD} & \textit{Stochastic Gradient Descent}\\
\textbf{SC} & \textit{Standard Convolution}\\
\textbf{DWSC} & \textit{DepthWise Separable Convolution}\\
\textbf{DWC} & \textit{DepthWise Convolution}\\
\textbf{PWC} & \textit{PointWise Convolution}\\
\textbf{BN} & \textit{Batch Normalization}\\
\textbf{GMP} & \textit{Global Max Pooling}\\
\textbf{GAP} & \textit{Global Average Pooling}\\
\textbf{NLP} & \textit{Natural Language Processing}\\
\textbf{PE} & \textit{Positional Encoder}\\
\textbf{APE} & \textit{Absolute Positional Encoding}\\
\textbf{LSTM} & \textit{Long Short-Term Memory}\\
\textbf{GRU} & \textit{Gated Recurrent Unit}\\
\textbf{MLP} & \textit{MultiLayer Perceptron}\\
\textbf{TimeCNN} & \textit{Time Convolutional Neural Network}\\
\textbf{FCN} & \textit{Fully Convolutional Network}\\
\textbf{ResNet} & \textit{Residual Netwrok}\\
\textbf{IN} & \textit{Instance Normalization}\\
\textbf{NNE} & \textit{Neural Network Ensemble}\\
\textbf{Disjoint-CNN} & \textit{Disjoint Convolutional Neural Network}\\
\textbf{ConvTran} & \textit{Convolutional Transformer}\\
\textbf{RPE} & \textit{Relative Positional Encoding}\\
\textbf{eRPE} & \textit{Efficient Relative Positional Encoding}\\
\textbf{CART} & \textit{Classification and Regression Tree}\\
\textbf{EBA} & \textit{Elastic Barycenter Averaging}\\
\textbf{DBA} & \textit{DTW Barycenter Averaging}\\
\textbf{MSM} & \textit{Move-Split-Merge}\\
\textbf{MBA} & \textit{MSM Barycenter Averaging}\\
\textbf{SoftDBA} & \textit{SoftDTW Barycenter Averaging}\\
\textbf{SBD} & \textit{Shape Based Distance}\\
\textbf{NCC} & \textit{Normalized Cross Correlation}\\
\textbf{AE} & \textit{Auto-Encoders}\\
\textbf{VAE} & \textit{Variational Auto-Encoders}\\
\textbf{DRNN} & \textit{Dilated-RNN}\\
\textbf{MSE} & \textit{Mean Squared Error}\\
\textbf{KL} & \textit{Kullback-Leibler}\\
\textbf{SSL} & \textit{Self Supervised Learning}\\
\textbf{SimCLR} & \textit{Simple Contrastive LeaRning}\\
\textbf{DCCNN} & \textit{Dilated Causal CNN}\\
\textbf{MCL} & \textit{Mixup Contrastive Learning}\\
\textbf{TimeCLR} & \textit{Time Series Self-Supervised Contrastive Learning framework for Representation}\\
\textbf{NHST} & \textit{Null Hypothesis Significance Testing}\\
\textbf{CDD} & \textit{Critical Difference Diagram}\\
\textbf{MCM} & \textit{Multi-Comparison Matrix}\\
\textbf{AR} & \textit{Average Rank}\\
\textbf{CD} & \textit{Critical Difference}\\
\textbf{DP} & \textit{Dirichlet Process}\\
\textbf{t-SNE} & \textit{t-distributed Stochastic Neighbor Embedding}\\
\textbf{CO-FCN} & \textit{Custom Only-Fully Convolutional Network}\\
\textbf{H-FCN} & \textit{Hybrid-Fully Convolutional Network}\\
\textbf{H-Inception} & \textit{Hybrid-Inception}\\
\textbf{IT} & \textit{InceptionTime}\\
\textbf{H-IT} & \textit{Hybrid-InceptionTime}\\
\textbf{PHIT} & \textit{Pre-trained H-InceptionTime}\\
\textbf{BNM} & \textit{Batch Normalizing Multiplexer}\\
\textbf{LITE:} & \textit{Light Inception with boosTing tEchniques}\\
\textbf{LITEMV} & \textit{LITE MultiVariate}\\
\textbf{FLOPS} & \textit{Floating Point Operations Per Second}\\
\textbf{LMVT} & \textit{LITEMVTime}\\
\textbf{LT} & \textit{LITETime}\\
\textbf{CT} & \textit{ConvTran}\\
\textbf{D-CNN} & \textit{Disjoint Convolutional Neural Network}\\
\textbf{RF} & \textit{Receptive Field}\\
\textbf{TRILITE} & \textit{TRIplet Loss In TimE}\\
\textbf{1LP} & \textit{1 Linear Perceptron}\\
\textbf{CAM} & \textit{Class Activation Map}\\
\textbf{GAP} & \textit{Global Average Pooling}\\
\textbf{FFT} & \textit{Fast Fourier Transform}\\
\textbf{RI} & \textit{Rand Index}\\
\textbf{ARI} & \textit{Adjusted Rand Index}\\
\textbf{TP} & \textit{True Positive}\\
\textbf{TN} & \textit{True Negative}\\
\textbf{FP} & \textit{False Positive}\\
\textbf{FN} & \textit{False Negative}\\
\textbf{FID} & \textit{Fréchet Inception Distance}\\
\textbf{APD} & \textit{Average Pair Distance}\\
\textbf{Ref} & \textit{Reference}\\
\textbf{RMSE} & \textit{Root Mean Squared Error}\\
\textbf{SVAE} & \textit{Supervised Variational Auto-Encoder}\\
\textbf{GAN} & \textit{Generative Adversarial Network}\\
\textbf{MoCoGAN} & \textit{Motion and Content decomposed Generative Adversarial Network}\\
\textbf{GPT} & \textit{Generative Pre-trained Transformer}\\
\textbf{PoseGPT} & \textit{Pose Generative Pre-trained Transformer}\\
\textbf{ACTOR} & \textit{Action-Conditioned TransfORmer}\\
\textbf{UM-CVAE} & \textit{Uncoupled-Modulation Conditional Variational Auto-Encoder}\\
\textbf{DDPM} & \textit{Denoising Diffusion Probabilistic Model}\\
\textbf{PCA} & \textit{Principal Component Analysis}\\
\textbf{rec} & \textit{reconstruction}\\
\textbf{cls} & \textit{classification}\\
\textbf{MOS} & \textit{Mean Opinion Scores}\\
\textbf{WPD} & \textit{Warping Path Diversity}\\
\textbf{IS} & \textit{Inception Score}\\
\textbf{FD} & \textit{Fréchet Distance}\\
\textbf{AOG} & \textit{Accuracy On Generated}\\
\textbf{ACPD} & \textit{Average per Class Pair Distance}\\
\textbf{MMS} & \textit{Mean Maximum Similarity}\\
\textbf{CConvVAE} & \textit{Conditional Convolutional Variational Auto-Encoder}\\
\textbf{CGRUVAE} & \textit{Conditional Gated Recurrent Unit Variational Auto-Encoder}\\
\textbf{CTransVAE} & \textit{Conditional Transformer Variational Auto-Encoder}\\
\textbf{KL} & \textit{Kullback-Leibler}\\
\textbf{KAN} & \textit{Kolmogorov-Arnold Network}\\
\end{abbreviations}


%
%
%


%
%
%
%




\mainmatter 

\pagestyle{thesis} 

%
%
\addchap{Résumé des chapitres}

%
%
\section*{Chapitre 1 : État de l'art pour l'analyse des séries temporelles : Apprentissage supervisé et non supervisé}
\addcontentsline{toc}{section}{Chapitre 1 : État de l'art pour l'analyse des séries temporelles : Apprentissage supervisé et non supervisé}

Ce chapitre traite de l'état de l'art dans l'analyse des séries temporelles, 
avec un focus sur deux types d'apprentissage: supervisé et non supervisé. 
L'analyse des séries temporelles est essentielle pour les scientifiques de données, 
car elle permet d'extraire des informations utiles à partir de données qui évoluent 
dans le temps. Il existe de nombreuses techniques dans ces deux approches, chacune 
adaptée à des contextes et des objectifs différents. Le chapitre commence par une 
introduction générale de ces concepts et passe ensuite en revue les méthodes les 
plus avancées dans ce domaine.

Dans l'apprentissage supervisé, l'objectif principal est de prédire des valeurs 
futures en se basant sur des observations passées. Cela inclut des tâches comme 
la régression extrinsèque, où l'on cherche à prédire des valeurs continues, par 
exemple le prix futur des actions, ou la classification, où l'on catégorise 
des événements futurs en fonction des observations historiques. Un 
exemple serait de classifier des emails comme "spam" ou "non-spam" 
en analysant les emails précédents. En revanche, l'apprentissage 
non supervisé consiste à découvrir des structures ou des motifs 
cachés dans les données sans utiliser de labels ou de catégories 
prédéfinis. Deux des tâches principales en apprentissage non supervisé 
sont le regroupement (clustering) et la détection d'anomalies. Ces 
méthodes sont utilisées pour identifier des comportements inhabituels 
ou des groupes de données similaires dans divers domaines, comme la 
segmentation des clients ou la détection de fraudes financières.

Le chapitre présente ensuite les techniques de classification des 
séries temporelles (\emph{Time Series Classification, TSC}), qui ont beaucoup 
évolué ces dernières décennies. Parmi les approches traditionnelles, 
on trouve les méthodes basées sur la distance (Plus Proche Voisin), qui utilisent des mesures 
de similarité comme le \emph{Dynamic Time Warping (DTW)}, une méthode capable 
de comparer deux séries temporelles en tenant compte des décalages 
temporels entre elles. Cette technique est très utile dans des domaines 
variés, allant de la reconnaissance d'activités humaines à la communication sans 
fil. Par exemple, dans une série de données temporelles, DTW permet de trouver 
la meilleure correspondance entre deux séries, même si elles ne sont pas parfaitement 
alignées dans le temps. C'est une avancée par rapport à des mesures de distance 
plus simples comme la distance euclidienne, qui ne tient pas compte des différences temporelles.

Un autre domaine important est celui des ``shapelets''~\cite{shapelets}. Ce sont des sous-séquences 
discriminatives d'une série temporelle qui permettent de distinguer différentes 
classes. Les shapelets sont souvent utilisées dans des domaines où la lisibilité 
et l'interprétabilité des modèles sont importantes, comme en médecine, 
où il est essentiel de comprendre pourquoi un certain diagnostic est fait 
à partir des données. Ces shapelets capturent des motifs locaux dans les séries 
temporelles qui sont représentatifs d'une classe spécifique.

Les approches basées sur des dictionnaires~\cite{boss} sont également abordées dans 
ce chapitre. Elles transforment les séries temporelles en séquences de symboles, 
ce qui permet de repérer des motifs répétés dans les données. L'une des techniques 
les plus connues dans ce domaine est SAX (\emph{Symbolic Aggregate approXimation})~\cite{sax}, 
qui simplifie les séries temporelles en les représentant par une séquence de symboles, 
rendant plus facile la détection de motifs répétitifs. Des méthodes dérivées comme 
BOSS (\emph{Bag-of-SFA-Symbols})~\cite{boss} permettent d'améliorer cette 
approche en analysant les séries temporelles sous forme de fenêtres qui se 
chevauchent, et en créant des ensembles de mots qui facilitent la 
classification des séries temporelles.

le chapitre aborde les méthodes hybrides qui combinent plusieurs approches pour 
améliorer les performances des modèles d'analyse de séries temporelles. Un modèle 
hybride célèbre est HIVE-COTE~\cite{hive-cote}, qui combine des techniques basées sur les distances, 
les dictionnaires, les shapelets et les intervalles, en les utilisant ensemble pour 
créer un modèle d'ensemble puissant et robuste. Ces approches combinées permettent 
souvent d'obtenir des résultats plus précis que l'utilisation d'une seule méthode, 
en exploitant les forces de chaque approche.

Les modèles plus récents pour la classification des séries temporelles 
sont de plus en plus basés sur l'apprentissage profond, en particulier les 
réseaux neuronaux convolutifs (\emph{Convolutional Neural Networks, CNN}). Ces 
réseaux utilisent des filtres qui capturent des motifs dans les données, 
comme des relations temporelles complexes, en parcourant les séries de 
manière glissante. Ils ont prouvé leur efficacité dans la reconnaissance 
de motifs dans les séries temporelles, souvent surpassant les méthodes 
traditionnelles.
Le chapitre consacre également une section importante à l'apprentissage profond 
pour l'analyse des séries temporelles. Les réseaux neuronaux, notamment les CNN 
et les réseaux récurrents (RNN), sont de plus en plus utilisés dans ce domaine.
Une particularité des réseaux neuronaux est leur capacité à extraire des 
caractéristiques complexes des séries temporelles sans nécessiter une étape 
explicite de sélection de caractéristiques, ce qui simplifie le processus 
d'apprentissage. Ces modèles peuvent paralléliser les calculs, ce qui 
les rend plus rapides que les approches traditionnelles pour les grands 
volumes de données. Les CNN, en particulier, sont efficaces pour la 
classification des séries temporelles, car ils permettent de détecter 
des motifs locaux importants dans les données.

La régression extrinsèque est une tâche importante de l'apprentissage 
supervisé, où l'objectif est de prédire une valeur continue à partir 
d'une série temporelle. Contrairement à la classification, où les résultats 
sont des catégories discrètes, la régression vise à estimer des valeurs 
numériques précises. Par exemple, elle peut être utilisée pour prédire la 
température d'une région en fonction des données climatiques historiques 
ou pour estimer les ventes futures en fonction des tendances des ventes passées.
L'approche traditionnelle consiste à utiliser des modèles de régression linéaire 
ou d'autres techniques classiques, mais avec les récentes avancées, les méthodes 
basées sur l'apprentissage profond gagnent en popularité. Les réseaux neuronaux 
peuvent capturer les relations non linéaires entre les observations passées et futures.

L'apprentissage non supervisé pour les séries temporelles ne vise pas à 
prédire des valeurs futures ou à classer des événements, mais plutôt à 
découvrir des structures cachées dans les données. Cette partie du chapitre 
se concentre sur troix techniques principales: le clustering, le prototypage
et l'apprentissage auto-supervisé. Ces méthodes sont particulièrement utiles 
lorsqu'on n'a pas, ou peu, de labels pour les données, ou lorsque les séries temporelles 
sont complexes et contiennent des motifs que l'on souhaite analyser sans 
définir de catégories précises à l'avance.

Le clustering (ou regroupement) est une méthode utilisée pour regrouper 
des séries temporelles similaires en fonction de leurs caractéristiques. 
Il permet d'organiser les données en groupes homogènes (\emph{clusters}) où les 
membres d'un même groupe partagent des propriétés communes. Par exemple, 
dans une analyse de séries temporelles de données clients, le clustering 
peut être utilisé pour segmenter les clients en fonction de leurs comportements 
d'achat, afin d'identifier différents types de consommateurs. Le chapitre 
présente plusieurs algorithmes pour le clustering, comme les méthodes 
basées sur la distance (ex : $k$-means), et explique comment ces techniques 
peuvent être adaptées aux séries temporelles.

L'un des défis du clustering dans les séries temporelles est de choisir 
la bonne mesure de similarité. Alors que les mesures comme la distance 
euclidienne sont simples à implémenter, elles ne prennent pas en compte 
les décalages dans le temps. Le \emph{Dynamic Time Warping (DTW)}, déjà mentionné 
dans la section sur la classification, est souvent utilisé pour cette raison. 
En outre, le chapitre mentionne des méthodes plus récentes qui combinent 
les approches traditionnelles avec des techniques d'apprentissage profond~\cite{deep-tscl-bakeoff} 
pour améliorer la précision du clustering.

Le prototypage~\cite{eamonn-prototyping} est une autre technique utile dans l'analyse non supervisée 
des séries temporelles, où l'objectif est de représenter un ensemble de séries 
temporelles par une série représentative appelée prototype. Cela permet de 
résumer efficacement de grands ensembles de données en une série qui capture 
l'essence des données. Ces prototypes peuvent être utilisés pour visualiser 
ou comprendre des groupes de séries temporelles, ou pour simplifier des modèles 
d'analyse complexes.

L'approche classique consiste à calculer une moyenne ou une médiane pour 
les séries temporelles d'un groupe, mais cette méthode simple peut échouer 
à capturer des motifs subtils. C'est pourquoi des méthodes plus sophistiquées, 
comme le $k$-means~\cite{kmeans-paper} avec le DTW ou d'autres mesures de similarité adaptées aux 
séries temporelles, sont utilisées pour obtenir des prototypes plus précis.

Le chapitre aborde également l'apprentissage auto-supervisé, qui est une méthode émergente 
dans l'analyse des séries temporelles. Contrairement à l'apprentissage supervisé 
classique où les modèles sont entraînés avec des labels, l'apprentissage auto-supervisé 
permet aux modèles d'apprendre à partir des données elles-mêmes, sans avoir 
besoin de labels explicites. Une technique commune consiste à créer des tâches 
d'apprentissage auxiliaires (par exemple, prédire une partie manquante de la 
série temporelle ou réorganiser des segments) pour permettre au modèle 
d'apprendre des représentations utiles des données. Ces représentations 
peuvent ensuite être réutilisées pour d'autres tâches comme la classification 
ou la régression.

L'apprentissage supervisé est particulièrement utile dans les situations où les 
labels sont rares ou coûteux à obtenir. Il permet d'exploiter efficacement 
les grands ensembles de données non étiquetés, ce qui est souvent le cas 
dans des secteurs comme la surveillance de la santé ou la gestion de grandes 
infrastructures industrielles.

Le chapitre conclut sur l'importance des progrès récents dans l'analyse des 
séries temporelles, que ce soit dans le domaine supervisé ou non supervisé. 
Il met en avant la richesse des méthodes disponibles et souligne que les 
techniques modernes, notamment celles basées sur l'apprentissage profond 
et l'apprentissage supervisé, sont essentielles pour résoudre les problèmes 
complexes que posent les séries temporelles aujourd'hui. Ces méthodes 
permettent de traiter des volumes de données de plus en plus importants, 
tout en offrant des prédictions plus précises et une meilleure capacité à 
identifier des comportements cachés ou inhabituels dans les données.

En résumé, le chapitre offre un panorama complet des techniques actuelles 
et des avancées dans l'analyse des séries temporelles. Que ce soit pour des 
tâches supervisées comme la classification et la régression, ou non supervisées 
comme le clustering et le prototypage, il souligne que la combinaison des 
approches traditionnelles avec les nouvelles techniques d'apprentissage profond 
offre des résultats prometteurs dans de nombreux domaines.

\section*{Chapitre 2 : Évaluation comparative des modèles d'apprentissage automatique sur les données de séries temporelles}
\addcontentsline{toc}{section}{Chapitre 2 : Évaluation comparative des modèles d'apprentissage automatique sur les données de séries temporelles}

L'évaluation comparative des modèles de machine learning est une pratique 
cruciale pour évaluer et améliorer les algorithmes. Elle permet de comparer 
la performance des modèles sur plusieurs jeux de données afin d'identifier 
les méthodes les plus performantes et de mieux comprendre les forces et 
faiblesses des différents modèles. Les méthodes traditionnelles comme le 
\emph{Wilcoxon signed-rank test}~\cite{wilcoxon-paper} et le \emph{Nemenyi test}~\cite{nemenyi-paper}
sont souvent utilisées, mais elles 
présentent des limites, par exemple elles peuvent être manipulées et ne donnent pas 
toujours une image complète des différences entre les modèles. Des méthodes plus récentes, 
comme les approches bayésiennes, sont proposées comme alternatives plus fiables pour les 
comparaisons multiples.

Dans ce chapitre, on se concentre sur les méthodes actuelles d'évaluation comparative, 
en particulier sur la tâche de classification des séries temporelles. L'un des outils 
les plus utilisés pour ces comparaisons est le
\emph{Critical Difference Diagram (CDD)}~\cite{demsar-cdd-paper,cdd-benavoli-paper}. 
Cependant, ce diagramme présente des limites importantes, comme l'instabilité dans les 
classements et la possibilité de manipulation. Une méthode nouvelle, appelée 
\emph{Multiple Comparison Matrix (MCM)}, est introduite pour offrir des comparaisons 
plus robustes et interprétables des modèles.

Le CDD résume les performances des modèles en les classant 
sur plusieurs jeux de données, comme les 128 de l'archive UCR~\cite{ucr-archive}.
Cependant, cette méthode 
ignore la magnitude des différences et peut être instable. Par exemple, ajouter ou retirer 
un modèle peut changer les conclusions sur les différences significatives.
Le \emph{Friedman test}~\cite{friedman-paper}
et le \emph{Nemenyi test}~\cite{nemenyi-paper} sont souvent utilisés,
mais ils ont des faiblesses similaires.

Le CDD simplifie les comparaisons, mais il présente trois problèmes majeurs :

\begin{itemize}
    \item \textbf{Instabilité des classements}: Le classement change quand on ajoute ou 
    retire des modèles.
    \item \textbf{Ignorance de la magnitude}: Le CDD ne prend pas en compte
    l'ampleur des gains ou pertes de performance.
    \item \textbf{Problèmes avec les tests statistiques}: Les \emph{p-values}
    ne reflètent pas toujours les différences réelles entre les modèles.
\end{itemize}

Pour résoudre ces problèmes, la \emph{Multiple Comparison Matrix (MCM)} est proposée 
dans ce chapitre 
comme une méthode plus fiable pour comparer les modèles. Cette méthode se concentre 
sur les comparaisons par paires des modèles, et elle garantit que les comparaisons 
restent invariantes à l'ajout ou au retrait d'autres modèles. Elle offre une vue 
plus détaillée des performances de chaque modèle avec des informations comme:

\begin{itemize}
    \item La différence moyenne de performance entre deux modèles.
    \item Un compte des victoires, égalités et défaites pour chaque paire de modèles.
    \item Une p-value issue d'un test \emph{Wilcoxon}.
\end{itemize}

En conclusion, l'évaluation comparative des modèles de machine learning joue un rôle 
central dans l'amélioration continue des algorithmes, en permettant d'identifier les 
meilleures méthodes et de mieux comprendre leurs forces et faiblesses. Bien que des 
approches traditionnelles comme les tests de Wilcoxon et de Nemenyi soient couramment 
utilisées, elles présentent des limites, notamment une instabilité dans les classements 
et une faible prise en compte des différences de magnitude entre les performances des 
modèles. Ces méthodes peuvent aussi être influencées par l'ajout ou le retrait de 
modèles, rendant les résultats moins fiables.

C'est dans ce contexte que des alternatives plus récentes, comme les approches 
bayésiennes ou la \emph{Multiple Comparison Matrix (MCM)}, ont été proposées 
pour offrir des comparaisons plus robustes et pertinentes. La MCM, en particulier, 
se concentre sur des comparaisons par paires entre modèles et garantit une 
meilleure stabilité des résultats, indépendamment des changements dans 
l'ensemble des modèles testés. Elle fournit aussi des informations plus détaillées 
et descriptives, notamment sur les différences de performance moyennes, ainsi 
que sur les victoires et défaites entre chaque paire de modèles.

Ce chapitre a mis en évidence les limites des outils traditionnels comme le 
\emph{Critical Difference Diagram (CDD)}, tout en proposant des solutions 
complémentaires avec la MCM pour pallier ces faiblesses. Ainsi, l'approche 
MCM permet d'offrir une vision plus précise et nuancée des différences entre 
les modèles, en garantissant des comparaisons plus cohérentes et informatives. 
Cela souligne l'importance de poursuivre l'exploration de méthodes d'évaluation 
robustes pour améliorer l'efficacité et la fiabilité des modèles de machine 
learning appliqués aux séries temporelles.

\section*{Chapitre 3 : Vers la recherche de modèles de fondation pour la classification des séries temporelles}
\addcontentsline{toc}{section}{Chapitre 3 : Vers la recherche de modèles de fondation pour la classification des séries temporelles}

Dans le domaine dynamique de la classification des séries temporelles
(\emph{Time Series Classification, TSC}), l'un des principaux défis consiste à développer 
des modèles robustes et adaptables à des jeux de données variés. Les modèles de fondation, qui 
sont de grands modèles pré-entraînés capables de généraliser sur plusieurs tâches, 
offrent une solution prometteuse. L'intérêt de ces modèles est qu'ils simplifient 
et accélèrent le processus d'ajustement pour des tâches spécifiques. Cela s'avère 
crucial dans des domaines comme la médecine, par exemple avec les signaux ECG, ou 
la gestion du trafic, où l'entraînement de modèles à partir de zéro est coûteux et 
long. Les modèles de fondation offrent un point de départ pré-entraîné qui comprend déjà
les motifs fondamentaux, ce qui permet de gagner en temps et en précision lors de l'ajustement 
à des jeux de données particuliers.

Ce chapitre présente deux contributions principales pour progresser vers 
les modèles de fondation profonds:

\begin{enumerate}
    \item La création de filtres convolutifs faits main pour améliorer 
    la généralisation des modèles.
    \item L'utilisation d'une méthodologie de pré-entraînement pour ajuster (\emph{fine tune})
    ces modèles à des tâches de classification spécifiques.
\end{enumerate}

Ces filtres sont conçus pour se concentrer sur des caractéristiques 
générales des données, indépendamment du domaine spécifique, 
afin d'améliorer l'adaptabilité des modèles sur diverses tâches.

Les modèles de deep learning pour la TSC sont souvent confrontés à des problèmes tels que:

\begin{itemize}
    \item Surapprentissage (\emph{overfitting}), où les modèles deviennent trop spécialisés 
    sur les données d'entraînement
    \item Complexité computationnelle
    \item Apprentissage de filtres redondants
\end{itemize}

Les réseaux neuronaux convolutifs (\emph{Convolutional Neural Networks, CNNs}) traditionnels 
apprennent les filtres grâce à la rétropropagation (\emph{backpropagation}), 
mais ce processus peut mener à 
des filtres trop spécifiques, manquant de généralité. Une solution consiste à créer des 
filtres manuellement, qui détectent des motifs génériques dans les données. Avant 
d'adopter cette approche, il faut supposer que les modèles CNN peuvent apprendre 
des filtres génériques communs à travers différents jeux de données.

Pour tester cette hypothèse, une analyse a été réalisée sur l'espace des filtres 
appris par les modèles CNN sur plusieurs ensembles de données ECG. L'analyse a 
montré qu'un certain nombre de filtres convolutifs coïncident dans l'espace, 
suggérant que certains filtres peuvent être partagés entre différents jeux de données.

Trois types de filtres faits main sont proposés pour capturer des motifs spécifiques 
dans les séries temporelles:

\begin{itemize}
    \item Filtre de détection de tendance croissante : construit pour détecter les 
    segments de séries temporelles où les valeurs augmentent
    \item Filtre de détection de tendance décroissante : construit pour 
    détecter les segments où les valeurs diminuent.
    \item Filtre de détection de pics : construit pour détecter les 
    changements brusques dans une série 
    temporelle, comme une augmentation suivie d'une diminution rapide.
\end{itemize}

Ces filtres ne sont pas ajustés pendant l'entraînement, permettant ainsi au modèle de se 
concentrer sur l'apprentissage de motifs plus complexes et nuancés. Ces filtres sont 
similaires à ceux utilisés en vision par ordinateur, tels que les filtres
Sobel~\cite{sobel-paper1,sobel-paper2}.

Pour évaluer l'impact des filtres faits main, trois architectures adaptées sont proposées :

\begin{itemize}
    \item CO-FCN (\emph{Custom Only-Fully Convolutional Network}): la premiere couche 
    de convolution dans FCN~\cite{fcn-resnet-mlp-paper} est 
    remplacé entièrement par les trois filtres créés manuellement.
    \item H-FCN (\emph{Hybrid-Fully Convolutional Network}): combine les filtres 
    créés manuellement et des filtres appris dans la première couche de convolution de FCN.
    \item H-Inception (\emph{Hybrid-Inception}): intègre les filtres faits main dans l'architecture 
    Inception~\cite{inceptiontime-paper}, qui est connue pour ses performances sur la TSC.
\end{itemize}

Les résultats expérimentaux sur 128 jeux de données de l'archive
UCR~\cite{ucr-archive} montrent que les modèles avec 
filtres faits main surpassent souvent leurs versions originales. Par exemple, 
CO-FCN a mieux performé que le modèle FCN d'origine sur la plupart des jeux de 
données. De même, les versions hybrides H-FCN et H-InceptionTime montrent 
des améliorations significatives.

Dans le reste de ce chapitre, nous abordons le développement d'un modèle de fondation pour 
la classification des séries temporelles en utilisant une tâche prétexte. 
L'objectif est de pré-entraîner un modèle sur une tâche générique avant de 
l'adapter à des jeux de données spécifiques pour des tâches de classification 
particulières. Cela permet non seulement de réduire le temps d'entraînement, 
mais aussi d'améliorer la capacité de généralisation du modèle. Le modèle de 
fondation pré-entraîné tire parti de l'architecture H-Inception et intègre des 
filtres convolutifs faits main, qui se sont révélés utiles pour capturer 
des motifs généraux dans les séries temporelles.

Dans de nombreuses applications du monde réel, comme la médecine ou la gestion 
industrielle, il est coûteux et laborieux de collecter et d'étiqueter de grandes 
quantités de données pour entraîner un modèle d' apprentissage profond à partir de zéro. 
Par exemple, pour la détection des maladies cardiaques à partir de signaux ECG, 
il est nécessaire de disposer d'énormes quantités de données annotées par des 
professionnels de santé, ce qui n'est souvent pas réalisable. De même, dans des 
domaines comme la maintenance prédictive, la collecte de données de capteurs 
nécessite un suivi à long terme et l'expertise d'ingénieurs pour annoter les 
modes de défaillance.

L'idée des modèles de fondation est de pré-entraîner un modèle 
robuste sur un large éventail de jeux de données similaires, ce qui permet 
ensuite de l'adapter plus facilement et plus efficacement à de nouveaux jeux 
de données spécifiques. Ce processus de \emph{fine tuning} est beaucoup plus rapide 
et permet d'éviter le risque de surapprentissage, un problème 
fréquent lorsque l'on travaille avec des ensembles de données de petite taille.

La méthode proposée dans ce chapitre repose sur une tâche prétexte construite pour 
entraîner un modèle à reconnaître les motifs généraux dans des séries temporelles 
à partir de différents jeux de données. Cette tâche prétexte consiste à apprendre 
au modèle à prédire le jeu de données d'origine de chaque échantillon de série 
temporelle. Ce processus permet au modèle de capturer des caractéristiques 
génériques applicables à plusieurs ensembles de données. Une fois ce modèle 
pré-entraîné, il est ensuite ajusté sur des tâches de classification propores
à chaque jeu de données.

L'architecture choisie pour ce modèle est basée sur H-Inception, proposé dans la premiere 
contribution de ce chapitre. Le processus global peut être résumé en deux étapes principales:
\begin{itemize}
    \item Étape 1: Pré-entraînement du modèle sur une tâche générique où il 
    doit prédire l'origine des séries temporelles.
    \item Étape 2: Ajustement du modèle sur des jeux de données 
    spécifiques pour des tâches de classification précises.
\end{itemize}

Les résultats expérimentaux montrent que le modèle proposé PHIT (\emph{Pre-trained H-InceptionTime}) 
dépasse la performance des approches traditionnelles d'ajustement. Le modèle a été 
comparé à des approches de pointe sur l'ensemble des 88 jeux de données de séries temporelles.

Une comparaison directe entre notre approche avec pré-entraînement et un modèle 
sans pré-entraînement a été réalisée. Les résultats montrent que PHIT 
offre de meilleures performances que le modèle de base dans $48$ jeux de données, 
tandis que le modèle de base n'en surpasse que $23$. L'analyse statistique de ces 
résultats à l'aide du \emph{Wilcoxon Signed-Rank Test}~\cite{wilcoxon-paper}
indique que l'amélioration apportée 
par PHIT est significative avec une p-value de $0,021$ (inférieure au seuil de $0,05$).

En conclusion, ce chapitre propose des avancées majeures dans la classification des 
séries temporelles en introduisant les modèles de fondation, qui permettent une 
généralisation plus efficace sur différentes tâches. Ces modèles pré-entraînés 
offrent un gain de temps et de précision, en particulier dans des domaines comme 
la médecine ou la gestion du trafic, où l'entraînement à partir de zéro est coûteux et complexe.

Les deux contributions principales de ce travail sont, d'une part, la 
création de filtres convolutifs créés manuellement, construits pour détecter des 
motifs génériques et améliorer la robustesse des modèles, et, d'autre 
part, l'utilisation d'une méthodologie de pré-entraînement. Cette dernière 
permet d'affiner les modèles sur des tâches spécifiques, en les rendant
plus adaptés à chaque jeux de données sans nécessiter un long processus 
d'apprentissage.

Les résultats expérimentaux montrent que les architectures telles que CO-FCN 
et H-Inception, intégrant ces filtres faits main, surpassent souvent les 
versions classiques sur un large éventail de jeux de données. De plus, 
la méthode de pré-entraînement basée sur une tâche générique améliore 
significativement les performances des modèles sur des tâches spécifiques, 
comme l'a montré le modèle PHIT.

En résumé, ce chapitre démontre l'efficacité des modèles de fondation dans la 
classification des séries temporelles, en offrant une meilleure capacité 
de généralisation et une réduction du 
temps d'entraînement, ce qui les rend particulièrement utiles pour des 
applications nécessitant des solutions rapides et performantes.

\section*{Chapitre 4 : Réduire la complexité des modèles d'apprentissage profond pour la classification des séries temporelles}
\addcontentsline{toc}{section}{Chapitre 4 : Réduire la complexité des modèles d'apprentissage profond pour la classification des séries temporelles}

Ce chapitre explore la réduction de la complexité des modèles d'apprentissage profond 
dans le cadre de la classification des séries temporelles. Traditionnellement, 
les modèles plus complexes et volumineux, comme InceptionTime~\cite{inceptiontime-paper} 
avec ses $2,1$ millions 
de paramètres, ont montré de bonnes performances. Cependant, leur complexité présente 
des défis lorsqu'il s'agit de les déployer dans des environnements à ressources 
limitées, comme les dispositifs embarqués ou mobiles.

Pour répondre à ces besoins, ce chapitre introduit le modèle LITE, qui vise à maintenir 
des performances compétitives tout en réduisant considérablement la taille et la complexité 
du modèle. LITE se base sur une version allégée de l'architecture Inception et intègre 
des techniques de boosting pour améliorer la capacité de généralisation sur divers jeux 
de données, tout en restant rapide à entraîner et économe en ressources.

Les principaux objectifs du modèle LITE incluent:

\begin{itemize}
    \item Architecture Inception allégée: réduction du nombre de paramètres et 
    de la complexité sans sacrifier la performance
    \item Techniques de boosting: intégration de techniques qui améliorent la 
    généralisation, réduisent le surapprentissage, et augmentent la précision
    \item Efficacité et adaptabilité: offrir un modèle adapté à des environnements 
    contraints en termes de ressources, tout en maintenant des performances 
    élevées avec un faible coût computationnel
\end{itemize}

L'architecture LITE est une version simplifiée du réseau Inception, construit pour maximiser 
l'efficacité tout en minimisant la complexité. Ses principaux éléments sont:

\begin{itemize}
    \item Filtres créés manuellement: Reprenant les contributions du chapitre précédent, 
    ces filtres sont utilisés dans la première couche pour capturer des motifs 
    génériques dès le début, tout en évitant un surapprentissage. Ils fonctionnent 
    en parallèle avec des convolutions apprises pour maximiser l'efficacité du modèle
    \item Multiplexing de convolution: Plusieurs convolutions avec différentes tailles 
    de filtres sont appliquées en parallèle dans les premières couches. Cela permet au 
    modèle de capturer divers motifs dans les séries temporelles, optimisant ainsi 
    l'extraction de caractéristiques importantes dès le début
    \item \emph{DepthWise Separable Convolutions(DWSC)}: Ces convolutions sont utilisées 
    dans les couches profondes de l'architecture. Elles permettent de réduire drastiquement 
    le nombre de paramètres et la charge computationnelle tout en conservant une forte 
    capacité d'extraction de caractéristiques pertinentes~\cite{mobilenets}
    \item Convolutions dilatées: Les couches profondes utilisent des convolutions 
    dilatées pour augmenter le champ réceptif du modèle sans ajouter de nouveaux 
    paramètres, ce qui permet au modèle d'apprendre des dépendances à plus long 
    terme dans les données~\cite{rocket}
    \item Pooling global: Un pooling global est appliqué dans les dernières 
    couches pour réduire la dimension des données avant la classification 
    finale, comme cela se fait dans les modèles classiques tels que
    FCN et ResNet~\cite{fcn-resnet-mlp-paper}
\end{itemize}

LITETime est une version ensemble du modèle LITE, similaire à
InceptionTime~\cite{inceptiontime-paper}, qui 
regroupe plusieurs modèles LITE pour améliorer les performances globales.

Les résultats expérimentaux montrent que LITE, avec ses moins de $10 000$ paramètres, 
surpasse des modèles bien plus volumineux comme FCN ($264 000$ paramètres) et ResNet 
($504 000$ paramètres). Il atteint une précision de $0,8304$, proche de celle des 
modèles plus complexes comme Inception ($0,8393$), tout en utilisant beaucoup 
moins de ressources et en étant beaucoup plus rapide à entraîner.

Lors des tests sur l'archive UCR avec 128 jeux de données~\cite{ucr-archive}, LITETime atteint 
une précision moyenne de 0,8462, tout en restant bien plus petit que InceptionTime. 
Il utilise seulement $2,34\%$ des paramètres d'InceptionTime, ce qui en fait une 
option beaucoup plus légère et efficace pour des environnements à ressources limitées.

Dans de nombreux cas réels, les séries temporelles sont multivariées, ce qui signifie 
qu'elles comportent plusieurs canaux de données (comme dans la santé, où des données 
ECG sont mesurées sur plusieurs axes). Pour s'adapter à cela, l'architecture LITEMV
a été développée, reposant sur la base du modèle LITE mais modifiée pour mieux gérer 
les séries temporelles multivariées.
LITEMV remplace les convolutions standards dans les premières couches par des convolutions 
\emph{DepthWise}, permettant de traiter chaque canal indépendamment avant de les 
combiner efficacement. Cela permet au modèle de mieux capturer les interactions 
entre les différents canaux de données.

Lors des tests sur 30 jeux de données multivariées de l'archive UEA~\cite{uea-archive}, LITEMVTime, 
l'ensemble de modèles basé sur LITEMV, a surpassé des modèles de pointe comme 
InceptionTime~\cite{inceptiontime-paper} et Disjoint-CNN~\cite{disjoint-cnn-paper}.
Dans certains cas, comme sur le jeu de données 
EigenWorms, LITEMVTime a obtenu une précision impressionnante de $93,89\%$, 
contre seulement $59,34\%$ pour ConvTran~\cite{convtran-paper},
un autre modèle de pointe pour la classification multivariée.

LITE et LITEMV représentent des avancées majeures dans la classification des séries temporelles. 
Grâce à leur faible complexité et leur efficacité énergétique, ces modèles sont parfaitement 
adaptés à des environnements à ressources limitées, tout en maintenant des performances 
compétitives face à des modèles bien plus complexes. De plus, l'approche d'ensemble, 
avec LITETime et LITEMVTime, montre que ces architectures peuvent offrir une précision 
encore plus élevée sans compromettre leur légèreté.

\section*{Chapitre 5 : Apprentissage semi-supervisé et auto-supervisé pour les données de séries temporelles avec un manque de labels}
\addcontentsline{toc}{section}{Chapitre 5 : Apprentissage semi-supervisé et auto-supervisé pour les données de séries temporelles avec un manque de labels}

Ce chapitre aborde les défis liés à la classification des séries temporelles dans les 
cas où les données annotées sont rares. Les méthodes traditionnelles de classification 
supervisée nécessitent des données largement annotées, ce qui est souvent difficile 
à obtenir en raison de la complexité et du besoin d'expertise pour annoter ces données. 
En réponse à cela, l'apprentissage semi-supervisé et l'apprentissage auto-supervisé 
émergent comme des solutions prometteuses. Ces techniques exploitent des données non 
annotées pour améliorer les performances des modèles.

L'approche proposée, nommée TRILITE (\emph{TRIplet Loss In TimE}), repose sur le concept 
de perte de triplet (\emph{triplet loss})~\cite{triplet-loss-paper}, 
une technique utilisée pour apprendre des 
représentations discriminatives à partir de données non annotées. TRILITE emploie 
une méthode d'augmentation de données spécifiquement adaptée aux séries temporelles, 
permettant au modèle d'apprendre des caractéristiques utiles sans avoir besoin de 
beaucoup de données annotées. Deux cas d'utilisation sont explorés:

\begin{itemize}
    \item L'amélioration des performances d'un classificateur supervisé 
    avec peu de données annotées
    \item Un contexte d'apprentissage semi-supervisé où une partie des 
    données est étiquetée et l'autre non
\end{itemize}

TRILITE est un modèle auto-supervisé qui utilise la perte de triplet pour apprendre 
des représentations significatives à partir de séries temporelles. Le modèle se compose 
de trois encodeurs partageant les mêmes poids, traitant les triplets d'entrée 
(référence, positif, négatif). Le mécanisme de triplet loss vise à rapprocher les 
échantillons similaires tout en éloignant les échantillons dissemblables~\cite{facenet-paper}.

TRILITE a été testé sur 85 jeux de données de l'archive UCR~\cite{ucr-archive}. 
Les expériences ont montré 
que TRILITE améliore les performances des classificateurs dans les deux scénarios 
explorés. Dans le cas de données annotées en faible quantité, TRILITE a aidé à fournir
des représentations complémentaires qui, combinées à des modèles supervisés comme 
FCN~\cite{fcn-resnet-mlp-paper}, 
améliorent significativement la précision. De plus, dans un contexte semi-supervisé, 
TRILITE, en utilisant à la fois des données annotées et non annotées, a surpassé les 
approches traditionnelles sur plusieurs jeux de données.

Ce chapitre montre que l'apprentissage auto-supervisé et semi-supervisé, via des 
approches comme TRILITE, offre des solutions efficaces lorsque les données annotées 
sont limitées. TRILITE utilise les données non annotées pour générer des 
représentations utiles, améliorant ainsi la performance des modèles de classification 
de séries temporelles. Ces résultats ouvrent la voie à des méthodes plus efficaces 
et moins dépendantes de l'annotation manuelle des données.

\section*{Chapitre 6 : Analyse de séries temporelles pour les données de mouvement humain}
\addcontentsline{toc}{section}{Chapitre 6 : Analyse de séries temporelles pour les données de mouvement humain}

L'analyse des mouvements humains à partir de données de squelettes est devenue une 
technique couramment utilisée dans des domaines variés, tels que la reconnaissance 
d'actions humaines~\cite{human-motion-example-paper}, la réhabilitation~\cite{kimore-paper}, 
et la génération de séquences de 
mouvements réalistes~\cite{action2motion-paper}. 
Ces données sont généralement capturées à l'aide de technologies 
comme Microsoft Kinect~\cite{kinect-paper} et les systèmes de capture de mouvement 
(MoCap)~\cite{mocap-paper}, qui 
enregistrent les positions des articulations du corps humain dans un espace 
tridimensionnel. Chaque articulation est représentée par des coordonnées X, Y, 
et Z dans le temps, formant une série temporelle multivariée (MTS).

Les MTS capturant des mouvements humains présentent un intérêt particulier car 
elles permettent d'extraire et d'analyser des caractéristiques spatiales et 
temporelles simultanément. Par exemple, dans la réhabilitation, il est 
essentiel de comprendre non seulement le déplacement des articulations 
individuelles au fil du temps, mais aussi la manière dont ces articulations se 
coordonnent pour réaliser des mouvements complexes.

Les données de mouvements humains présentent plusieurs avantages pour 
l'analyse de séries temporelles:

\begin{itemize}
    \item Elles sont souvent bien structurées et capturent les 
    dynamiques des articulations en mouvement
    \item Elles peuvent être directement utilisées dans de nombreux 
    algorithmes de l'apprentissage automatique et l'apprentissage profond 
    pour des tâches comme la classification, la régression et la génération.
    \item Elles sont particulièrement adaptées aux modèles qui exploitent la 
    relation entre les différentes dimensions des séries temporelles, 
    comme les réseaux neuronaux convolutifs (CNN) et les réseaux neuronaux récurrents (RNN).
\end{itemize}

Les modèles d'apprentissage profond, en particulier, se sont avérés être 
des outils puissants pour traiter et analyser les séries temporelles 
multivariées provenant de mouvements humains. Les architectures de 
réseaux neuronaux convolutifs ont montré leur efficacité dans l'extraction 
automatique des caractéristiques complexes des séries temporelles, offrant ainsi 
des performances supérieures à celles des méthodes manuelles traditionnelles. 
Ce chapitre se concentre sur plusieurs méthodes avancées d'analyse des MTS, 
notamment dans le domaine de la réhabilitation et de la génération de mouvements.

L'un des domaines d'application les plus importants pour l'analyse des mouvements 
humains est la réhabilitation. Dans ce contexte, les données de séries temporelles 
issues des mouvements humains peuvent être utilisées pour évaluer la progression 
des patients au cours de leurs séances d'exercices physiques. Traditionnellement, 
cette évaluation est réalisée par des experts humains, qui observent et notent 
la qualité des mouvements. Cependant, ce processus peut être subjectif, coûteux 
et manquer de précision. Les modèles d'apprentissage profond, en revanche, 
offrent une solution pour automatiser cette évaluation, fournissant des 
résultats rapides, cohérents et basés sur des données objectives.

Le modèle LITEMVTime, proposé dans chapitre 4 a été testé sur le jeu de données 
Kimore~\cite{kimore-paper}, un ensemble de données capturant des séquences 
de mouvements humains pendant des exercices de réhabilitation. Le jeu de données 
contient des enregistrements de patients sains et malades, chacun effectuant 
plusieurs exercices physiques sous la supervision d'experts humains. Chaque mouvement 
est annoté avec un score de qualité allant de $0$ (très mauvaise performance) à $100$ 
(excellente performance), attribué par des professionnels de la réhabilitation.

Le modèle LITEMVTime a été entraîné pour classer la qualité des mouvements en ``bon'' 
ou ``mauvais'' en utilisant les annotations d'experts comme vérité de terrain. 
Les résultats expérimentaux montrent que LITEMVTime surpasse d'autres architectures 
d'apprentissage profond telles que FCN, ResNet~\cite{fcn-resnet-mlp-paper}
et InceptionTime~\cite{inceptiontime-paper}, à la fois en termes de précision et 
de vitesse d'exécution. Grâce à sa conception légère, LITEMVTime peut être facilement 
intégré dans des systèmes cliniques en temps réel, fournissant ainsi des retours 
immédiats aux patients et aux cliniciens pendant les sessions de réhabilitation.

L'un des plus grands défis dans l'entraînement des modèles de deep learning sur des 
données médicales est le manque de données annotées. Les mouvements humains capturés 
pour des études médicales sont souvent limités, et leur annotation nécessite des experts 
spécialisés, ce qui en fait une ressource rare et coûteuse. Ce manque de données annotées 
peut entraîner des problèmes de surapprentissage, où les modèles d'apprentissage profond 
deviennent trop spécialisés sur les données d'entraînement et ne parviennent pas à 
bien généraliser sur de nouvelles données.
Pour répondre à ce problème, ce chapitre propose une méthode de prototypage de 
séries temporelles appelée ShapeDBA (\emph{Shape Dynamic Time Warping Barycenter Averaging}). 
Cette méthode permet de créer des prototypes qui représentent des moyennes barycentriques 
des séries temporelles, à partir desquelles de nouvelles séquences synthétiques peuvent 
être générées. Ces séquences synthétiques, qui conservent les propriétés essentielles 
des mouvements humains capturés, peuvent être ajoutées aux jeux de données 
d'entraînement pour augmenter artificiellement la taille du jeu de données 
et améliorer ainsi la généralisation des modèles.

Le prototypage de séries temporelles est une technique précieuse, en particulier 
pour les applications où les données réelles sont limitées. En créant des prototypes 
barycentriques, il est possible de générer des mouvements synthétiques qui imitent 
les mouvements réels des patients tout en offrant une plus grande diversité. Cela 
permet aux modèles de deep learning d'apprendre des motifs plus robustes et de 
mieux se généraliser à de nouveaux patients et à de nouvelles tâches de réhabilitation.

Les expérimentations menées sur le jeu de données Kimore~\cite{kimore-paper} montrent 
que l'ajout de données synthétiques générées par ShapeDBA améliore considérablement 
la performance des modèles d'apprentissage supervisé utilisés pour évaluer la qualité 
des mouvements des patients. Les modèles, lorsqu'ils sont entraînés à la fois sur 
des données réelles et synthétiques, produisent des prédictions plus précises sur 
la qualité des mouvements, réduisant les erreurs de prédiction mesurées par la MAE 
(erreur absolue moyenne) et la RMSE (erreur quadratique moyenne). Cela montre 
que ShapeDBA est une méthode efficace pour augmenter les jeux de données limités 
et améliorer les performances globales des modèles de régression.

Si le prototypage de séries temporelles est une méthode efficace pour étendre 
les jeux de données de manière synthétique, il existe une autre approche complémentaire, 
basée sur l'utilisation des modèles génératifs profonds. Les modèles génératifs, tels que 
les Auto-Encodeurs Variationnels (\emph{Variational Auto-Encoder, VAE})~\cite{vae-paper} et 
les Réseaux adverbiaux génératifs (\emph{Generative Adversarial Networks, GAN})~\cite{gan-paper}. 
Ces modèles peuvent apprendre des 
distributions complexes de mouvements et ensuite générer de nouvelles 
séquences qui ressemblent aux données d'entraînement d'origine.

Dans ce chapitre, on explore l'utilisation des VAE pour la génération 
de mouvements humains. Les VAE sont une classe de modèles génératifs qui apprennent 
à encoder des données d'entrée dans un espace latent de faible dimension, 
à partir duquel de nouvelles données peuvent être générées. Dans le cas des mouvements 
humains, les VAE peuvent capturer les dynamiques des articulations et générer 
des mouvements réalistes qui imitent les séquences observées dans les données d'entraînement.

Dans ce chapitre on propose le SVAE (\emph{Supervised Variational Autoencoder}) 
est une amélioration par rapport au VAE classique, car il intègre une tâche de classification 
dans l'espace latent du modèle. Cela permet au modèle de non seulement générer des 
séquences de mouvements humains réalistes, mais aussi de les classifier selon des 
catégories prédéfinies, ce qui renforce à la fois ses capacités génératives et discriminatives.

L'architecture du SVAE se compose de trois parties principales : l'encodeur, l'espace 
latent, et le décodeur, similaires à un VAE classique:
\begin{itemize}
    \item Entrée (séquence de squelettes): Les données d'entrée sont des 
    séquences de mouvements humaines capturées sous forme de coordonnées 
    3D des articulations squelettiques.
    \item Encodeur: L'encodeur prend les séquences de mouvements 
    comme entrée et apprend une représentation latente sous la 
    forme d'une distribution gaussienne (paramétrée par une moyenne et une variance). 
    L'objectif est d'apprendre une distribution latente compacte qui capture les 
    caractéristiques essentielles des mouvements humains.
    \item Supervision dans l'espace latent: Contrairement au VAE traditionnel, 
    le SVAE introduit une tâche de classification dans l'espace latent. Cette 
    supervision permet au modèle d'apprendre une séparation plus claire entre 
    les différentes actions (par exemple, marcher, courir, lever les bras). 
    Cela permet d'améliorer la cohérence entre la génération de séquences et 
    la classe d'action correspondante.
    \item Décodeur: Le décodeur prend un échantillon de l'espace latent et 
    reconstruit la séquence originale. En même temps, un classificateur est 
    intégré dans le modèle pour reconnaître l'action à partir de la représentation latente.
\end{itemize}

Le modèle SVAE permet ainsi de réaliser deux tâches simultanées :
\begin{itemize}
    \item Reconnaissance d'action: Prédire l'action associée à une séquence de mouvement.
    \item Génération de séquences réalistes: Créer des séquences de mouvements 
    réalistes en générant des exemples à partir de l'espace latent.
\end{itemize}

Ce double usage améliore à la fois les capacités génératives du modèle 
(production de nouvelles séquences de mouvements) et ses capacités discriminatives 
(classement des séquences dans la bonne catégorie).

Les expériences se basent sur le jeux de données de reconnaissance d'action
HumanAct12~\cite{action2motion-paper}. Ce jeux de données contient des 
séquences de mouvements humains où les positions des articulations sont 
enregistrées en 3D sur plusieurs frames.

Le modèle est testé sur sa capacité à générer des séquences de mouvements réalistes 
et variés à partir de l'espace latent. La qualité et diversité des séquences générées est mesurée 
à l'aide de metrics telles que la \emph{Fréchet Inception Distance (FID)}
et la \emph{Average Paired Distance (APD)}, 
qui évalue la similitude entre les distributions des données réelles et des données générées.

Les séquences générées par le SVAE sont plus réalistes que celles générées par des 
modèles VAE traditionnels. Le SVAE capture mieux les variations dans les mouvements 
humains et évite les artefacts communs des méthodes traditionnelles. Cela est reflété 
par des scores FID plus bas, indiquant une plus grande similarité entre les séquences 
générées et réelles.
Le modèle génère non seulement des séquences réalistes, mais aussi diversifiées. 
Cela est important pour les applications où des variations réalistes de mouvements 
sont requises, comme dans les jeux vidéo ou la réhabilitation médicale.

Ce chapitre a exploré plusieurs techniques avancées pour l'analyse des séries 
temporelles appliquées aux mouvements humains. Les modèles comme LITEMVTime ont 
montré leur efficacité pour évaluer la qualité des mouvements dans des contextes 
de réhabilitation, tandis que des approches comme ShapeDBA et le SVAE ont permis 
de surmonter les limites liées à la rareté des données annotées en générant des 
données synthétiques. Ces avancées offrent des perspectives prometteuses pour des 
applications en temps réel, non seulement dans le domaine médical, mais aussi dans 
des domaines créatifs comme le cinéma et les jeux vidéo.

Les technologies décrites dans ce chapitre démontrent que les données 
de mouvements humains capturées via des capteurs comme le Kinect ont 
le potentiel de révolutionner de nombreux domaines, en combinant l'analyse de 
séries temporelles avec des modèles d'apprentissage profond performants et des 
méthodes de génération de données synthétiques.

\section*{Chapitre 7 : Métriques d'évaluation pour la génération de mouvement humain}
\addcontentsline{toc}{section}{Chapitre 7 : Métriques d'évaluation pour la génération de mouvement humain}

Ce chapitre aborde les métriques d'évaluation des modèles génératifs appliqués à 
la génération de mouvements humains. Contrairement aux modèles discriminatifs, 
où la comparaison avec des données réelles est directe, les modèles génératifs 
posent un défi plus complexe~\cite{reliable-fidelity-diversity}, 
car il faut évaluer la fidélité des échantillons 
générés en fonction de leur ressemblance avec des données réelles et leur diversité. 
L'évaluation repose donc sur deux dimensions clés: la fidélité et la diversité. La 
fidélité mesure à quel point les données générées sont proches des données réelles, 
tandis que la diversité s'assure que le modèle génératif peut produire une variété 
d'échantillons.

Les méthodes traditionnelles d'évaluation, comme le \emph{Mean Opinion Scores (MOS)}~\cite{mos-paper}, 
ne sont pas adaptées aux modèles génératifs, car elles présupposent une perception 
uniforme de l'utilisateur, ce qui est souvent irréaliste. Par conséquent, 
l'évaluation quantitative devient essentielle pour juger la performance des 
modèles génératifs. Le chapitre souligne qu'il est difficile de trouver une 
métrique unique pour évaluer à la fois la fidélité et la diversité, d'où la 
nécessité d'un cadre unifié d'évaluation.

Un aspect crucial des données de mouvement humain est leur dépendance temporelle. 
Les distorsions temporelles, telles que les changements de fréquence ou les décalages 
dans le temps, jouent un rôle important dans l'évaluation des séquences de mouvements. 
Pourtant, de nombreuses métriques d'évaluation ne tiennent pas compte de ces aspects 
temporels, se concentrant davantage sur les caractéristiques latentes. Pour remédier 
à ce problème, une nouvelle métrique, appelée \emph{Warping Path Diversity (WPD)}, est 
introduite. Cette métrique permet de mesurer la diversité des distorsions 
temporelles dans les données réelles et générées, offrant ainsi une évaluation 
plus précise des modèles génératifs de séquences temporelles.

Les métriques de fidélité décrites dans ce chapitre incluent la \emph{Fréchet Inception 
Distance (FID)}~\cite{fid-original-paper}, qui évalue la différence entre les distributions des données réelles 
et générées. Plus la FID est basse, plus les données générées ressemblent aux données 
réelles. Une autre métrique importante est l'\emph{Accuracy on Generated (AOG)}, qui mesure la 
capacité du modèle à générer des échantillons conformes aux étiquettes de classes 
définies (par exemple, générer des mouvements de course lorsque la classe ``courir''  
est donnée). Enfin, la métrique de \emph{Density}~\cite{reliable-fidelity-diversity} évalue combien d'échantillons 
générés correspondent aux données réelles en mesurant la proximité entre ces 
deux ensembles dans l'espace des caractéristiques.

Les métriques de diversité permettent d'évaluer à quel point les données générées 
sont variées. La \emph{Average Pair Distance (APD)}~\cite{action2motion-paper}, 
par exemple, mesure la distance moyenne 
entre des paires d'échantillons générés, indiquant si le modèle évite la production de 
résultats trop similaires (un problème appelé mode collapse).
La \emph{Coverage}~\cite{reliable-fidelity-diversity} est une autre 
métrique qui mesure la proportion d'échantillons réels couverts par les échantillons 
générés, assurant que les données générées couvrent bien l'ensemble de l'espace des 
données réelles.

Un autre concept important introduit est celui de la Mean Maximum Similarity (MMS)~\cite{msm-paper}. 
Cette métrique évalue la nouveauté des échantillons générés en mesurant la distance 
entre les échantillons générés et les plus proches voisins dans l'ensemble des données 
réelles. Une valeur élevée de MMS indique que les échantillons générés sont non 
seulement variés, mais aussi nouveaux par rapport aux données d'entraînement.

Le \emph{Warping Path Diversity (WPD)}, une nouvelle métrique qu'on propose, est présentée 
pour évaluer les distorsions temporelles. Utilisant l'algorithme 
\emph{Dynamic Time Warping (DTW)}~\cite{dtw-paper}, cette métrique mesure comment 
les séquences générées diffèrent temporellement des séquences réelles. Par exemple, 
dans une séquence de mouvements comme ``boire avec la main gauche'', les échantillons 
réels peuvent commencer à différents moments, tandis que les échantillons générés 
peuvent ne pas varier suffisamment en termes de timing. WPD quantifie cette diversité 
dans les distorsions temporelles, offrant ainsi une évaluation plus fine.

Les expériences menées dans ce chapitre reposent sur l'utilisation de modèles \emph{Conditional 
Variational Auto-Encoders (CVAE)} pour la génération de mouvements humains. 
Ces modèles sont évalués sur plusieurs métriques en fonction de leurs architectures 
(CNN, RNN ou Transformer) et de différents hyperparamètres. Les résultats montrent 
que certains modèles, comme le CConvVAE (CVAE base sur les CNN), excellent en termes 
de fidélité et de 
diversité, mais les performances dépendent fortement des configurations de paramètres. 
Par exemple, le CConvVAE obtient les meilleurs résultats en fidélité (mesurée par la FID) 
dans plusieurs cas, mais les résultats peuvent varier lorsque l'on change les 
paramètres d'entraînement ou l'architecture.

L'analyse des résultats montre qu'il est impossible de trouver un modèle unique 
qui surpasse tous les autres sur toutes les métriques. Chaque métrique capture 
un aspect différent de la qualité des échantillons générés, et en fonction 
des besoins (diversité dans les jeux vidéo ou fidélité dans la réhabilitation médicale), 
on peut être amené à privilégier une métrique sur une autre.

En conclusion, ce chapitre propose un cadre d'évaluation unifié pour les modèles 
génératifs appliqués à la génération de mouvements humains, avec plusieurs métriques 
permettant d'évaluer à la fois la fidélité et la diversité des modèles. Le 
\emph{Warping Path Diversity (WPD)} ajoute une dimension temporelle essentielle à 
cette évaluation, en tenant compte des distorsions temporelles dans les séquences 
générées. Ce cadre d'évaluation contribue à améliorer la comparaison entre différents 
modèles génératifs et facilite l'avancement de la recherche dans le domaine de la 
génération de mouvements humains.

\section*{Chapitre 8 : Recherche reproductible}
\addcontentsline{toc}{section}{Chapitre 8 : Recherche reproductible}

Ce chapitre traite de l'importance de la reproductibilité dans la recherche 
scientifique, en particulier dans le contexte de l'analyse des séries temporelles et de 
l'apprentissage profond. La reproductibilité garantit que les travaux peuvent 
être reproduits et adaptés par d'autres chercheurs, renforçant ainsi la confiance 
dans les résultats et favorisant l'innovation future. Ce chapitre met en lumière 
les efforts entrepris pour assurer que les travaux présentés dans cette thèse 
respectent les normes les plus élevées en matière de reproductibilité.

Un élément clé de ce chapitre est l'introduction du paquet 
\emph{aeon}~\cite{aeon-paper}, une bibliothèque 
open-source en Python construit pour effectuer diverses tâches d'apprentissage 
automatique sur les séries temporelles. Le développement de ce paquet a permis 
d'intégrer les contributions issues de cette recherche dans une plateforme 
accessible à la communauté scientifique. En rendant ces outils disponibles à 
tous, le projet encourage la reproductibilité et l'utilisation plus large des 
méthodes développées au cours de ce travail.

La documentation détaillée et le code ouvert jouent un rôle crucial dans 
la reproductibilité. Toutes les expériences décrites dans les chapitres 
précédents sont soutenues par du code public, permettant ainsi aux chercheurs 
de reproduire les expériences, de valider les résultats et de construire de 
nouveaux modèles basés sur ce travail. Ce code est accompagné de descriptions 
claires, facilitant la prise en main et l'adaptation du projet par d'autres 
chercheurs. En fournissant des instructions détaillées et en tenant compte des 
commentaires de la communauté, l'objectif est d'améliorer constamment la 
reproductibilité et la fiabilité des recherches.

Le paquet \emph{aeon}~\cite{aeon-paper} est au cœur de cet effort, 
offrant des outils pour diverses tâches comme la classification, la 
régression, la détection d'anomalies, et la segmentation des séries 
temporelles. En tant que développeur principal, j'ai contribué 
à la conception et à l'extension de ce paquet pour intégrer des modèles 
d'apprentissage profond, notamment ceux utilisés dans la classification 
des séries temporelles. Les modèles tels que InceptionTime, H-InceptionTime, 
et LITETime ont été inclus, ainsi que de nouveaux modules en développement 
pour des tâches comme le clustering des séries temporelles.

Les efforts pour garantir la reproductibilité ne se limitent pas à la mise à 
disposition du code. Le maintien de ce cadre logiciel implique également la 
correction des bugs, l'amélioration de la documentation, et l'ajout de nouvelles 
fonctionnalités pour répondre aux besoins évolutifs de la communauté scientifique. 
Par ailleurs, l'utilisation des tests unitaires permet de s'assurer que les 
nouveaux développements n'affectent pas la performance du code existant.

Une section clé du chapitre concerne les principes fondamentaux d'un travail 
reproductible. Cela inclut une documentation soignée, la fourniture des dépendances 
nécessaires, et une architecture de code claire et modulaire. Le code doit être facile 
à comprendre et à modifier, permettant ainsi à d'autres chercheurs de l'étendre 
pour ajouter de nouveaux modèles ou fonctionnalités. Des bonnes pratiques telles que 
l'utilisation de noms de variables explicites et une organisation efficace des fichiers 
sont également mises en avant pour améliorer la lisibilité et la maintenabilité du code.

On souligne également l'importance d'utiliser des outils comme Docker pour 
faciliter la gestion des environnements de développement et garantir que 
le code fonctionne de manière cohérente sur différentes machines. L'utilisation 
de conteneurs Docker permet de simplifier l'intégration des dépendances, notamment 
les configurations CUDA nécessaires pour l'utilisation des GPU, assurant ainsi 
une reproduction facile des expériences dans des environnements informatiques complexes.

Un autre aspect abordé dans le chapitre concerne la visualisation et la publication 
des résultats sous forme d'outils interactifs, tels que des pages web et des figures 
dynamiques. Ces outils permettent de mieux comprendre les résultats obtenus et 
d'interagir avec les données générées par les modèles. Par exemple, des visualisations 
du chemin de distorsion temporelle et de l'espace des filtres convolutifs ont été mises 
à disposition pour aider les chercheurs à mieux analyser les modèles de classification 
des séries temporelles.

En conclusion, ce chapitre met l'accent sur la nécessité de garantir la 
transparence et la reproductibilité dans la recherche scientifique. En 
publiant le code, en documentant les processus et en fournissant des outils 
interactifs, je contribue à renforcer la fiabilité de la recherche et 
à encourager la collaboration au sein de la communauté scientifique. Le 
développement continu de la plateforme \emph{aeon} et la mise en place 
de ressources accessibles montrent un engagement fort envers la création 
d'un écosystème de recherche ouvert et reproductible. Ces efforts garantissent 
que le travail présenté dans cette thèse peut servir de base solide pour de 
futures avancées dans le domaine de l'analyse des séries temporelles.
%
%
\addchap{Introduction}
\label{introduction}

Time series refers to sequential data where data points are equally spaced in time, 
with each point corresponding to a specific timestamp. Time series data is prevalent 
across a wide range of applications, including medical fields such as electrocardiograms 
(ECGs)~\cite{ecg-example-paper} and electroencephalograms (EEGs)~\cite{eeg-example-paper},
human motion~\cite{human-motion-example-paper}, stock
market trends~\cite{stock-exchange-example-paper}, and 
telecommunications signals~\cite{telecom-example-paper} between
base stations and users, among others.
The term ``time'' in time series does not imply that only temporal ordering is relevant; 
rather, any data with a necessary sequence or order can be treated similarly to temporal 
data. For instance, in applications like image contour
extraction~\cite{image-countour-shapelets-paper}, the data has an inherent 
order that, if disrupted, such as by shuffling the image into a jigsaw puzzle, would result 
in a loss of meaningful information. Therefore, this data is treated similarly to
time series data.   
While time series data can, to some extent, be represented in a tabular format, where 
each row corresponds to a time series sample and each column represents a variable at 
a particular timestamp, its analysis cannot be adequately performed using standard tabular 
data analysis methods. Traditional tabular methods fail to consider the sequential nature 
of the data, focusing only on relationships between variables, rather than the critical 
ordering of those variables.

In 2006, time series analysis was recognized as one of the top 10 challenges in the field 
of data mining~\cite{data-mining-challenges-paper}. This area of study involves a 
broad spectrum of tasks that are crucial for understanding temporal data patterns. 
These tasks can be effectively tackled using a variety of approaches, including 
traditional statistical methods, modern machine learning techniques, and advanced 
deep learning models.

\textbf{Forecasting} is a specialized regression task in the domain of time series data, 
aiming at predicting future segments of the input series by utilizing the temporal patterns 
present in the data~\cite{monash-forecasting}.
This task is essential to a wide range of applications, such as weather 
forecasting~\cite{weather-forecasting} and stock market 
prediction~\cite{stock-exchange-example-paper}.

\begin{figure}
    \centering
    \caption{Time Series Extrinsic Regression (TSER) is the task of predicting continuous 
    labels of the time series samples.}
    \label{fig:task-tser}
    \includegraphics[width=\textwidth]{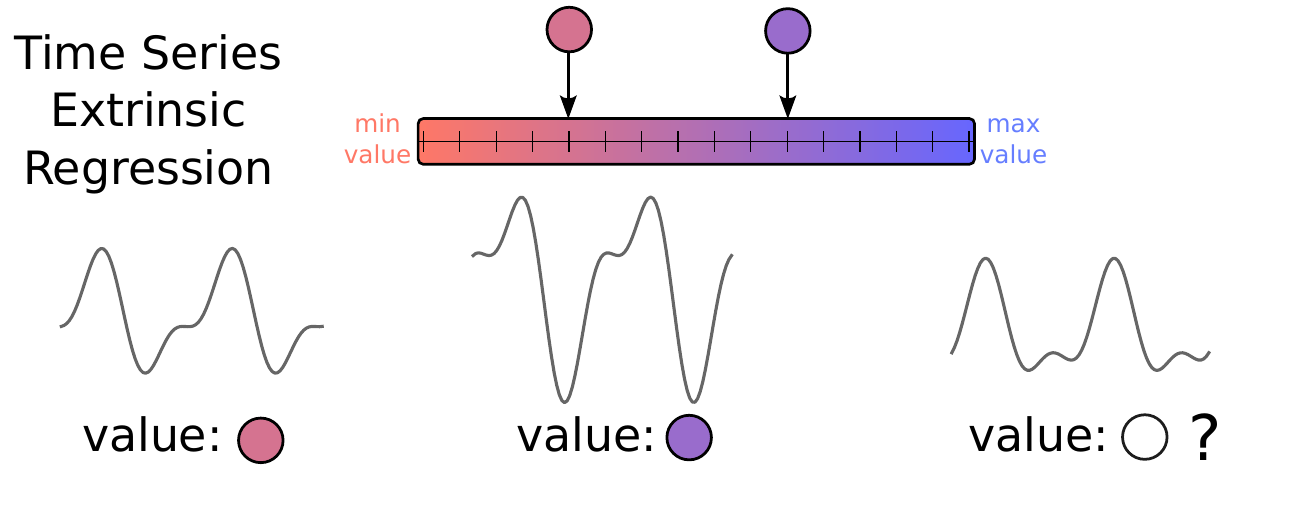}
\end{figure}

\textbf{Extrinsic regression}~\cite{tser-archive}, presented in 
Figure~\ref{fig:task-tser} differs significantly from 
time series forecasting. In this task, the goal is to predict a continuous value 
that is not a future point in the input series, but rather a value generated by 
a random variable that depends on the entire time series, including its trends and values.
This type of regression is commonly used in applications such as 
live fuel moisture content estimation~\cite{tser-live-fuel-moisture,deep-live-fuel} 
and human rehabilitation motion 
assessment~\cite{kimore-paper}.

\textbf{Anomaly detection}~\cite{anomaly-detection-review} in time series 
focuses on identifying data points or patterns that deviate from the expected norm. 
This task is essential for spotting unusual events that may indicate issues like system 
failures or fraud. It is widely applied in areas such as network
security monitoring~\cite{network-anomaly-detection} 
and healthcare diagnostics~\cite{healthcare-anomaly-detection}, where early
detection of anomalies can prevent significant problems.

\begin{figure}
    \centering
    \caption{Time Series CLustering (TSCL) is the task of discovering common information 
    between samples of time series in order to group them into clusters.}
    \label{fig:task-tscl}
    \includegraphics[width=\textwidth]{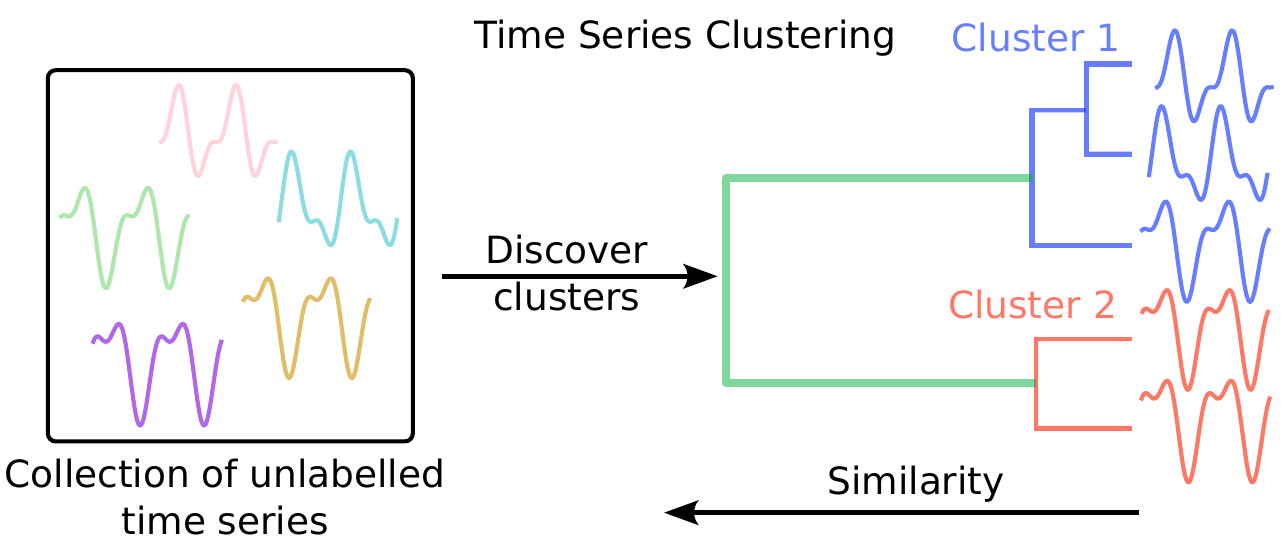}
\end{figure}

\textbf{Clustering}~\cite{deep-tscl-bakeoff}, presented in Figure~\ref{fig:task-tscl},
involves identifying and extracting patterns within the input series to categorize them 
into distinct groups, which are defined by the nature of the data. This approach is 
applied in various fields, such as detecting daily patterns in stock market 
data~\cite{timeseries-clustering-application} and identifying specific 
behaviors in solar magnetic wind~\cite{wind-clustering}.

\begin{figure}
    \centering
    \caption{Time Series Prototyping (TSP) is the task of finding a representative 
    of a collection of time series of a similar group.}
    \label{fig:task-tsp}
    \includegraphics[width=\textwidth]{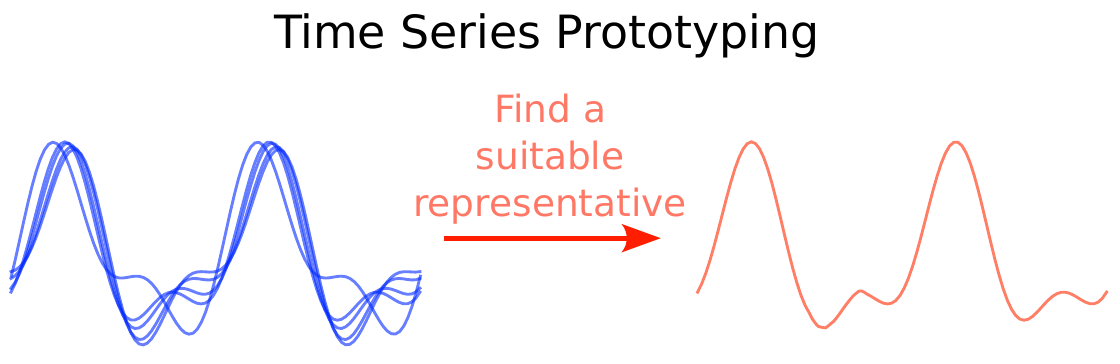}
\end{figure}

\textbf{Prototyping}~\cite{dba-paper}, presented in Figure~\ref{fig:task-tsp}, is 
the process of identifying a representative 
time series from a group of similar series. This task is particularly useful in time 
series clustering, as it helps to summarize and simplify the data by selecting a central 
or typical example from each cluster.
In healthcare, time series prototyping can be valuable for creating a summarized exemplar 
of patients with common health conditions, thereby facilitating comparisons with new
patients~\cite{time-series-snippets}.

\begin{figure}
    \centering
    \caption{Time Series Classification (TSC) is the task of predicting 
    a discrete label of the time series samples.}
    \label{fig:task-tsc}
    \includegraphics[width=\textwidth]{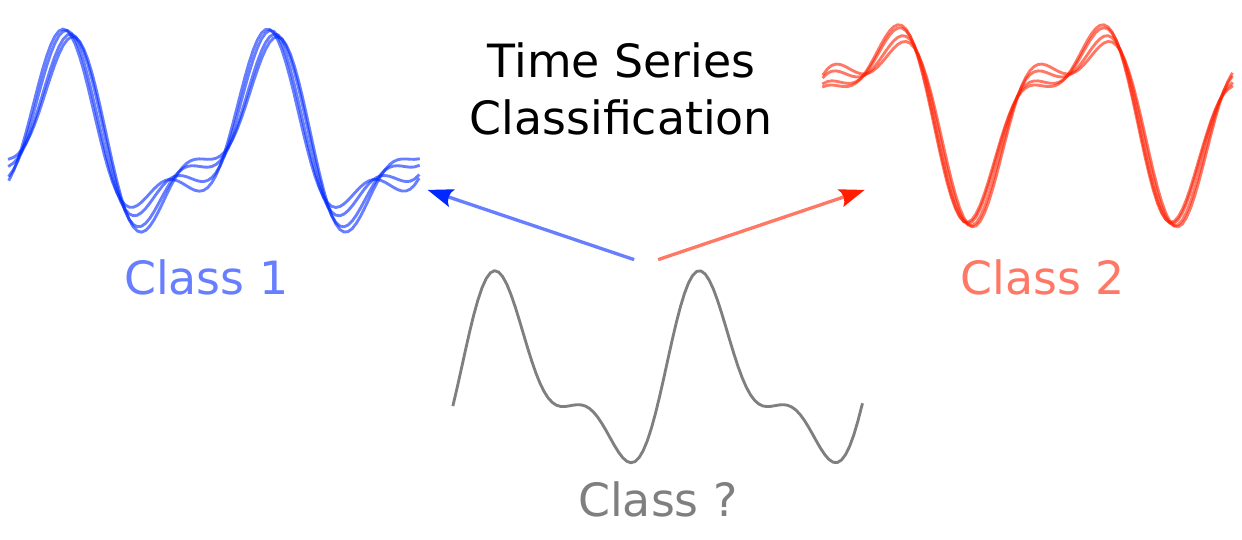}
\end{figure}

\textbf{Classification}~\cite{bakeoff-tsc-2}, is a discretized version 
of extrinsic regression as presented in Figure~\ref{fig:task-tsc}, where
the goal is to predict a discrete class label for each time series with prior knowledge of 
the number of possible classes, which distinguishes it from clustering. This task is 
widely applied in various fields, including human activity 
recognition~\cite{human-motion-example-paper} and medical diagnostics, such as 
classifying heart diseases from ECG signals~\cite{ecg-example-paper}.

In this thesis, we mainly focus on four tasks for time series data: classification, clustering,
extrinsic regression and prototyping.
These tasks are particularly relevant to the application of human motion analysis, where the 
input time series consists of sequences of recorded kinematic skeleton data at each time step.
Such data can be used for classification tasks to predict the activity of a subject and 
for extrinsic regression to predict a continuous value associated to rehabilitation motion to assess 
patients' performance.
More details on this application and the contributions in such domain are presented in 
Chapter~\ref{chapitre_6}.

\begin{figure}
    \centering
    \caption{The number of research papers mentioning ``deep learning'' 
    and ``time series classification''
    increased rapidly in the last years.}
    \label{fig:intro-trend-dl4tsc}
    \includegraphics[width=\textwidth]{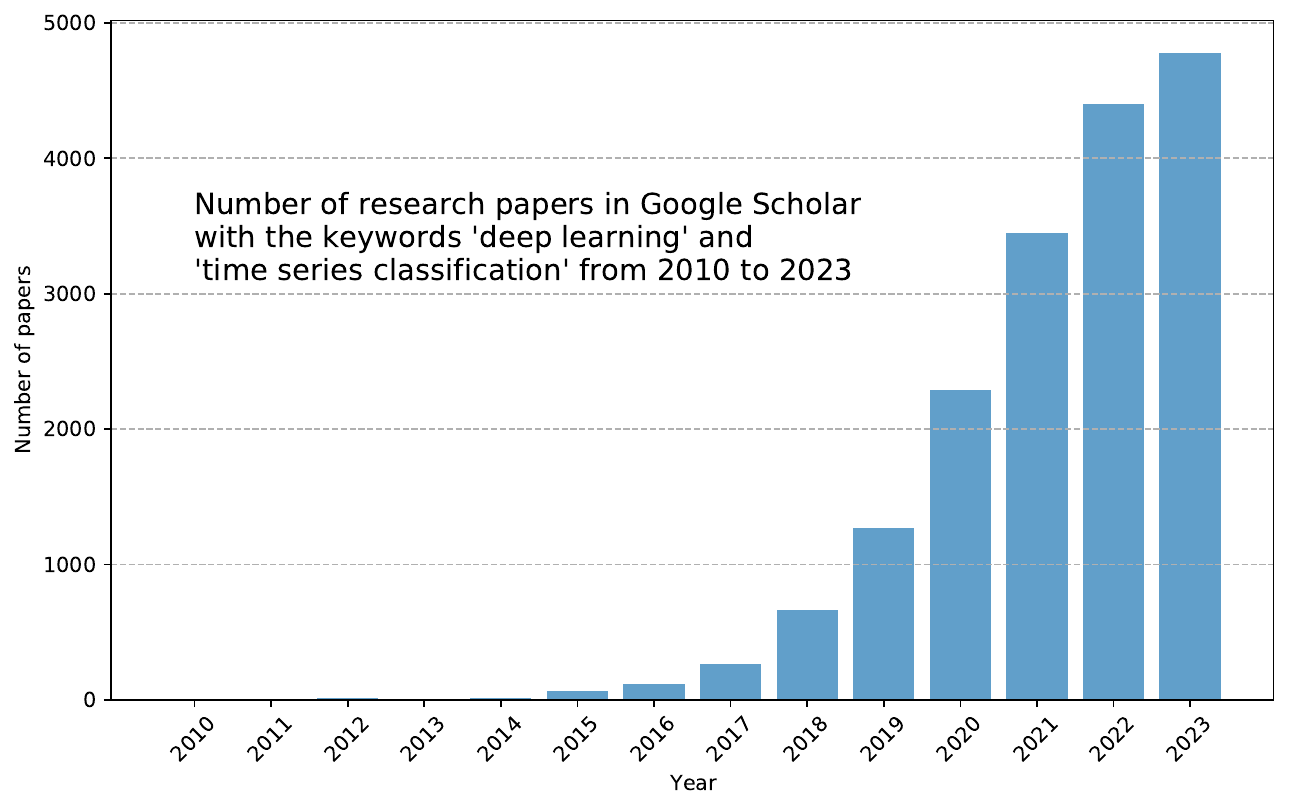}
\end{figure}

In order to address the above mentioned applications, we addressed the approached 
perspective of time series analysis, starting with a study of
traditional methods that have been used for years to solve Time Series Classification (TSC).
However, it was shown in the first 2017 TSC bake-off~\cite{bakeoff-tsc-1} that using 
the traditional techniques does not achieve state-of-the-art performance, instead the authors found
that hybrid approaches work much better.
Moreover, the domain of TSC was then extended by different methods proposed in between, ranging from convolution
methods~\cite{rocket} to bag-of-words methods~\cite{weasel2}.
After the publication of the 2017 bake-off, researchers began to question the role of deep learning 
models in this domain, especially given their significant performance in image 
classification~\cite{lecun2015deep,alexnet-paper}. The number of related papers in deep learning for TSC started to 
increase rapidly, leading to the 2019 deep learning for TSC review~\cite{dl4tsc}. The 2019 
review demonstrated that the best deep learning model achieved performance comparable to 
the state-of-the-art non-deep learning model.
\citet{dl4tsc} paved the way for new research in deep learning for time series classification (TSC).
This trend, illustrated in Figure~\ref{fig:intro-trend-dl4tsc},
also extends to addressing other tasks within the time series domain.
For instance, deep learning methods for Time Series Clustering (TSCL) was reviewed in~\cite{deep-tscl-bakeoff}
showcasing that deep TSCL methods can outperform traditional elastic methods or shape based methods~\cite{kshape-paper}.
Additionally, the usage of deep learning emerged for the task of Time Series Prototyping (TSP)
with the usage of multi-task Auto-Encoders~\cite{deep-averaging-paper}.
Finally,~\citet{deep-tsc-tser} showed that numerous research papers are addressing 
the task of deep learning for Time Series Extrinsic Regression (TSER).
A significant amount of literature work on these topics are presented in Chapter~\ref{chapitre_1}.

Given the growing interest and proven effectiveness of deep learning in time series analysis, we 
employ
this approach to tackle the four tasks of TSC, TSCL, TSP, and TSER. Deep learning's ability 
to capture complex patterns and dependencies in sequential data makes it well-suited for addressing 
these challenges.
However, before presenting any contribution in this thesis in these four tasks
, we start 
by addressing the evaluation of discriminative models.
The current evaluation framework, even though having its own advantages,
presents some limitations that are beneficial to any research work that want to 
manipulate the ``view'' of the results to make it seem better than other approaches.
For this reason, we propose, not a replacement, but a complimentary tool
for such an evaluation framework, that we present in Chapter~\ref{chapitre_2}.

After establishing the evaluation framework and recognizing the growing interest in developing  
foundation models for time series data~\cite{foundation-models-example-paper}, we introduce 
in Chapter~\ref{chapitre_3} two key
contributions. These 
contributions converge to form 
a novel approach aiming at defining such a foundation model, specifically tailored 
for the task of TSC. Chapter~\ref{chapitre_3} not only outlines the foundation model but also 
explains how it tackles these unique challenges through the engineering of 
hand-crafted features, paving the way for more robust and generalized models in this domain.

Focusing on the carbon footprint of such complex deep learning models for time series data,
we propose in Chapter~\ref{chapitre_4} to reduce model complexity while keeping the performance 
statistically non significantly worse than the state-of-the-art.
This is done by the proposal of LITE,
the smallest deep learning model for time series data found in the 
literature, that is proven to be very effective in the 
second TSC bake-off~\cite{bakeoff-tsc-2}.

Moreover, acquiring labeled data in time series can be challenging. To address this, 
in Chapter~\ref{chapitre_5}, we propose an unsupervised framework
designed to handle 
situations where only a limited number of labeled samples are available. This 
unsupervised framework, based on representation learning, can be applied to various 
downstream tasks involving time series data.

\begin{table}[h!]
    \hspace*{-1.6cm}
    \centering
    \begin{tabular}{|c|c|c|c|}
    \hline
    \textbf{Contribution}  & \textbf{Task} & \textbf{Chap.} & \textbf{GitHub} \\ \hline
    \cite{mcm-paper}  & Evaluation & \ref{chapitre_2} & \href{https://github.com/MSD-IRIMAS/Multi_Comparison_Matrix}{Multi\_Comparison\_Matrix} \\ \hline
    \cite{hand-crafted-filters-paper}  & Classification & \ref{chapitre_3} & \href{https://github.com/MSD-IRIMAS/CF-4-TSC}{CF-4-TSC} \\ \hline
    \cite{pretext-task-paper}  & Classification & \ref{chapitre_3} & \href{https://github.com/MSD-IRIMAS/DomainFoundationModelsTSC}{DomainFoundationModelsTSC} \\ \hline
    \cite{lite-paper}  & Classification & \ref{chapitre_4} & \href{https://github.com/MSD-IRIMAS/LITE}{LITE} \\ \hline
    \cite{lite-extension-paper}  & Classification & \ref{chapitre_4} & \href{https://github.com/MSD-IRIMAS/LITE}{LITE} \\ \hline
    \cite{trilite-paper}  & \begin{tabular}{c}Self-Supervised/ \\ Classification \end{tabular} & \ref{chapitre_5} & \href{https://github.com/MSD-IRIMAS/TRILITE}{TRILITE} \\ \hline
    \cite{shape-dba-paper}  & Prototyping & \ref{chapitre_6} & \href{https://github.com/MSD-IRIMAS/ShapeDBA}{ShapeDBA} \\ \hline
    \cite{weighted-shape-dba-paper}  & \begin{tabular}{c}Prototyping/\\Regression\end{tabular} & \ref{chapitre_6} & \href{https://github.com/MSD-IRIMAS/Weighted-ShapeDBA-4-Rehab}{Weighted-ShapeDBA-4-Rehab} \\ \hline
    \cite{svae-paper}  & \begin{tabular}{c}Generation/\\Classification \end{tabular} & \ref{chapitre_6} & \href{https://github.com/MSD-IRIMAS/SVAE-4-HMG}{SVAE-4-HMG} \\ \hline
    \cite{metrics-paper}  & \begin{tabular}{c}Generation/\\Evaluation\end{tabular} & \ref{chapitre_7} & \href{https://github.com/MSD-IRIMAS/Evaluating-HMG}{Evaluating-HMG} \\ \hline
    \cite{aeon-paper}  & All & \ref{chapitre_8} & \href{https://github.com/aeon-toolkit/aeon}{aeon} \\ \hline
    \cite{ismail-fawaz2024elastic-vis}  & Visualization & \ref{chapitre_8} & \href{https://github.com/MSD-IRIMAS/Elastic_Warping_Vis}{Elastic\_Warping\_Vis} \\ \hline
    \cite{Ismail-Fawaz2023weighted-ba}  & Prototyping & \ref{chapitre_8} & \href{https://github.com/MSD-IRIMAS/Augmenting-TSC-Elastic-Averaging}{Augmenting-TSC-Elastic-Averaging} \\ \hline
    \cite{Ismail-Fawaz2023kan-c22-4-tsc}  & Classification & \ref{chapitre_8} & \href{https://github.com/MSD-IRIMAS/Simple-KAN-4-Time-Series}{Simple-KAN-4-Time-Series} \\ \hline
    
    \end{tabular}
    \caption{List of contributions including 11 papers and 3 open source published work
    with the companion GitHub repository.}
    \label{tab:intro-contributions}
\end{table}

Given that this thesis is conducted within the framework of the ANR JCJ DELEGATION (more details in
Section~\nameref{financements}), 
which targets the analysis of human motion, Chapter~\ref{chapitre_6} addresses the specific challenges 
of human motion data. It highlights the unique characteristics of this data and 
demonstrates how our contributions are particularly effective in this domain, in 
line with the project's objectives.
We demonstrate how using small 
models like LITE, optimized for human motion, can achieve state-of-the-art performance 
in rehabilitation assessment within a classification framework, while also being 
effective in terms of medical explainability. Additionally, we introduce a novel 
TSP approach used as a generative method 
for enhancing extrinsic regression models 
in rehabilitation motion assessment tasks.
Moreover, considering the rise 
generative models for human motion~\cite{action2motion-paper,actor-paper} and 
the strong performance of CNNs in the time series domain~\cite{inceptiontime-paper},
we propose a deep 
generative model with a CNN backbone for human motion data
that nearly matches state-of-the-art results.
Given we focus on the evaluation framework for discriminative models in Chapter~\ref{chapitre_2}, 
we argue in Chapter~\ref{chapitre_7} for the necessity of a unified framework specifically for 
generative models, particularly in the context of human motion data.

In Chapter~\ref{chapitre_8}, we conclude by discussing the importance of 
reproducible research, offering 
a professional perspective on this critical aspect of scientific inquiry. We highlight the 
contributions of this thesis to the open-source Python package \textit{aeon}~\cite{aeon-paper}, 
as well as several other open-source projects developed during the course of this research, 
with or without accompanying publications.

In Table~\ref{tab:intro-contributions} we present all of the contributions in this thesis,
including 11 papers and 3 open source projects.
All of our research work is based on using publicly available datasets, including the UCR
archive~\cite{ucr-archive} for univariate setups, the UEA archive~\cite{uea-archive}
for multivariate setups and both 
the HumanAct12~\cite{action2motion-paper} and Kimore~\cite{kimore-paper} datasets
for human motion applications of activity recognition and 
rehabilitation assessment.

\section*{Publications}


\subsection*{International Journals}

\begin{itemize}
    \item Middlehurst, Matthew, \textbf{Ali Ismail-Fawaz}, Antoine Guillaume, 
    Christopher Holder, David Guijo Rubio, Guzal Bulatova, Leonidas Tsaprounis, Lukasz Mentel, 
    Martin Walter, Patrick Schäfer, Anthony Bagnall
    (2024) ``aeon: a Python toolkit for learning from time series''.
    In \textcolor{orange}{Journal Machine Learning Research (JMLR), open source software track}.
    doi: \url{https://arxiv.org/abs/2406.14231}
\end{itemize}

\subsection*{International Conferences and Workshops}
\begin{itemize}
    \item \textbf{Ismail-Fawaz, Ali}, Maxime Devanne, 
    Jonathan Weber, and Germain Forestier. (2022). 
    ``Deep learning for time series classification using 
    new hand-crafted convolution filters''. In \textcolor{orange}{IEEE International Conference on Big Data (Big Data)}
    (pp. 972-981). IEEE. doi: \url{https://doi.org/10.1109/BigData55660.2022.10020496}
    
    \item \textbf{Ismail-Fawaz, Ali}, Maxime Devanne, Jonathan Weber, 
    and Germain Forestier. (2023). ``Enhancing time series classification with self-supervised learning''.
    In \textcolor{orange}{International Conference on Agents and Artificial Intelligence (ICAART)}
    (pp. 40-47). SCITEPRESS-Science and Technology Publications. doi: \url{https://doi.org/10.5220/0011611300003393}
   
    \item \textbf{Ismail-Fawaz, Ali}, Hassan Ismail Fawaz, François Petitjean, 
    Maxime Devanne, Jonathan Weber, Stefano Berretti, 
    Geoffrey I. Webb, and Germain Forestier. 
    (2023). ``ShapeDBA: generating effective time series prototypes using 
    shapeDTW barycenter averaging''. In \textcolor{orange}{International Workshop on Advanced Analytics and 
    Learning on Temporal Data; in conjunction with the European Conference on Machine Learning and Principles 
    and Practice of Knowledge Discovery in Databases} (pp. 127-142). Cham: Springer Nature Switzerland.
    doi: \url{https://doi.org/10.1007/978-3-031-49896-1_9}

    \item \textbf{Ismail-Fawaz, Ali}, Maxime Devanne, Stefano Berretti, 
    Jonathan Weber, and Germain Forestier. (2023). 
    ``Lite: Light inception with boosting techniques for time series 
    classification''. In \textcolor{orange}{IEEE 10th International Conference on Data Science and Advanced 
    Analytics (DSAA)} (pp. 1-10). IEEE. doi: \url{https://doi.org/10.1109/DSAA60987.2023.10302569}
    
    \item \textbf{Ismail-Fawaz, Ali}, Maxime Devanne, Stefano Berretti, 
    Jonathan Weber, and Germain Forestier. (2024). 
    ``Finding foundation models for time series classification with a pretext task''.
    In \textcolor{orange}{Pacific-Asia Conference on Knowledge Discovery and Data Mining} 
    (pp. 123-135). Singapore: Springer Nature Singapore. doi: \url{https://doi.org/10.1007/978-981-97-2650-9_10}
    
    \item \textbf{Ismail-Fawaz, Ali}, Maxime Devanne, Stefano Berretti, 
    Jonathan Weber, and Germain Forestier. (2024). 
    ``A Supervised Variational Auto-Encoder for Human Motion 
    Generation using Convolutional Neural Networks''. 
    In \textcolor{orange}{4th International Conference on Pattern 
    Recognition and Artificial Intelligence (ICPRAI)}
    
    \item \textbf{Ismail-Fawaz, Ali}, Maxime Devanne, Stefano Berretti, 
    Jonathan Weber, and Germain Forestier. (2024). 
    ``Weighted Average of Human Motion Sequences for Improving Rehabilitation Assessment''.
    In \textcolor{orange}{International Workshop on Advanced Analytics and 
    Learning on Temporal Data (AALTD); in conjunction with the European Conference on Machine Learning and Principles 
    and Practice of Knowledge Discovery in Databases(ECML/PKDD)}.
    doi: \url{https://ecml-aaltd.github.io/aaltd2024/articles/Fawaz_AALTD24.pdf}
\end{itemize}

\subsection*{National Conferences}
\begin{itemize}
    \item \textbf{Ismail-Fawaz, Ali}, Maxime Devanne, 
    Jonathan Weber, and Germain Forestier. (2023). ``Apprentissage en Profondeur pour la Classification des Séries Temporelles 
    à l'aide de Nouveaux Filtres de Convolution Créés Manuellement''. In \textcolor{orange}{ORASIS}.
    doi: \url{https://hal.science/hal-04219450/document}
    
    \item \textbf{Ismail-Fawaz, Ali}, Maxime Devanne, Stefano Berretti, 
    Jonathan Weber, and Germain Forestier. (2024). ``LITE: Light Inception avec des Techniques de Boosting pour la Classification 
    de Séries Temporelles''. In \textcolor{orange}{Extraction et Gestion des Connaissances (EGC)} (pp. 377-384).
    doi: \url{https://editions-rnti.fr/?inprocid=1002948}
    
    \item \textbf{Ismail-Fawaz, Ali}, Maxime Devanne, Stefano Berretti, 
    Jonathan Weber, and Germain Forestier. (2023). ``Reducing Complexity of Deep Learning for Time Series Classification 
    Using New Hand-Crafted Convolution Filters''. \textcolor{orange}{Upper Rhine Artificial Intelligence Symposium (URAI)}.
    doi: \url{https://maxime-devanne.com/publis/ismail-fawaz_urai2023.pdf}
\end{itemize}

\section*{Under Submission}

\subsection*{International Journals}
\begin{itemize}
    \item \textbf{Ismail-Fawaz, Ali}, Maxime Devanne, Stefano Berretti, 
    Jonathan Weber, and Germain Forestier. ``Look Into the LITE in Deep Learning for Time Series Classification''.
    In \textcolor{orange}{International Journal of Data Science and Analytics}.
    doi: \url{https://arxiv.org/abs/2409.02869}
    \item \textbf{Ismail-Fawaz, Ali}, Maxime Devanne, Stefano Berretti, 
    Jonathan Weber, and Germain Forestier. ``Establishing a Unified Evaluation Framework for Human Motion Generation:
    A Comparative Analysis of Metrics''. In \textcolor{orange}{Computer Vision and Image Understanding}.
    doi: \url{https://arxiv.org/abs/2405.07680}
\end{itemize}

\section*{Arxiv}
\begin{itemize}
    \item \textbf{Ismail-Fawaz, Ali}, Angus Dempster, Chang Wei Tan, 
    Matthieu Herrmann, Lynn Miller, Daniel F. Schmidt, 
    Stefano Berretti, Jonathan Weber, Maxime Devanne,
    Germain Forestier, and Geoffrey I. Webb. (2023). ``An approach to multiple comparison benchmark evaluations that 
    is stable under manipulation of the comparate set''. doi: \url{https://arxiv.org/abs/2305.11921}
\end{itemize}

%
%
\chapter{State Of The Art For Time Series Analysis: Supervised and Unsupervised Learning} 
\label{chapitre_1}

%
\newcommand{\keyword}[1]{\textbf{#1}}
\newcommand{\tabhead}[1]{\textbf{#1}}
\newcommand{\code}[1]{\texttt{#1}}
\newcommand{\file}[1]{\texttt{\bfseries#1}}
\newcommand{\option}[1]{\texttt{\itshape#1}}
\newcounter{mydefinitioncounter}
\newcommand\mydefinition{\stepcounter{mydefinitioncounter}\paragraph*{Definition \arabic{mydefinitioncounter}}}
%

%
%

\section{Introduction}

Time series analysis, a critical aspect of data science, leverages both supervised and unsupervised
learning methods to extract meaningful insights from time-dependent data. In supervised learning, 
the goal is to predict future values based on past observations. This includes tasks such as extrinsic regression, 
where continuous future values are forecasted, and classification, where future events are categorized 
into predefined classes. Examples include predicting stock prices or classifying email as spam or regular 
based on historical data.
Unsupervised learning, on the other hand, involves discovering inherent structures or patterns within 
the data without predefined labels. Key tasks include clustering, where similar data points are grouped 
together, and anomaly detection, which identifies unusual patterns that deviate from the norm. 
Applications of these techniques range from segmenting customers based on purchasing behavior 
to detecting fraudulent transactions in financial systems. By employing both supervised and unsupervised 
learning, time series analysis can effectively address a wide array of predictive and descriptive tasks, 
driving informed decision-making across various fields.

In the rest of this chapter, we will detail the state-of-the-art literature for both cases.
For supervised learning, we will explore classification and extrinsic regression techniques, while for 
unsupervised learning, we will focus on clustering, prototyping, and self-supervised methods. 
This comprehensive review aims to provide a thorough understanding of the latest advancements 
and applications in time series analysis.

\section{Supervised Learning: Time Series Classification and Extrinsic Regression}

This subsection covers two main tasks: classification, which categorizes time series data into predefined 
classes, and extrinsic regression, which predicts continuous values. We will review state-of-the-art models and 
techniques for these tasks, discussing their applications, strengths, and limitations.

\subsection{Time Series Classification}

The task of TSC has been addressed for the last three decades in various approaches ranging from distance
based approaches to recent deep learning methods.
Such type of data can be found in various domains ranging from human activity recognition~\cite{human-motion-example-paper} to wireless communication~\cite{bertalanivc2022resource}.
With the availability of new TSC datasets, a significant amount of models has been proposed in the literature.
Collecting such data and preprocessing them to become available for benchmarking is not a simple task, for this reason
the UCR/UEA~\cite{ucr-archive,uea-archive} archives had such a significant impact in the last decade on the amount of research in the TSC field.

In this section, we present the prerequisite definitions needed to understand all the materials.
We follow these definitions by an extensive detail view over some state-of-the-art models in both deep and non deep learning methods for TSC.

\mydefinition A Univariate Time Series (UTS) $\textbf{x} = \{x_1,x_2,\ldots,x_L\}$ is a sequence of ordered real values.
The length of this sequence is $L$.
\mydefinition A Multivariate Time Series (MTS) of $M$ dimensions (also referred to as channels) $\textbf{x}=\{\textbf{x}^1,\textbf{x}^2,\ldots,\textbf{x}^M\}$ is a set of $M$ univariate time series of length $L$,
where $\textbf{x}^m=\{x_1^m,x_2^m,\ldots,x_L^m\}$ is a univariate series of length $L$
and $\textbf{x}_t=\{x_t^1,x_t^2,\ldots,x_t^M\}$ a one dimensional vector of shape $(M,)$, $m~\in~[1,M]$ and $t=[1,L]$.
\mydefinition A TSC dataset $\mathcal{D}=\{\textbf{x}_i,\textbf{y}_i\}_{i=1}^N$
is a collection of $N$ pairs of time series and their corresponding label $\textbf{y}_i$ where
$\textbf{x}_i=\{\textbf{x}_{i,1},\textbf{x}_{i,2},\ldots,\textbf{x}_{i,L}\}$ is an MTS of $M$ dimensions and length $L$.
The label $\textbf{y}_i$ is a vector of length $C$ where $C$ is the number of possible classes in $\mathcal{D}$.
Each element $c~\in~[1,C]$ in $\textbf{y}_i$ is one if $\textbf{x}_i$ belongs to class $c$ and zero otherwise.

The task of TSC comes down to constructing a model $\mathcal{F}$ that can achieve correct predictions of 
labels associated to each time series in the dataset.
This is done by teaching the model how to predict a discrete probability distribution of $C$ elements with the goal of having the highest probability assigned to the correct class.
\begin{equation}\label{equ:classif-proba}
    \mathcal{F}(\textbf{x}) = [p_1,p_2,\ldots,p_C]
\end{equation}
\noindent where $\sum_{c=1}^C p_c = 1$ and $0 \leq p_c \leq 1$.

\begin{figure}
    \centering
    \caption{Eight sections of \protect\mycolorbox{0,66,58,0.25}{Time Series Classification} models from the literature:
    \protect\mycolorbox{181,133,199,0.35}{shapelet based},
    \protect\mycolorbox{243,102,74,0.35}{convolution based},
    \protect\mycolorbox{0,189,199,0.35}{distance based},
    \protect\mycolorbox{0,189,87,0.35}{feature based},
    \protect\mycolorbox{0,58,199,0.35}{interval based},
    \protect\mycolorbox{191,130,120,0.35}{dictionary based},
    \protect\mycolorbox{243,120,199,0.35}{hybrid based} and 
    \protect\mycolorbox{255,0,0,0.35}{deep learning based}.
    }
    \includegraphics[width=\textwidth]{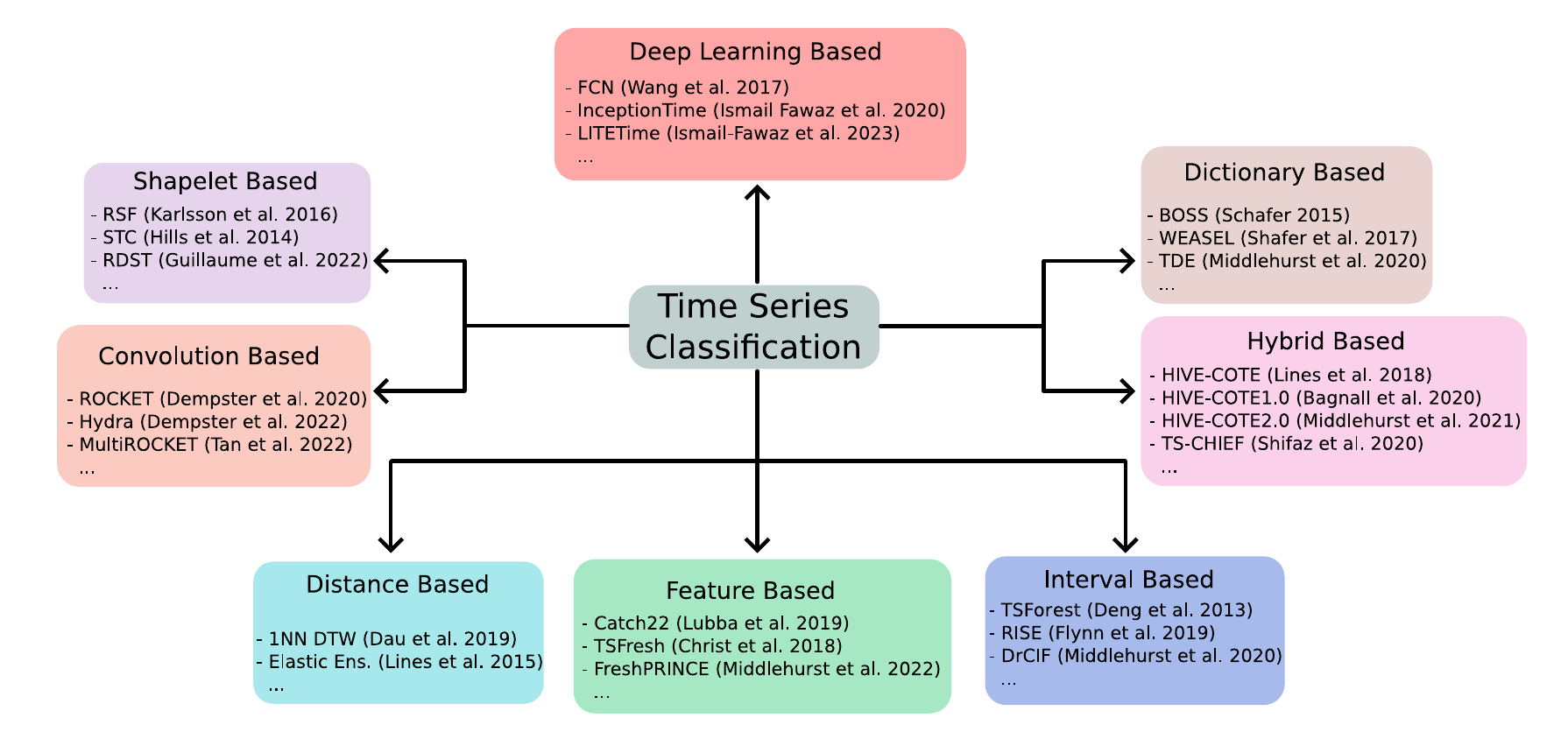}
    \label{fig:bakeoff}
\end{figure}

For many years, the most famous approach known in the literature to address TSC was the use of Nearest Neighbor (NN) coupled
with Dynamic Time Warping (DTW) similarity measure and was used as a baseline~\cite{bakeoff-tsc-1}.
Some work also tried to address an adaptation of Support Vector Machines (SVMs)~\cite{svm-paper} for TSC, such as the usage 
of edit distance kernels~\cite{marteau2014recursive, cuturi2007kernel}.
Ever since the release of the first TSC review by~\cite{bakeoff-tsc-1}, much more classifiers have been published.
With the rise of available data,~\cite{dl4tsc} presents a detailed review over all deep learning models from the literature
addressed for the task of TSC and evaluated them on the UCR/UEA archives.
The 2019 deep learning for TSC review~\cite{dl4tsc} highlighted the importance of deep learning models that were 
missed in the TSC review of~\cite{bakeoff-tsc-1}, showcasing their competitive performance
with non-deep learning models.
Moreover, the number of TSC models have increased significantly, which led to the second TSC review~\cite{bakeoff-tsc-2}.
The models in the literature can be divided into eight different sections based on the method used to solve the TSC task.
These sections, presented in Figure~\ref{fig:bakeoff}, are: \textbf{distance based methods}, \textbf{feature based methods}, \textbf{interval based methods},
\textbf{dictionary based methods}, \textbf{convolution based methods}, \textbf{shapelet based methods}, \textbf{hybrid based methods}
and \textbf{deep learning based methods}.

In this section, we go through some approaches of solving the task of TSC with non-deep learning methods.

\subsubsection{Distance Based Methods}\label{sec:tsc-distance}

In this section we present the distance based methods to solve the TSC task which utilizes measures such as DTW or MSM~\cite{msm-distance} etc.

\paragraph{$k$-Nearest Neighbor - Dynamic Time Warping ($k$-NN-DTW) and Variants}

As mentioned before, the most famous method to solve TSC was based on the $k$-NN algorithm coupled with a similarity measure.
While for other types of data $k$-NN is coupled with the Euclidean Distance (ED) presented in Eq.~\ref{equ:ED}, it does not capture the temporal
aspect of time series.
\begin{equation}\label{equ:ED}
    ED(\textbf{x}_1,\textbf{x}_2) = \sqrt{\sum_{t=1}^L\sum_{m=1}^M (x^m_{1,t}-x^m_{2,t})^2}
\end{equation}
For instance if we have two time series $\textbf{x}_1=[1,1,0,0,0,1,0]$ and $\textbf{x}_2=[0,1,1,0,0,0,1]$, ED would 
produce a value of $2$ however the series are identical with a simple shift of one time stamp between them.
For this reason, DTW was proposed in order to capture this kind of temporal distortion.
DTW finds the optimal alignment path between two time series before applying the Minkowski (ED if $q=2$) over the aligned series.
The mathematical formulation of DTW is as follows:
\begin{equation}\label{equ:dtw}
    DTW_q(\textbf{x}_1,\textbf{x}_2) = \min_{\pi\in\mathcal{A}(\textbf{x}_1,\textbf{x}_2)}(\sum_{(t_1,t_2)\in\pi}\sum_{m=1}^M(x^m_{1,t_1}-x^m_{2,t_2})^q)^{1/q}
\end{equation}
\noindent where $\pi$ is an alignment path of length $L_{\pi}$ and is a sequence of $L_{\pi}$ pairs of indices
$[(t_{11},t_{21}),(t_{12},t_{22}),\ldots,(t_{1,L_{\pi}},t_{2,L_{\pi}})]$.
$\mathcal{A}(\textbf{x}_1,\textbf{x}_2)$ is the set of all acceptable paths between the two series.
A path $\pi$ is considered acceptable if:
\begin{enumerate}
    \item Start and ending point match the ones of the series:
    \begin{itemize}
        \item $\pi_1 = (1,1)$
        \item $\pi_{L_{\pi}} = (L_1,L_2)$
    \end{itemize}
    \item The sequence is monotonically increasing:
    \begin{itemize}
        \item $t_{1,l-1} \leq i_l \leq t_{1,l-1}+1$
        \item $t_{2,l-1} \leq j_l \leq t_{2,l-1}+1$
    \end{itemize}
    where $l\in[1,L_{\pi}]$
\end{enumerate}
The distance is usually set to the squared error so $q=2$.
A detailed view of the DTW algorithm is presented in Algorithm~\ref{alg:dtw}.
As presented in the detailed algorithm, for each element in the distance matrix, the squared error between the current time stamps first fills the matrix's cell.
Second, at each cell, the smallest element between its three neighbors is added, the upper neighbor indicates inserting a time stamp from one series to another, the left neighbor indicates deleting an element from one series and the bottom neighbor indicates that these two time stamps are aligned so no need for an operation.
The time complexity of the DTW algorithm is $\mathcal{O}(L_1.L_2)$ and $\mathcal{O}(L^2)$ if both series are of the same length.
This complexity is considered very high, and when coupled with NN the whole complexity is $\mathcal{O}(N_{train}.N_{test}.L^2)$,
however some work has optimized such complexity by defining a lower bound for DTW~\cite{lb-keogh-paper,lb-webb-paper}.

\begin{algorithm}
    \caption{Dynamic Time Warping (DTW)}
    \label{alg:dtw}
    \begin{algorithmic}[1]
        \REQUIRE Two Time Series $\textbf{x}_1$ and $\textbf{x}_2$ of length $L$ and dimension $M$
        \ENSURE DTW measure between \(\textbf{x}_1\) and \(\textbf{x}_2\)
        
        \STATE $D = array [L+1,L+1]$
        \FOR{\(t_1 = 1\) to \(L+1\)}
            \FOR{\(t_2 = 1\) to \(L+1\)}
                \STATE $D[t_1,t_2] = +\infty$
            \ENDFOR
        \ENDFOR

        \STATE D[0,0] = 0.0

        \FOR{\(t_1 = 2\) to \(L+1\)}
            \FOR{\(t_2 = 2\) to \(L+1\)}
                \STATE $cost = \sum_{m=1}^M(x^m_{1,t_1-1}-x^m_{2,t_2-1})^2$
                \STATE $up\_insertion = D[t_1-1,t_2]$
                \STATE $left\_deletion = D[t_1,t_2-1]$
                \STATE $diagonal\_match = D[t_1-1,t_2-1]$
                \STATE $D[t_1,t_2] = cost + min(up\_insertion,left\_deletion,diagonal\_match)$
            \ENDFOR
        \ENDFOR

        \STATE \textbf{Return:} $D[L+1,L+1]$
        
    \end{algorithmic}
\end{algorithm}

\begin{figure}
    \centering
    \caption{Example of DTW alignment path computation between two series 
    (\protect\mycolorbox{255,0,0,0.5}{in red} and 
    \protect\mycolorbox{0,0,255,0.5}{in blue}) from the 
    \texttt{ItalyPowerDemand} dataset of the UCR archive.
    The DTW optimal alignment path between both series 
    is presented \protect\mycolorbox{128,128,128,0.7}{in gray}.
    }
    \label{fig:dtw-example-uni}
    \includegraphics[width=0.7\textwidth]{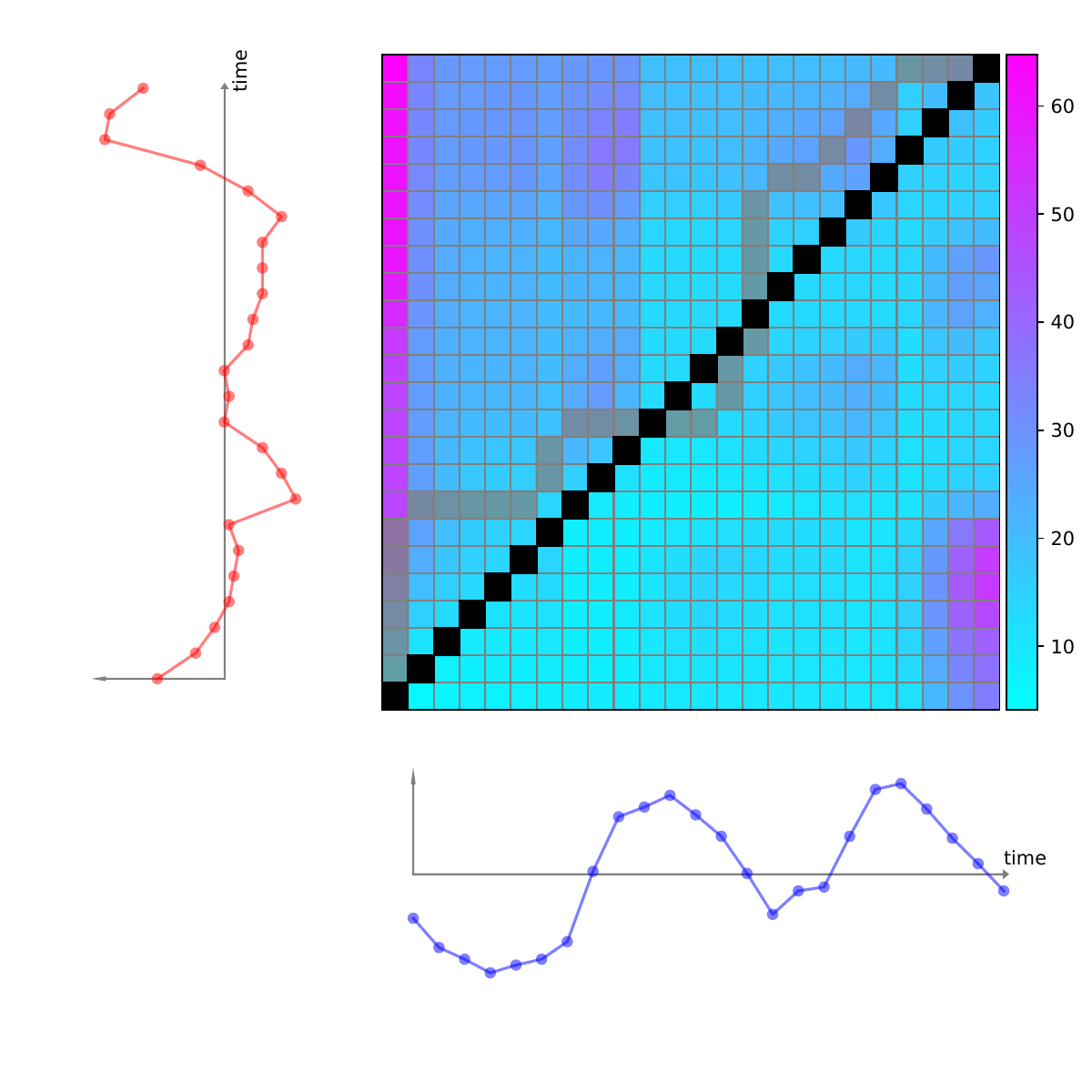}
\end{figure}

\begin{figure}
    \centering
    \caption{DTW optimal \protect\mycolorbox{0,128,0,0.7}{alignment} vs the ED's assumption of a perfect \protect\mycolorbox{0,128,0,0.7}{alignment} on two time series of the \texttt{ItalyPowerDemand} dataset of the UCR archive.}
    \label{fig:dtw-ed-alignment}
    \subfloat[Alignment using ED]{\includegraphics[width=0.5\textwidth]{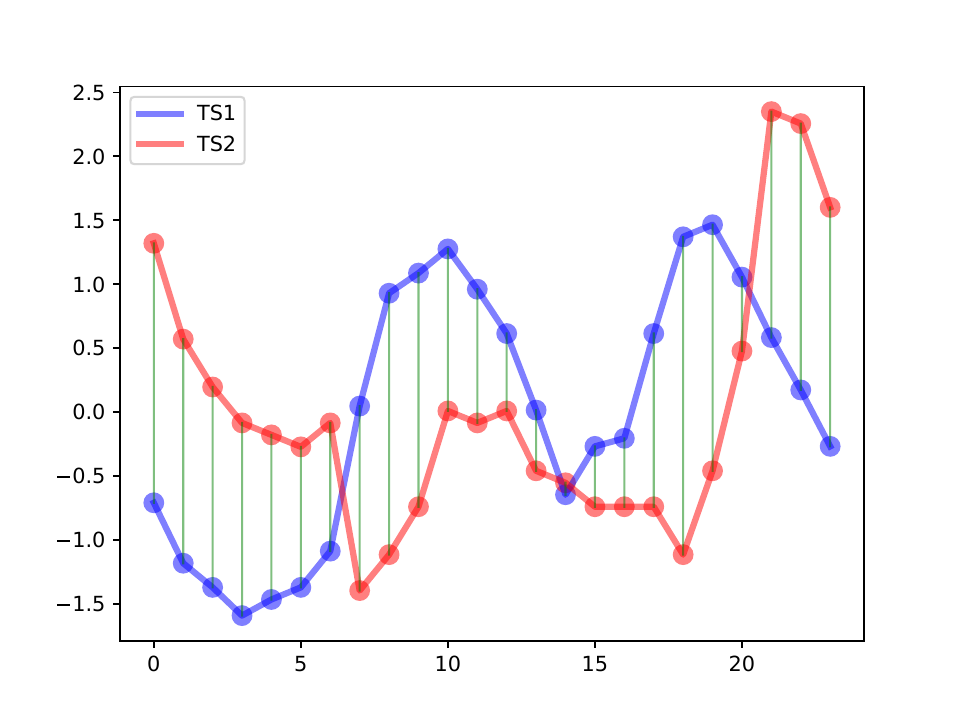}}
    \subfloat[Alignment using DTW]{\includegraphics[width=0.5\textwidth]{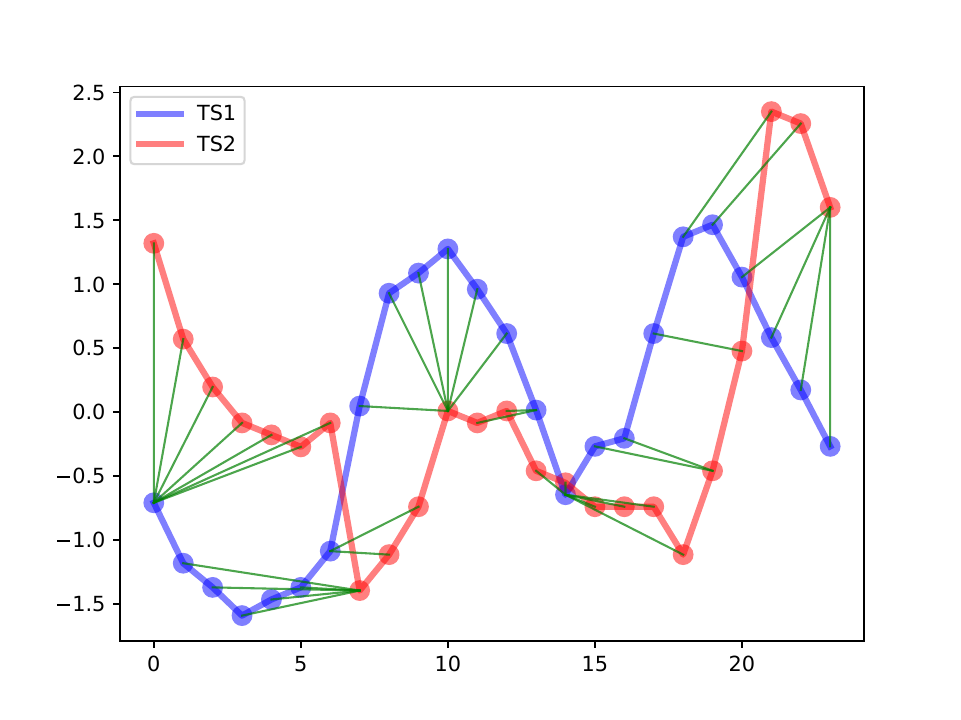}}
\end{figure}

An example of DTW alignment path computation between two time series of the \texttt{ItalyPowerDemand} dataset of the UCR archive~\cite{ucr-archive} is presented in Figure~\ref{fig:dtw-example-uni}.
To showcase the need of a DTW alignment instead of simply using a Euclidean Distance that assume a perfect alignment, we present in Figure~\ref{fig:dtw-ed-alignment} for the same series used in Figure~\ref{fig:dtw-example-uni} both the perfect alignment that ED assumes vs the DTW alignment.

The DTW measure is then utilized to calculate the similarity between each testing sample to all training samples, the predicted label for the test sample is the same as its $k$ nearest neighbors following the used similarity measure.
This algorithm has been developed over the years by simply fine-tuning the parameters of DTW, or by changing the similarity measure.
For instance, many versions of DTW have been proposed over the years, such as SoftDTW~\cite{soft-dtw-distance} and ShapeDTW~\cite{shape-dtw-distance}.

The SoftDTW~\cite{soft-dtw-distance} version addresses the issue of differentiability of DTW
especially because of the minimization step in Algorithm~\ref{alg:dtw}.
The authors of SoftDTW~\cite{soft-dtw-distance} argues the need of a differentiable 
DTW in order to be able to construct an optimization problem used for many applications such as clustering and deep learning.
SoftDTW solved this issue by replacing the $hard-\min$ operation by a $soft-\min$ operation.
The $soft-\min$ operation is used as follows:
\begin{equation}\label{equ:soft-min}
    soft-\min~^{\gamma}(a_1,a_2,\ldots,a_N) = -\gamma.\log\sum_{i=1}^N e^{-a_i/\gamma}
\end{equation}
\noindent where $\gamma$ is the smoothing factor, and as it tends to the value $0^{+}$, then the $soft-\min$ becomes the $hard-\min$ hence SoftDTW becomes the original DTW.

In~\cite{shape-dtw-distance}, a variation of DTW was introduced, which aligns transformations of
sub-sequences within time series instead of aligning all time series simultaneously.
This approach aims to maintain the consideration of neighborhood structure when
aligning timestamps across different time series. To define ShapeDTW mathematically,
let $\mathcal{F}$ be a descriptor function, $\textbf{x}_1$ and $\textbf{x}_2$ be two MTS of length $L$ and dimension $M$.
The process begins by extracting sub-sequences over all channels of length
$r$ (referred to as reach) and transform them using a descriptor $\mathcal{F}: \mathds{R}^r~\to~\mathds{R}^d$.
This results in two new MTS $\mathcal{D}_1$ and $\mathcal{D}_2$ of length $L$ and dimension $M.d$, associated to $\textbf{x}_1$
and $\textbf{x}_2$ respectively.

The DTW alignment path is then computed on the transformed version of the series $\mathcal{D}_1$ and $\mathcal{D}_2$,
followed by the optimal path being transferred onto the original series space to calculate the measure between
the original time stamps instead of the sub-sequences.
In this manner, the DTW algorithm will calculate the distance between time stamps following their neighborhood alignments.
The ShapeDTW measure can be formulated as the following optimization problem:
\begin{equation}\label{equ:shape-dtw}
    ShapeDTW_q(\textbf{x}_1, \textbf{x}_2) = (\sum_{(t_1,t_2)\in\pi^{*}}\sum_{m=1}^M(x^m_{1,t_1}-x^m_{2,t_2})^q)^{1/q}
\end{equation}
\noindent where $\pi^{*}$ is the optimal path obtained by the DTW alignment path between transformed series as follows:
\begin{equation}\label{equ:shape-dtw-path}
    \pi^{*} = arg\min_{\pi\in\mathcal{A}(\mathcal{D}_1,\mathcal{D}_2)}(\sum_{(\tilde{t}_1,\tilde{t}_2)\in\pi}\sum_{m=1}^M(\mathcal{D}^m_{1,\tilde{t}_1}-\mathcal{D}^m_{2,\tilde{t}_2})^q)^{1/q}
\end{equation}
The distance is usually set to the squared error so $q=2$.

\paragraph{Elastic Ensemble}

Given the high number of similarity measures for time series data,~\cite{elastic-ensemble} proposed to do a weighted Elastic Ensemble (EE) of $11$ NN classifiers each using a different similarity measure.
Below, we define what an ensemble of classifiers is, a concept that will be used throughout the rest of this work.
\mydefinition Ensembling different classifiers is motivated by the idea that combining multiple opinions 
often leads to a more robust decision. Since each classifier generates a probability distribution for 
each series across the possible classes, the ensemble method averages these probability distributions 
from all classifiers.

\paragraph{Proximity Forest and Proximity Forest2.0}

Proximity Forest (PF)~\cite{proximity-forest} is a Random Forest (RF)~\cite{random-forest} classifier adaptation for the time series classification task.
PF utilizes the same $11$ similarity measures that EE~\cite{elastic-ensemble} uses, however it randomly sets at each branch one of the similarity measures to be used for the fed time series.
Until 2023, PF was the state-of-the-art distance based classifier for the task of TSC on the UCR archive~\cite{ucr-archive} following the recent TSC review~\cite{bakeoff-tsc-2}.
However, recently, the same group that developed PF upgraded the algorithm and developed
PF2.0~\cite{proximity-forest-2} and it was significantly better than PF.
PF2.0 differs from the original PF by three main features: (1) being efficiently better than much 
faster, (2) the addition of a new similarity measure Amerced Dynamic Time Warping (ADTW)~\cite{adtw-distance}
and (3) tuning the parameters of the cost function.

\subsubsection{Feature Based Methods}\label{sec:tsc-feature}

Using traditional machine learning classifiers such as RF~\cite{random-forest} or RIDGE classifier~\cite{hoerl1970ridge}
is insufficient on raw time series data as these classifiers are constructed to use tabular input.
In order to overcome this issue, some feature based classifiers were proposed for TSC which consist 
of a pipeline of feature extraction methods followed by a simple classifier designed for tabular data.
In what follows, we present briefly some state-of-the-art feature based methods of TSC.

\paragraph{The Canonical Time Series Characteristics (Catch22)}

Building on the original work of~\cite{hctsa}, which proposed the
\emph{Highly Comparative Time-Series Analysis (hctsa)} tool to extract
around $7700$ features from each time series, the authors in~\cite{catch22} did an extensive amount of 
experiment in order to identify the most effective $22$ \emph{hctsa} features.
This new set of features proposed in~\cite{catch22} is called Catch22, which is then followed by an 
RF classifier~\cite{random-forest}.

\paragraph{Time Series Feature Extraction based on Scalable Hypothesis Tests (TSFresh) and The FreshPRINCE}

TSFresh~\cite{tsfresh} is a set of around $800$ features that are extracted from each time series.
This set of features are not all utilized for the classification, instead, the authors in~\cite{tsfresh} proposed
the usage of a feature selection method called FRESH~\cite{fresh}.
The selected features are then fed into an RF classifier or an AdaBoost classifier~\cite{adaboost}.
The TSFresh features were then used recently by~\cite{fresh-prince} where the authors removed the feature selection method
and utilize a Rotation Forest (RotF) classifier~\cite{rotation-forest} on top of the TSFresh features, to produce the FreshPRINCE feature-based classifier.

Until now, the FreshPRINCE classifier is the state-of-the-art feature-based method for TSC on the UCR archive following the recent TSC review~\cite{bakeoff-tsc-2}.

\subsubsection{Convolutional Based Methods}\label{sec:tsc-convolution}

Convolution based approaches have shown to be very effective on image classification since the birth of 
Convolutional Neural Networks (CNNs)~\cite{lecun2015deep}.
In the case of images, convolution filters are two-dimensional operation where the filter slides all over the image
in order to extract some meaningful features.
However, in the case of time series, the convolution operation is one dimensional and slides all over the temporal axis
of the time series in order to extract temporal features and local dependencies.

\mydefinition A one-dimensional convolution operation over a univariate time series $\textbf{x}$ of
length $L$ with a kernel $\textbf{w}=\{w_1,w_2,\ldots,w_K\}$ of length $K$ is defined as follows:
\begin{equation}\label{equ:convolution1d-uni}
    o_t = \sum_{k=1}^K x_{t+k-1}.w_k
\end{equation}
\noindent with $t~\in~[1,L-K+1]$ and $\textbf{o}=\textbf{x}*\textbf{w}=\{o_1,o_2,\ldots,o_{L-K+1}\}$ is the output
series of the one-dimensional convolution and $*$ is the convolution operator.
Whenever a value in $\textbf{o}$, representing a segment in $\textbf{x}$ of length $K$, is positive, it is referred to 
as the convolution kernel being activated at that segment in $\textbf{x}$, thus a pattern is detected.
\emph{It is important to note that the above definition of a 1d convolution operation uses a stride of $1$, which represents the amount 
of time stamps the convolutional kernel shifts when sliding on the temporal axis.
By default, in the rest of this work, all convolutional operations use a stride of $1$.}

In the above approach, the convolution is being applied to a consecutive set of time stamps.
However, it can be interesting to extract features of time stamps with wider temporal distance between them.
This can be done by simply increasing the length of the convolution kernel, however it would 
increase the number of parameters used.
A more constructive approach is to use dilated convolution to increase the view of 
the kernel over the time series sample.

\mydefinition Dilated one-dimensional convolution between a series $\textbf{x}$ and a kernel $\textbf{w}$ of
lengths $L$ and $K$ respectively with a dilation rate $d > 1$ is defined as follows:
\begin{equation}\label{equ:convolution1d-dilation}
    o_t = \sum_{k=1}^K x_{t+(k-1).d}.w_k
\end{equation}
\noindent where $t~\in~[1, L-(K-1).d]$ and $\textbf{o}$ the output
series of the one-dimensional dilated convolution.
If $d = 1$ then this comes down to applying the convolution as in Eq.~\ref{equ:convolution1d-uni}.
The dilation rate allows the convolution operation to skip some elements in the input series to detect longer patterns.

\paragraph{RandOm Convolutional KErnel Transform (ROCKET)}
\cite{rocket}
proposed ROCKET, a convolution based model that randomly generates a large set of kernels
following a standard Gaussian distribution $\mathcal{N}(0,1)$, with random dilation rates and random 
biases sampled from a uniform distribution $\mathcal{U}(-1,1)$.
This set of filters is then applied on each of training time series samples followed by two aggregation functions.
The first aggregation is choosing the maximum value of the convolution output and the second is taking the Proportion
of Positive Values (PPVs).
The PPV is obtained as detailed in Eq.~\ref{equ:ppv}:
\begin{equation}\label{equ:ppv}
    PPV(\textbf{o} = \textbf{x}*\textbf{w}) = \dfrac{1}{L-K+1}\sum_{t=1}^{L-K+1} \mathds{1}[\textbf{o}_t > 0]
\end{equation}
\noindent $\mathds{1}[condition]$ is the indicator function defined as:
\begin{equation}
    \mathds{1}[condition]=
    \begin{cases}
        1 & if\text{ }condition\text{ }is\text{ }True\\
        0 & if\text{ }condition\text{ }is\text{ }False
    \end{cases}
\end{equation}

Assuming that ROCKET uses $\mathcal{K}$ convolution filters, the output space dimension is 
$2\mathcal{K}$ per time series sample. This latent space of the training set is then used 
to optimize the parameters of a RIDGE classifier~\cite{hoerl1970ridge}. A unique feature of 
ROCKET, compared to other classifiers, is that it is entirely independent of the training 
dataset during its feature extraction phase,
with its parameters being randomly generated. ROCKET's computational runtime is 
significantly smaller than that of other state-of-the-art classifiers, and it has consistently 
been one of the top-performing models in the literature across widely used community 
benchmarks~\cite{ucr-archive,uea-archive}.

\paragraph{MiniROCKET \& MultiROCKET}

The same authors of ROCKET~\cite{rocket} proposed in 2021 a new version called MiniROCKET~\cite{mini-rocket} that is almost
a deterministic version of the original model in order to reduce its randomness and add some dependency with the input data.
The key differences between ROCKET and MiniROCKET can be summarized in the following:
\begin{enumerate}
    \item MiniROCKET fixed the length of the filters to $9$
    \item MiniROCKET randomly generates the values of the filters from a discrete set of values $\{-1,2\}$ instead of using
a Gaussian distribution
    \item MiniROCKET drops the maximum aggregation and utilizes only the PPV
    \item MiniROCKET samples the bias values from the quantiles of the convolution output, making it dependent on the training data
    \item MiniROCKET fixed the number of possible dilation rates from $1$ to $\log_2(\dfrac{L-1}{K-1})$ where $L$ is the time series
    length and $K$ the kernel length.
\end{enumerate}
MiniROCKET highlights that by reducing the degree of freedom of ROCKET, than both the accuracy and efficiency can increase.

The same group proposed MultiROCKET in the following years in~\cite{multi-rocket}, that utilizes the MiniROCKET setup however
it applies the transformation over the original time series and its first order derivative.
MultiROCKET does not rely only on the PPV features however it produces three new features:
\begin{enumerate}
    \item Mean of Positive Values (MPV) that averages the positive values of the convolution output
    \item Mean of Indices of Positive Values (MIPV) that averages the indices of the positive values of the convolution output
    \item Longest Stretch of Positive Values (LSPV) that finds the length of the longest subsequences containing positive consecutive values in the convolution otuput
\end{enumerate}
MultiROCKET adds some computation complexity to MiniROCKEt, however it achieved state-of-the-art results over the UCR archive
for the TSC task in 2022.

\paragraph{HYbrid Dictionary-Rocket Architecture}

In 2023, a new adaptation of ROCKET based framework was proposed in~\cite{hydra} called HYDRA.
Unlike ROCKET, HYDRA does not rely on the actual output activation of the filter, instead it leverages over how many times
a convolution kernel is activated the most between a set of kernels.
In other words, HYDRA randomly defines a set of kernels, called group and applies the convolution such as in ROCKET
followed by assigning each kernel in this group the number of time stamps it is activated the most in the group.
HYDRA employs $\mathcal{G}$ groups with $\mathcal{K}$ kernels in each group, resulting in an output feature space of dimension
$\mathcal{G}$x$\mathcal{K}$ containing integer values.
This feature space is subsequently used to train a RIDGE classifier~\cite{hoerl1970ridge}.
\cite{hydra} concluded that by combining the feature space of HYDRA with the feature space of MultiROCKEt,
resulting in HydraMR (HYDRA-MultiROCKET), achieves state-of-the-art performance for the task of TSC on the UCR archive~\cite{ucr-archive}.

HydraMR is currently one of the state-of-the-art models for TSC, not only in convolution based methods, 
but overall as well.

\subsubsection{Shapelet Based Methods}\label{sec:tsc-shapelet}

Shapelet-based time series classification methods focus on identifying and using small, discriminative
subsequences, known as shapelets, to distinguish between different classes. These methods extract shapelets
that capture local patterns highly indicative of the target class, providing interpretable and precise models.
This approach is particularly useful in applications like medical diagnosis, where specific patterns in data can
be crucial for accurate classification. Shapelet-based classifiers are valued for their robustness and interpretability,
making them a powerful tool in time series analysis.

Shapelets were first introduced in~\cite{image-countour-shapelets-paper,shapelets} as discriminative subsequences used within decision tree
classifiers for time series classification. Since then, the research community in TSC has extensively developed
and expanded this concept, leading to a variety of advanced algorithms and applications that leverage shapelets
for improved accuracy, interpretability, and computational efficiency.

\paragraph{Shapelet Transform Classifier (STC)}

The STC~\cite{stc} is a two steps classifier. First the model searches for shapelets in the set 
of training samples and transforms the series
to a vector of distances between the shapelet and a set of other shapelets from the series itself.
Second, a decision tree classifier is trained on top of the transformed space of the series.

\paragraph{Random Dilated Shapelet Transform (RDST)}

The RDST model~\cite{rdst}, motivated by ROCKET~\cite{rocket}, leverages from the randomness techniques to select the shapelets 
from the training samples.
Instead of learning the shapelets, RDST randomly selects a high number of shapelets from the training data.
Similar to convolution based methods, RDST employs the dilation technique to enrich the shapelet transform.
The transformed space of RDST is then used to train a RIDGE classifier~\cite{hoerl1970ridge}.

Currently, RDST is still the state-of-the-art shapelet based methods for TSC evaluated on the UCR archive.

\subsubsection{Dictionary Based Methods}\label{sec:tsc-dictionary}

This approach of solving TSC is based on finding discriminative patterns in the time series and counting the number of times 
it was repeated, followed by using this information to train a classifier.
The patterns detected are not from the raw input, instead the time series is transformed first into a discrete space using a
symbolic transformation.
A very famous symbolic transformation proposed in~\cite{sax} called Symbolic Aggregate approXimation (SAX) defines a set of
discrete symbols that represent a segment of the time series.
SAX employs this symbolic transformation as follows:
\begin{itemize}
    \item \textbf{First}, each time series $\textbf{x}$, supposing being univariate, of length $L$ is z-normalized to have a zero mean and unit standard deviation.
    \item \textbf{Second}, the time series is divided into non-overlapping segments of length $l$ each:
    $\{\textbf{x}[(t-1).l:t.l]\}_{t=1}^{\lfloor L/l \rfloor}$
    \item \textbf{Third}, each segment is replaced by its mean value following the Piecewise Aggregate Approximation (PAA)~\cite{paa}
    dimensionality reduction technique to obtain:
    $\{p_t=mean(\textbf{x}[(t-1).l:t.l])\}_{t=1}^{\lfloor L/l \rfloor}$
    \item \textbf{Fourth}, the dictionary of symbols $\mathcal{D}_{ict}=\{s_1,s_2,\ldots,s_{\alpha}\}$ is defined for specific number of alphabet $\alpha$ (a SAX hyper-parameter)
    using the percent point function of the standard Gaussian distribution to obtain $\alpha-1$ breakpoints $\{b_j\}_{j=1}^{\alpha-1}$
    \item \textbf{Finally}, for each segment $\textbf{x}[(t-1).l:t.l]$ for $i~\in~[1,\lfloor L/l \rfloor]$
    replaced by its mean value, a one-to-one mapping function is used to choose 
    the replacement symbol from the dictionary as follows:
    \begin{equation}\label{equ:sax}
        SAX(p_i) = 
        \begin{cases}
            s_1 & if~-\infty < p_i \leq b_1\\
            s_2 & if~b_1 < p_i \leq b_2\\
            \ldots & \\
            s_{\alpha} & if~b_{\alpha-1} < p_i < +\infty
        \end{cases}
    \end{equation}
\end{itemize}

The above steps can be applied in the same way on MTS data, by going through each dimension independently.
The core idea of SAX is simply representing each segment of the series by a discrete 
symbol to form a word (sequence of symbols) that is chosen following
the Gaussian distribution, this is argued by the authors in~\cite{sax} by saying
``\emph{$\ldots$ normalized time series have a Gaussian distribution}``.
This symbolic representation was then used for classification in~\cite{sax-vsm} with the first dictionary based classifier 
for time series, called SAX Vector Space Model (SAX-VSM).
SAX-VSM begins by generating the Symbolic Aggregate approXimation (SAX) representation for all time series 
within each class, while also preserving the frequency of the symbol sequences. When presented with a new, 
unlabeled time series, it undergoes SAX transformation to obtain its symbolic sequence. Then, by comparing 
the frequencies derived from the precomputed set of symbol sequences, the unlabeled series is assigned to 
the class that best matches its sequence frequencies.

In~\cite{sfa}, the authors proposed a novel version, six years after the breakthrough of SAX, called 
the Symbolic Fourier Approximation (SFA).
First, SFA decomposes the series into segments and then z-normalize the sub-sequences instead of normalizing them prior to
the decomposition.
Second, SFA utilizes the Discrete Fourier Transform as a dimensionality reduction technique instead of PAA~\cite{paa}.
Third, SFA uses a binning technique proposed in~\cite{sfa} called Multiple Coefficient Binning (MCB) that is based on the 
distributions of real and imaginary values of the Fourier Transform.
Finally, Those distributions go through the binning mechanism to generate the symbols.

\paragraph{Bag-of-SFA-Symbols (BOSS)}

In~\cite{boss}, the authors proposed a novel approach for dictionary based TSC called BOSS that utilizes on SFA.
BOSS applies the SFA transformation on overlapping windows of the time series instead of considering the whole time series at once.
This results in a sequence of words for each series instead of producing a sequence of symbol (one word only).
A BOSS classifier utilizes a non-symmetric distance in the setup of a NN classifier, and multiple BOSS classifiers are finally ensembled
to form the final BOSS model.
Until 2015, BOSS was the state-of-the-art dictionary based method for TSC evaluated on the UCR archive.

\paragraph{Word Extraction for Time Series Classification (WEASEL1.0 and WEASEL2.0)}

WEASEL1.0 is a novel dictionary based model proposed in~\cite{weasel} for TSC, that in contrary to BOSS,
its goal is to identify meaningful words in the output transformation of SFA.
This is done by applying the transformation using SFA on a large set of possible parameters, followed by a Chi-squared test
to identify the words that have the highest power and discard the words that have a power lower than a specified threshold.
The output space is then used to train a RIDGE classifier~\cite{hoerl1970ridge}.

Although WEASEL1.0 have seen to outperform BOSS on the UCR archive, it still however suffer from the dimensionality curse
and runtime curse because of the large grid search space.
The same authors~\cite{weasel2} proposed a new version denoted by WEASEL2.0 that utilizes the randomness technique of ROCKET
and randomly generate a set of parameters for the SFA transformation thus controlling the searching space.
WEASEL2.0 sets a random dilation rate as well for the windowing phase of the workflow, motivated from the impact of dilation 
on the ROCKET transformation.
WEASEL2.0 became the state-of-the-art dictionary based method for TSC in terms of both accuracy and efficiency.

\subsubsection{Interval Based Methods}\label{sec:tsc-interval}

Interval based methods, first proposed in~\cite{tsf} as an RF~\cite{random-forest} based classifier, is a technique of ensembling 
different classifiers trained on different transformations of extracted intervals from the time series samples.
Most approaches randomly generate the intervals' bounds that are used throughout all the samples in the dataset.
The motivation of using such technique instead of feature based methods where the transformation is done over all the series at 
the same time, is to avoid noisy features that will lead in miss-classification and confusing the classifier.

\paragraph{Time Series Forest (TSF)}

The TSF~\cite{tsf} model employs for each decision tree $\sqrt{L}$ intervals, where $L$ is the time series length, of randomly selected bounds.
For each of the selected intervals, TSF extracts the mean, variance and slope and concatenate them into one feature vector that is then
used to build the decision tree.
All the decision trees are then ensembled through a voting mechanism to form the TSF classifier.

\paragraph{Canonical Interval Forest and Diverse Representation Canonical Interval Forest (CIF and DrCIF)}

Similar to TSF~\cite{tsf}, CIF~\cite{cif} is an ensemble of decision tree classifiers, however it utilizes the Catch22 features
alongside the mean, variance and slope features of TSF.
The concatenated vector is then used to build the decision tree.
CIF leverages over TSF by being suitable for multivariate time series, as it the number of intervals for each tree is
$\sqrt{L}.\sqrt{M}$, where $L$ and $M$ are the length and number of channels of the time series samples respectively.
In order to keep the selected intervals in the one-dimensional space, CIF randomly assigns a channel for each of the selected 
intervals.

In~\cite{hive-cote2.0}, the same authors proposed DrCIF by incorporating two new extracted features alongside the ones of CIF:
(1) the periodograms to identify and quantify the frequency components present within the time series and (2) 
the first order derivative.

\paragraph{QUANT}

In 2023, the QUANT model, as referenced in~\cite{quant}, discarded all previously employed
features identified through interval based methods.
Instead, QUANT relies on quantiles that represent the empirical distribution of the intervals.
However, it extracts quantiles over four different representations including the time series itself, the first and second order
derivative, the Fourier transform.
The choice of the intervals is not random in QUANT, instead it is fixed and dyadic (defined based on the powers of $2$).
For each interval of length $l$, QUANT defines sub-intervals of length $l/4$ and extracts two features called quantiles from 
each sub-interval.
These two features are the median of the sub-interval and the median of zero centered sub-interval (mean of the sub-interval is 
extracted before finding the median).
The output features of all intervals are then concatenated representing a new quantized version of the input time series.
QUANT utilizes extremely randomized trees~\cite{extremely-randomized-trees} for the classification task, where the transformed quantized space is used to train the 
tree based classifier.

In the recent TSC review~\cite{bakeoff-tsc-2}, it has been shown that QUANT is the current 
state-of-the-art interval based method
for TSC evaluated on the UCR archive~\cite{ucr-archive} in both accuracy and efficiency 
as it is significantly faster than other interval based methods.

\subsubsection{Hybrid Based Methods}\label{sec:tsc-hybrid}

Given that time series data does not have a unified approach to address its classification task, it is most of the time a
difficult challenge to choose from the pool of methods.
For this reason, hybrid models have been proposed throughout the literature in a way to combine different methods, e.g.
distance based and interval based methods.

\paragraph{Time Series Combination of Heterogeneous and Integrated Embedding Forest (TS-CHIEF)}

TS-CHIEF~\cite{ts-chief-paper} is a tree-based ensemble method where the nodes within each 
tree perform splits using three distinct feature criteria: distance-based, dictionary-based, 
and spectral interval-based. The parameters are randomly initialized to ensure diversity 
within the ensemble. The distance-based splits are derived from the
EE\cite{elastic-ensemble}, the dictionary-based splits are inspired by BOSS\cite{boss}, 
and the interval-based splits are based on the Random Interval Spectral Ensemble 
(RISE)~\cite{rise}.

\paragraph{HIVE-COTE1.0 and HIVE-COTE2.0}

The HIVE-COTE (HC) method has been developed throughout the years, starting with the Collective Of
Transformation-based Ensemble (COTE)~\cite{cote} which is an ensemble of $35$ time series classifiers of different approaches.
This model was developed to the HIerarchical VotE Collective Of Transformation Ensemble (HIVE-COTE)~\cite{hive-cote} that only utilizes five
classifiers and are ensembled through the  Cross-validation Accuracy Weighted Probabilistic Ensemble (CAWPE)~\cite{cawpe}.
HIVE-COTE utilizes a distance based, dictionary based, shapelet based, interval based and spectral based classifiers.
However, the distance based model used in HIVE-COTE is computationally expensive, for this reason it was dropped in HIVE-COTE1.0~\cite{hive-cote1}
and a more performing dictionary based classifiers is used.
The most recent HC based model is HIVE-COTE2.0~\cite{hive-cote2.0} which changed the set of classifiers to more recent ones that 
are much more performing.
A unique feature of HC2 over HC1.0, HC and COTE, is that the classifiers used in its hybrid ensemble are suitable for multivariate 
datasets.
HC2 is currently one of the state-of-the-art models for TSC evaluated over the UCR archive, not only in hybrid based methods, 
but overall as well.

\subsubsection{Deep Learning Methods}\label{sec:tsc-deep}

Although the previously presented methods for TSC are performing well on the available benchmarks~\cite{ucr-archive,uea-archive},
most of them lack the capability of parallelization of their calculation over GPUs, which can decrease their efficiency.
Another critical limitation of most of these classifiers is their lack of explainability, which is increasingly 
important for understanding model decisions, ensuring transparency, and gaining trust in applications where 
decision-making is crucial.
For these reasons, deep learning methods can be a suitable solution, however we do not claim it should 
be the only solution for TSC 
as there are still some models that can achieve better performance compared to deep learning models 
but are less suitable in terms of scalability.

Deep learning~\cite{lecun2015deep} methods leverage over all the previous techniques with the ability of
parallelization over multiple GPUs making them much faster during training and
inference.
Moreover, deep learning methods can conduct two steps including the feature extraction
and the classification task at the same time instead of manually constructing the features phase
and only training the classifier.
Deep learning methods consist on many neural network architectures, such as Convolutional
Neural Networks (CNNs), Recurrent Neural Networks (RNNs) and Transformers.
However, an extensive review of deep learning methods for TSC has been conducted in~\cite{dl4tsc}
highlighted that CNNs outperform other architectures.

A deep learning model for TSC consists on applying $\Lambda$ parametrized layers of different
characteristics.
Each of the layers $\lambda_i$ where $i~\in~[1,\Lambda]$ represents a function $f_{\lambda_i}$ parametrized by
a set of parameters $\theta_{\lambda_i}$.
Each layer $\lambda_i$ takes as input the output of the previous layer $\lambda_{i-1}$ and applied a non-linear transformation
over it that is controlled by $\theta_i$.
Given an input time series $\textbf{x}$, feeding to a neural network of $\Lambda$ layers comes down to the following pipeline:
\begin{equation}\label{equ:dl-pipeline}
    f_{\Lambda}(\theta_{\Lambda},\textbf{x}) = f_{\Lambda-1}(\theta_{\Lambda-1},f_{\Lambda-2}(\theta_{\Lambda-2},\ldots,f_{1}(\theta_{1},\textbf{x})))
\end{equation}
The above pipeline is referred to in the community as the feed-forward propagation.

Since the task at hand is classification, the last layer of the deep learning model
outputs a probability distribution for each sample belonging to each of the possible
classes.
The parameters of all the layers are then optimized using the back-propagation algorithm~\cite{rumelhart1986learning}.
In what follows, we present the different types of layers than are used for time series in the literature.

\paragraph{Types of Layers}

In this section, we go through some layer types used in the literature's architectures.
These layers are based on non-linear transformations that consists on either extracting information, detecting some patterns
or combining some features in the time series samples.

\paragraph*{Fully Connected (FC) Layers}
The FC layers are simply a linear transformation followed by applying a non-linear activation
such as ReLU, sigmoid etc.
This linear transformation is computed using matrix multiplication.
For instance, if the input dimension is $n$ and the output dimension is $m$, then the FC
layer consists on weight matrix $W$ of shape $(m,n)$ and the output of the FC layer is computed as follows:
\begin{equation}\label{equ:fc-layer}
    \textbf{o} = \sigma(W\odot \textbf{x} + \textbf{b})
\end{equation}
\noindent where $\textbf{x}$ is the input, $\textbf{o}$ is the output, $W$ is the transformation matrix,
$\textbf{b}$ is the bias vector of dimension $m$, $\sigma(.)$ is a non-linear activation function
and $\odot$ is the matrix multiplication operation.

This type of layer is almost always used as the last layer in a deep learning model for a classification
task while setting the activation function to the softmax function and the output dimension to $C$, the number of possible classes.
This function ensures that the output vector is a probability distribution and each element in the output vector $\textbf{o}$ is computed as follows:
\begin{equation}\label{equ:softmax}
    o_c = \dfrac{e^{W[c,:]\odot\textbf{x} + b_c}}{\sum_{\tilde{c}=1}^{C}e^{W[\tilde{c},:]\odot\textbf{x} + b_{\tilde{c}}}}
\end{equation}
\noindent where $o_c$ is the probability of $\textbf{x}$ belonging to class $c~\in~[1,C]$

In order to find the optimal weights of Eq.~\ref{equ:fc-layer} and~\ref{equ:softmax},
we can use an optimization algorithm to minimize the error in the model's predictions.
This error is measured through a loss function, that should be differentiable given that the
optimization algorithm is gradient based.
The common loss function to be used for the classification tasks is the categorical cross 
entropy, that measures the difference between two probability distributions, defined as follows
on the $i_{th}$ example of the dataset:
\begin{equation}\label{equ:cross-entropy}
    \mathcal{L}_i(\textbf{y}_i,\hat{\textbf{y}}_i) = -\sum_{c=1}^{C}y_{i,c}.\log_{2}(\hat{y}_{i,c})
\end{equation}
\noindent where $C$ is the total number of classes in the dataset, $\textbf{y}_i$ is the ground truth
label of the $i_{th}$ series, denoted as a one hot encoding, e.g. if the ground truth label is $C_2$ out of a set
$\{C_1,C_2,C_3\}$ then $\textbf{y}_i=[0,1,0]$ representing a discrete deterministic probability distribution.
$\hat{\textbf{y}}_i$ is a vector of length $C$ representing a discrete probability distribution where 
each element $\hat{y}_{i,c}$~$c~\in~[1,C]$ is the probability of the $i_{th}$ sample belonging to class $c$.

The total loss over a batch of $N$ samples is the average loss over all the samples in the batch:
\begin{equation}\label{equ:batch-loss}
    \mathcal{L} = \dfrac{1}{N} \sum_{i=1}^{N}\mathcal{L}_i(\textbf{y}_i,\hat{\textbf{y}}_i)
\end{equation}

In order to update the weights of Eq.~\ref{equ:fc-layer} and~\ref{equ:softmax}, a gradient based 
optimizer can be used such as Stochastic Gradient Descent (SGD) as follows:
\begin{equation}\label{equ:sgd}
    W = W - \alpha.\dfrac{\partial \mathcal{L}}{\partial W}
\end{equation}
\noindent where $\alpha$ is the learning rate hyper-parameter controlling the step size of the 
optimization algorithm.

In the current literature, deep learning models consist of a very high number of layers on top 
of each other, in this case, the partial derivative of Eq.~\ref{equ:sgd} cannot be calculated.
Instead, for the last 20 years, neural networks utilize the derivative chain rule, the core idea 
of the back-propagation algorithm~\cite{back-prop}.

\paragraph*{Convolution Layers}

The convolution operation, as explained in Eq.~\ref{equ:convolution1d-uni}, is applied the same way 
in a convolution layer, where the optimization algorithm learns the best weights of the convolution 
kernel.
A convolution layer applies $\mathcal{K}$ filters $\{\textbf{w}\}_{j=1}^{\mathcal{K}}$ of the same
length $K$ and same dilation rate over the input 
time series.
If the input time series is univariate $\textbf{x}$ of length $L$, the output of the convolution layer is 
a multivariate time series computed as follows:
\begin{equation}\label{equ:conv-layer-uni}
    \textbf{o} = concat(\{\textbf{x} * \textbf{w}_j\}_{j=1}^{\mathcal{K}})
\end{equation}
\noindent where $concat$ is the concatenation operation and $\textbf{o}$ is a multivariate time series 
of $\mathcal{K}$ dimensions with length $L-K+1$ each.

In the case $\textbf{x}$ is a multivariate time series of $M$ channels of length $L$ each, and the
target output dimension is $\mathcal{K}$ (the chosen number of filters), then in reality, the number 
of filters to learn is $M.\mathcal{K}$.
This is done by simply learn $M$ filters, one for each of the input dimensions and summing the output.
This is repeated $\mathcal{K}$ times and the output sums are concatenated to produce the output MTS $\textbf{o}$.
The mathematical formulation of the above operation is defined as follows:
\begin{equation}\label{equ:conv-layer-multi}
    \textbf{o} = concat(\{\sum_{m=1}^{M}\textbf{x}^m*\textbf{w}_{m,j}\}_{j=1}^{\mathcal{K}})
\end{equation}
\noindent where the above summation in Eq.~\ref{equ:conv-layer-multi} is over the temporal axis of all
the series inside the sum, producing after each summation a univariate time series.

\begin{figure}[t]
\centering
\caption{Standard Convolution applied on a 
\protect\mycolorbox{0,0,255,0.5}{multivariate input time series }
of dimensions $3$, convoluted with two times with three different 
\protect\mycolorbox{0,125,0,0.5}{convolutional filters}, producing 
a \protect\mycolorbox{255,161,0,0.5}{convolutional output} per filter,
that are then summed together to produce two \protect\mycolorbox{255,0,0,0.5}{final outputs}.
The convolution operation starts with an element wise
\protect\mycolorbox{138,74,171,0.7}{multiplication}
followed by a \protect\mycolorbox{202,120,115,0.7}{summation operation}.}
\label{fig:std-conv}
\includegraphics[width=\textwidth]{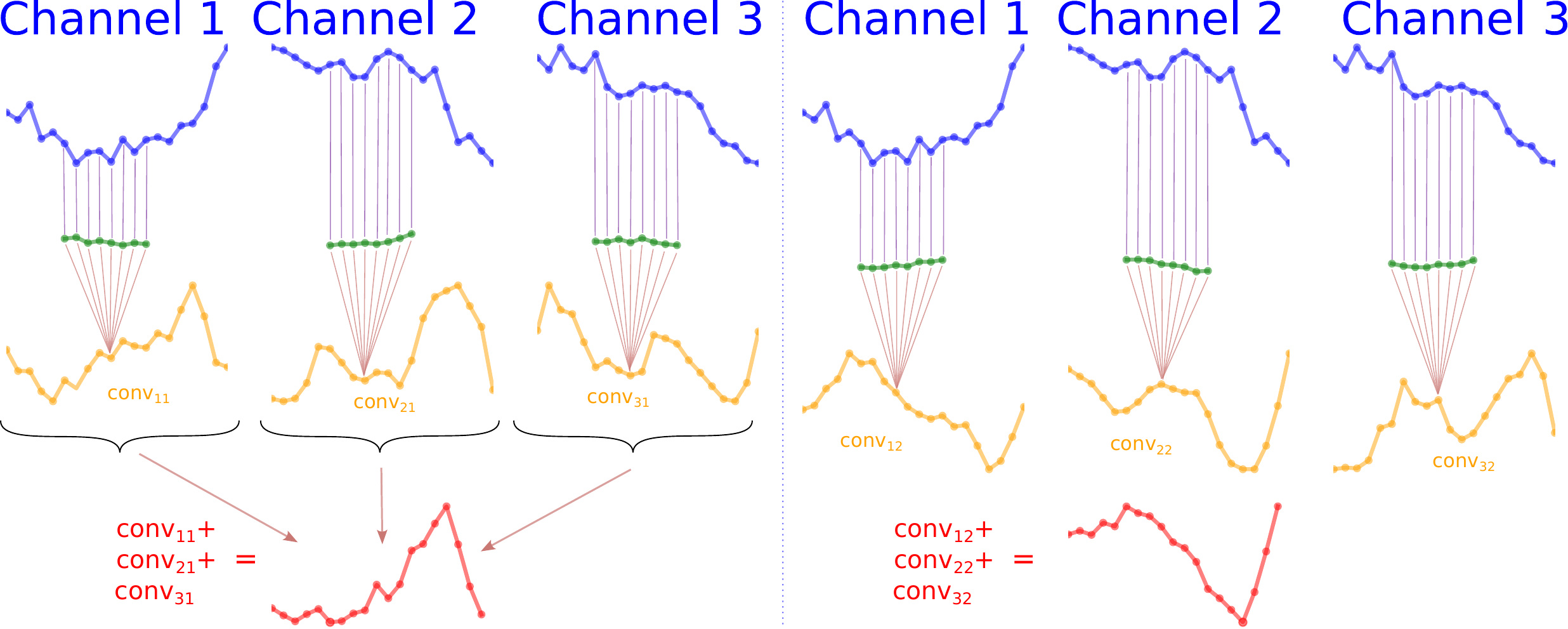}
\end{figure}

This type of convolution layer is referred to in the rest of this work as Standard Convolution (SC) layer.
A visualization of the SC layer with a chosen number of filters set to $2$ with a kernel size of $8$
is presented in Figure~\ref{fig:std-conv} applied on an input MTS of dimension $3$.
It can be seen from this figure that the total number of filters to learn is $6$ instead of $2$ (the chosen output dimension).

\paragraph*{DepthWise Separable Convolution (DWSC)}

First used for image classification in MobileNets~\cite{mobilenets}, this type of convolution layer has a unique feature 
of having a very low number of trainable parameters.
DWSCs are in fact a pipeline made of two different convolution layers: (1) DepthWise Convolution (DWC) followed by (2) PointWise
Convolution (PWC).
DWSC are more common to be used on MTS input data.
The DWC layer (first phase of DWSCs) consists on learning $M$ filters where $M$ 
is the dimension of the input MTS, hence the reason to why DWSCs are
commonly used on multivariate input, or else we learn only one filter.
For instance, if the input raw MTS $\textbf{x}$
has $M$ dimensions of length $L$,
applying a DWC layer with kernel size $K$ is defined as follows:
\begin{equation}\label{equ:depthwise-conv}
    \textbf{o} = concat(\{\textbf{x}^m*\textbf{w}_m\}_{m=1}^{M})
\end{equation}
\noindent where $\textbf{o}$ is the output of the DWC layer, also with $M$ dimensions of length $L-K+1$

\begin{figure}[t]
    \centering
    \caption{DepthWise Separable convolution
    \protect\mycolorbox{0,0,255,0.5}{multivariate input time series }
of dimensions $3$, convoluted one time with three different 
\protect\mycolorbox{0,125,0,0.5}{convolutional filters}, producing 
a \protect\mycolorbox{255,161,0,0.5}{convolutional output} per filter,
that go through a weighted summed  to produce two \protect\mycolorbox{255,0,0,0.5}{final outputs}.
The convolution operation starts with an element wise
\protect\mycolorbox{138,74,171,0.7}{multiplication}
followed by a \protect\mycolorbox{202,120,115,0.7}{summation operation}.}
    \label{fig:sep-conv}
    \includegraphics[width=\textwidth]{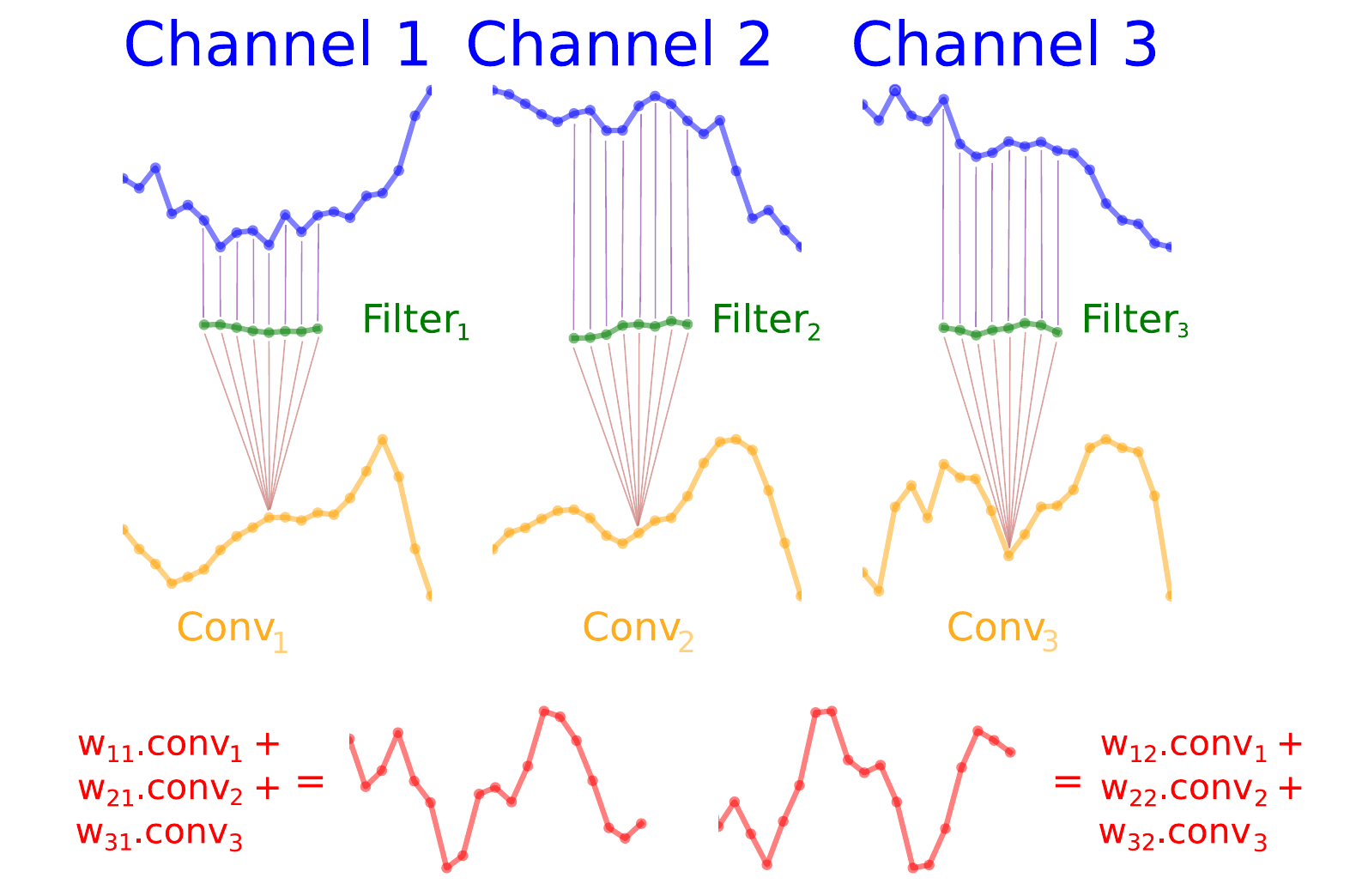}
\end{figure}

The PWC layer, also referred to as bottleneck layer, consists on a change of dimensionality through a standard convolution layer 
with a kernel size of $1$.
Applying a PWC layer with target dimension $\mathcal{K}$ on an input MTS $\textbf{x}$ of $M$ dimensions of length $L$
is defined as follows:
\begin{equation}\label{equ:pointwise-conv}
    \textbf{o} = concat(\{\sum_{m=1}^{M}\textbf{x}^m.w_{m,j}\}_{j=1}^{\mathcal{K}})
\end{equation}
\noindent where $W=\{\{w_{m,j}\}_{m=1}^{M}\}_{j=1}^{\mathcal{K}}$ is a two-dimensional matrix of real values and $\textbf{o}$
is the output of the PWC layer with the same length as the input and $\mathcal{K}$ dimensions.

Finally, the pipeline of DWSC layer of kernel size $K$ and target dimension $\mathcal{K}$
applied on an input MTS $\textbf{x}$ of $M$ dimensions of length $L$ can be defined as follows:
\begin{equation}\label{equ:depthwise-sep-conv}
    \textbf{o} = concat(\{\sum_{m=1}^{M}concat(\{\textbf{x}^m*\textbf{w}_m\}_{m=1}^{M})^m.w_{m,j}\}_{j=1}^{\mathcal{K}})
\end{equation}
\noindent where $\{\textbf{w}_m\}_{m=1}^{M}$ is a set of convolution kernels of length $K$ each and
$\{\{w_{m,j}\}_{m=1}^{M}\}_{j=1}^{\mathcal{K}}$ is a set of real values and $\textbf{o}$ is the output of the DWSC layer 
with $\mathcal{K}$ dimensions of length $L-K+1$.

A visualization of the DWSC layer with a chosen number of target dimension set to $2$ with a kernel size of $8$
is presented in Figure~\ref{fig:sep-conv} applied on an input MTS of dimension $3$.
It can be seen from this figure that the total number of filters to learn is $3$ instead of $6$ (SC layer)
with additional $6$ real values to learn (the chosen output dimension).

The total number of parameters learned by an SC and DWSC layers are $M.\mathcal{K}.K$ and $M.K + M.\mathcal{K}$ respectively.

\paragraph*{Residual Connections}

Deep learning models sometimes suffer from a common issue referred to as the vanishing gradient.
This issue is more common when the network's depth gets higher, resulting in a very deep model,
and during the backward phase of the optimization, the gradient may become zero.
To avoid this issue, the authors in~\cite{resnet-images} proposed the residual connections,
where instead of having one branch, the network gets divided into two branches.
The first branch serves as the encoding and feature extraction, and the second serves as the skip 
branch, where it simply uses an almost identity like function.
Both branches meet after a specific number of layers in an element-wise addition operation, resulting 
in what follows at layer $\lambda_l$:
\begin{equation}\label{equ:resnet}
    f_{l}(\theta_{l},g_{l-1}) = f_{1,l}(\theta_{1,l},g_{l-1}) + f_{2,l}(\theta_{2,l},g_{l-1})
\end{equation}
\noindent where $g_{l-1}$ is the output of the $l-1_{th}$ layer,
$f_{1,l}$ is an identity like function with parameters $\theta_{1,l}$,
and $f_{2,l}$ is a stack of layers with parameters $\theta_{2,l}$.

\paragraph*{Batch Normalization}

Batch Normalization (BN) is a technique used to improve the training speed and stability of neural networks 
by normalizing the inputs of each layer to have a mean of zero and a standard deviation of one.
This is necessary in order to reduce the chance of gradient exploding due to different range of values
in the features going from one layer to another.
This is done by calculating the mean and variance of the inputs within each mini-batch during 
training and then scaling and shifting the inputs based on these statistics. In the context of 
time series analysis, especially when used after 1D convolution layers, batch normalization can 
be particularly beneficial. Since 1D convolution layers are often used to extract temporal features 
from the time series data, the distribution of features across different time steps may vary significantly. 

Similar to other neural network layers, BN has two trainable parameters, called $\gamma$ and $\beta$.
Following the first z-normalization step that produces zero-mean and unit variance features, the BN layer 
learns how to shift and scale all the features to a new mean and variance.
Supposing a batch of $B$ MTS $\{\textbf{x}_i\}_{i=1}^{B}$ produced
by a convolution layer with $M$ channels of length $L$, applying the BN layer with parameters
$\gamma_1,\gamma_2,\ldots,\gamma_M$ and $\beta_1, \beta_2, \ldots, \beta_M$ is defined as follows:
\begin{equation}\label{equ:batch-norm-train}
    \textbf{o}_i = concat(\{\gamma_m.\dfrac{\textbf{x}_i^m - \mu_m}{\sigma_m} + \beta_m\}_{m=1}^M)
\end{equation}
\noindent where $\{\textbf{o}_i\}_{i=1}^{B}$ is a set of MTS of same shape as $\textbf{x}_i$, $\mu_m$ is the average of the channel $m$ of all samples in the batch
over the temporal axis, and $\sigma_m$ is its standard deviation.

Moreover, the BN layer also has two non-trainable parameters called $\mu_{mov}$ and $\beta_{mov}$ which contains a moving average 
of the mean and standard deviation of the input data $\boldsymbol{\mu}$ and $\boldsymbol{\sigma}$, for each dimension separately.
These two non-trainable are then used during inference to scale the features of new unseen samples.
These two parameters are calculated during training as follows:
\begin{equation}\label{equ:batch-norm-mvg1}
    \mu_{mov} = concat(\{\alpha_{BN}.\mu_{mov_m} + (1-\alpha_{BN}.\mu_{m})\}_{m=1}^M)
\end{equation}
\begin{equation}\label{equ:batch-norm-mvg2}
    \beta_{mov} = concat(\{\alpha_{BN}.\beta_{mov_m} + (1-\alpha_{BN}.\beta_{m})\}_{m=1}^M)
\end{equation}
\noindent where $\alpha_{BN}$ is the moving average parameter and $0 < \alpha_{BN} < 1$.

\paragraph*{Local Pooling (Max and Average)}

Pooling operations have been shown to be very effective for images throughout the years ever since the birth of deep learning~\cite{lecun2015deep}.
The motivation of doing local pooling operations is to reduce dimensionality resulting in a focus on more local important 
features extracted by the network.

There exist many local pooling layers, however two types are more common in the time series classification 
literature, the first being max pooling and the second being average pooling.
For the max pooling layer, such as for convolution layers, a kernel is defined with a specific length, however in this case 
the kernel does not have any trainable weights.
Instead, the kernel slides over the temporal axis of the series, and the values seen by the kernel are replaced by their maximum 
value.
The operation done by the max pooling layer with kernel size $K$ is defined below on an input univariate series $\textbf{x}$
of dimensions $(L)$:
\begin{equation}\label{equ:max-pooling}
    o_t = \max(x[t:t+k])
\end{equation}
\noindent with $t$ $\in~[1,L-K+1]$ and $\textbf{o}=\{o_1,o_2,\ldots,o_{L-K+1}\}$ is a univariate series with length $L-K+1$.

The average pooling layer is defined in the same way as the max pooling, however instead of choosing the maximum between 
the values seen by the kernel, the average value replaces them.
The operation done by the average pooling layer with kernel size $K$ is defined below on an input univariate series $\textbf{x}$
of dimensions $(L)$:
\begin{equation}\label{equ:avg-pooling}
    o_t = \dfrac{1}{K}\sum_{k=1}^{K} x[t:t+k]
\end{equation}
\noindent with $t$ $\in~[1,L-K+1]$ and $\textbf{o}=\{o_1,o_2,\ldots,o_{L-K+1}\}$ is a univariate series with length $L-K+1$.

Similar to convolution layers, max and average pooling layers can both be applied on MTS, however in this case the operation is 
applied on each channel independently and the output dimension will be the same as the input.
Moreover, dilation rates can also be used for local pooling layers similar to convolutions, see Eq.~\ref{equ:convolution1d-dilation}.

\emph{It is important to note, that by default all local pooling layers utilize a stride equals to the kernel size, unless 
specified to use another stride. The above output length calculations ($L-K+1$) is in the case where strides are set to $1$,
however in default mode, the output length is $\lceil \dfrac{L - K + 1}{K}\rceil$.}

\paragraph*{Global Pooling (Max and Average)}

Global pooling is a powerful technique often employed in neural network architectures 
for dimensionality reduction and feature summary. Unlike local pooling, which focuses on local 
features within specific regions, global pooling computes the summary of feature maps across the entire spatial 
dimensions, providing a global perspective of the input data.

Similar to local pooling, two main global pooling layers are used for time series classification in deep learning models,
the first being Global Max Pooling (GMP) and the second being Global Average Pooling (GAP).
In the case of time series, the global pooling layers are mostly used posterior to all the feature extraction layers.

The GMP layer receives an input dimension of $(L,M)$, where $L$ is the length of the series and $M$ is its dimensions,
and outputs a vector, per series, of dimension $(M,)$, where each point of the vector is the maximum value over all the time 
axis of each dimension.
We define below the GMP operation over an input time series $\textbf{x}$ of dimension $(L,M)$:
\begin{equation}\label{equ:gmp-layer}
    \textbf{v} = concat(\{\max(\textbf{x}^m[1:L])\}_{m=1}^{M})
\end{equation}
\noindent where $\textbf{v} = \{v_1,v_2,\ldots,v_M\}$ is a vector of dimension $(M,)$.

Similar to the GMP layer, the GAP layer receives the same input dimension and outputs a vector $\textbf{v}$ also of dimension 
$(M,)$, however each point of the vector is the average value over the time axis of each dimension.
In simpler ways, we define in what follows the operation done in the GAP layer over a series $\textbf{x}$ of dimensions $(L,M)$:

\begin{equation}\label{equ:gap-layer}
    \textbf{v} = concat(\{\dfrac{1}{L}\sum_{t=1}^{L}x_t^m\}_{m=1}^{M})
\end{equation}
\noindent where $\textbf{v} = \{v_1,v_2,\ldots,v_M\}$ is a vector of dimension $(M,)$.

As mentioned above, global pooling layers are used at the last feature extraction step of the network,
this is because it can be now fed  to an FC layer (see Eq.~\ref{equ:fc-layer}) with a softmax activation
for the classification task.

\paragraph*{Temporal Self-Attention}

The Self-Attention mechanism has shown to have a significant impact in Natural Language Processing (NLP) ever since the 
birth of Transformers for language translation in the paper \emph{Attention Is All You Need}~\cite{attention-all-you-need}.
The Self-Attention mechanism, adapted from the original Attention mechanism~\cite{attention-mechanism}, allows the model 
to learn about the dependency between features spread along a temporal axis.
This information is then used to transform the input features into a new space that is more compact and contains denser information 
about important features.

To explain how does the Self-Attention layer is able to do the operation mentioned above, we will assume again an input time series $\textbf{x}$
of shape $(L,M)$, supposing that this time series is actually the output of previous feature extraction layers such as CNNs.
We detail below each step of the Self-Attention mechanism:

\textbf{First}, Self-Attention mechanism has a unique feature of being order invariant, for this reason we use Positional Encoders (PEs)
to add position information to each element in the sequence. Sinusoidal functions generate these encoding,
which are added (element-wise) to the input embeddings, ensuring the model can use the sequence order. The PE for position 
$pos~\in~[1,L]$:

\begin{equation}\label{equ:pe-sin}
PE_{(pos, 2k)} = \sin\left(\frac{pos}{w_k}\right)
\end{equation}
\begin{equation}\label{equ:pe-cos}
PE_{(pos, 2k+1)} = \cos\left(\frac{pos}{w_k}\right)
\end{equation}
\noindent where $k~\in~[0,\dfrac{d_{model}}{2}]$, the frequency 
$w_k=10000^{2k/d_{model}}$ and $d_{model}$ is the dimension of the embeddings.

These encoding ensure the sequence includes both content and positional information.
This type of PE is commonly referred to in the literature as Absolute Positional Encoding (APE). 

\textbf{Second}, the Self-Attention layers transforms each time stamp of the input time series $\textbf{x}$ from dimension $M$ to 
dimension $d_{model}$, a hyper-parameter of the Self-Attention layer.
This first step is done three times independently to produce three different representations of the input series, referred to as:
(1) Query $\textbf{Q}$, (2) Key $\textbf{K}$ and (3) Value $\textbf{V}$.
The Query and Key are used to find the dependency information between time stamps in $\textbf{x}$, for which this information 
is then used to transform the Value to a new more compact space.
This is done by simply defining three FC layers (see Eq.~\ref{equ:fc-layer}) with weight matrices $W_{\textbf{Q}}$,
$W_{\textbf{K}}$ and $W_{\textbf{Q}}$, for the Query, Key and Value respectively of shape $(M,d_{model})$ each.
The three matrices are used to transform each time stamp of $\textbf{x}$ to a new space of different dimensions, as defined below 
to produce $\textbf{Q}$, $\textbf{K}$ and $\textbf{V}$ of shape $(L,d_{model})$ each:
\begin{equation}\label{equ:attention-query}
    \textbf{Q} = concat_{temporal~axis}(\{\textbf{x}_t\circ W_{\textbf{Q}}\}_{t=1}^{L})
\end{equation}
\begin{equation}\label{equ:attention-key}
    \textbf{K} = concat_{temporal~axis}(\{\textbf{x}_t\circ W_{\textbf{K}}\}_{t=1}^{L})
\end{equation}
\begin{equation}\label{equ:attention-value}
    \textbf{V} = concat_{temporal~axis}(\{\textbf{x}_t\circ W_{\textbf{V}}\}_{t=1}^{L})
\end{equation}
\noindent where the concatenation is over the temporal axis and the matrix multiplication $\circ$ is over the dimension axis.

\textbf{Third}, given that both $\textbf{Q}$ and $\textbf{V}$ are different representations but of the same input sequence 
$\textbf{x}$, the Self-Attention layer utilizes these two sequences in order to find dependency information between each time stamp 
and all the other time stamps.
This is done by calculating the attention score matrix as follows:
\begin{equation}\label{equ:attention-score}
    Att = soft\max(\dfrac{Q\circ K^{T}}{\sqrt{d_{model}}})
\end{equation}
\noindent where $Att$ is called the attention score matrix of shape $(L,L)$ and the $soft\max$ operation is performed over the 
column's axis, producing per row a probability distribution how much the row time stamp is correlated with all the column time 
stamps.
The scaling factor $1/\sqrt{d_{model}}$ is utilized to avoid high values produced in the dot products, resulting in values close 
to the $soft\max$ limits where the gradient can be very small.

\textbf{Fourth}, the above attention matrix $Att$ is then used to transform the sequence $\textbf{V}$ into a new representation,
making the output sequence more compact in terms of dependency information between time stamps.
The transformed sequence goes through a dimension change using another FC layer with weight matrix $W_{\textbf{o}}$ with shape 
$(d_{model},M)$ to change back to the original dimension of $\textbf{x}$:
\begin{equation}\label{equ:attention-transform}
    \textbf{o} = (Att\circ\textbf{V}) \circ W_{\textbf{o}}
\end{equation}
\noindent where $\textbf{o}$ has the same dimension as $\textbf{x}$.

It is common to use the concept of multi-head attention, where the same procedure described above is repeated independently 
$H$ times (in parallel), where $H$ is the number of heads.
The outputs of each head are finally concatenated and the final transformation matrix $W_{\textbf{o}}$ is used on the concatenated 
transformations of all heads.

\paragraph*{Recurrent Layers}

Recurrent Neural Networks (RNNs)~\cite{rnn-paper} are specialized neural networks designed for processing sequential data. 
They maintain a hidden state that captures information from previous inputs, making them ideal for tasks 
like language modeling, speech recognition~\cite{graves2013speech}, and sequence-to-sequence learning~\cite{seq-to-seq}
RNNs are effective at learning patterns and dependencies in sequences, leveraging their ability to remember context over time.

There exists three different recurrent layers that have been proposed during the last three decades:
(1) Simple RNN~\cite{rnn-paper}, (2) Long Short-Term Memory (LSTM)~\cite{lstm-paper} and Gated Recurrent Unit (GRU)~\cite{gru-paper}.
In what follows, we present each of these layers briefly with their mathematical formulation.

\begin{enumerate}
    \item \textbf{Elman Recurrent Neural Network (Simple RNN)}, proposed in~\cite{rnn-paper}, is one of the simplest
    forms of RNNs. It consists of a single hidden layer that maintains a recurrent connection to itself, allowing it
    to capture sequential dependencies. For an input time series $\textbf{x}$ of length $L$ and dimension $M$, applying 
    once recurrence step $t$ where $t~\in~[1,L]$ using the Simple RNN is defined as:
    \begin{equation}\label{equ:rnn-cell-hidden}
        \textbf{h}_t = \sigma(W_{hx}\circ\textbf{x}_t + W_{hh}\circ\textbf{h}_{t-1}+\textbf{b}_{h})
    \end{equation}
    \begin{equation}\label{equ:rnn-cell-output}
        \textbf{o}_t = \sigma(W_{oh}\circ\textbf{h}_t + \textbf{b}_{o})
    \end{equation}
    \noindent where $\textbf{h}_t$ is the hidden state of time stamp $t$ of dimension $d_{hidden}\neq M$,
    $W_{hx}$ is the input-to-hidden transformation matrix of shape $(d_{hidden},M)$, 
    $W_{hh}$ is the hidden-to-hidden transformation matrix of shape $(d_{hidden},d_{hidden})$,
    $b_{h}$ is the hidden layer bias vector of dimension $d_{hidden}$,
    $W_{oh}$ is the hidden-to-output transformation matrix of shape $(M,d_{hidden})$,
    $\textbf{b}_{o}$ is the output bias vector of dimension $M$,
    $\sigma$ is the activation function commonly a sigmoid or hyperbolic tangent function
    and $\textbf{o}$ is the output series of length $L$ and dimension $M$

    \item \textbf{Long Short-Term Memory (LSTM)}, proposed in~\cite{lstm-paper}, addresses the vanishing gradient 
    problem faced by traditional RNNs, enabling them to capture long-range dependencies more effectively. 
    LSTM introduces a gating mechanism that regulates the flow of information, allowing the network to 
    selectively remember or forget information over time.
    The mathematical formulation of the LSTM layer is defined as follows:
    \begin{equation}\label{equ:lstm-cell-forget}
        \textbf{f}_t = sigmoid(W_\textbf{f}\circ concat(\textbf{h}_{t-1},\textbf{x}_t) + \textbf{b}_f)
    \end{equation}
    \begin{equation}\label{equ:lstm-cell-input}
        \textbf{e}_t = sigmoid(W_\textbf{e}\circ concat(\textbf{h}_{t-1},\textbf{x}_t) + \textbf{b}_e)
    \end{equation}
    \begin{equation}\label{equ:lstm-cell-output}
        \textbf{o}_t = sigmoid(W_\textbf{o}\circ concat(\textbf{h}_{t-1},\textbf{x}_t) + \textbf{b}_o)
    \end{equation}
    \begin{equation}\label{equ:lstm-cell-candidate}
        \tilde{\textbf{C}}_t = \tanh(W_\textbf{C}\circ concat(\textbf{h}_{t-1},\textbf{x}_t) + \textbf{b}_C)
    \end{equation}
    \begin{equation}\label{equ:lstm-cell-state}
        \textbf{C}_t = \textbf{f}_t\odot\textbf{C}_{t-1} + \textbf{e}_t\odot\tilde{\textbf{C}}_t
    \end{equation}
    \begin{equation}\label{equ:lstm-cell-hidden}
        \textbf{h}_t = \textbf{o}_t\odot\tanh(\textbf{C}_t)
    \end{equation}
    \noindent where $\textbf{h}_t$, is the hidden state vector of dimension $d_{hidden}$,
    $\textbf{f}_t$, $\textbf{e}_t$ and $\textbf{o}_t$ are the forget, input and output gates vectors respectively
    of dimension $d_{hidden}$ with
    $W_\textbf{f}$, $W_\textbf{e}$, $W_\textbf{o}$ as their respective transformation matrices
    of shape $(d_{hidden},M+d_{hidden})$ each and
    $\textbf{b}_\textbf{f}$, $\textbf{b}_\textbf{e}$, $\textbf{b}_\textbf{o}$ their respective bias vectors
    of dimension $d_{hidden}$ each.
    $\tilde{\textbf{C}}_i$ is the candidate cell state vector, $\textbf{C}_i$ is the cell state vector both of 
    dimension $d_{hidden}$ and finally $W_\textbf{C}$ is the state cell candidate transformation matrix of shape 
    $(d_{hidden}, M)$ and $\textbf{b}_\textbf{C}$ is the cell state bias vector of dimension $d_{hidden}$.
    $\odot$ denotes element-wise and $\circ$ denotes the matrix multiplication operation.

    \item \textbf{Gated Recurrent Unit (GRU)}, proposed in~\cite{gru-paper}, is a variation of the LSTM network 
    that simplifies its architecture while maintaining comparable performance. The GRU combines the forget and 
    input gates into a single update gate, reducing the number of parameters and computational complexity. 
    The mathematical formulation of the GRU layer is defined as follows:
    \begin{equation}\label{equ:gru-update-gate}
        \textbf{z}_t = \sigma(W_\textbf{z} \circ concat(\textbf{h}_{t-1}, \textbf{x}_t) + \textbf{b}_\textbf{z})
    \end{equation}
    \begin{equation}\label{equ:gru-reset-gate}
        \textbf{r}_t = \sigma(W_\textbf{r} \circ concat(\textbf{h}_{t-1}, \textbf{x}_t) + \textbf{b}_\textbf{r})
    \end{equation}
    \begin{equation}\label{equ:gru-candidate-activation}
        \tilde{\textbf{h}}_t = \tanh(W_\textbf{h} \circ concat(\textbf{r}_t \odot \textbf{h}_{t-1}, \textbf{x}_t) + \textbf{b}_\textbf{h})
    \end{equation}
    \begin{equation}\label{equ:gru-hidden-state}
        \textbf{h}_t = (\textbf{1}_{d_{hidden}} - \textbf{z}_t) \odot \textbf{h}_{t-1} + \textbf{z}_t \odot \tilde{\textbf{h}}_t
    \end{equation}
    \noindent where $\textbf{z}_t$ and $\textbf{r}_t$ are the update and reset gates respectively,
     $\tilde{\textbf{h}}_t$ is the candidate activation, and $\textbf{h}_t$ is the hidden state, of dimension $d_{hidden}$.
     $\odot$ denotes element-wise and $\circ$ denotes the matrix multiplication operation.
     $W_\textbf{z}$, $W_\textbf{r}$, and $W_\textbf{h}$ are weight matrices of shape $(d_{hidden}, M + d_{hidden})$,
     and $\textbf{b}_\textbf{z}$, $\textbf{b}_\textbf{r}$, and $\textbf{b}_\textbf{h}$ are bias terms of 
     dimension $d_{hidden}$. $\textbf{1}_{d_{hidden}}$ is a column vector of $1$s of dimension $d_{hidden}$.
\end{enumerate}

\paragraph{Different Neural Network Architectures for Time Series Classification}

During the last decade, a significant amount of architectures has been proposed addressing the task of TSC.
A detailed benchmark paper~\cite{dl4tsc} questioned the need of a fair comparison between most of these architectures
over all the datasets of the UCR/UEA archives~\cite{ucr-archive,uea-archive}.
Their choice of architectures to include in the benchmark depended on their ability to reproduce the model from scratch.
This benchmark, on some level, became a starting point of addressing TSC tasks with deep learning methods.
In this section, we present some architectures used in this benchmark~\cite{dl4tsc} as well as new architectures proposed
for both univariate and multivariate TSC ever since 2019 (the publication year of the benchmark).

\paragraph*{Multi Layer Perceptron (MLP)}

\begin{figure}
    \centering
    \caption{The MultiLayer Perceptron (MLP) architecture~\cite{fcn-resnet-mlp-paper} for Time Series Classification.}
    \label{fig:mlp}
    \includegraphics[width=\textwidth]{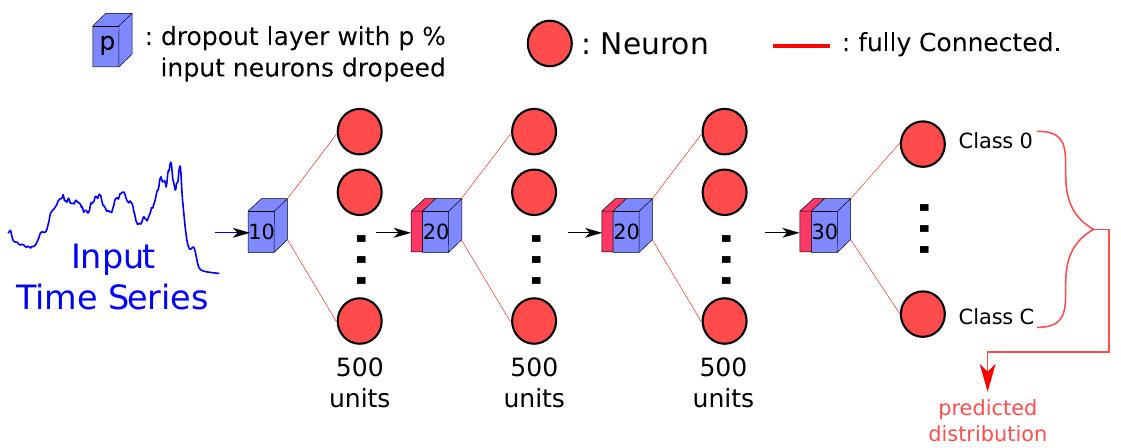}
\end{figure}

The concept of Multilayer Perceptrons (MLPs) originated from the field of artificial neural networks, where they 
were developed as a class of feedforward neural networks consisting of multiple layers of nodes \cite{mlp-original-paper}. 
Researchers began exploring the application of MLPs to time series classification due to their ability to model complex 
relationships between input features.
The authors in~\cite{fcn-resnet-mlp-paper} proposed an MLP architecture, presented in Figure~\ref{fig:mlp}, for TSC.
The architecture is made of three hidden FC layers each followed by a ReLU activation function and a Dropout layer.
Dropout layers simply uses a random $p \%$ of the input neurons and set them to zero and scale the non-dropped input neurons 
by $\dfrac{1}{1-p}$, used to avoid overfitting the model on training data, where $p$ is the drop rate parameter.
The last Dropout layer's output is then fed to an FC layer for the classification task, see Figure~\ref{fig:mlp} for a detailed 
view on the parameters of the MLP architecture.
However, a significant limitation of MLPs in this context is their inability to
effectively capture local temporal dependencies within the data, as they process input data in a fixed manner without 
considering the sequential nature of time series data. This limitation led to the exploration of other neural network
architectures better suited for temporal data, such as Recurrent Neural Networks (RNNs) and Convolutional Neural Networks 
(CNNs).

\paragraph*{Time Convolutional Neural Network (TimeCNN)}

\begin{figure}
    \centering
    \caption{Time-CNN~\cite{time-cnn} architecture for Time Series Classification.}
    \label{fig:time-cnn}
    \includegraphics[width=\textwidth]{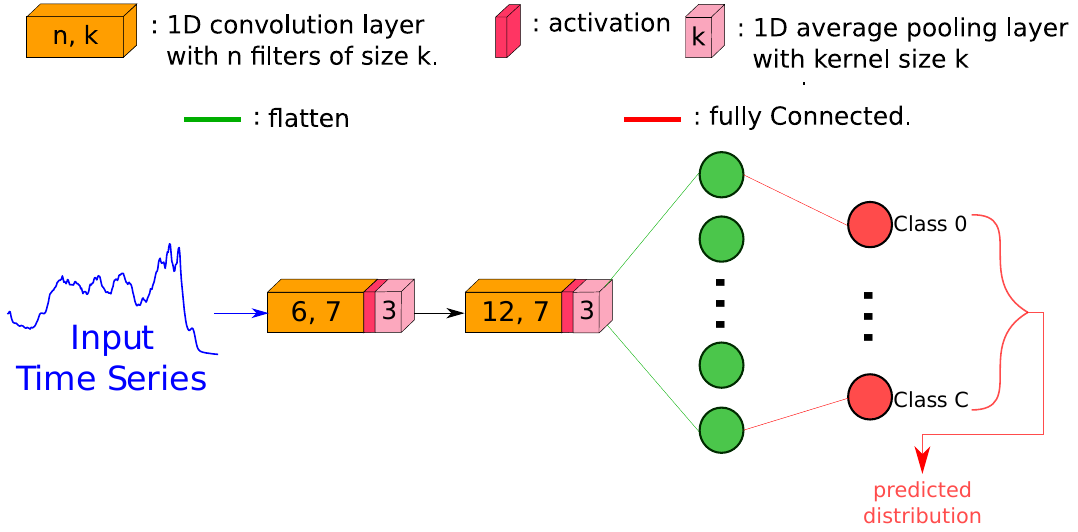}
\end{figure}

Ever since AlexNet~\cite{alexnet-paper} has been released in 2012 and highlighted the performance of deep CNN models on image 
classification~\cite{imagenet-paper}, a significant amount of researchers started to wonder on the need to include deep learning 
into other applications.
The authors in~\cite{time-cnn} proposed a CNN architecture, based on the image classification CNN in~\cite{lecun2015deep},
on one dimensional temporal data.
The architecture, presented in Figure~\ref{fig:time-cnn}, consists of two convolution blocks each containing a 1D convolution 
layer followed by a sigmoid activation function and a local average pooling layer of default strides (see Figure~\ref{fig:time-cnn} for detailed 
view on the parameters of TimeCNN).
Following the second convolution block, the output is an MTS of $7$ channels and the length depending on the input time series 
characteristics.
This output MTS is flattened to form a large one dimensional vector that is then fed to an FC layer for the classification 
task.

\paragraph*{Fully Convolutional Network (FCN)}

\begin{figure}
    \centering
    \caption{The Fully Convolutional Network (FCN)~\cite{fcn-resnet-mlp-paper} architecture for Time Series Classification.}
    \label{fig:fcn}
    \includegraphics[width=\textwidth]{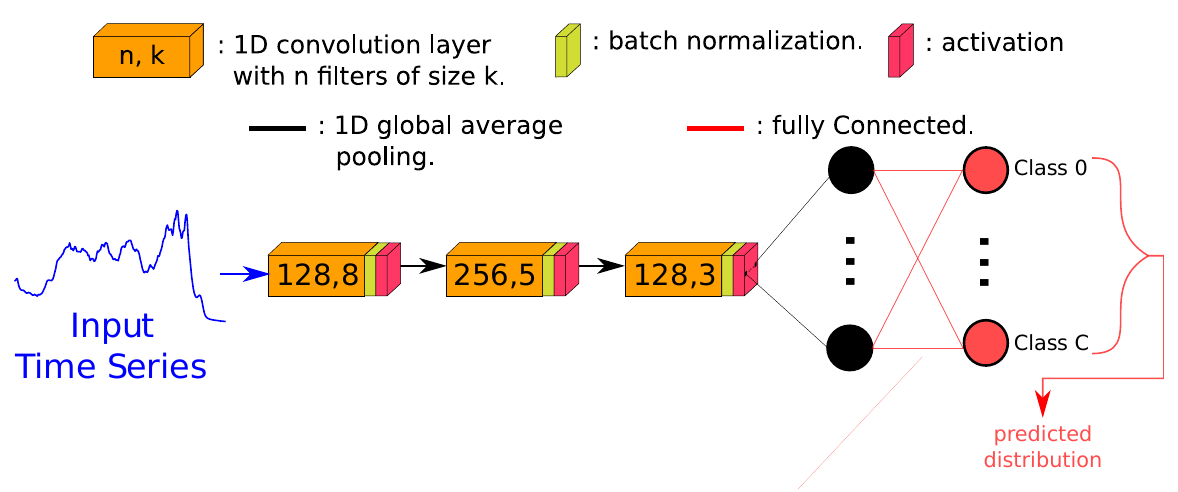}
\end{figure}

The authors in~\cite{fcn-resnet-mlp-paper} questioned the need of local pooling layers and proposed instead a 
Fully Convolutional Network (FCN), composed of three convolution blocks, each containing a one dimensional convolution 
layer followed by a batch normalization layer and a ReLU activation function.
The FCN architecture is presented in Figure~\ref{fig:fcn} including all the parameter setup proposed in~\cite{fcn-resnet-mlp-paper}.
The authors of FCN argues that the replacement of local pooling layers by the batch normalization not only enhances the 
performance given that local pooling can lose some information, but increases the speed of convergence of the model as well.
The convolution layers used in FCN applies a zero-padding on the input, hence the length of the series is preserved throughout 
the network.
This padding operation ensures that the network can detect some patterns on the edges of the series.
The FCN architecture feeds the last activation layer to a global pooling layer, specifically a GAP, instead of flattening
in order to reduce the number of parameters to learn in the last classification FC layer.

\paragraph*{Residual Network (ResNet)}

\begin{figure}
    \centering
    \caption{The Residual Network (ResNet) architecture~\cite{fcn-resnet-mlp-paper} for Time Series Classification.}
    \label{fig:resnet}
    \includegraphics[width=\textwidth]{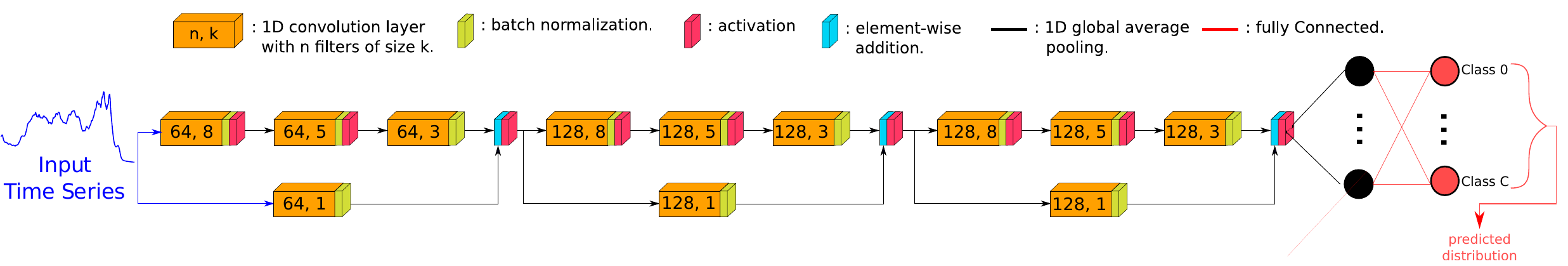}
\end{figure}

Since the impact of residual connections have been successful for image classification~\cite{resnet-images}, the authors 
in~\cite{fcn-resnet-mlp-paper} proposed to enhance the FCN architecture with this kind of operations.
The authors in~\cite{fcn-resnet-mlp-paper} argues the need of residual connections given that neural networks also may suffer 
from the vanishing gradient problem for TSC.
For this reason, the authors proposed ResNet for TSC, presented in Figure~\ref{fig:resnet} with its parameter setup.
This architecture consists of three residual blocks, where each block is an FCN architecture without the GAP and classification 
layer.
Each residual block contains an element-wise addition between its input layer and output layer, with the residual connection 
including a bottleneck layer (PWC see Eq.~\ref{equ:pointwise-conv}) to adjust dimensions.
The convolution layers, such as in FCN, utilize a zero-padding to ensure the edge pattern detection, resulting as well in 
equal length input/output at the beginning and end of the network.
The last activation layer of ResNet is fed to a GAP layer followed by an FC layer for classification.

\paragraph*{Encoder}

\begin{figure}
    \centering
    \caption{The Encoder architecture~\cite{encoder-paper} for Time Series Classification.}
    \label{fig:encoder}
    \includegraphics[width=\textwidth]{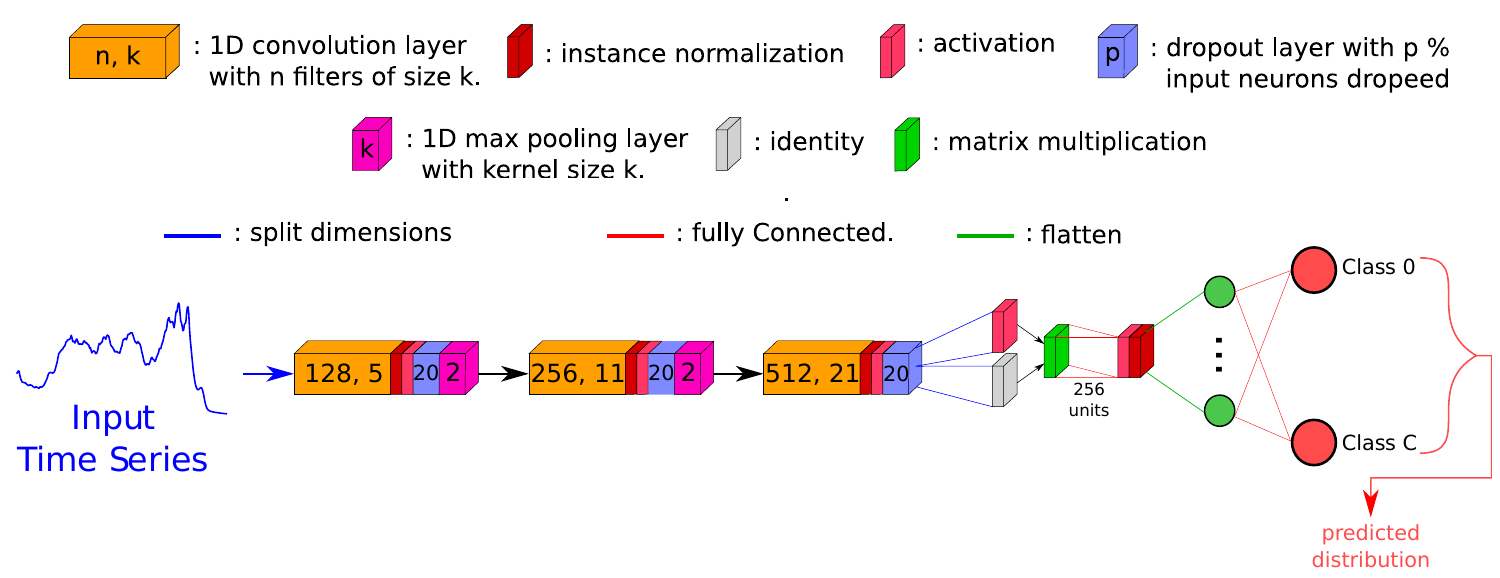}
\end{figure}

Motivated by FCN~\cite{fcn-resnet-mlp-paper}, the authors in~\cite{encoder-paper} proposed a novel hybrid deep learning model, 
Encoder, that replaces the GAP layer by a slightly different version of Self-Attention.
The Encoder architecture, presented in Figure~\ref{fig:encoder}, consists on three convolution blocks, each containing a one dimensional convolutional
layer, followed by an Instance Normalization (IN) layer instead of a BN layer, a Parametric ReLU activation function, a dropout layer and 
finally a local max pooling layer.
The IN layer consists on using the same normalization concept of BN however it is done per example in the batch instead of averaging
statistics over all samples in the batch.
The third convolution block however does not contain a local max pooling layer, instead the outputs are split on the dimension axis 
into two parts used for a Self-Attention mechanism.
The output of the attention layer goes through an FC transformation layer followed by a flattening and the last FC classification layer.

\paragraph*{Neural Network Ensemble (NNE)}

Ensemble models in machine learning~\cite{hive-cote2.0} has shown to have 
a significant impact, for instance until the year 2020, the state-of-the-art 
hybrid ensemble model, HIVE-COTE~\cite{hive-cote}, consisted of $36$ classifiers 
ensembled.
The authors in~\cite{nne-paper} studied the impact of ensembling in deep 
learners given that HIVE-COTE~\cite{hive-cote} ensembles non-deep learners.
In~\cite{nne-paper}, the authors proposed the Neural Network Ensemble (NNE)
consisting of $6$ different deep learning architectures with $10$ different initialization,
hence a total of $60$ models, ensembled posterior to training.
NNE highlighted that it can achieve the performance of HIVE-COTE over the UCR 
archive as well as outperform significantly an ensemble of any other architecture 
alone, highlighting that the importance of hybrid ensembles.

\paragraph*{InceptionTime}

\begin{figure}
    \centering
    \caption{The Inception architecture~\cite{inceptiontime-paper}
    for Time Series Classification.}
    \label{fig:inception}
    \includegraphics[width=\textwidth]{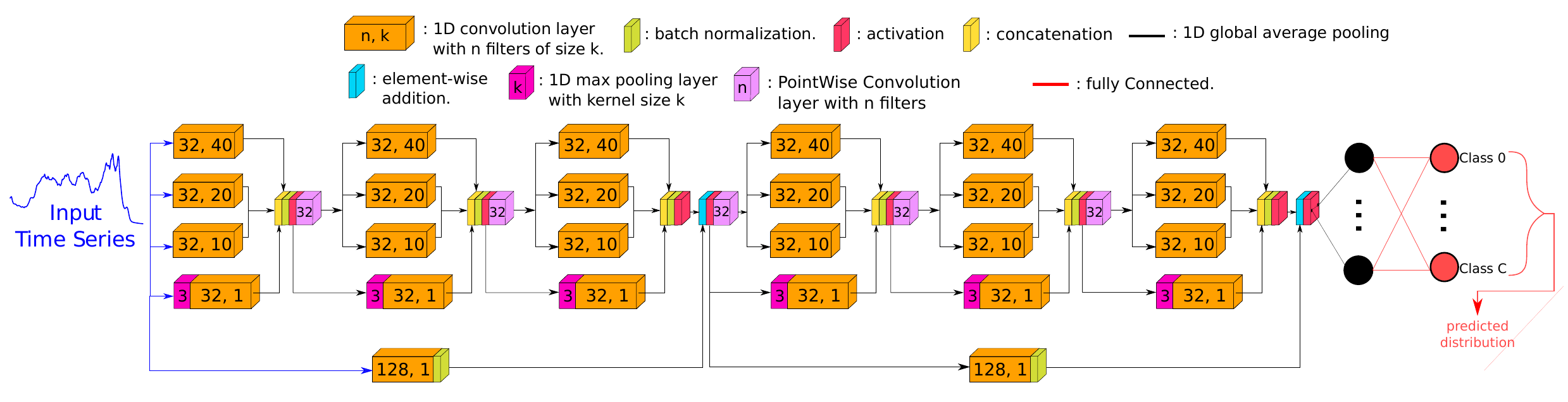}
\end{figure}

In~\cite{inceptiontime-paper}, the authors argued the need to \emph{find the 
AlexNet for Time Series Classification} given the increase in number of datasets 
available and the high similarity that exists between them.
For this reason, a deeper architecture should be proposed to outperform the 
current state-of-the-art deep learning model ResNet~\cite{fcn-resnet-mlp-paper,dl4tsc}.
Motivated from the impact of Inception architecture for image classification,
the authors in~\cite{inceptiontime-paper} proposed an adaptation of the Inception
architecture for TSC, specifically the authors were based on the fourth version 
of Inception on image classification~\cite{inceptionv4-paper}.
The Inception architecture adapted for time series data is presented in 
Figure~\ref{fig:inception} with a detailed view on its parameters' setup.

The Inception architecture consists of two Inception-blocks each containing a
residual connection~\cite{resnet-images} connecting their input and output.
Within each Inception-block, there is three Inception-modules connected in series,
each containing three convolution layers in parallel (this is referred to later as multiplexing convolution)
of different kernel size applied 
on the same input, and a local max pooling layer followed by a PWC layer for dimension adjustment.
Each Inception-module, if its input has dimension higher to $1$, applies a PWC layer 
in order to reduce the number of filters to learn in the following convolution layers.
The output of the three convolution and the max pooling layers are concatenated on the 
channel axis and fed to a BN layer and a ReLU activation function.
The last activation layer goes through a global pooling operation, specifically a GAP layer,
before being fed to an FC classification layer.

The authors in~\cite{inceptiontime-paper}, seeing the impact of ensemble models,
proposed InceptionTime, an ensemble of five Inception architectures each trained with 
different initialization.
In 2020, InceptionTime became the state-of-the-art deep learning model for TSC and shown to 
have even less difference in performance, statistically, with HIVE-COTE2.0~\cite{hive-cote2.0}.
InceptionTime, not only highlighted its ability to achieve HC2.0 performance, but as well as it being 
more efficient in terms of training runtime in function of both training dataset size and length of
time series.

\paragraph*{Disjoint Convolutional Neural Network (Disjoint-CNN)}

All the previously presented architectures were originally proposed for a 
general setup of TSC, meaning they were not constructed in a manner to address 
univariate time series or multivariate time series specifically, and they can be applied 
to both and have been evaluated on both.
However, some researchers argued that handling MTS data is not the same as handling 
UTS data and questioned the way convolution operations are being done over MTS.

\begin{figure}
    \centering
    \caption{The Disjoint-CNN architecture~\cite{disjoint-cnn-paper} for Time 
    Series Classification.}
    \label{fig:disjoin-cnn}
    \includegraphics[width=\textwidth]{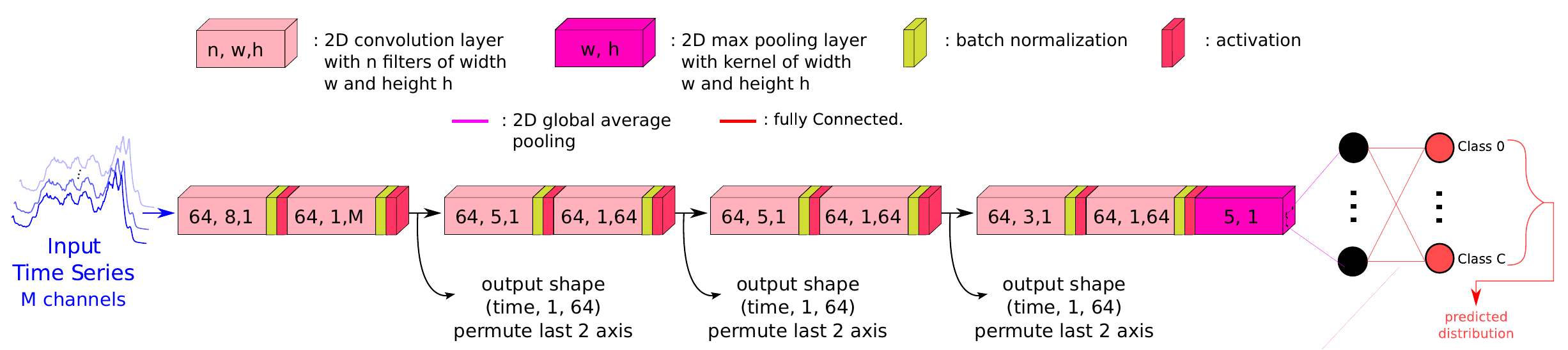}
\end{figure}

In~\cite{disjoint-cnn-paper}, the authors proposed the Disjoint Convolutional 
Neural Network (Disjoint-CNN), composed of, following the naming of the authors,
1+1D convolution layers (see Figure~\ref{fig:disjoin-cnn} for a detailed view on the architecture).
The 1+1D convolution layer are two convolution operations operated in series,
the first being a temporal convolution and the second being a spatial convolution.
Given an input time series of $M$ dimensions and length $L$, the temporal convolution layer 
is, in other words, a 2D convolution layer with a kernel of width $K>1$ and height of $1$.
This ensures that the convolution layer will not sum up the outputs as done on MTS data 
with 1D convolution layers (see Eq.~\ref{equ:conv-layer-multi}).
The spatial convolution is as well a 2D convolution layer however with a kernel of 
width $1$ and height $M$, hence learning a linear combination of temporal features over
different dimensions.
The core idea of 1+1D convolution blocks is very similar to what 1D DWSC does, however in this case 
more parameters are learned in the network.

Each of the temporal and spatial convolution layers is followed by a BN layer 
and an ELU activation function, forming a 1+1D convolution block.
The Disjoin-CNN architecture is made of four 1+1D convolution block in series,
followed by a local max pooling operation and finishing, just like Inception~\cite{inceptiontime-paper},
by a GAP layer and an FC classifier.

\paragraph*{Convolutional Transformer (ConvTran)}

\begin{figure}
    \centering
    \caption{The ConvTran architecture~\cite{convtran-paper} for Time 
    Series Classification.}
    \label{fig:conv-tran}
    \includegraphics[width=\textwidth]{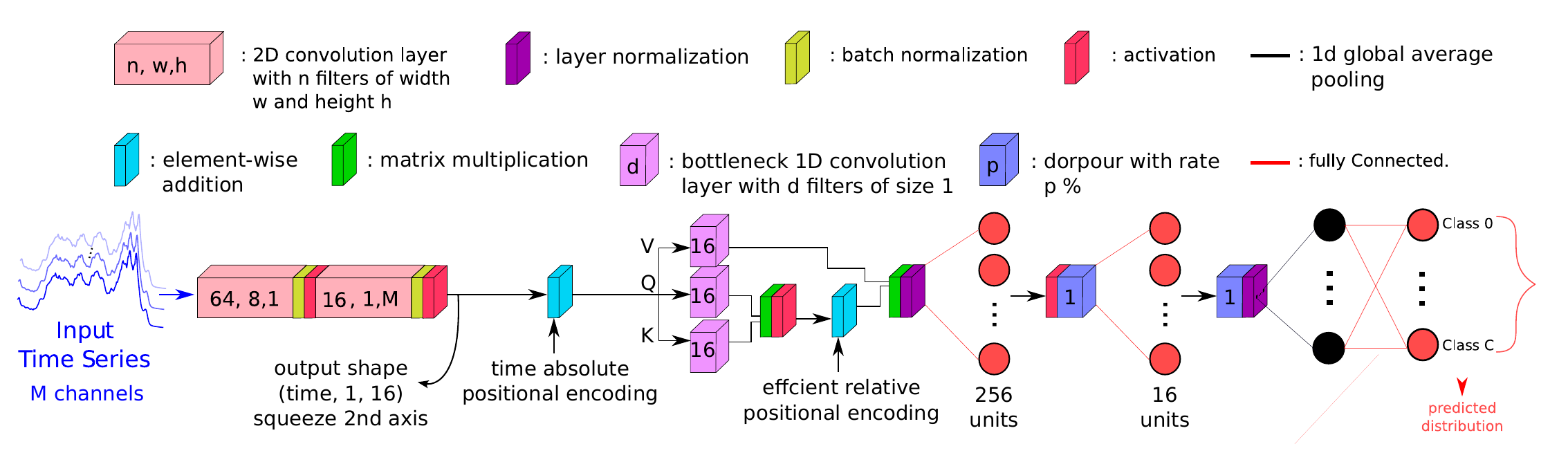}
\end{figure}

Researchers wondered the impact of Transformers and Self-Attention~\cite{attention-all-you-need}
when addressing the task of TSC.
Although no work has been published addressing univariate data, the authors in~\cite{convtran-paper}
proposed the first working transformer on multivariate TSC.
The proposed network, the Convolutional Transformer (ConvTran), consists of two 
phases, the time series encoder and the Self-Attention mechanism.
We present in Figure~\ref{fig:conv-tran} a detailed view on the ConvTran architecture with its detailed parameters' setup.
The time series encoder ensures that the attention mechanism is being applied 
over a space in which each time stamp represents one patch of the input MTS space.
The encoder used in ConvTran is a 1+1D convolution block~\cite{disjoint-cnn-paper}.

The ConvTran has two more contributions as well, as it adapts the frequency 
of the $sin$ and $cos$ functions of the Positional Encoder in
Eqs.~\ref{equ:pe-sin} and~\ref{equ:pe-cos}.
This frequency adjustment is a form of normalization to the input length and 
dimension before feeding the embeddings to the Self-Attention layer.
This normalization step is essential given that the original PE~\cite{attention-all-you-need}
was proposed for language models presenting a high dimensionality, which is not the case on average 
with the MTS datasets available in the literature.
The authors in~\cite{convtran-paper} showcased that the unnormalized PE suffers from 
its lack of ability to reflect similarity between different time stamps.
The proposed normalization to the Absolute Positional Encoder (APE) is simply the following:
\begin{equation}\label{equ:tape}
    w_k = \dfrac{w_k . d_{model}}{L}
\end{equation}
\noindent where $w_k$ is the frequency of the $cos$ and $sin$ functions in 
Eqs.~\ref{equ:pe-sin} and~\ref{equ:pe-cos}, $d_{model}$ is the dimension of the input
embedding and $L$ is the length of the series.
This normalized version of APE is referred to as Time APE (tAPE).

The second contribution of the ConvTran~\cite{convtran-paper} is an adaptation of the Relative 
Positional Encoding (RPE), which was first proposed on language models~\cite{relative-pe-paper1}.
The RPE is applied at the query and keys space.
RPE in self-attention is motivated by the need 
to encode the relative positions of tokens, rather than absolute positions, 
to better capture the relationships between tokens irrespective of their 
absolute positions in the sequence. This approach improves the model's 
ability to generalize across different sequence lengths and better handles 
long-range dependencies. It enhances performance in tasks where the relative 
positioning of words or tokens is crucial for understanding context and 
meaning.
ConvTran utilizes a shift based RPE instead of an index based one, taking into consideration 
that there should be a unique positional encoding for indices with a specific shift between them.
This results in a set of scalars $w_{\delta=|i-j|}$, learnable, where $i$ and $j$ 
$\in~[1,L]$ and $L$ is the length of the embedded series.
This would reduce the number of parameters to learn in the RPE to $2L-1$ instead of $(2L-1)d_{model}$.
This proposed RPE is referred to as Efficient RPE (eRPE).

In 2023, ConvTran became the state-of-the-art deep and non-deep learning model 
for multivariate TSC evaluated on the multivariate TSC UEA archive~\cite{uea-archive},
outperforming the last state-of-the-art models,
InceptionTime and ROCKET.

\subsection{Time Series Extrinsic Regression}

Time Series Extrinsic Regression (TSER) stands as a significant advancement in the 
domain of time series analysis, offering a departure from traditional 
intrinsic methods by focusing on predicting continuous scores rather 
than discrete classes. Unlike its classification counterpart, which 
aims to assign time series data into predefined categories, extrinsic 
regression is concerned with forecasting continuous values based on both 
temporal dynamics and external factors.
In the last decade, many models have been proposed to address the task 
of TSER.
These models leverage a variety of techniques, ranging from classical 
linear regression to more sophisticated algorithms such as ensemble 
methods and neural networks.

Extrinsic regression exhibits a wide array of applications, extending 
its reach across domains such as finance, healthcare, and environmental 
science. For example, it serves as a vital tool in satellite image analysis, 
where its application involves estimating live fuel moisture content in 
vegetation~\cite{tser-live-fuel-moisture} to mitigate the risk of wildfires. Additionally, in healthcare, 
extrinsic regression plays a pivotal role, particularly in predicting heart 
rates using electrocardiogram (ECG) signals~\cite{heart-rate-estimattion} from patients, aiding in the 
diagnosis and management of cardiovascular conditions.

As more data becomes available and computational methods advance, 
extrinsic regression in time series analysis remains an active and 
promising field. Researchers and practitioners are constantly seeking 
new methods and applications, pushing the limits of predictive modeling 
and empirical analysis.
This impact was particularly significant when Monash University published 
the TSER archive~\cite{tser-archive}, which includes 19 different TSER 
datasets spanning applications from healthcare to energy monitoring.
Released in 2021, the TSER archive features 4 univariate and 15 multivariate 
time series datasets. Similar to the UCR/UEA TSC archives, the TSER archive 
facilitates benchmarking, enabling researchers to evaluate their contributions 
in TSER across a comprehensive set of datasets.

More recently, the authors in~\cite{bakeoff-tser} contributed $44$ new TSER 
datasets including $24$ univariate and $20$ multivariate, resulting in a total 
of $63$ TSER datasets when combined with the original archive~\cite{tser-archive}.
The authors in~\cite{bakeoff-tser} adapted as well some classification based 
algorithms in~\cite{bakeoff-tsc-2} to work with regression tasks.
The authors concluded in~\cite{bakeoff-tser} that the best regressors available 
now are feature based algorithms, especially DrCIF~\cite{hive-cote2.0} and 
FreshPRINCE~\cite{fresh-prince}.

In this section, we define the task at hand and detail briefly some alternations
done over some classification models to work with regression problems.

\mydefinition A TSER dataset $\mathcal{D}=\{\textbf{x}_i,y_i\}_{i=1}^{N}$ is a
a collection of $N$ multivariate time series of $M$ dimensions and length $L$ 
$\textbf{x}_i$ and their corresponding continuous real label $y_o$.

The task of TSER comes down to constructing a model $\mathcal{F}$ that can 
achieve correct continuous predictions as accurate as possible.
Unlike in TSC, the task is done by learning the parameters of a model 
$\mathcal{F}$ to correctly predict real values instead of categorized classes.
\begin{equation}\label{equ:tser}
    \mathcal{F}(\textbf{x}) = \hat{y}~\in~\mathds{R}
\end{equation}

\subsubsection{Distance Based Methods: $k$-NN}\label{tser-distance}

Similar to classification (Section~\ref{sec:tsc-distance}), distance based 
methods can be used for the task of TSER as well.
For instance, the $k$-NN model coupled with any similarity 
measure, where for each new test sample, the predicted label is simply the 
arithmetic mean of the labels of the $k$ nearest neighbors as follows:

\begin{equation}\label{equ:tser-knn}
    \hat{y} = \dfrac{1}{k}\sum_{i=1}^{k}y_{neighbor_{i}}
\end{equation}

\subsubsection{Convolution Based Methods: ROCKET and MultiROCKET}\label{sec:tser-convolution}

In~\cite{tser-archive}, the authors adapted the ROCKET~\cite{rocket} transformation model 
to work with TSER by simply replacing the RIDGE classification model 
by a RIDGE regression model.
This was done as well for MultiROCKET~\cite{multi-rocket}, the newest
adaptation of ROCKET for TSC (see Section~\ref{sec:tsc-convolution}), to work on TSER in the same way as ROCKET.

\subsubsection{Feature Based Methods: FreshPRINCE}\label{sec:tser-feature}

Feature based approach in the case of TSER should use an unsupervised feature 
extraction method as the label space, unlike in TSC in Section~\ref{sec:tsc-feature}, is not discrete.
The FreshPRINCE, consisting of a pipeline of TSFresh transformation~\cite{tsfresh}
followed by a Rotation Forrest~\cite{rotation-forest}, is adapted to TSER in~\cite{bakeoff-tser}
by replacing the C4.5 methods of tree generation by the 
Classification and Regression Tree (CART)~\cite{cls-and-res-tree}.
The prediction of all trees are finally averaged and produce the predicted value for new test samples.
The TSFresh phase of FreshPRINCE does not change.

\subsubsection{Interval Based Methods: DrCIF}\label{sec:tser-interval}

For TSC, interval based methods require \textbf{first} to extract phase 
independent intervals from the time series, \textbf{second} to extract features 
from each interval, \textbf{third} to train a classifier per features per interval and 
\textbf{fourth} to ensemble the classifiers trained.
In the case of TSER, the same pipeline is used, however the interval selection
must be purely unsupervised as the label space is not discrete, and the classifier 
is replaced by a regressor.
In the case of DrCIF (Section~\ref{sec:tsc-interval}) the regressors used are 
tree regressors and the ensemble is simply the average predicted value from each 
tree.

\subsubsection{Deep Learning Methods}\label{sec:tser-deep}

For TSC, the deep learning models in Section~\ref{sec:tsc-deep} are trained to predict 
a discrete probability distribution of each sample belonging to each class.
In the case of TSER, the label space is not discrete but rather continuous, for this reason
the deep learning model should predict one real value instead of a vector of $C$ values with a softmax activation.
To make this alternation, the last FC layer in all deep learning architectures is changed to have one output neuron
with no activation (linear by default) given there is no assumption of constraints over the label values.

In TSC, the cost function used to train the deep learning model's parameters is the categorical cross entropy 
(see Eq.~\ref{equ:cross-entropy}) as the predicted and ground truth values are probability distributions.
In the case of TSER, given the predicted and ground truth values are in fact real values, the cost function used 
is the Squared Error as such for sample $i~\in~[1,N]$ in the dataset:

\begin{equation}\label{equ:se-loss}
    \mathcal{L}_i(y_i,\hat{y}_i) = (y_i - \hat{y}_i)^2
\end{equation}

The total loss over a batch of $N$ samples is simply the average loss in Eq.~\ref{equ:se-loss}:
\begin{equation}\label{equ:se-loss-batch}
    \mathcal{L} = \dfrac{1}{N}\sum_{i=1}^{N}\mathcal{L}_i
\end{equation}

\section{Unsupervised Learning: Prototyping, Clustering and Self-Supervised}

Unsupervised learning techniques are crucial in time series analysis, 
especially when labeled data is scarce or unavailable. These methods 
facilitate the discovery of inherent patterns and structures within 
time series data. Key approaches include prototyping, clustering, and 
Self-Supervised Learning (SSL). Prototyping involves creating representative 
samples or profiles of time series, aiding in data summarization and 
visualization. Clustering groups similar time series together, enabling 
the identification of common behaviors and anomalies. Self-supervised 
learning, a form of representation learning in deep learning, leverages 
the intrinsic structure of the data to learn meaningful features, 
enhancing the performance of subsequent tasks such as classification 
or forecasting. These techniques collectively expand the toolkit for 
analyzing complex time series datasets, offering valuable insights 
across various applications.

In this section, we explore the three aforementioned unsupervised tasks: 
prototyping, clustering, and SSL. 
We will provide the necessary background material for understanding 
these concepts and discuss some state-of-the-art approaches 
addressing each task.

\subsection{Time Series Prototyping (TSP)}~\label{sec:ts-prototyping}

\begin{figure}
    \centering
    \caption{
        Time Series Prototyping comes down to finding a
        \protect\mycolorbox{255,0,0,0.45}{good representative} of the
        \protect\mycolorbox{0,40,255,0.5}{input set of time series}.
        This example uses the ECG5000 dataset of the UCR archive~\cite{ucr-archive}.
    }
    \label{fig:tsp}
    \includegraphics[width=\textwidth]{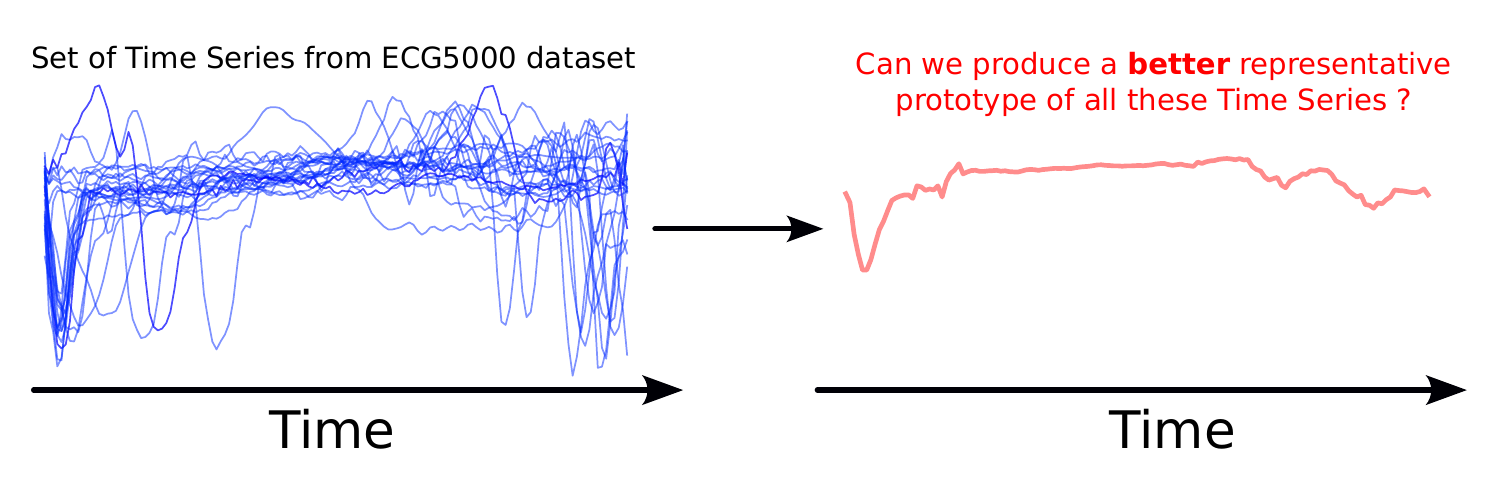}
\end{figure}

Time Series Prototyping (TSP)~\cite{eamonn-prototyping} involves creating 
representative profiles or 
prototypes of time series data, as summarized in Figure~\ref{fig:tsp}, 
which can simplify the analysis and 
interpretation of large datasets. This technique is particularly useful 
in applications such as anomaly detection, pattern recognition, and data 
summarization. For instance, in healthcare, having a representative time 
series for each disease based on ECG data can significantly speed up the 
classification of new patients. By comparing a new patient's ECG time 
series to these prototypes, medical professionals can quickly identify 
the most likely diagnosis.

The task of finding a prototype for a group of time series involves 
identifying the series that minimizes the average dissimilarity to 
the others in the group.

\mydefinition Given a group of $N$ time series $\{\textbf{x}_i\}_{i=1}^{N}$,
finding the group prototype comes down to solving the following:
\begin{equation}\label{equ:tsp-task}
    \textbf{x}^{*} = arg\min_{\textbf{x}}\dfrac{1}{N}\sum_{i=1}^{N}d(\textbf{x},\textbf{x}_i)
\end{equation}
\noindent where $d(.,.)$ represents any similarity measure between two time series.

In the rest of this section, we will present some traditional ways of 
prototyping as well as developed methods presented throughout the years.

\subsubsection{Arithmetic Mean}\label{sec:arithmetic-mean}

A naive approach to time series prototyping involves calculating the 
arithmetic mean of corresponding data points from multiple time series. 
This method, often referred to as the ``mean prototype'' does not take 
into account the temporal alignment or any variations in the time series 
but simply averages the values at each time point.

Given a group of $N$ time series $\{\textbf{x}_i\}_{i=1}^{N}$
of $M$ dimensions and length $L$, the mean prototype $\textbf{x}_p$
is calculated as follows:

\begin{equation}\label{equ:mean-prototype}
    x^m_{p,t} = \dfrac{1}{N}\sum_{i=1}^{N}x^m_{i,t}
\end{equation}
\noindent where $t~\in~[1,L]$ and $m~\in~[1,M]$.

This method is straightforward but often fails to capture important 
temporal dynamics and variations in the data, making it less effective 
for applications where the temporal order and shape of the time series 
are crucial.

\subsubsection{Piece-wise Linear Segmentation and Weighting}

The approach proposed in~\cite{eamonn-prototyping} introduces an enhanced method 
for time series prototyping that combines piece-wise linear segmentation 
with a weighting scheme to capture the importance of different segments. 
This method, referred to as ``Weighted Piece-wise Linear Segmentation'', 
involves representing each time series as a series of linear segments 
and assigning weights to these segments based on their relevance.

Prior to prototype mining, each series $\textbf{x}_i$ goes through a segmentation 
step and $S$ segments are extracted:
\begin{equation}\label{equ:segmented}
    \{(\textbf{x}_{t_{1l}},\textbf{x}_{t_{1r}}),
    (\textbf{x}_{t_{2l}},\textbf{x}_{t_{2r}}),\ldots,
    (\textbf{x}_{t_{2Sl}},\textbf{x}_{t_{Sr}})\}
\end{equation}
\noindent where $(\textbf{x}_{t_{sl}},\textbf{x}_{t_{sr}})$ denotes the start (left: l) and end (right: r) 
of a segment under the constraint: $t_{sl} < t_{sr}$.
Each segment is assigned, by the segmentation algorithm, a weight $w_s$
and $s~\in~[1,S]$.

To be able to merge two series of the group: $\textbf{x}_i$ and $\textbf{x}_j$ where
$i$ and $j~\in~[1,N]$, the following steps are taken for each segment $s~\in~[1,S]$
and each dimension $m~\in~[1,M]$,

\begin{itemize}
    \item \textbf{Step 1:} Compute the $sign$ and $mag$ (magnitude difference):
    \begin{equation}\label{equ:eamonn-prot-step1-sign}
        sign = 
        \begin{cases} 
        -1 & \text{if } w_{i,s} \cdot w_{j,s} < 0 \\
        1 & \text{otherwise}
        \end{cases}
    \end{equation}
    \begin{equation}\label{equ:eamonn-prot-step1-mag}
        mag = \dfrac{\min(|w_{i,s}|,|w_{j,s}|)}{\max(|w_{i,s}|,|w_{j,s}|)}
    \end{equation}
    \item \textbf{Step 2:} Compute the combined segment values:
    \begin{equation}\label{equ:eamonn-prot-step2-left}
        x^m_{p,t_s} = \dfrac{x^m_{i,t_{sl}}.w_{i,s} + x^m_{j,t_{sl}}.w_{j,s}}{w_{i,s}+w_{j,s}}
    \end{equation}
    \begin{equation}\label{equ:eamonn-prot-step1-right}
        x^m_{p,t_{s+1}} = \dfrac{x^m_{i,t_{sr}}.w_{i,s} + x^m_{j,t_{sr}}.w_{j,s}}{w_{i,s}+w_{j,s}}
    \end{equation}
    \item \textbf{Step 3:} Compute the weight for the segment:
    \begin{equation}\label{equ:eamonn-prot-step3-weight}
        w_{p,s} = (w_{i,s}.w_{j,s}).(1+\dfrac{sign.mag}{1+d})
    \end{equation}
    \noindent where $d$ is a scale factor calculated as follows:
    \begin{equation}\label{equ:eamonn-prot-step3-scale}
        d = |\dfrac{(x^m_{i,t_{sl}} - x^m_{j,t_{sl}}) - (x^m_{i,t_{sr}} - x^m_{j,t_{sr}})}{t_{sr} - t_{sl}}|.norm
    \end{equation}
    \noindent and $norm$ is calculated as follows:
    \begin{equation}\label{equ:eamonn-prot-step3-norm}
        norm = \max(\max(\textbf{x}^m_{i,:l}), \max(\textbf{x}^m_{i,:r})) - \min(\min(\textbf{x}^m_{i,:l}),\min(\textbf{x}^m_{i,:r}))
    \end{equation}
\end{itemize}

To reconstruct the time series from the new segments, the authors in~\cite{eamonn-prototyping}
used the piece-wise linear representation created by the merging process.

\subsubsection{Elastic Barycenter Averaging Methods}

Elastic Barycenter Averaging (EBA) is a technique first addressed by~\cite{dba-paper} that proposed a combination between 
the elastic similarity measure DTW (see Alg.~\ref{alg:dtw}) in order to find a prototype of a group of time series.
The core difference between the arithmetic Mean in Section~\ref{sec:arithmetic-mean} and the first proposed method based on
EBA: DTW Barycenter Averaging (DBA)~\cite{dba-paper} is that DBA takes into consideration the alignment information between 
all samples in the group.
This results in a prototype representing the \emph{average warping} as well as the average amplitude, on contrary with 
arithmetic Mean which considers all series are aligned.
Another powerful unique feature with EBA methods, starting with DBA in~\cite{dba-paper}, is their ability to find 
prototypes over a group of unequal length time series samples.

The detailed working of DBA is presented in Algorithm~\ref{alg:dba}.
DBA initializes a prototype by randomly choosing one series in the group, and iteratively optimizes this prototype by 
finding for each of its time stamps, the aligned time stamps with all other series in the group (referred to as associates of the 
prototype's time stamp).
The value of the time stamp of the current prototype is then replaced by the barycenter (arithmetic mean) of the aligned values with it.
Theoretically, DBA works with any value of $q$ in $DTW_q$ (see Eq.~\ref{equ:dtw}), however the authors in~\cite{neares-centroid-dba-paper}
proved that in the case of $q=2$, DBA converges and the optimal solution is using the arithmetic mean over aligned points.

\begin{algorithm}
    \caption{DTW Barycenter Averaging (DBA)}
    \label{alg:dba}
    \begin{algorithmic}[1]
        \REQUIRE Group of $N$ Time Series $\{\textbf{x}\}_{i=1}^{N}$ of length $L$ and dimension $M$ each
        \REQUIRE Initial prototype series $\textbf{x}_p$ of length $L$ and dimension $M$
        \ENSURE Time Series of length $L$ and dimension $M$: The DBA prototype representative of the group

        \STATE $CountAssociates\gets zeros(shape=(L,))$

        \FOR{$i=1~\to~N$}
            \STATE $\pi \gets DTW_{path}(\textbf{x}_p,\textbf{x}_i)$ 

            \FOR{$j=1~\to~len(\pi)$}
                \STATE $t_1\gets \pi_{j,1}$
                \STATE $t_2\gets \pi_{j,2}$
                \STATE $\textbf{x}_{p,t_1} = \textbf{x}_{p,t_1} + \textbf{x}_{i,t_2}$
                \STATE $CountAssociates_{t_1}\gets CountAssociates_{t_1} + 1$
            \ENDFOR
        \ENDFOR

        \STATE \textbf{Return:} $\textbf{x}_p / CountAssociates$

    \end{algorithmic}
\end{algorithm}

More recently, a new version of DBA has been proposed in~\cite{mba-paper} that replaces the DTW similarity measure 
by the Move-Split-Merge (MSM) measure~\cite{msm-distance}, given that it has been seen to outperform DTW for 
clustering~\cite{elastic-clustering-review}.
The proposed method, MSM Barycenter Averaging (MBA) in~\cite{mba-paper} outperformed the usage of DBA for clustering.


Elastic similarity measures such as DTW, as explained in Algorithm~\ref{alg:dtw}, uses three moves on the cost matrix,
diagonal move, horizontal move and vertical move.
In terms of DTW, the algorithm penalizes a miss alignment by simply producing longer warping path with higher number 
of non-diagonal alignments.
This can be problematic in some cases, as we need the algorithm to penalize directly the non-diagonal alignment itself 
rather than utilizing the outcome as one global penalty.
Moreover, edit distances such as MSM~\cite{msm-distance} considers that a diagonal move is a match,
and vertical/horizontal moves are considered as insertion/deletion (split/merge) and are penalized instantly when the move 
is made.
This penalty in MSM depends on a hyper-parameter called $c$ which represents the penalty minimum cost.
For a match movement (diagonal), the MSM uses the absolute difference between matched values instead of the squared error such as 
in DTW.
However, when two values are not matched (horizontal or vertical), the MSM uses a different functionality.
For instance assuming two series $\textbf{x}_1$ and $\textbf{x}_2$ for which the MSM is finding the alignment path between them,
the cost for a split (insertion/vertical) movement is calculated as such on dimension $m~\in~[1,M]$:
\begin{equation}\label{equ:msm-split}
    C(x^m_{1,t_1},x^m_{1,t_1-1},x^m_{2,t_2},c) = 
    \begin{cases}
        c  \hspace*{2cm}\textbf{if } x^m_{1,t_1-1}\leq x^m_{1,t_1} \leq x^m_{2,t_2}\\
        c  \hspace*{2cm}\textbf{if } x^m_{1,t_1-1}\geq x^m_{1,t_1} \geq x^m_{2,t_2}\\
        c + \min(|x^m_{1,t_1}-x^m_{1,t_1-1}|,|x^m_{1,t_1}-x^m_{2,t_2}|)  \textbf{  otherwise}
    \end{cases}
\end{equation}
\noindent where the above operation consists on inserting $x^m_{1,t_1}$ between $x^m_{1,t_1-1}$ and 
$x^m_{2,t_2}$.

For the deletion operation (horizontal movement), its the inverse of the insertion operation as the series swap in Eq.~\ref{equ:msm-split}
by calling $C(x^m_{1,t_2},x^m_{2,t_2-1},x^m_{1,t_1},c)$.

The MSM algorithm is detailed in Algorithm~\ref{alg:msm}.

\begin{algorithm}[t]
    \caption{Move-Split-Merge (MSM)}
    \label{alg:msm}
    \begin{algorithmic}[1]
        \REQUIRE Two Time Series $\textbf{x}_1$ and $\textbf{x}_2$ of length $L$ and dimension $M$
        \REQUIRE The cost penalty parameter $c$
        \ENSURE MSM measure between \(\textbf{x}_1\) and \(\textbf{x}_2\)
        
        \STATE $\tilde{D} \gets array [L,L] = \{0.0\}$
        \FOR{$m=1~\to~M$}
        \STATE $D \gets array [L,L] = \{0.0\}$
            \STATE $D[1,1]\gets |x^m_{1,1} - x^m_{2,1}|$
            \FOR{$t=2~\to~L$}
                \STATE $D[t,1]\gets D[t-1,1] + C(x^m_{1,t},x^m_{1,t-1},x^m_{2,1},c)$
            \ENDFOR
            \FOR{$t=2~\to~L$}
                \STATE $D[1,t]\gets D[1,t-1] + C(x^m_{2,t},x^m_{2,t-1},x^m_{1,1},c)$
            \ENDFOR

            \FOR{$t_1=2~\to~L$}
                \FOR{$t_2=2~\to~L$}
                    \STATE $move\gets D[t_1-1,t_2-2] + |x^m_{1,t_1} - x^m_{2,t_2}|$
                    \STATE $split\gets D[t_1-1,t_2] + C(x^m_{1,t_1},x^m_{1,t_1-1},x^m_{2,t_2},c)$
                    \STATE $merge\gets D[t_1,t_2-1] + C(x^m_{2,t_2},x^m_{2,t_2-1},x^m_{1,t_1},c)$

                    \STATE $D[t_1,t_2] = \min(move,split,merge)$
                \ENDFOR
            \ENDFOR
            \STATE $\tilde{D} = \tilde{D} + D$
        \ENDFOR

        \STATE \textbf{Return:} $\tilde{D}[L,L]$
        
    \end{algorithmic}
\end{algorithm}

\subsubsection{Differentiable EBA: SoftDTW Barycenter Averaging (SoftDBA)}

As mentioned in Section~\ref{sec:tsc-distance}, the authors in~\cite{soft-dtw-distance} argue the need of a differentiable
version of DTW and proposed SoftDTW.
However, this is not the only contribution of~\cite{soft-dtw-distance}, as in fact the authors proposed a novel version of
DBA~\cite{dba-paper} which does not simply replace DTW by SoftDTW, but learns the optimal barycenter using a gradient based 
optimization algorithm.
The reason that makes this possible is the fact that SoftDTW is differentiable and the solution to Eq.~\ref{equ:tsp-task}
can be found through a gradient optimization approach, where the gradient to be found is the following:
\begin{equation}\label{equ:softdtw-gradient}
    \nabla_{\textbf{x}} DTW^{\gamma}(\textbf{x},\textbf{x}_i)
\end{equation}
\noindent where $\textbf{x}_i$ is one series in the group of size $N$ to be prototyped $i~\in~[1,N]$ and $DTW^{\gamma}$ 
is the SoftDTW algorithm with smoothness parameter $\gamma$.

\subsubsection{Time Elastic Kernel Averaging (TEKA)}

\cite{marteau2019times} proposes a probabilistic approach, TEKA, to time series averaging and denoising based 
on time-elastic kernels. By interpreting kernel alignment matrices probabilistically, the method 
introduces a stochastic alignment automaton to compute the centroid of a set of time series. This 
process effectively captures both temporal dynamics and structural shape, allowing for robust 
averaging and noise reduction. Empirical evaluations across 45 datasets highlight its effectiveness, 
showing significant performance improvements for centroid-based classifiers over medoid-based counterparts. 
Moreover, the method proves valuable in reducing training set sizes for applications such as gesture 
recognition, demonstrating its utility in both denoising and efficient representation (condensing).

\subsection{Time Series CLustering (TSCL)}\label{sec:tscl}

Clustering is a fundamental technique in data analysis that involves grouping a set of objects in such a way that objects in 
the same group (or cluster) are more similar to each other than to those in other groups. This similarity is measured based 
on certain features or characteristics of the data. Clustering is widely used in various fields such as marketing for customer 
segmentation, biology for classifying species, and document clustering in text analysis.
Time Series Clustering (TSCL) specifically deals with temporal data, where the objective is to group time series that exhibit 
similar behaviors or patterns over time. Unlike traditional clustering, time series clustering must handle the unique 
characteristics of time-dependent data, such as temporal ordering, trends, and seasonality. Effective clustering of 
time series can reveal important insights in fields like finance, healthcare, and climate science.
In this section we review three common approaches in the literature of TSCL: (1) $k$-means clustering 
algorithm coupled with elastic similarity measures and EBA methods, (2) shape based algorithms and (3) deep learning methods.

\subsubsection{$k$-means with Elastic Barycenter Averaging}

$k$-means with Elastic Barycenter Averaging (EBA) is an enhanced version of the traditional $k$-means algorithm~\cite{kmeans-paper} 
designed for time series data. EBA addresses the alignment and averaging challenges of time series by allowing 
temporal distortions during the centroid calculation. This makes $k$-means more robust and accurate for clustering 
time series data, as it can handle shifts and variations in the time sequences.
A detailed view on how $k$-means work for TSCL is presented in Algorithm~\ref{alg:kmeans-eba}.
$k$-means initializes random centroids $\{\textbf{s}_j\}_{j=1}^k$ at the beginning and then updates the centroids following the set of nearest samples in the 
data $\mathcal{D} = \{\textbf{x}_i\}_{i=1}^N$.
This algorithm converges when the inertia presents a small non-significant change, compared to a threshold.
The inertia is computed as the sum of distances between each sample and its nearest centroid:
\begin{equation}\label{equ:inertia-kmeans}
    inertia = \sum_{j=1}^k~\sum_{\textbf{x}_i~\in~NN(\textbf{s}_j,\mathcal{D},d)}~d(\textbf{x}_i,\textbf{s}_j)
\end{equation}
\noindent where $d(.,.)$ is a similarity measure between two time series and $NN(\textbf{s}_j,\mathcal{D})$ is the set of time 
series in $\mathcal{D}$ that are nearest to centroid $\textbf{s}_j$ following the measure $d(.,.)$.

The authors of~\cite{dba-paper} proposed the setup of $k$-means coupled with DTW similarity measure and DBA averaging method.
Through extensive experiments on different datasets, the proposed setup for $k$-means outperformed the original setup 
which relies on using Euclidean Distance and Arithmetic Mean.

This setup was changed by~\cite{soft-dtw-distance} by using the SoftDTW similarity measure and SoftDBA averaging method 
for $k$-means and showcased how it can outperform the previous setup with DBA proposed in~\cite{dba-paper}.
The authors of \cite{soft-dtw-distance} argue that the optimization steps of SoftDBA, facilitated by the differentiability 
of SoftDTW, effectively eliminate noise in the time series samples. This process yields more accurate centroids, 
thereby enhancing clustering performance.

\begin{algorithm}[H]
\caption{$k$-means with Elastic Barycenter Averaging}
\label{alg:kmeans-eba}
\begin{algorithmic}[1]
\REQUIRE $N$ time series samples $\mathcal{D}=\{\textbf{x}_i\}_{i=1}^N$
\REQUIRE Number of clusters $k$
\REQUIRE Similarity measure $d(.,.)$ between two time series 
\REQUIRE Averaging method $A()$ between a set of time series
\REQUIRE Maximum number of iterations $max_{itr}$ in case of no convergence
\REQUIRE Threshold $\epsilon$ for inertia convergence check
\ENSURE Cluster centroids $\{\textbf{s}_j\}_{j=1}^k$ and cluster assignments
\FOR{$j=1~\to~k$}
\STATE $\textbf{s}_j~\gets~random\_choice(\mathcal{D})$
\ENDFOR
\STATE $previous\_inertia~\gets~\infty$
\FOR{$itr=1~\to~max_{itr}$}
    \STATE $previous\_inertia~\gets~current\_inertia$
    \STATE $assigned\_cluster~\gets~zeros(size=(N,))$
    \FOR{$i=1~\to~N$}
        \STATE $distances~\gets~zeros(size=(k,))$
        \FOR{$j=1~\to~k$}
            \STATE $distances[j]~\gets~d(\textbf{x}_i,\textbf{s}_j)$
        \ENDFOR
        \STATE $assigned\_cluster[i]~\gets~arg\min_{j~\in~[1,k]} distances$
        \STATE $current\_inertia~\gets~current\_inertia + \min(distances)$
    \ENDFOR
    \IF{$|current\_inertia - previous\_inertia| < \epsilon$}
        \STATE $break$
    \ENDIF
    \STATE $previous\_inertia~\gets~current\_inertia$
    \FOR{$j=1~\to~k$}
        \STATE $\textbf{s}_j~\gets~A(\{\textbf{x}_i~|~assigned\_cluster[i] = j\})$
    \ENDFOR
\ENDFOR
\STATE \textbf{Returns:} $\{\textbf{s}_j\}_{j=1}^k$, $assigned\_cluster$
\end{algorithmic}
\end{algorithm}

\subsubsection{Shape Based Method: $k$-shape}

The $k$-shape algorithm proposed in~\cite{kshape-paper} not only outperformed previous approaches with $k$-means,
but also presented a significantly faster TSCL approach.
For instance $k$-shape does not rely on using elastic similarity measures that suffer from time complexity,
instead it utilizes a Shape Based Distance (SBD).
Given two time series $\textbf{x}_1$ and $\textbf{x}_2$ of length $L$, the SBD used in~\cite{kshape-paper} relies on the
Normalized Cross Correlation (NCC) between $\textbf{x}_1$ and the shifted version of $\textbf{x}_2$.
The shifted version of $\textbf{x}_2$ of shift $s$ is computed as follows:
\begin{equation}\label{equ:kshape-shift}
    \textbf{x}_2(s) = 
    \begin{cases}
        \{\overbrace{0,\ldots,0}^{length=s},x_{2,1},x_{2,2},\ldots,x_{2,L-s}\} & if~s > 0\\
        \{x_{2,1-s},x_{2,2-s},\ldots,x_{2,L},\overbrace{0,\ldots,0}^{length=s}\} & if~s < 0
    \end{cases}
\end{equation}
And the NNC between both series is computed as follows:
\begin{equation}\label{equ:kshape-ncc}
    NCC(\textbf{x}_1,\textbf{x}_2) = \max_{s~\in~\mathcal{S}}~
    \dfrac{\sum_{t=1}^L~x_{1,t}.x_{2,t+s}}{L.module(\textbf{x}_1).module(\textbf{x}_2)}
\end{equation}
\noindent with the assumption that both series are normalized to zero mean and unit standard deviation.
The $module(.)$ operation is simply the sum of squares of all elements in the series and
$\mathcal{S}=\{-L+1,-L+2,\ldots,L-2,L-1\}$.
The final SBD is calculated as follows:
\begin{equation}\label{equ:kshape-sbd}
    SBD(\textbf{x}_1,\textbf{x}_2) = 1 - NCC(\textbf{x}_1,\textbf{x}_2)
\end{equation}
\noindent For the case of MTS data, the SBD is simply the aggregation of SBD applied on each dimensions independently.
The SBD returns as well the aligned version of $\textbf{x}_2$ (referred to as $\textbf{x}_2^{'}$) following Eq.~\ref{equ:kshape-shift} using the optimal shift
from Eq.~\ref{equ:kshape-ncc}.

The centroid finding technique used in $k$-shape over a set of time series $\mathcal{D}=\{\textbf{x}_i\}_{i=1}^N$
is the solution to the following equation:
\begin{equation}\label{equ:kshape-equation-avg}
    \textbf{s}_j^* = arg\max_{\textbf{s}_j}~
    \sum_{\textbf{x}_i~\in~NN(\textbf{s}_j,\mathcal{D},SBD)}NNC(\textbf{x}_i,\textbf{s}_j)^2
\end{equation}
\noindent where $i~\in~[1,N]$ and $j~\in~[1,k]$.

The authors in~\cite{kshape-paper} mentioned that the above optimization problem, after some alternations,
can be solved using the problem of maximization of the Rayleigh Quotient~\cite{golub2013matrix}.
One iteration of the $k$-shape centroid extraction (shape extraction as referred to in the paper) 
phase is presented in Algorithm~\ref{alg:kshape-centroid}.

\begin{algorithm}[H]
\caption{$k$-shape: Shape Extraction, One Iteration}
\label{alg:kshape-centroid}
\begin{algorithmic}[1]

\REQUIRE Set of Time Series in the same cluster $\textbf{X}=\{\textbf{x}_i\}_{i=1}^N$, assumed z-normalized
\REQUIRE The current centroid of this cluster $\textbf{s}$
\ENSURE A more accurate centroid $\textbf{s}^{'}$

\STATE $\textbf{X}^{'}$ = [~]
\STATE $\textbf{O}~\gets~[[1~for~o=1~\to~L]~for~oo=1~\to~L]$
\FOR{$i=1~\to~N$}
    \STATE $distance, \textbf{x}_i^{'}~\gets~SBD(\textbf{s},\textbf{x}_i)$
    \STATE $\textbf{X}^{'}.append(\textbf{x}_i^{'})$
\ENDFOR
\STATE $\textbf{A}~\gets~\textbf{X}^{'T}\circ \textbf{X}^{'}$
\STATE $\textbf{Q}~\gets~I_L - \dfrac{1}{L}.\textbf{O}$
\STATE $\textbf{M}~\gets~\textbf{Q}^{T}\circ \textbf{A}\circ \textbf{Q}$
\STATE $\textbf{s}^{'}~\gets EigVectors(\textbf{M})$
\STATE \textbf{Returns:} $\textbf{s}^{'}$

\end{algorithmic}
\end{algorithm}

\subsubsection{Deep Learning Methods}

While deep learning has revolutionized Time Series Classification~\cite{dl4tsc}, its potential for Time Series Clustering is still 
being explored. Recognizing the success of deep learning in classification tasks, researchers have hypothesized that 
it can also significantly enhance clustering methods. Consequently, various deep learning-based approaches for time 
series clustering have been proposed and reviewed in recent literature~\cite{deep-tscl-bakeoff}.
In the mentioned review~\cite{deep-tscl-bakeoff}, the authors not only compared deep learning models for TSCL on the 
architecture level, however they also compared different methods to train these models for better clustering downstream task.

Deep learning models, such as Recurrent Neural Networks (RNNs), Convolutional Neural Networks (CNNs), and Auto-Encoders (AEs), 
offer several advantages for clustering tasks. These models can automatically learn complex features and patterns directly 
from raw time series data, bypassing the need for manual feature extraction. RNNs, particularly Long Short-Term Memory 
(LSTM) networks, are adept at capturing long-term dependencies in sequential data, which are often overlooked by traditional 
clustering methods. CNNs, on the other hand, can efficiently process large-scale time series data in parallel, making 
them suitable for handling extensive datasets.

The adaptability and scalability of deep learning models make them promising for time series clustering. 
Their ability to model non-linear relationships and intricate temporal dependencies leads to more accurate 
and meaningful clustering results. As researchers continue to explore and refine these methods, deep learning 
is poised to offer robust solutions for clustering complex time series data, paralleling its success in 
classification tasks~\cite{dl4tsc}.

\begin{figure}
    \centering
    \caption{AE based architecture with a reconstruction loss
    for Time Series CLustering. The \protect\mycolorbox{255,0,0,0.47}{first step} is to train 
    the AE architecture to reconstruction the \protect\mycolorbox{0,40,255,0.5}{input time series}.
    The \protect\mycolorbox{189,109,226,0.47}{second step} is to generate the 
    \protect\mycolorbox{255,165,9,0.6}{latent features} and apply a clustering algorithm on top
    of these feature.}
    \label{fig:auto-encoding-clustering}
    \includegraphics[width=\textwidth]{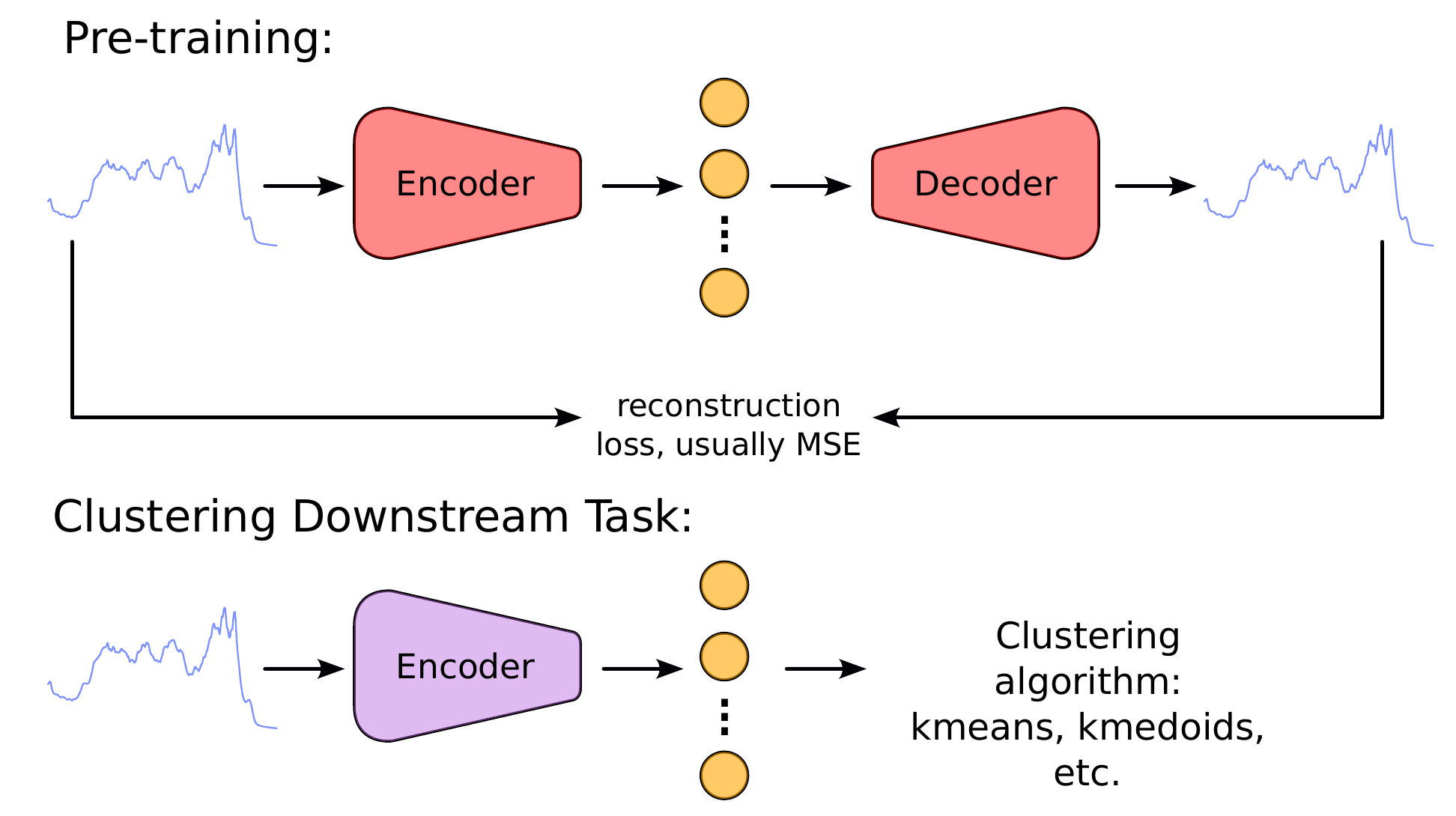}
\end{figure}

In this section, we will detail different architectural methods for deep clustering. Deep clustering can be performed 
using AEs or even standalone encoders with various pretext losses. These pretext losses include traditional 
reconstruction loss, multi-reconstruction~\cite{multi-rec-paper} loss between encoder and decoder,
triplet loss~\cite{triplet-loss-paper}, clustering losses~\cite{clustering-loss-paper}, and more.
An overview of AE architectures used for TSCL is presented in Figure~\ref{fig:auto-encoding-clustering}.
The review~\cite{deep-tscl-bakeoff} concluded that the best-performing architectures were a 
ResNet-based~\cite{fcn-resnet-mlp-paper} 
AE with multi-reconstruction loss 
and no clustering loss for univariate time series, and a non-symmetrical Dilated Recurrent Neural 
Network (DRNN)~\cite{drnn-paper} AE 
with multi-reconstruction loss and no clustering loss for multivariate time series.

\paragraph{Background on Auto-Encoders and Variational Auto-Encoders}

Auto-Encoders (AEs) are a type of artificial neural network initially 
introduced in the 1980s by~\cite{kramer1991nonlinear} for the purpose of unsupervised learning, 
particularly for tasks like dimensionality reduction and feature 
learning. An AE consists of two main parts: an encoder and 
a decoder. The encoder maps the input series $\textbf{x}$ of length $L$ and dimension $M$ into a latent space 
representation $\textbf{z}$, and the decoder reconstructs the input data from 
this latent representation.
This structure allows the AE to learn efficient codings of 
the data by minimizing the reconstruction error between the original 
input and its reconstructed output. The mathematical representation 
is as follows:
\begin{enumerate}
    \item Encoder: $\textbf{z} = E_{\theta}(\textbf{x})$
    \item Decoder: $\hat{\textbf{x}} = D_{\phi}(\textbf{z})$
\end{enumerate}
\noindent where each of the encoder $E_{\theta}(.)$ and decoder $D_{\theta}(.)$ is parametrized 
by a set of parameters for each of their layers, $\theta$ and $\phi$ respectively.

The objective is to minimize the reconstruction error, 
typically measured by the Mean Squared Error (MSE):

\begin{equation}\label{equ:ae-mse}
    \mathcal{L}_{mse}(\textbf{x},\hat{\textbf{x}}) = 
    \dfrac{1}{L.M}\sum_{t=1}^L~\sum_{m=1}^M (x^m_t - \hat{x}^m_t)^2
\end{equation}

However, traditional AEs face limitations in generating new 
data samples because they do not provide a probabilistic framework 
for the latent space.
This is where Variational Auto-Encoders (VAEs)~\cite{vae-paper} come into play. 
VAEs introduce a probabilistic approach to the latent space, 
allowing for both data reconstruction and generation of new data 
samples. By leveraging the principles of variational inference, 
VAEs model the distribution of the latent variables explicitly, 
which makes them highly effective for generative tasks.

The  (VAE) consists, like the traditional AE, of an encoder-decoder framework. However, unlike in AEs, the 
latent space in VAEs is regularized to follow a specific distribution, 
usually a Gaussian distribution. In traditional AEs, the latent space 
is not constrained to follow any distribution, which can lead to 
significant diversity in the placement of latent variables, potentially 
causing model collapse.

VAEs address this issue by incorporating a Gaussian projection step in 
the latent space. Specifically, the encoder in a VAE outputs two vectors: 
one representing the mean and the other representing the variance of a 
Gaussian distribution. This Gaussian distribution is then used to randomly 
sample points in the latent space before feeding them to the decoder. 
This sampling ensures continuity and smoothness in the latent space, 
which helps in generating coherent outputs.
To prevent the encoder from producing a high-variance Gaussian distribution, 
which could also lead to model collapse, VAEs include a regularization loss. 
This regularization loss is the Kullback-Leibler (KL) divergence between 
the learned Gaussian distribution $q_{\theta}(\textbf{z}|\textbf{x})$ 
and the standard normal distribution $\mathcal{N}(0,1)$. 
The KL divergence loss ensures that the learned distribution 
remains close to the prior distribution, maintaining the stability 
and structure of the latent space:
\begin{equation}\label{equ:vae-kl}
\begin{split}
    \mathcal{L}_{KL} &= D_{KL}(q_{\theta}(\textbf{z}|\textbf{x}),\mathcal{N}(0,1))\\
    &= -\dfrac{1}{2}~\sum_{d=1}^\textbf{d}(1+\log\sigma_d^2-\mu_d^2-\sigma_d^2)
\end{split}
\end{equation}
\noindent where $\boldsymbol{\mu}$ and $\log~\boldsymbol{\sigma}^2$ are both the mean
and $\log$ of the variance learned by the encoder part of the VAE,
and $\textbf{d}$ is the dimension of the latent space.

The total loss of the VAE used to optimize its parameters is a weighted sum of both
the reconstruction and KL loss:
\begin{equation}\label{equ:vae-total-loss}
    \mathcal{L}_{vae} = (1-\beta).\mathcal{L}_{mse} + \beta.\mathcal{L}_{KL}
\end{equation}
\noindent where $\beta$ is a hyperparameter between $0$ and $1$ used 
to control the amount of impact of the KL loss on the training phase, which was introduced 
in~\cite{beta-vae-paper}.

\paragraph{Using AEs and VAEs for TSCL}

\begin{figure}
    \centering
    \caption{AE based architecture with a multi-reconstruction loss~\cite{multi-rec-paper}
    for Time Series CLustering. The first step is to train 
    the AE architecture to reconstruction the \protect\mycolorbox{0,40,255,0.5}{input time series}
    as well as \protect\mycolorbox{255,0,0,0.47}{each layer of the AE network}.
    The second step is the same as in Figure~\ref{fig:auto-encoding-clustering}}
    \label{fig:multi-rec-auto-encoding-clustering}
    \includegraphics[width=\textwidth]{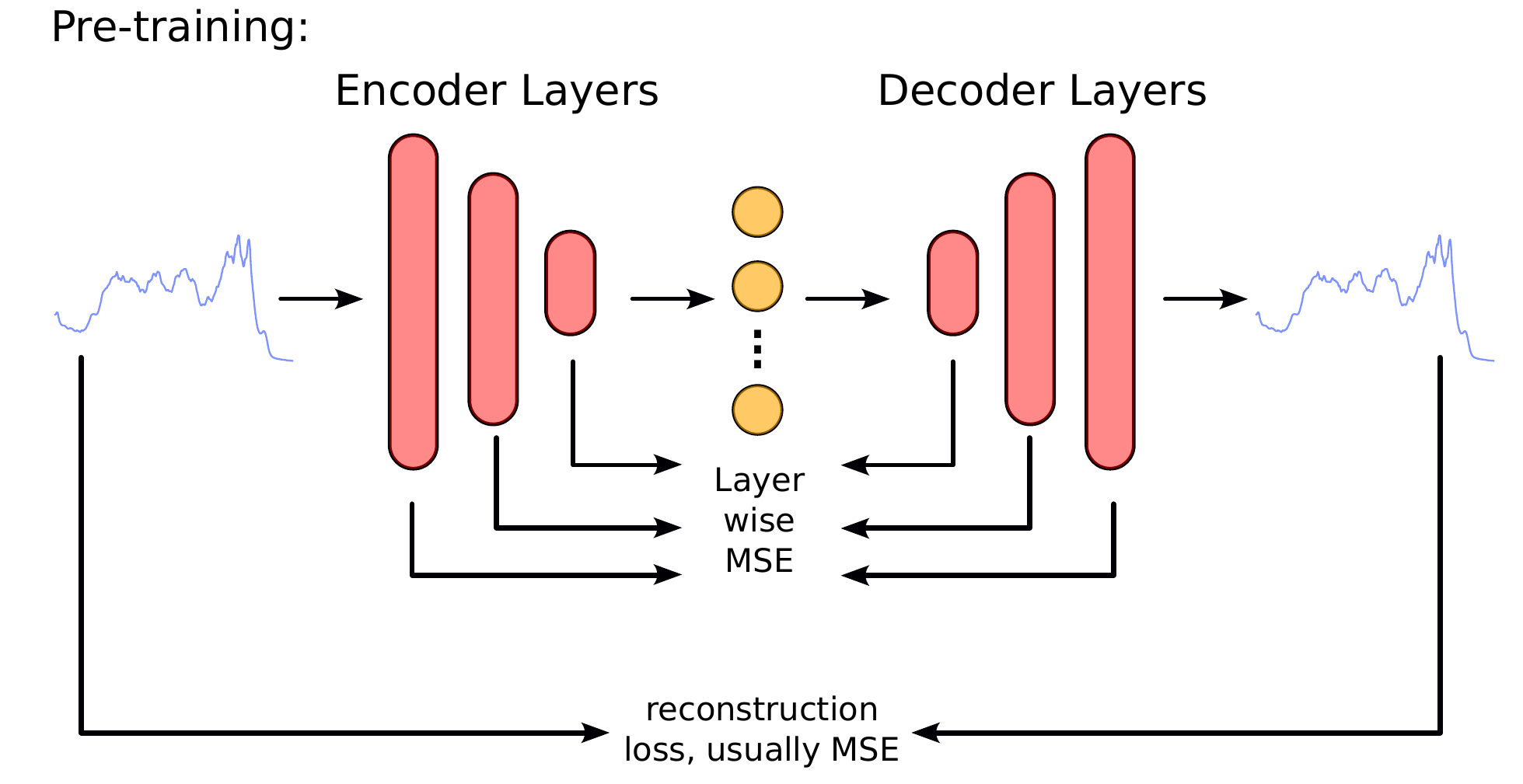}
\end{figure}

The review in~\cite{deep-tscl-bakeoff} presented multiple approaches 
from the literature that addressed the task of TSCL through deep learning.
The best winning approaches are AE and VAE based networks that uses a
multi-reconstruction loss.
The multi-reconstruction loss has the same functionality as the reconstruction 
loss in Eq.~\ref{equ:ae-mse}, however it is applied between each depth of the 
encoder with its symmetrical depth in the decoder.
This is illustrated in Figure~\ref{fig:multi-rec-auto-encoding-clustering}.

In the case of univariate datasets, the review~\cite{deep-tscl-bakeoff}
highlights that the winning model is an AE based architecture that utilizes
ResNet~\cite{fcn-resnet-mlp-paper} as a backbone network.
In order to define a ResNet decoder, a symmetrical architecture is defined that 
replaces the standard convolutions by transpose convolutions.
\mydefinition
The one dimensional transpose convolution increases the length of the input series 
instead of decreasing it.
This is done through a de-convolution step.
Given an input univariate series $\textbf{x}$ of length $L$ and a kernel $\textbf{w}$ of length $K$,
the one dimensional transpose convolution is applied as follows:
\begin{equation}\label{equ:1d-transpose-convolution}
    o_t = \sum_{t^{'}=1}^L~\sum_{k=1}^K~x_{t^{'}}.w_k.\delta(t,s.t^{'}+k)
\end{equation}
\noindent where $\textbf{o}=\{o_1,o_2,\ldots,o_{(L-1).s+k}\}$ is the output series 
of length $(L-1).s+k$ and $s$ is the number of strides.
In the case of multivariate input series and multiple kernels used, transpose convolution 
follow the same protocol as standard convolutions, see Eq.~\ref{equ:conv-layer-multi}.

Using the above transpose convolution, the ResNet based AE network can be defined.
Posterior to training the AE network, the latent space is subsequently used to train 
a simple $k$-means cluster using the arithmetic mean and ED as parameters.

The review of deep TSCL methods~\cite{deep-tscl-bakeoff} showed however that using the VAE 
regularization can degrade the clustering performance.

Moreover, for multivariate datasets, the winning approach was an RNN based 
AE using  the Dilated-RNN (DRNN) architecture~\cite{drnn-paper}.
The DRNN AE architecture consists of three bidirectional GRU layers stacked 
on top of each other in the encoder part with a single GRU layer for the decoder.
This architecture performed as the best deep clustering model for MTS data 
coupled with the reconstruction loss, unlike ResNet with the multi-reconstruction loss
in the case of UTS data.
The main reason to why the multi-reconstruction loss was not utilized in the case of 
DRNN AE network is that its a non-symmetrical AE architecture.

\subsection{Self-Supervised Learning for Time Series Analysis}

\begin{figure}
    \centering
    \caption{The difference between using deep learning for Time Series Classification,
    Extrinsic Regression and Representation Learning (Self-Supervised Learning, SSL).}
    \label{fig:deep-ssl}
    \includegraphics[width=\textwidth]{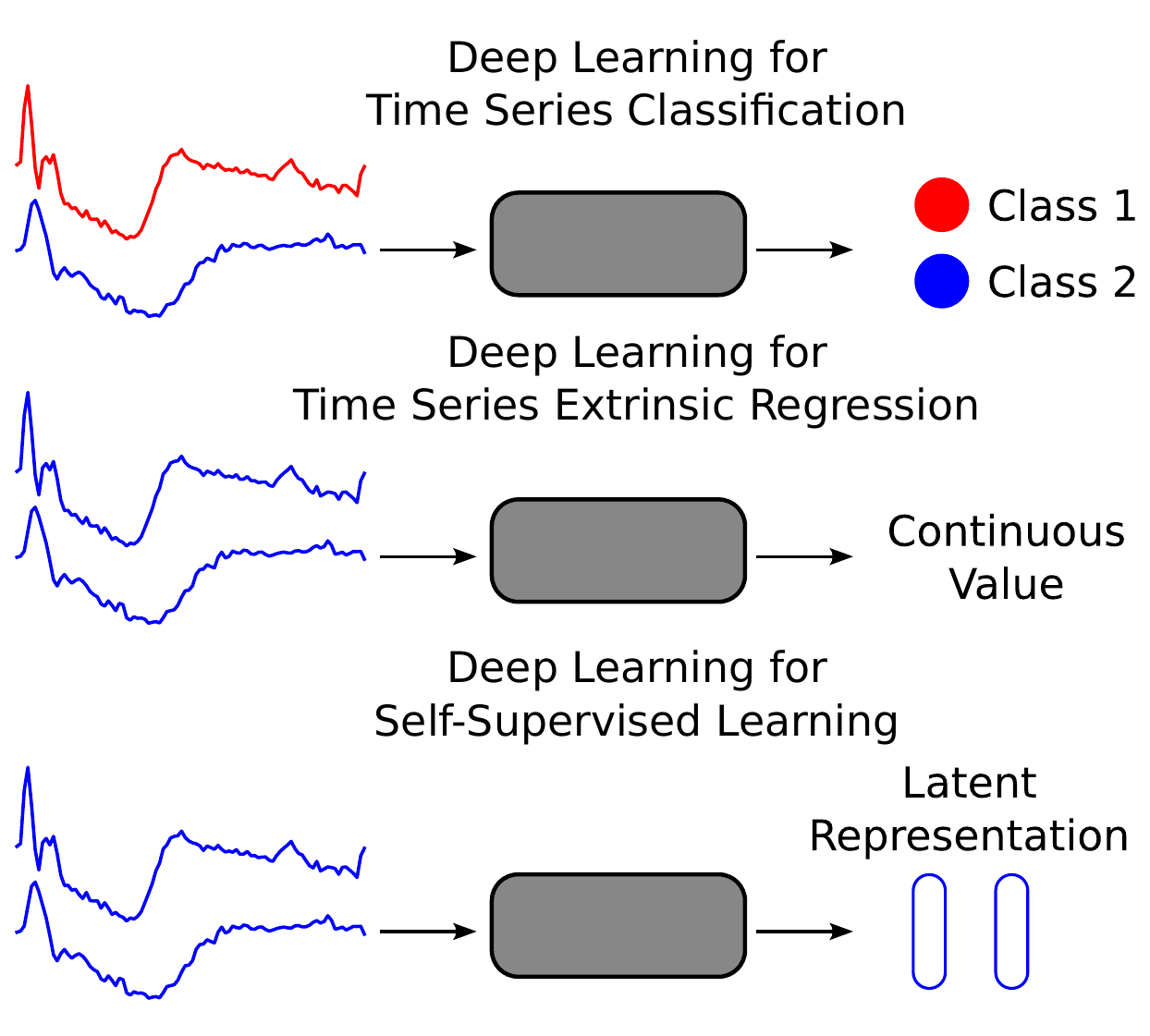}
\end{figure}

SSL, sometimes referred to as representation learning, 
is a machine learning paradigm where the model learns a compact representation 
of its input data without relying on labeled samples.
The primary goal is to learn useful representations from the input data 
that can be utilized for various downstream tasks. By learning from large 
volumes of unlabeled data, SSL enables models to 
capture intricate patterns and structures, reducing the dependency on 
extensive labeled datasets.
Figure~\ref{fig:deep-ssl} summarizes the pipeline of using a deep learning model for SSL,
and the difference with other supervised tasks.
Unlike supervised models, where deep learning models need to predict target values that are 
commonly known, deep learnign for SSL tries to find the best latent representation of the input data.

In the context of time series data, SSL can be 
particularly beneficial.
Time series data, which involves patterns and dependencies over time, can be 
difficult to label accurately and thoroughly
By employing self-supervised techniques, 
models can learn to understand these patterns and temporal structures 
by predicting future points from past points or filling in missing data 
segments etc. This pre-training process results in robust feature 
representations that can significantly enhance the performance 
of downstream tasks such as forecasting, anomaly detection, and 
classification, even when labeled data is scarce.

Pre-trained models from SSL can greatly enhance 
generalization in downstream tasks, especially in scenarios with 
limited labeled data. These models, having been trained on large 
unlabeled datasets, possess a rich understanding of the underlying 
data distribution and can transfer this knowledge to specific tasks. 
This transfer learning approach is particularly useful in semi-supervised 
setups, where only a small fraction of the data is labeled. 
By fine-tuning pre-trained models on the available labeled data, we 
can achieve superior performance compared to training models from scratch. 
Thus, SSL not only alleviates the challenge of 
data scarcity but also promotes better generalization and adaptability 
in various real-world applications.

In this section, we present briefly some starting work of self-supervised 
learning for time series analysis, mostly used for downstream classification 
tasks.

\subsubsection{Background on Siamese Neural Networks and SimCLR}

Siamese networks, introduced in the early 1990s~\cite{siamese-networks}, 
are a type of neural network architecture used to determine the 
similarity between two input samples. These networks consist of 
two identical subnetworks that share the same parameters and weights. 
They are trained using pairs of inputs, where the goal is to learn a 
function that can effectively distinguish between similar and 
dissimilar pairs. This method is particularly well-known for its 
application in signature verification and face verification tasks.

The original approach to training a Siamese network consists on defining 
for each sample a positive and negative representation of it.
For instance in the case of signature 
verification task, it would consist on the same signature written twice, once by the same anchor subject 
and the second forged by a second subject.
First, $\textbf{x}_1$ and $\textbf{x}_2$, are fed to the same 
network to obtain $\textbf{z}_1$ and $\textbf{z}_2$, two feature representations
of the two input samples.
Second, the ED is calculated between these two feature vectors.
Third, the contrastive loss is computed as follows:
\begin{equation}\label{equ:contrastive-siamese}
    \mathcal{L}_{contrastive} = \dfrac{1}{2}.(1-y).ED(\textbf{z}_1,\textbf{z}_2)^2 
    + \dfrac{1}{2}.y.\max(0, \alpha-ED(\textbf{z}_1,\textbf{z}_2)^2)
\end{equation}
\noindent where $y$ indicates if the pair of samples are an anchor with its positive 
representation or an anchor with its negative representation.
$\alpha$ is the boundary parameter that represent a penalty for the negative pairs.
The goal of the Siamese network is to learn how to increase the distance between an anchor 
sample and its negative representation and decrease the distance between the anchor and its positive 
representation.

This SSL approach was then adapted in~\cite{simCLR-paper}
to use an unlabeled setup for computing the contrastive loss.
The authors in~\cite{simCLR-paper} proposed a Simple Contrastive LeaRning 
(SimCLR) self-supervised model for image representation.
SimCLR does not rely on a labeled pair of samples such as in the original 
Siamese network.
Instead, SimCLR takes as input two augmented versions of the same sample, and never 
the original sample, and minimizes the distance between the feature representation 
of the two augmented versions.
This ensures that the latent space provides the same distance based information 
as the original space, hence can be used for downstream task e.g. simple 
linear classifier in the latent space for the classification task.

\subsubsection{Self-Supervised Learning Models for Time Series Analysis}\label{sec:self-supervised}

The above explained approach has been used as a base for almost all SSL research work 
and has been adapted to almost all domains in data science, such as time series.
In what follows, we present briefly some of these adapted work in the last five years for time series data.

\paragraph{Dilated Causal CNN (DCCNN) with Triplet Loss}

The first ever work to address SSL for time series data 
was in~\cite{triplet-loss-paper}.
The authors proposed a new architecture that they used for the self-supervised 
setup.
The proposed architecture, DCCNN, consists on multiple dilated causal convolutions 
stack on top of each other with residual networks between them.
Causal convolutions are a type of convolution that ensures predictions at any point in time only use current
and past data, never future data, which is important for time-based sequences.
However, the main contribution of the paper was not the architecture, instead it was the proposal
of a novel pretext loss, the triplet loss~\cite{facenet-paper}.
The triplet loss mechanism differs from the contrastive loss detailed in 
Eq.~\ref{equ:contrastive-siamese}, by computing the loss on both positive and negative pairs 
at the same time, without using the labeling of positive/negative pairs $y$.
The proposed loss in~\cite{triplet-loss-paper} is computed as follows:
\begin{equation}\label{equ:frans-triplet-loss}
    \mathcal{L}_{triplet} = -\log(\sigma(\textbf{f}(\textbf{x}_{ref})^T\odot\textbf{f}(\textbf{x}_{pos})))
    - \sum_{n_{neg}=1}^{N_{neg}}~\log(\sigma(-\textbf{f}(\textbf{x}_{ref})^T\odot\textbf{f}(\textbf{x}_{n_{neg}})))
\end{equation}
\noindent where $\sigma$ is the sigmoid function, $\textbf{f}$ 
is a deep model used to encode feature vectors, $\odot$ is the matrix 
multiplication operation, $^T$ is the transpose operation, $N_{neg}$
is the number of negative samples $\{\textbf{x}_{n_{neg}}\}_{n_{neg}=1}^{N_{neg}}$ per reference sample $\textbf{x}_{ref}$, and 
$\textbf{x}_{pos}$ is the positive representation of $\textbf{x}_{ref}$.

Given the loss defined in Eq.\ref{equ:frans-triplet-loss}, there should be a way 
to define the triplets $(\textbf{x}_{ref},\textbf{x}_{pos},\{\textbf{x}_{n_{neg}}\}_{n_{neg}=1}^{N_{neg}})$.
In~\cite{triplet-loss-paper}, the authors proposed the following approach to construct these triplets:
\begin{enumerate}
    \item Choose one reference series $\textbf{r}$
    \item Define $\textbf{x}_{ref}$ as a random subsequence from $\textbf{r}$
    \item Define $\textbf{x}_{pos}$ as another random subsequence from $\textbf{r}$
    \item Define $\{\textbf{x}_{n_{neg}}\}_{n_{neg}=1}^{N_{neg}}$ as being $N_{neg}$ random subsequences from another 
    series $\textbf{r}^{'}\neq\textbf{r}$
\end{enumerate}

\paragraph{Mixup Contrastive Learning (MCL)}

Instead of relying on subsequences, the authors in~\cite{mixing-up-paper}
proposed to use a novel approach to learn a compact representation of time 
series data.
This approach, Mixup Contrastive Learning (MCL), does not rely on the concept 
of negative and positive representations
directly.
Instead, given two input series $\textbf{x}_1$ and $\textbf{x}_2$,
the proposed approach defines a weighted average $\bar{\textbf{x}}$ of these two series as follows:
\begin{equation}\label{equ:mixing-up-augment}
    \bar{\textbf{x}} = \lambda.\textbf{x}_1 + (1-\lambda).\textbf{x}_2
\end{equation}
\noindent where $\lambda$ is a real value between $0$ and $1$, representing
the amount of mixing up from each of the two series.
This parameter follows a beta distribution.

The self-supervised setup in Mixing Up~\cite{mixing-up-paper} tries to predict 
the amount of contribution from each of the two input series.
Assuming an input batch of $N$ series, this batch is then randomly shuffled to two new batches 
$\{\textbf{x}_1^{(1)},\textbf{x}_2^{(1)},\ldots,\textbf{x}_N^{(1)}\}$ and $\{\textbf{x}_1^{(2)},\textbf{x}_2^{(2)},\ldots,\textbf{x}_N^{(2)}\}$.
These two new batches now produces a set of weighted averages:
$\{\bar{\textbf{x}}_1,\bar{\textbf{x}}_2,\ldots,\bar{\textbf{x}}_N\}$.
The contrastive loss is then computed as follows for each sample in the batch:

\begin{equation}\label{equ:mixing-up-loss}
\begin{split}
    l_i &= -\lambda.\log\dfrac{exp(D_c(\textbf{f}(\bar{\textbf{x}}_i),\textbf{f}(\textbf{x}_i^{(1)}))/\tau)}{\sum_{j=1}^N(exp(D_c(\textbf{f}(\bar{\textbf{x}}_i),\textbf{f}(\textbf{x}_j^{(1)}))/\tau) + exp(D_c(\textbf{f}(\bar{\textbf{x}}_i),\textbf{f}(\textbf{x}_j^{(2)}))/\tau))}\\
    &- (1-\lambda).\log\dfrac{exp(D_c(\textbf{f}(\bar{\textbf{x}}_i),\textbf{f}(\textbf{x}_i^{(2)}))/\tau)}{\sum_{j=1}^N(exp(D_c(\textbf{f}(\bar{\textbf{x}}_i),\textbf{f}(\textbf{x}_j^{(1)}))/\tau) + exp(D_c(\textbf{f}(\bar{\textbf{x}}_i),\textbf{f}(\textbf{x}_j^{(2)}))/\tau))}
\end{split}
\end{equation}
\noindent where $\textbf{f}$ is the deep feature extractor, 
$\tau$ is the smoothness temperature parameter and $D_c(.,.)$ refers to the 
cosine similarity function.
The final loss over all the batch of samples is the average loss over each 
sample as follows:
\begin{equation}\label{equ:mixing-up-total-loss}
    \mathcal{L}_{MixingUp} = \dfrac{1}{N}~\sum_{i=1}^{N}~l_i
\end{equation}

The core idea of the above loss proposed in~\cite{mixing-up-paper} is to 
somehow predict the amount of information each series contributed into the
new weighted average series, however in the latent space instead of the original
one.
The backbone architecture used in the Mixing Up model is the FCN~\cite{fcn-resnet-mlp-paper}.

\paragraph{Time Series Self-Supervised
Contrastive Learning framework for Representation (TimeCLR)}

Instead of basing the triplet generation on subsequences or weighted averages,
TimeCLR~\cite{time-clr-paper} defines, prior to training a self-supervised 
model, an AE network to be used for the generation of new samples.
This AE network is trained to approximate the DTW distance between two raw 
series, by the ED between their latent representation extracted by the AE.
The AE takes as input two series, and is trained to reconstruct both of them, such as 
in Eq.~\ref{equ:ae-mse} using the MSE loss.
However, the MSE loss is also computed between the DTW measure between those 
two series and the ED between their latent representation, as follows:
\begin{equation}\label{equ:time-clr-dtw}
    \mathcal{L}_{distance} = (DTW(\textbf{x}_1,\textbf{x}_2) - ED(\textbf{z}_1,\textbf{z}_2))^2
\end{equation}
\noindent where $\textbf{x}_1$ and $\textbf{x}_2$ are the two input series 
with their latent representations (encoder's output of the AE) $\textbf{z}_1$
and $\textbf{z}_2$ respectively.

This AE is then used to perform some augmentation of the input series.
For instance, posterior to training the AE, each input series can now be 
transformed to a new series by simply extracting the latent feature vector from 
the encoder part of the AE, and feeding the decoder a noisy version of this 
vector.
The output of the decoder is now an augmented version of the original input series.
The above pipeline over one input series $\textbf{x}$ is summarized as follows:
\begin{equation}\label{equ:time-clr-augment}
    \hat{\textbf{x}} = D(E(\textbf{x}) + \mathcal{N}(0,1))
\end{equation}
\noindent where $E(.)$ and $D(.)$ are both the encoder and decoder of the pre-trianed 
AE network.

This augmentation method is used to generated two views of each series in 
the dataset, for which these two views are fed to a deep learning model, 
with Inception~\cite{inceptiontime-paper} as a backbone, to be used in a 
contrastive learning setup.
For instance, each series $\textbf{x}_i$ in a batch of $N$ samples goes through the above augmentation method two times
to obtain $\hat{\textbf{x}}_i^{(1)}$ and $\hat{\textbf{x}}_i^{(2)}$.
These two augmented series are fed to the Inception network, with no final task 
layer, and the model's parameters are optimized using the following contrastive loss on each sample in the batch:
\begin{equation}\label{equ:time-clr-contrastive}
    l_i = -\log\dfrac{exp(D_c(\textbf{f}(\hat{\textbf{x}}_i^{(1)}),\textbf{f}(\hat{\textbf{x}}_i^{(2)}))/\tau)}{
        \sum_{j\neq i}(exp(D_c(\textbf{f}(\hat{\textbf{x}}_i^{(1)}),\textbf{f}(\hat{\textbf{x}}_j))/\tau) + exp(D_c(\textbf{f}(\hat{\textbf{x}}_i^{(2)}),\textbf{f}(\hat{\textbf{x}}_j))/\tau))
    }
\end{equation}
\noindent where $D_c(.,.)$ is the cosine similarity, 
$\textbf{f}$ is the deep learning feature extractor with the Inception architecture, and $\tau$ is the smoothness temporature parameter.
The final loss over the whole batch is the average loss over all $N$ samples:
\begin{equation}\label{equ:time-clr-total-loss}
    \mathcal{L}_{TimeCLR} = \dfrac{1}{N}~\sum_{i=1}^{N}~l_i
\end{equation}

The TimeCLR model learns how to represent two series that are an augmentation 
of the same original series, as much as close in the feature space,
hence learning a compact representation space.

\section{Conclusion}

This Chapter has provided an in-depth exploration of the state-of-the-art 
methodologies in time series analysis, a crucial field in data science 
that focuses on extracting meaningful insights from time-dependent data. 
This chapter has navigated through both supervised and unsupervised 
learning techniques, each offering unique advantages and applications.

In the domain of supervised learning, we examined two primary tasks: 
Time Series Classification and Extrinsic Regression. Time Series 
Classification has seen a range of approaches over the years, 
from traditional distance-based methods like k-Nearest Neighbor 
with Dynamic Time Warping (k-NN-DTW) to modern deep learning architectures. 
Distance-based methods, while foundational, have evolved with innovations 
like SoftDTW and ShapeDTW, enhancing their ability to handle temporal 
distortions. Feature-based methods such as Catch22 and TSFresh have 
streamlined the process of extracting significant characteristics from 
time series data, facilitating their use in various classifiers. 
Interval-based methods, dictionary-based methods like BOSS and WEASEL, 
and convolution-based methods such as ROCKET and its variants have 
all contributed to the growing arsenal of tools for Time Series 
Classification, each addressing different aspects of the problem. 
Notably, hybrid models like HIVE-COTE have demonstrated the power 
of combining multiple approaches to achieve superior performance.

Deep learning methods have emerged as a dominant force in Time Series 
Classification, leveraging the parallelization capabilities of GPUs 
and the comprehensive feature extraction capabilities of architectures 
like Convolutional Neural Networks (CNNs), Recurrent Neural Networks 
(RNNs), and Self-Attention. These models have shown remarkable performance 
improvements, driven by their ability to learn complex temporal patterns 
directly from the data.

For unsupervised learning, the chapter covered essential tasks 
such as clustering, prototyping, and SSL. 
Clustering methods group similar time series, aiding in tasks like 
customer segmentation and anomaly detection. Prototyping techniques 
create representative examples of time series, simplifying the analysis 
of large datasets. SSL methods, which leverage 
the data itself to generate supervisory signals, have opened new avenues 
for extracting valuable insights without the need for labeled data.


In summary, this chapter has laid a comprehensive foundation for understanding 
the current landscape of time series analysis. This foundation sets the 
stage for the subsequent chapters, which will explore deeper into specific 
methodologies, their applications, and the nuances of implementing these 
advanced techniques in real-world scenarios. In the next chapter, we will 
explore the methods for comparing these models against each other by 
utilizing a wide range of different datasets to benchmark their performance.
%
%
\chapter{Benchmarking Machine Learning Models on Time Series Data}
\label{chapitre_2}

\section{Introduction}~\label{sec:chap-mcm-intro}

Benchmarking machine learning models is a critical practice in the field 
of computer science and machine learning. This process involves comparing 
the performance of various algorithms across multiple datasets to determine 
their relative effectiveness and identify the state-of-the-art methods. 
Effective benchmarking is essential for understanding the strengths and 
weaknesses of different models, guiding future research, and improving 
algorithm design.
In this chapter, we will explore the methodologies for benchmarking 
machine learning models on time series data, with a focus on 
the classification task.

A crucial aspect of benchmarking involves hypothesis testing 
and the use of post-hoc tests for p-values and Null Hypothesis 
Significance Testing (NHST). Traditional methods like the Wilcoxon 
signed-rank test~\cite{wilcoxon-paper} and the Nemenyi test~\cite{nemenyi-paper}
are commonly used but have 
significant limitations~\cite{pvalues-pitfal-paper},
such as being prone to manipulation and 
providing limited insight into the true differences between model.
Recent studies advocate for Bayesian methods~\cite{bayseian-benavoli-paper}
as more reliable alternatives for multiple comparisons.

We explore in this chapter the current method used for benchmarking, 
such as the Critical Difference Diagram (CDD) introduced by~\cite{demsar-cdd-paper}, 
and discuss its evolution~\cite{demsar-extension-paper,cdd-benavoli-paper}
and limitations. These traditional methods often suffer 
from issues such as instability in ranking and susceptibility to 
manipulation. To address these concerns, we will introduce the Multiple 
Comparison Matrix (MCM), a novel approach designed to provide more robust 
and interpretable comparisons.

\section{Background and Current Benchmarking Methods}\label{sec:background-benchmark}

Following our exploration of the limitations inherent in traditional benchmarking methods, 
this section explores the current approaches used to evaluate classifiers.
We address the challenge of summarizing the evaluation outcomes of $m$ 
comparates $\mathcal{C} = \{c_1, \ldots, c_m\}$ across multiple datasets 
$\mathcal{D} = \{d_1, \ldots, d_n\}$ using a single performance measure 
$\gamma: \mathcal{C} \times \mathcal{D} \rightarrow \mathbb{R}$, that 
assesses the performance of a comparate $c \in \mathcal{C}$ on a task 
$d \in \mathcal{D}$. Each task (dataset) involves training a classifier $\lambda$ on a 
time series training set, with performance measured by the accuracy 
of $\lambda$ on a corresponding time series test set. 
In this context, the comparates $\mathcal{C}$ are the time series 
classifiers, the tasks $\mathcal{T}$ are the classification problems 
derived from 128 datasets in the UCR archive~\citep{ucr-archive}, and 
the performance measure $\gamma$ is the classification accuracy.

\subsection{Ranking Comparates}\label{sec:average-rank}

\begin{figure}
    \centering
    \caption{Critical Difference Diagram between five state-of-the-art models of 
    TSC evaluated on 128 datasets 
    of the UCR archive highlighting the \protect\mycolorbox{0,0,255,0.7}{\textcolor{white}{average rank}}
    of each model over all datasets. If a \protect\mycolorbox{255,0,0,0.7}{\textcolor{white}{clique}} 
    is formed between two models it means that no conclusion can be made 
    on the statistical significance in difference of performance between this pair of 
    models.}
    \label{fig:cdd-example}
    \includegraphics[width=\textwidth]{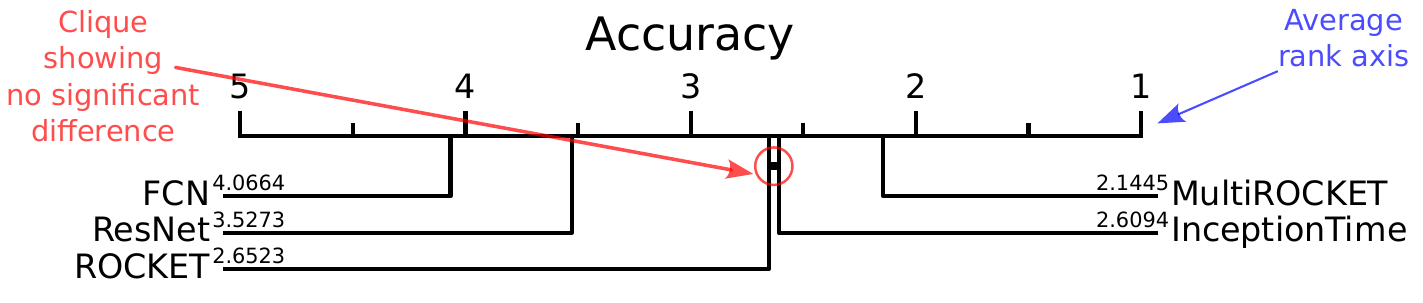}
\end{figure}

A common approach to summarize such evaluation is through ranking each of the comparate independently for each dataset
for which we can produce an average rank per comparate overall datasets.
The Critical Difference Diagram (CDD)~\citep{demsar-cdd-paper} is currently the primary method 
used for multi-comparate and multi-dataset benchmarking. This diagram provides 
two types of comparisons: (1) a group-wise comparison using the mean rank 
of each comparate, and (2) a pairwise comparison indicating which pairs 
of comparates show significant performance differences. An example is 
shown in Figure~\ref{fig:cdd-example} where each comparate is assigned an average rank overall datasets
and if a clique (straight line) exists amongst a set of comparates it highlights that no conclusion can be found 
on the statistical significance in difference of performance between these comparates.

The mean rank is computed as follows. Each comparate $c_i$ is assigned a 
rank $R_{c_i}^{d_k}$ on each task $d_k$ with $k~\in~[1,n]$ based on its relative performance score 
$\gamma$:
\begin{equation}
    R_{c_i}^{d_k} = 1 + \left|\{c_j \in \mathcal{C} \setminus c_i : \gamma(c_j, d_k) \succ \gamma(c_i, d_k)\}\right| + \tfrac{1}{2} \cdot \left|\{c_j \in \mathcal{C} \setminus c_i : \gamma(c_i, d_k) = \gamma(c_j, d_k)\}\right|,
\end{equation}

\noindent
where $\succ$ means \emph{better than}. For example, if $\gamma$ represents 
the accuracy on the test set, then a higher accuracy means a \emph{better}
model, however if $\gamma$ represents the error rate, then a lower error 
rate means a \emph{better} model.
Each comparate is then assigned an 
Average Rank (AR) by averaging its ranks over all $n$ datasets in 
$\mathcal{D}$,
\begin{equation}
    \mathrm{AR}^{\mathcal{D}}_{c_i} = \frac{\sum_{k=1}^n R_{c_i}^{d_k}}{n}.
\end{equation}

The lower the AR value, the better the performance assessment relative to competing comparators.
The placement of the comparates in the diagram in Figure~\ref{fig:cdd-example}
follows their AR.

\subsection{Pairwise Comparisons with the CDD}

Originally, the CDD proposed 
in~\cite{demsar-cdd-paper} also emphasizes the significance of 
performance differences between each pair of comparates. 
This significance test is crucial for determining whether it is 
necessary to adopt the \emph{better} model with the lowest AR (AR$_1$).
However, there may be another model with an AR 
(AR$_2$) very close to AR$_1$ (i.e., AR$_2 = $ AR$_1 + \epsilon$, 
where $0 < \epsilon << 1$) that is significantly faster and less 
complex than the \emph{winning} model (see example between InceptionTime and ROCKET in Figure~\ref{fig:cdd-example}).

The method to assess this statistical significance in difference of performance 
has been changed throughout the years.
For instance, the original CDD utilized the Nemenyi test~\cite{nemenyi-paper},
based on the actual values of the AR.
However,~\cite{cdd-benavoli-paper} argued that using a post-hoc test following
the AR values may be miss-leading and proposed the usage of the Wilcoxon Signed 
Rank Test for pairwise significance comparison and the Friedman test~\cite{friedman-paper}
for group-wise significance comparison.

\subsubsection{Friedman Test}\label{sec:friedman}

The Friedman test~\cite{friedman-paper} is a non-parametric statistical test used to detect differences in 
treatments across multiple test attempts. It is particularly useful for comparing multiple 
algorithms over multiple datasets. Given $m$ algorithms and $n$ datasets, this test ranks the performance of algorithms for each 
dataset and then analyzes these ranks to determine if there are statistically significant 
differences between the algorithms.

Let $R_{c_i}^{d_k}$ be the rank of the $i_{th}$ comparate $c_i$ on the $k_{th}$ dataset $d_k$.
The test's objective is to determine if at least one of the $m$ comparates performs 
significantly better than all other comparates.
This is determined by following these steps below:

\begin{enumerate}
    \item Compute the rank sums for each comparate:
    \begin{equation}\label{equ:friedman-step1}
        R_{c_i} = \sum_{k=1}^{n}R_{c_i}^{d_k}
    \end{equation}
    \item Calculate the Friedman test statistic:
    \begin{equation}\label{equ:friedman-step2}
        \chi_{F}^2 = \dfrac{12}{n.m.(m+1)}\sum_{i=1}^{m}R_{c_i}^2~ - 3n.(m+1)
    \end{equation}
    \item Determine the p-value:
    The test statistic $\chi_{F}^2$ approximately follows a chi-squared distribution with 
    $m.(m+1)$ degrees of freedom. The p-value is computed based on this distribution. 
    A low p-value (typically less than $\alpha=0.05$) indicates that at least one of the algorithms 
    performs significantly differently from the others.
\end{enumerate}

While the Friedman test is useful for identifying differences between multiple algorithms,
it has several limitations:

\begin{itemize}
    \item \textbf{Magnitude Ignored}: The test only considers the ranks of the algorithms,
    not the magnitude of the differences in performance. As a result, small and large
    differences are treated equally.
    \item \textbf{Instability}: The results of the Friedman test can be sensitive to the set of 
    comparates included in the analysis. Adding or removing an algorithm can change the 
    conclusions about the relative performance of the remaining algorithms.
    \item \textbf{Post-hoc Analysis Needed}: To determine which specific algorithms differ from each
    other, a post-hoc test (such as the Nemenyi test) is required, adding complexity to 
    the analysis.
\end{itemize}

\subsubsection{Nemenyi Test}\label{sec:nemenyi}

The Nemenyi test~\cite{nemenyi-paper} is a post-hoc statistical test used to determine whether 
the performance differences between pairs of algorithms are statistically 
significant. The Nemenyi test is typically applied after conducting a Friedman 
test, which assesses whether there are any overall differences among 
multiple algorithms across multiple datasets.
The Nemenyi test compares the mean 
ranks of all pairs of algorithms and determines if the differences 
in their ranks exceed a critical value, which would indicate a 
statistically significant difference in performance.
Mathematically, the Critical Difference (CD) for the Nemenyi 
test is calculated as follows:

\begin{equation}
    CD = q_{\alpha} \sqrt{\frac{m(m+1)}{6n}}
\end{equation}
\noindent where $q_{\alpha}$ is the critical value from the Studentized range distribution, 
which depends on the desired significance level $\alpha$, usually set to 
$0.05$ and the number of comparates $m$.

Two algorithms $c_i$ and $c_j$ are considered to have a statistically 
significant difference in performance if the absolute difference 
in their ARs exceeds the critical difference:

\begin{equation}
|AR_{c_i}^{\mathcal{D}} - AR_{c_j}^{\mathcal{D}}| > CD
\end{equation}
\noindent where $AR_{c_i}^{\mathcal{D}}$ and $AR_{c_j}^{\mathcal{D}}$
are the ARs of both comparates $c_i$ and $c_j$ respectively on all datasets $\mathcal{D}$.
If this condition is met, it can be concluded that the performance 
of the two algorithms differs significantly at the given significance level 
$\alpha$. However, if the condition is not met, it does not imply 
that the two comparates are equivalent in performance. Rather, it 
suggests that, given the set of $n$ datasets $\mathcal{D}$, there is insufficient 
evidence to conclude a statistically significant difference in 
performance between the two comparates.

However, this test has several limitations~\cite{cdd-benavoli-paper}:
\begin{itemize}
    \item \textbf{Rank-Only Consideration}: It only considers the ranks of the comparates, 
    focusing solely on the number of tasks where one comparate performs better or worse than others.
    \item \textbf{Ignoring Magnitude Differences}: The test does not take into account the magnitude of performance 
    differences, treating small and large differences equally.
    \item \textbf{Results Instability}: The test's results are unstable with respect to 
    the set of comparates included in the evaluation.
    \item \textbf{Sensitivity to Comparate Changes}: The inclusion or exclusion of a single comparate can significantly 
    alter the pairwise conclusions drawn for the remaining comparates.
\end{itemize}

\subsubsection{Wilcoxon Signed-Rank Test}\label{sec:wilcoxon}

The Wilcoxon signed-rank test is a non-parametric statistical test used to compare two related samples, 
matched samples, or repeated measurements on a single sample. It is the non-parametric alternative to the 
Nemenyi test (Section~\ref{sec:nemenyi}), where~\cite{cdd-benavoli-paper} questions the following:
\emph{``Should we really use post-hoc tests based on mean-ranks?''}. This test was proposed in~\cite{wilcoxon-paper} 
as a means to test for differences in the median values of two related groups without assuming that the
differences follow a normal distribution.

Given $n$ datasets, the Wilcoxon signed-rank test can be used to compare the performance
of two specific comparates, $c_i$ and $c_j$ evaluated across these $n$ datasets. The steps are as follows:

\begin{enumerate}
    \item Compute the differences between the performance of the two comparates for each dataset:
    \begin{equation}\label{equ:wilcoxon-step1}
        D_k = \gamma(c_i,d_k) - \gamma(c_j,d_k) \hspace{1cm} for~k~\in~[1,n]
    \end{equation}
    \noindent where $\gamma(c_i,d_k)$ and $\gamma(c_j,d_k)$ are the performance of both 
    comparates $c_i$ and $c_j$ evaluated on the test set of 
    the $k_{th}$ dataset $d_k~\in~\mathcal{D}$.
    \item Rank the absolute values of the differences, ignoring the signs, and assign ranks $RD_k$.
    If there are ties, assign the average rank to both values.
    \item Restore the signs to the ranks, resulting in signed ranks $sRD_i$
    \item Calculate the test statistic $W^{+}$, which is the sum of the positive ranks:
    \begin{equation}\label{equ:wilcoxon-step2}
        W^{+} = \sum_{k:~sRD_k > 0}~|sRD_k|
    \end{equation}
    \item Calculate the test statistic $W^{-}$, which is the sum of the negative ranks:
    \begin{equation}\label{equ:wilcoxon-step3}
        W^{-} = \sum_{k:~sRD_k < 0}~sRD_k
    \end{equation}
    \item Use the smaller of the two sums, $W=\min(W^{+},W^{-})$ as the test statistic
    \item Determine the p-value:
    The p-value is calculated based on the distribution of $W$. For large sample sizes (typically $n > 30$),
    the distribution of $W$ approaches a normal distribution, and a z-score can be used. For smaller samples,
    exact tables or software can be used to find the p-value.
    If this p-value is lower than a threshold $\alpha$ (usually set to $0.05$), then the difference of performance between 
    both comparates $c_i$ and $c_j$ on the $n$ datasets is statistically significant.
    However if the p-value is higher than $\alpha$, then the $n$ datasets are not enough to find a conclusion
    on the statistical significance in difference of performance between $c_i$ and $c_j$.
\end{enumerate}

While the Wilcoxon signed-rank test is robust and widely applicable, it has several limitations:
\begin{itemize}
    \item \textbf{Sensitivity to Outliers}: Like many non-parametric tests, the Wilcoxon signed-rank test can be sensitive
    to outliers, which can disproportionately influence the results.
    \item \textbf{Dependent Observations}: The test assumes that the pairs of observations are independent of each other.
    If there is dependence, the results may not be valid.
    Such dependency exists when two comparates are the same approach but one being a weaker version than the other.
\end{itemize}

\paragraph*{Holm Correction for Multiple Pairwise Comparison}
When we need to find the p-values for multiple paired comparates among $m$ comparates, the Wilcoxon
signed-rank test is often used in conjunction with the Holm correction~\cite{holm-correction-paper}.
The Holm correction is a method for controlling the family-wise error rate when performing multiple comparisons.
It adjusts the significance levels for each hypothesis test to account for the multiple comparisons being made.

Given $m$ comparates and the significance threshold $\alpha$, we have $\hat{m}=\dfrac{m.(m-1)}{2}$ p-values between all possible pairs of comparates:
$p_1,p_2,\ldots,p_{\hat{m}}$.
The Holm correction is applied as follows:
\begin{enumerate}
    \item Order the p-values from smallest to largest
    \item Compute the adjusted significance level for each ordered p-value $p_v$ as such:
    \begin{equation}\label{equ:holm-adjust}
        \alpha_v = \dfrac{\alpha}{\hat{m}-v+1}
    \end{equation}
    \item Compare each ordered p-value $p_v$ to its corresponding adjusted significance level $\alpha_v$:
    \begin{itemize}
        \item Reject the null hypothesis for $p_v$ if $p_v\leq\alpha_v$
        \item Stop testing as soon as you fail to reject a null hypothesis (i.e. when $p_v>\alpha_v$)
    \end{itemize}
\end{enumerate}

\section{Limitations of the CDD}~\label{sec:limitations-cdd}

A significant benefit of the CDD is its ability to distill a large volume of information into a format that
is easy to understand. However, this simplification introduces several shortcomings. We will discuss three key 
issues with the CDD: (1) the inconsistency of the average rank, (2) the failure to adequately account 
for the magnitude of performance differences and (3) the adverse effects of applying multiple testing corrections.

\subsection{Instability of the Average Rank}~\label{sec:instability-average-rank}

\begin{figure}
    \centering
    \includegraphics[width=0.85\linewidth]{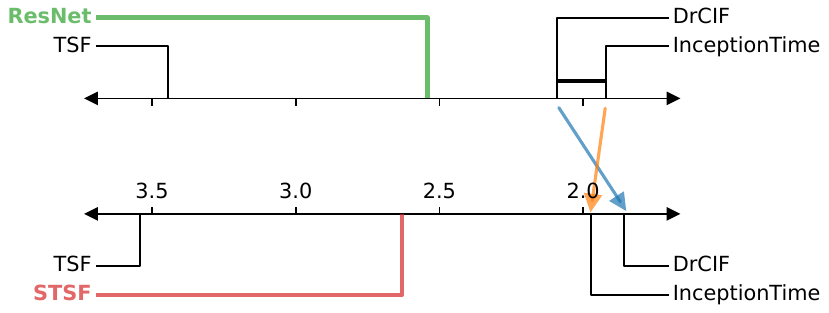}
    \caption{Manipulation of the ranks of DrCIF and InceptionTime and the statistical 
    significance of their 
    pairwise differences by inclusion of similar comparates. When \protect\mycolorbox{44,160,44,0.7}{ResNet} is 
    replaced by \protect\mycolorbox{214,39,40,0.7}{STSF}, DrCIF moves from 
    a ``\protect\mycolorbox{31,119,180,0.7}{worse}'' to a ``\protect\mycolorbox{255,127,14,0.7}{better}''
    rank, and the pairwise differences between DrCIF and InceptionTime change from being 
    not statistically significant to statistically significant.}
    \label{fig:cdd-swap-1}
\end{figure}

\begin{figure}
    \centering
    \includegraphics[width=0.85\linewidth]{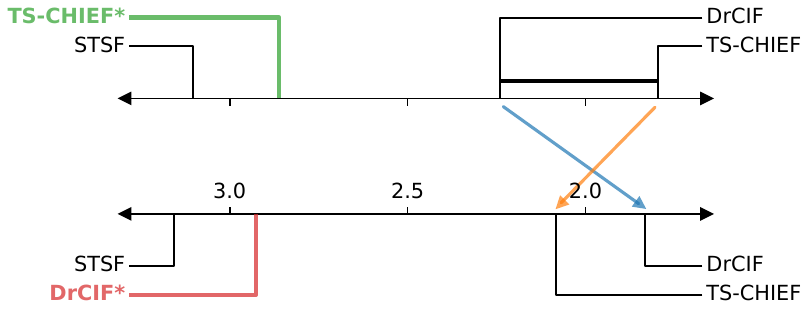}
    \caption{Manipulation of the ranks of DrCIF and TS-CHIEF and the 
    statistical significance of their pairwise 
    differences by inclusion of weakened comparates. When 
    a weakened variant of TS-CHIEF (\protect\mycolorbox{44,160,44,0.7}{TS-CHIEF*})
    is replaced by a weakened 
    variant of DrCIF (\protect\mycolorbox{214,39,40,0.7}{DrCIF*}), 
    DrCIF moves from a 
    ``\protect\mycolorbox{31,119,180,0.7}{worse}'' to a ``\protect\mycolorbox{255,127,14,0.7}{better}'' 
    rank and the pairwise differences between DrCIF 
    and TS-CHIEF change from being not statistically 
    significant to statistically significant.}
    \label{fig:cdd-swap-2}
\end{figure}

The CDD arranges comparates based on their average rank (see Section~\ref{sec:average-rank}). However, this average 
rank can vary with the addition or removal of comparates. The relative ranking of a group of comparates
$\mathcal{C}$, can 
shift when one or more comparates are added or removed, as illustrated in Figures~\ref{fig:cdd-swap-1} and~\ref{fig:cdd-swap-2}. 

In particular, Figure~\ref{fig:cdd-swap-1} demonstrates that by replacing ResNet with STSF in the set of comparates, 
DrCIF changes its rank relative to InceptionTime,from a worse to a better rank. This shift also alters the pairwise 
significance between DrCIF and InceptionTime from not significant to significant. (In this scenario, ResNet is a 
weaker deep learning algorithm compared to InceptionTime, and STSF is a weaker interval method compared to DrCIF.)

Figure~\ref{fig:cdd-swap-2} shows how adding a weaker version of a comparate (e.g., a different version of a classifier 
with fewer parameters or different hyperparameter tuning) can influence the mean rank and pairwise statistical significance.
Adding a weaker version of a 
comparate can significantly elevate the rank of the original comparate and change the statistical significance of 
pairwise differences between comparates. The original comparate, being more accurate than its weaker version, will 
have a better mean rank on many tasks, resulting in a lower p-value in the Wilcoxon signed-rank test, which can 
shift pairwise differences to become statistically significant under the Holm correction.

Therefore, we suggest that comparates should be ordered using a statistical measure that remains stable regardless 
of the addition or removal of other comparates. This approach ensures that the relative ranking of different comparates 
is consistent across studies and not susceptible to manipulation, either intentionally or unintentionally.

\subsection{Insufficient Attention to the Magnitude of Wins and Losses}~\label{sec:wtl-not-sufficient}

The mean rank evaluates the frequency with which a comparate outperforms or underperforms others across multiple tasks, 
such as achieving higher or lower classification accuracy. It does not consider the extent of these performance 
differences. Therefore, a comparate might achieve a low average rank through several minor wins while also experiencing 
significant losses. For instance, if a comparate $c_i$ has a $90\%$ likelihood of a slight loss against $c_j$ but a $10\%$ 
likelihood of a substantial gain, most would prefer $c_i$. Nevertheless, mean rank would heavily favor $c_j$.

While the Wilcoxon test somewhat considers the magnitude of wins, it does not fully resolve the issue when comparisons 
are primarily based on rank, as illustrated in the CDD of Figures~\ref{fig:cdd-swap-1} and~\ref{fig:cdd-swap-2}.

\subsection{Null Hypothesis Significance Testing}~\label{sec:issue-nhst}

The practice of using statistical significance tests to evaluate performance on benchmark tasks is facing growing
criticism.~\cite{bayseian-benavoli-paper} asserts that null hypothesis significance testing (NHST), such as the Wilcoxon test, 
is not optimal for benchmarking scenarios involving multiple comparates and multiple tasks for four main reasons. 
This viewpoint is also echoed by~\cite{pvalues-pitfal-paper}.

\textbf{First}, Null hypothesis significance testing (NHST) does not provide the probability of the alternative hypothesis 
(the hypothesis that there is a difference between comparates given the observed results). Instead, NHST calculates the 
likelihood of observing the outcomes $O$ assuming the null hypothesis $H_0$ is true, 
represented as $p(O|H_0)$, meaning there is no difference between comparates.~\cite{bayseian-benavoli-paper} suggests that 
a more meaningful assessment would be $p(H_0|O)$, the probability of the null hypothesis being \textit{true}
given the observed outcomes.

\textbf{Second},~\cite{bayseian-benavoli-paper} highlights that the null hypothesis 
can always be rejected by simply increasing the number of 
examples, a point we revisit in Section~\ref{sec:holm-manipulation}.

\textbf{Third},~\cite{bayseian-benavoli-paper} claims that NHST fails to indicate the 
magnitude of differences between comparates,
regardless of how small the p-value might be. For instance, if one comparate's accuracy is only $10^{-4}$ 
higher than another's on most benchmark tasks, the p-value (e.g., from a Wilcoxon test) could be extremely low, 
misleadingly suggesting a significant difference in accuracy. However, the p-value does not provide any insight 
into the actual magnitude of the performance difference between comparates. This problem is related to, but separate from, 
the issue discussed in Section~\ref{sec:wtl-not-sufficient}, where ranks do not reflect the magnitude of performance 
differences between comparates.

\textbf{Finally}, because the p-value indicates the likelihood of observing the given outcomes under the assumption 
that the null hypothesis is true, it does not directly reflect the probability that the null hypothesis itself 
is true. Therefore, a large p-value (which might suggest that differences are not statistically significant) 
does not actually confirm or refute the null hypothesis~\cite{lecoutre2014significance}.

\subsubsection{Inferential vs Descriptive Statistics}
The p-value derived from a statistical hypothesis test is often utilized as an inferential statistic rather than 
a descriptive one. Descriptive statistics accurately summarize the empirical properties of the results, whereas 
inferential statistics attempt to draw conclusions about the population from which the data was sampled. Practically, 
inferential statistics aim to predict or quantify how a comparate would perform on new or unseen data randomly drawn 
from the same source. Inferential statistics make stronger claims about the results by relying on stronger assumptions.

For instance, when applying the Wilcoxon test to two comparates $c_i$ and $c_j$ over $n$ tasks 
$\mathcal{D} = \{d_1, \ldots, d_n\}$, we obtain performance measures $\gamma(c_i,d_1), \ldots, \gamma(c_i,d_n)$ 
and $\gamma(c_j,d_1), \ldots, \gamma(c_j,d_n)$. The two-tailed Wilcoxon test returns the probability $p$ that the 
differences $\gamma(c_i,d_1) - \gamma(c_j,d_1), \ldots, \gamma(c_i,d_n) - \gamma(c_j,d_n)$ 
would be observed if these values were an independent and identically 
distributed (iid) sample from a distribution $\Omega$ that is symmetric around zero. 
As an inferential statistic, we reject the null hypothesis that the distribution of these differences is symmetric 
if $p-value \leq \alpha$, where $\alpha$ is the chosen significance level. The p-value represents the probability of 
obtaining a test statistic as extreme as, or more extreme than, the one observed, purely by chance, if the comparates 
were selected without prior knowledge of the data.

In traditional scientific experiments, this approach is feasible because data is typically collected freshly for each 
experiment. However, in many benchmarking scenarios, researchers select comparates based on algorithms and hyper-parameters 
that perform well on the given benchmark. They do not collect new benchmark data by re-sampling from the problems that 
define the benchmark. Therefore, it is difficult to identify a meaningful distribution from which the performance scores 
could be considered an iid sample, making the use of the test statistic and p-value for inferential purposes problematic.

Despite these issues, test statistics and p-values are valuable for measuring divergence between two or more sets 
of data points (e.g., the classification accuracies of two comparates). They provide a quantitative descriptive measure 
of performance differences between comparates.

\subsection{The Use of Multiple Test Corrections}~\label{sec:holm-manipulation}

When performing multiple tests, such as Wilcoxon tests between all pairs of comparates, it is generally accepted that the 
significance level (typically $0.05$) should be adjusted 
\cite{demsar-cdd-paper,bayseian-benavoli-paper,pvalues-pitfal-paper}.
Multiple test corrections aim to manage the risk of incorrectly rejecting any null hypothesis when several null 
hypotheses are tested simultaneously. The Holm correction~\cite{holm-correction-paper} is a widely used method for this 
purpose (see Section~\ref{sec:wilcoxon} for details on the Holm correction).

We identify that it is more crucial to control the risk for each pair of comparates rather than for the entire 
study. We also show that multiple testing corrections can introduce undesirable effects, potentially leading 
to the manipulation of benchmark results. This undermines efforts to provide a consistent and stable comparison 
of multiple comparates across tasks.

A significant issue is that the number of comparates in a study and the specific comparates included can affect the 
likelihood of one comparate outperforming another~\cite{bayseian-benavoli-paper}.
This enables the manipulation of the statistical significance of differences between comparates, whether
intentionally or unintentionally, by including or excluding certain comparates from the comparison.

\begin{figure}[ht]
    \begin{subfigure}{0.5\linewidth}
        \centering
        \includegraphics[width=0.995\linewidth]{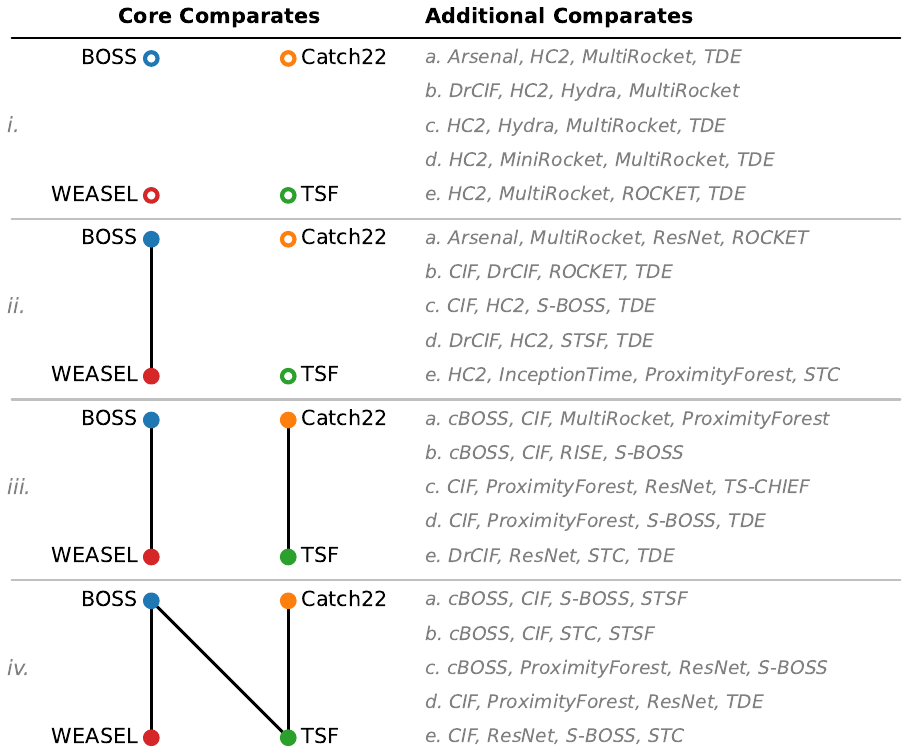}
        \caption{\null}
    \end{subfigure}
    \begin{subfigure}{0.5\linewidth}
        \centering
        \includegraphics[width=0.995\linewidth]{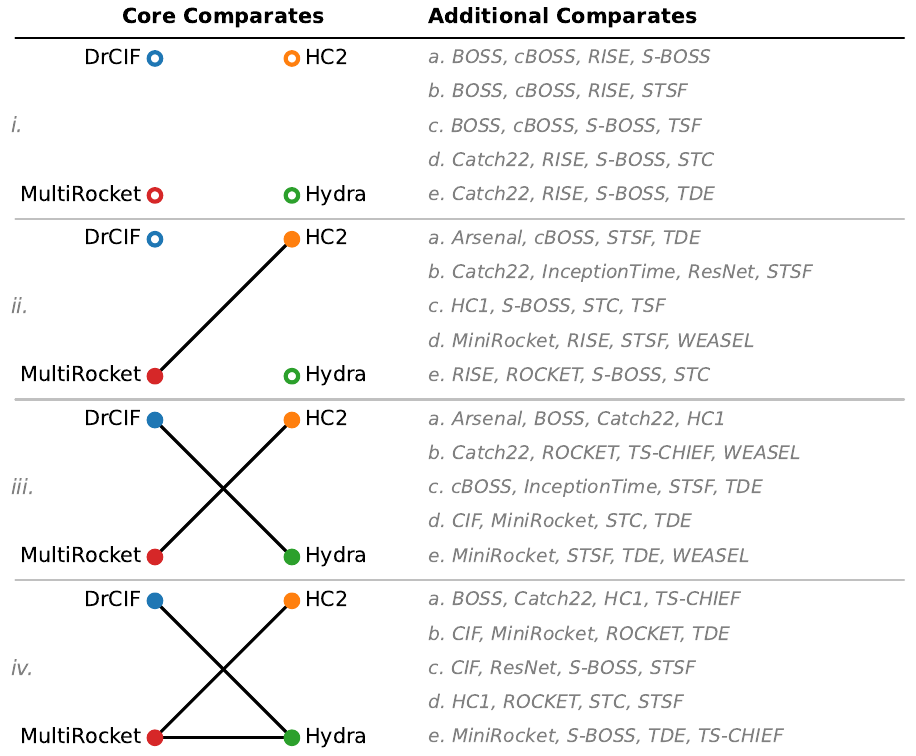}
        \caption{\null}
    \end{subfigure}
    \caption{
        Two examples demonstrate the instability of pairwise significance under the Holm correction: (a) for the 
        comparates BOSS, Catch22, TSF, and WEASEL; and (b) for the comparates DrCIF, HC2, Hydra, and MultiRocket. 
        The statistical significance of pairwise differences between comparates is influenced by the additional 
        comparates included in the comparison. In each case, four different patterns of statistically significant 
        pairwise differences,(i), (ii), (iii), and (iv),are shown in the left column (pairs with non-significant 
        differences according to the Wilcoxon signed-rank test with Holm correction are connected by a black line). 
        Randomly chosen examples of additional comparates that result in the given pattern of statistically 
        significant pairwise differences are shown in the right column.
    }
    \label{fig:stability-holm}
\end{figure}

To highlight the instability of pairwise significance when using the Holm correction, we illustrate how different sets 
of comparates influence the results. Figure~\ref{fig:stability-holm} shows two examples of this instability, demonstrating how the statistical 
significance of pairwise differences can change with different sets of comparates. In each example, we begin with a 
core set of four comparates and repeatedly combine this core set with different sets of four additional comparates. 
For each combination, the Wilcoxon test with Holm correction was applied to all pairs, noting which pairwise differences 
within the core set were statistically significant. Non-significant pairwise differences within the core set are indicated 
by lines connecting the comparates.

Figure~\ref{fig:stability-holm} presents results for core sets consisting of (a) BOSS, Catch22, TSF, and WEASEL; and (b) DrCIF, 
HC2, Hydra, and MultiRocket. These examples are based on results for 23 different comparates across 108 datasets from 
the UCR archive. For instance, in Figure~\ref{fig:stability-holm}(a), combining comparates BOSS, Catch22, TSF, and WEASEL 
with any of the five sets of comparates listed for pattern (ii),such as Arsenal, MultiRocket, ResNet, and Rocket,produces 
the pattern shown on the left, where the pairwise differences between BOSS and WEASEL are not statistically significant. 
However, combining the same core comparates with any of the sets listed in (i), (iii), or (iv) results in different 
patterns of statistical significance.

The examples in Figure~\ref{fig:stability-holm} illustrate that the statistical significance of pairwise differences can 
be influenced by adding or removing comparates, whether intentionally or unintentionally. In many instances, various
different sets of additional comparates can result in the same pattern of statistical significance for pairwise differences. 
For instance, in Figure~\ref{fig:stability-holm}(b), there are 123 different sets of additional comparates that produce pattern 
(i), 1,876 sets that produce pattern (ii), 680 sets that produce pattern (iii), and 1,197 sets that produce pattern (iv). 
For simplicity, only five randomly selected combinations of additional comparates are shown for each pattern in Figure~\ref{fig:stability-holm}.

This issue occurs because a multiple testing correction, such as the Holm correction, adjusts the threshold for statistical 
significance based on the p-values of all pairs of comparates. Consequently, the significance of pairwise differences for a given 
pair of comparates depends on the p-values of all other pairs. Adding or removing comparates with small 
p-values can shift pairwise differences above or below the threshold for significance. For a given pair of comparates, 
including or excluding other comparates can turn an otherwise significant difference into a non-significant one, and vice versa.

Another issue is that multiple test corrections aim to prevent any algorithm from being incorrectly found superior by chance. 
However, this comes at the expense of increasing the risk of overlooking true findings of superiority. It is debatable why one 
risk should be prioritized over the other. The ability to claim that a new algorithm is not significantly less effective 
than the current state-of-the-art allows proponents to add enough algorithms to a comparison to achieve such a claim.

To avoid the ``data dredging'' problem, the adjustment for multiple testing should ideally consider the total number of 
comparates, including all variations of an algorithm that were tested and discarded during development. However, this is 
often impractical, suggesting that such efforts may be futile.

Moreover, different studies might produce varying results for pairwise comparisons of the same two competitors depending 
on the number of comparates included. A study with fewer comparates might reject the null hypothesis and find a significant 
difference between comparates $c_i$ and $c_j$, whereas a study with more comparates might fail to reject the null hypothesis, 
finding no significant difference based on the same evidence. This inconsistency undermines the reliability of such evidence 
bases.

\section{An Alternative Approach}

As previously mentioned, recent studies have attempted to tackle some of these issues, especially regarding the statistical 
significance testing of pairwise differences between comparates. Notably, we emphasize the approach proposed
by~\cite{bayseian-benavoli-paper}.

\cite{bayseian-benavoli-paper} argued that the Wilcoxon test, or similar tests, should be replaced by a new Bayesian test 
modeled after the Wilcoxon test. For comparing two comparates $c_i$ and $c_j$, the Bayesian signed rank test generates 
a probability distribution indicating the likelihood that $c_i$ is significantly better than $c_j$, $c_j$ is significantly 
better than $c_i$, or $c_i$ and $c_j$ are not significantly different. 

Let $\textbf{z} = [z_0, z_1, \dots, z_i, \dots, z_q]$ be the vector of performance differences between $c_i$ and $c_j$ 
across $q$ tasks, including a pseudo observation $z_0$, a hyperparameter of the Dirichlet Process (DP). Assuming 
$\textbf{z}$ follows a DP, the resulting probability distribution is given by:

\begin{equation}
    \begin{split}
        \theta_l &= \sum_{i=0}^q\sum_{j=0}^q \omega_i \omega_j \mathbf{I}_{(-\infty,-2r)}(z_i+z_j)\\
        \theta_e &= \sum_{i=0}^q\sum_{j=0}^q \omega_i \omega_j \mathbf{I}_{(-2r,2r)}(z_i+z_j)\\
        \theta_r &= \sum_{i=0}^q\sum_{j=0}^q \omega_i \omega_j \mathbf{I}_{(2r,\infty)}(z_i+z_j),
    \end{split}
\end{equation}

\noindent
where $\mathbf{I}_A(x) = 1$ if $x \in A$, the weights $\omega_i$ follow a Dirichlet distribution $D(s,1,1,\ldots,1)$ and 
$r$ is the ``rope'' value that sets the interval of which the two classifiers have no significant difference.
Given that the distribution $\theta_l$, $\theta_e$,and $\theta_r$ does not have a closed form solution, the probability 
distribution is generated using a Monte Carlo sampling on the weights $\omega_i$.

\begin{figure}
    \centering
    \includegraphics[width=0.65\linewidth]{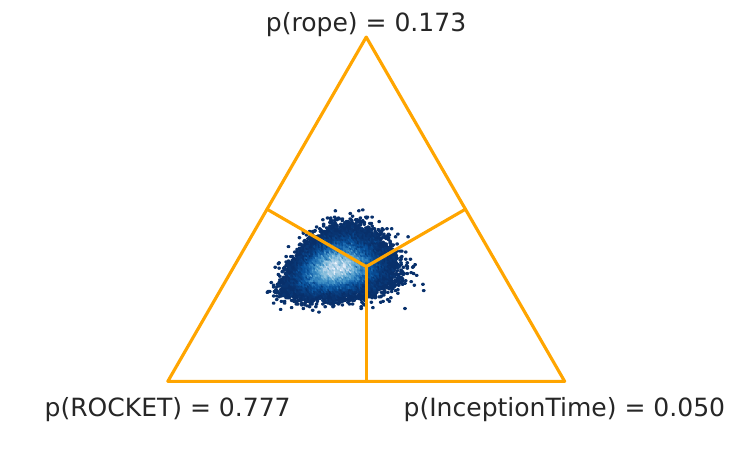}
    \caption{A visualization of the Bayesian Signed Rank Test proposed in~\cite{bayseian-benavoli-paper} as a 
    replacement for the Wilcoxon Signed Rank Test is provided. As illustrated, the Bayesian test offers information 
    on the probability that the null hypothesis is true, given the performance metrics of both InceptionTime and 
    ROCKET on the 108 tasks from the UCR archive.
    }
    \label{fig:bayesian_triangle}
\end{figure}

An example of this Bayesian test is shown in Figure~\ref{fig:bayesian_triangle}, comparing InceptionTime~\cite{inceptiontime-paper} 
and ROCKET~\cite{rocket} on 108 datasets from the UCR archive~\cite{ucr-archive}. The triangle illustrates that there is a 
$77.7\%$ probability that ROCKET outperforms InceptionTime, with only a $17.3\%$ probability that the classifiers are not 
meaningfully different.

Considering the limitations discussed in this chapter, there is a clear need for an alternative approach instead 
of the CDD.
The method proposed by~\cite{bayseian-benavoli-paper} highlights the importance of measuring the statistical 
significance of pairwise differences between classifiers as an alternative to the Wilcoxon signed-rank test. 
This Bayesian approach is not mutually exclusive with the method described in the following section; instead, 
we consider it complementary. These efforts are geared towards enhancing the robustness of statistical testing 
for differences between comparates. Instead of relying on p-values derived from the Wilcoxon signed-rank test, 
the probabilities computed via the Bayesian signed-rank test could be utilized in the alternative approach 
detailed in the subsequent section.

\section{The Multi-Comparison Matrix}\label{sec:MCM}

Our goal is to develop methods for assessing $m$ comparates $\mathcal{C}$ across multiple datasets $\mathcal{D}$
using a single performance measure $\gamma$ and pairwise comparison measure $\delta$ that:

\begin{itemize}
    \item prioritizes pairwise comparisons between comparates;
    \item focuses on descriptive statistics over statistical hypothesis testing;
    \item ensures that pairwise comparisons $\delta(c_i,c_j)$ between any two comparates $c_i\in\mathcal{C}$ and 
    $c_j\in\mathcal{C}$ are invariant to $\mathcal{C}\backslash\{c_i,c_j\}$ (i.e., no pairwise comparison will 
    change with the addition or deletion of other comparates, maintaining consistency across studies);
    \item orders comparates such that the relative order of any two comparates $c_i\in\mathcal{C}$ and $c_j\in\mathcal{C}$ 
    is invariant to $\mathcal{C}\backslash\{c_i,c_j\}$ (i.e., the order of $c_i$ and $c_j$ will not change with the addition 
    or deletion of other comparates, ensuring stability across studies);
    \item provides a good balance between the amount of information presented and the informativeness of that information.
\end{itemize}

To achieve this, we propose a grid of pairwise comparison statistics, as shown in
Figures~\ref{fig:mcm-example} and~\ref{fig:mcm-example-row}. 
The proposed Multi-Comparison Matrix (MCM) maintains the pairwise comparisons between each pair of comparates 
$c_i$ and $c_j$, ordering the comparates by default based on the average performance measure $\gamma$. 
Each cell of this matrix contains three pairwise statistics between $c_i$, the comparate for the row, and $c_j$, 
the comparate for the column. 
These three statistics are:

\begin{itemize}
    \item The mean of $\gamma(c_i,d)-\gamma(c_j,d)$ over all $d\in\mathcal{D}$;
    \item A Win Tie Loss count for $c_i$ against $c_j$ over all the tasks (datasets) in $\mathcal{D}$;
    \item A p-value ($p$) for a Wilcoxon Signed Rank Test ($\delta$).
\end{itemize}
Note that despite concerns about using statistical significance testing for benchmarking, we continue to use 
the Wilcoxon test and the associated $p$ value. While there are arguments for excluding formal statistical hypothesis tests, 
we recognize that this may be too drastic for many. Therefore, we prioritize descriptive statistics. However, considering 
the issues discussed above, we encourage interpreting the $p$ value as a \textit{descriptive} statistic, a measure of the 
"strength" of the difference between comparates, rather than as an inferential statistic. In other words, the $p$ value 
should not be viewed as indicating the likelihood or probability of observing a similar difference in accuracy between 
comparates on new or unseen data (i.e., ``out of benchmark'').

By default, the MCM is generated with all comparates present in both the rows and the columns,
resulting in $m \times (m-1) / 2$ comparisons. Alternatively, separate lists of comparates for the rows and columns
can be specified, $\mathcal{C}_{\mathrm{row}}$ and $\mathcal{C}_{\mathrm{col}}$. In this case there are
$|\mathcal{C}_{\mathrm{row}}|\times|\mathcal{C}_{\mathrm{col}}|-|\mathcal{C}_{\mathrm{row}}\cap\mathcal{C}_{\mathrm{col}}|$
comparisons.

\subsection{MCM Examples}\label{sec:heatmap}
\begin{figure}
    \centering
    \includegraphics[width=1.0\linewidth]{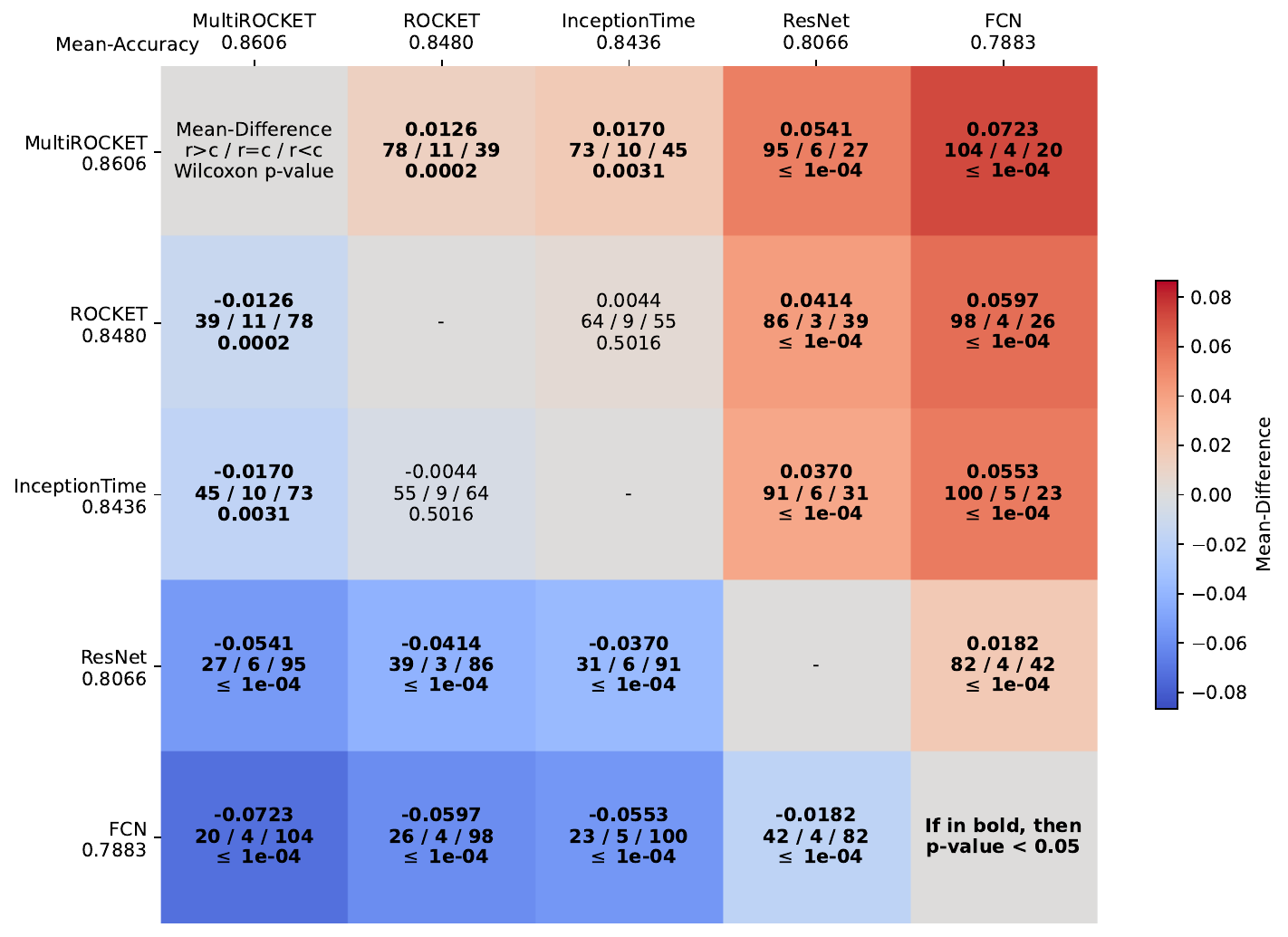}
    \caption{MCM showing all pairwise comparisons between MultiROCKET, ROCKET, InceptionTime, ResNet, and
    FCN on the 128 datasets of the UCR archive. In this setup, the full pairwise comparison is presented.}
    \label{fig:mcm-example}
\end{figure}

By default, all pairwise comparisons between comparates are displayed. The average performance 
measure $\gamma$ (e.g., classification accuracy) across all tasks $\mathcal{D}$ is shown next to each 
comparate label. An example illustrating the results for five comparates over 108 datasets from the UCR 
archive~\cite{ucr-archive} is provided in Figure~\ref{fig:mcm-example}. In the Heat Map, colors represent 
the mean difference in $\gamma$. A positive difference (red) indicates that the row comparate outperforms 
the column comparate on average. For instance, in Figure~\ref{fig:mcm-example}, the top right cell is red, 
demonstrating that MultiROCKET (row) is generally more accurate than ResNet (column). Conversely, a 
negative difference (blue) means that the column comparate outperforms the row comparate on average. 
Text in each cell is in \textbf{BOLD} if the $p$ value is below a specified threshold (e.g., $0.05$).

This format of the MCM is highly effective for presenting comparisons in a benchmark review, such as in~\cite{dl4tsc}. 
In such reviews, detailed information on all pairwise comparisons is essential to highlight the strengths and weaknesses 
of each comparate. For example, this MCM format can illustrate how a comparate that performs poorly against the winning 
comparate might still excel on certain datasets compared to other state-of-the-art comparates.

\begin{figure}
    \centering
    \includegraphics[width=0.75\linewidth]{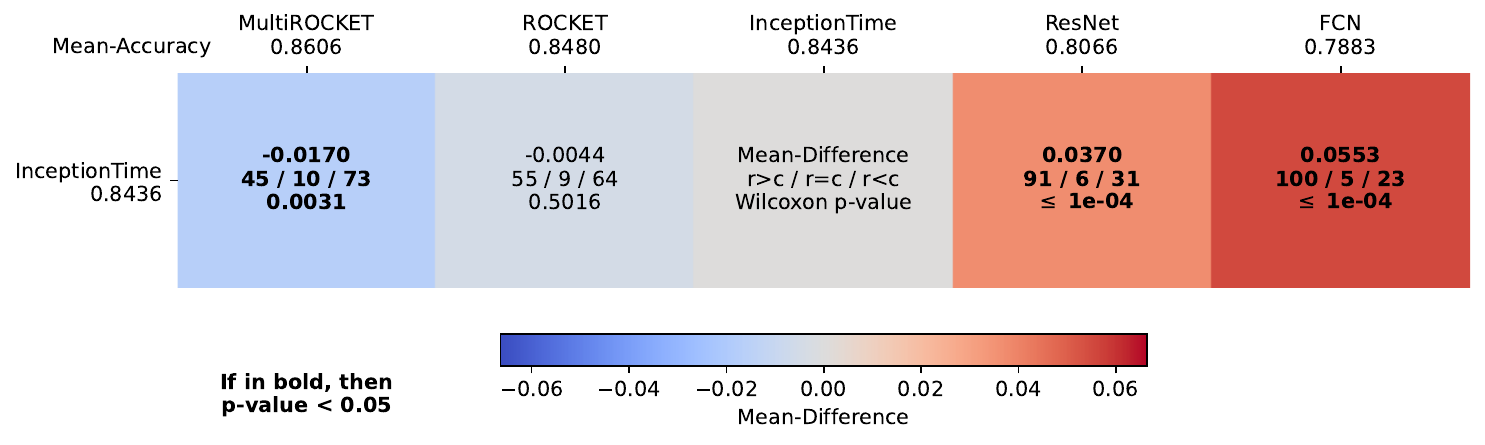}
    \caption{MCM showing pairwise comparisons between InceptionTime and each of ROCKET, 
    MultiROCKET, FCN and ResNet on the 128 datasets of the UCR archive.}
   \label{fig:mcm-example-row}
\end{figure}

However, if a study focuses on a few specific comparates, such as when introducing a new algorithm, it is often more beneficial 
to highlight comparisons between these few and many existing alternatives. In these scenarios, the MCM can be adjusted to display 
only the necessary results for comparing the proposed comparates with the current state-of-the-art. This allows the reader 
to focus on the most pertinent comparisons. The proposed comparates are listed in either a row or a column of the matrix, 
as illustrated in Figure~\ref{fig:mcm-example-row}.

\section{Conclusion}

In this chapter, we have explored the critical aspects of benchmarking machine learning models on time series data. 
The discussion began with an examination of the significance of rigorous benchmarking and the various challenges 
associated with it, including the instability of pairwise comparisons and the inherent limitations of traditional 
statistical significance tests. These challenges highlight the necessity for developing more robust and reliable 
benchmarking techniques.

We detailed the Critical Difference Diagram (CDD), a widely utilized tool for visualizing the performance of 
multiple comparates across various tasks. While the CDD provides valuable insights, it also has significant 
limitations, such as its sensitivity to the addition or removal of comparates and its reliance on average ranks that 
may not adequately reflect performance differences.

To address these limitations, we proposed the Multi-Comparison Matrix (MCM) as a novel approach for presenting 
pairwise comparisons between comparates. The MCM preserves the integrity of pairwise comparisons and ensures that 
these comparisons remain consistent regardless of the inclusion or exclusion of other comparates. This method offers 
a more stable and comprehensive view of comparative performance, enabling researchers to make more informed decisions 
about model selection and evaluation.

The usage of MCM is not intended to act as a ``replacement'' for the CDD, as the CDD remains and will 
continue to be more ``attractive'' due to its simplicity. Instead, we propose that the MCM should serve as a complementary 
analysis tool alongside the CDD in future research, such as we do in the following Chapter.
 
%
%
\chapter{Towards Finding Foundation Models for Time Series Classification} 
\label{chapitre_3}

\section{Introduction}

In the dynamic field of TSC, the quest for developing models that are robust and adaptable across
diverse datasets remains a significant challenge.
Foundation models, which are large pre-trained models capable of generalizing across various tasks, 
offer a promising solution to this problem.
The necessity of foundation models arises from their ability to simplify and expedite the fine-tuning process. 
In many real-world applications, such as predicting heart conditions from ECG signals or forecasting traffic patterns, 
starting model training from scratch is time-consuming and computationally expensive. Foundation models mitigate these 
challenges by offering a pre-trained base that already understands the fundamental patterns within a domain. Consequently, 
fine-tuning becomes a matter of adapting this base model to the nuances of a specific dataset, leading to faster and 
more efficient training with improved performance.
For instance, in the medical field, a model pre-trained on various ECG datasets can be fine-tuned to detect 
arrhythmias with greater accuracy and speed. Similarly, in traffic management, a model pre-trained on traffic 
data from multiple cities can be fine-tuned to predict congestion patterns in a specific city. This allows 
for more accurate and efficient traffic control solutions tailored to local conditions.

This chapter introduces two key contributions aimed at advancing towards
deep foundation models: the creation of hand-crafted convolution filters
to enhance model generalization, and the utilization of a pre-training
methodology
to fine-tune these models for specific classification tasks.
These filters are designed to shift the model's focus from specific features to more fundamental, general 
characteristics that are independent of any particular domain. This ensures that the model can identify 
and leverage intrinsic patterns within 
the data, making it more adaptable across various tasks.
Building on this, our second contribution incorporates a model that utilizes our hand-crafted filters
to develop a preliminary foundation model. This model undergoes extensive pre-training on multiple datasets within the
same domain, such as ECG or traffic data, aiming to predict the original dataset of each series. This pre-training equips
the model with a broad understanding of domain-specific patterns, providing a strong foundation for subsequent fine-tuning.
Once pre-trained, the model is fine-tuned on individual datasets to address specific classification tasks. 

By combining the strengths of hand-crafted filters with a robust pre-training and fine-tuning methodology, we aim to 
construct a foundation model that is both versatile and powerful. These contributions not only enhance the model's 
performance across varied TSC tasks but also represent a significant step towards the development 
of deep foundation models in this domain.

\section{Hand-Crafted Convolution Filters}~\label{sec:hcf}

The development of deep learning models for time series classification often encounters significant challenges, 
such as overfitting, computational complexity, and redundancy in learned filters. Traditional CNNs~\cite{inceptiontime-paper}
typically learn filters through back-propagation, where filters are initialized randomly and refined 
during training. While effective, this process can lead to several issues:

\begin{itemize}
    \item \textbf{Overfitting}: Learned filters may become overly specialized to the training data, reducing the model's
    ability to generalize to new, unseen data
    \item \textbf{Time Spent on Learning ``Easy'' Features}: During training, models may spend significant time learning simple,
    generic features that could have been predefined, rather than focusing on complex features that are more difficult
    to construct by hand.
    \item \textbf{Difficulty in Finding Generic Filters}: While learning simple filters from scratch is feasible, 
    the process becomes significantly more challenging for complex filters due to error propagation, making it 
    difficult to generalize filters that work effectively across various datasets.
\end{itemize}

\subsection{Are There Any Common Learned Convolution Filters Between Datasets?}\label{sec:common-filters}

One approach to address issues like overfitting, excessive focus on ``easy'' features, and the difficulty of finding
generic filters is to construct hand-crafted convolution filters that detect generic patterns in time series data. 
Before constructing these filters, we must assume that if such generic convolution filters can be manually created, 
deep CNN models should be able to learn them across different datasets.

\begin{figure}
    \centering
    \includegraphics[width=\textwidth]{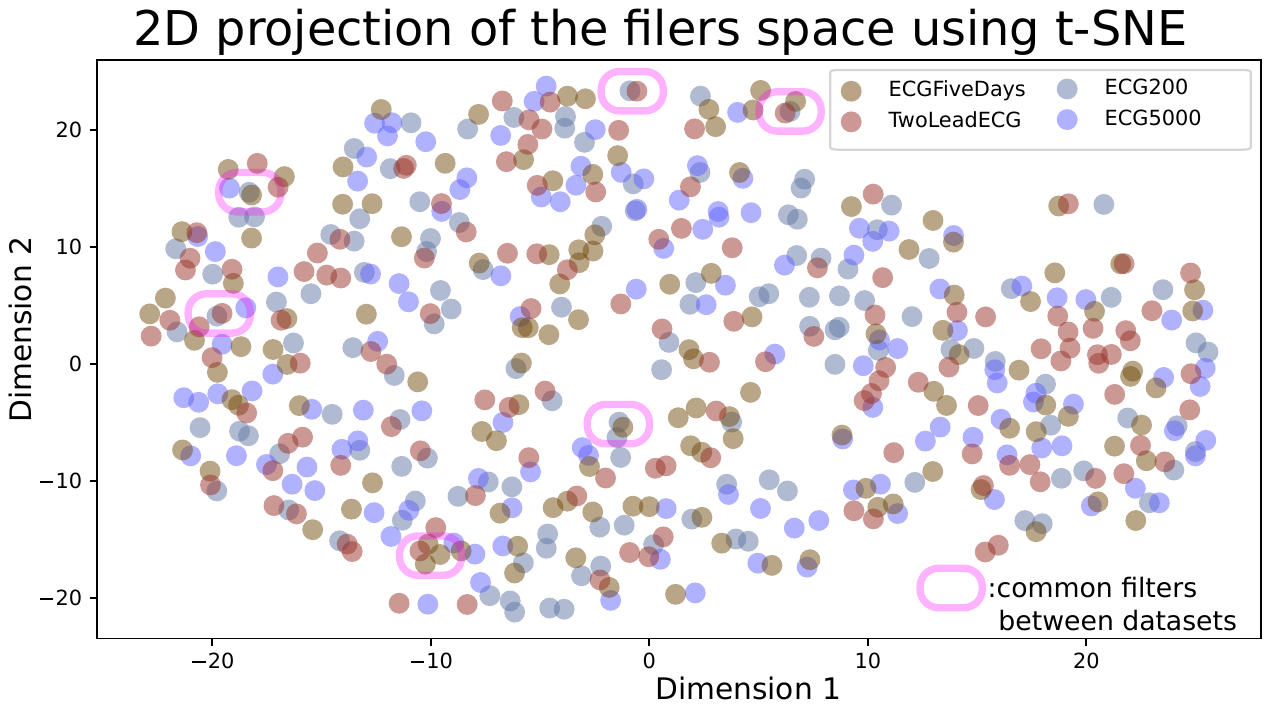}
    \caption{t-SNE two-dimensional projection of the first-layer 
    convolution filters learned by CNN models on
    four ECG datasets (
        \protect\mycolorbox{98,120,166,0.5}{ECG200}, 
        \protect\mycolorbox{98,100,255,0.5}{ECG5000}, 
        \protect\mycolorbox{115,77,16,0.5}{ECGFiveDays}, 
        \protect\mycolorbox{153,50,42,0.5}{TwoLeadECG}). 
    The \protect\mycolorbox{255,0,255,0.3}{clustering of filters} in the t-SNE space
    suggests that deep CNN models can identify common, 
    generic filters across different datasets.}
    \label{fig:common-filters}
\end{figure}

To test this assumption, we analyzed the  t-distributed Stochastic Neighbor Embedding (t-SNE)~\cite{t-sne-paper}
two-dimensional projection of the filter space learned by CNN models, 
as shown in Figure~\ref{fig:common-filters}. We focused on the filters from the first layer since it is more practical 
to identify generic patterns in raw data than in the deeper feature spaces. The t-SNE space was generated using 
the DTW similarly measure to ensure shift independence between the normalized filters.

Figure~\ref{fig:common-filters} shows that a significant number of convolution filters coincide in the t-SNE space across 
four different ECG-based datasets with varying characteristics (training size, time series length). This suggests that certain 
filters may be shared or common across different datasets.
Furthermore, we can consider the model's ability to find optimal solutions as having a specific amount of energy,
$\mathcal{E}_{\text{total}}$. This energy is split into two parts:

\begin{equation}\label{equ:energy-analogy}
    \mathcal{E}_{\text{total}} = \mathcal{E}_1 + \mathcal{E}_2
\end{equation}
\noindent where $\mathcal{E}_1$ is the energy used to find a set of simple filters, and $\mathcal{E}_2$ is the energy used to 
find the remaining filters. By providing the model with some untrained hand-crafted convolution filters that it would 
have found on its own, we effectively set $\mathcal{E}_1$ to 0. This allows the model to focus all its energy
$\mathcal{E}_2$ on finding other filters, reducing overfitting and improving performance.

This idea aligns with our approach of introducing hand-crafted convolution filters specifically designed to detect 
fundamental patterns in time series data. These filters are fixed and not adjusted during training, allowing the model 
to focus on learning more complex and nuanced patterns. This approach has already been addressed in the computer vision 
community with the construction of untrained Sobel convolution filters~\cite{sobel-paper1,sobel-paper2}.
In the following section, we detail the construction of the proposed hand-crafted convolution filters.

\subsection{Construction of Hand-Crafted Filters}~\label{sec:construction-hcf}

The hand-crafted filters are designed to capture specific types of patterns that are common and crucial in
time series data. We propose three types of filters: (1) increasing trend detection, (2) decreasing trend detection 
and (3) peak detection, for which we define as such:

\mydefinition{Increasing Trend Detection Convolution Filter:}

An increasing trend is a sub-sequence of a time series $\textbf{x}$ where the values are strictly increasing in time.
The filter detecting this trend is designed to detect subsequences where values are strictly increasing over time.
An increasing trend detection filter of length $K$ is defined as:

\begin{equation}\label{equ:increasing-trend-filter}
    \textbf{w}_{I_K} = \{(-1)^{k}\}_{k=1}^{K}
\end{equation}

\begin{theorem}[Increasing Trend Dection Convolution Filter]\label{the:increasing-trend-filters}
    Let $K$ be an even positive integer, a convolutional filter 
    $\textbf{w}_{I_K}=[(-1)^{k}$ for $k \in \{1,...,K\}]$ is an 
    increasing trend detection filter of time series, i.e. it only activates
    (produces positive values) on increasing trend segments.
\end{theorem}

\textit{proof}:
Given a time series $\textbf{x}$ of length $L$ and univariate for simplicity,
that contains increasing, decreasing and stationary trends, we will prove in what follows
that the increasing trend detection filter only activates on increasing trends.

For stationary trends: Suppose $\exists (t_0,t_1), \epsilon$ 
where $t_1 > t_0$ and $\forall t \in [t_0,t_1]$ we have 
$|x_{t+1} - x_t| < \epsilon$.
By convolving this segment of the time series $\textbf{x}$ with the filter $w_{I_K}$ we get:
\begin{equation}
    \begin{split}
        \forall t \in [t_0,t_1], s[t] &= \sum_{k=1}^{K} x_{t+k-1}.w_k\\
        &= \underbrace{-x_t + x_{t+1}}_{\textstyle < \epsilon} \underbrace{- x_{t+2} + x_{t+3}}_{\textstyle < \epsilon}- \dots \underbrace{-x_{t+k-1} +x_{t+k}}_{\textstyle < \epsilon}\\
        &\approx 0~\text{ not activated}
    \end{split}
\end{equation}

\textit{continued proof}:
For decreasing trends: Suppose $\exists (t_0,t_1), \epsilon$ 
where $t_1 > t_0$ and $\forall t \in [t_0,t_1]$ we have 
$x_{t+1} < x_t $.
By convolving this segment of the time series $\textbf{x}$ with the filter $w_{I_K}$ we get:
\begin{equation}
    \begin{split}
        \forall t \in [t_0,t_1], s[t] &= \sum_{k=1}^{K} x_{t+k-1}.w_k\\
        &= \underbrace{-x_t + x_{t+1}}_{\textstyle < 0} \underbrace{- x_{t+2} + x_{t+3}}_{\textstyle < 0}- \dots \underbrace{-x_{t+k-1} +x_{t+k}}_{\textstyle < 0}\\
        &< 0~\text{ not activated}
    \end{split}
\end{equation}

\textit{continued proof}:
For increasing trends: Suppose $\exists (t_0,t_1), \epsilon$ 
where $t_1 > t_0$ and $\forall t \in [t_0,t_1]$ we have 
$x_{t+1} > x_t $.
By convolving this segment of the time series $\textbf{x}$ with the filter $w_{I_K}$ we get:
\begin{equation}
    \begin{split}
        \forall t \in [t_0,t_1], s[t] &= \sum_{k=1}^{K} x_{t+k-1}.w_k\\
        &= \underbrace{-x_t + x_{t+1}}_{\textstyle > 0} \underbrace{- x_{t+2} + x_{t+3}}_{\textstyle > 0}- \dots \underbrace{-x_{t+k-1} +x_{t+k}}_{\textstyle > 0}\\
        &> 0~\text{ activated}
    \end{split}
\end{equation}

\mydefinition{Decreasing Trend Detection Convolution Filter:}

A decreasing trend is a sub-sequence of a time series $\textbf{x}$ where the values are strictly decreasing in time.
The filter detecting this trend is designed to detect subsequences where values are strictly decreasing over time.
A decreasing trend detection filter of length $K$ is defined as:

\begin{equation}~\label{equ:decreasing-trend-filter}
    \textbf{w}_{D_K} = \{(-1)^{k+1}\}_{k=1}^{K}
\end{equation}

\begin{theorem}[Decreasing Trend Dection Convolution Filter]\label{the:decreasing-trend-filters}
    Let $K$ be an even positive integer, a convolutional filter 
    $\textbf{w}_{D_K}=[(-1)^{k+1}$ for $k \in \{1,...,K\}]$ is a 
    decreasing trend detection filter of time series, i.e. it only activates
    (produces positive values) on decreasing trend segments.
\end{theorem}

\textit{proof}: The proof of Theorem~\ref{the:decreasing-trend-filters}
follows the same methodology of the proof of Theorem~\ref{the:increasing-trend-filters}.

\mydefinition{Peak Detection Convolution Filter:}

A peak is a sub-sequence of a time series $\textbf{x}$ where the values changed with a large variation increasingly and then
decreasingly.
To detect peaks, we use a filter inspired by the shape of the negative second derivative of a Gaussian function.
The filter mimics this shape using a squared parabolic function divided into three parts: a negative parabolic segment,
a positive parabolic segment, and another negative parabolic segment.
For example, a peak detection filter of length 12 is:
\begin{equation}\label{equ:peak-filter-example}
    \textbf{w}_{P_{12}} = \{-0.25,-1,-1,-0.25,0.5,2,2,0.5,-0.25,-1,-1,-0.25\}
\end{equation}

\begin{figure}
    \centering
    \caption{Three \protect\mycolorbox{0,128,0,0.5}{hand-crafted filters} detecting: 
    (1) increasing trends, (2) decreasing
    trends and (3) peaks in a time series. The 
    \protect\mycolorbox{255,165,0,0.5}{orange points} indicates on which
    time stamps the filters are activated after being convolved 
    with an \protect\mycolorbox{0,0,255,0.5}{input time
    series} from the Meat dataset of the UCR Archive.}
    \label{fig:hcf-summary}
    \includegraphics[width=\textwidth]{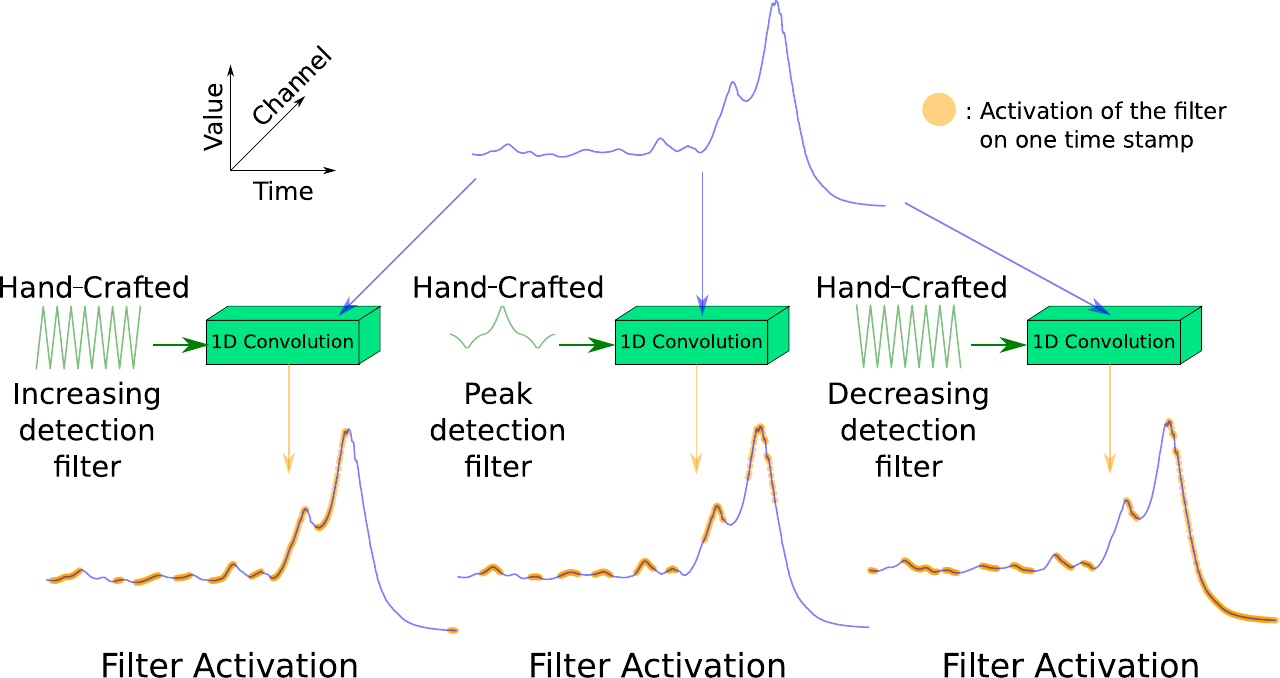}
\end{figure}

These hand-crafted filters are generic and applicable across various datasets without modification,
making them robust tools for initial feature extraction in time series classification tasks.
An example of these three filters can be seen in Figure~\ref{fig:hcf-summary} on the Meat dataset of
the UCR archive~\cite{ucr-archive}.
The output convolution between the input series and each of the hand-crafted filters go through a ReLU activation,
this operation will filter out the negative outcomes, keeping the parts where the filters are activated (the target patterns).

\subsection{Integration Into Deep Learning Architectures}\label{sec:hcf-archis}

To evaluate the impact of these hand-crafted filters, we integrate them into existing deep learning architectures,
mainly FCN~\cite{fcn-resnet-mlp-paper} and Inception~\cite{inceptiontime-paper}.
We propose two adapted versions for FCN and one for Inception:

\subsubsection{Custom Only-Fully Convolutional Network (CO-FCN)}\label{sec:co-fcn}

\begin{figure}
    \centering
    \caption{The CO-FCN architecture, applied non trainable hand-crafted convolution filters on the input data,
    followed by the rest of the FCN architecture~\cite{fcn-resnet-mlp-paper}.}
    \label{fig:co-fcn}
    \includegraphics[width=\textwidth]{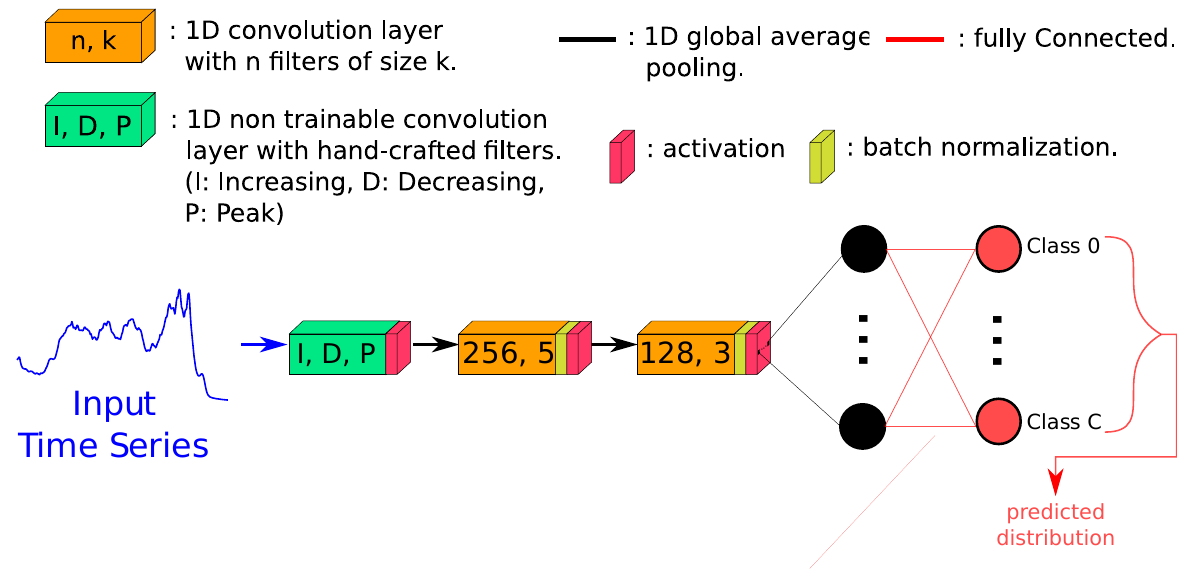}
\end{figure}

In this architecture, the first convolution layer of the standard FCN is replaced entirely by the hand-crafted filters.
This adaptation ensures that the initial feature extraction is driven by these fixed filters, allowing the subsequent
layers to focus on learning more complex patterns.
We refer to this architecture as Custom Only-Fully Convolutional Network (CO-FCN), which evaluates the usage of the 
proposed hand-crafted filters alone with no learnable layers.
The details of this architecture are presented in Figure~\ref{fig:co-fcn}.
To avoid choosing a specific length for hand-crafted filters, we apply a set of different lengths for each of the three 
proposed filters.
We retained the parameters setup used in the last two layers of FCN as detailed in Figure~\ref{fig:fcn}
of Chapter~\ref{chapitre_1}.

\subsubsection{Hybrid-Fully Convolutional Network (H-FCN)}\label{sec:h-fcn}

\begin{figure}
    \centering
    \caption{The H-FCN architecture using non trainable hand-crafted filters in parallel to trainable 
    convolution filters.}
    \label{fig:h-fcn}
    \includegraphics[width=\textwidth]{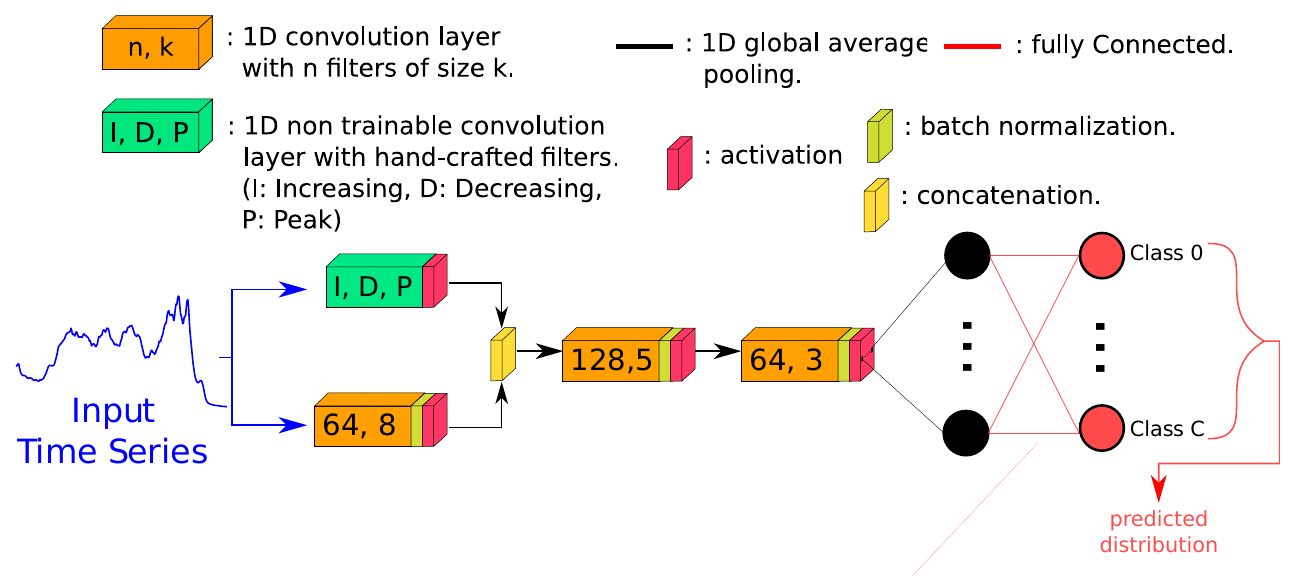}
\end{figure}

The H-FCN architecture enhances the standard FCN by incorporating both hand-crafted and trainable filters in the 
first convolution layer. Features extracted by the hand-crafted filters are concatenated with those from the trainable 
filters, enabling the model to leverage the strengths of both approaches.
The details of this architecture are presented in Figure~\ref{fig:h-fcn}, and such as CO-FCN, we utilize a set of different 
lengths for the hand-crafted filters.
The rest of the FCN architecture is used, however,
unlike the original FCN, given we incorporate hand-crafted filters, we reduce the number of filters throughout all the 
rest of the network by half.
This is due to the fact that the model does not need many filters now given the presence of the hand-crafted filters,
increasing the model's efficiency.

\subsubsection{Hybrid-Inception (H-Inception)}\label{sec:hinceptiontime}

\begin{figure}
    \centering
    \caption{The H-Inception architecture using non trainable hand-crafted filters in parallel to trainable 
    convolution filters of Inception~\cite{inceptiontime-paper}.}
    \label{fig:h-inception}
    \includegraphics[width=\textwidth]{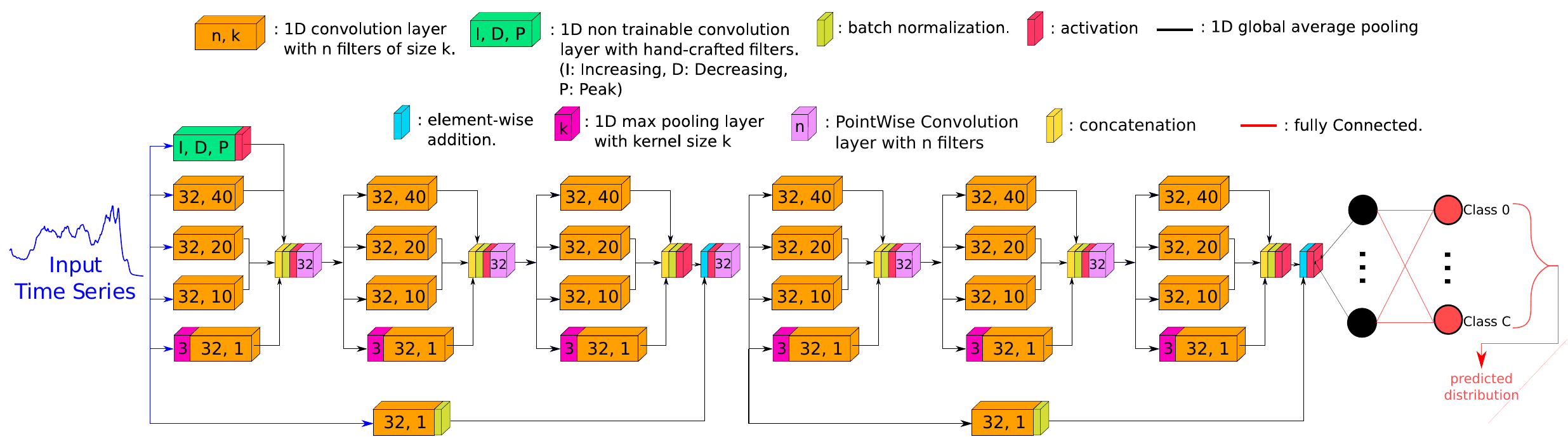}
\end{figure}

The Hybrid-Inception (H-Inception) architecture integrates hand-crafted filters into the Inception model~\cite{inceptiontime-paper}
(see Figure~\ref{fig:inception} of Chapter~\ref{chapitre_1} for the network details), which is known for 
its superior performance in TSC. Similar to the H-FCN, the H-Inception network combines 
the features captured by the hand-crafted filters with those extracted by the first Inception block. This concatenation 
occurs before the data is processed by the remaining layers of the Inception network. Additionally, we construct H-InceptionTime 
which leverages an ensemble of five H-Inception models (similar to InceptionTime) to further enhance its performance.
The integration of hand-crafted filters in this complex architecture allows it to maintain its depth and 
capacity to capture diverse patterns, while the ensemble approach ensures robustness and improved accuracy 
across various datasets.
Figure~\ref{fig:h-inception} presents the detailed architecture of H-Inception.
The rest of the Inception architecture and parameters are retained the same.
The reason to why in H-Inception we do not reduce the number of filters to learn, is that the number of filters per 
convolution is small, and much smaller than the one set in FCN (32 < 128).

In the following section, we present a detailed evaluation of these three proposed architectures on 128 TSC datasets 
of the UCR archive~\cite{ucr-archive}, proving that the hand-crafted filters help the models to generalize, paving
the way to constructing a foundation generic model for TSC tasks.

\subsection{Experimental Setup}

The evaluation of the proposed architectures was conducted using the UCR Archive, which includes 128 labeled univariate 
time series datasets. Each time series in these datasets undergoes z-normalization to achieve a zero-mean and
unit-standard-deviation.
The performance of each model was measured by comparing their accuracy on these datasets. The models were trained 
using the Adam optimizer with a learning rate decay, and the best-performing model based on training loss was selected 
for evaluation on the test set. To ensure robustness, the training process was repeated five times with different initialization, 
and the results were averaged.

We compared the performance of our adapted architectures with hand-crafted filters to the original models. 
For each pair of models, we compared the accuracy on each dataset and computed the number of wins, ties, and losses. 
These comparative results are presented using Win/Tie/Loss one-vs-one plots, showing the Win/Tie/Loss count between 
two different classifiers on the 128 datasets of the UCR Archive. Each point in the plots represents a single dataset from the UCR Archive. 
The axes display the accuracy of each classifier between $0$ and $1$. Additionally, to assess the significance of the comparisons, 
the Wilcoxon Signed Rank Test~\cite{wilcoxon-paper} was performed for each pair of classifiers. 
The resulting statistical measure, the p-value, is shown in the legend of each plot. 
We set the $\alpha$ threshold for the p-values to $0.05$, as done in the literature~\cite{dl4tsc}.

\subsection{Experimental Results}

In this section we compared the proposed architectures to their original network, as well as to other deep learning models 
for TSC such as ResNet (see Figure~\ref{fig:resnet}).
We follow this comparison by looking into the standing of our proposed models to non-deep learning state-of-the-art methods.
Finally, we present a detailed analysis into the changes the FCN variants go through when using the hand-crafted filters.

\subsubsection{Comparing To Original Networks}

\begin{figure}
    \centering
    \caption{Results over 128 datasets of the UCR archive~\cite{ucr-archive} presented in a 1v1 scatter plot format
    between FCN, InceptionTime (IT) and their variants CO-FCN, H-FCN and H-InceptionTime (H-IT) using the
    hand-crafted convolution filters.}
    \begin{subfigure}{0.32\linewidth}
        \centering
        \includegraphics[width=\linewidth]{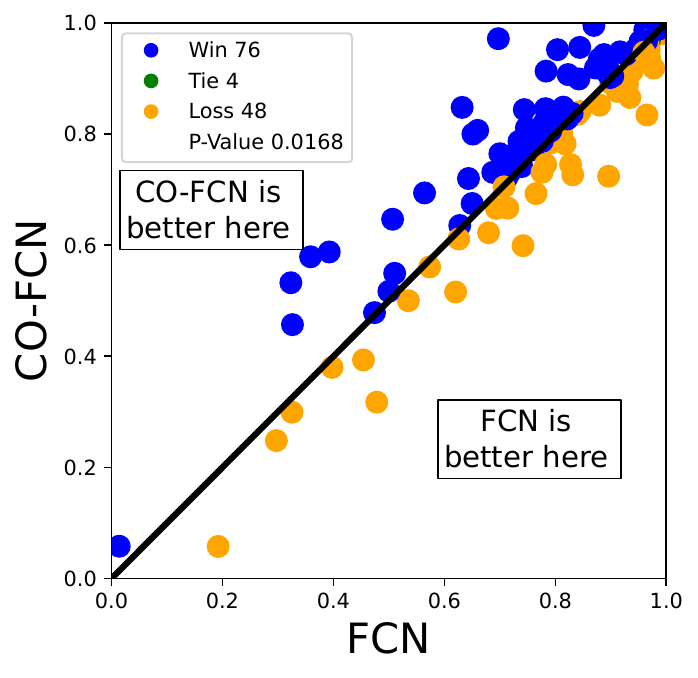}
        \caption{\null}
        \label{fig:cofcn-vs-fcn}
    \end{subfigure}
    \begin{subfigure}{0.32\linewidth}
        \centering
        \includegraphics[width=\linewidth]{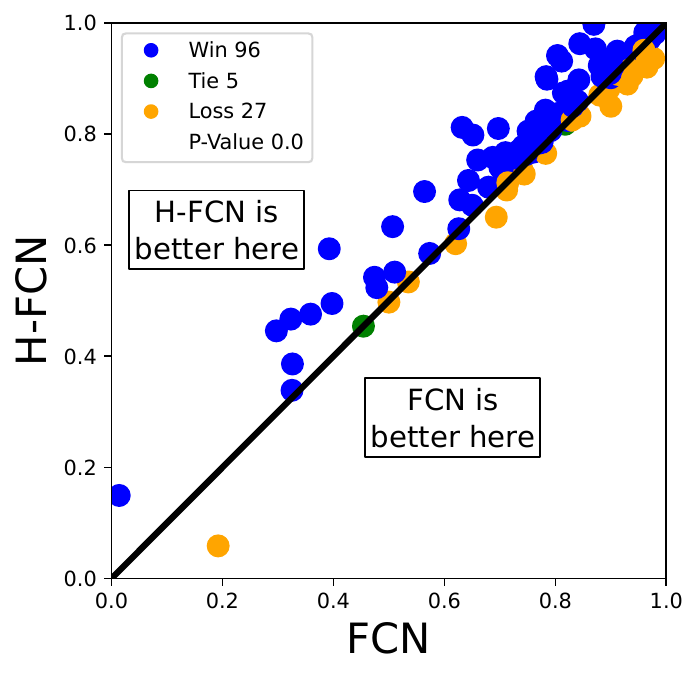}
        \caption{\null}
        \label{fig:hfcn-vs-fcn}
    \end{subfigure}
    \begin{subfigure}{0.32\linewidth}
        \centering
        \includegraphics[width=\linewidth]{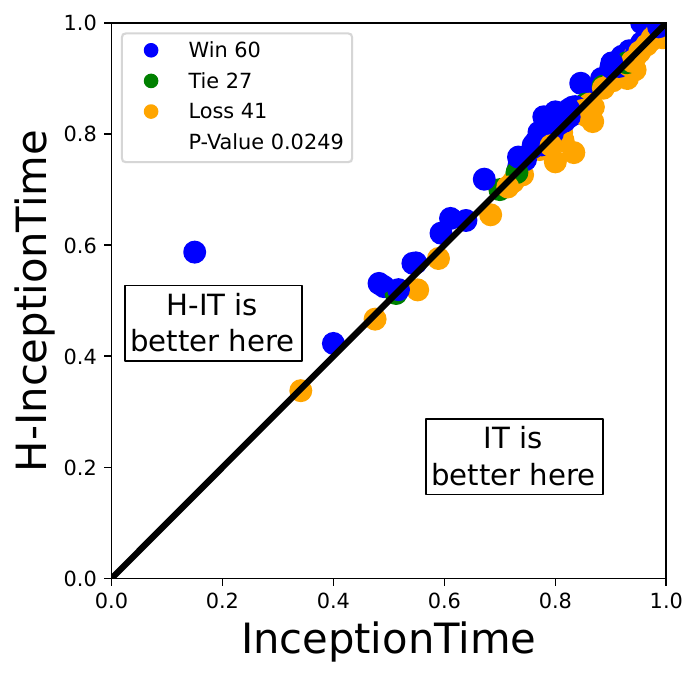}
        \caption{\null}
        \label{fig:hit-vs-it}
    \end{subfigure}
\end{figure}

The CO-FCN model, which replaces the first convolutional layer of the FCN with hand-crafted filters, demonstrated 
improved performance over the original FCN in most cases. The Win/Tie/Loss analysis, presented in Figure~\ref{fig:cofcn-vs-fcn}, showed that CO-FCN outperformed 
the original FCN on a majority of datasets, indicating that hand-crafted filters can effectively replace learned filters 
in the initial layer of the network.
The p-value between the performances of CO-FCN and FCN is $0.0168<0.05$ indicating that on the 128 datasets of the UCR archive,
CO-FCN outperforms FCN with a, statistically, significant difference of performance.
However, there were instances where the original FCN performed better, suggesting that
a hybrid approach could be beneficial, highlighting again the need to quantify the win/loss margin between two models.

The H-FCN model, which combines hand-crafted and learnable filters, showed significant improvements over the original FCN 
model. The Win/Tie/Loss analysis, presented in Figure~\ref{fig:hfcn-vs-fcn}, highlighted that H-FCN achieved better
accuracy in a substantial number 
of datasets compared to these models.
The performance difference between FCN and H-FCN is statistically significant (p-value$<0.05$), with FCN having a very
small winning margin in terms of accuracy values. 
This demonstrates that by adding just a few hand-crafted filters, the performance of FCN can be significantly enhanced, 
while also being more efficient, as H-FCN ($77~440$ parameters) reduces the number of parameters by 
almost three times compared to FCN ($264~704$ parameters).

The H-InceptionTime model, variant of the InceptionTime model with hand-crafted filters, 
demonstrated significant performance improvements. As seen in Figure~\ref{fig:hit-vs-it}, H-InceptionTime surpassed
the InceptionTime ensemble in terms of accuracy,
with a statistical significance in terms of difference of performance as the p-value is less than the threshold. 
This makes H-InceptionTime the newest state-of-the-art deep learning model for TSC.

\subsubsection{Comparing To Other Networks}\label{sec:hfc-between-models-compare}

\begin{figure}
    \centering
    \caption{Critical Difference Diagram~\cite{cdd-benavoli-paper} presenting a comparison between three proposed 
    networks and three state-of-the-art deep learning models for TSC, aggregated over the 128 datasets of the UCR archive.}
    \label{fig:hcf-cdd-deep}
    \includegraphics[width=\textwidth]{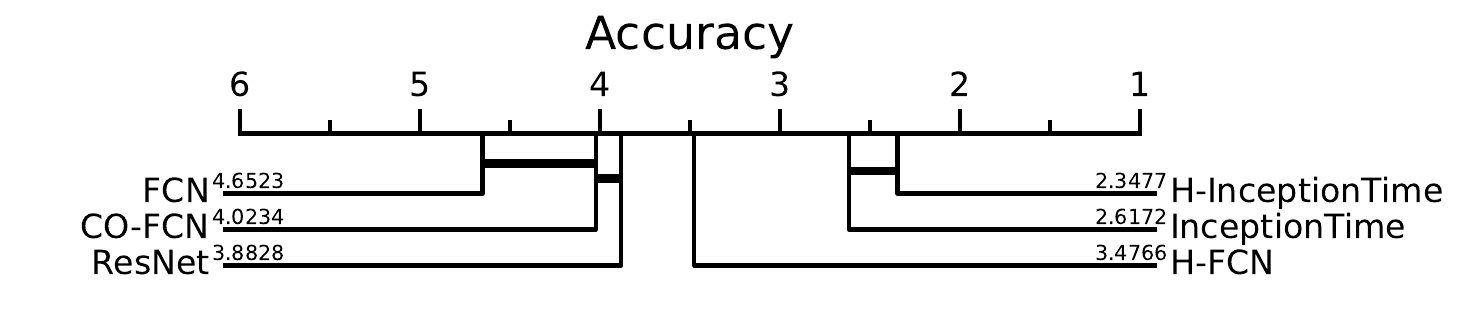}
\end{figure}

After demonstrating that hand-crafted filters can enhance a network's performance on the UCR archive, as shown in the previous section, 
we now present a cross-comparison between models. 
For instance, Figure~\ref{fig:hcf-cdd-deep} illustrates the CD diagram from~\cite{cdd-benavoli-paper}, implemented 
in\footnote{https://github.com/hfawaz/cd-diagram}. 
This CD diagram provides a multi-classifier comparison between the three proposed networks and three state-of-the-art baseline networks: FCN, ResNet, and InceptionTime.

The CD diagram highlights that H-FCN significantly outperforms ResNet,
while ResNet significantly outperforms FCN, 
demonstrating the substantial impact of the hand-crafted filters. 
Furthermore, the diagram shows that, in terms of average rank on the accuracy metric, H-InceptionTime is the top performer. 
However, the diagram also indicates a non-significant difference in performance between H-InceptionTime and InceptionTime, 
which contradicts the 1v1 scatter plot in Figure~\ref{fig:hit-vs-it}, highlighting the issues associated with Holm correction 
discussed in Chapter~\ref{chapitre_2}, Section~\ref{sec:holm-manipulation}. 
A similar argument applies to the comparison between CO-FCN and FCN. This issue can also lead to misleading conclusions about 
the non-significance in difference of performance between CO-FCN and ResNet,whether it exists due to Holm correction.

\begin{figure}
    \centering
    \caption{Multi-Comparison Matrix (Chapter~\ref{chapitre_2}) presenting a comparison between three proposed 
    networks and three state-of-the-art deep learning models for TSC, aggregated over the 128 datasets of the UCR archive.}
    \label{fig:hcf-mcm-deep}
    \includegraphics[width=\textwidth]{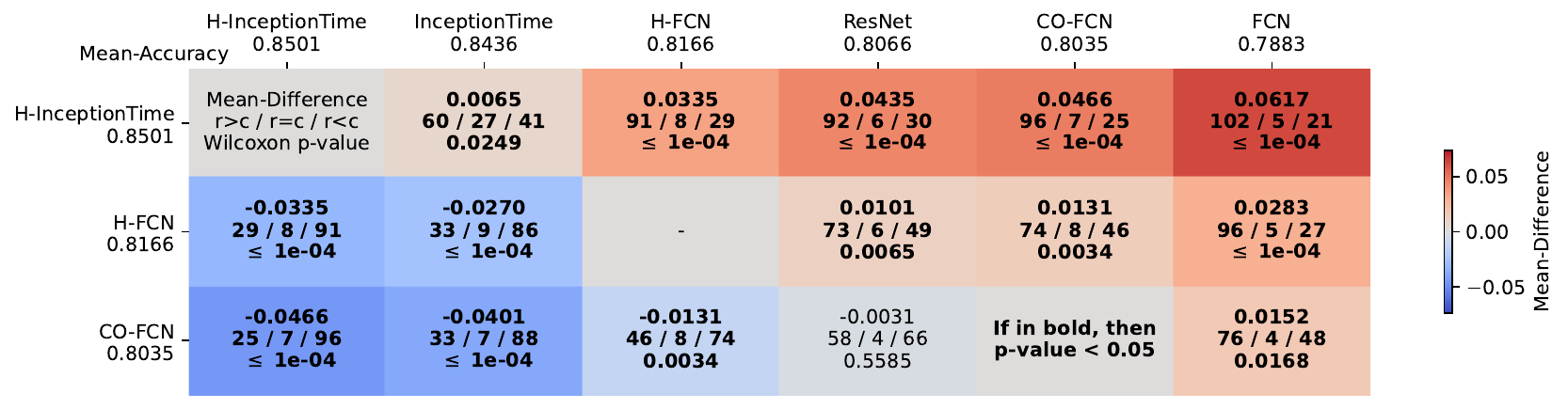}
\end{figure}

For these reasons, in this section and the rest of this thesis, we utilize the MCM (Chapter~\ref{chapitre_2}) for both 1v1 and multiple
model comparisons, as shown in Figure~\ref{fig:hcf-mcm-deep}.
In this MCM, it is clear that CO-FCN significantly outperforms FCN, with no significant difference compared to ResNet.
The MCM also shows that H-FCN outperforms ResNet in terms of average accuracy.
These findings are crucial, as ResNet is a complex model with nearly $500,000$ parameters to train, 
while CO-FCN and H-FCN have $122,496$ and $77,440$ parameters, respectively.
This questions the direct relationship between the number of parameters and performance, indicating that adding 
hand-crafted filters can significantly improve a ``poorly performing'' model like FCN, enabling it to surpass a state-of-the-art model like ResNet.

\subsubsection{Comparison With Non-Deep Models}\label{sec:hcf-results-non-deep}

\begin{figure}
    \centering
    \caption{Multi-Comparison Matrix (Chapter~\ref{chapitre_2}) presenting a comparison between H-InceptionTime, InceptionTime,  
    and two state-of-the-art non-deep learning models for TSC, ROCKET and MultiROCKET, aggregated over the 128
    datasets of the UCR archive.}
    \label{fig:hcf-mcm-non-deep}
    \includegraphics[width=\textwidth]{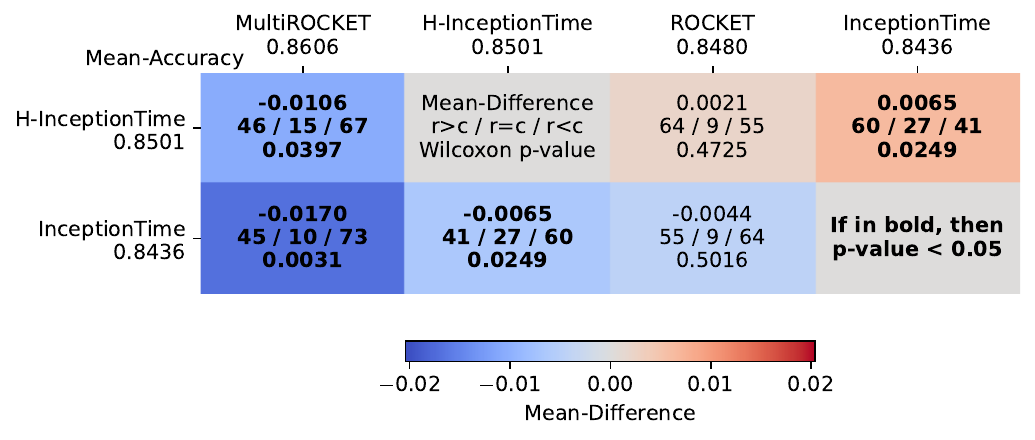}
\end{figure}

This section demonstrates how hand-crafted filters boost InceptionTime to achieve better average accuracy compared to ROCKET and 
approach the performance of MultiROCKET. 
The MCM in Figure~\ref{fig:hcf-mcm-non-deep} highlights that the p-value between H-InceptionTime and MultiROCKET 
is closer to the threshold than the p-value between InceptionTime and MultiROCKET.
We conclude that this performance boost is attributed to the hand-crafted convolution filters.

\subsection{Analysis}

To verify the hypothesis that the original models could identify shapes similar to our proposed filters, we first analyzed if 
the original models were able to find similar patterns. Additionally, we needed to verify that models using 
the hand-crafted filters did not redundantly learn the same filters, as this would be ineffective. Finally, 
we aimed to understand why the hand-crafted filters enhance model performance, specifically whether they help 
the model generalize better. By examining these aspects, we sought to confirm the value of hand-crafted filters 
in capturing critical patterns and improving the model's robustness across different datasets.

\subsubsection{Learned vs Hand-Crafted Convolution Filters}\label{sec:hcf-learned-vs-hcf}

To evaluate the effectiveness of our hand-crafted filters, we compared them to the learned filters in the original models. 
Our primary goal was to determine whether the original models were learning filters similar to our hand-crafted ones. 
In this experiment, we analyzed the 128 filters learned by the first layer of the original FCN model on the CinCECGTorso 
dataset. For each hand-crafted filter, we identified the closest learned filter by computing the DTW distance for each pair, 
after Z-normalizing the learned filters.

\begin{figure}
    \centering
    \includegraphics[width=0.6\textwidth]{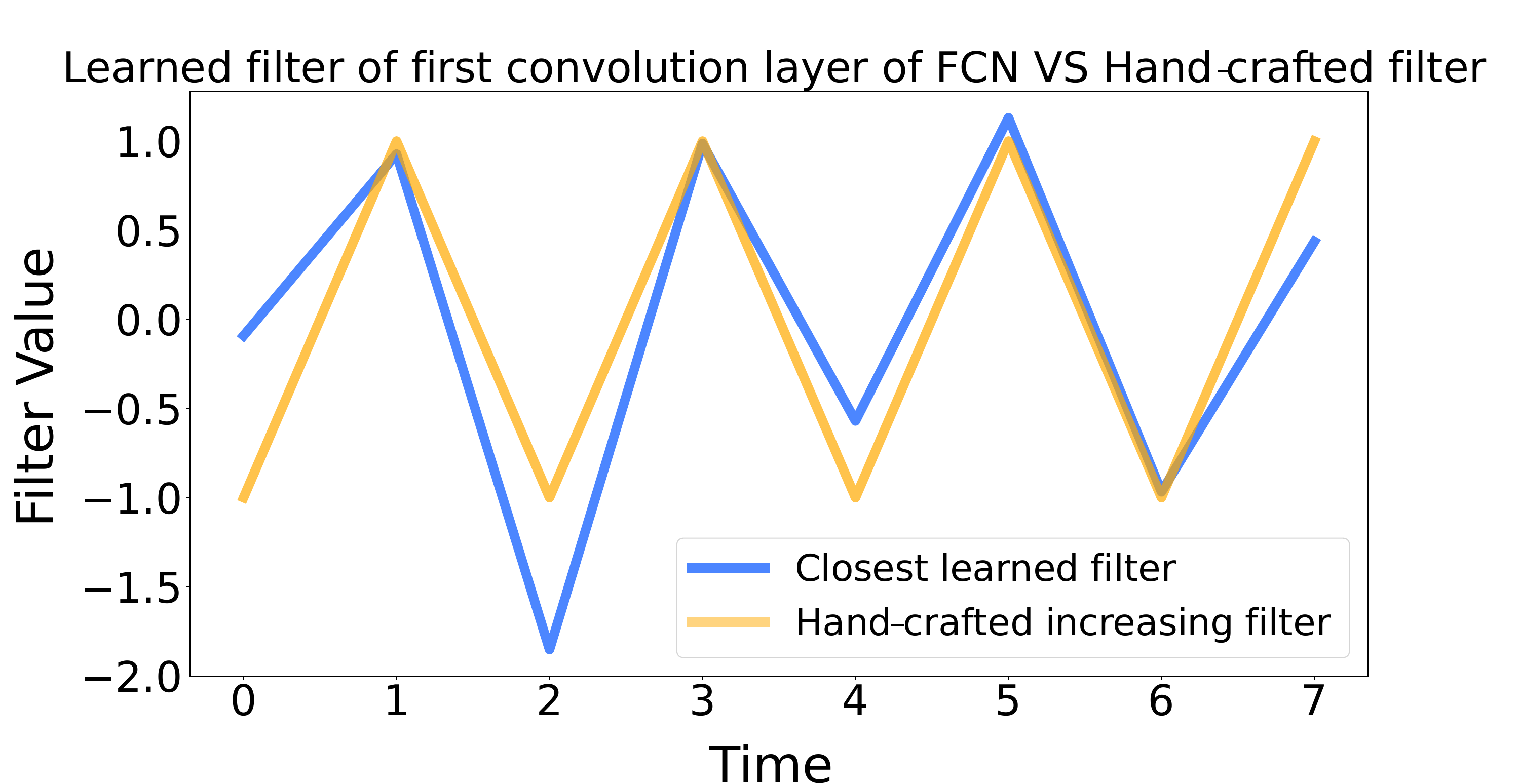}
    \caption{\protect\mycolorbox{0,82,255,0.7}{Hand-crafted increasing trend detection filter} 
    of size $K=8$ and its \protect\mycolorbox{255,169,0,0.7}{closest learned filter} on 
    the CinCECGTorso dataset. The learned filter is from the first layer of the original FCN.}
    \label{fig:hcf-fixed_vs_learned}
\end{figure}

For example, the hand-crafted increasing filter of size $K=8$ and its closest learned filter on the CinCECGTorso 
dataset are shown in Figure~\ref{fig:hcf-fixed_vs_learned}. The original FCN model learned a similar filter for detecting 
increasing trends in the time series. The learned filter is a weighted version of our hand-crafted filter. 
Additionally, the first part of the learned filter resembles the peak detection filter, while the rest matches the 
increasing trend detection. This suggests that the FCN model learned to construct filters capturing multiple patterns 
simultaneously, aligning with our approach in the H-FCN model that combines hand-crafted and learned filters.

\begin{figure}
    \centering
    \includegraphics[width=0.8\textwidth]{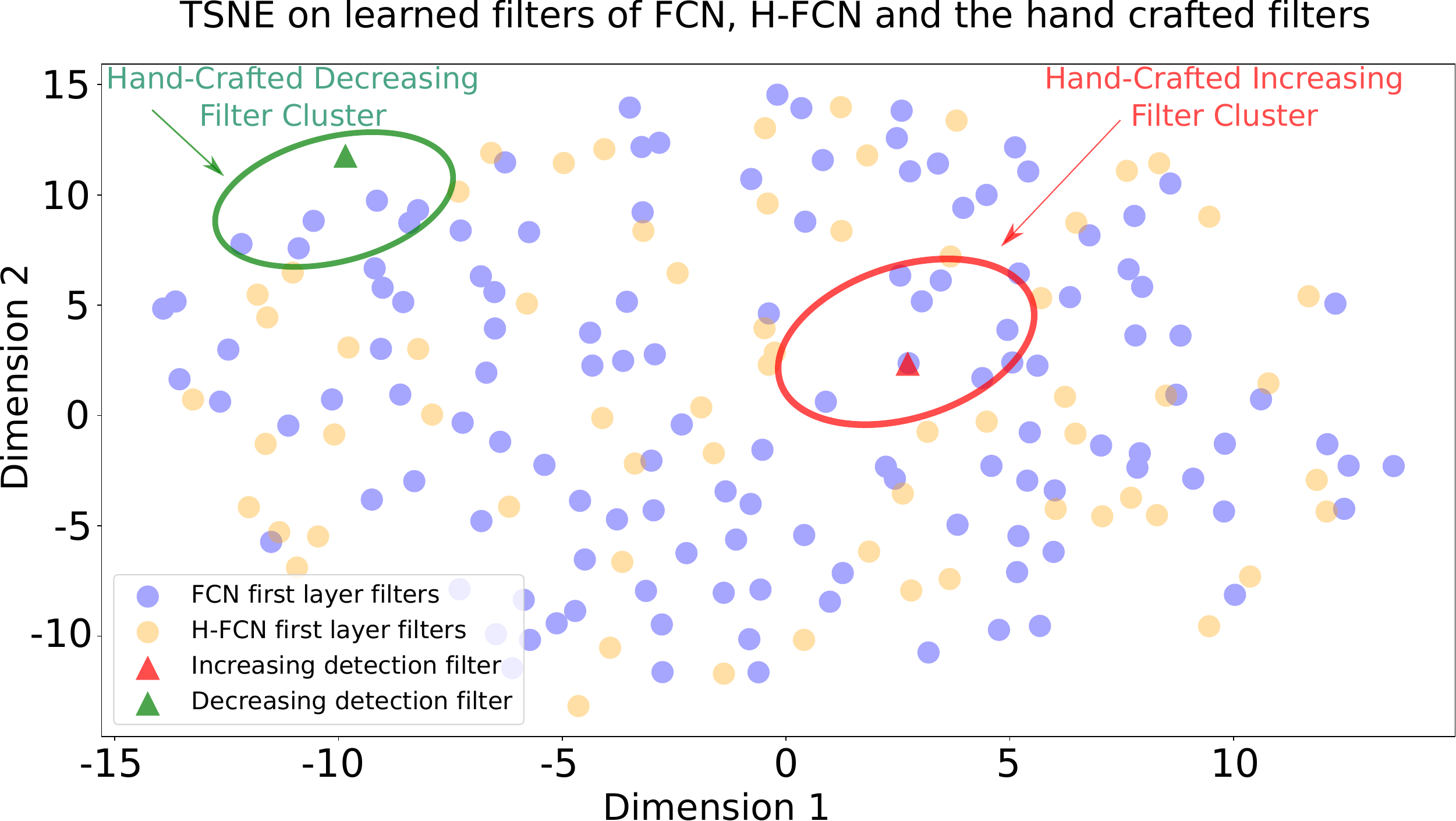}
    \caption{T-SNE two-dimensional projection of the 128 filters 
    learned by the first layer of the original 
    \protect\mycolorbox{0,0,255,0.2}{FCN} and the 64 filters learned by the first layer of the 
    \protect\mycolorbox{255,165,0,0.2}{H-FCN} on the CricketY dataset of the UCR Archive. 
    The two hand-crafted \protect\mycolorbox{255,0,0,0.7}{increasing} and 
    \protect\mycolorbox{0,128,0,0.7}{decreasing} trend 
    detection filters are also projected in the two-dimensional 
    space.}
    \label{fig:hcf-tsne_filters}
\end{figure}

We further assessed the impact of incorporating hand-crafted filters into the H-FCN model. We compared the 64 
learned filters in the first layer of H-FCN with the 128 filters in the first layer of the original FCN, including 
our hand-crafted increasing and decreasing filters. After computing the DTW distance for each pair of filters, 
we projected them into a two-dimensional space using t-SNE~\cite{t-sne-paper}, 
as shown in Figure~\ref{fig:hcf-tsne_filters} on the CricketY dataset.

We can see that the filters learned by H-FCN (in orange) are quite similar to those learned by FCN (in blue). 
However, there are noticeable gaps in the H-FCN filter distribution, marked by red and green ellipsoids. 
These gaps correspond to the hand-crafted increasing filter (red triangle) and decreasing filter (green triangle).
These hand-crafted filters serve as representative prototypes for several FCN learned filters. 
By integrating these hand-crafted filters, H-FCN reduces the number of learned filters needed, allowing 
the model to concentrate on learning other important patterns.

In the following analysis, we examine the model's behavior during training, both with and without our filters, 
to determine if the inclusion of hand-crafted filters improves the model's ability to generalize.

\subsubsection{Do Hand-Crafted Filters Help Models Generalize?}

\begin{figure}
    \centering
    \caption{
        Training phase of the FCN and H-FCN architectures on the 
        FiftyWords dataset while monitoring the validation loss on the test set.
    The \protect\mycolorbox{224,171,172,0.6}{training loss of H-FCN} converges faster than 
    the \protect\mycolorbox{0,128,156,0.6}{training loss of FCN}.
    Also, the \protect\mycolorbox{226,128,156,0.6}{validation loss of H-FCN} is always 
    below the \protect\mycolorbox{121,195,156,0.6}{validation loss of FCN}.
    This shows that H-FCN generalizes better than FCN 
    on this dataset.
    }
    \label{fig:hcf-generalization}
    \includegraphics[width=0.8\textwidth]{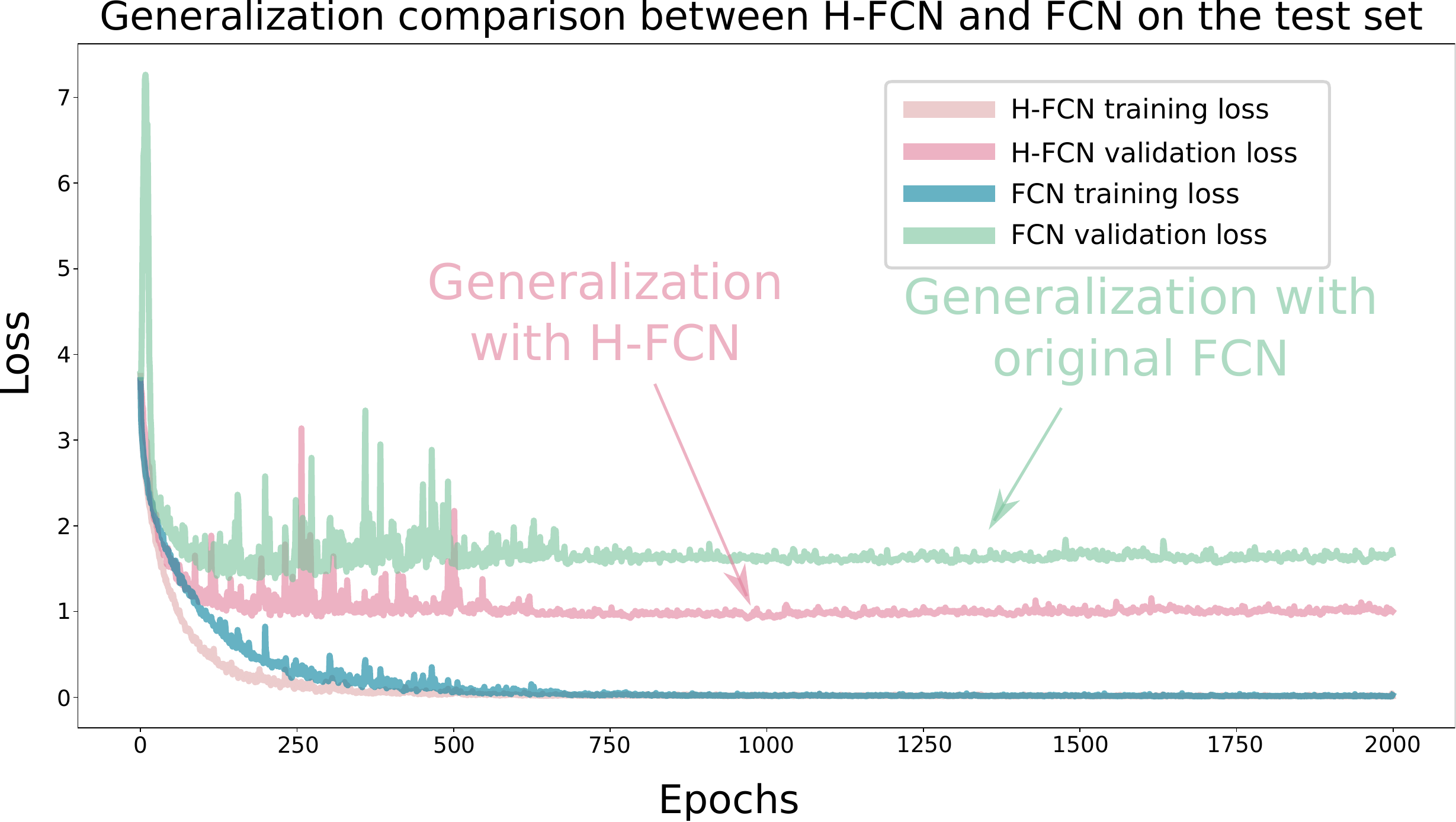}
\end{figure}

Experimental results in Section~\ref{sec:hcf-learned-vs-hcf} demonstrated that adding hand-crafted filters to Deep
Learning models 
improves performance. However, as shown in Section~\ref{sec:hcf-learned-vs-hcf}, some of these hand-crafted filters 
in H-FCN resemble 
filters learned by the original FCN. This raises a critical question: if FCN can independently learn these filters, 
why does including hand-crafted filters in H-FCN enhance accuracy?

The answer likely lies in the superior generalization ability of hand-crafted filters. In models like FCN 
without hand-crafted filters, the optimization process focuses on fitting the training data as closely as possible, 
which increases the risk of overfitting,a common problem in Deep Learning for TSC. Conversely, the inclusion of generic 
hand-crafted filters in H-FCN helps the model capture broader patterns that are more applicable to unseen test data.

To test this hypothesis, we compared the training and validation loss curves for both FCN and H-FCN models using the 
FiftyWords dataset, as illustrated in Figure~\ref{fig:hcf-generalization}.
Since the UCR Archive provides only training and test sets, 
we used the test set to compute the validation loss, solely for monitoring generalization and not for tuning 
hyper-parameters. The results show that the validation loss for H-FCN converges to a significantly lower value 
than that of FCN, indicating better generalization. This explains why H-FCN achieves a $15\%$
higher accuracy on the test set compared to FCN.

\subsubsection{Can We Use Hand-Crafted Filters To Construct Foundation Models?}

In conclusion, hand-crafted filters significantly aid in generalization because they are independent of specific datasets, 
allowing models to capture broad patterns applicable across various data. This approach is a step toward developing 
foundation models, which aim to be versatile and effective across multiple tasks and datasets. The next logical step is 
to explore constructing a foundation model by leveraging the new state-of-the-art model, H-InceptionTime, 
which incorporates hand-crafted filters. This will be discussed in the following section, where we will dig 
into the methodology and potential benefits of using H-InceptionTime as a foundation model.

\section{Finding Foundation Models for Time Series Classification Using A Pretext Task}\label{sec:pretext-task}

In this section, we introduce a foundation model for Time Series Classification (TSC) that leverages the H-Inception 
architecture, incorporating the benefits of hand-crafted filters. The core idea behind foundation models is to develop 
robust, pre-trained models that can generalize across diverse datasets, significantly enhancing the performance 
and efficiency of deep learning models in TSC.

In many real-world applications, starting from scratch with a deep learning model can be a significant downside. 
Collecting large amounts of labeled data is often costly and time-consuming, and in many cases, acquiring such 
extensive datasets is impractical. For instance, in the medical field, gathering sufficient data for training 
models to detect heart diseases from ECG signals involves not only extensive time and financial resources but 
also the expertise of medical professionals to annotate the data accurately. Similarly, in industrial applications 
like predictive maintenance, collecting sensor data from machinery involves prolonged monitoring periods and expert 
annotation to identify failure modes.

This is where pre-trained foundation models prove to be invaluable. By starting with a robust, pre-trained model 
that has already learned generalizable features from a wide array of datasets, we can significantly reduce the 
amount of new data needed. Fine-tuning these models on small, domain-specific datasets can lead to better performance 
without the risk of overfitting, which is a common issue when training from scratch with limited data.

Our foundation model, based on the H-Inception architecture, addresses these challenges effectively.
The H-Inception architecture (see Section~\ref{sec:hinceptiontime}) integrates hand-crafted filters (see
Section~\ref{sec:construction-hcf}), which have shown strong generalization capabilities, 
making them independent of specific datasets. This attribute aligns well with the objective of creating foundation 
models that perform consistently across different data sources.

To construct our foundation model, we benefit from the extensive UCR archive~\cite{ucr-archive},
which includes 128 datasets divided 
into 8 distinct domains, such as ECG, sensor data, and motion. This diverse collection allows us to train a model 
on a common task across datasets within the same domain. By leveraging these varied datasets, the pre-trained model 
learns to identify patterns that are relevant across different, yet related, data sources. This pre-training on a 
broad set of tasks enhances the model's ability to generalize and perform well when fine-tuned on specific datasets 
within each domain.

\begin{figure}
    \centering
    \caption{Summary of the proposed pretext task approach.
    Given an archive of $N$ datasets, the first step is to train a
    \protect\mycolorbox{0,175,255,0.6}{\emph{pre-trained} model}
    on all of the datasets, where the classification task is to 
    \protect\mycolorbox{255,123,0,0.6}{predict
    the dataset each time series belongs to}.
    The second step is to copy the \protect\mycolorbox{0,175,255,0.6}{\emph{pre-trained} model}
    and follow it with an \protect\mycolorbox{0,175,57,0.6}{\emph{addon} 
    model randomly initialized}.
    The second step is done for each of the $N$ datasets of the archive independently.
    After constructing the $N$ new models, they are fine-tuned on each dataset 
    depending on \protect\mycolorbox{255,0,0,0.6}{\textcolor{white}{the task of each one}}.}
    \label{fig:pretext-summary}
    \includegraphics[width=\textwidth]{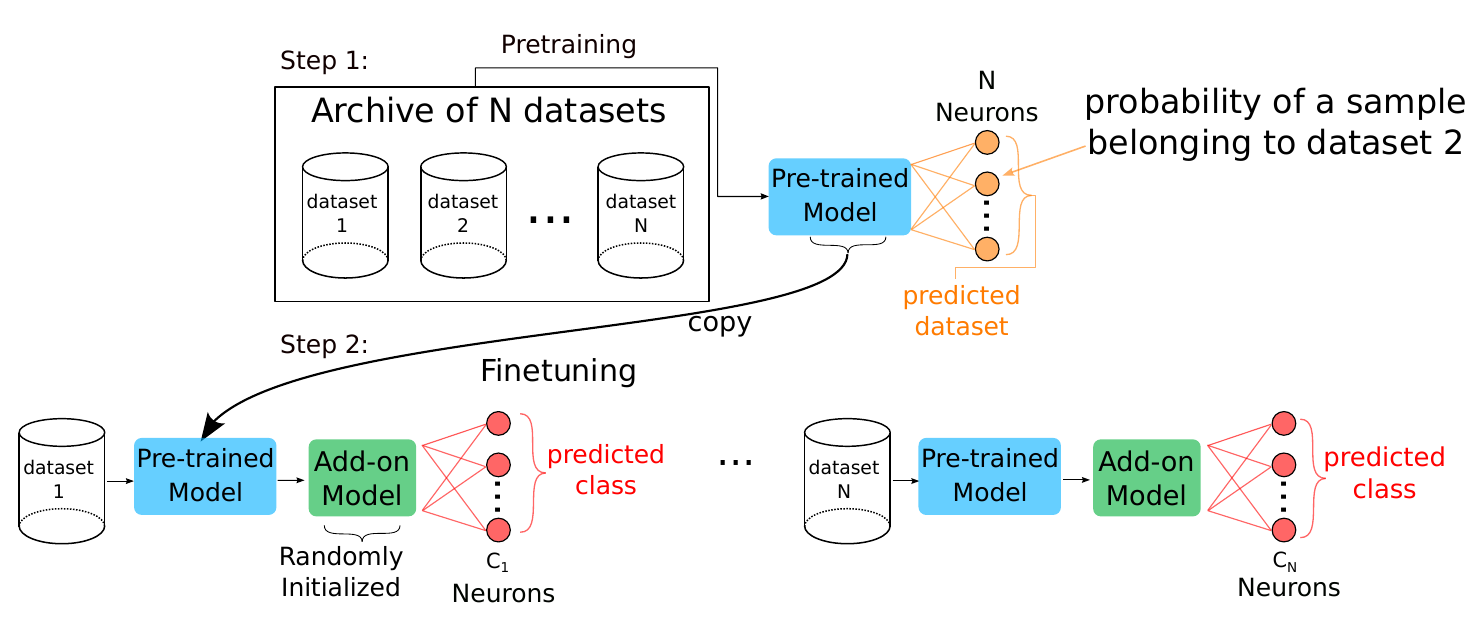}
\end{figure}

The contribution of this part of the chapter is twofold as detailed in Figure~\ref{fig:pretext-summary}: Firstly,
we propose a novel pre-training strategy that involves
a pretext task designed to predict the originating dataset of each time series sample. This approach enables the
model to learn generic features that are applicable across multiple datasets. Secondly, we fine-tune the pre-trained
model on specific datasets, enhancing its ability to adapt to the unique characteristics of each dataset while retaining
the generalized knowledge acquired during pre-training.

After fully training the pre-trained model on the pretext task, we can proceed with fine-tuning through two different 
approaches. The first approach involves directly fine-tuning the pre-trained model, followed by adding a classification 
layer tailored to the dataset's specific task. The second approach fine-tunes the pre-trained model by cascading it with 
additional deeper layers before adding the classification layer, allowing for the extraction of more intricate features. 
Previous research~\cite{transfer-learning-paper} employed the first approach to study transfer learning for TSC,
but the results were suboptimal 
due to target datasets being highly sensitive to the source dataset used. In our work, we opted for the second approach, 
believing it to be more robust. This decision stems from the understanding that the first method assumes the pre-trained 
model has already identified the optimal convolution filters, potentially neglecting deeper, dataset-specific features 
during fine-tuning. By incorporating deeper layers, our approach ensures that the model refines its feature extraction 
capabilities, leading to better generalization and performance across diverse datasets.

\subsection{Foundation Model Architecture Construction}\label{sec:pretext-construction}

Given a backbone deep learning model for TSC (H-Inception in our case)
consisting of $\Lambda$ layers, we divided the model into two distinct sub-models. 
The first sub-model, referred to as the pre-trained model, is designed to learn a pretext task. The second sub-model, 
which is randomly initialized, serves as an extension to the pre-trained model and focuses specifically on the TSC task.

The pretext task selected for this work involves the pre-trained model predicting the dataset of origin for each sample 
from a set of $N$ datasets (see Algorithm~\ref{alg:pretext-step1}). While it might seem more straightforward to combine all
datasets and 
classes into a single large class distribution for prediction, this approach has significant limitations. When there is 
no correlation between classes from different datasets, the combined class distribution would lack meaningful representation. 
Thus, using a pretext task to first train the model ensures a more structured and effective learning process.

\begin{algorithm}
\caption{Train the Pre-Trained Model on pretext Task}
\label{alg:pretext-step1}
\begin{algorithmic}[1]
    \REQUIRE
        $\mathcal{D}=\{\mathcal{D}_1,\mathcal{D}_2\ldots\mathcal{D}_N\}$ N datasets where
        $\mathcal{D}_i=\{\textbf{x}_{ij},y_{ij}\}_{j=1}^{M_i}$, the number of layers for the pre-trained mode $L_{PT}$
    \ENSURE A pre-trained model $PT(.)$ trained on the pretext task over all the datasets in $\mathcal{D}$
    
    \STATE Define $M = sum(M_1,M_2,\ldots,M_{N})$
    \STATE Define $\mathcal{D}_{PT} = empty List$
    \STATE Build $PT(.)$ a neural network with $\Lambda_{PT}$ layers and $M$ output units with $softmax$ activation
    \FOR{$i=1$ to $N$}
        \FOR{$j=1$ to $M_i$}
            \STATE $\mathcal{D}_{PT}.append([\textbf{x}_{ij},i])$
        \ENDFOR
    \ENDFOR
    \STATE $PT.train(\mathcal{D}_{PT})$
    \STATE \textbf{Return:} $PT(.)$
    
\end{algorithmic}
\end{algorithm}

Upon completing the training of the pre-trained model, we enhance it by integrating a randomly initialized sub-model.
This newly constructed composite model, consisting of the pre-trained and the new sub-model, is subsequently
fine-tuned for the TSC task on each dataset independently (refer to Algorithm~\ref{alg:pretext-step2}).

\begin{algorithm}
\caption{Fine Tuning on Each Dataset}
\label{alg:pretext-step2}
\begin{algorithmic}[1]
    \REQUIRE $\mathcal{D}=\{\mathcal{D}_1,\mathcal{D}_2\ldots\mathcal{D}_N\}$ N datasets
    where $\mathcal{D}_i=\{\textbf{x}_{ij},y_{ij}\}_{j=1}^{M_i}$, a pre-trained model $PT(.)$ of $\Lambda_{PT}$ layers
    trained on the pretext task, the number of layers of an addon model while fine tuning $\Lambda_{FT}$
    \ENSURE $\{FT_1(.),FT_2(.),\ldots FT_N(.)\}$ $N$ fine tuned models of $\Lambda_{PT}+\Lambda_{FT}$ layers trained
    on the task of each dataset independently
    
    \STATE Build $\{FT_1(.),FT_2(.),\ldots, FT_N(.)\}$ neural networks of $\Lambda_{PT}+\Lambda_{FT}$ layers with
    output nodes respecting the number of classes of each dataset in $\mathcal{D}$ respectively
    \STATE Fill the first $\Lambda_{PT}$ layers in $\{FT_1(.),FT_2(.),\ldots,FT_N(.)\}$ by the learned parameters from the 
    feature extraction part of $PT(.)$
    \FOR{$i=1$ to $N$}
        \STATE $FT_i.train(\mathcal{D}_i)$
    \ENDFOR
    
    \STATE \textbf{Return:} $\{FT_1(.),FT_2(.),\ldots, FT_N(.)\}$
    
\end{algorithmic}
\end{algorithm}

\subsection{Backbone Selection}\label{sec:pretext-backbone}

\begin{figure}
    \centering
    \caption{
        The architecture of H-Inception divided into two sub-models.
    The \protect\mycolorbox{0,121,30,0.6}{first model} 
    is the pre-trained model, trained on the pretext 
    task, while the second model is the \protect\mycolorbox{255,0,0,0.6}{randomly initialized 
    add-on model}.
    The H-Inception model is made of six Inception modules, 
    where each module contains three \protect\mycolorbox{255,160,0,1.0}{convolution layers}
    and a \protect\mycolorbox{252,0,190,1.0}{Max Pooling layer} followed by a
    \protect\mycolorbox{255,221,55,1.0}{concatenation}, a
    \protect\mycolorbox{210,221,55,1.0}{batch normalization layer}
    and an \protect\mycolorbox{255,55,100,1.0}{activation function}.
    Each Inception module, except the first one, is proceeded 
    by a \protect\mycolorbox{240,148,255,1.0}{bottleneck layer}
    to reduce the dimensionality and hence the number of parameters.
    The first Inception module contains the hybrid addition, which 
    is the \protect\mycolorbox{0,232,132,1.0}{hand-crafted convolution filter}.
    \protect\mycolorbox{0,212,247,1.0}{Residual connections} exist between 
    the input and the third module, as well as between the third module
    and the output.
    }
    \label{fig:pretext-archi}
    \includegraphics[width=\textwidth]{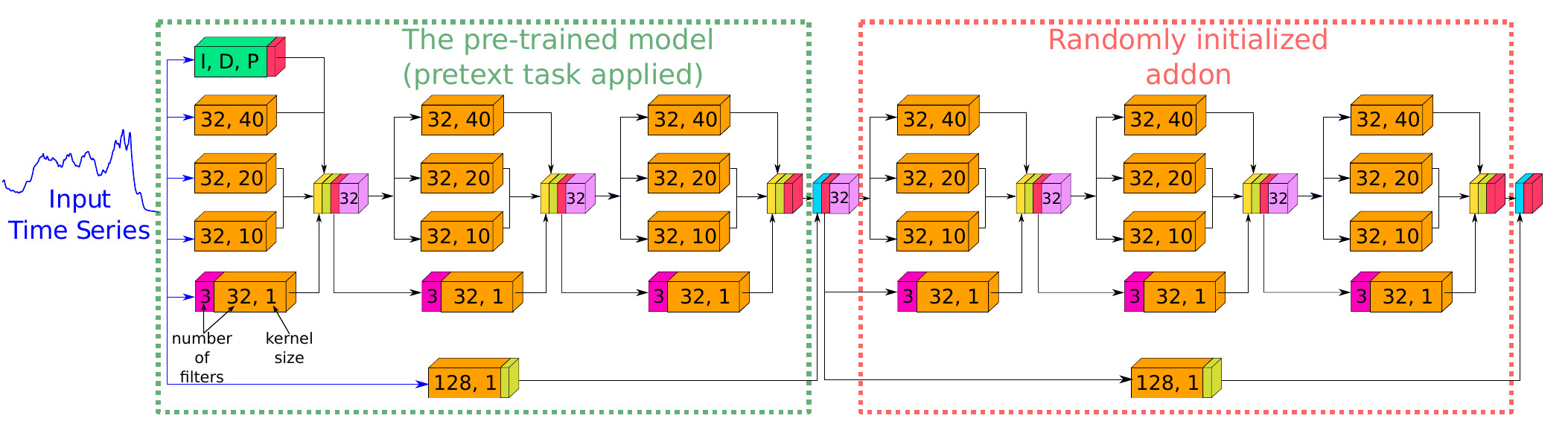}
\end{figure}

Our model is based on the state-of-the-art deep learning architecture for TSC, the H-Inception network
(Section~\ref{sec:hinceptiontime}).
Figure~\ref{fig:pretext-archi} illustrates how the H-Inception backbone is divided for our approach. The original H-Inception
architecture has six Inception modules. We designate the first three modules for the pre-trained model
and the last three for the fine-tuning phase.
H-InceptionTime is an ensemble of five H-Inception models with different initializations.
Thus, we adopt the H-Inception architecture as our backbone and use model ensemble, consistent with the original
works~\cite{inceptiontime-paper}.
We call this approach Pre-trained H-InceptionTime (PHIT).

\subsection{Does It Make Sense To Use Batch Normalization On Different Datasets?}\label{sec:pretext-batchnorm}

Most cutting-edge deep learning models for TSC~\cite{dl4tsc}, which excel on the UCR archive~\cite{ucr-archive},
are convolution-based
architectures that utilize Batch Normalization layers (Chapter~\ref{chapitre_1} Section~\ref{sec:tsc-deep})
to speed up training. In the H-Inception model we selected, each convolution layer is followed
by Batch Normalization. This process adjusts the batch samples to achieve 
zero mean and unit variance. However, this approach can be problematic when the batch samples come from different 
distributions, such as different datasets, which is the scenario for our pre-trained model.

To mitigate this issue, we introduce multiple Batch Normalization layers, each dedicated to a specific dataset,
instead of the single Batch Normalization layer typically used in CNN architectures for TSC. This setup requires
the model to appropriately connect each sample in the batch to its corresponding Batch Normalization layer.

\begin{figure}
    \centering
    \caption{
        An example using the proposed
        \protect\mycolorbox{0,164,0,0.6}{Batch Normalizing Multiplexer (BNM)} 
        that solves the problem of 
        learning a batch normalization layer on multiple samples of 
        different distributions (datasets).
    The BNM is made of \protect\mycolorbox{210,221,55,1.0}{multiple batch normalization layers} 
    (with \protect\mycolorbox{0,0,255,1.0}{\textcolor{white}{blue}}
    and \protect\mycolorbox{255,0,0,1.0}{\textcolor{white}{red}} contours) 
    proceeded by a multiplexer.
    This multiplexer has three different nodes: (a) input node, 
    where the input time series goes through, 
    (b) the control node, where the information about the dataset this 
    input time series belong to goes through, 
    and (c) the output node.
    The path selected for the output node is controlled by the node (b).
    It is important to note that the BNM, such as the traditional batch 
    normalization layer, learns on the whole batch.
    The only difference is that more than one batch normalization 
    layer will be fed by parts of this batch, which 
    intuitively means the flow of information is slower when using the BNM.
    }
    \label{fig:pretext-batchnorm}
    \includegraphics[width=\textwidth]{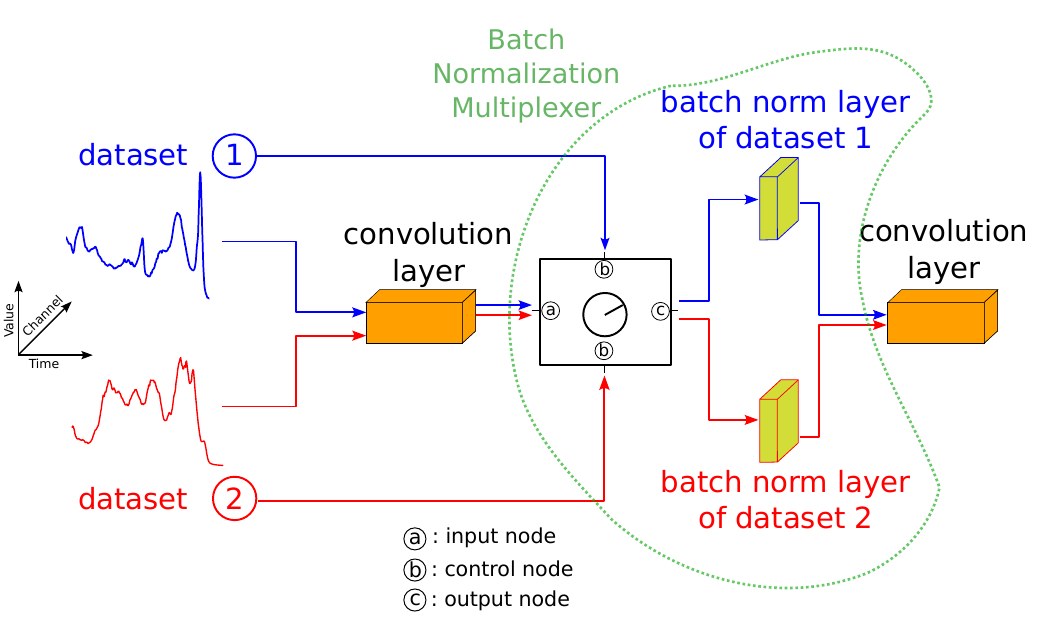}
\end{figure}

Our innovative Batch Normalization Multiplexer (BNM) is depicted in Figure~\ref{fig:pretext-batchnorm}.
The BNM takes as input the output
from the preceding layer, along with the dataset information for each series being processed. This dataset information,
which the model is also attempting to predict, guides the control node of the BNM to select the correct Batch
Normalization layer for the output node. This design ensures that proper normalization is applied, even when dealing
with diverse datasets, thereby enhancing the pre-trained model's robustness and performance.

\subsection{Experimental Setup}

\subsubsection{Datasets}
To evaluate the performance of our proposed approach, we conducted a series of experiments using the UCR archive dataset,
which consists of 128 datasets. Due to redundancies, we narrowed our study to 88 datasets. For example, some datasets 
appear multiple times with different train-test splits for various classification tasks, which could interfere with our 
model's objective of predicting the source dataset of a sample. Including identical series from different datasets in the 
training set could confuse the model. Additionally, some datasets were excluded because they only varied in class counts 
or were truncated versions of others. A detailed explanation of the exclusions is provided in Table~\ref{tab:pretext-ucr-not-used}.

To ensure consistency, all datasets were z-normalized before training, achieving a zero mean and unit variance. 
Since the sample lengths varied, we applied zero padding within each batch (instead of before training) to match 
the length of the longest series in that batch. This approach maintains the integrity of the model's training and 
evaluation processes.

\begin{table*}
    \centering
    \caption{Excluded datasets from the UCR archive in this study.
    Each dataaset is followed by its information and a reason for its exclusion.
    }
    \label{tab:pretext-ucr-not-used}
    \scriptsize
    \resizebox{\columnwidth}{!}{
    \begin{tabular}{|c|c|c|c|c|c|}
    \hline
    \textbf{Type} &
      \textbf{Dataset} &
      \textbf{Train Samples} &
      \textbf{Test Samples} &
      \textbf{Length} &
      \textbf{reason (if excluded)} \\ \hline
    EOG &
      EOGHorizontalSignal &
      362 &
      362 &
      1250 &
      \begin{tabular}[c]{@{}c@{}}same datasets multivariate\\ with 2 channels divided\\ into 2 univariate datasets\end{tabular} \\ \cline{2-5}
     &
      EOGVerticalSignal &
      362 &
      362 &
      1250 &
       \\ \hline
    EPG &
      InsectEPGRegularTrain &
      62 &
      249 &
      601 &
      \begin{tabular}[c]{@{}c@{}}same test set, different\\ train set size, a combination\\ of both train sets is better\\ than doing a pretext task\end{tabular} \\ \cline{2-5}
     &
      InsectEPGSmallTrain &
      17 &
      249 &
      601 &
       \\ \hline
     &
      PigAirwayPressure &
      104 &
      208 &
      2000 &
       \\ \cline{2-5}
    Hemodynamics &
      PigArtPressure &
      104 &
      208 &
      2000 &
      \begin{tabular}[c]{@{}c@{}}correlation unclear between\\ these three datasets\end{tabular} \\ \cline{2-5}
     &
      PigCVP &
      104 &
      208 &
      2000 &
       \\ \hline
    HRM &
      Fungi &
      18 &
      186 &
      201 &
      Only one dataset in this type \\ \hline
     &
      DistalPhalanxOutlineAgeGroup &
      400 &
      139 &
      80 &
      \begin{tabular}[c]{@{}c@{}}same samples as DistalPhalanxTW\\ with different classification and\\ train test split\end{tabular} \\ \cline{2-5}
     &
      DistalPhalanxOutlineCorrect &
      600 &
      276 &
      80 &
       \\ \cline{2-6} 
     &
      FaceAll &
      560 &
      1690 &
      131 &
      \begin{tabular}[c]{@{}c@{}}same as FacesUCR with different\\ train test split\end{tabular} \\ \cline{2-6} 
     &
      FiftyWords &
      450 &
      455 &
      270 &
      \begin{tabular}[c]{@{}c@{}}same as WordSynonyms with\\ more classes\end{tabular} \\ \cline{2-6} 
    Image &
      MiddlePhalanxOutlineAgeGroup &
      400 &
      154 &
      80 &
      Same reason as DistalPhalanx \\ \cline{2-5}
     &
      MiddlePhalanxOutlineCorrect &
      600 &
      291 &
      80 &
       \\ \cline{2-6} 
     &
      ProximalPhalanxOutlineAgeGroup &
      400 &
      205 &
      80 &
      Same reason as DistalPhalanx \\ \cline{2-5}
     &
      ProximalPhalanxOutlineCorrect &
      600 &
      291 &
      80 &
       \\ \cline{2-6} 
     &
      MixedShapesRegularTrain &
      500 &
      2425 &
      1024 &
      \begin{tabular}[c]{@{}c@{}}Bigger version of\\ MixedShapesSmallTrain\end{tabular} \\ \cline{2-6} \hline
     &
      GunPoint &
      50 &
      150 &
      150 &
      \begin{tabular}[c]{@{}c@{}}GunPointAgeSpan is the new\\ version with more samples\end{tabular} \\ \cline{2-6} 
     Motion &
      WormsTwoClass &
      181 &
      77 &
      900 &
      \begin{tabular}[c]{@{}c@{}}Same as Worms with different\\ number of classes\end{tabular} \\ \cline{2-6}
     &
      GunPointMaleVersusFemale &
      135 &
      316 &
      150 &
      \begin{tabular}[c]{@{}c@{}}Same as AgeSpan version with\\ different train test split\end{tabular} \\ \cline{2-5}
     &
      GunPointOldVersusYoung &
      136 &
      315 &
      150 &
       \\ \hline
    Power &
      PowerCons &
      180 &
      180 &
      144 &
      Only one dataset for this type \\ \hline
     &
      AllGestureWiimoteX &
      300 &
      700 &
      Vary &
       \\ \cline{2-5}
     &
      AllGestureWiimoteY &
      300 &
      700 &
      Vary &
      \begin{tabular}[c]{@{}c@{}}too much Variable length datasets\\ to handle in this type for the pretext\\ task which already has the variable\\ length issue/instability\end{tabular} \\ \cline{2-5}
     &
      AllGestureWiimoteZ &
      300 &
      700 &
      Vary &
       \\ \cline{2-6} 
     &
      DodgerLoopDay &
      78 &
      80 &
      288 &
       \\ \cline{2-5}
     &
      DodgerLoopGame &
      20 &
      138 &
      288 &
      \begin{tabular}[c]{@{}c@{}}All dodger datasets are the same\\ with different train test split\\ with too many missing values\end{tabular} \\ \cline{2-5}
    Sensors &
      DodgerLoopWeekend &
      20 &
      138 &
      288 &
       \\ \cline{2-6} 
     &
      FreezerRegularTrain &
      150 &
      2850 &
      301 &
      \begin{tabular}[c]{@{}c@{}}Same as FreezerSmallTrain with\\ more training examples\end{tabular} \\ \cline{2-6} 
     &
      GesturePebbleZ1 &
      132 &
      172 &
      Vary &
       \\ \cline{2-5}
     &
      GesturePebbleZ2 &
      146 &
      158 &
      Vary &
      \begin{tabular}[c]{@{}c@{}}too much Variable length datasets\\ to handle in this type for the pretext\\ task which already has the variable\\ length issue/instability\end{tabular} \\ \cline{2-5}
     &
      PickupGestureWiimoteZ &
      50 &
      50 &
      Vary &
       \\ \cline{2-5}
     &
      ShakeGestureWiimoteZ &
      50 &
      50 &
      Vary &
       \\ \hline
     &
      Rock &
      20 &
      50 &
      2844 &
      Not a time series \\ \cline{2-6} 
     &
      SemgHandGenderCh2 &
      300 &
      600 &
      1500 &
       \\ \cline{2-5}
    Spectrum &
      SemgHandMovementCh2 &
      450 &
      450 &
      1500 &
      \begin{tabular}[c]{@{}c@{}}Same datasets different split\\ if we include one of them we end up\\ with one dataset for this type\end{tabular} \\ \cline{2-5}
     &
      SemgHandSubjectCh2 &
      450 &
      450 &
      1500 &
       \\ \hline
     &
      GestureMidAirD1 &
      208 &
      130 &
      Vary &
       \\ \cline{2-5}
    Trajectory &
      GestureMidAirD2 &
      208 &
      130 &
      Vary &
      \begin{tabular}[c]{@{}c@{}}Only datasets of variable length.\\ The three datasets are from the same\\ distribution of a 3D multivariate\\ time series with each being a dimension\\ a more suitable approach is to combine\\ the three datasets and solve\\ a multivariate TSC task\end{tabular} \\ \cline{2-5}
     &
      GestureMidAirD3 &
      208 &
      130 &
      Vary &
       \\ \hline
\end{tabular}
    }
\end{table*}

\subsubsection{Division of the Datasets into Types}

The goal of using a pre-trained model is to boost the performance of deep learning classifiers on small datasets 
by utilizing knowledge gained from larger datasets. This strategy is particularly effective when there is some 
shared basic information between the large and small datasets. To explore this, we conducted eight different 
pretext experiments, each corresponding to a different type of dataset in the UCR archive. For each experiment, 
we trained a pre-trained model using all datasets of a specific type, such as ECG, and then fine-tuned the model 
on each dataset individually. The eight dataset types and their corresponding numbers of datasets are as follows:

\begin{itemize} 
    \item Electrocardiogram (ECG): 7 datasets
    \item Sensors: 18 datasets
    \item Devices: 9 datasets
    \item Simulation: 8 datasets
    \item Spectrogram: 8 datasets
    \item Motion: 13 datasets
    \item Traffic: 2 datasets
    \item Image Contours: 23 datasets
\end{itemize}

\subsubsection{Implementation Details}

We maintained the same parameters for the H-Inception model as in the original study (first contribution of this chapter).
Each experiment 
was conducted with five different initializations, covering both the pre-trained and fine-tuned models. 
We aggregated the results from these multiple runs, selecting the best-performing model based on training loss for evaluation.

To optimize training, we utilized a learning rate decay with the ReduceLROnPlateau function in Keras~\cite{keras-book}, 
which halves the learning rate when the training loss stabilizes. All models were trained with a batch 
size of 64. Both the pre-trained and fine-tuned models were trained for 750 epochs each, ensuring that 
the total training duration did not exceed the 1500 epochs used for the baseline model in the original
study (first contribution in this chapter).


\subsection{Experimental Results}

In this section, we present the results of PHIT compared to the baseline model, followed by a comparison with 
state-of-the-art deep and non-deep models for TSC on the UCR archive.

\subsubsection{Comparing With Baseline With No Pre-Training}

\begin{figure}
    \centering
    \caption{
        A 1v1 scatter plot that compares the performance of H-InceptionTime (baseline) and PHIT following the accuracy
metric.
    }
    \label{fig:pretext-1v1-baseline}
    \includegraphics[width=0.5\textwidth]{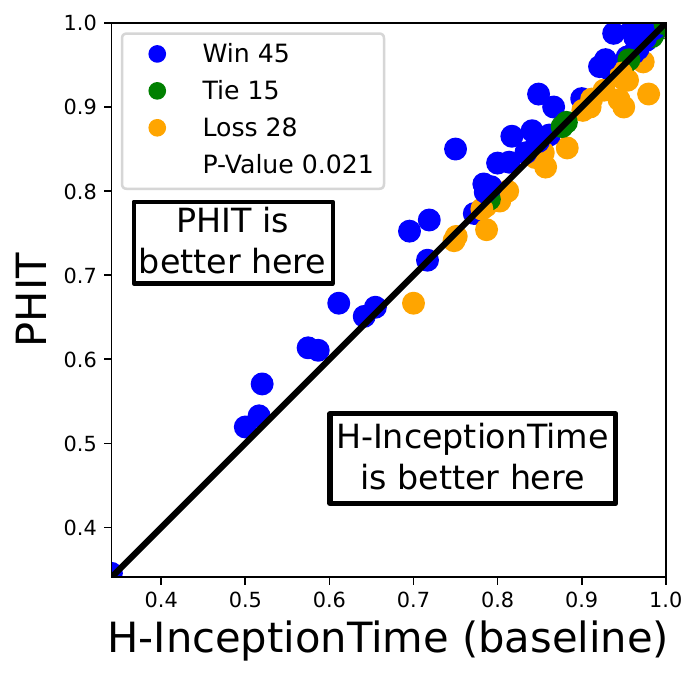}
\end{figure}

In this section, we present a direct comparison between our pre-training approach using the H-Inception 
architecture and the baseline model, both evaluated in their ensemble forms.
Figure~\ref{fig:pretext-1v1-baseline} illustrates this comparison with a scatter 1v1 plot.
The x-axis indicates the accuracy of H-InceptionTime, while the y-axis shows the accuracy of PHIT, 
both measured on the test sets.
Our findings show that PHIT outperforms on average the baseline across 88 datasets, with PHIT achieving higher accuracy in 
48 datasets compared to the baseline's 23. To assess the statistical significance of this difference, we used 
the Wilcoxon Signed-Rank Test to produce a p-value reflecting the confidence level of the performance 
difference. With a p-value of approximately $0.021 < 0.05$, it is clear that PHIT significantly outperforms the baseline.
This demonstrates that the pre-trained model was able to generalize to the test set better, a result that 
we will analyze in more detail in Section~\ref{sec:pretext-analyze}.

\subsubsection{Comparing To State-Of-The-Art}

\begin{figure}
    \centering
    \caption{
        A Multi-Comparison Matrix (MCM) representing the comparison between the proposed approach PHIT with the
state-of-the-art approaches. The winning approach following the average performance is MultiROCKET and in second comes
our approach. No conclusion can be found on the difference of performance between MultiROCKET and PHIT given the
high p-value.
    }
    \label{fig:pretext-mcm-row}
    \includegraphics[width=\textwidth]{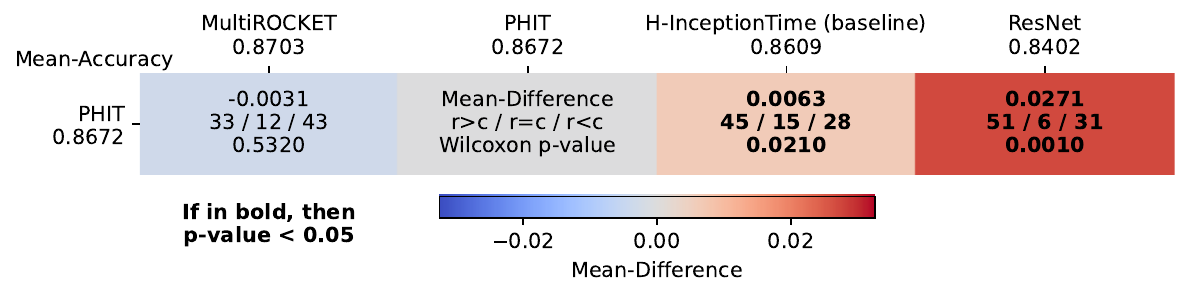}
\end{figure}

Figure~\ref{fig:pretext-mcm-row} presents the MCM (Chapter~\ref{chapitre_2} Section~\ref{sec:MCM})
comparing PHIT with state-of-the-art approaches, encompassing both deep and non-deep 
learning models. The results demonstrate that PHIT outperforms all deep learning approaches based on the average 
performance metric across the 88 datasets in the UCR archive. Moreover, the MCM also reveals that there is no 
statistically significant difference in performance between PHIT and the state-of-the-art MultiROCKET model, 
indicating that while PHIT shows strong performance, it matches rather than surpasses MultiROCKET in statistical significance.

\subsection{Analysis}\label{sec:pretext-analyze}

In this section, we present a detailed analysis aimed at understanding why the pre-training phase was able to 
outperform the baseline model. We achieve this by examining the performance differences between the pre-trained 
model and the baseline across various domains. Additionally, we analyze the impact of training set size for 
each dataset within these domains. Finally, we visualize the filter space to observe the effects of training 
for the three methods: baseline, pre-trained, and fine-tuned.

\subsubsection{Analysing Performance Per Domain}\label{sec:pretext-per-domain}

\begin{table}
    \centering
    \caption{The Win/Tie/Loss count between the proposed PHIT approach and the baseline (H-InceptionTime) per dataset domain.
    The first column presents the number of datasets included per domain followed by the number of Wins for PHIT, number of Ties, and number of Wins for the baseline.
    We include as well the percentage of number of losses and the average difference in accuracy (PHIT - baseline).
    A positive value in the last column indicates that on average of all datasets in a specific domain, PHIT performs better than the baseline on the accuracy metric (lowest value 0.0 and highest value 1.0).}
    \label{tab:pretext-wins-per-type}
    \resizebox{\columnwidth}{!}{
    \begin{tabular}{|c|c|c|c|c|c|c|}
    \hline
    \scriptsize
    \textbf{\begin{tabular}[c]{@{}c@{}}Dataset\\ Type\end{tabular}} &
      \textbf{\begin{tabular}[c]{@{}c@{}}Number of\\ Datasets\end{tabular}} &
      \textbf{\begin{tabular}[c]{@{}c@{}}Wins of\\ PHIT\end{tabular}} &
      \textbf{\begin{tabular}[c]{@{}c@{}}Ties of\\ PHIT\end{tabular}} &
      \textbf{\begin{tabular}[c]{@{}c@{}}Losses of\\ PHIT\end{tabular}} &
      \textbf{\begin{tabular}[c]{@{}c@{}}Difference in\\ Average Accuracy\\ (PHIT - Baseline)\end{tabular}} &
      \begin{tabular}[c]{@{}c@{}}\textbf{Percentage}\\ \textbf{of Losses}\end{tabular} \\ \hline
    \textbf{Devices}    & 9  & 4  & 0 & \textbf{5} & +0.0046 & \textbf{55.55 \%} \\ \hline
    \textbf{ECG}        & 7  & \textbf{3}  & 2 & 2 & +0.0012 & \textbf{28.57 \%} \\ \hline
    \textbf{Images}     & 23 & \textbf{14} & 2 & 7 & +0.0087 & \textbf{30.43 \%} \\ \hline
    \textbf{Motion}     & 13 & \textbf{11} & 1 & 1 & +0.0179 & \textbf{07.69 \%} \\ \hline
    \textbf{Sensors}    & 18 & \textbf{7}  & 5 & 6 & +0.0002 & \textbf{33.33 \%} \\ \hline
    \textbf{Simulation} & 8  & \textbf{3}  & \textbf{3} & 2 & +0.0051 & \textbf{25.00 \%} \\ \hline
    \textbf{Spectro}    & 8  & \textbf{3}  & 2 & \textbf{3} & +0.0115 & \textbf{37.50 \%} \\ \hline
    \textbf{Traffic}    & 2  & 0  & 0 & 2 & -0.0333 & \textbf{100.0 \%} \\ \hline
    \end{tabular}
    }
\end{table}

In Table~\ref{tab:pretext-wins-per-type}, we provide a comprehensive analysis of the PHIT approach's 
performance compared to the baseline 
across different dataset domains. For each domain in the UCR archive, we list the total number of datasets, 
the Win/Tie/Loss count, and the average difference in performance in the final column. A positive value in 
this column indicates that, on average, PHIT surpasses the baseline in terms of accuracy. Additionally, the 
last column shows the percentage of datasets where PHIT performed worse than the baseline.

The table reveals that the percentage of losses for PHIT exceeds $50\%$ in only two instances, and the 
average performance difference is positive for all domains except Traffic, which contains only two datasets. 
These results highlight that PHIT generally outperforms the baseline across most domains in the UCR archive.

This analysis demonstrates that fine-tuning a pre-trained model on a common task shared by multiple datasets 
is significantly more effective than the traditional approach. In the following section, we will investigate deeper 
into the scenarios where the pre-trained model outperforms the baseline by examining the size of the training sets.

\subsubsection{Larger Datasets Helping Smaller Datasets}\label{sec:pretext-large-help-small}

\begin{figure}
    \centering
    \caption{
        Comparing the performance of the proposed approach and its change 
        with respect to the training set size. The
curve represents the \protect\mycolorbox{0,0,255,0.6}{\textcolor{white}{difference in 
performance between the proposed approach and the baseline}}. A positive value
(above the \protect\mycolorbox{255,0,0,0.6}{tie line})
represents a win for the pre-training approach. For each plot, 
we show this comparison on the datasets of the same type in the
UCR archive. The $x$-axis represents the number of training 
examples (in $\log_{10}$ scale). The y-axis represents the difference
of accuracy between the usage of our pre-training approach and the baseline.
    }
    \label{fig:pretext-per-training-size}
    \includegraphics[width=\textwidth]{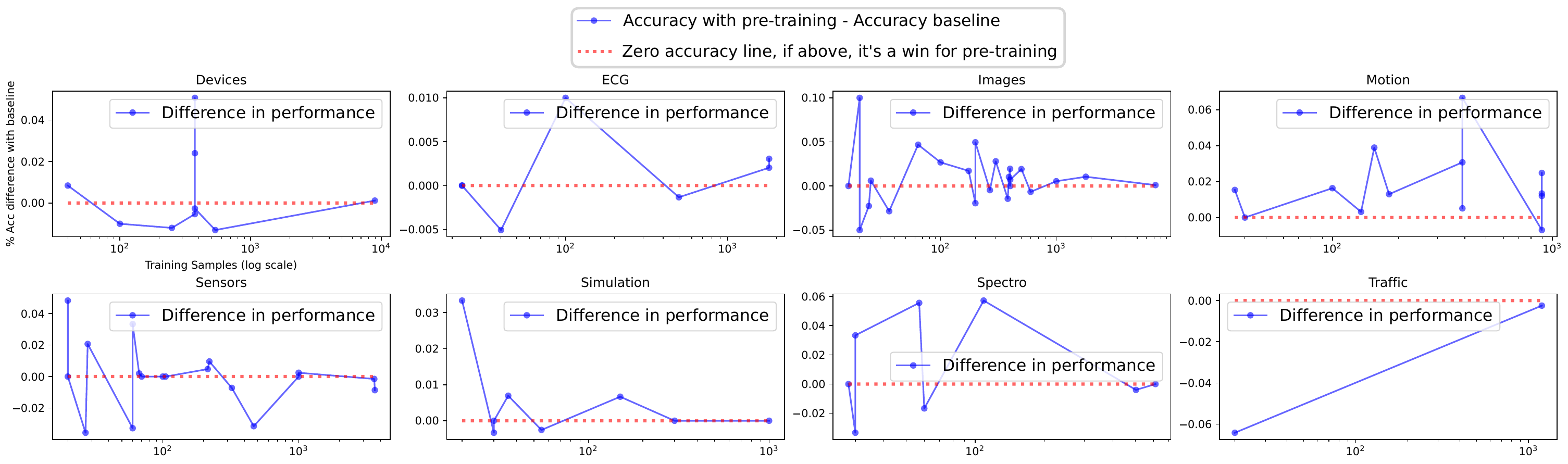}
\end{figure}

As outlined earlier (Section~\ref{sec:pretext-task}), the primary aim of the pretext task is to
enhance the performance of deep learning models on 
TSC tasks, particularly when faced with datasets that have a limited number of training samples. In this section, 
we explore the effect of the pretext task on each of the 8 dataset types, considering the number of training samples 
available. This analysis is illustrated in Figure~\ref{fig:pretext-per-training-size},
where the $y$-axis represents the difference in accuracy between 
PHIT and the baseline, and the $x$-axis (in $\log_{10}$ scale) denotes the training set size. The study is presented across 8 
distinct plots, one for each dataset type. Positive values in the blue curves indicate that PHIT outperforms the baseline.

Our observations reveal that, on average, the pretext task significantly benefits datasets with fewer than $10^3$ 
training samples. This is evident in most cases, though not in every case. We hypothesize that this effect arises 
because the pretext task allows the model to glean more knowledge from larger datasets, which can then be effectively 
transferred to smaller ones. This transfer process provides the fine-tuning stage with rich, informative data for 
small datasets while introducing a degree of noise for larger ones. Larger datasets require the model's full attention 
on their specific tasks, whereas smaller datasets gain a crucial boost from the additional information, which the model 
alone might struggle to learn without external guidance.

\subsubsection{Analyzing The Filters Space}\label{sec:pretext-filters}

\begin{figure}
    \centering
    \caption{
        A two dimensional representation of the filters coming 
        from the \protect\mycolorbox{0,62,255,0.6}{first Inception module of the baseline}, 
        \protect\mycolorbox{255,16,0,0.6}{pre-trained} and 
        \protect\mycolorbox{0,127,0,0.6}{fine tuned} models.
    The used datasets in this study are ECG200 (left) 
    and NonInvasiveFetalECGThorax1 (right).
    The two dimensional representation is done using $t$-SNE coupled with DTW to 
    as a distance measure.
    Some \protect\mycolorbox{255,100,255,0.6}{areas} can be seen to be in common 
    between the three models in the case of large datasets (right) however it is not 
    the case for small datasets (left).
    }
    \label{fig:pretext-filters}
    \includegraphics[width=\textwidth]{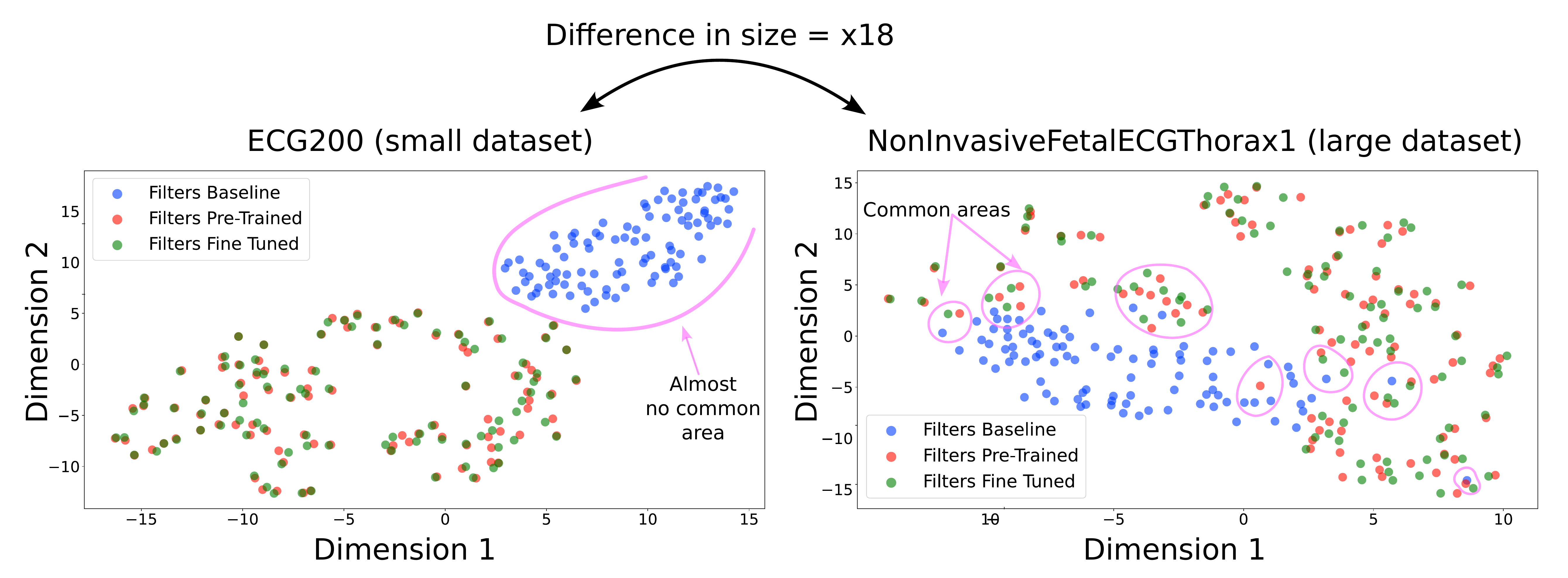}
\end{figure}

Given our focus on CNNs, we can compare the learned filter spaces to understand the impact of our pre-training approach. 
To visualize this, we employed the t-SNE technique~\cite{t-sne-paper}, reducing the 
dimensionality of the filters to a two-dimensional space. We used DTW for the t-SNE technique as explained before, 
to have a shift-invariant two-dimensional projection.
We visualized the filters from the first Inception module of the baseline, pre-trained, and fine-tuned models in 
Figure~\ref{fig:pretext-filters}, focusing on the ECG datasets: ECG200 and NonInvasiveFetalECGThorax1 from the UCR 
archive~\cite{ucr-archive}.
These datasets were chosen deliberately 
due to their differing training set sizes, with ECG200 having $100$ training examples and
NonInvasiveFetalECGThorax1 having $1800$.

Figure~\ref{fig:pretext-filters} displays the filter distributions for the baseline, pre-trained,
and fine-tuned models for each dataset. 
One prominent observation is that the blue points (baseline filters) are markedly different from the red and 
green points (pre-trained and fine-tuned filters). This demonstrates that the pre-training followed by fine-tuning 
leads to the learning of different convolution filters compared to the traditional baseline approach.
Another key observation is the variation between the two plots. For ECG200 (left plot), there is minimal overlap 
between the filters of the three models, indicating distinct learning outcomes. In contrast, for 
NonInvasiveFetalECGThorax1 (right plot), there are numerous overlapping areas among the filters of 
different models. This supports our earlier argument in Section 4.3 that larger datasets tend to refine 
existing knowledge rather than discovering new features. However, the presence of new regions for the 
pre-trained and fine-tuned filters (green and red) indicates that even large datasets can benefit from 
the new filters explored during pre-training, leveraging insights gained from other datasets.

In summary, the filter distributions show that pre-training allows models to learn different and sometimes 
more complex filters than the baseline approach. While smaller datasets like ECG200 encourage the discovery 
of new, unique features, larger datasets like NonInvasiveFetalECGThorax1 tend to refine knowledge, though 
they can still benefit from the exploration of new filters during pre-training.

\section{Conclusion}\label{sec:pretext-conclusion}

In this chapter, we have explored the integration of hand-crafted convolution filters into deep learning architectures 
for time series classification (TSC). We started by examining the rationale and construction of these filters, 
demonstrating their potential to improve model performance significantly. Our experiments showed that hand-crafted 
filters could generalize across various datasets, enhancing the robustness and accuracy of the state-of-the-art 
deep learning model InceptionTime by proposing it's new hybrid version H-InceptionTime.

We then introduced a pre-training approach using the H-Inception architecture, leveraging the generalization 
capabilities of hand-crafted filters. The pretext task, designed to predict the dataset of origin for each time 
series, proved effective in transferring knowledge from larger to smaller datasets. This methodology helped 
mitigate the common issue of overfitting, particularly when dealing with limited training samples.

The experimental results validated our approach, with the Pre-trained H-InceptionTime (PHIT) model 
consistently outperforming the baseline across different dataset domains. Our analysis also highlighted 
the importance of fine-tuning with additional layers to capture deeper features, further boosting performance.

In summary, this chapter demonstrates that incorporating hand-crafted filters and a strategic pre-training 
approach can significantly enhance the performance of deep learning models for TSC. These findings pave 
the way towards developing foundation models that can generalize well across diverse datasets, reducing 
the need for extensive data collection and training from scratch.

One of the key insights from this chapter was that simply adding hand-crafted filters to the FCN "poorly performing" 
model enabled it to achieve state-of-the-art performance. This raises an intriguing question as we move forward: 
can we reduce the complexity of all these models while maintaining or even enhancing their performance? 
This question will be the focus of the next chapter. 
\chapter{Reducing Complexity in Deep Learning Models for Time Series Classification}\label{chapitre_4}

\section{Introduction}\label{sec:lite-intro}

In Chapter~\ref{chapitre_3}, we explored the significant impact of integrating hand-crafted filters into small, 
non-complex models in terms of number of parameters (FCN), demonstrating that such models can outperform more complex,
state-of-the-art models (ResNet), as presented in Section~\ref{sec:hfc-between-models-compare}. 
This finding challenges the conventional wisdom that increasing model complexity and the number of parameters 
inherently leads to better performance. Instead, it suggests that strategic simplicity and careful feature engineering 
can yield superior results.
For instance, the groundbreaking work presented in~\cite{inceptiontime-paper} introduced an innovative deep learning model,
InceptionTime (Chapter~\ref{chapitre_1} Section~\ref{sec:tsc-deep}), for 
TSC, significantly advancing the role of Convolutional Neural Networks (CNNs) in this field. Despite its 
impressive performance, InceptionTime, with nearly 2.1 million parameters distributed across five Inception models, 
exemplifies a large and complex architecture.
This complexity poses challenges for deployment in real-world applications, 
particularly those requiring small, resource-constrained devices.

Building on these foundations, this chapter introduces and looks into the innovative approach presented in here. 
The core motivation behind this work is to develop a more efficient and effective model, LITE, for TSC
by leveraging lightweight Inception-based architectures and boosting techniques.
This results in a model that is not only powerful but also efficient and adaptable.

The key contributions of the LITE approach are multifaceted:

\begin{itemize}
    \item \textbf{Lightweight Inception Architecture}: The LITE model employs a streamlined version of the
    Inception (Figure~\ref{fig:inception}) architecture, designed to reduce computational complexity without
    compromising performance. This makes the model more accessible for applications with limited computational resources.
    \item \textbf{Boosting Techniques}: To further enhance performance, LITE integrates boosting techniques that 
    improve the model's ability to generalize across diverse datasets. Boosting helps in mitigating overfitting,
    enhancing its predictive accuracy.
    \item \textbf{Efficiency and Adaptability}: The combination of a lightweight architecture and boosting 
    techniques results in a model that is both efficient and adaptable, capable of performing well across 
    various TSC datasets with reduced training times, lower computational demands and carbon footprint.
\end{itemize}

In real-world scenarios, the LITE model has significant implications. For example, in the healthcare sector, 
it can be utilized for rapid and accurate diagnosis of heart conditions from ECG signals, even in resource-constrained 
environments. In traffic management, LITE can be deployed to predict congestion patterns and optimize traffic flow 
with minimal computational overhead.
\begin{figure}
    \centering
    \caption{
        Difficulties in deploying a high-parameter deep learning model, such as FCN, on a Sony 
        robot for ground type classification. The extensive computational resources and memory 
        required by FCN present significant challenges for resource-constrained devices.
    }
    \label{fig:lite-deployment}
    \includegraphics[width=\textwidth]{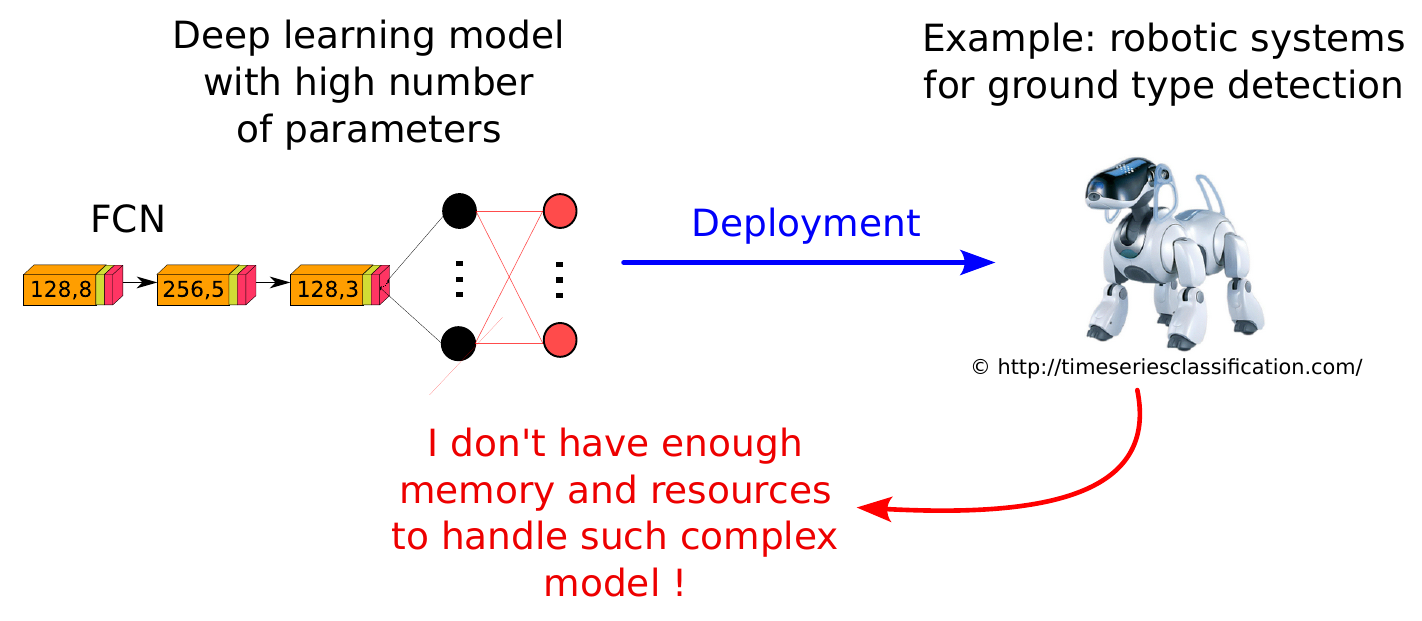}
\end{figure}
Figure~\ref{fig:lite-deployment} illustrates a real-world scenario where deploying a model with over $200,000$ 
parameters, such as FCN, on a Sony robot for ground type classification encounters significant difficulties. 
The high parameter count not only demands substantial computational resources but also strains the device's 
memory and processing capabilities. To address this issue, we propose constructing a lightweight model: LITE.

This chapter will provide an in-depth exploration of the LITE model, discussing its architecture, the integration 
of boosting techniques, and its performance across different TSC tasks. By building on the insights gained from the 
previous chapter regarding the efficacy of hand-crafted filters in simplifying and enhancing model performance, 
we will see how the LITE approach takes these principles further to achieve state-of-the-art results in a 
lightweight and efficient manner.
We also propose an adaptation of the proposed LITE network specifically for the case of multivariate time series,
LITE MultiVariate (LITEMV).
We support the findings of this chapter with extensive experiments on both the UCR and UEA
archives~\cite{ucr-archive,uea-archive}.

\section{The LITE Architecture}\label{sec:lite-archis}

The LITE architecture is a streamlined version of the Inception network (Figure~\ref{fig:inception}), designed to maintain high 
performance while significantly reducing computational overhead. The LITE
architecture, presented in Figure~\ref{fig:lite} with a detailed parametric view, includes:

\begin{figure}
    \centering
    \caption{The proposed LITE architecture for Time Series Classification.}
    \label{fig:lite}
    \includegraphics[width=\textwidth]{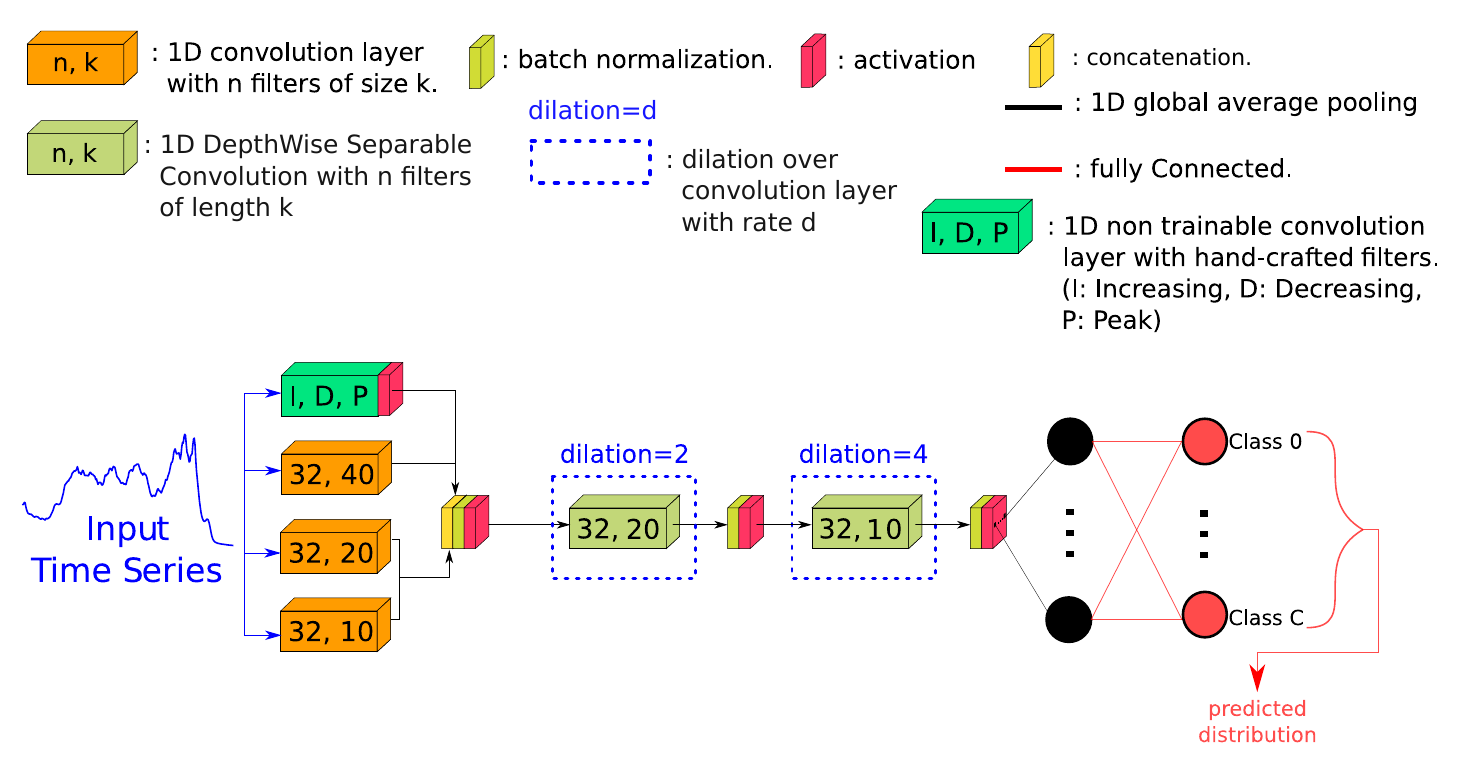}
\end{figure}

\begin{itemize}
    \item \textbf{Hand-Crafted Convolution Filters}: Recognizing the significant impact of hand-crafted filters 
    (Chapter~\ref{chapitre_3} Section~\ref{sec:construction-hcf}) on the performance of the simple FCN network,
    we employ these filters in the first layer of LITE. This approach mirrors the implementations
    in H-FCN and H-Inception (Chapter~\ref{chapitre_3} Section~\ref{sec:hcf-archis}).
    This is the \textbf{first} boosting technique used by LITE.

    \item \textbf{Multiplexing Convolution}: The core of the LITE architecture is built upon Inception modules,
    which apply multiple 
    convolution operations with different filter sizes simultaneously. This allows the model to capture various 
    types of patterns within the time series data. This is referred to as multiplexing convolution, the \textbf{second} boosting 
    technique used in LITE, and it is used only on the raw data and not in the rest of the network such as in
    Inception~\cite{inceptiontime-paper}.
    The convolution layers used in this part of the network are the standard convolution operations.
    The output convolution of these three layers, as well as the hand-crafted filters (with an activation),
    are concatenated on the channels axis and go through a Batch Normalization layer and an activation layer.

    \item \textbf{Efficient Convolutions}: The architecture employs efficient convolution techniques, DWSCs
    (Chapter~\ref{chapitre_1} Section~\ref{sec:tsc-deep}), in the second and third layers of the LITE network. 
    This approach drastically reduces the computational cost and memory footprint 
    while maintaining the model's ability to extract meaningful features from the data.
    It is important to notice that for the first layer, standard convolutions are used instead of DWSC. This is due to the
    fact that as the input time series is univariate, DWSC will learn only one filter.
    The output of these DWSC layers go through a Batch Normalization layer and an activation layer.

    \item \textbf{Dilation}: The LITE network uses a dilation rate for the DWSC 
    layers in the second and third depths. This increases the receptive field without increasing the kernel size, 
    thus reducing parameters, unlike Inception, which does not use dilation. This is the 
    \textbf{third} boosting technique used by LITE.
    Notably, dilation is not used in the first standard convolution layers to avoid missing crucial 
    input data.

    \item \textbf{Global Average Pooling}: Similar to the state-of-the-art networks, e.g. FCN, ResNet and Inception,
    the LITE network applies a GAP operation over the last activation layer, transforming the output MTS to a vector,
    before being fed to the classification FC layer.

\end{itemize}

Table~\ref{tab:lite-params} highlights how much LITE is less complex than the three state-of-the-art networks: FCN, ResNet and 
Inception, in terms of both number of trainable parameters and the number of FLoat-point Operation Per Second (FLOPS).
The number of parameters shown is the number of trainable parameters of the architecture without the last
classification Fully Connected layer because it depends on each dataset (number of classes).
The table shows that the smallest model in terms of number of parameters is the LITE with $9, 814$
parameters. This is mainly due to the usage of DWSC instead of standard ones.

\subsection{LITETime: An Ensemble Approach}\label{sec:litetime}

Similar to InceptionTime, which is an ensemble of five Inception models, we propose LITETime, an ensemble 
of five LITE models. The goal of an ensemble is to reduce the variance in the model's performance. Thus, 
the more sensitive a model is, the greater the impact an ensemble will have.

Given the compact architecture of LITE, with approximately $10k$ parameters compared to the nearly $400k$ parameters 
of Inception, we believe that ensembling multiple LITE models will have a significantly higher impact. 
This approach leverages the efficiency and simplicity of LITE, amplifying its performance through ensembling 
to achieve robust and reliable results.

\begin{table}
    \centering
    \caption{Comparing LITE, FCN, ResNet and Inception in terms of number of trainable parameters and number of 
    FLoat-point Operation Per Second (FLOPS).}
    \label{tab:lite-params}
    \begin{tabular}{c|c|c|c|c|}
    \cline{2-5}
     & FCN & ResNet & Inception & LITE \\ \hline
    \multicolumn{1}{|c|}{\begin{tabular}[c]{@{}c@{}}Number\\ of trainable\\ parameters\end{tabular}} &
    264,704 & 504,000 & 420,192 & 9,814 \\ \hline
    \multicolumn{1}{|c|}{FLOPS} & 266,850 & 507,818 & 424,414 & 10,632 \\ \hline
    \end{tabular}
    \end{table}

\subsection{Experimental Setup}

We utilize the 128 datasets of the UCR archive~\cite{ucr-archive} to evaluate the performance of LITE and LITETime 
compared to existing deep learning models.
All datasets were z-normalized prior to training and testing.

We meticulously measured the 
training time, inference time, CO2 emissions, and energy consumption using the CodeCarbon python package~\cite{codecarbon}.
The best-performing model during training, 
determined by monitoring the training loss, was selected for testing. The Adam optimizer with Reduce on Plateau 
learning rate decay method was employed, using TensorFlow's~\cite{tensorflow-paper} default parameter settings.
Each LITE model in the LITETime ensemble was trained with a batch size of 64 for 1500 epochs, similarly to Inception.


\subsection{Experimental Results}

In this section, we present the experimental results of LITE in terms of performance and efficiency 
compared to other complex deep learning models, notably FCN, ResNet, and Inception.

\subsubsection{Comparing To State-Of-The-Art}\label{sec:litetime-vs-deep}

\begin{figure}
    \centering
    \caption{MCM (Chapter~\ref{chapitre_2}) showing the comparison between (LITE, LITETime) and the rest of the state-of-the-art 
    deep learning models for time series classification.}
    \label{fig:lite-mcm}
    \includegraphics[width=\textwidth]{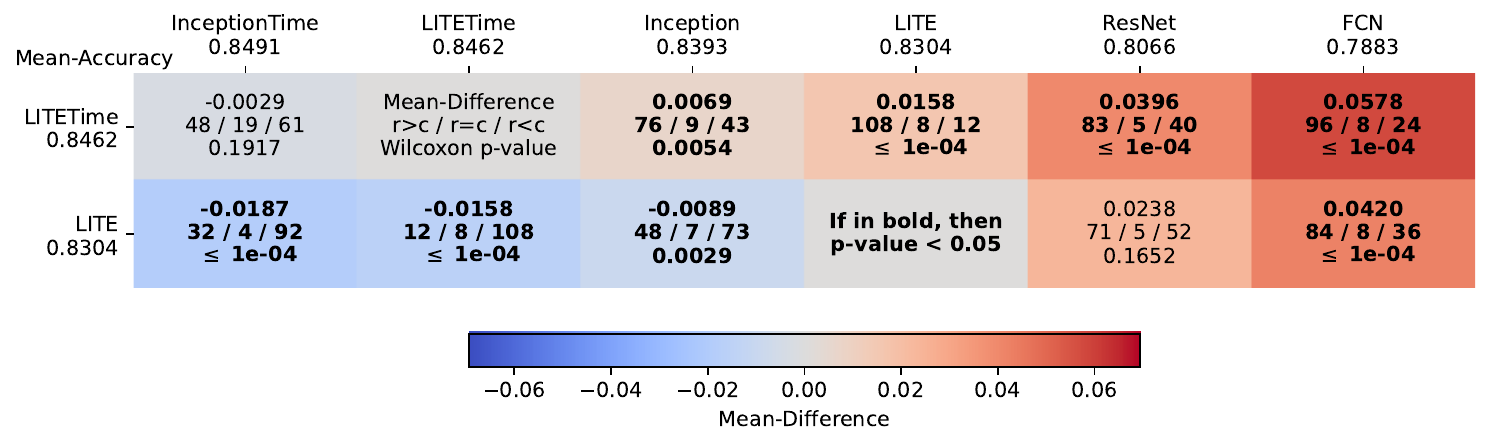}
\end{figure}

Our proposed LITE model and its ensemble, LITETime, demonstrate competitive performance on the UCR archive, 
particularly when considering their significantly smaller size compared to other deep learning models. 
The LITE model, with approximately $10k$ parameters, and LITETime, which ensembles five LITE models, 
show impressive results in terms of accuracy, as presented in Figure~\ref{fig:lite-mcm}.

LITE achieves a mean accuracy of 0.8304, outperforming traditional models such as ResNet (0.8066) and FCN (0.7883), 
which have considerably more parameters. Notably, LITETime further improves this performance, reaching a mean 
accuracy of 0.8462. While slightly below the performance of InceptionTime (0.8491), the advantage of LITE and 
LITETime lies in their efficiency and lower computational requirements.

Examining the significance in performance differences, it is noteworthy that LITE alone significantly 
outperforms FCN and shows no significant difference in performance compared to ResNet. This is a revolutionary 
finding given that LITE has a significantly smaller number of parameters compared to these two models.
Furthermore, LITETime presents no significant difference in performance compared to InceptionTime, despite 
having only $2.34\%$ of the parameters of InceptionTime. This underscores the efficiency and effectiveness 
of LITE and LITETime, demonstrating that smaller, well-optimized models can achieve competitive results 
with much lower computational requirements.

The smaller parameter size of LITE compared to InceptionTime, which has nearly 400,000 parameters, highlights 
the effectiveness of our approach. The compact architecture of LITE, combined with boosting techniques and the 
efficiency of ensembling in LITETime, allows for robust performance with reduced computational costs. 
This makes LITE and LITETime particularly suitable for deployment in resource-constrained environments, 
where model size and inference time are critical factors.

\begin{figure}
    \centering
    \caption{
        For each model, the y-axis shows accuracy on the FreezerSmallTrain dataset, and the x-axis shows FLOPS in 
        a~$\log_{10}$ scale. Circle diameter represents the number of trainable parameters. The smallest model, 
        LITE (ours), has~$10k$ parameters and the lowest FLOPS (4 in $\log_{10}$ scale), while achieving the 
        highest test accuracy.
    }
    \label{fig:lite-tradeoff}
    \includegraphics[width=0.6\textwidth]{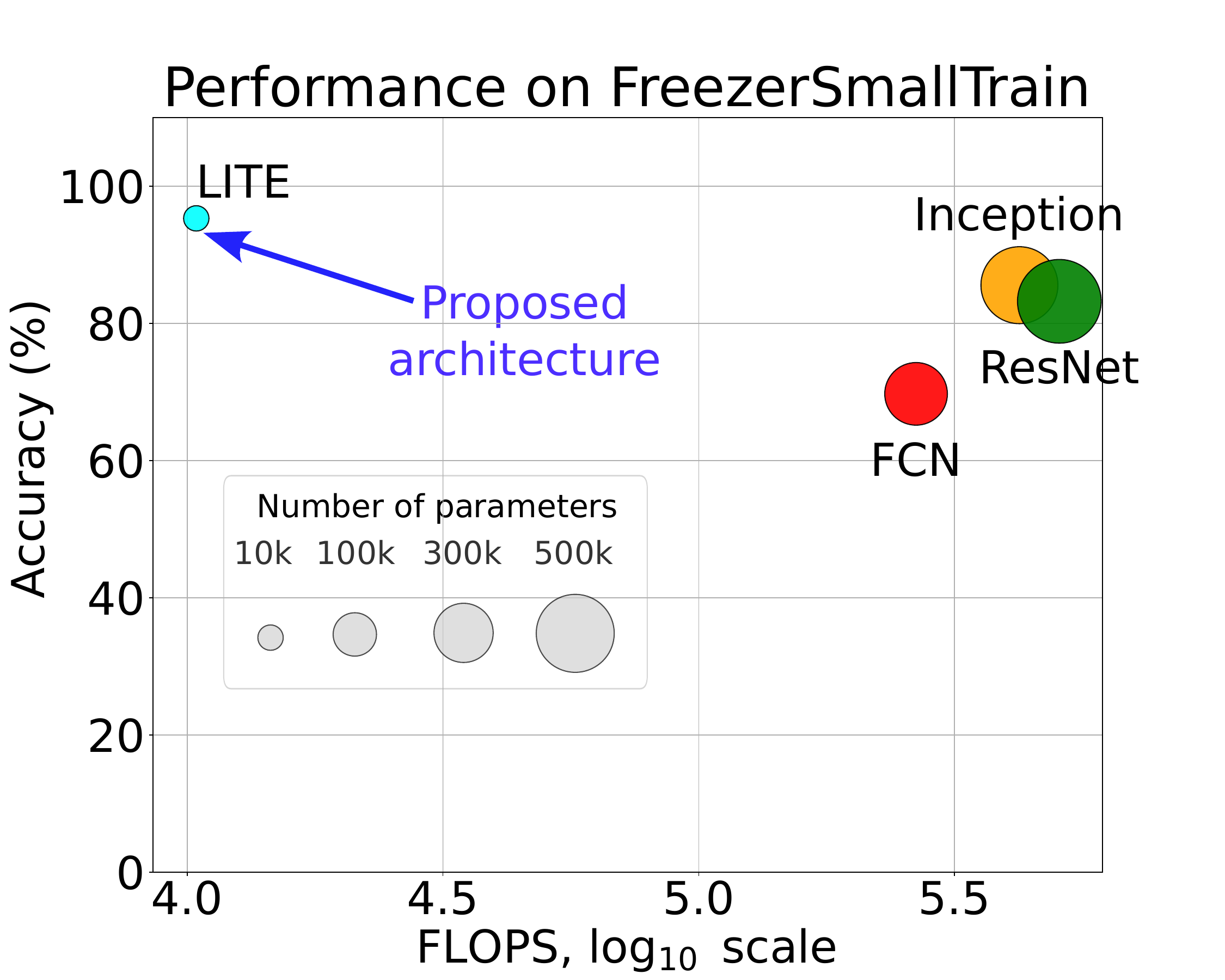}
\end{figure}

Overall, the results validate the hypothesis that smaller, well-optimized models like LITE can achieve 
high performance comparable to larger models, offering a viable and efficient alternative for TSC tasks.
An example showcasing the trade off between performance, FLOPS and number of parameters can be seen in
Figure~\ref{fig:lite-tradeoff}.
In this figure we present the performance of LITE, FCN, ResNet and Inception over the test set of the 
FreezerSmallTrain dataset~\cite{ucr-archive}, as well as the number of FLOPS needed for one inference 
of each model per sliding window.
We also present each model in a form of circle where its radius represents the number of trainable parameters 
of each model.
It can be seen form Figure~\ref{fig:lite-tradeoff} that LITE is the most accurate model in terms of performance as well 
as the most efficient in terms of FLOPS and number of parameters, with a large gap in difference of efficiency with 
the other deep learning models.

\subsubsection{Efficiency Comparison}\label{sec:lite-efficiency}

\begin{table}[h!]
    \hspace*{-1.6cm}
    \centering
    \begin{tabular}{|c|c|c|c|c|c|c|}
        \hline
        Models &
        \begin{tabular}[c]{@{}c@{}}Number of\\ parameters\end{tabular} &
        FLOPS &
        Training Time &
        Testing Time &
        CO2 (g) &
        Energy (Wh) \\ \hline
    Inception &
      420,192 &
      424,414 &
      \begin{tabular}[c]{@{}c@{}}145,267 seconds\\ 1.68 days\end{tabular} &
      \begin{tabular}[c]{@{}c@{}}81 seconds\\ 0.0009 days\end{tabular} &
      0.2928 g &
      0.6886 Wh \\ \hline
      ResNet &
      504,000 &
      507,818 &
      \begin{tabular}[c]{@{}c@{}} 165,089 seconds\\  1.91 days\end{tabular} &
      \begin{tabular}[c]{@{}c@{}} 62 seconds\\  0.0007 days\end{tabular} &
      0.3101 g &
      0.7303 Wh \\ \hline
    FCN &
    264,704 &
    266,850 &
    \begin{tabular}[c]{@{}c@{}}149,821 seconds\\ 1.73 days\end{tabular} &
    \begin{tabular}[c]{@{}c@{}}27 seconds\\ 0.00031 days\end{tabular} &
    0.2623 g &
    0.6176 Wh \\ \hline
    \textbf{LITE} &
    \textbf{9,814} &
    \textbf{10,632} &
    \begin{tabular}[c]{@{}c@{}}\textbf{53,567 seconds}\\ \textbf{0.62 days}\end{tabular} &
    \begin{tabular}[c]{@{}c@{}}\textbf{44 seconds}\\ \textbf{0.0005 days}\end{tabular} &
    \textbf{0.1048 g} &
    \textbf{0.2468 Wh} \\ \hline
\end{tabular}
\caption{Comparison between the proposed methods with FCN, ResNet and Inception without ensemble.}
\label{tab:lite-efficiency}
\end{table}

Table~\ref{tab:lite-efficiency} summarizes the number of parameters, the number of
(FLOPS), training time, inference time, CO2, and power consumption using CodeCarbon~\cite{codecarbon}
across the 128 datasets of the 
UCR archive~\cite{ucr-archive}. The values are aggregated over the 128 datasets and averaged over five different runs.

Compared to FCN, ResNet, and Inception, LITE has only $3.7\%$, $1.95\%$, and $2.34\%$ of their 
respective number of parameters. Moreover, LITE is the fastest model during the training phase, 
with a training time of $0.62$ days. This makes LITE $2.79$, $3.08$, and $2.71$ times faster than 
FCN, ResNet, and Inception, respectively.

Additionally, LITE consumes the least amount of CO2 and energy, at $0.1048$ g and $0.2468$ Wh, respectively. 
This demonstrates that LITE is not only the fastest but also the most environmentally friendly model for 
TSC compared to FCN, ResNet, and Inception. Given these factors, we believe that LITE is highly suitable 
for deployment in small devices, such as mobile phones.

\subsection{Ablation Studies}

In this section, we present an extensive study on the various features of the LITETime classifier. 
LITETime leverages three key features: (1) boosting techniques in each LITE model, (2) Depthwise 
Separable Convolution layers, and (3) the ensemble approach. In the following three sections, 
we examine the significance of each of these features, providing a detailed analysis of their 
contributions to the overall performance of LITETime.

\subsubsection{Impact Of Boosting Techniques}

\begin{figure}
    \centering
    \caption{
        The Heat Map shows the one-vs-one comparison between the Striped-LITE and the three variants: (1) Add-Custom-Filters, (2) Add-Multiplexing-
Convolution and (3) Add-Dilated-Convolution. The colors of the Heat Map follow the value of the first line in each cell. This value is the difference between
the value of the first line (average accuracy when winning/losing). The second line represents the Win/Tie/Loss count between the models in question (wins
for the column model). The last line is the statistical P-Value between the two classifier using the Wilcoxon Signed Rank Test
    }
    \label{fig:lite-ablation}
    \includegraphics[width=\textwidth]{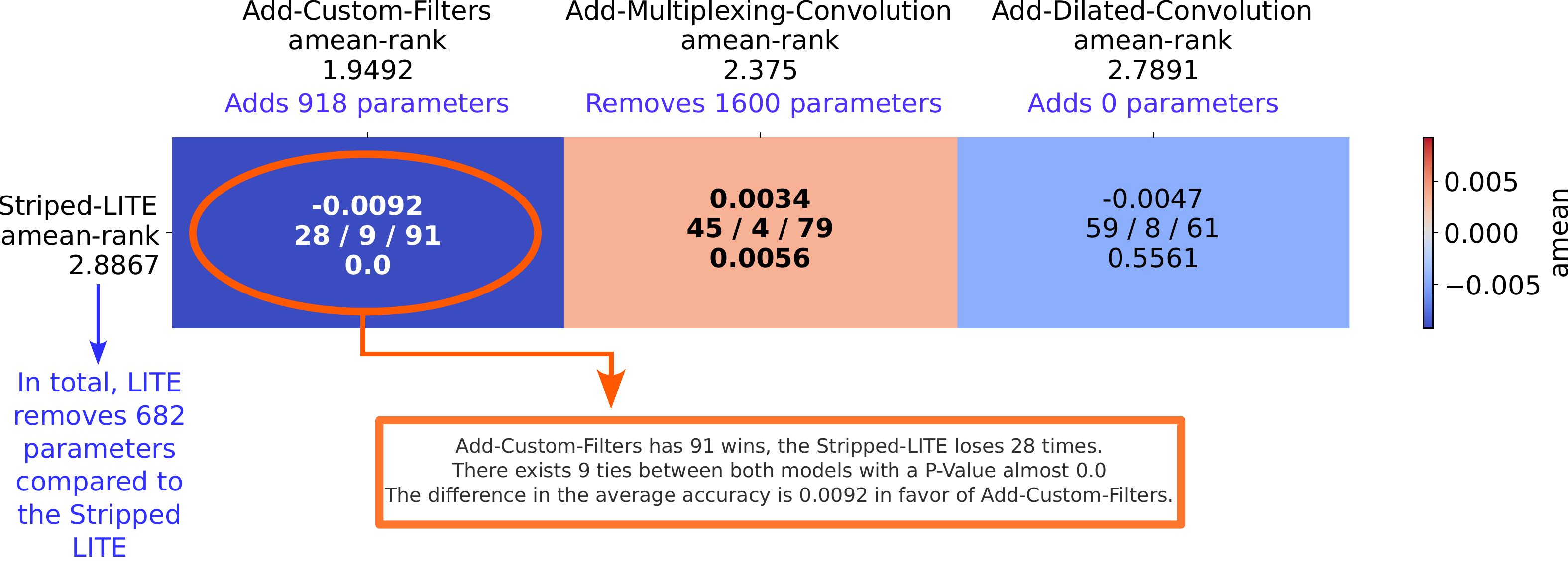}
\end{figure}

The LITE architecture leverages several advanced techniques to boost its performance. To demonstrate the 
individual impact of each technique, we conduct a comprehensive ablation study.

\begin{figure}
    \centering
    \caption{The stripped version of the LITE architecture for Time Series Classification.}
    \label{fig:stripped-lite}
    \includegraphics[width=\textwidth]{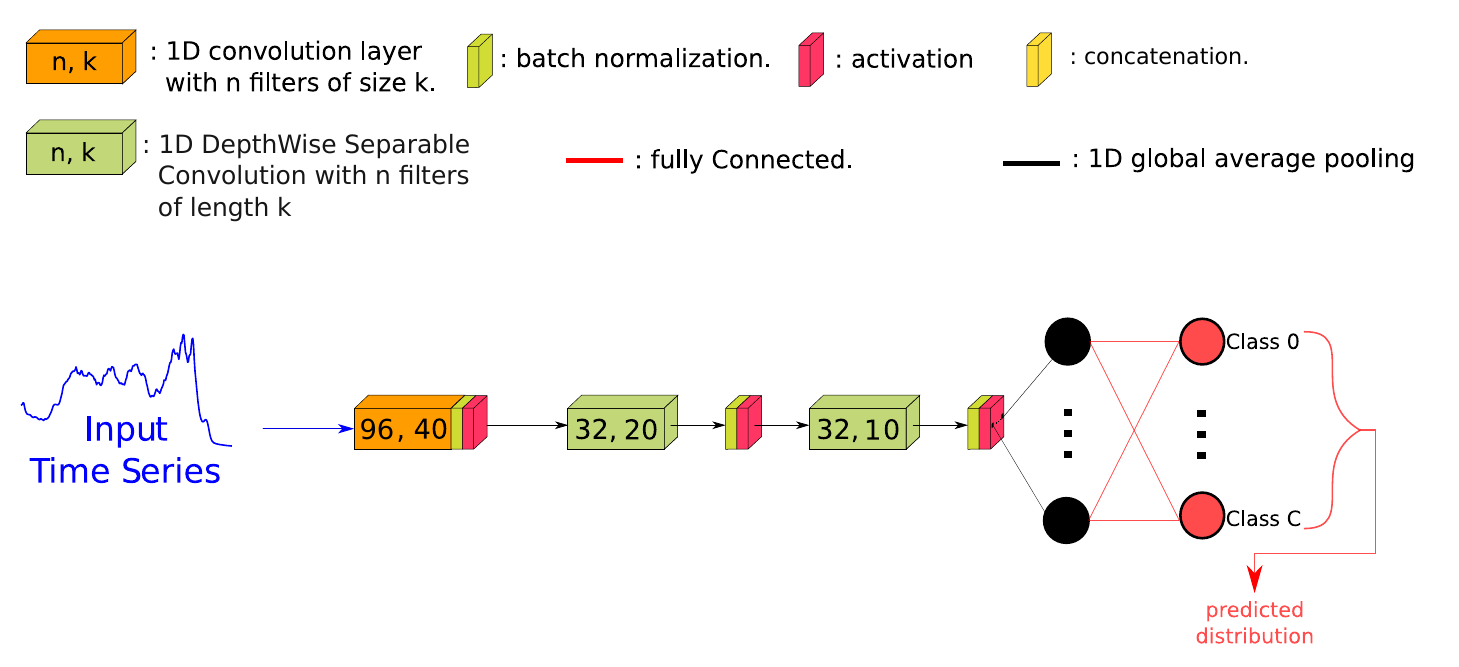}
\end{figure}

Initially, we strip the LITE model of its three key techniques: dilation, multiplexing, and hand-crafted filters. 
In the multiplexing convolutions performed in the first layer, there are three layers with $32$ filters, 
resulting in the stripped-down LITE learning a total of $96=3$x$32$ filters for the first depth. The remaining 
architecture is kept the same, utilizing DWSCs without dilation.
The detailed architecture of the Striped-LITE is presented in Figure~\ref{fig:stripped-lite}.

Following this, we reintroduce each boosting technique one at a time to the stripped LITE model and evaluate 
its performance. The results of this ablation study are illustrated in Figure~\ref{fig:lite-ablation} using an MCM.
The results demonstrate that integrating hand-crafted filters in the first layer and utilizing multiplexing convolutions 
significantly enhance the LITE model's performance. The MCM shows that hand-crafted filters positively impact 
average accuracy, although they do introduce additional parameters. In contrast, multiplexing convolutions achieve 
minor performance gains.
The minor average impact of $0.34\%$ is outweighed by the benefits, as multiplexing significantly reduces the 
number of parameters and consistently outperforms across the majority of datasets.

While the addition of 
dilated convolutions does not yield statistically significant improvements (p-value $> 0.05$), the average 
accuracy differences indicate that dilated convolutions generally enhance performance. This is particularly 
relevant for large datasets, as dilation expands the receptive field without increasing parameter count. 
Occasionally, dilation may negatively affect performance on datasets that do not require an extensive receptive field.

In summary, the LITE model, equipped with these boosting techniques, features fewer parameters than the 
stripped-down version while delivering performance on par with state-of-the-art models. The reduction in 
parameter count is primarily due to multiplexing, which offsets the additional parameters introduced by hand-crafted filters.

Figure~\ref{fig:lite-ablation} also presents the average rank of the models, similar to the CD Diagram. 
The model with custom filters ranks the highest, indicating the best performance, while the stripped LITE 
ranks the lowest, marking it as the least effective. Thus, the Stripped-LITE model, without any boosting techniques,
is the weakest among the configurations shown in the MCM.

\subsubsection{Impact Of DepthWise Separable Convolutions}\label{sec:lite-impact-dwsc}

\begin{figure}
    \centering
    \caption{One-vs-one comparison between LITETime and
    LITETime with Standard convolutions over the 128
    datasets of the UCR archive~\cite{ucr-archive}.}
    \label{fig:lite-dwsc-impact}
    \includegraphics[width=0.5\textwidth]{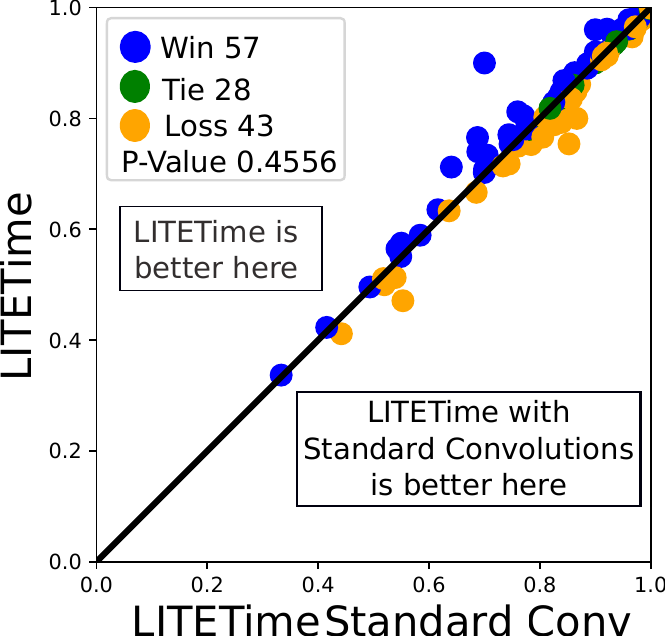}
\end{figure}

To further investigate the impact of DepthWise Separable Convolutions (DWSC), we replaced them with standard 
convolutions followed by a BottleNeck layer.
The reason we add a bottleneck layer is it simulates the reduction of number of filters needed to learn, 
as motivated from the Inception architecture~\cite{inceptiontime-paper}.
To ensure a fair and accurate comparison, we employed ensemble 
techniques, which are crucial in this context due to the significant disparity in the number of parameters, LITE 
has only about $11\%$ of the parameters compared to the alternative model.
Since the alternative model has about $85,000$ parameters, LITE would have around 9,350 parameters ($11\% of 85,000$).

Figure~\ref{fig:lite-dwsc-impact} showcases a one-on-one comparison between LITETime and LITETime utilizing standard convolutions. 
The findings reveal that incorporating DWSC does not substantially influence performance, as indicated 
by a high P-Value of 0.4556. This high p-value suggests that the performance difference is not statistically 
significant, highlighting that DWSC can achieve comparable results with a significantly reduced parameter count.

\subsubsection{Impact Of Number of LITE Models in the Ensemble}\label{sec:lite-number-heads}

\begin{figure}
    \centering
    \caption{A Critical Difference diagram showcasing the
    comparison of performance of LITETime when more or less
    LITE models are used in the ensemble.}
    \label{fig:lite-ensembles-cdd}
    \includegraphics[width=\textwidth]{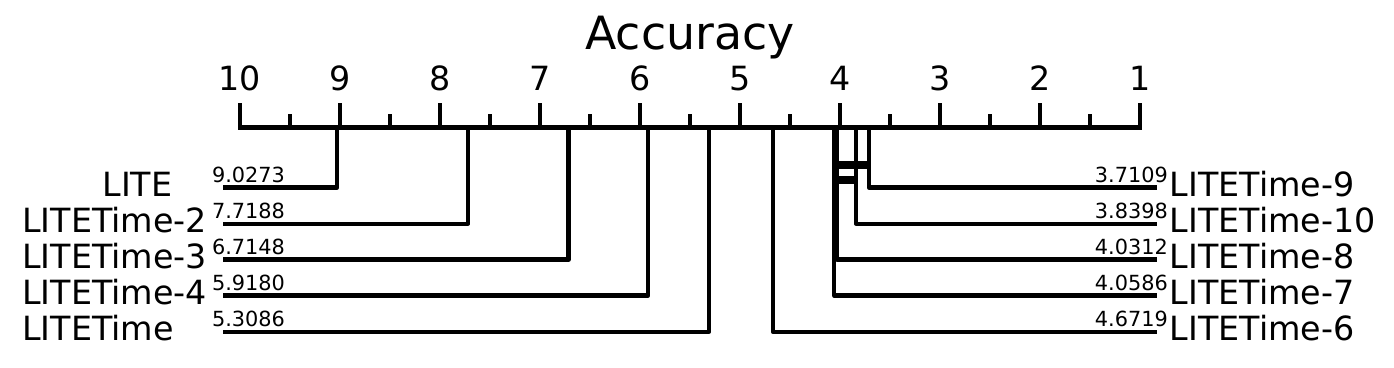}
\end{figure}

\begin{figure}
    \centering
    \caption{A Multi-Comparison Matrix (Chapter~\ref{chapitre_2}) showcasing the
    comparison of performance of LITETime-10,9 and 8 with other LITETime models with varying
    number of LITE heads.}
    \label{fig:lite-ensembles-mcm}
    \includegraphics[width=\textwidth]{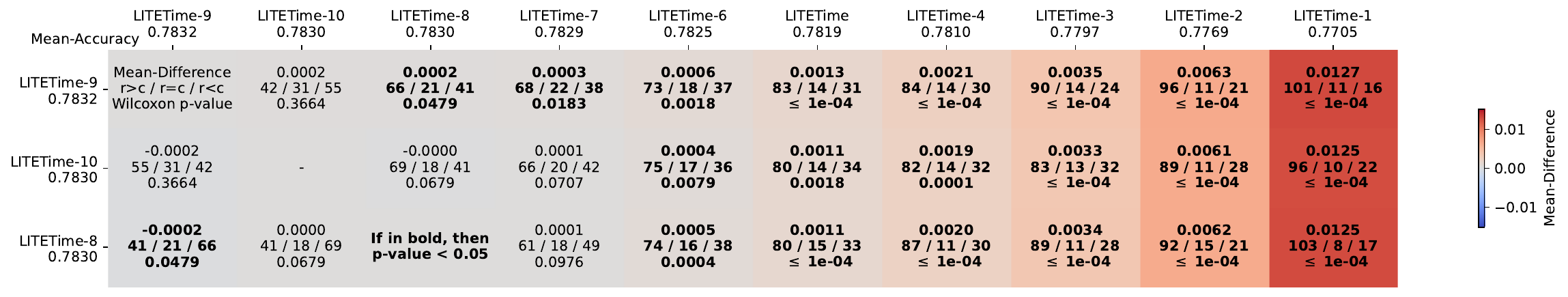}
\end{figure}

In previous experiments, we used five LITE models in LITETime to ensure a 
fair comparison with InceptionTime, which is an ensemble of five Inception models. The original work of
InceptionTime~\cite{inceptiontime-paper} has 
shown that no significant performance improvement is observed on the UCR archive when the ensemble size of 
InceptionTime exceeds five models. However, due to the smaller and more lightweight architecture of LITE, 
it may exhibit greater variance and reduced robustness compared to Inception.

To address this, we believe that increasing the number of LITE models in the LITETime ensemble could enhance performance. 
To test this hypothesis, we trained ten different LITE models on the UCR archive and constructed ensembles of 
varying sizes (from 1 to 10 models) by averaging all possible ensemble combinations. For example, to create LITETime-3 
(an ensemble of three LITE models), we combined all possible sets of three models from the ten trained models. 
The results are displayed in the CD diagram in Figure~\ref{fig:lite-ensembles-cdd} and the detailed MCM in 
Figure~\ref{fig:lite-ensembles-mcm}.

As illustrated in Figure~\ref{fig:lite-ensembles-cdd}, LITETime-5 (LITETime)
(an ensemble of five LITE models) is not the optimal limit for LITE. 
Instead, LITETime-7 proves to be more effective. This is due to LITE's compact size, approximately $42$ times 
smaller than Inception, which allows for greater scalability within the ensemble framework. Consequently, 
LITETime can enhance accuracy while maintaining significantly lower complexity compared to InceptionTime. 
Specifically, LITETime-5, with five models, utilizes only about $2.34\%$ of InceptionTime's trainable parameters, 
and LITETime-7 increases this to just $3.27\%$.
However, the MCM in Figure~\ref{fig:lite-ensembles-mcm} demonstrates that LITETime-9 significantly outperforms
LITETime-8, which contrasts with the conclusions drawn from the CD diagram in Figure~\ref{fig:lite-ensembles-cdd}.
Additionally, LITETime-10 does not show a significant difference compared to LITETime-9, indicating that the optimal
ensemble size for LITETime lies between nine and ten models. Remarkably, even with ten models, LITETime-10
remains approximately 21 times smaller than the ensemble InceptionTime.

\begin{figure}
    \centering
    \caption{A Comparison on the Beef dataset of the UCR
    archive~\cite{ucr-archive} between the ensemble of LITE and Inception models.
    The $x$-axis represents the number of models used in
    each ensemble and the $y$-axis the performance of the ensemble on the test set of the Beef dataset.}
    \label{fig:lite-ensembles-beef}
    \includegraphics[width=0.7\textwidth]{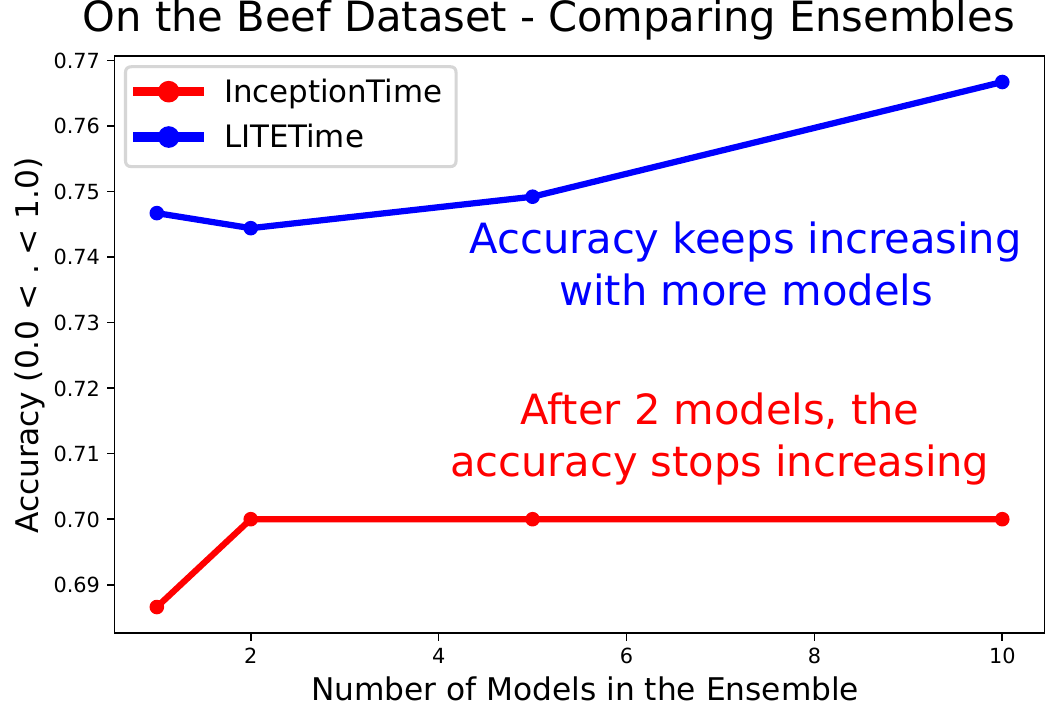}
\end{figure}

Moreover, Figure~\ref{fig:lite-ensembles-beef} provides a concrete example using the Beef dataset from the UCR archive,
illustrating how the 
performance on unseen data varies with the number of models in both the LITETime and InceptionTime ensembles. 
This figure underscores the scalability and efficiency of the LITETime approach, highlighting its potential 
for superior performance with minimal computational overhead.

\paragraph*{LITETime for Multivariate Time Series?}

Given the groundbreaking findings of this study with LITETime, its suitability for real-world applications becomes 
evident due to its remarkably small parameter count and memory footprint. However, practical applications often 
involve multivariate TSC tasks. This raises an important question: Can LITETime be effectively 
applied to multivariate scenarios, or does it require adaptation?

In the following section, we address this question by introducing a novel multivariate deep learning model, LITEMV 
(LITE MultiVariate). This new model is specifically designed to extend the capabilities of LITETime to handle the 
complexities of multivariate TSC, ensuring its applicability across a broader range of real-world tasks.

\section{LITEMV: Addressing Multivariate Time Series Classification}

As we explore the potential applications of the LITE architecture, it becomes evident that many real-world 
scenarios involve multivariate time series data and not only univariate data. Examples include medical diagnostics 
using multiple biosignals, financial forecasting with various economic indicators, and industrial monitoring with 
multiple sensor readings. Efficient and effective handling of multivariate data is essential for advancing the 
applicability of TSC models where channel dependency is crucial for capturing discriminative patterns.

While the LITE model has demonstrated exceptional performance and efficiency for univariate TSC, 
its design requires adaptation to fully leverage the information contained in multivariate datasets. The standard 
convolution approach used in the first layer of LITE for univariate data does not fully exploit the potential of 
multivariate inputs, where interactions between different channels can provide critical insights.

The proposed LITEMV (LITE MultiVariate) architecture is designed to address these challenges. By learning 
a filter per channel and combining them effectively, LITEMV ensures that the unique contributions of 
each channel are preserved and utilized to enhance classification performance. This adaptation allows 
LITEMV to maintain the efficiency and performance advantages of LITE while being optimized for the 
complexities of multivariate time series data.

\subsection{Model Adaptation For Multivariate Time Series}\label{sec:litemv-archi}

To address the issue of effectively handling MTS data, we propose adapting the LITE 
architecture by replacing the three standard convolution layers at the beginning of the network with 
DWSC layers. DWSCs allow each channel to be processed independently, 
preserving the unique information in each channel before combining them.

Additionally, the hand-crafted convolution filters used at the beginning of the network, originally 
implemented as standard convolutions, are also replaced with DepthWise Convolutions
(DWCs, Chapter~\ref{chapitre_1} Section~\ref{sec:tsc-deep}). While we continue 
to use the same hand-crafted filters, the outputs of these DWCs are concatenated 
rather than summed. This ensures that the information from each channel is retained and effectively 
utilized in the subsequent layers of the network.

This adaptation enhances the model's ability to manage the complexities of multivariate time series data, 
leading to improved overall performance and accuracy. We refer to this enhanced multivariate version of 
LITE as LITEMV and its ensemble version LITEMVTime.

In what follows we present extensive experiments to highlight the contribution of this adaptation
and its placement compared to the state-of-the-art deep learning models for multivariate TSC tasks, notably
ConvTran~\cite{convtran-paper} and Disjoint-CNN~\cite{disjoint-cnn-paper} (Chapter~\ref{chapitre_1} Section~\ref{sec:tsc-deep}).

\subsection{Experimental Setup}

We utilize the 30 datasets of the UEA archive~\cite{uea-archive} to evaluate the
performance of LITETime and LITEMVTime compared to existing deep learning models for multivariate TSC.
All datasets were z-normalized prior to training and testing independently on each channel.
The best-performing model during training, determined by monitoring the training
loss, was selected for testing. The Adam optimizer with a Reduce on Plateau learning
rate decay method was employed, using TensorFlow's~\cite{tensorflow-paper} default
parameter settings. Similar to LITE, each LITEMV model in the LITEMVTime ensemble was trained with
a batch size of 64 for 1500 epochs.

\subsection{Experimental Results}

\begin{table}
    \centering
    \caption{Accuracy performance in $\%$ of LITEMVTime (LMVT), LITETime (LT), ConvTran (CT), InceptionTime (IT), Disjoint-CNN (D-CNN), FCN and ResNet on $30$ datasets of the UEA archive. The datasets are ordered by their average number of training samples per class. The accuracy of the best model for each dataset is presented in bold and of the second best is underlined.}
    \label{tab:lite-results-uea}
    \resizebox{\columnwidth}{!}{
    \begin{tabular}{|c|c|c|c|c|c|c|c|c|}
    \hline
    Dataset &
      \begin{tabular}[c]{@{}c@{}}Train Size\\ per Class\end{tabular} &
      LMVT &
      LT &
      CT &
      IT &
      D-CNN &
      FCN &
      ResNet \\ \hline
    FaceDetection             & 2945 & 61.01          & {\ul 62.37}    & \textbf{67.22} & 58.85          & 56.65          & 50.37          & 59.48          \\ \hline
    InsectWingbeat            & 2500 & 61.72          & 39.79          & \textbf{71.32} & {\ul 69.56}    & 63.08          & 60.04          & 65.00          \\ \hline
    PenDigits                 & 750  & \textbf{98.86} & {\ul 98.83}    & 98.71          & 97.97          & 97.08          & 98.57          & 97.71          \\ \hline
    SpokenArabicDigits        & 660  & 98.59          & {\ul 98.77}    & \textbf{99.45} & 98.72          & 98.59          & 98.36          & 98.32          \\ \hline
    LSST                      & 176  & \textbf{66.42} & {\ul 62.85}    & 61.56          & 44.56          & 55.59          & 56.16          & 57.25          \\ \hline
    FingerMovements           & 158  & \textbf{56.00} & 44.00          & \textbf{56.00} & \textbf{56.00} & {\ul 54.00}    & 53.00          & {\ul 54.00}    \\ \hline
    MotorImagery              & 139  & 53.00          & 51.00          & \textbf{56.00} & 53.00          & 49.00          & {\ul 55.00}    & 52.00          \\ \hline
    SelfRegulationSCP1        & 134  & 73.04          & 75.09          & \textbf{91.80} & 86.34          & {\ul 88.39}    & 78.16          & 83.62          \\ \hline
    Heartbeat                 & 102  & 61.46          & 67.80          & \textbf{78.53} & 62.48          & 71.70          & 67.80          & {\ul 72.68}    \\ \hline
    SelfRegulationSCP2        & 100  & {\ul 55.00}    & 53.89          & \textbf{58.33} & 47.22          & 51.66          & 46.67          & 50.00          \\ \hline
    PhonemeSpectra            & 85   & 15.81          & 17.45          & \textbf{30.62} & 15.86          & {\ul 28.21}    & 15.99          & 15.96          \\ \hline
    CharacterTrajectories     & 72   & \textbf{99.58} & {\ul 99.51}    & 99.22          & 98.81          & 99.45          & 98.68          & 99.45          \\ \hline
    EthanolConcentration      & 66   & \textbf{69.20} & {\ul 67.30}    & 36.12          & 34.89          & 27.75          & 32.32          & 31.55          \\ \hline
    HandMovementDirection     & 40   & 35.14          & 21.62          & {\ul 40.54}    & 37.83          & \textbf{54.05} & 29.73          & 28.38          \\ \hline
    PEMS-SF                   & 39   & 79.19          & 82.66          & 82.84          & \textbf{89.01} & \textbf{89.01} & {\ul 83.24}    & 73.99          \\ \hline
    RacketSports              & 38   & 73.68          & 78.29          & \textbf{86.18} & 82.23          & {\ul 83.55}    & 82.23          & 82.23          \\ \hline
    Epilepsy                  & 35   & \textbf{99.28} & {\ul 98.55}    & {\ul 98.55}    & \textbf{99.28} & 88.98          & \textbf{99.28} & \textbf{99.28} \\ \hline
    JapaneseVowels            & 30   & 96.49          & 97.30          & \textbf{98.91} & 97.02          & {\ul 97.56}    & 97.30          & 91.35          \\ \hline
    NATOPS                    & 30   & 90.00          & 88.89          & \textbf{94.44} & 91.66          & {\ul 92.77}    & 87.78          & 89.44          \\ \hline
    EigenWorms                & 26   & {\ul 93.89}    & \textbf{95.42} & 59.34          & 52.67          & 59.34          & 41.98          & 41.98          \\ \hline
    UWaveGestureLibrary       & 15   & 84.68          & 85.00          & {\ul 89.06}    & \textbf{90.93} & {\ul 89.06}    & 85.00          & 85.00          \\ \hline
    Libras                    & 12   & {\ul 89.44}    & 87.78          & \textbf{92.77} & 87.22          & 85.77          & 85.00          & 83.89          \\ \hline
    ArticularyWordRecognition & 11   & 97.33          & 97.67          & {\ul 98.33}    & \textbf{98.66} & \textbf{98.66} & 98.00          & 98.00          \\ \hline
    BasicMotions &
      10 &
      \textbf{100.0} &
      {\ul 95.00} &
      \textbf{100.0} &
      \textbf{100.0} &
      \textbf{100.0} &
      \textbf{100.0} &
      \textbf{100.0} \\ \hline
    DuckDuckGeese             & 10   & 18.00          & 24.00          & \textbf{62.00} & 36.00          & {\ul 50.00}    & 36.00          & 24.00          \\ \hline
    Cricket                   & 9    & {\ul 98.61}    & 97.22          & \textbf{100.0} & {\ul 98.61}    & 97.72          & 93.06          & 97.22          \\ \hline
    Handwriting               & 6    & \textbf{40.00} & 36.82          & 37.52          & 30.11          & 23.72          & {\ul 37.60}    & 18.00          \\ \hline
    ERing                     & 6    & 84.44          & 89.63          & \textbf{96.29} & {\ul 92.96}    & 91.11          & 90.37          & {\ul 92.96}    \\ \hline
    AtrialFibrillation        & 5    & 13.33          & 06.67          & \textbf{40.00} & 20.00          & \textbf{40.00} & {\ul 33.33}    & {\ul 33.33}    \\ \hline
    StandWalkJump             & 4    & \textbf{66.67} & {\ul 60.00}    & 33.33          & 40.00          & 33.33          & 40.00          & 40.00          \\ \hline
    \end{tabular}
    }
\end{table}

\begin{figure}
    \centering
    \caption{A Multi-Comparison Matrix (MCM) showcasing
    the performance of LITEMVTime, LITETime, InceptionTime,
    Disjoint-CNN and ConvTran on the 30 datasets of the UEA
    archive.}
    \label{fig:lite-uea-mcm}
    \includegraphics[width=\textwidth]{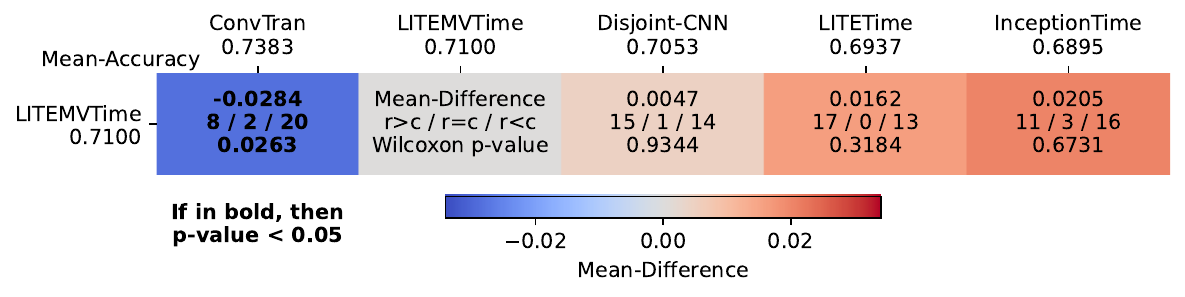}
\end{figure}

In Table~\ref{tab:lite-results-uea}, we compare the accuracy performance of LITEMVTime, LITETime, and five other
leading models: ConvTran, 
InceptionTime, Disjoint-CNN, FCN, and ResNet. The accuracy metrics for these competitors are sourced from the 
ConvTran paper~\cite{convtran-paper}. LITEMVTime shows a significant performance edge when it surpasses other models,
as evidenced 
by the substantial gaps in accuracy. This is further illustrated in the MCM plot in Figure~\ref{fig:lite-uea-mcm},
where LITEMVTime's 
performance is benchmarked against the other models.

LITEMVTime ranks second in overall performance, outperforming LITETime, Disjoint-CNN and InceptionTime. 
Although LITEMVTime does not exceed ConvTran on more than eight datasets, a detailed examination 
of Table~\ref{tab:lite-results-uea} reveals that its victories are often by a wide margin. For example, on the EigenWorms 
dataset, ConvTran achieves a top accuracy of $59.34\%$, whereas LITEMVTime attains an impressive $93.89\%$, 
and LITETime achieves an even higher accuracy of $95.42\%$.
In the following section, we dig into the analysis into the common characteristics of the datasets where LITEMVTime 
wins with a significant margin compared to ConvTran.
Moreover, a crucial limitation of ConvTran is its sensitivity to the length of the input series. As a Self-Attention 
based network, ConvTran must store an attention score matrix of size $(L,L)$, where $L$ is the length of the 
input series, with a runtime complexity of $\mathcal{O}(L^2)$. This can create issues for long time series, 
as the model may face out-of-memory errors during training despite its architectural simplicity.

\subsection{Analysis On Dataset Characteristics When Comparing LITEMVTime and ConvTran}

\begin{figure}
    \centering
    \caption{Difference of performance between LITEMVTime and ConvTran with respect to the number
    of training samples in log scale.}
    \label{fig:litemv-train-size}
    \includegraphics[width=0.8\textwidth]{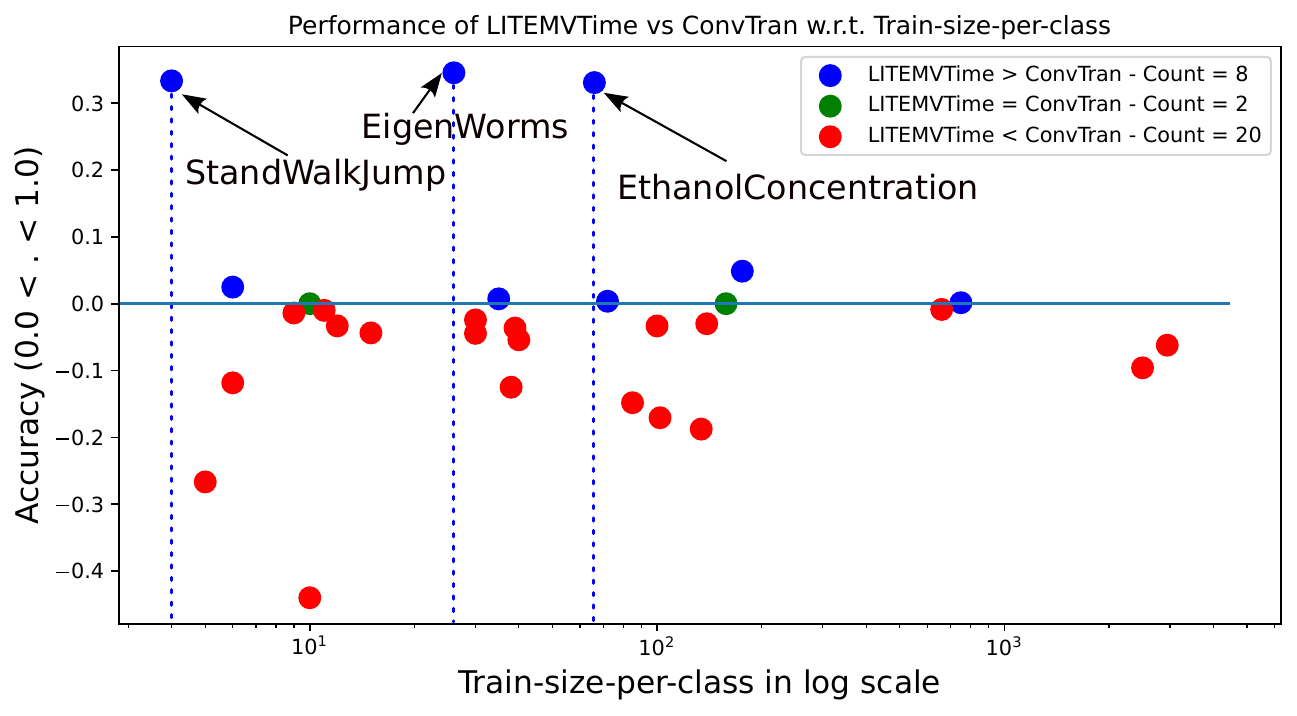}
\end{figure}

To gain a deeper understanding of the scenarios where LITEMVTime significantly outperforms ConvTran, 
we analyzed the performance differences between these two models in relation to the number of training 
samples per class. This analysis aims to uncover any patterns or commonalities in the datasets where 
LITEMVTime demonstrates superior performance.

Figure~\ref{fig:litemv-train-size} illustrates the performance gaps between LITEMVTime and
ConvTran. The most pronounced differences 
are observed in three datasets: StandWalkJump, EigenWorms, and EthanolConcentration. These datasets exhibit 
considerable variation in the number of training examples. For instance, StandWalkJump has a small training 
set with only 4 samples per class and a total of 12 training samples. On the other hand, the EthanolConcentration 
dataset contains 66 training samples per class, amounting to 261 samples in total.

This analysis indicates that LITEMVTime's enhanced performance is not confined to datasets with a specific 
size but extends across datasets with varying numbers of training examples. This highlights LITEMVTime's 
robustness and versatility in handling diverse dataset conditions, making it a reliable choice for a wide range 
of time series classification tasks.

\begin{figure}
    \centering
    \caption{Difference of performance between LITEMVTime and ConvTran with respect to the number
    of channels in log scale.}
    \label{fig:litemv-channels}
    \includegraphics[width=0.8\textwidth]{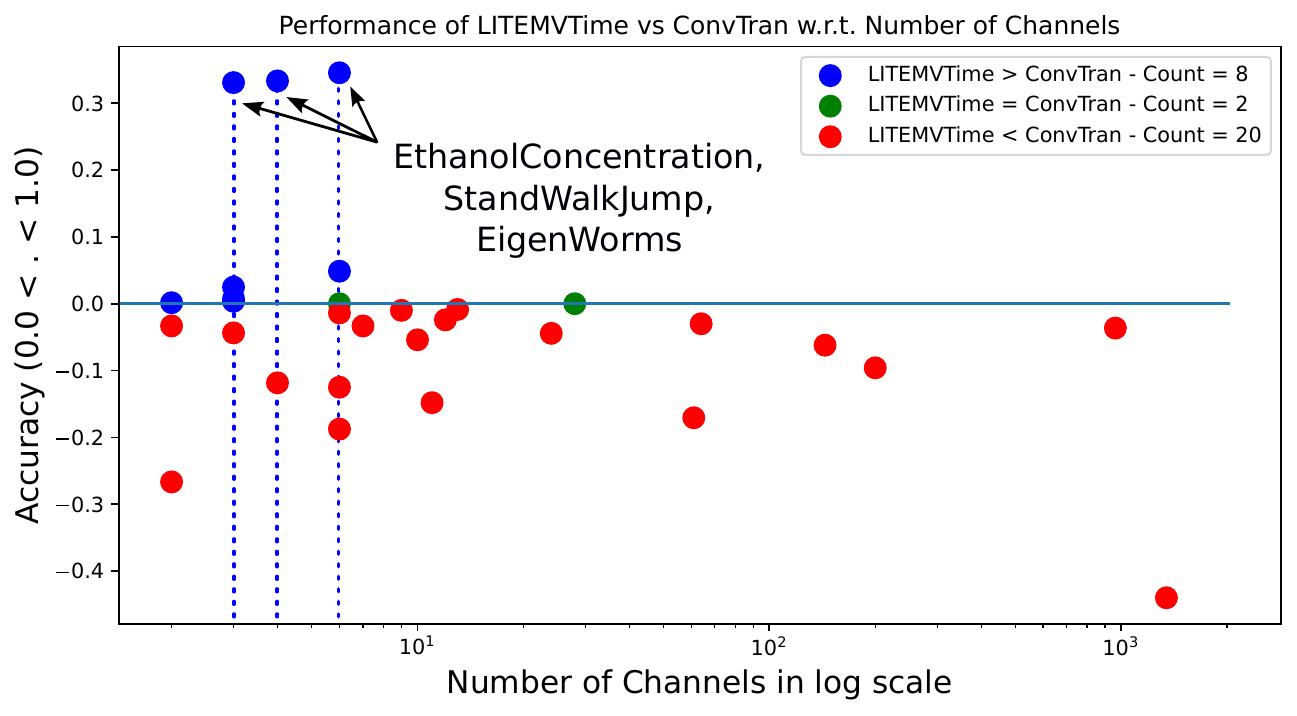}
\end{figure}

To further investigate why LITEMVTime performs better than ConvTran on certain datasets, we can analyze the 
impact of the number of dimensions in multivariate time series data. Figure~\ref{fig:litemv-channels}
illustrates the same performance 
differences shown in Figure~\ref{fig:litemv-train-size}, but this time as a function of the number of dimensions in the datasets.
The analysis highlights that the datasets where LITEMVTime shows significant superiority,StandWalkJump, 
EigenWorms, and EthanolConcentration,all have a relatively small number of dimensions. However, 
ConvTran consistently outperforms LITEMVTime as the number of dimensions increases.

Figures~\ref{fig:litemv-train-size} and~\ref{fig:litemv-channels} illustrate that LITEMVTime is more suitable
for scenarios with small training data
and a limited number of channels. This limitation of ConvTran was also discussed in the original paper~\cite{convtran-paper}.

\section{Discussion Over Limitations of LITE and LITEMV}

LITE and LITEMV are designed with low complexity, making them efficient compared to other architectures. However, 
this simplicity may pose a limitation when dealing with very large datasets. For instance, these models might not 
perform optimally with training sets comprising millions of samples. This limitation can be mitigated by increasing 
the number of filters in the DWSC layers, which, thanks to the efficient convolution application of DWSCs, would 
result in only a slight increase in computational cost.
Assuming an input MTS of $M$ channels and the output target dimension we want is $M^{'}$ produce by a convolution layer 
with kernel size $K$, then the ratio between number of parameters needed in the cases of standard and DWSC layers is:
\begin{equation}\label{equ:lite-std-vs-dwsc-params}
\begin{split}
    Ratio &= \dfrac{number~of~parameters~standard}{number~of~parameters~DWSC}\\
    &= \dfrac{M.M^{'}.K}{M.K + M.M^{'}}\\
    &= \dfrac{M^{'}.K}{K+M^{'}}
\end{split}
\end{equation}

\begin{figure}
    \centering
    \caption{Number of parameters of Standard Convolutions and DWSCs in function of number of convolution filters 
    to learn and their kernel size.}
    \label{fig:params-convolution-ctr}
    \includegraphics[width=\textwidth]{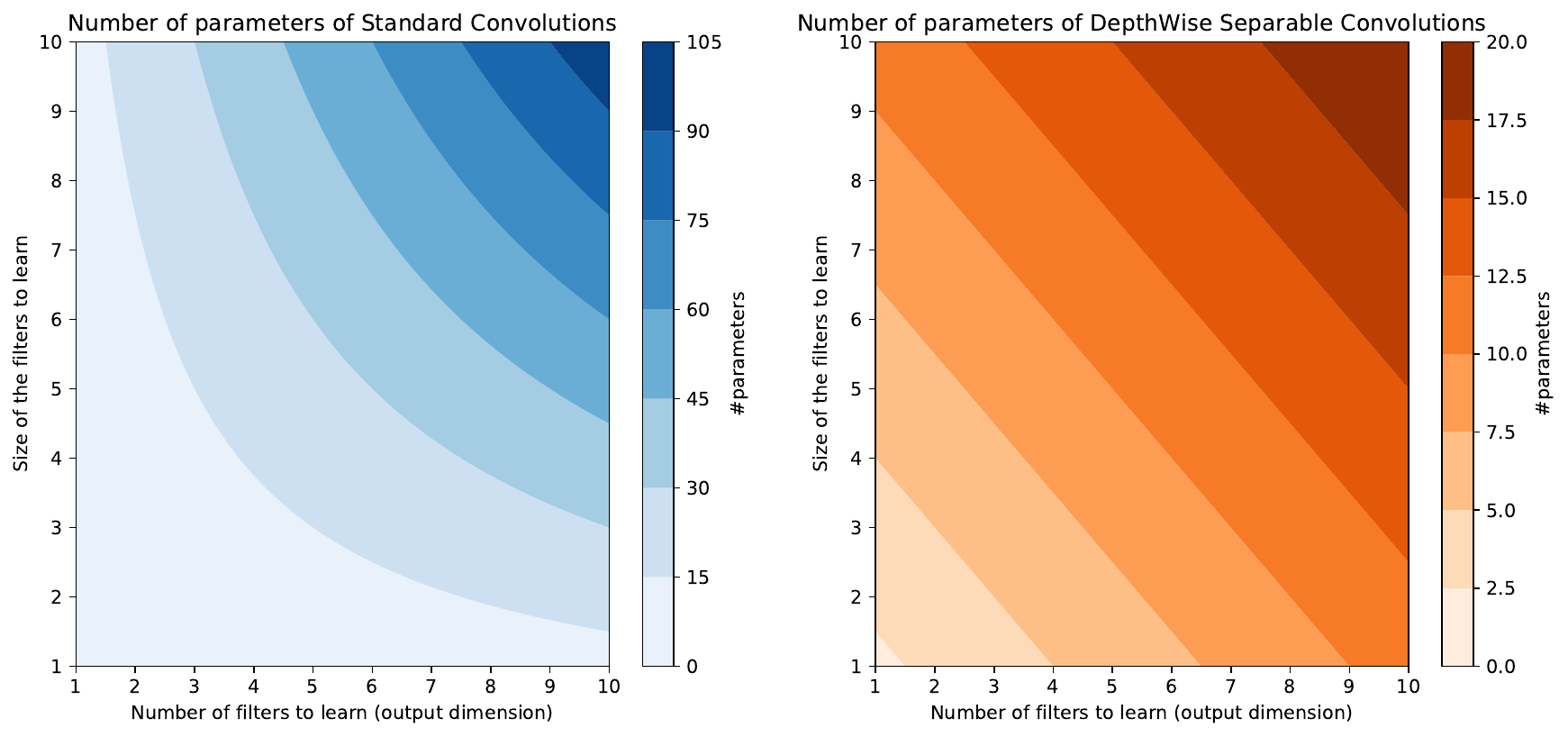}
\end{figure}

As seen in Eq.~\ref{equ:lite-std-vs-dwsc-params} and illustrated in Figure~\ref{fig:params-convolution-ctr},
independently of the number of input channels $M$, the number of parameters 
of the Standard Convolutions increases much faster compared to DWSCs.
For instance, the number of parameters in Standard Convolutions increases quadratically, while Depthwise Separable
Convolutions (DWSCs) exhibit a linear increase.

A second limitation common to all architectures discussed in this work relates to the length of the time series 
samples. This can be addressed by enhancing the CNN's Receptive Field (RF), which determines the length of 
the input visible to the CNN at the last layer. For a CNN with $\Lambda$ convolution layers, each with kernel size $K_i$
and dilation rate $d_i$ where $i~\in~[1,\Lambda]$, the RF is calculated as:

\begin{equation}\label{equ:receptive-field}
    RF = 1+\sum_{i=1}^{\Lambda}~d_i.(K_i-1)
\end{equation}

The RF varies between different CNN models. For example, the RF for FCN~\cite{fcn-resnet-mlp-paper}
is $14=1+7+4+2$, which is relatively small 
compared to the time series lengths in the UCR archive. In contrast, ResNet has an RF of $40$, and Inception extends 
this to $235$. For LITE and LITEMV, the RF is $114$, which is sufficient for state-of-the-art performance on the UCR archive. 
However, for datasets with much longer time series, this RF needs to be increased.

The RF can be expanded by either increasing the filter lengths or adding more layers to deepen the model. 
While this typically leads to a substantial increase in network complexity for conventional CNNs, LITE and 
LITEMV can accommodate these adjustments without significant complexity increases, making them well-suited for 
handling longer time series while maintaining their efficiency and performance.

\section{Conclusion}

In this chapter we introduced LITE: a lightweight, Inception-based architecture enhanced with boosting techniques for TSC.
Through rigorous experimentation and analysis, we demonstrated that LITE achieves competitive 
performance while maintaining a significantly lower number of parameters compared to more complex models like
Inception, FCN, and ResNet.

The ensemble approach, LITETime, further capitalizes on the strengths of LITE, reducing variance and enhancing robustness. 
We explored the impact of boosting techniques such as hand-crafted convolution filters, multiplexing convolutions, and dilated 
convolutions through comprehensive ablation studies, confirming their contributions to the model's overall efficacy.

Moreover, recognizing the importance of multivariate time series classification in real-world applications, 
we introduced LITEMV. This adaptation preserves the efficiency of LITE while extending its capabilities to
handle the complexities of multivariate data. Our experiments confirmed that LITEMV performs exceptionally 
well in its ensemble version LITEMVTime, often surpassing more memory-intensive models like ConvTran in various scenarios.

We also addressed the limitations of our proposed architectures, particularly in handling extremely large 
datasets. We discussed potential solutions such as increasing the number of filters in 
DWSC layers and expanding the receptive field, which LITE and LITEMV can achieve with minimal increase in complexity.

In summary, LITE and LITEMV present a significant advancement in time series classification, offering a 
balance of efficiency, performance, and adaptability. These models are well-suited for deployment in 
resource-constrained environments, making them practical for a wide range of real-world applications.
\chapter{Semi-Supervised and Self-Supervised Learning for Time Series Data with a Lack of Labels}\label{chapitre_5}

\section{Introduction}

The challenge of effectively classifying time series data has garnered significant attention within the 
field of machine learning. Previous chapters have discussed traditional approaches, which most of the time
primarily rely on supervised learning techniques demanding the availability of labeled data. However, acquiring sufficient 
labeled time series data is often prohibitively difficult due to the need for expert annotation and the inherent 
complexity of the data itself. Consequently, there is a growing interest in methodologies that can leverage 
limited labeled data while making the most of the abundant unlabeled data available.

This chapter explores semi-supervised and self-supervised learning techniques as promising methods to address 
the issue of time series classification when labeled data is scarce. Semi-supervised learning and 
self-supervised learning methods aim to reduce the dependency on labeled data by utilizing unlabeled 
data to improve model performance. These techniques have shown considerable potential in 
various domains, and are becoming increasingly popular within the time series data mining research community.

In this chapter, we propose a novel self-supervised approach for enhancing TSC.
Our method, 
named TRIplet Loss In TimE (TRILITE), is built upon the concept of triplet loss, a mechanism traditionally used 
in tasks like face recognition to learn effective representations without the need for extensive labeled data
(see Chapter~\ref{chapitre_1} Section~\ref{sec:self-supervised}). 
TRILITE employs a specialized augmentation technique adapted to the characteristics of time series data, allowing 
the model to learn discriminative features from unlabeled data.

We investigate two specific use cases to evaluate the efficacy of TRILITE. The first scenario considers the augmentation 
of a supervised classifier's performance when only a small amount of labeled data is available. Here, TRILITE acts as a 
booster, enhancing the classifier by providing additional, meaningful representations. The second scenario addresses a 
semi-supervised learning context, where the dataset comprises both labeled and unlabeled samples.In this scenario, TRILITE 
is utilized to effectively harness the unlabeled data, resulting in improved overall classification accuracy.

Through extensive experiments conducted on 85 datasets from the UCR archive, we demonstrate the potential of 
TRILITE in both scenarios. The results indicate that our approach not only boosts performance in low-labeled 
data settings but also effectively incorporates unlabeled data to create more robust classifiers. This chapter 
outlines the methodology, experimental setup, and findings, contributing to the broader understanding of 
semi-supervised and self-supervised learning in time series classification.

By addressing the limitations of traditional supervised learning models and harnessing the power of unlabeled data, 
this work paves the way for more efficient and scalable solutions in the analysis and classification of time series data.

\section{TRILITE: TRIplet Loss In TimE}

This section presents the proposed self-supervised learning approach for time series classification, named TRIplet 
Loss In TimE (TRILITE). Our approach leverages triplet loss to learn meaningful representations from time series 
data without requiring extensive labeled data. We describe the architecture of the TRILITE model, the triplet 
loss mechanism, and the specific data augmentation techniques employed to generate effective triplets.
The term \emph{data augmentation} in this chapter does not mean training on more samples, it simply means 
transforming the input series to a new series that is somehow similar to the reference series.

\subsection{Model Construction}

\begin{figure}
    \centering
    \caption{Overview of our TRILITE model with a triplet input example taken form the Beef dataset 
    of the UCR archive~\cite{ucr-archive}.}
    \label{fig:trilite-model}
    \includegraphics[width=0.6\textwidth]{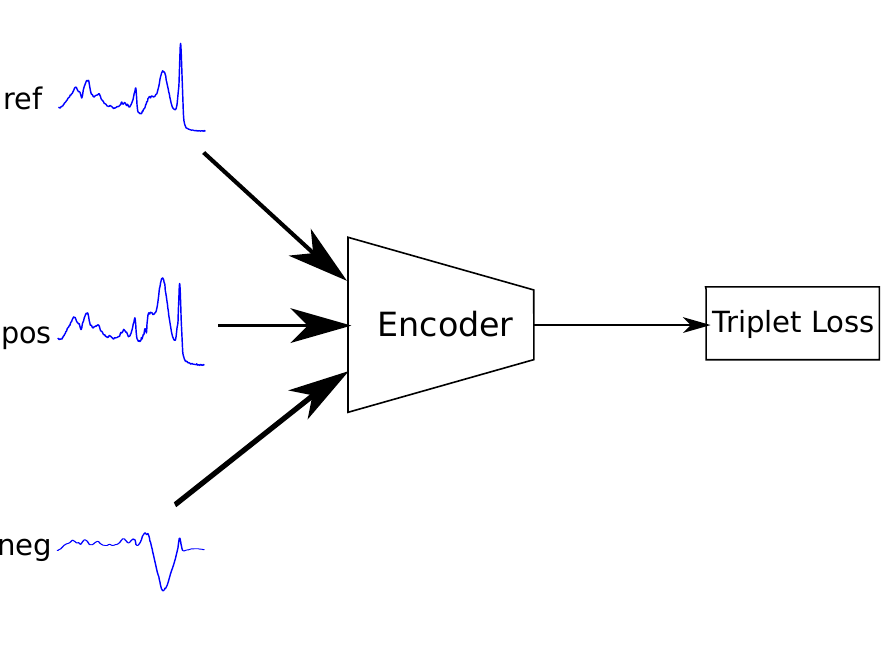}
\end{figure}

Our TRILITE model features a trio of encoders, all sharing identical weights to ensure consistency. 
This configuration effectively functions as a single encoder processing the generated triplets. We have 
adopted the FCN architecture~\cite{fcn-resnet-mlp-paper}, but modified it by removing the classification layer 
to suit our self-supervised learning framework. Each component of the triplet, the reference $ref$, 
positive $pos$, and negative $neg$ samples, is input into the model, producing their respective latent 
representations ($ref_l$, $pos_l$, and $neg_l$).
In this case, positive and negative samples refer to similar and dissimilar representations of the anchor (reference)
sample.
These representations are streamlined to a fixed size of 128 
dimensions, enabling robust and efficient feature extraction.
The three representations are then used to calculate the triplet loss, defined in the following section,
as presented in Figure~\ref{fig:trilite-model}.

\subsection{Triplet Loss}\label{sec:trilite-loss}

\begin{figure}
    \centering
    \caption{Schema of the relaxed spaced controlled by the
    margin $\alpha$ with an input example taken form the Beef dataset 
    of the UCR archive~\cite{ucr-archive}.}
    \label{fig:trilite-margin}
    \includegraphics[width=0.6\textwidth]{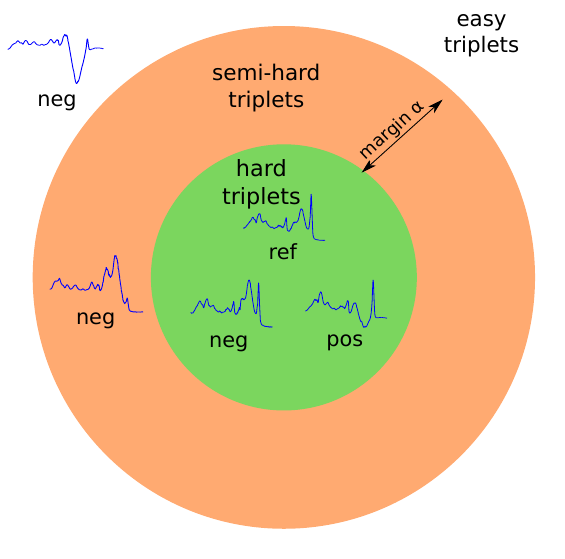}
\end{figure}

The core of our approach is the triplet loss~\cite{facenet-paper}, which is designed to create a discriminative latent space 
by minimizing the distance between a reference sample and its positive representation while maximizing the 
distance between the reference and a negative representation. Formally, the triplet loss for a given triplet 
(reference, positive, and negative) is defined as:

\begin{equation}\label{equ:trilite-loss}
    \mathcal{L}_{triplet}(ref_l,pos_l,neg_l) = \max(0, \alpha+d(ref_l,pos_l) - d(ref_l,neg_l))
\end{equation}

where $d(.,.)$ is the Euclidean Distance and $\alpha$ is a margin hyperparameter that controls the separation between 
positive and negative pairs. The loss encourages the model to learn embeddings where similar samples are close 
together, and dissimilar samples are far apart.

The objective of the triplet loss function is to maximize the distance between the reference latent representation 
$ref_l$ and the negative latent representation $neg_l$, while minimizing the distance between the reference and 
positive latent representations $pos_l$. 
Consequently, we can identify three situations for triplets
as illustrated in Figure~\ref{fig:trilite-margin}:

\begin{itemize}
    \item \textbf{Easy Triplet}: The loss equals $0$ because $d(ref_l,pos_l)+\alpha<d(ref_l,neg_l)$
    \item \textbf{Hard Triplet}: The negative representation is closer to the reference than the positive representation,
    i.e. $d(ref_l,neg_l) < d(ref_l,pos_l)$
    \item \textbf{Semi-Hard Triplet}: The positive representation is closer than the negative, i.e. 
    $d(ref_l,pos_l) < d(ref_l,neg_l)$, but there is still a strictly positive loss.
\end{itemize}

Setting $\alpha$ to $0$ would limit us to identifying only easy and hard triplets. Exclusively using easy triplets 
would likely cause the model to overfit, while relying solely on hard triplets could lead to underfitting. 
Therefore, the introduction of the $\alpha$ hyperparameter is crucial, as it facilitates the creation of semi-hard 
triplets, striking a balance between these extremes. Moreover, the incorporation of the max operation in the 
loss function ensures that the optimization problem remains convex, promoting more effective and stable training.

In the following section, we detail the proposed method of generating these triplets $ref$, $pos$ and $neg$.

\subsection{Triplet Generation}\label{sec:trilite-triplets}

\begin{algorithm}
\caption{Triplet\_Generation}
\label{alg:trilite-triplets}
\begin{algorithmic}[1]
    \REQUIRE~A time series dataset $\mathcal{D}$ of $N$ samples of length $L$ each
    \ENSURE~Three sets of triplets $ref$ $pos$ and $neg$ of same shape as $\mathcal{D}$ each
    \STATE~$shuffle(\mathcal{D})$
    \STATE~$w \gets random(0.6,1)$ 
    \FOR{$i$~: $0 \rightarrow N$}
        \STATE~$ref[i]  \gets data[i]$
        \STATE~$ts_1 \gets random\_sample(data)$
        \STATE~$ts_2 \gets random\_sample(data)$
        \STATE~$pos[i] \gets w.ref[i] + (\frac{1-w}{2}).(ts_1+ts_2)$
        \STATE~$ts_3 \gets random\_sample(data)$
        \STATE~$ts_4 \gets random\_sample(data)$
        \STATE~$ts_5 \gets random\_sample(data)$
        \STATE~$neg[i] = w.ts_3 + (\frac{1-w}{2}).(ts_4+ts_5)$
        \STATE~$pos[i],neg[i] \gets Mask(pos[i],neg[i])$
    \ENDFOR
    \STATE~$pos \gets Znormalize(pos)$
    \STATE~$neg \gets Znormalize(neg)$
    \STATE~\textbf{Return} $ref,pos,neg$
\end{algorithmic}
\end{algorithm}

\begin{algorithm}
\caption{Mask}
\label{alg:trilite-masking}
\begin{algorithmic}[1]
    \REQUIRE~Two input time series $x$ and $y$ of length $L$
    \ENSURE~Masked version of $x$ and $y$
    \State~$l \gets len(x)$
    \State~$start \gets random\_randint(0,L-1)$
    \State~$stop \gets random\_randint(start+\frac{L-1-start}{10},start+\frac{L-1-start}{2.5})$
    \State~$x[0:start] \gets noise$
    \State~$x[stop+1:] \gets noise$
    \State~$y[0:start] \gets noise$
    \State~$y[stop+1:] \gets noise$
    \State~\textbf{Return} $x,y$
\end{algorithmic}
\end{algorithm}

Generating effective triplets is crucial for the success of the TRILITE model. We combine two main strategies for triplet
generation: mixing up and masking.
In~\cite{mixing-up-paper}, a mixing-up strategy is employed, while in~\cite{triplet-loss-paper},
a masking approach is utilized.
The proposed triplet generation setup is detailed in
Algorithms~\ref{alg:trilite-triplets} and~\ref{alg:trilite-masking}.

First, we generate a positive sample $pos$ by computing a weighted sum of three time series, including the reference
$ref$. For the negative sample $neg$, the reference is excluded from the weighted sum. We limit the mixed samples
to three, ensuring each sample contributes significantly. The process can be represented by the following equations:

\begin{equation}\label{equ:trilite-pos-gen}
    pos = w.ref+\dfrac{1-w}{2}.(ts_1+ts_2)
\end{equation}
\begin{equation}\label{equ:trilite-neg-pos}
    neg = w.ts_3+\dfrac{1-w}{ts_4+ts_5}
\end{equation}

\noindent where $ts_1$ and $ts_2$ are randomly selected time series distinct from the reference, and the contribution 
weight $w$ is randomly chosen between $0.6$ and $1.0$. This ensures the positive sample has a greater influence 
from the reference compared to $ts_1$ and $ts_2$.
For the negative sample, three distinct samples, $ts_3$, $ts_4$ and $ts_5$ are randomly chosen from the training dataset 
excluding the $ref$ sample.

Next, we apply a random-length mask to both the positive and negative samples. This masking strategy simplifies 
the training process by allowing the model to focus on learning specific segments of the representations 
rather than the entire sequence.

\begin{figure}
    \centering
    \caption{
        A \protect\mycolorbox{255,165,0,0.6}{mixed up $pos$} is built from
        \protect\mycolorbox{0,90,255,0.6}{three 
        time series including the $ref$}.
        The resulting time series is close to
        the $ref$ except some areas as highlighted in the 
        \protect\mycolorbox{255,30,0,0.6}{red circle}. A \protect\mycolorbox{0,164,0,0.6}{mask}
        is then applied on the
        \protect\mycolorbox{255,165,0,0.6}{mixed up $pos$} 
        to generate the \protect\mycolorbox{0,164,0,0.6}{final
        sample}, where the 
        unmasked parts are replaced by a Gaussian noise.
    }
    \label{fig:trilite-pos-gen}
    \includegraphics[width=\textwidth]{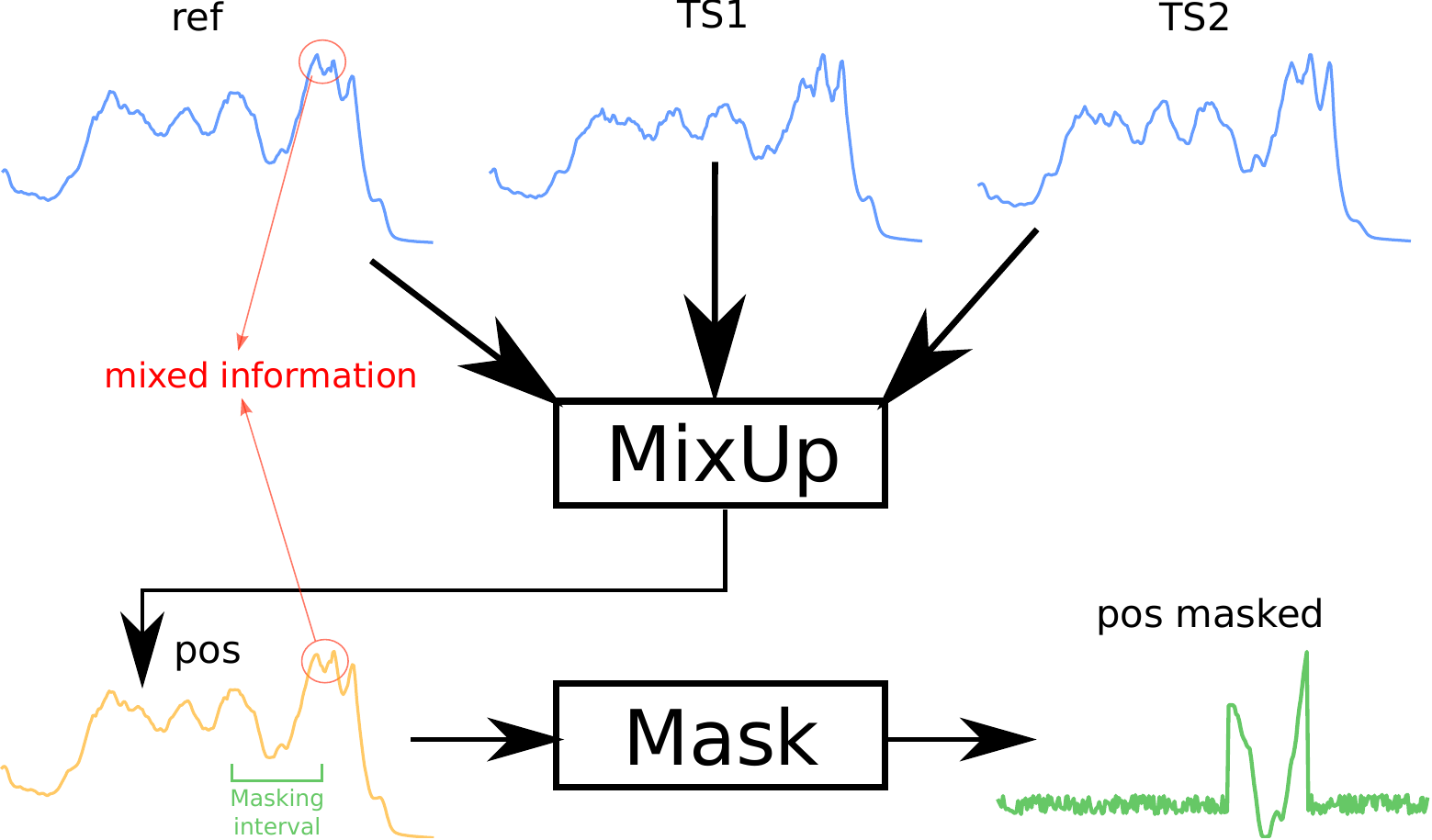}
\end{figure}

\begin{figure}
    \centering
    \caption{
        A \protect\mycolorbox{255,165,0,0.6}{mixed up $neg$} is built from
        \protect\mycolorbox{0,90,255,0.6}{three 
        time series excluding the $ref$} used to generate the $pos$ sample.
        The resulting time series is close to
        the $not~ref$ except some areas as highlighted in the 
        \protect\mycolorbox{255,30,0,0.6}{red circle}. A \protect\mycolorbox{0,164,0,0.6}{mask}
        is then applied on the
        \protect\mycolorbox{255,165,0,0.6}{mixed up $neg$} 
        to generate the \protect\mycolorbox{0,164,0,0.6}{final
        sample}, where the 
        unmasked parts are replaced by a Gaussian noise.
    }
    \label{fig:trilite-neg-gen}
    \includegraphics[width=\textwidth]{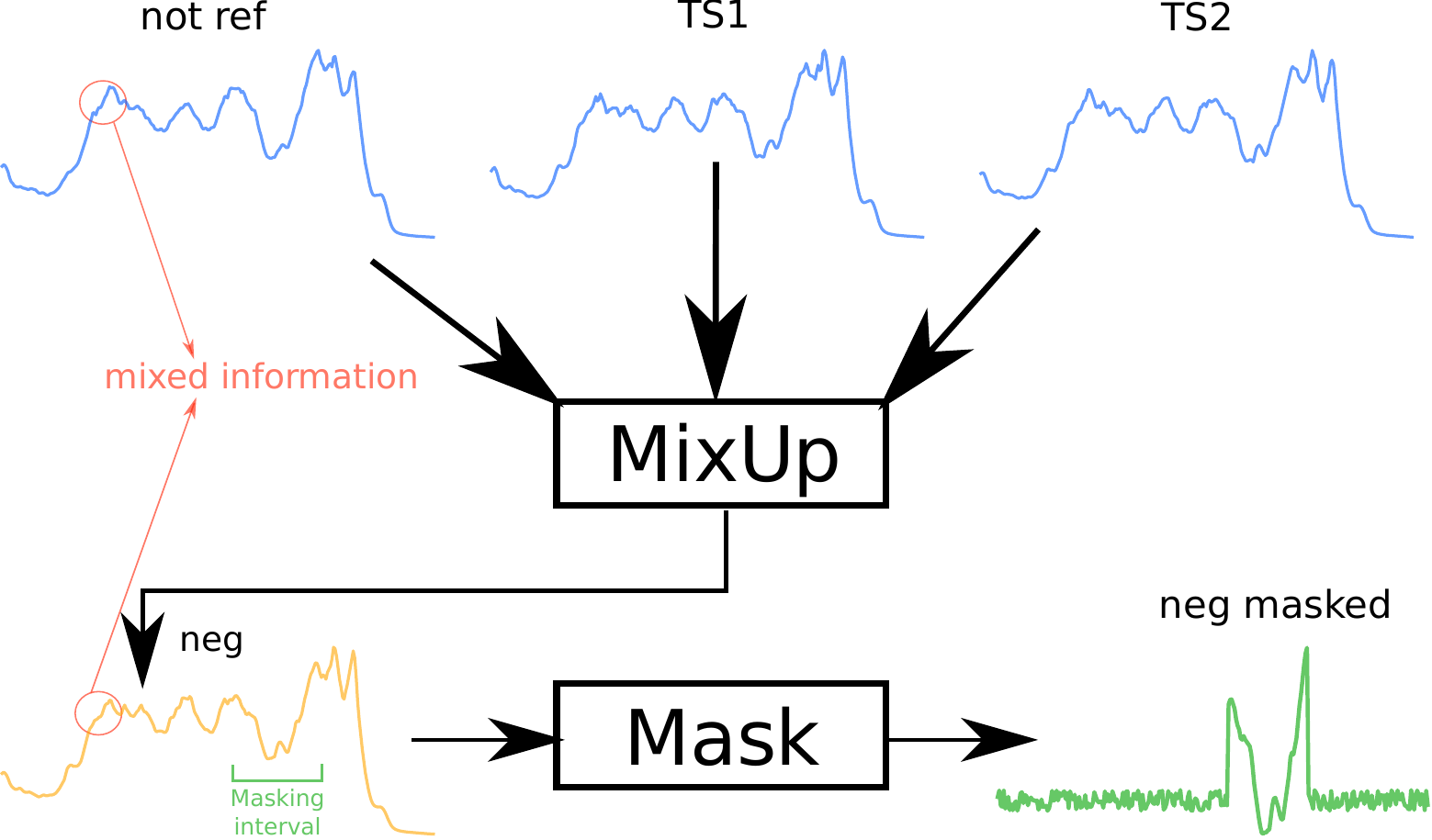}
\end{figure}

Finally, the unmasked segments of the time series are replaced with random Gaussian noise, enhancing the 
robustness of the model. Figures~\ref{fig:trilite-pos-gen} and~\ref{fig:trilite-neg-gen} provides 
a visualization of the positive and negative samples generation. Importantly, 
triplet generation occurs online during each training epoch, promoting better generalization of the model.


\section{Experimental Setup}

For our experiments, we utilized 85 datasets from the UCR
archive~\cite{ucr-archive}~\footnote{More experiments will be done for this technique.}.
All datasets were z-normalized 
prior to training. We employed the Adam optimizer with an initial learning rate of $10^{-3}$. Triplet generation 
occurred online for each epoch to ensure robust generalization. For evaluation on the test set, we used the final 
trained model.
The models were trained for $1000$ epochs with batch size $32$.

To identify the optimal hyperparameter $\alpha$, we conducted a thorough exploration across a range of values:
$\{10^{-6},10^{-5},10^{-4},10^{-3},10^{-2},10^{-1}\}$.
By meticulously visualizing the resulting latent representations, we discerned that modifying $\alpha$
primarily affected the scale of these representations without altering the fundamental classification of 
triplet types (as elaborated in Section~\ref{sec:trilite-loss} of this chapter).
This nuanced understanding led us to conclude that an $\alpha$ value 
of $10^{-2}$ struck the right balance. This value was meticulously fine-tuned using a representative subset 
of the UCR archive, ensuring robustness and generation in our model's performance.

\section{Experimental Results}

\subsection{Comparing To State-Of-The-Art}

\begin{figure}
    \centering
    \caption{Comparing the proposed TRILITE approach to two state-of-the-art SSL models: DCNN~\cite{triplet-loss-paper}
    and MCL~\cite{mixing-up-paper}. We compare the classification of TRILITE latent spaces to MCL's latent space using 
    a 1NN and to DCNN's latent space using both 1NN and a Support Vector Machine classifier~\cite{svm-paper}
    following the original work of both comparates.}
    \label{fig:trilite-vs-sota}
    \begin{subfigure}{0.32\linewidth}
        \centering
        \includegraphics[width=\linewidth]{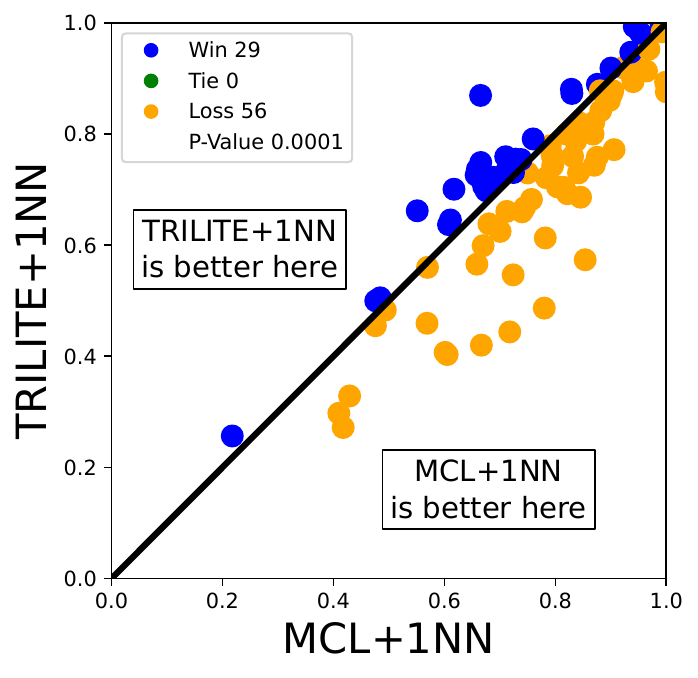}
        \caption{\null}
    \end{subfigure}
    \begin{subfigure}{0.32\linewidth}
        \centering
        \includegraphics[width=\linewidth]{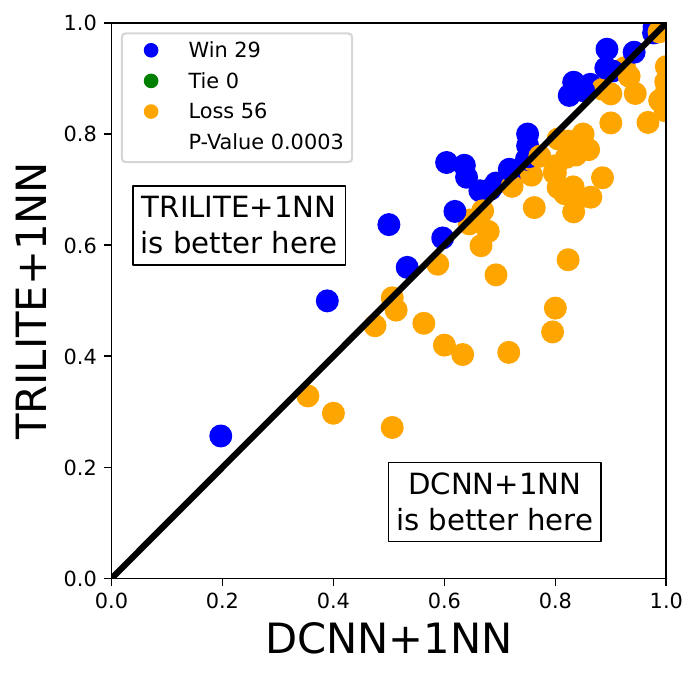}
        \caption{\null}
    \end{subfigure}
    \begin{subfigure}{0.32\linewidth}
        \centering
        \includegraphics[width=\linewidth]{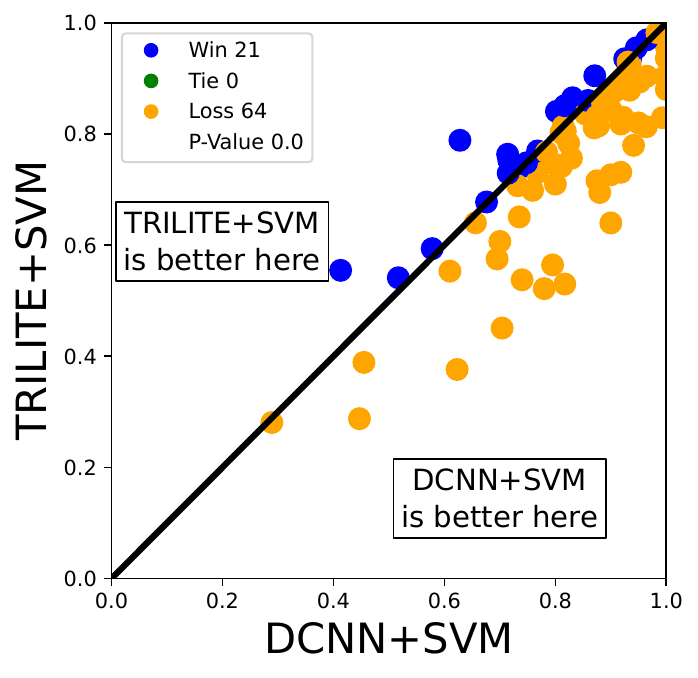}
        \caption{\null}
    \end{subfigure}
\end{figure}

Comparing our approach to the two state-of-the-art models, Dilated Convolutional Neural Network (DCNN) with its 
version of triplet loss~\cite{triplet-loss-paper} and Mixup Contrastive Learning (MCL)~\cite{mixing-up-paper}, 
is essential as this work draws significant motivation from their methodologies.
However, the goal was not to outperform these models as our objectives differ.
In this study, the aim is to demonstrate how self-supervised models can enhance supervised learning, particularly
in scenarios with limited data and a lack of labeled data.
For each self-supervised 
model, DCNN, MCL, and TRILITE (ours), we applied a classifier to their latent features to evaluate performance,
posterior to training. 
Specifically, we compared TRILITE and MCL using a 1NN classifier with Euclidean distance, consistent with the 
evaluation method used in~\cite{mixing-up-paper}. Additionally, we compared TRILITE and DCNN
with both 1NN and SVM classifiers, as presented in~\cite{triplet-loss-paper}.
The choice of classifier when comparing to each of the two comparators aligns with those used in the original
experiments of DCNN and MCL.
We present in Figure~\ref{fig:trilite-vs-sota} the three 1v1 scatter plots illustrating these comparisons.
Although our model does not consistently surpass the performance of DCNN and MCL across 
all datasets, it does perform comparably on several datasets. This indicates that TRILITE has potential. Now, 
we will present the two cases addressing the challenges of small labeled datasets and the lack of labeled samples.
In scenarios with small labeled datasets, the primary challenge is that models tend to overfit, as they struggle 
to learn patterns that generalize well beyond the limited training data. Such scenario results in poor performance on unseen 
data, limiting the model's utility. Moreover, the lack of labeled samples poses a significant challenge 
in training supervised models, as they rely on labeled data to learn associations between inputs and outputs. 
Without sufficient labeled data, models cannot effectively learn or make accurate predictions, leading to 
unreliable outcomes in practical applications.

\subsection{Use Case 1: Small Annotated Time Series Datasets}

\begin{figure}
    \centering
    \caption{
        Scatter plots comparing the proposed TRILITE model to FCN.
    }
    \label{fig:trilite-vs-fcn-and-concat}
    \begin{subfigure}{0.45\linewidth}
        \centering
        \includegraphics[width=\textwidth]{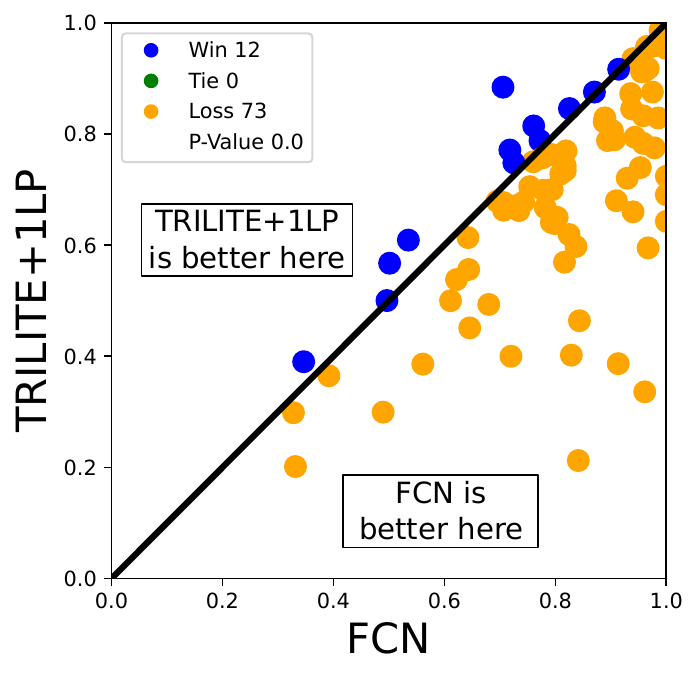}
        \caption{TRILITE+1LP VS FCN}
        \label{fig:trilite-vs-fcn}
    \end{subfigure}
    \begin{subfigure}{0.45\linewidth}
        \centering
        \includegraphics[width=\textwidth]{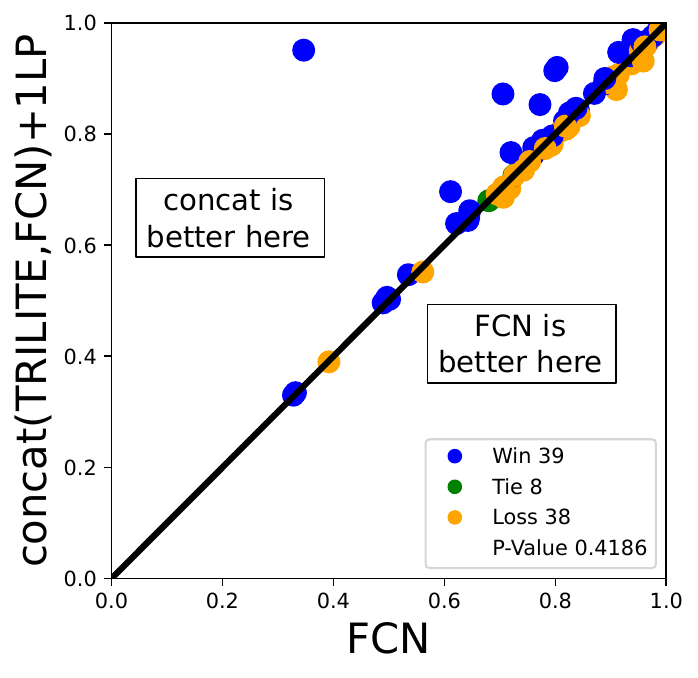}
        \caption{concat(TRILITE, FCN)+1LP VS FCN}
        \label{fig:trilite-concat-vs-fcn}
    \end{subfigure}
\end{figure}

To address the first case of having a small annotated time series dataset, we compared the TRILITE model, 
followed by a fully connected layer with $softmax$ activation (denoted as TRILITE+1-LP), against the fully supervised 
FCN model.
It is important to note that the TRILITE+1LP approach is a two step training, first the TRILITE is trained on the self-supervised 
task and this is subsequently followed by training a 1LP classifier on the pre-trained TRILITE's latent space.
The 1v1 scatter plots are reported in Figure~\ref{fig:trilite-vs-fcn}.
As expected, the supervised model generally 
outperforms the self-supervised one. However, for certain datasets, the self-supervised features notably improve 
classification accuracy. This observation motivated us to explore the contribution 
of self-supervised features within a supervised learning context.

To do this, we concatenated the latent representations from the self-supervised TRILITE model (each of size $128$) 
with those from the supervised FCN model (also of size $128$) for both the training and test sets. The concatenated 
features were then used to train a classifier comprising a single fully connected layer with softmax activation (1LP).
Subsequently, this pipeline is evaluated on the concatenated features of the test set. 
We compared this approach, denoted as concat(TRILITE,FCN)+1LP, against the fully supervised FCN in
Figure~\ref{fig:trilite-concat-vs-fcn}.
This Win-Tie-Loss comparison highlights that the concatenation method is never significantly worse than the single FCN, 
in terms of magnitude of accuracy difference. 
This can be attributed to the fact that supervised features are not negatively impacted by the SSL features; in the worst 
case, the linear classifier can simply ignore the SSL features if they do not aid classification.

\begin{figure}
    \centering
    \caption{The MCM (Chapter~\ref{chapitre_2}) comparing concat(TRILITE,FCN)+1LP to FCN and TRILITE+1LP
    over the 85 datasets of the UCR archive.}
    \label{fig:trilite-mcm}
    \includegraphics[width=\textwidth]{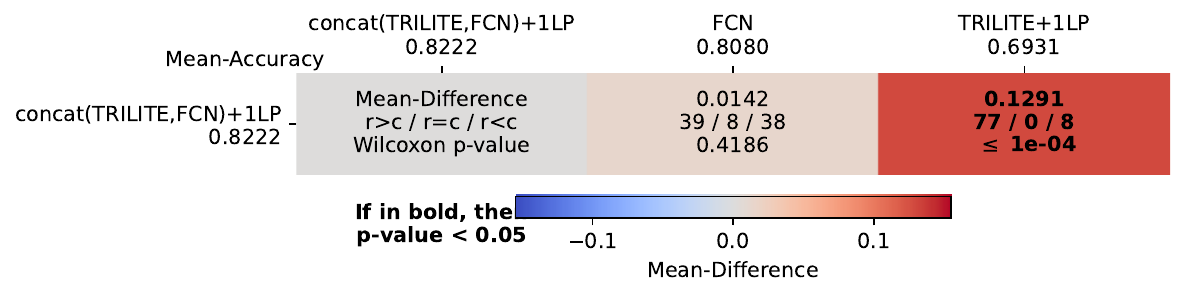}
\end{figure}

Furthermore, we present in Figure~\ref{fig:trilite-mcm} the MCM comparing the three models: TRILITE+1LP, FCN, 
and the concatenation method concat(TRILITE, FCN)+1LP. The MCM highlights that the concatenation method 
is nearly $2\%$ better in terms of average accuracy, demonstrating the boosting effect.
This indicates that SSL generates features distinct from those produced by supervised learning. Consequently, 
the combination of both sets of features enhances classification performance. Notably, the peaks in performance 
improvement are observed in datasets with a small number of samples, such as DiatomSizeReduction, which has only 
16 samples in the training set.

\subsection{Use Case 2: Partially Annotated Time Series Datasets}

\begin{figure}
    \centering
    \caption{
        Comparison of experiment 1 and experiment
2. In experiment 1, the TRILITE model is trained only
on the labeled subset ($30\%$ of the data). On the contrary, in
experiment 2, the TRILITE model is trained on the whole
train set. The evaluation is done on the whole test set.
    }
    \label{fig:trilite-semi-supervised}
    \includegraphics[width=0.45\textwidth]{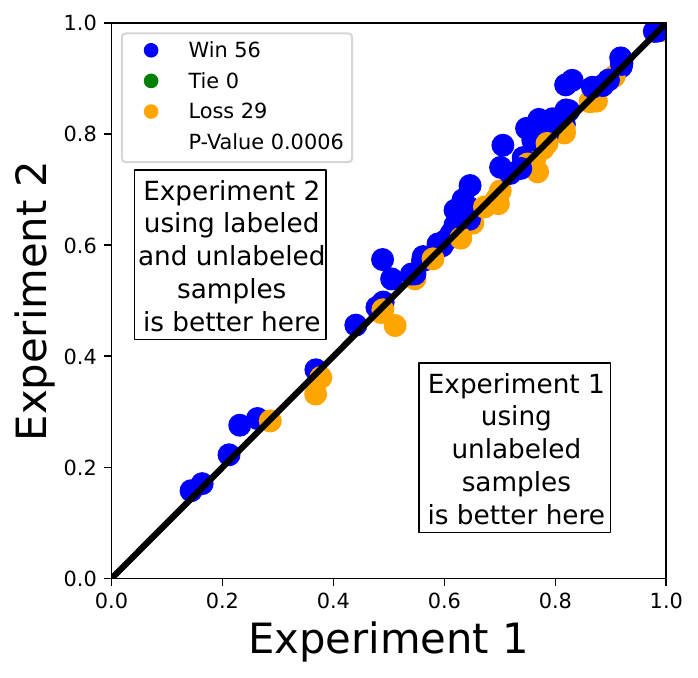}
\end{figure}

In this second case, we explore a semi-supervised scenario where only a portion of the data is labeled. 
Our objective is to assess how self-supervised learning can address the challenge of limited labels. Assuming that only $30\%$ of 
the training set is labeled, we proceed with the following steps:

\begin{enumerate}
    \item \textbf{Self-supervised training}: We generate self-supervised latent representations by training our TRILITE model:
    \begin{itemize}
        \item \textbf{Experiment 1}: Training is conducted solely on the labeled subset.
        \item \textbf{Experiment 2}: Training is conducted on the entire training set,
        including both labeled and unlabeled data.
    \end{itemize}
    \item \textbf{Supervised learning}: The latent representations derived from the labeled set 
    (from either Experiment 1 or Experiment 2) are fed into a Ridge classifier~\cite{hoerl1970ridge}.
    \item \textbf{Evaluation}: The performance of the trained classifier is then evaluated on the test set.
\end{enumerate}

To ensure the reliability and robustness of our results, these steps are repeated across 25 runs, 
with the average accuracy calculated for each run.
The high number of experiments motivated the usage of Ridge classifier instead of SVM 
as proposed by~\cite{triplet-loss-paper} given its fast training time.
The same labeled subset is utilized for both experiments 
within each run. The 1v1 scatter plot comparison between \textbf{Experiment 1} and \textbf{Experiment 2}
is illustrated in Figure~\ref{fig:trilite-semi-supervised}.

The comparison reveals that Experiment 2 outperforms \textbf{Experiment 1} more frequently.
On average, when \textbf{Experiment 2} 
prevails, the accuracy difference is 2.12 ± 2.13. Conversely, when Experiment 1 has better performance, 
the accuracy difference averages 1.17 ± 1.21. 
This disparity underscores the effectiveness of self-supervised learning in cultivating more nuanced 
and comprehensive latent representations. By integrating both labeled and unlabeled data, 
\textbf{Experiment 2} is able to capture a wider spectrum of underlying patterns and structures within the data. 
This setup not only enhances the overall performance of the model but also demonstrates that self-supervised methods
contribute to greater stability and adaptability across diverse datasets. These findings emphasize the potential 
of self-supervised learning to bridge gaps in data quality and quantity, leading to models that are better 
equipped to generalize across different tasks and challenges.
\section{Conclusion}

In this chapter, we have explored innovative approaches in semi-supervised and self-supervised learning for 
TSC. The primary 
goal of our research was to utilize self-supervised models to enhance supervised models in two specific cases: 
first, when there is a lack of data but all available data is labeled, and second, when there is a scarcity of 
labeled data in a semi-supervised learning context.

Our TRILITE model, based on triplet loss, was developed to generate meaningful latent representations from 
time series data. We conducted extensive experiments comparing TRILITE with state-of-the-art models such as 
DCNN~\cite{triplet-loss-paper} and MCL~\cite{mixing-up-paper}.
While TRILITE did not consistently outperform these models, it demonstrated competitive 
performance on several datasets. This indicates the potential of self-supervised learning in improving 
TSC, particularly in challenging scenarios with limited labeled data.

In the first use case, where only a small annotated time series dataset is available, we showed that 
incorporating self-supervised features with supervised learning models can enhance classification 
accuracy. Specifically, by concatenating the latent representations from the TRILITE model with those 
from a supervised FCN model, we observed significant improvements in performance. This approach, 
termed concat(TRILITE, FCN), consistently achieved comparable results to the single FCN model and better performance 
on small datasets, showcasing 
the complementary nature of self-supervised and supervised features.

In the second use case, involving a semi-supervised scenario with only a portion of the data labeled, 
we demonstrated how self-supervised learning can effectively address the lack of labels. By training 
the TRILITE model on both the labeled subset and the entire training set, we obtained more robust latent 
representations that improved the performance of the downstream classifier. The results from repeated 
experiments confirmed that leveraging unlabeled data in self-supervised training leads to more meaningful 
feature extraction, which in turn enhances classification accuracy.

The findings in this chapter underscore the importance of integrating self-supervised learning techniques 
to enhance supervised learning models, especially in the context of time series data with limited labeled 
samples. The potential of TRILITE to generate useful features from both labeled and unlabeled data opens 
new avenues for future research. Moving forward, further refinement of these techniques could lead to more 
robust and efficient models, advancing the field of TSC, and potentially other tasks, addressing the persistent 
challenges of data scarcity and annotation costs.
\chapter{Time Series Analysis For Human Motion Data}\label{chapitre_6}

\section{Introduction}

Human motion data, particularly skeleton-based data, plays a crucial role in various applications such as 
action recognition~\cite{human-motion-example-paper}, rehabilitation assessment~\cite{kimore-paper},
prediction/forecasting~\cite{hm-prediction-paper}, and generation~\cite{actor-paper}
for cinematic and gaming systems. 
This chapter digs into the unique aspects of Multivariate Time Series (MTS) analysis when applied to 
skeleton-based human motion sequences.

\begin{figure}
    \centering
    \caption{Example of one sample per action (12 total) taken from the HumanAct12 
    action recognition
    skeleton based dataset~\cite{action2motion-paper}. Each skeleton is made of five body parts,
    \protect\mycolorbox{161,135,88,1.0}{right arm},
    \protect\mycolorbox{203,182,126,1.0}{left arm},
    \protect\mycolorbox{108,158,157,1.0}{spine \& neck},
    \protect\mycolorbox{125,116,130,1.0}{right leg} and
    \protect\mycolorbox{178,168,183,1.0}{left leg}.
    }
    \label{fig:example-humanact}
    \includegraphics[width=\textwidth]{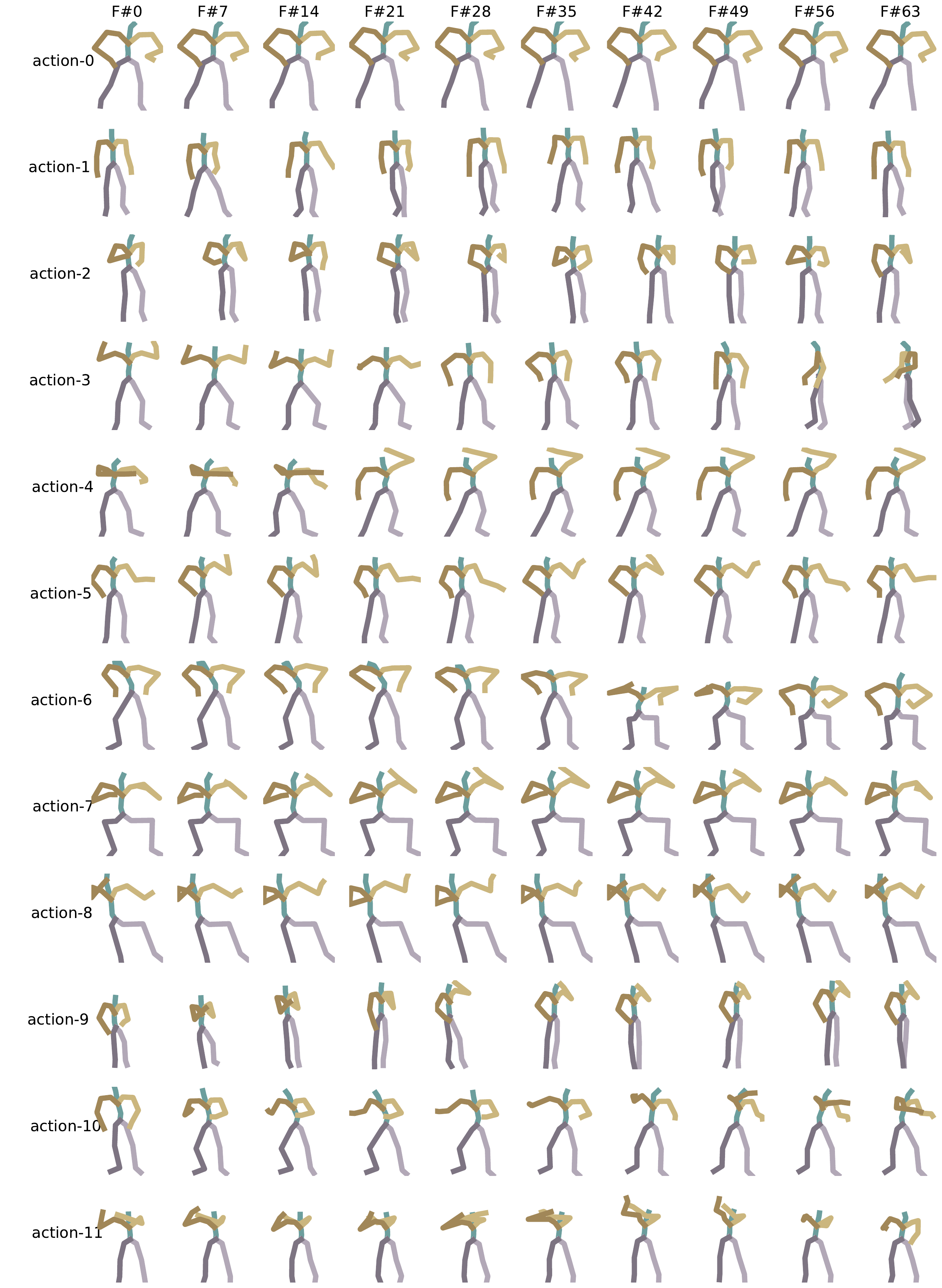}
\end{figure}

Skeleton-based human motion data is primarily extracted using advanced sensing technologies like the Microsoft 
Kinect~\cite{kinect-paper,kinect-survey-paper}, which captures 3D spatial coordinates
of body joints. This data provides a simplified yet informative 
representation of human motion by tracking the positions of key skeletal joints over time. Other technologies, 
such as motion capture systems~\cite{mocap-paper},
also offer detailed skeletal data by using markers placed on the body to record 
joint movements with high precision. These datasets are invaluable for various tasks due to their ability to 
encapsulate the complexity of human motion in a structured format.
One significant advantage of Kinect sensors over traditional MoCap technologies is their inherent synchronization. 
With Kinect, all joints are detected through a single camera operating at a consistent sampling frame rate, 
ensuring uniformity and temporal coherence in the captured data. In contrast, MoCap systems assign individual 
sensors to each joint, which can lead to potential desynchronization issues between the sensors. Moreover, Kinect 
sensors offer the benefits of being both low-cost and non-intrusive, enhancing their accessibility and ease of use. 
These attributes make Kinect an ideal choice for our study, and hence, we have exclusively utilized Kinect-based data 
in this chapter.
An example of such sequences are presented in Figure~\ref{fig:example-humanact}.

The versatility of skeleton-based human motion data lends itself to numerous tasks:

\begin{itemize}
    \item \textbf{Action Recognition}: Identifying specific actions or activities performed by individuals.
    \item \textbf{Motion Assessment}: Monitoring and analyzing patient movements to aid in physical therapy and recovery.
    \item \textbf{Prediction}: Forecasting future movements based on past motion patterns.
    \item \textbf{Generation}: Assessment Creating realistic human movements for use in cinematic productions, gaming environments,
    medical research etc.
\end{itemize}

\begin{figure}
    \centering
    \caption{Example of one sample taken from the HumanAct12 action recognition
    skeleton based dataset~\cite{action2motion-paper} represented as a Multivariate
    Time Series.
    For the sake of visualization we consider only five joints:
    \protect\mycolorbox{0,30,255,0.6}{head}, 
    \protect\mycolorbox{0,0,0,0.3}{left wrist},
    \protect\mycolorbox{255,163,0,0.6}{right wrist},
    \protect\mycolorbox{255,30,0,0.6}{left ankle},
    and \protect\mycolorbox{0,127,0,0.6}{right ankle}, 
    each in a 3D space, resulting in an MTS of $15$ dimensions.}
    \label{fig:example-humanact-to-mts}
    \includegraphics[width=\textwidth]{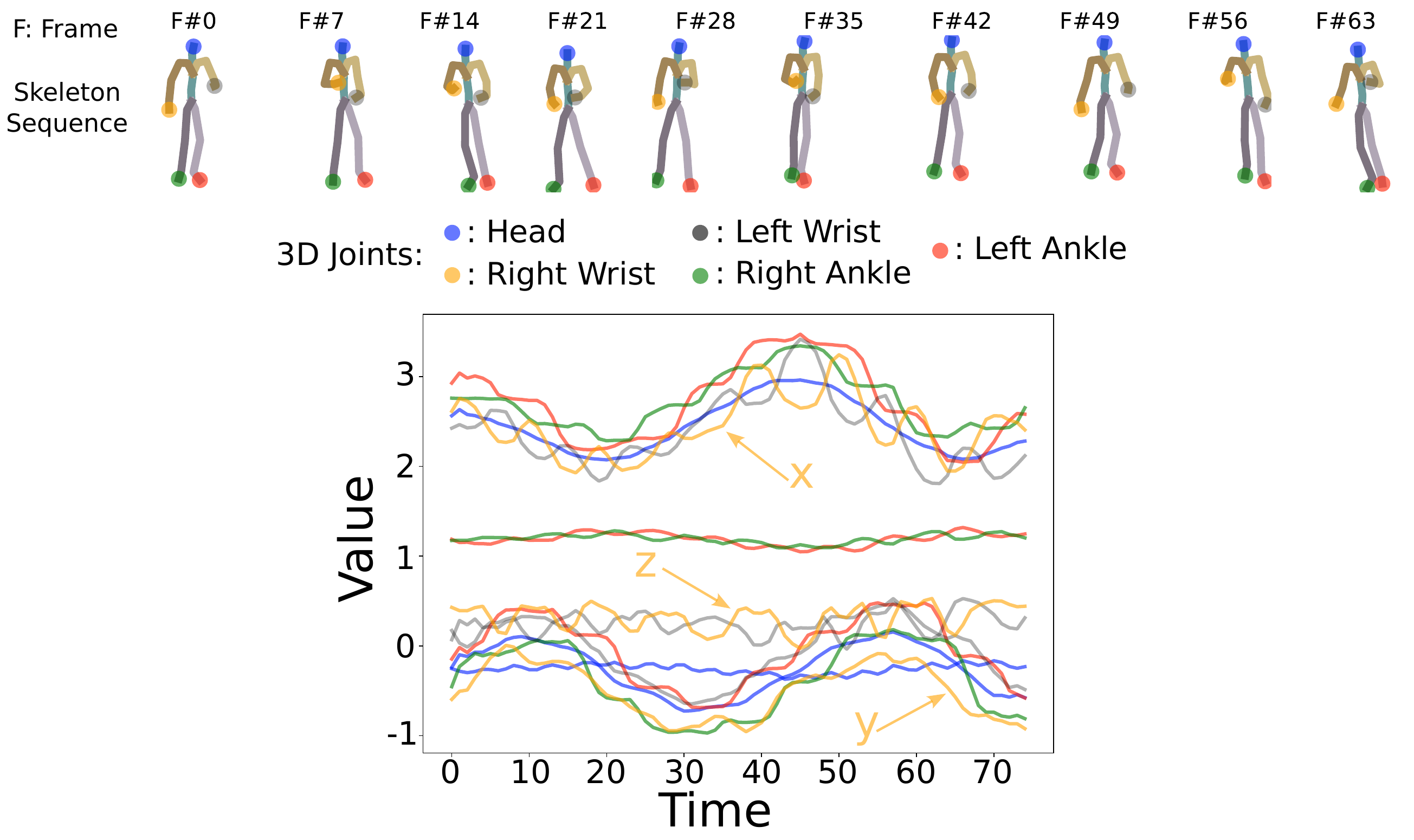}
\end{figure}

Skeleton-based human motion sequences can be effectively represented as MTS. Typically, 
a skeleton sequence has a shape of $(time, number~of~joints, dimensions~per~joint)$. For instance, with 25 
joints tracked in 3D space ($x$, $y$, $z$ coordinates), each time step is characterized by a $75$-dimensional 
vector. This structure can be transformed into a more conventional MTS format of $(time, 75)$. This transformation enables 
the application of standard time series analysis techniques to the data.
An example of this representation is illustrated in Figure~\ref{fig:example-humanact-to-mts}.

In previous chapters, we have explored various deep learning methodologies for time series analysis. 
This chapter will extend that exploration to the domain of skeleton-based human motion data. We will 
assess the effectiveness of deep learning models in addressing specific tasks related to this type of data.

In this chapter, we will explore several key contributions related to the analysis of skeleton-based human motion data.
\textbf{First}, we will investigate the use of deep learning models to assess the quality of
a patient's movement for rehabilitation exercises.
\textbf{Second}, to mitigate overfitting, we will explore techniques to prototype and extend medical datasets, 
ensuring robust model performance. This will be done through a novel Time Series Prototyping (TSP) approach, notably 
ShapeDBA (ShapeDTW Barycenter Average).
\textbf{Third}, the chapter will cover the use of deep generative models to create new, realistic 
motion sequences, expanding the potential applications in action recognition tasks
by proposing a novel CNN-based VAE model.

We will use two publicly available datasets for all our work in this chapter, notably the HumanAct12~\cite{action2motion-paper}
dataset for action recognition and the Kimore~\cite{kimore-paper} dataset for the medical rehabilitation assessment.

\section{Advancing Human Motion Rehabilitation Assessment with LITEMVTime}

In the domain of rehabilitation assessment, accurately evaluating a patient's performance 
during physical exercises is paramount. Traditional methods often rely on subjective 
judgments or handcrafted features, which can be both time-consuming and inconsistent. 
The advent of deep learning models, particularly for time series classification, 
has opened new avenues for enhancing the precision and efficiency of these assessments.

Deep learning techniques, such as Convolutional Neural Networks (CNNs) and Recurrent Neural 
Networks (RNNs), have shown remarkable success in classifying time series data, including 
human motion sequences captured via 3D skeleton tracking~\cite{sensors-activity-recog}. 
These methods leverage the sequential and spatial characteristics of the data, providing a 
more nuanced understanding of the movements.

In this section, we utilize LITEMVTime, the multivariate extension of the previously 
developed LITETime model (Chapter~\ref{chapitre_4}). LITEMVTime has been specifically
designed to address MTS data for classification tasks, such as the task of 
human motion rehabilitation assessment. It is lightweight in terms of 
both the number of parameters and computational requirements, making it highly 
suitable for real-time applications in medical settings.
This efficiency is crucial for deployment in clinical environments, 
where timely and accurate feedback is essential for both patients and 
medical practitioners.

We showcase in this section that LITEMVTime outperforms other architectures on this task.
The model's superior performance is attributed to its innovative architecture, which 
effectively captures the temporal dynamics and spatial configurations of the 
human skeleton during rehabilitation exercises. Unlike conventional models that 
may require extensive computational resources, LITEMVTime's streamlined design 
ensures it can operate on standard medical clinic hardware without compromising on performance.

Furthermore, the model's efficiency ensures that it can be integrated into existing 
clinical workflows without the need for extensive computational infrastructure. 
This integration can enhance the overall quality of care by enabling more frequent 
and detailed assessments, ultimately contributing to better patient 
outcomes~\cite{gait-analysis-wearable,activity-recog-wearable}.

A crucial aspect of deploying machine learning models in medical applications is 
the explainability of their decisions. Medical doctors often pose the question, 
``\emph{Why should I trust you?}''~\cite{lime-paper} when presented with automated assessment results. To 
address this concern, we have integrated Class Activation Maps (CAMs)~\cite{original-cam-paper} into our 
framework. CAMs help in visualizing the regions of the input data that are most 
influential in the model's decision-making process, thereby providing insights 
into which features are most impactful for a given prediction. This is particularly 
important for our CNN-based LITEMVTime model.

\subsection{Experimental Setup \& Dataset Preprocessing}\label{sec:kimore-explain}

\begin{figure}
    \centering
    \caption{Visualization of one sample from the Kimore skeleton based human rehabilitation
    dataset~\cite{kimore-paper}, per exercise.}
    \label{fig:example-kimore}
    \includegraphics[width=\textwidth]{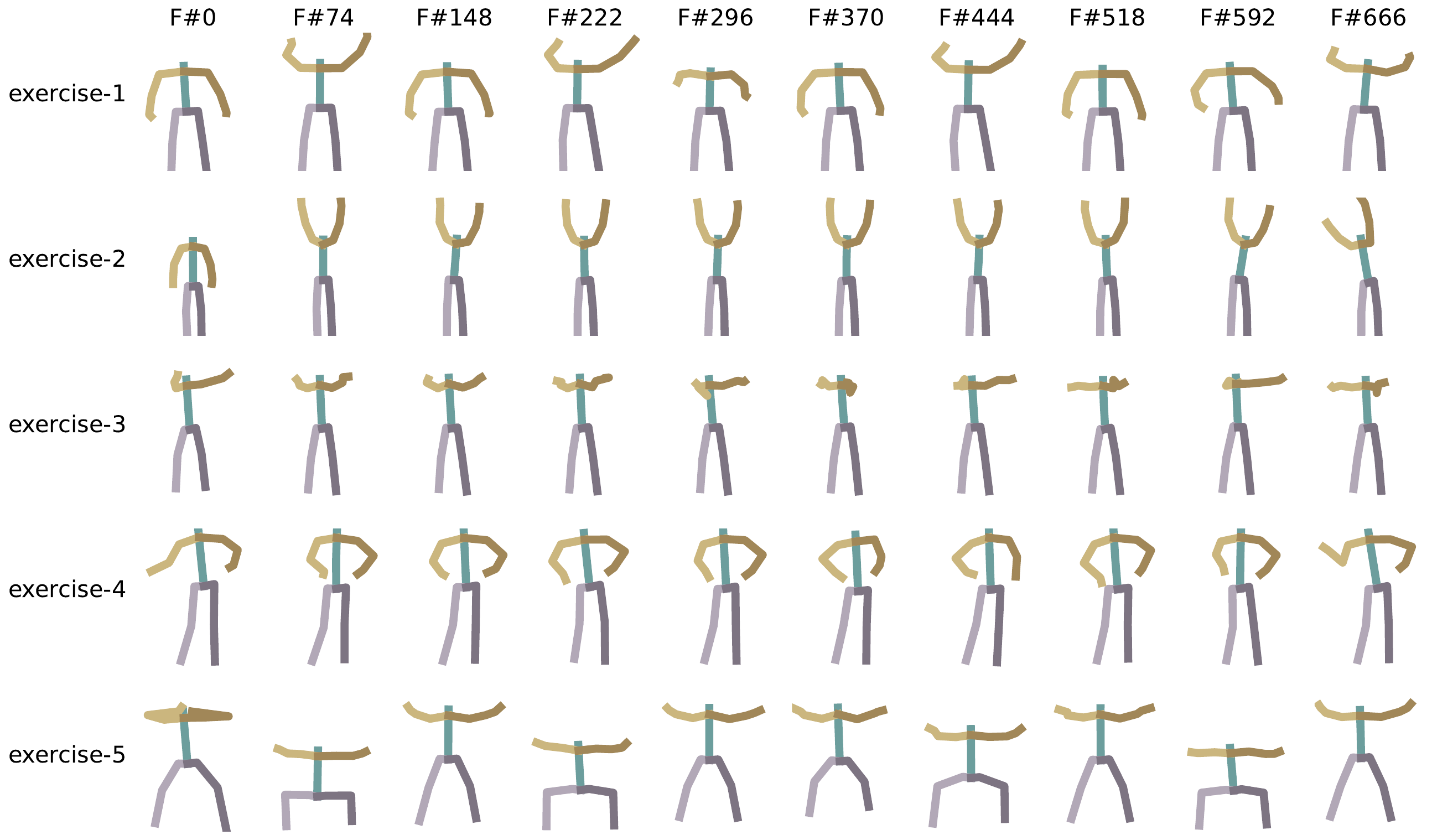}
\end{figure}

For this experiment, we utilized the Kimore dataset~\cite{kimore-paper},
which includes video sequences of 
patients performing rehabilitation exercises, captured and converted into numerical 
3D sequences using Kinect v2 sensors~\cite{kinect-paper}. The dataset comprises recordings 
from both healthy and unhealthy subjects executing five distinct rehabilitation exercises. 
These exercises are the following: (1) lifting of the arms, 
(2) lateral tilt of the trunk with the arms in extension,
(3) trunk rotation, (4) Pelvis rotations on the transverse plane
and (5) Squatting.
An example sequence of each exercise is represented in Figure~\ref{fig:example-kimore}.
The skeletons contain $18$ joints each in a three dimensional space $x$, $y$ and $z$.

Each sequence of patient's movement in every exercise is evaluated by a human expert,
who assigns a quality score ranging from $0$
(poor performance) to $100$ (excellent performance).
The dataset comprises $71$ subjects, with $40$ being healthy and $31$ unhealthy. 
Each subject performs at least five repetitions of each exercise, and all repetitions 
are recorded as individual samples, resulting in $71$ samples per exercise.

\begin{figure}
    \centering
    \caption{
        The distribution of the scores given by experts to
        \protect\mycolorbox{0,30,255,0.6}{healthy}
        and \protect\mycolorbox{255,30,0,0.6}{unhealthy}
        patients when performing each of the five different exercises.
        The \protect\mycolorbox{0,128,0,0.6}{threshold} set to discretize 
        these scores is chosen to be the middle point
        posed at $50$.
    }
    \label{fig:kimore-scores}
    \includegraphics[width=\textwidth]{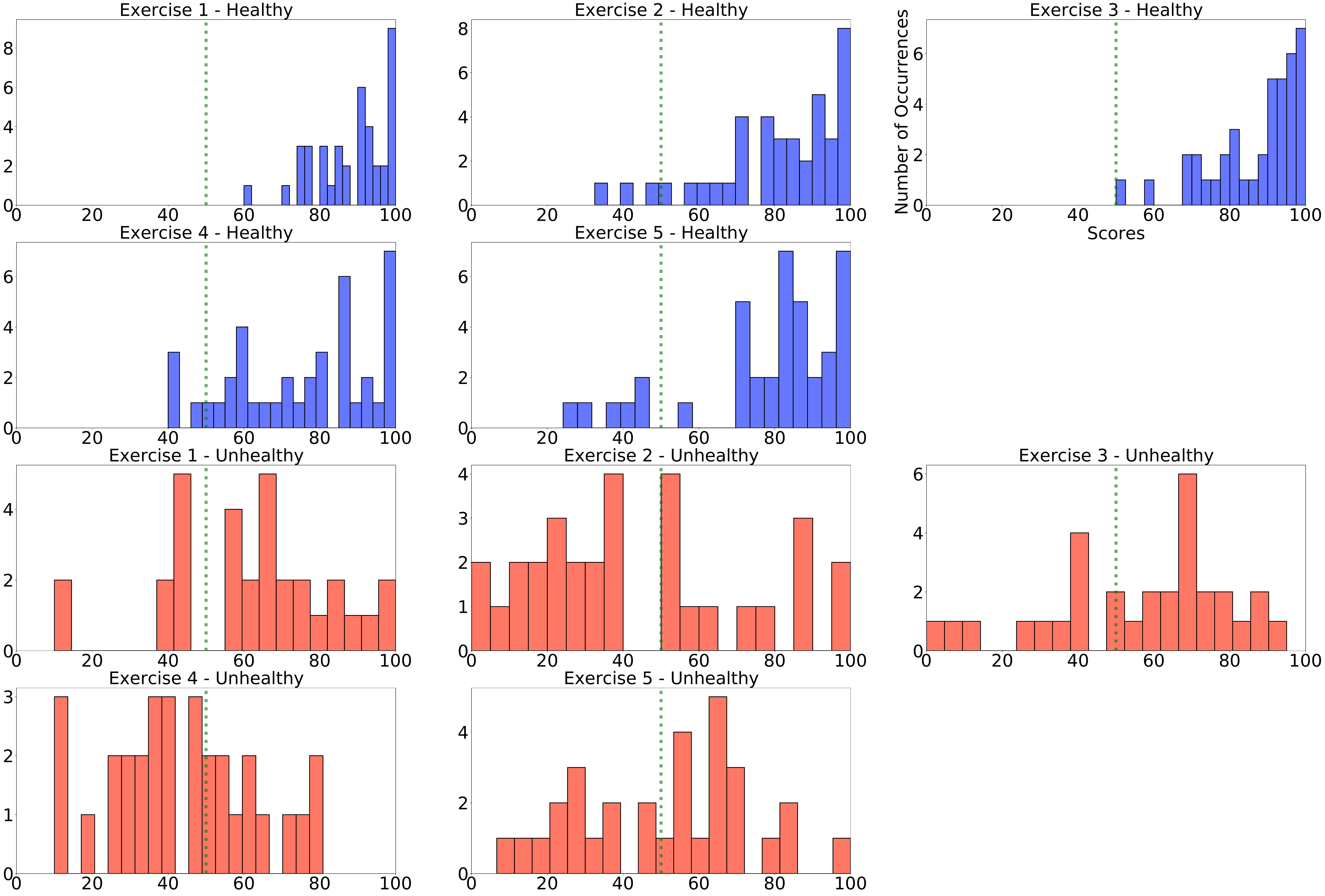}
\end{figure}

Figure~\ref{fig:kimore-scores} illustrates the distribution of
performance scores for each exercise, differentiated 
between healthy and unhealthy subjects. Typically, unhealthy subjects tend to receive 
lower scores, whereas healthy subjects achieve higher scores.
However, this is not always the case, as seen in Figure~\ref{fig:kimore-scores}.
This discrepancy arises because even if some subjects are considered as unheathy
in the dataset, their injury could not limit to perform some exercise
Despite the regression 
nature of the dataset, we reframed the task to evaluate
the performance of subjects irrespective of their health status. The evaluation criteria 
are defined as follows:

\begin{itemize}
    \item Scores below $50$ indicate a poorly performed exercise.
    \item Scores above $50$ indicate a well-performed exercise.
\end{itemize}

The dataset features sequences of varying lengths, which necessitated resampling all 
sequences to a common length, determined to be 748 frames (the average length) using
the Fourier resampling method in \textit{scipy} Python package~\cite{scipy-paper}.
This is due to the fact that skeleton-based motion sequences extracted using 
kinect cameras are sampled using a uniform sample rate, which is a condition 
to be able to use this resampling approach.
We split the dataset 
into an $80\%-20\%$ train-test set, ensuring stratification to maintain a balanced 
representation of good and bad performances in both sets.

Each 3D human motion sequence is transformed to an MTS and each exercise is utilized 
as an independent dataset, given that one score value does not represent the same 
thing from one exercise to another.
The dataset were z-normalized prior to training and testing independently on each channel.
The best-performing model during training, determined by monitoring the training
loss, was selected for testing. The Adam optimizer with a Reduce on Plateau learning
rate decay method was employed, using TensorFlow's~\cite{tensorflow-paper} default
parameter settings.
The same parameters are used as presented in Chapter~\ref{chapitre_4} for LITEMV,
as well as its ensemble version LITEMVTime.

\subsection{Competitor Models}

We evaluated the performance of various deep learning models on this dataset, 
including Fully Convolutional Networks (FCN), ResNet, and InceptionTime, 
along with our proposed model, LITEMVTime. Additionally, we included a baseline classifier, 
1-Nearest Neighbor Dynamic Time Warping (1-NN-DTW) following~\cite{bakeoff-tsc-2}.
For all the competitors we utilize the same parameter setup used in~\cite{dl4tsc}.

\subsection{Experimental Results}

\begin{table}
    \caption{
    Accuracy of the baseline, 1-NN-DTW, three state-of-the-art deep learning models, 
    FCN ResNet and InceptionTime compared to \textbf{our} LITEMVTime on the Kimore 
    human rehabilitation exercise.
    We present for each of the five exercises the accuracy of the models on the test unseen split.
    The performance of the winning model for each exercise is shown in \textbf{bold}
    and the second best is shown in \underline{underline}.}
    \label{tab:kimore-cls-results}
    \resizebox{\columnwidth}{!}{
    \begin{tabular}{c|ccccc}
    Kimore Exercise & \textbf{1-NN-DTW} & \textbf{FCN} & \textbf{ResNet} & \textbf{InceptionTime} & \textbf{LITEMVTime} \\ \hline
    \textbf{Exercise 1}       & 60.00          & 84.00          & {\ul 85.33} & 78.67          & \textbf{86.67} \\
    \textbf{Exercise 2}       & 46.67          & 72.00          & 69.33       & {\ul 78.67}    & \textbf{80.00} \\
    \textbf{Exercise 3}       & 86.67          & \textbf{92.00} & 86.67       & {\ul 88.00}    & 86.67          \\
    \textbf{Exercise 4}       & \textbf{66.67} & {\ul 65.33}    & 60.00       & 57.33          & \textbf{66.67} \\
    \textbf{Exercise 5}       & 73.33          & 66.67          & {\ul 81.33} & \textbf{84.00} & 80.00          \\ \hline
    \textbf{Average Accuracy} & 66.67          & 76             & 76.53       & {\ul 77.33}    & \textbf{80.00} \\ \hline
    \textbf{Average Rank}     & 3.6            & 2.8            & 2.8         & {\ul 2.6}      & \textbf{1.8} 
    \end{tabular}
    }
\end{table}

The results, summarized in Table~\ref{tab:kimore-cls-results},
indicate that LITEMVTime outperforms the other models 
in both average performance and average rank across all exercises. This demonstrates 
that LITEMVTime, despite its small size, is highly effective in classifying the 
quality of exercise performance based on recorded sequences. Consequently, 
LITEMVTime proves to be a valuable tool for assessing patient rehabilitation 
exercises, providing reliable classifications that can support clinical decision-making.

\subsection{Enhancing Trust and Transparency in Human Rehabilitation Assessment With LITEMV}

In the field of rehabilitation assessment, achieving high performance with deep learning 
models is essential. However, the ability to understand the decision-making process of 
these models is equally important, especially in medical applications where the stakes 
are high. Over the past decade, there has been a significant focus on model interpretability, 
particularly in Time Series Classification (TSC) over the last five
years~\cite{explainability-tsc}.

Class Activation Maps (CAM) are a powerful technique for interpreting the decisions of 
deep Convolutional Neural Networks (CNNs), which are often perceived as black-box models. 
Initially introduced by~\cite{original-cam-paper} for image data, CAMs have since been adapted 
for time series data, providing a way to visualize which parts of the input data 
contribute most to the model's decisions.
This technique got first adapted to time series classification in~\cite{fcn-resnet-mlp-paper}.

CAMs require a global representative layer before the softmax classification layer, 
such as Global Average Pooling (GAP). This setup is used in various architectures 
including FCN, ResNet, Inception, LITE, and LITEMVTime. In the context of TSC, the 
output of a CAM is a univariate time series where each timestamp indicates the importance 
of that specific point in the input series for the model's decision.

Mathematically, CAM is defined as follows:

\begin{itemize}
    \item Let $\textbf{O}(t)=\{\textbf{o}^1(t),\textbf{o}^2(t),\ldots,\textbf{o}^M(t)\}$
    represent the output of the last convolutional layer, 
    an $MTS$ with $M$ variables (the number of filters). 
    Thus, $\textbf{o}^m(t)$ is the output univariate time series of filter $m~\in~[1,M]$.
    \item Let $\textbf{w}^{c}=\{w_{1}^{c},w_{2}^{c},\ldots,w_{M}^{c}\}$
    be the weight vector connecting the GAP output to the neuron of the
    winning class $c$ (the class with the highest probability value).
\end{itemize}

The CAM output is then:
\begin{equation}\label{equ:CAM}
    CAM(t) = \sum_{m=1}^M~w_m^c.\textbf{o}^m(t)
\end{equation}

This output is normalized using min-max normalization. For two given timestamps, 
the one with the highest CAM score has contributed more significantly to the 
decision of the black-box model.

\begin{figure}
    \centering
    \caption{
        Explainability of the LITEMV model using the Class Activation Map (CAM) on the 
        feature of the last DWSC layer.
        The colorbar values represent the normalized (between $0$ and $1$) 
        scores of the CAM.
        Five samples each from one of the five exercises are presented with 
        the CAM scores on different time stamps.
        A higher CAM score indicates the importance of a time stamp for 
        the decision making of LITEMV.
    }
    \label{fig:kimore-cam-examples}
    \includegraphics[width=\textwidth]{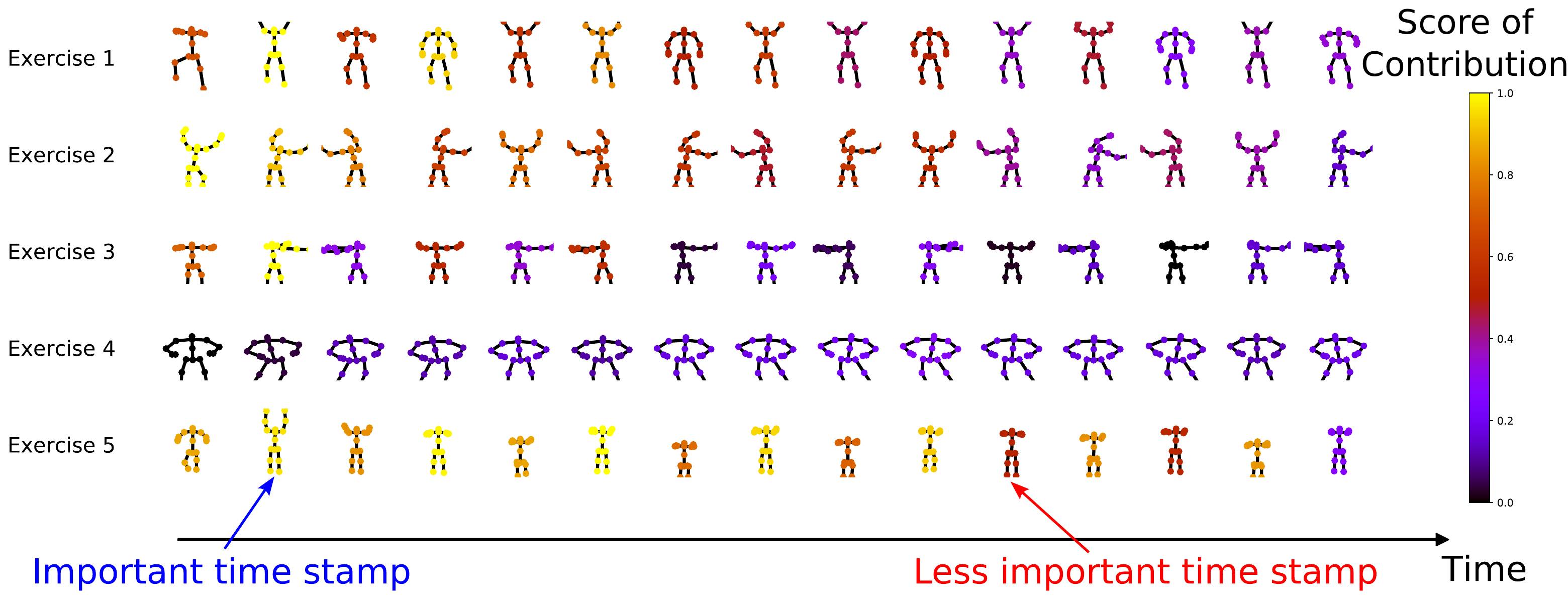}
\end{figure}

In our study, we apply CAM to the LITEMVTime model to interpret its decisions 
on the Kimore dataset, which includes human rehabilitation exercises. 
We analyze five different examples from each exercise and generate CAM 
outputs using a LITEMVTime model trained for each exercise 
classification task presented in Figure~\ref{fig:kimore-cam-examples}.

\begin{figure}
    \centering
    \caption{
        Explainability of the LITEMV model using the Class Activation Map (CAM) 
        on the feature of the last DWSC layer.
        Two samples from the test split of the same exercise are presented, the first 
        (top) having a ground truth of class 1, and the second (bottom) having a ground 
        truth of class 0.
        LITEMV correctly classifies the first sample but incorrectly the second.
        It can be seen that the important time stamp in the case of the correctly classified 
        sample has higher color intensity, so higher CAM score, compared to the same time stamp 
        from the incorrectly classified sample.
    }
    \label{fig:kimore-cam-diff}
    \includegraphics[width=\textwidth]{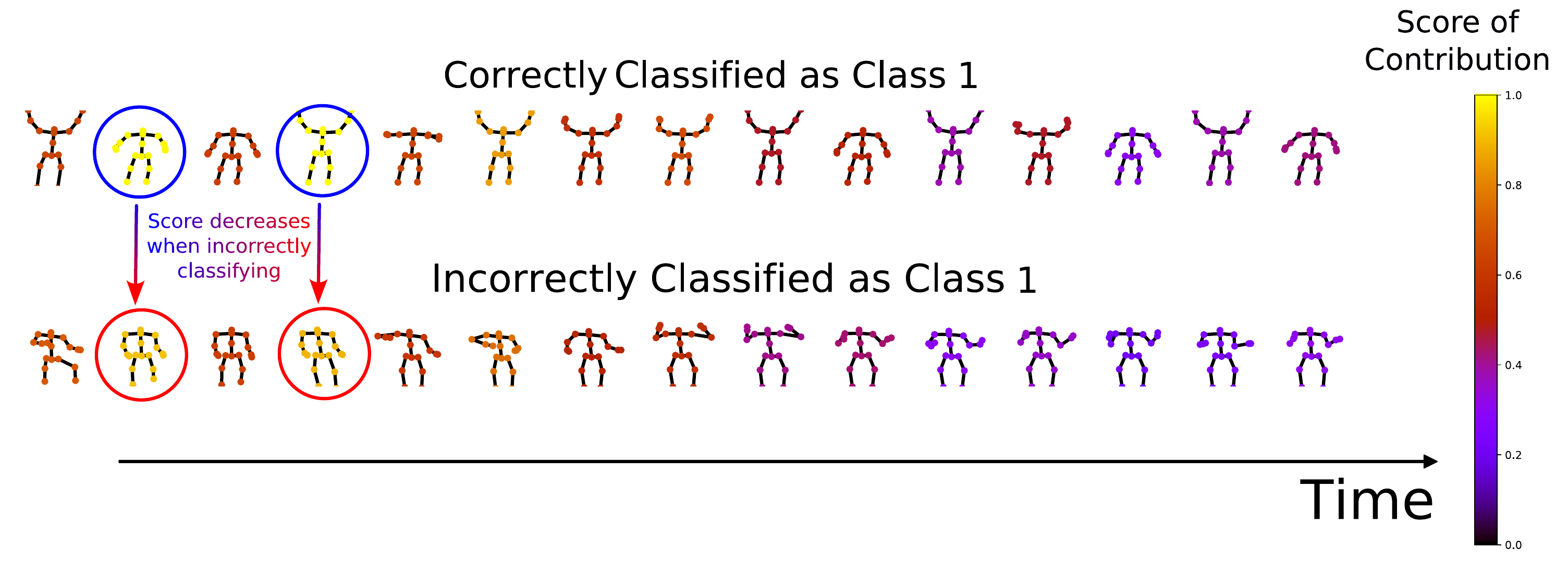}
\end{figure}

Given that human skeleton data forms a multivariate time series, the CAM values 
represent the temporal axis, with each timestamp's CAM score indicating the 
contribution of that particular pose to the classification. To further explore 
the variability in CAM scores, we compare two CAM explanations for two samples 
of the same exercise in Figure~\ref{fig:kimore-cam-diff}:
one correctly classified as class 1 ($score > 50$) and 
another incorrectly classified as class 1 when it should be class 0 ($score < 50$). 
The higher intensity of CAM colors for the correctly classified sample indicates 
higher contribution of important timestamps, while the misclassified sample shows 
lower scores, reflecting the influence of the incorrect class weights.

By utilizing CAM, we can provide clear explanations for the LITEMVTime model's 
decisions, demonstrating the specific data points that influenced its classifications. 
This transparency is crucial for integrating deep learning models into clinical 
workflows, ensuring that healthcare professionals can rely on these tools with confidence.

\section{Extending Human Motion Rehabilitation Data With Time Series Prototyping}

Human motion rehabilitation data is inherently sensitive and challenging to acquire, particularly 
due to privacy concerns and the complex nature of medical data. This scarcity often results in 
limited datasets that can lead to overfitting in machine learning models used for patient assessment, 
whether for classification or regression tasks. To mitigate this issue, we propose generating synthetic 
data using time series prototyping techniques~\cite{dba-paper}.
By creating synthetic data, we can extend existing datasets, making models 
more resilient to overfitting and improving their generalization.

One effective approach for generating synthetic data is through time series prototyping. 
We introduce an innovative approach called ShapeDBA (Shape Dynamic Time Warping Barycenter Average)
for time series prototyping (Chapter~\ref{chapitre_1} Section~\ref{sec:ts-prototyping}).
Following its application in standard prototyping, we leverage ShapeDBA 
for advanced weighted prototyping. This method builds upon the sophisticated principles of weighted 
elastic averages as delineated by \cite{weighted-dba-paper}. This method involves creating prototypes 
that capture the essential characteristics of a set of time series data, which can then be 
used to generate new, synthetic sequences.

In the following sections, we will present the ShapeDBA method in detail and provide extensive 
experimental results to demonstrate that it is now the state-of-the-art prototyping method
for time series data. Subsequently, we will introduce the weighted ShapeDBA setup tailored for 
regression tasks, specifically focusing on human rehabilitation assessment using the Kimore dataset~\cite{kimore-paper}.
We will extend the Kimore dataset with synthetic data generated by the ShapeDBA method and demonstrate 
that this augmented dataset significantly enhances the performance of deep supervised regression models 
compared to using the original dataset alone.

\subsection{Generating Effective Time Series Prototypes With ShapeDBA}

Prototyping time series data is a critical task in various domains, including medical diagnostics, 
human motion analysis, and satellite imagery interpretation. Traditional methods for generating time 
series prototypes often fall short in preserving the inherent patterns and nuances of the data, leading 
to out-of-distribution artifacts. This discrepancy is primarily due to the reliance on conventional 
DTW measure~\cite{dba-paper}, which emphasize absolute similarities over neighborhood similarities. 
These artifacts can significantly impact the accuracy and reliability of subsequent analyses, such as 
clustering~\cite{elastic-clustering-review}, classification~\cite{neares-centroid-dba-paper}
or explainability~\cite{prototyping-explainability}.
Therefore, there is a pressing need for a more robust and representative 
method to generate prototypes that faithfully capture the underlying data distribution, ensuring more 
accurate and meaningful insights from time series analysis.

\subsubsection{ShapeDBA Methodology}

Our proposed prototyping method, ShapeDBA (Shape Dynamic Time Warping Barycenter
Averaging), is an advanced method 
designed to generate more accurate and representative time series prototypes. 
The key innovation in ShapeDBA is the integration of the ShapeDTW~\cite{shape-dtw-distance}
(Chapter~\ref{chapitre_1} Section~\ref{sec:tsc-distance})
measure, which considers the structural similarities within the neighborhoods of 
time series data points, thus overcoming the limitations of traditional DTW methods.

The ShapeDBA algorithm follows these steps:

\begin{enumerate}
    \item \textbf{Initialization}: Start with an initial average time series, which can be 
    randomly selected from the dataset.
    \item \textbf{Alignment}: For each time stamp in the average time series, find the aligned 
    points (using ShapeDTW measure) in all the time series samples.
    This involves creating a set of associated 
    time stamps $assoc_t$ for each time stamp $t$ in the average series.
    \item \textbf{Averaging}: Calculate the barycenter for each time stamp $t$ by averaging 
    all the aligned points in $assoc_t$. The barycenter is computed as:
    \begin{equation}\label{equ:shape-dba-barycenter}
        ShapeDBA_barycenter(assoc_t) = \dfrac{1}{|assoc_t|}\sum_{i=1}^{|assoc_t|}assoc_t^i
    \end{equation}
    \item \textbf{Iteration}: Repeat steps 2 and 3 until convergence, i.e.
    until the changes in the average time series are minimal.
\end{enumerate}

\subsubsection{Reach Value Control}

The ``reach'' hyperparameter in ShapeDTW defines the neighborhood size around each 
time stamp for alignment purposes. By adjusting this parameter, ShapeDTW can 
emulate different similarity measures. When the reach is set to 1, ShapeDTW 
operates like traditional DTW, focusing solely on individual time stamps. 
However, when the reach is set to a very large value, it behaves similarly 
to the Euclidean distance, as the neighborhood extends across the entire 
time series. This flexibility allows ShapeDTW, and consequently
ShapeDBA, to balance between local and 
global alignment, adapting to the specific requirements of the data.

\subsubsection{Qualitative Evaluation}

\begin{figure}
\centering
\caption{A qualitative evaluation of the proposed average technique compared to
other approaches on a GunPoint dataset. The ShapeDBA algorithm is the only
approach to not generate out-of-distribution artifacts.}
\label{fig:shape-dba-averaging-compare}
\includegraphics[width=\textwidth]{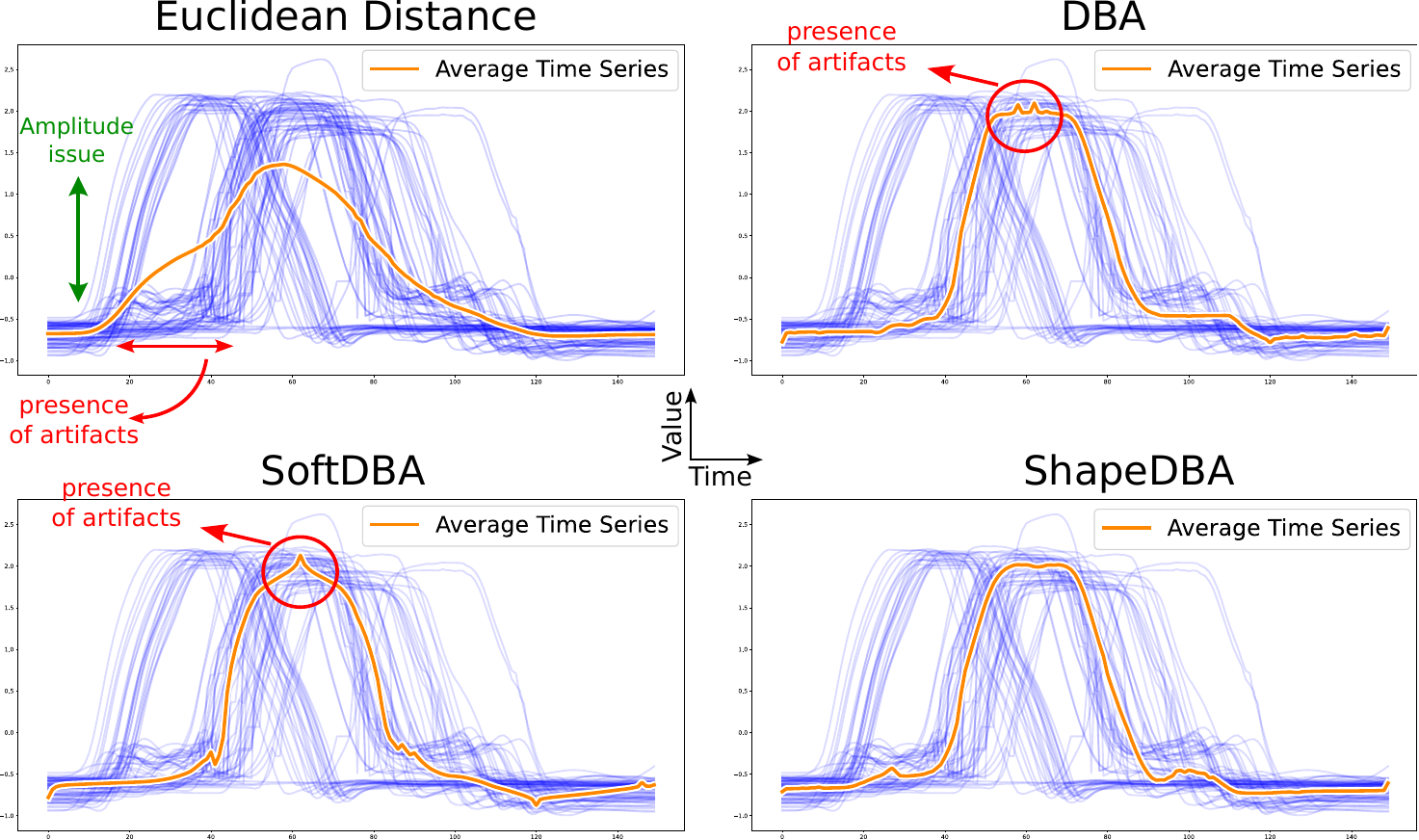}
\end{figure}

In the literature, the artimetic mean and two notable TSP approaches with elastic measures,
DBA~\cite{dba-paper} and SoftDBA~\cite{soft-dtw-distance}.
Figure~\ref{fig:shape-dba-averaging-compare}
compares prototype calculations using these methods and our ShapeDBA 
on the GunPoint dataset of the UCR archive~\cite{ucr-archive}.
The figure reveals that the arithmetic mean is 
unsuitable for temporal data, particularly when samples are shifted. 
It introduces spatial and temporal artifacts; the temporal placement of the prototype 
skews toward the most frequent occurrence, and the amplitude values become out-of-distribution 
due to averaging misaligned values. DBA and SoftDBA improve temporal alignment but still 
produce peak artifacts because of the rigid point-to-point alignment of DTW and SoftDTW, 
which do not account for amplitude differences. ShapeDBA, however, offers the best 
of both worlds. It leverages DTW's temporal alignment and ShapeDTW's neighborhood 
alignment~\cite{shape-dtw-distance}, avoiding point-to-point issues.
Consequently, the prototype generated by 
ShapeDBA in Figure~\ref{fig:shape-dba-averaging-compare} is free from peak 
out-of-distribution artifacts.

\subsubsection{Quantitative Evaluation}

To quantitatively evaluate the effectiveness of ShapeDBA in time series prototyping, 
we coupled ShapeDBA and ShapeDTW as the averaging method and distance measure in 
the $k$-means algorithm for clustering. Clustering, particularly using the $k$-means 
algorithm, serves as a robust evaluation metric for TSP methods based on 
elastic similarity measures~\cite{elastic-clustering-review}.

We compared ShapeDBA against four distance-based methods: (1) $k$-means with 
the default setup (arithmetic mean and Euclidean Distance), referred to as 
MED; (2) $k$-means with SoftDBA and SoftDTW; (3) $k$-means with DBA and DTW; and (4) $k$-shape.
Given that all other methods iteratively find prototypes, we applied the same iterative 
approach to the MED method. Instead of using a simple arithmetic mean of all samples, 
we iteratively calculated the mean over aligned points for each time stamp in the 
prototype, similar to DBA. However, for MED, we assumed ideal alignment without any warping.

\paragraph{Experimental Setup}

We conducted our experiments on 123 datasets from the UCR archive~\cite{ucr-archive}. Out of 
the 128 available datasets since 2018, five were excluded due to their 
high time series length, which would have been computationally prohibitive 
given the quadratic time complexity of the considered algorithms. All samples in each dataset 
were z-normalized to ensure a zero mean and unit standard deviation. The clustering 
algorithms were trained on the combined train-test splits of these 123 datasets. 
While some UCR datasets are merely different train-test splits of the same original dataset, 
this occurs infrequently, so the same data might be clustered multiple times.
We set the value of the ``reach'' hyperparameter to $15$ resulting in a sliding
window size of $31$, following the original work of ShapeDTW~\cite{shape-dtw-distance}.

In machine learning, non-deterministic estimators often suffer from performance biases 
related to their initial setup, such as the initialization of weights in deep learning 
models. This bias is particularly relevant in clustering tasks, where the starting 
positions of clusters can significantly influence the results. To address this in 
our experiments, we ran each clustering algorithm five times, each with different 
initial cluster configurations, and averaged the results. However, using different 
initial clusters for each method could introduce another layer of bias. To ensure a 
fair comparison, we used the same set of five initial clusters across all clustering 
algorithms for each dataset. This approach eliminated variability due to initial cluster 
selection and allowed us to present unbiased average performance metrics, accurately 
reflecting the effectiveness of each clustering method.

\paragraph{Evaluation Metric: Adjusted Rand Index ($ARI$)}

The Adjusted Rand Index ($ARI$)~\cite{ari-paper} is an enhanced version of the Rand Index ($RI$), 
addressing the limitations of the original metric. The $RI$ 
measures the similarity between true labels $\textbf{y}$ and predicted labels $\hat{\textbf{y}}$
from a clustering algorithm using the formula:

\begin{equation}\label{equ:ri}
    RI(\textbf{y},\hat{\textbf{y}}) = \dfrac{TP+TN}{TP+FP+FN+TN}
\end{equation}

Here, $TP$ (True Positive) and $TN$ (True Negative) denote correctly clustered pairs, 
while $FP$ (False Positive) and $FN$ (False Negative) denote incorrectly clustered pairs. 
The $RI$ calculates the proportion of pairwise agreements between the true and predicted clusters.
However, the $RI$ can be misleading because it may indicate high similarity for 
clusters that are randomly generated, particularly when the number of clusters is large. 
This occurs because the expected $RI$ value varies between random clusters.

To overcome this, the $ARI$ adjusts the $RI$ to account for chance, normalizing the score so 
that random clustering yields an $ARI$ of $0.0$. The $ARI$ is defined as:

\begin{equation}\label{equ:ari}
    ARI(\textbf{y},\hat{\textbf{y}}) = \dfrac{RI(\textbf{y},\hat{\textbf{y}}) - E[RI]}{1.0 - E[RI]}
\end{equation}

where $E[RI]$ is the expected value of the $RI$ for random clustering. The $ARI$ ranges from
$-0.5$ (indicating no similarity) to $1.0$ (indicating perfect agreement),
providing a more reliable measure of clustering performance by correcting for random chance.

\paragraph{Implementation Efficiency}

\begin{figure}
    \centering
    \caption{Calculation of the ShapeDTW measure between two
    \protect\mycolorbox{255,128,255,0.6}{time} \protect\mycolorbox{0,30,255,0.6}{series}.
    The \protect\mycolorbox{135,86,51,0.49}{overlapping area} between the 
    \protect\mycolorbox{194,224,194,0.84}{two} 
    \protect\mycolorbox{255,165,0,0.2}{sliding windows} is recomputed.}
    \label{fig:shapedtw-recompute}
    \includegraphics[width=\textwidth]{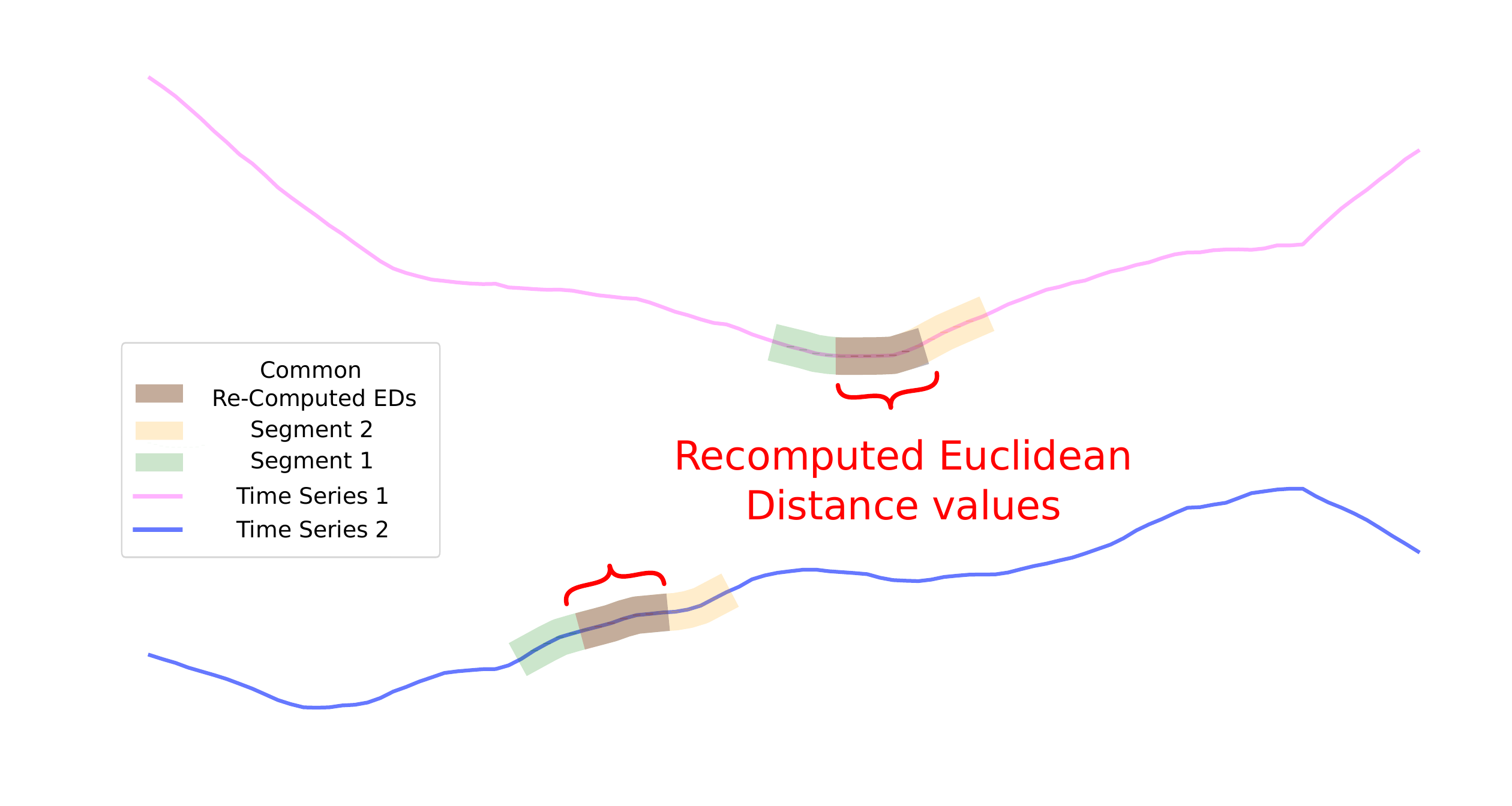}
\end{figure}

The ShapeDTW algorithm modifies the original DTW similarity measure by transforming 
the input time series into a multivariate format. In the univariate case with the 
``identity'' descriptor, each time stamp's neighborhood is converted into a Euclidean 
vector, creating a multivariate time series. Applying DTW to this transformed series 
involves computing the Euclidean distance between the channel vectors of paired time 
stamps, which can lead to computational inefficiency when sliding the reach window, 
as depicted in Figure~\ref{fig:shapedtw-recompute}.
This inefficiency is specific to the identity transformation.

\begin{figure}
    \centering
    \caption{A more efficient implementation of the ShapeDTW measure with the 
    identity descriptor involves sliding a window over the time stamp pairwise Euclidean
    matrix between the two time series, instead of applying DTW directly 
    on their multivariate transformation. The data from each window position 
    is collected into a zero-initialized matrix, which is then processed using 
    the DTW algorithm, significantly reducing computational overhead.}
    \label{fig:shape-dtw-efficient}
    \includegraphics[width=\textwidth]{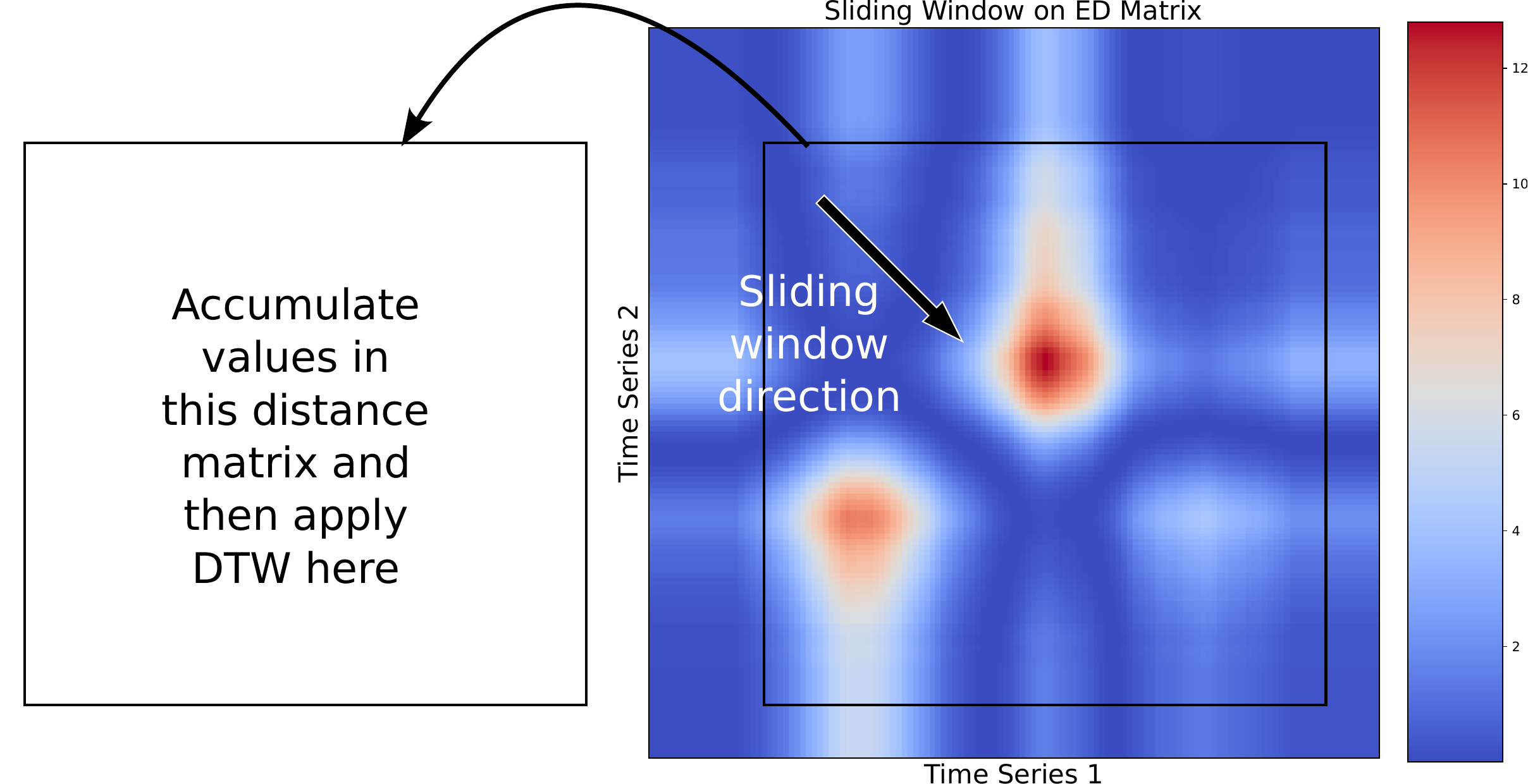}
\end{figure}

To optimize this process, we first calculate the Euclidean pairwise distances 
between the two time series, resulting in a distance matrix. This matrix is 
then padded with edge values equal to half the reach. A window, with dimensions 
matching the length of the time series, slides diagonally across the distance matrix. 
The results are accumulated into a zero-initialized matrix. The DTW algorithm is 
subsequently applied to this new matrix, thereby avoiding unnecessary computations 
and improving efficiency. Figure~\ref{fig:shape-dtw-efficient}
illustrates this streamlined implementation of ShapeDTW.

\paragraph{Experimental Results}

\begin{figure}
    \centering
    \caption{An MCM comparing ShapeDBA to other averaging approaches, coupled with $k$-means and 
    their associated similarity measure, and $k$-shape, on the ARI metric.}
    \label{fig:shapedba-mcm-ari}
    \includegraphics[width=\textwidth]{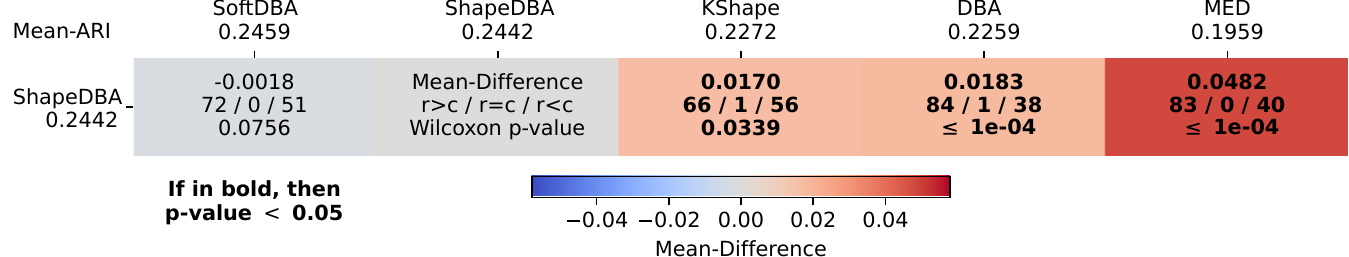}
\end{figure}

In Figure~\ref{fig:shapedba-mcm-ari} we present the MCM (Chapter~\ref{chapitre_2}) between ShapeDBA, $k$-shape
and its competitors when coupled with $k$-means following the ARI metric on $123$ datasets of 
the UCR archive~\cite{ucr-archive}.
The MCM showcases that our proposed ShapeDBA outperforms MED and the original DBA
work~\cite{dba-paper},
significantly, as well as significantly outperforming $k$-shape, the fastest TSCL algorithm
in the literature. The winning approach in terms of average performance ranking is
SoftDBA~\cite{soft-dtw-distance}.
However, comparing ShapeDBA and SoftDBA showcases that no conclusion can be made between 
the performance of both algorithms given the high p-value.
We show however in Section~\ref{sec:shapedba-runtime} that ShapeDBA is way faster than
SoftDBA.

\begin{figure}
    \centering
    \subfloat[\centering \label{fig:shapedba-1v1-med} 1V1 with Mean Euclidean Distance]{\includegraphics[width=0.4\textwidth]{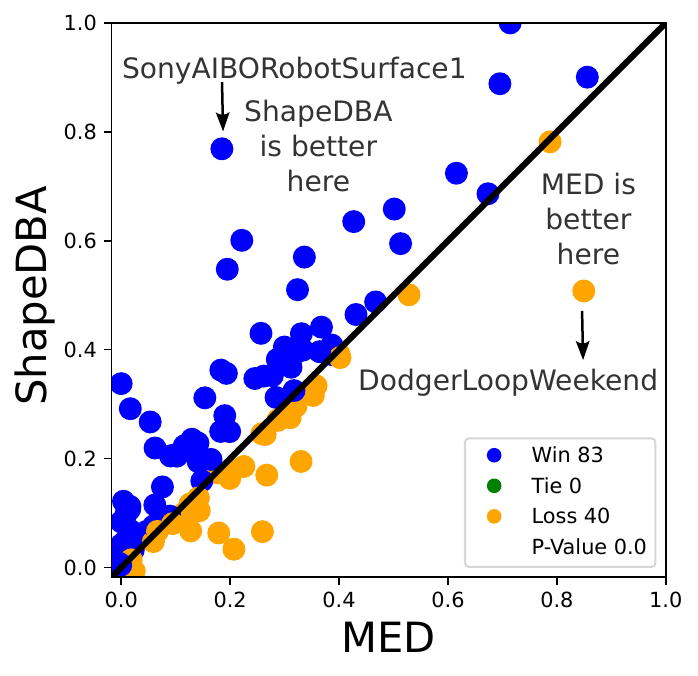}}
    \subfloat[\centering \label{fig:shapedba-1v1-dbadtw} 1V1 with DBA using DTW as a metric]{\includegraphics[width=0.4\textwidth]{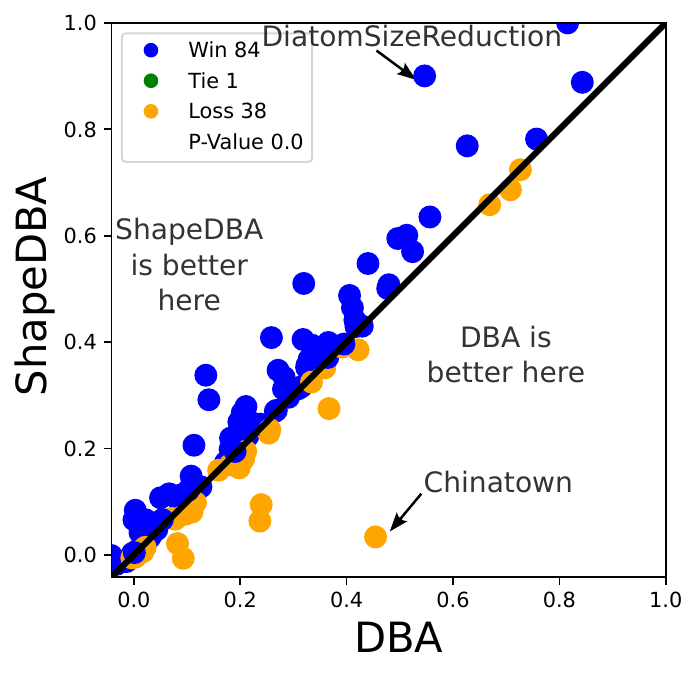}}\\
    
    \subfloat[\centering \label{fig:shapedba-1v1-kshape} 1V1 with $k$-shape]{\includegraphics[width=0.4\textwidth]{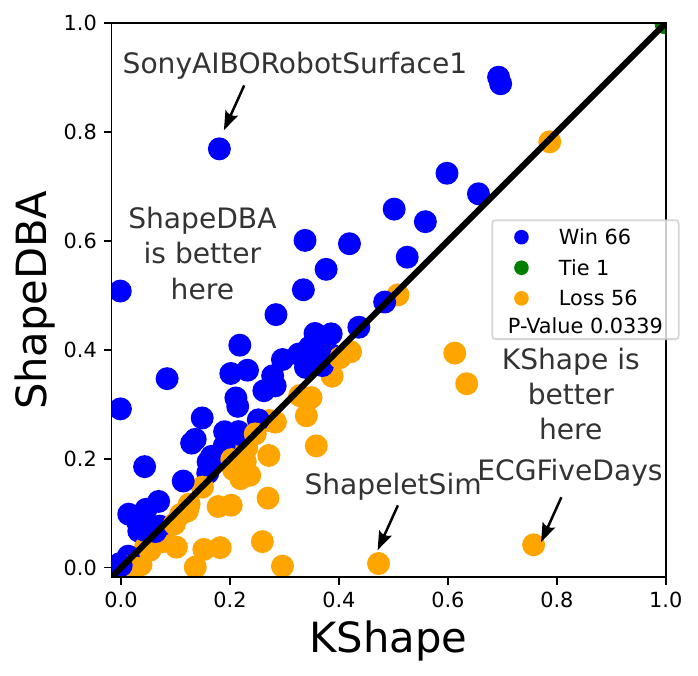}}
    \subfloat[\centering \label{fig:shapedba-1v1-softdbasoftdtw} 1V1 with SoftDBA using SoftDTW as a metric]{\includegraphics[width=0.4\textwidth]{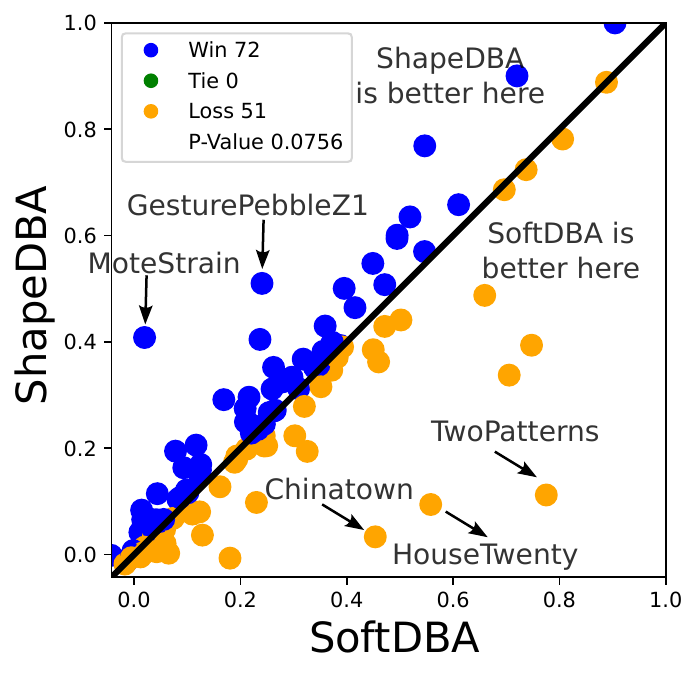}}
    \caption{1v1 Comparison between using $k$-means with ShapeDBA-ShapeDTW and other
    approaches from the literature using the Adjusted Rand Index clustering metric.}
    \label{fig:shapedba-1v1}
\end{figure}

\begin{figure}
    \centering
    \caption{The ECGFiveDays dataset from the UCR archive provides
    \protect\mycolorbox{0,128,0,0.6}{two} \protect\mycolorbox{255,30,0,0.6}{examples} 
    from each class. In this dataset, the majority of the time stamps are 
    \protect\mycolorbox{255,106,55,0.6}{noisy}, 
    with the critical information localized in the 
    \protect\mycolorbox{0,30,255,0.6}{central section of the time series}.
    ShapeDBA does not perform well on this dataset, with an ARI score of almost $0.042$.}
    \label{fig:shape-dba-ecgfivedays}
    \includegraphics[width=\textwidth]{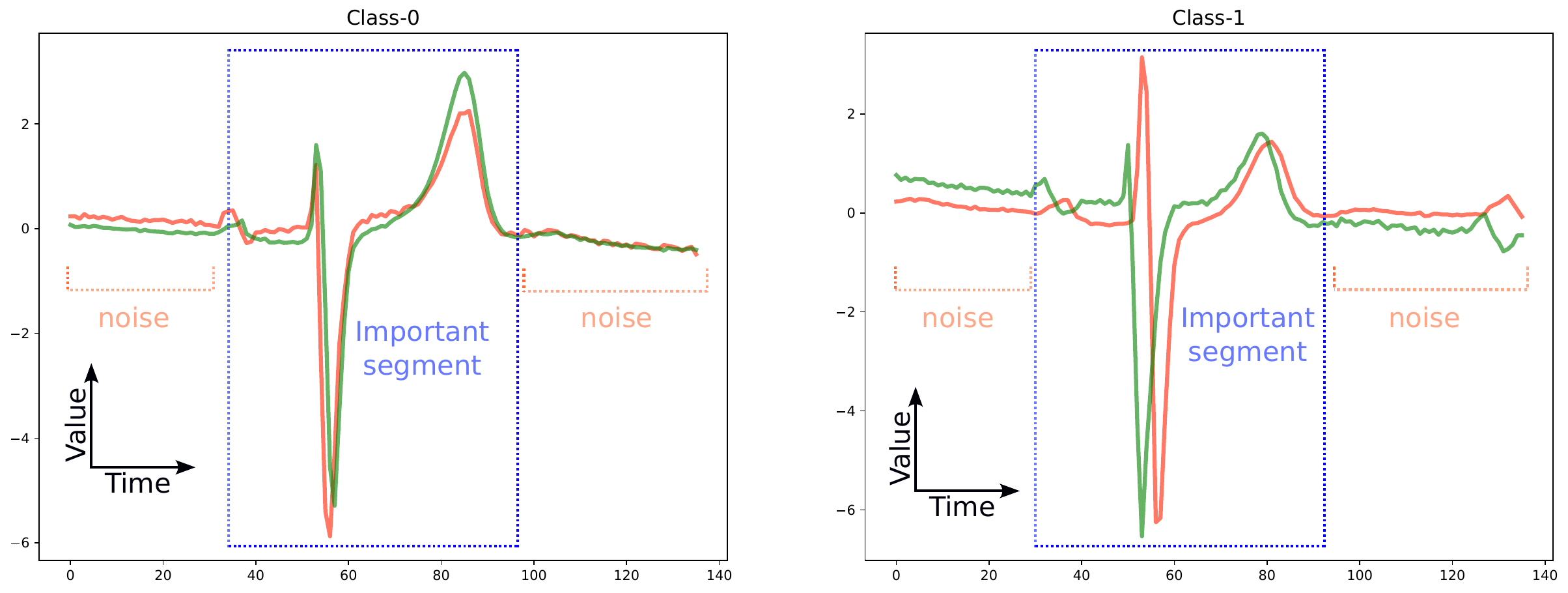}
\end{figure}

A detailed 1v1 scatter plot of the comparisons between ShapeDBA and the three other comparates
is presented in Figure~\ref{fig:shapedba-1v1}.
Certain outliers in the One-vs-One scatter plots distinctly favor either ShapeDBA 
or the other methods. For example, ShapeDBA shows lower performance (low $ARI$) 
compared to $k$-shape on the ShapeletSim and ECGFiveDays datasets. The ShapeletSim 
dataset, being a simulation of random data, does not provide meaningful conclusions. 
However, the ECGFiveDays dataset, as shown in Figure~\ref{fig:shape-dba-ecgfivedays},
uniquely illustrates a  limitation of ShapeDBA.
The ECGFiveDays dataset consists mainly of noisy time stamps, with critical 
information compressed into the middle segments of the time series, as depicted 
in Figure~\ref{fig:shape-dba-ecgfivedays}.
This noise introduces challenges during the optimization steps of 
ShapeDTW. On the other hand, ShapeDBA significantly outperforms $k$-shape on the 
SonyAIBORobotSurface1 dataset, with an ARI difference of nearly $0.6$. However, 
it is important to note that this might reflect $k$-shape's underperformance since 
MED, DBA, and SoftDBA also show better results on this dataset.
When comparing ShapeDBA to DBA, ShapeDBA exhibits a distinct advantage on the 
DiatomSizeReduction dataset, which struggles with having only four samples per 
class label. This highlights ShapeDBA's effectiveness in handling datasets with 
sparse training data.

\paragraph{Computational Runtime}\label{sec:shapedba-runtime}

\begin{figure}
    \centering
    \caption{An MCM (Chapter~\ref{chapitre_2}) showing ShapeDBA's (ours) duration
    (in seconds) compared to other approaches to finalize the clustering task on $123$
    datasets of the UCR archive~\cite{ucr-archive}.}
    \label{fig:shape-dba-mcm-time}
    \includegraphics[width=\textwidth]{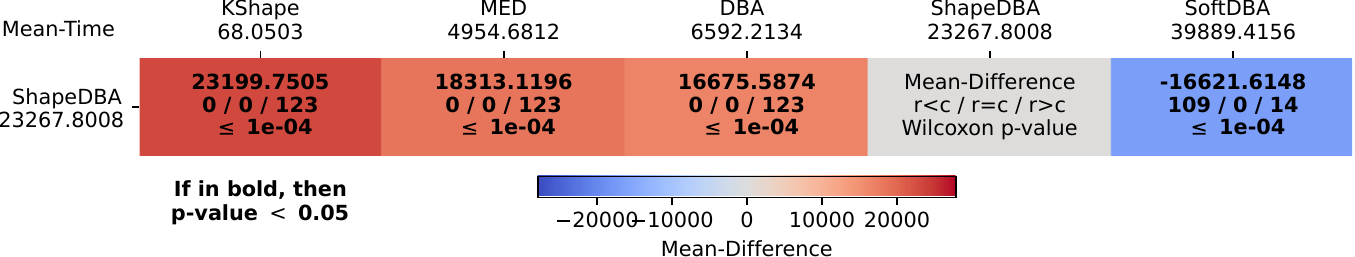}
\end{figure}

All experiments were conducted on the same machine and under identical conditions, 
ensuring a fair comparison of computational times. We recorded the total computation 
time for each clustering method, averaging the results over five initializations. This 
approach allowed us to apply the same comparison techniques as for the ARI.
In Figure~\ref{fig:shape-dba-mcm-time}, 
the MCM from Chapter~\ref{chapitre_2} illustrates the computational runtime comparison between ShapeDBA and the 
other comparates. To maintain 
consistency, we inverted the values (multiplying by -1) since lower times are preferable. 
The MCM reveals that $k$-shape is the fastest method, primarily due to its use of the Fast 
Fourier Transform (FFT) for cross-correlation, while SoftDBA is the slowest because of its 
computational gradient based optimization step.
Across 123 datasets, ShapeDBA is on average 1.7 times faster than SoftDBA, 
with 109 wins in terms of computational runtime.
Given that no definitive conclusion can be drawn regarding the performance difference 
between ShapeDBA and SoftDBA, and considering that ShapeDBA is significantly faster, 
these extensive experiments highlight ShapeDBA as the more suitable state-of-the-art 
method for the task of TSP.

In the following section, we propose using ShapeDBA in a weighted setup to generate 
new synthetic samples. In this approach, weights determine the amount of information 
drawn from each sample. This method will be applied to human motion data, specifically 
in the medical field of human rehabilitation, with the goal of extending a regression dataset.

\subsection{Weighted Average of Human Motion Sequences
for Improving Rehabilitation Assessment}

The collection and annotation of rehabilitation sequences~\cite{kimore-paper}
are complex, time-consuming, 
and require clinical expertise, which limits the size of available datasets. 
Conventional data augmentation methods for human motion data, simply adding noise,~\cite{human-motion-data-aug-noise},
although beneficial in other areas, 
tend to produce unrealistic motion sequences that do not capture the intricate 
temporal dynamics of human movement. This inadequacy necessitates innovative 
approaches to data generation that can create meaningful and representative synthetic sequences.

Another approach to generating synthetic sequences is the use of deep generative
models~\cite{actor-paper}, 
which have shown success in creating realistic sequences for human motion tasks. However, 
these models face a significant challenge: the lack of sufficient data to train a deep 
supervised model also means inadequate data for training a deep generative model. 
To address this issue, researchers in the time series domain often turn to prototyping 
techniques~\cite{weighted-dba-paper}.
These techniques create representative average sequences from existing 
training samples, offering a practical solution for augmenting datasets when data 
availability is limited.

Incorporating our previously proposed prototyping method, we propose to utilize
a weighted version of ShapeDBA
presented in the previous section, tailored to multivariate time series representing 
rehabilitation motions. By incorporating weights, this method ensures that 
the generated synthetic sequences maintain the essential characteristics of 
the original data, enhancing the realism and variability of the dataset. Our 
approach not only compensates for the limited data but also improves the 
generalization capability of models trained for rehabilitation assessment. 
This study utilizes the Kimore regression dataset~\cite{kimore-paper}
to validate our method, highlighting 
its effectiveness in producing coherent synthetic data that can significantly 
aid in the evaluation and personalization of rehabilitation treatments.

Moreover, we employ a weighted version of the ShapeDBA approach to generate 
diverse synthetic average sequences, subsequently used to enhance the training phase 
of deep learning models for downstream tasks. Due to the time-consuming nature of this 
averaging method, it cannot be computed at each epoch of the training phase, 
as is common in data augmentation. Instead, we generate several average 
sequences beforehand to expand the original training dataset, a process 
we refer to as data extension to distinguish it from traditional data augmentation.

\begin{figure}
    \centering
    \caption{
        We determine the $N$ nearest neighbors for each reference in 
        the dataset using Dynamic Time Warping (DTW). Each neighbor 
        is assigned a weight according to its DTW distance from the 
        reference. These weights are then utilized to compute a weighted 
        average sequence through ShapeDBA. Subsequently, a weighted score 
        is calculated, allowing us to expand the regression training dataset 
        with this new synthetic sample.
    }
    \label{fig:w-sdba-summary}
    \includegraphics[width=\textwidth]{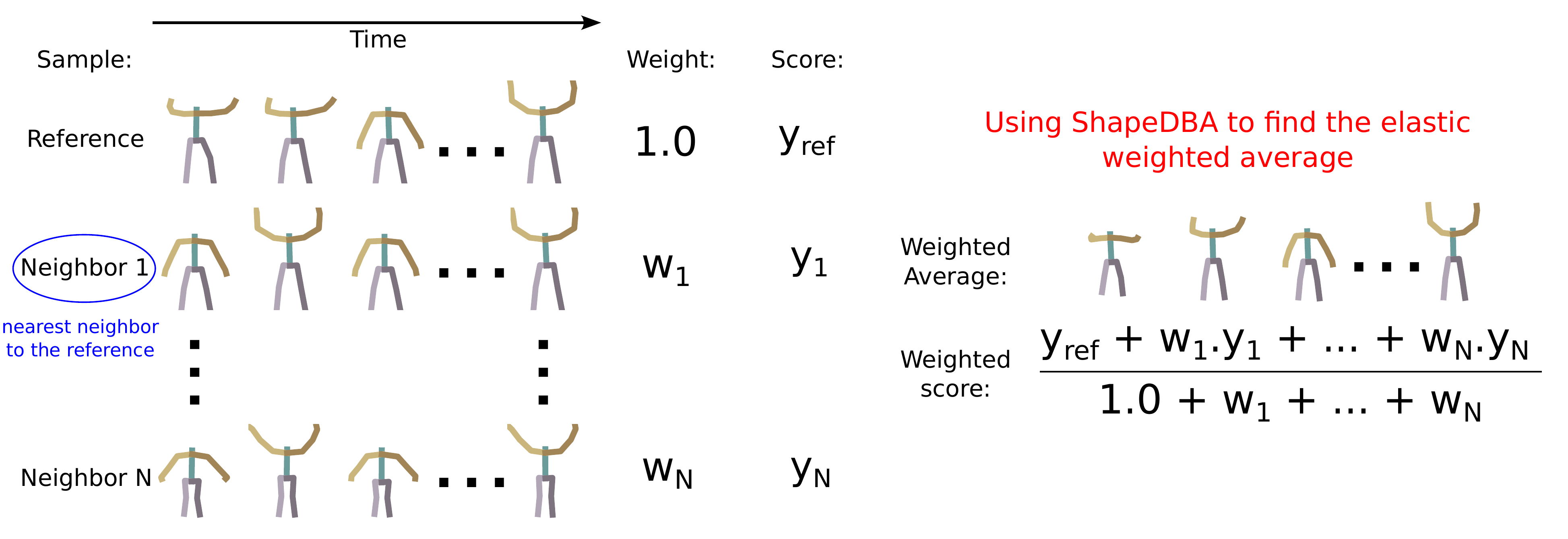}
\end{figure}

Finaly, we address the challenge of rehabilitation assessment, which is 
often approached as an extrinsic regression problem. The goal is to predict a 
continuous performance score for each rehabilitation sequence. Current state-of-the-art 
methods primarily use data augmentation or data extension for classification tasks, 
where synthetic samples are given discrete labels matching the original samples. 
However, assigning continuous labels to synthetic data presents a more complex problem. 
To overcome this, we propose a novel approach that employs a weighting strategy to 
calculate weighted continuous labels based on the true labels of a set of samples. 
This method enhances the realism and accuracy of the generated data, as detailed in
Figure~\ref{fig:w-sdba-summary}, for which we detail in the following section.

\subsubsection{Methodology}

Since rehabilitation motion sequences can be sparse, using the original ShapeDBA 
method to compute an average may lead to meaningless or incoherent results. To 
better capture the distribution of these sequences, a weighted average is more 
suitable~\cite{weighted-dba-paper}.

For each reference motion sequence, we generate a synthetic version by considering 
a neighborhood of $N$ motion sequences. The reference sequence $S_{ref}$ is given 
a weight of $1$, while each neighboring sequence $S_i$ is weighted based on its 
similarity to the reference. The weight $w_i$ for each neighbor is calculated as follows:

\begin{equation}\label{equ:w-sdba-weights}
    w_i = e^{ln(0.5).\dfrac{DTW(S_i,S_{ref})}{d_{NN}}}
\end{equation}
\noindent where $d_{NN}$ is the DTW distance between $S_{ref}$ and its nearest neighbor.
This weighting emphasizes the influence of sequences that are more similar to the reference.

\begin{figure}
    \centering
    \caption{
        Averaging six human motion sequences
        (\protect\mycolorbox{0,125,0,0.6}{Examples $0$ to $5$} shown on the top left/right) 
        with uniformly distributed weights results in an unrealistic example (bottom right 
        skeleton sequence). The 2D t-SNE~\cite{t-sne-paper}
        projection of these sequences (bottom left) 
        illustrates that the \protect\mycolorbox{255,30,0,0.6}{averaged sequence} 
        falls outside the \protect\mycolorbox{0,30,255,0.6}{manifold} of 
        the \protect\mycolorbox{0,125,0,0.6}{original sequences}.
    }
    \label{fig:w-sdba-sparse}
    \includegraphics[width=\textwidth]{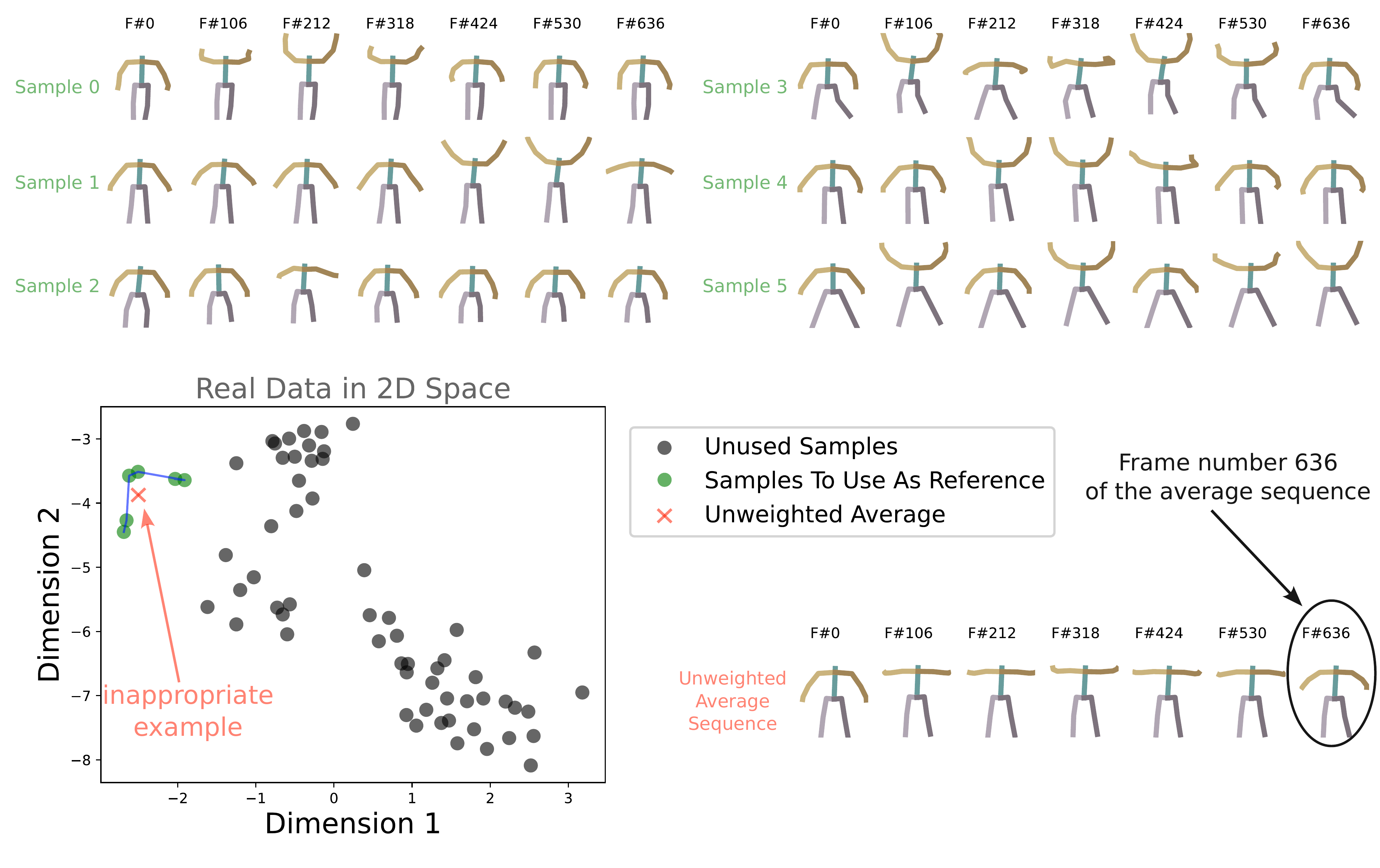}
\end{figure}

Figure~\ref{fig:w-sdba-sparse} demonstrates the significance of
using a weighted average in cases where 
the data manifold is sparse and the distribution is non-spherical.

For each reference sequence $S_{ref}$, the weighted ShapeDBA computation produces 
a corresponding synthetic sequence $\hat{S}$. To assign a continuous label (score) 
to the synthetic sequence, the same weights are normalized using $\min-\max$ normalization, 
ensuring the labels remain within the original range of $0$ to $1$. The continuous label 
$\hat{y}$ for the synthetic sequence $\hat{S}$ is calculated as:

\begin{equation}\label{equ:w-sdba-weighted-score}
    \hat{y} = \sum_{i=1}^{N+1} \bar{w}_i.y_i
\end{equation}
\noindent where $\bar{w}_i$ is the normalized weight of the $i_{th}$ sequence in the set 
of the $N+1$ samples (including the reference and its $N$ nearest neighbor sequences).

\subsubsection{Experimental Evaluation}

\paragraph{Experimental Setup}

We conducted a two-fold evaluation of our proposed approach. First, we examined the 
coherence of the synthetic sequences through both qualitative and quantitative 
analyses. Second, we investigated the utility of these synthetic sequences in 
improving a deep learning model's performance in extrinsic regression, aiming 
to predict continuous scores for rehabilitation sequences.

\paragraph*{Dataset}

We utilize the Kimore dataset~\cite{kimore-paper}, for which the pre-processing of the sequences 
are presented in Section~\ref{sec:kimore-explain}, however we use the regression task here.
We implemented a 5-fold cross-validation protocol with a unique adaptation to more 
accurately reflect real-world scenarios by including only unhealthy subjects in 
the test phase. Unlike the standard approach, we divided the sequences of unhealthy 
subjects into 5 folds. For each iteration, all sequences from healthy subjects and 4 
folds of unhealthy subjects were used for training, while the remaining fold of unhealthy 
subjects was reserved for testing. This method ensures that the evaluation is focused 
on the performance for unhealthy subjects, providing a more realistic assessment. 

\paragraph*{Comparative Sets of Rehabilitation Sequences}

Our comparative study involves several sets of rehabilitation sequences. The reference set, 
consisting of $D$ sequences, includes the original sequences from the Kimore dataset. 
As a baseline, we introduced random noise $\mathcal{N}(0,0.1)$ to each reference sequence, 
creating a noisy set of the same size. Beyond this, we generated five additional 
sets of synthetic sequences using our proposed weighted ShapeDBA method. For each 
reference sequence, we created synthetic versions by applying weighted ShapeDBA 
with neighborhood sizes ranging from $N = 1$ to $N = 5$. This process resulted in five 
distinct sets, each containing $D$ sequences, which we labeled as $ShapeDBA NN1$, 
$ShapeDBA NN2$, $ShapeDBA NN3$, $ShapeDBA NN4$, and $ShapeDBA NN5$.

\paragraph{Evaluation of Synthetic Data Coherence}\label{sec:evaluation-wsdba}

Evaluating generative models typically requires multiple methods. For human motion data, 
one way is to visually inspect the realism of the generated sequences. However, 
visual inspection alone is not enough, as it lacks quantitative objectivity,
as argued in~\cite{reliable-fidelity-diversity}. Thus, 
it is crucial to use specific metrics to assess the reliability of the generated 
samples. In this section, we introduce a comprehensive evaluation strategy that 
includes both visual and numerical assessments to ensure a thorough analysis of 
the generated sequences.

\paragraph*{Qualitative Analysis: Visualizing Real vs Generated}

\begin{figure}
    \centering
    \caption{
        Visualization of three examples: the top row shows a real sample, the 
        middle row displays a noisy sample, and the bottom row features a 
        sample generated using the weighted ShapeDBA method. Each sequence 
        is represented by 10 frames, arranged sequentially from left to right.
    }
    \label{fig:w-sdba-vs-noisy}
    \includegraphics[width=\textwidth]{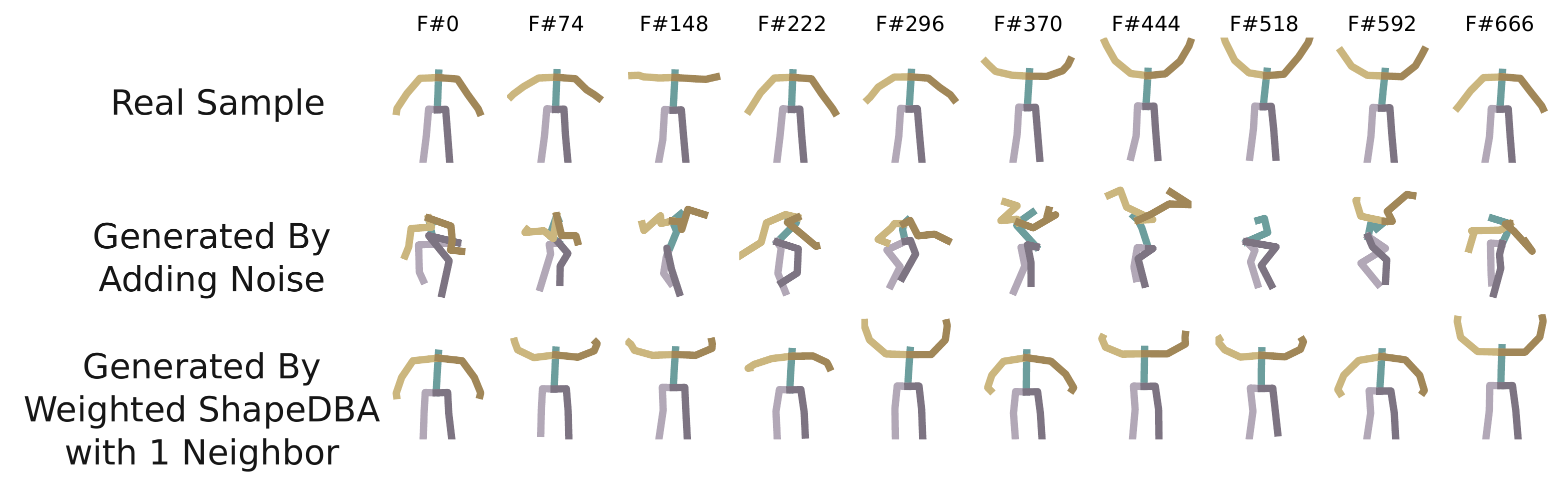}
\end{figure}

Figure~\ref{fig:w-sdba-vs-noisy} showcases three sequences:
the first is an original sequence from the Kimore 
dataset, the second is created by adding random noise to this original sequence, 
and the third is generated using our weighted ShapeDBA method. 
Visually, the noisy sequence appears less realistic compared to the sequence produced 
by the weighted ShapeDBA approach.

\begin{figure}
    \centering
    \caption{
        Visualization of two real sequences (top and bottom) and their corresponding 
        weighted ShapeDBA sequence (middle). The generated average sequence preserves 
        the temporal alignment of the top sequence, which has a higher weight, while 
        also incorporating features like the patient's height from the bottom sequence.
    }
    \label{fig:w-sdba-generation}
    \includegraphics[width=\textwidth]{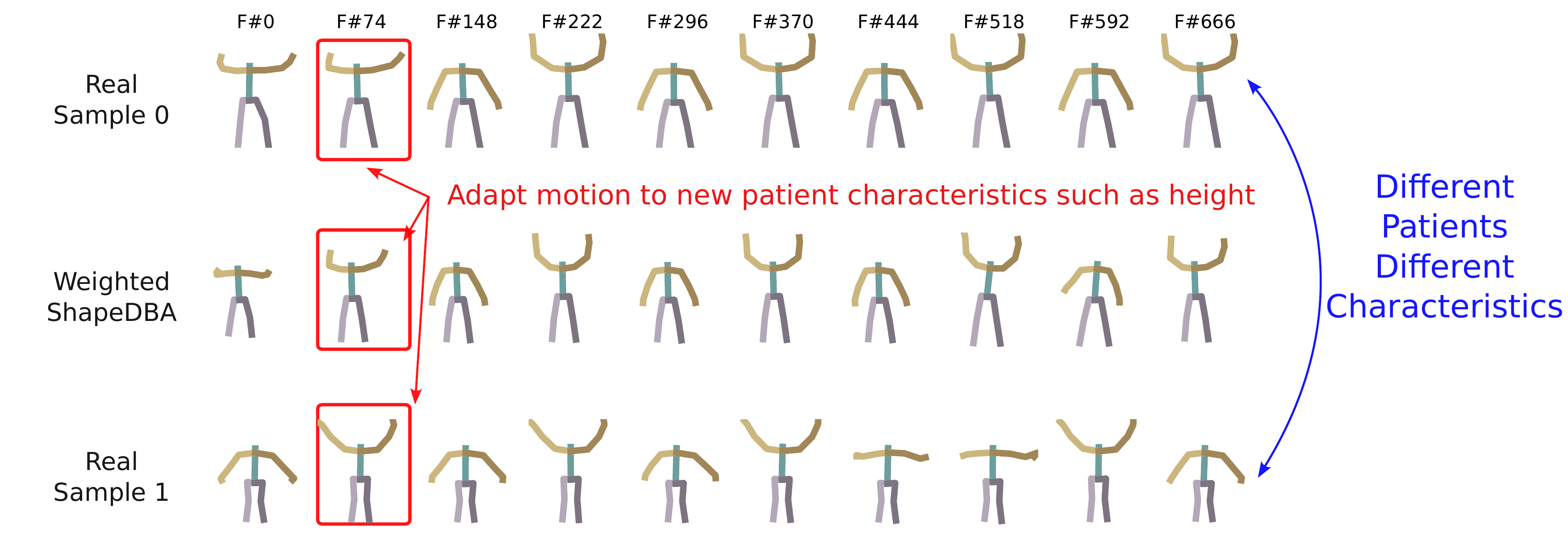}
\end{figure}

To illustrate the effectiveness of weighted ShapeDBA, Figure~\ref{fig:w-sdba-generation}
presents two sequences 
and their resulting weighted ShapeDBA sequence. The higher weight given to the first 
sequence (the reference) ensures that the temporal motion aligns closely with it. 
Simultaneously, the weighted ShapeDBA sequence incorporates attributes from the second 
sequence, such as the patient's height and form, effectively blending characteristics 
from both sequences to create a more coherent and realistic synthetic sequence.

\paragraph*{Quantitative Analysis: Fidelity and Diversity}

When extending a dataset, even using non-deep learning methods, it is crucial to 
evaluate both the fidelity and diversity of the generated samples. Fidelity metrics 
determine how closely the generated samples match real ones, with higher fidelity 
indicating more reliable samples for real-world use. Diversity metrics, on the other 
hand, assess the variability among samples, ensuring that both real and generated 
samples are distinct from one another. The goal of a generative model is typically 
to create a generated space that is as diverse as the original data.

In this study, we used two common metrics to evaluate fidelity and diversity: the 
Fréchet Inception Distance (FID) and the Average Pair Distance (APD), define below:

\begin{itemize}
    \item \textit{Fréchet Inception Distance} (FID): The FID metric evaluates the similarity 
    between the distributions of real and generated samples. To compute FID, both real 
    and generated samples first undergo latent feature extraction using a pre-trained 
    deep learning model $\mathcal{F}$ (in this study, trained on a regression task). The mean and 
    covariance of these latent features are then calculated for both sets of samples. 
    FID measures the distance between these distributions by assuming they follow a 
    Gaussian distribution. The mathematical formula for FID is:
    \begin{equation}\label{equ:fid}
        FID(\mathcal{P}_1,\mathcal{P}_2)^2 = \textit{trace}(\Sigma_1+\Sigma_2-2(\Sigma_1.\Sigma_2)^{1/2}) + \sum_{i=1}^{f}(\mu_{1,i}-\mu_{2,i})^2
    \end{equation}
    \noindent where, $\mathcal{P}_1$ and $\mathcal{P}_2$ represent the distributions of 
    real and generated samples, respectively. $f$ is the dimension of the latent space of 
    $\mathcal{F}$. $\mu_1$ and $\mu_2$ are vectors of dimension $f$, representing
    the means of the real and 
    generated samples' feature spaces, respectively. $\Sigma_1$ and $\Sigma_2$ are the covariance 
    matrices of dimensions $(f,f)$ for the real and generated samples, respectively.

    \item \textit{Average Pair Distance} (APD): The APD metric evaluates the average Euclidean Distance 
    (ED) between randomly selected samples from both real and generated data within the 
    feature space, utilizing $\mathcal{F}$ as the latent feature extractor. To compute APD, two 
    randomly selected sets of samples, $\mathcal{S}_1$ and $\mathcal{S}_2$,
    each containing $S_{apd}$ samples, 
    are defined. The average distance between these sets is calculated as:
    \begin{equation}\label{equ:apd}
        APD(\mathcal{S}_1,\mathcal{S}_2) = \dfrac{1}{S_{apd}}\sum_{i=1}^{S_{apd}} \sqrt{\sum_{j=1}^{f} (\mathcal{S}_{i,j} - \mathcal{S}^{'}_{i,j})^2} ,
    \end{equation}
    This process is repeated across multiple random sets of $\mathcal{S}_1$ and $\mathcal{S}_2$ 
    where the final APD value is the average of these calculations, which helps 
    to eliminate bias from set selection.

In this study, we utilized open-source software to compute the FID and APD metrics. 
The results were averaged over different initializations of our pre-trained deep 
regression model, which was trained solely on the training set. For the APD metric, 
we used $S_{apd} = 20$ for the size of the randomly selected sets.
\end{itemize}

\begin{table}
    \centering
    \caption{The FID values of different augmentation methods and the real dataset, over different resamples of each exercise.
    The presented FID values include the average and standard deviation over all resamples per exercise and different initialization of the pre-trained feature extractor.}
    \label{tab:w-sdba-fid}
    \resizebox{\columnwidth}{!}{%
    \begin{tabular}{c@{\quad}|@{\quad}c@{\quad}|@{\quad}c@{\quad}|@{\quad}c@{\quad}|@{\quad}c@{\quad}|@{\quad}c}
                 & \textbf{Exercise 1}                 & \textbf{Exercise 2}                 & \textbf{Exercise 3}                 & \textbf{Exercise 4}                 & \textbf{Exercise 5}                 \\ \hline
    Real         & 02.18$E^{-6}$~$\pm$~06.66$E^{-7}$ & 02.48$E^{-6}$~$\pm$~09.82$E^{-7}$ & 02.19$E^{-6}$~$\pm$~08.18$E^{-7}$ & 05.00$E^{-6}$~$\pm$~02.09$E^{-6}$ & 03.06$E^{-6}$~$\pm$~01.30$E^{-6}$ \\ \hline
    Noisy        & \textbf{00.07~$\pm$~00.03}             & \textbf{00.08~$\pm$~00.03}             & \textbf{00.09~$\pm$~00.03}             & \textbf{00.24~$\pm$~00.10}             & \textbf{00.07~$\pm$~00.03}             \\
    ShapeDBA NN1 & {\ul 01.94~$\pm$~00.81}             & {\ul 04.12~$\pm$~01.86}             & {\ul 02.28~$\pm$~01.10}             & {\ul 04.19~$\pm$~02.95}             & {\ul 03.39~$\pm$~01.92}             \\
    ShapeDBA NN2 & 03.15~$\pm$~01.06             & 06.62~$\pm$~02.56             & 04.01~$\pm$~02.70             & 05.95~$\pm$~04.07             & 05.75~$\pm$~02.34             \\
    ShapeDBA NN3 & 03.77~$\pm$~01.21             & 07.98~$\pm$~02.59             & 05.16~$\pm$~03.39             & 07.16~$\pm$~04.51             & 07.34~$\pm$~02.91             \\
    ShapeDBA NN4 & 04.31~$\pm$~01.24             & 09.38~$\pm$~03.05             & 05.96~$\pm$~03.52             & 08.01~$\pm$~04.64             & 08.51~$\pm$~03.76             \\
    ShapeDBA NN5 & 04.62~$\pm$~01.28             & 10.21~$\pm$~03.34             & 06.45~$\pm$~03.69             & 08.90~$\pm$~05.07             & 09.23~$\pm$~04.12            
    \end{tabular}%
    }
\end{table}

\begin{table}
    \centering
    \caption{The APD values of different augmentation method and the real dataset, over different resamples of each exercise.
    The presented APD values include the average and standard deviation over all resamples per exercise, different randomly selected sets of size $S_{apd}=20$ and different initialization of the pre-trained feature extractor.}
    \label{tab:w-sdba-apd}
    \resizebox{\columnwidth}{!}{%
    \begin{tabular}{c@{\quad}|@{\quad}c@{\quad}|@{\quad}c@{\quad}|@{\quad}c@{\quad}|@{\quad}c@{\quad}|@{\quad}c}
     & \textbf{Exercise 1} & \textbf{Exercise 2} & \textbf{Exercise 3} & \textbf{Exercise 4} & \textbf{Exercise 5} \\ \hline
    Real & 06.09~$\pm$~00.63 & 06.61~$\pm$~00.83 & 05.95~$\pm$~00.78 & 08.24~$\pm$~01.24 & 07.10~$\pm$~01.10 \\ \hline
    Noisy & \textbf{06.05~$\pm$~00.57} & \textbf{06.55~$\pm$~00.79} & \textbf{05.90~$\pm$~00.78} & \textbf{08.14~$\pm$~01.18} & \textbf{07.00~$\pm$~01.03} \\
    ShapeDBA NN1 & {\ul 05.21~$\pm$~00.68} & {\ul 05.32~$\pm$~00.63} & {\ul 05.06~$\pm$~00.68} & {\ul 07.14~$\pm$~01.12} & {\ul 06.11~$\pm$~01.08} \\
    ShapeDBA NN2 & 04.93~$\pm$~00.69 & 04.96~$\pm$~00.66 & 04.77~$\pm$~00.65 & 06.90~$\pm$~01.16 & 05.70~$\pm$~00.98 \\
    ShapeDBA NN3 & 04.78~$\pm$~00.75 & 04.64~$\pm$~00.54 & 04.53~$\pm$~00.66 & 06.84~$\pm$~01.20 & 05.52~$\pm$~00.96 \\
    ShapeDBA NN4 & 04.67~$\pm$~00.74 & 04.47~$\pm$~00.55 & 04.34~$\pm$~00.63 & 06.59~$\pm$~01.16 & 05.35~$\pm$~00.91 \\
    ShapeDBA NN5 & 04.61~$\pm$~00.67 & 04.35~$\pm$~00.53 & 04.30~$\pm$~00.61 & 06.50~$\pm$~01.13 & 05.31~$\pm$~00.90
    \end{tabular}%
    }
\end{table}

For each augmentation method—noisy and weighted ShapeDBA, we present the average FID 
and APD values along with their standard deviations. These metrics are calculated over 
various resamples of the dataset for each exercise, as well as different initializations
of the pre-trained deep regression model.

Table~\ref{tab:w-sdba-fid} displays the FID values
for each exercise using the noisy augmentation method and 
five variations of our weighted ShapeDBA (with neighborhood sizes ranging from $1$ to $5$). 
The FID for real samples is also included as a baseline. Ideally, the FID of any generative 
method should be close to, but slightly higher than, the FID of the real data, as generated 
samples cannot exceed the fidelity of the actual data. As seen in Table~\ref{tab:w-sdba-fid},
the noisy method has a lower FID compared 
to the weighted ShapeDBA method. On one hand, this difference can be explained by the fact that the dataset exists 
in a sparse space. Adding a small amount of noise generates a point in the encoder's latent space that 
is very close to the real sample. On the other hand, generating data with ShapeDBA introduces new points 
into this sparse space. Given the small amount of data, the FID metric may appear ``better'' 
with noise augmentation.
Despite this, Figure~\ref{fig:w-sdba-vs-noisy}
clearly 
shows that the noisy augmentation produces unrealistic sequences. This highlights the 
importance of using multiple evaluation methods to draw comprehensive conclusions about 
generative models, as there is no single best approach.

The same trend is observed with the APD values in Table~\ref{tab:w-sdba-apd}. The noisy method has an 
APD closest to that of real samples, outperforming the weighted ShapeDBA method 
in quantitative evaluations.

\paragraph{Evaluation of Synthetic Sequences for Data Extension}

In our second experiment, we assess the effectiveness of the proposed method as 
a data extension technique for rehabilitation assessment. This task is framed 
as an extrinsic regression problem, where the aim is to predict a continuous 
performance score associated with each rehabilitation sequence.
We use the Fully Convolutional Network (Chapter~\ref{chapitre_1} Figure~\ref{fig:fcn})
as our backbone model (same used as feature extractor
for generation metrics in previous section).

To compare the predicted scores from our models with the clinical scores provided by 
experts, we used two metrics as outlined by~\cite{regression-metrics-paper}:
the Root Mean Square Error (RMSE) 
and the Mean Absolute Error (MAE). Given two sets of $N$ scores, $\textbf{y}$ (the ground truth) 
and $\hat{\textbf{y}}$ (the predictions), the MAE and RMSE are calculated as follows:
\begin{equation}\label{equ:rmse}
    MAE(\textbf{y},\hat{\textbf{y}}) = \dfrac{1}{N}~\sum_{i=1}^N~|y_i-\hat{y}_i|,
\end{equation}
\begin{equation}\label{equ:mae}
    RMSE(\textbf{y},\hat{\textbf{y}}) = \sqrt{\dfrac{1}{N}~\sum_{i=1}^N~(y_i-\hat{y}_i)^2}.
\end{equation}

\begin{table}
    \caption{MAE and RMSE errors obtained for all compared approaches on each exercise separately. Best values are emphasized in bold, while second best values are underlined.}
    \centering
    \label{tab:w-sdba-mae-rmse}
    \resizebox{\columnwidth}{!}{
        \begin{tabular}
            {c@{\quad}|@{\quad}c@{\quad}|@{\quad}c@{\quad}|@{\quad}c@{\quad}|@{\quad}c@{\quad}|@{\quad}c}
            \multicolumn{1}{c|}{Training Set} & \multicolumn{1}{c|}{Exercise 1} & \multicolumn{1}{c|}{Exercise 2} & \multicolumn{1}{c|}{Exercise 3} & \multicolumn{1}{c|}{Exercise 4} & Exercise 5 \\ \hline
            \multicolumn{6}{c}{MAE} \\ \hline
            
            \multicolumn{1}{@{\quad}c@{\quad}|}{Ref.} & \multicolumn{1}{@{\quad}c@{\quad}|}{0.206 $\pm$ 0.069} & \multicolumn{1}{@{\quad}c@{\quad}|}{0.202 $\pm$ 0.037} & \multicolumn{1}{@{\quad}c@{\quad}|}{0.204 $\pm$ 0.055} & \multicolumn{1}{@{\quad}c@{\quad}|}{0.184 $\pm$ 0.068} & {\ul 0.224 $\pm$ 0.058} \\
            
            \multicolumn{1}{@{\quad}c@{\quad}|}{Ref. + Noise} & \multicolumn{1}{@{\quad}c@{\quad}|}{0.186 $\pm$ 0.065} & \multicolumn{1}{@{\quad}c@{\quad}|}{\textbf{0.172 $\pm$ 0.040}} & \multicolumn{1}{@{\quad}c@{\quad}|}{0.203 $\pm$ 0.045} & \multicolumn{1}{@{\quad}c@{\quad}|}{0.185 $\pm$ 0.073} & 0.229 $\pm$ 0.069 \\
            
            \multicolumn{1}{@{\quad}c@{\quad}|}{Ref. + ShapeDBA NN1} & \multicolumn{1}{@{\quad}c@{\quad}|}{ {\ul 0.167 $\pm$ 0.070}} & \multicolumn{1}{@{\quad}c@{\quad}|}{{\ul 0.175 $\pm$ 0.030}} & \multicolumn{1}{@{\quad}c@{\quad}|}{\textbf{0.182 $\pm$ 0.051}} & \multicolumn{1}{@{\quad}c@{\quad}|}{\textbf{0.141 $\pm$ 0.062}} & \textbf{0.208 $\pm$ 0.079} \\
            
            \multicolumn{1}{@{\quad}c@{\quad}|}{Ref. + ShapeDBA NN2} & \multicolumn{1}{@{\quad}c@{\quad}|}{0.169 $\pm$ 0.057} & \multicolumn{1}{@{\quad}c@{\quad}|}{ 0.177 $\pm$ 0.041} & \multicolumn{1}{@{\quad}c@{\quad}|}{{\ul 0.194 $\pm$ 0.041}} & \multicolumn{1}{@{\quad}c@{\quad}|}{{\ul 0.168 $\pm$ 0.056}} & 0.226 $\pm$ 0.066 \\
            
            \multicolumn{1}{@{\quad}c@{\quad}|}{Ref. + ShapeDBA NN3} & \multicolumn{1}{@{\quad}c@{\quad}|}{0.173 $\pm$ 0.063} & \multicolumn{1}{@{\quad}c@{\quad}|}{0.183 $\pm$ 0.047} & \multicolumn{1}{@{\quad}c@{\quad}|}{0.199 $\pm$ 0.058} & \multicolumn{1}{@{\quad}c@{\quad}|}{0.168 $\pm$ 0.083} & 0.225 $\pm$ 0.055 \\
            
            \multicolumn{1}{@{\quad}c@{\quad}|}{Ref. + ShapeDBA NN4} & \multicolumn{1}{@{\quad}c@{\quad}|}{0.168 $\pm$ 0.059} & \multicolumn{1}{@{\quad}c@{\quad}|}{0.179 $\pm$ 0.043} & \multicolumn{1}{@{\quad}c@{\quad}|}{0.199 $\pm$ 0.043} & \multicolumn{1}{@{\quad}c@{\quad}|}{0.180 $\pm$ 0.080} & 0.231 $\pm$ 0.060 \\
            
            \multicolumn{1}{@{\quad}c@{\quad}|}{Ref. + ShapeDBA NN5} & \multicolumn{1}{@{\quad}c@{\quad}|}{\textbf{0.166 $\pm$ 0.067}} & \multicolumn{1}{@{\quad}c@{\quad}|}{0.185 $\pm$ 0.043} & \multicolumn{1}{@{\quad}c@{\quad}|}{0.201 $\pm$ 0.050} & \multicolumn{1}{@{\quad}c@{\quad}|}{0.182 $\pm$ 0.089} & 0.226 $\pm$ 0.061 \\ \hline        
            
            \multicolumn{6}{@{\quad}c@{\quad}}{RMSE} \\ \hline
            
            \multicolumn{1}{@{\quad}c@{\quad}|}{Ref.} & \multicolumn{1}{@{\quad}c@{\quad}|}{0.251 $\pm$ 0.083} & \multicolumn{1}{@{\quad}c@{\quad}|}{0.247 $\pm$ 0.045} & \multicolumn{1}{@{\quad}c@{\quad}|}{0.248 $\pm$ 0.065} & \multicolumn{1}{@{\quad}c@{\quad}|}{0.230 $\pm$ 0.083} & {\ul 0.267 $\pm$ 0.073} \\
            
            \multicolumn{1}{@{\quad}c@{\quad}|}{Ref. + Noise} & \multicolumn{1}{@{\quad}c@{\quad}|}{ 0.203 $\pm$ 0.078} & \multicolumn{1}{@{\quad}c@{\quad}|}{{\ul 0.226 $\pm$ 0.043}} & \multicolumn{1}{@{\quad}c@{\quad}|}{ 0.238 $\pm$ 0.046} & \multicolumn{1}{@{\quad}c@{\quad}|}{ 0.227 $\pm$ 0.090} & 0.274 $\pm$ 0.092 \\
            
            \multicolumn{1}{@{\quad}c@{\quad}|}{Ref. + ShapeDBA NN1} & \multicolumn{1}{@{\quad}c@{\quad}|}{{\ul 0.199 $\pm$ 0.087}} & \multicolumn{1}{@{\quad}c@{\quad}|}{\textbf{0.226 $\pm$ 0.036}} & \multicolumn{1}{@{\quad}c@{\quad}|}{\textbf{0.214 $\pm$ 0.054}} & \multicolumn{1}{@{\quad}c@{\quad}|}{\textbf{0.178 $\pm$ 0.074}} & \textbf{0.251 $\pm$ 0.094} \\
            
            \multicolumn{1}{@{\quad}c@{\quad}|}{Ref. + ShapeDBA NN2} & \multicolumn{1}{@{\quad}c@{\quad}|}{0.203 $\pm$ 0.075} & \multicolumn{1}{@{\quad}c@{\quad}|}{0.232 $\pm$ 0.052} & \multicolumn{1}{@{\quad}c@{\quad}|}{{\ul 0.226 $\pm$ 0.044}} & \multicolumn{1}{@{\quad}c@{\quad}|}{{\ul 0.210 $\pm$ 0.074}} & 0.268 $\pm$ 0.083 \\
            
            \multicolumn{1}{@{\quad}c@{\quad}|}{Ref. + ShapeDBA NN3} & \multicolumn{1}{@{\quad}c@{\quad}|}{0.205 $\pm$ 0.082} & \multicolumn{1}{@{\quad}c@{\quad}|}{0.235 $\pm$ 0.050} & \multicolumn{1}{@{\quad}c@{\quad}|}{0.240 $\pm$ 0.062} & \multicolumn{1}{@{\quad}c@{\quad}|}{0.214 $\pm$ 0.105} & 0.268 $\pm$ 0.066 \\
            
            \multicolumn{1}{@{\quad}c@{\quad}|}{Ref. + ShapeDBA NN4} & \multicolumn{1}{@{\quad}c@{\quad}|}{\textbf{0.198 $\pm$ 0.071}} & \multicolumn{1}{@{\quad}c@{\quad}|}{0.235 $\pm$ 0.050} & \multicolumn{1}{@{\quad}c@{\quad}|}{0.234 $\pm$ 0.048} & \multicolumn{1}{@{\quad}c@{\quad}|}{0.230 $\pm$ 0.105} &  0.279 $\pm$ 0.070 \\
            
            \multicolumn{1}{@{\quad}c@{\quad}|}{Ref. + ShapeDBA NN5} & \multicolumn{1}{@{\quad}c@{\quad}|}{0.202 $\pm$ 0.079} & \multicolumn{1}{@{\quad}c@{\quad}|}{0.230 $\pm$ 0.049} & \multicolumn{1}{@{\quad}c@{\quad}|}{0.244 $\pm$ 0.057} & \multicolumn{1}{@{\quad}c@{\quad}|}{0.231 $\pm$ 0.109} & \multicolumn{1}{@{\quad}c@{\quad}}{0.280 $\pm$ 0.080}         
            \end{tabular}
    }
\end{table}

We evaluated the performance of the FCN model for extrinsic regression by training 
it on various sets of rehabilitation sequences 
combined with the reference set. The average MAE and RMSE errors ($\pm$ standard deviation) 
are reported in Table~\ref{tab:w-sdba-mae-rmse}.

Our first observation is that adding noisy sequences to the training set 
($Ref. + Noise$) generally leads to better performance compared to using only 
the original training set ($Ref.$). This suggests that the FCN model tends to 
overfit the original data, and incorporating noisy sequences helps to improve 
its generalization capabilities.

Moreover, Table~\ref{tab:w-sdba-mae-rmse} highlights that the best
error values for both metrics are achieved 
when the FCN model is trained on the rehabilitation set extended with ShapeDBA-generated 
averages. This indicates that synthetic average sequences not only mitigate overfitting 
but also capture realistic rehabilitation motion patterns, allowing the FCN model to 
more effectively learn the variations in rehabilitation exercises.

Finally, we note that for three out of five exercises, the optimal performance is 
obtained when the training data includes average sequences computed using a single 
neighbor ($Ref. + ShapeDBA NN1$). This suggests that using a larger neighborhood may 
sometimes result in less coherent average sequences, as the neighbors might not 
reside within a continuous subspace.

The experiments demonstrated that incorporating synthetic sequences generated by the 
weighted ShapeDBA method significantly enhances the performance of rehabilitation 
assessment models. This approach mitigates overfitting and captures realistic motion 
patterns, leading to more accurate predictions. However, the reliance on real data 
distribution and the computational expense of finding nearest neighbors and applying 
ShapeDBA can be problematic in real-world scenarios where instant generation is required.
In the next section, we address this challenge by exploring the use of deep learning models 
for generation tasks, aiming to mitigate the limitations of dataset distribution and provide 
a more efficient solution.

\section{Exploring Deep Generative Models for Human Motion Generation}

In this section, we investigate the application of deep generative models 
for the task of human motion generation. The need for generating 
realistic and diverse motion sequences efficiently has spurred 
significant interest in this domain. Traditional prototyping 
methods, such as ShapeDBA, though effective, can be computationally 
intensive and time-consuming, particularly during the inference 
(generation) phase. This limitation makes them less suitable for 
real-time applications where quick data generation is crucial. 
Deep generative models, on the other hand, offer a substantial 
advantage in terms of speed and efficiency during the inference 
phase, making them highly suitable for these scenarios.

Generative models can revolutionize various fields, including the 
cinematic and gaming industries, by enabling the creation of 
lifelike and varied character
animations~\cite{gta-dataset-paper,generating-cut-scenes-paper}.
In medical rehabilitation~\cite{rehab-generation-paper}, 
these models can assist in creating realistic motion sequences for 
better patient assessment and treatment planning. The flexibility 
and scalability of deep generative models allow them to generate 
large volumes of high-quality motion sequences, essential for these 
applications. Additionally, fine-tuning these models~\cite{motiongpt-fine-tune-paper}
on small 
datasets helps avoid overfitting, ensuring the generated sequences 
are diverse and realistic without requiring extensive training data.

\begin{figure}
    \centering
    \caption{
        Summary of our proposed Supervised VAE (SVAE) architecture: The 
        \protect\mycolorbox{0,30,255,0.6}{input 
        skeleton sequences} are processed by the \protect\mycolorbox{102,220,230,0.6}{encoder} 
        to learn a \protect\mycolorbox{255,165,0,0.6}{Gaussian 
        distribution in the latent space}. From this distribution, a
        \protect\mycolorbox{255,30,0,0.6}{random sample} 
        is drawn and utilized in two ways: it is fed into a
        \protect\mycolorbox{255,100,255,0.6}{classifier} for the 
        action recognition task and also into the
        \protect\mycolorbox{48,255,0,0.6}{decoder} to reconstruct the 
        \protect\mycolorbox{0,30,255,0.6}{original sequence}. 
        This dual functionality enhances both the generative 
        and discriminative capabilities of the model.
    }
    \label{fig:svae-summary}
    \includegraphics[width=0.6\textwidth]{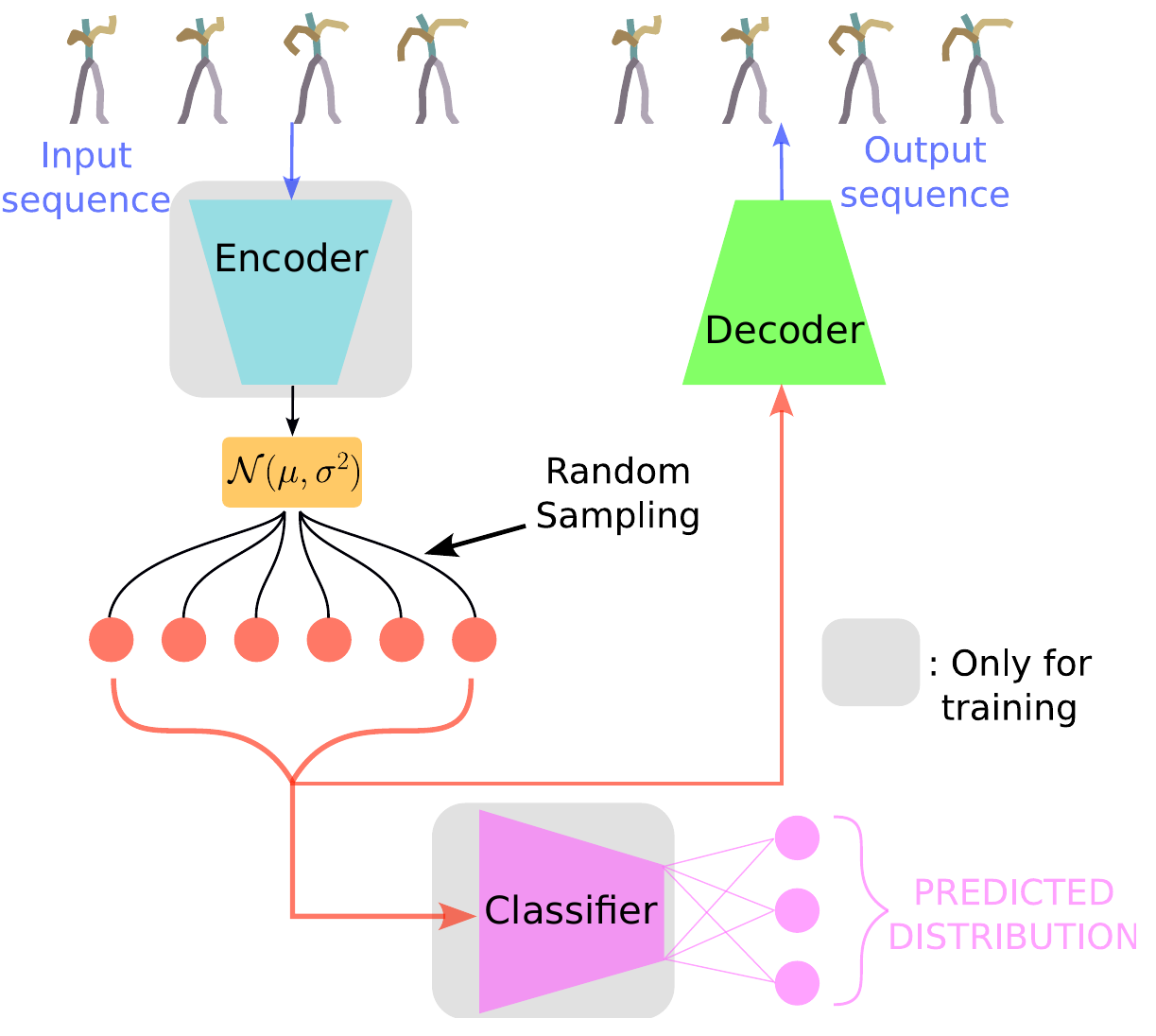}
\end{figure}

Convolutional Neural Networks (CNNs) have proven highly effective in TSC,
as seen in chapters~\ref{chapitre_3} and~\ref{chapitre_4},
yet their application to human motion generation remains relatively 
unexplored. Despite this, CNNs offer distinct advantages over Recurrent Neural 
Networks (RNNs)~\cite{action2motion-paper} and Transformers~\cite{actor-paper},
particularly regarding inference time and computational 
efficiency (measured in FLOPS). To harness these benefits, we propose a CNN-based 
Variational Auto-Encoder (VAE) generative model tailored for human motion data, 
specifically for action recognition tasks as it is directly conditioned with labels.
The traditional approach to incorporating a conditioning aspect into a VAE model is 
by simply concatenating a one-hot encoded vector, representing the action label, with 
the latent space before feeding it to the decoder. However, this technique may have 
limitations in terms of sensitivity, as the model is not explicitly trained to recognize 
that the action label is significantly different from another.
To improve the conditioning and effectively separate the latent space, we introduce a 
classification task within the latent space by proposing a Supervised VAE (SVAE),
summarized in Figure~\ref{fig:svae-summary}.
Our model demonstrates competitiveness with state-of-the-art methods in terms of 
fidelity and diversity metrics. Additionally, we address the challenge of balancing 
training datasets concerning class label distribution, thereby enhancing downstream 
classification performance.

In the following sections, we detail some of the background work on human motion 
generation using deep learning models, the proposed architecture and its specific 
features, followed by the experimental results using the publicly available 
HumanAct12 dataset~\cite{action2motion-paper} (Figure~\ref{fig:example-humanact}).

\subsection{Background Work}

Recent advancements in deep generative models have brought significant 
improvements in the generation of human motion sequences. This 
subsection provides an overview of the key state-of-the-art models, 
highlighting their unique contributions and advantages.
The current state-of-the-art in this field lies within the 
capabilities of VAEs, GANs, and diffusion models.

\subsubsection{Generative Adversarial Networks (GANs)}

Generative Adversarial Networks (GANs)~\cite{gan-paper} employ a
dual-network structure consisting of 
a generator and a discriminator. The generator creates synthetic data, while the 
discriminator evaluates the authenticity of the generated data against real data. 
Through this adversarial process, GANs learn to produce highly realistic data that 
can often be indistinguishable from real samples. The ability to generate 
high-quality data quickly makes GANs especially appealing for applications 
requiring rapid generation of human motion sequences, such as in real-time 
game character animation or virtual reality environments.
Before the availability of 3D human motion datasets, researchers addressed the motion 
generation problem at a 2D level. One notable approach is the two-stage Generative 
Adversarial Network (GAN) introduced in~\cite{two-stage-gan-paper},
which generates 2D videos of human motion. 
This model works in two phases: first, it creates a human skeleton from random noise, 
and then it transforms this skeleton into an image, repeating this for multiple 
frames to produce a coherent video sequence. Another significant model is
MoCoGAN~\cite{mocogan-paper}, 
a GAN-based approach that generates motion by conditioning on specific content, learning 
two distinct spaces—one for content representation and the other for frame sequences. 
These innovative models, Two-Stage GAN and MoCoGAN, were later adapted for 3D human motion 
sequences, as demonstrated in~\cite{action2motion-paper},
paving the way for more advanced and realistic 3D human 
motion generation techniques.

\subsubsection{Variational Auto-Encoders (VAEs)}

Variational Auto-Encoders (VAEs)
(Chapter~\ref{chapitre_1} Section~\ref{sec:tscl})
learn to encode data into a Gaussian latent 
space and then decode it back into the data space, effectively 
capturing the underlying distribution of the training data. 
This allows VAEs to generate new data points that are coherent 
and representative of the original data distribution. 
The capability of VAEs to produce high-quality, diverse 
data points makes them particularly useful in applications 
such as the cinematic and gaming industries, where creating 
lifelike and varied character animations is essential.
With the availability of 3D human motion datasets, researchers in~\cite{action2motion-paper}
highlighted the 
limitations of existing datasets and introduced HumanAct12, a new dataset derived from 
the PHSPD dataset~\cite{PHSPD-dataset-paper,PHSPD-dataset-paper},
which utilizes a polarization camera and three Kinect v2 cameras~\cite{kinect-paper}.
Additionally, the authors in~\cite{action2motion-paper}
proposed Action2Motion, a VAE model, called Action2Motion,
designed to generate skeleton-based human motion sequences. This auto-regressive VAE 
consists of two encoders, a prior and a posterior encoder, and one decoder. The model approximates 
a latent representation of the prior and posterior time frames while minimizing the 
Kullback-Leibler (KL) divergence between their distributions to ensure regularization. 
Importantly, Action2Motion is a conditional VAE, with conditioning based on the action label
and the timestamp to differentiate between prior and posterior pose frames.
Other researchers have also explored conditional VAE modeling. For example~\cite{LCP-VAE-paper}
introduced a conditional VAE that learns a latent representation of the condition
instead of adding the label directly. This model comprises two VAEs, CS-VAE and 
LCP-VAE, one encoding the prior knowledge (condition) and the other encoding the 
future sequence (posterior knowledge). The model can be applied in two ways: using 
the action label as prior knowledge or using a part of the training sequence as 
prior knowledge. The latter approach has shown better performance, as using a real 
human motion sequence as prior knowledge increases the diversity of the generated samples.
Transformers have recently shown significant impact in translation
models~\cite{attention-all-you-need}
and image recognition~\cite{vision-transformer-paper}.
Building on this, the authors in~\cite{actor-paper} developed ACTOR, 
a Transformer-based VAE model to generate 3D human motion sequences.
Moreover,~\cite{posegpt-paper} introduced a Quantized VAE~\cite{vq-vae-paper}
to generate 3D human motion sequences. 
This model employs a Generative Pre-trained Transformer (GPT)-like model~\cite{gpt-1-paper}
in the 
latent space after quantization to predict latent indices. More recently, 
researchers in~\cite{um-cvae-paper} proposed a VAE that encodes the input human motion sequence 
in two streams simultaneously, resulting in action-agnostic and action-aware 
representations, known as UM-CVAE.

\subsubsection{Denoising Diffusion Probabilistic Models (DDPMs)}

Recently, the advancement of Denoising Diffusion Probabilistic Models
(DDPMs)~\cite{ddpm-paper}
has added another dimension to the 
field of generative models. DDPMs gradually transform simple 
initial data, such as Gaussian noise, into complex data distributions through 
a series of steps. This approach allows for high-quality data generation and 
has shown impressive results in generating detailed and realistic motion sequences. 
Diffusion models have advanced the field by providing a robust framework for 
generating complex data, further enhancing the ability of generative models to
create lifelike human motion sequences efficiently.
For instance,~\cite{motion-diffuse-paper} introduced MotionDiffuse, a
DDPM for human motion generation. This diffusion model employs an attention 
mechanism architecture for the denoising process, incorporating residual 
connections instead of the U-Net~\cite{u-net-paper}
architecture used in the original DDPM model~\cite{ddpm-paper}.

\subsection{Proposed Model}

\begin{figure}
    \centering
    \caption{
        The Variational Auto-Encoder (VAE) model for human motion generation 
        leverages an FCN backbone in both the Encoder and Decoder. In the Decoder, 
        \protect\mycolorbox{0,230,130,1.0}{standard one-dimensional convolutions} 
        are replaced with \protect\mycolorbox{255,199,20,1.0}{transposed convolutions}. 
        Each convolution layer is followed by batch normalization and a ReLU activation 
        function. The SVAE model further includes
        \protect\mycolorbox{0,158,255,0.57}{Fully Connected layers} 
        to enhance its functionality.
    }
    \label{fig:svae-detailed}
    \includegraphics[width=0.75\textwidth]{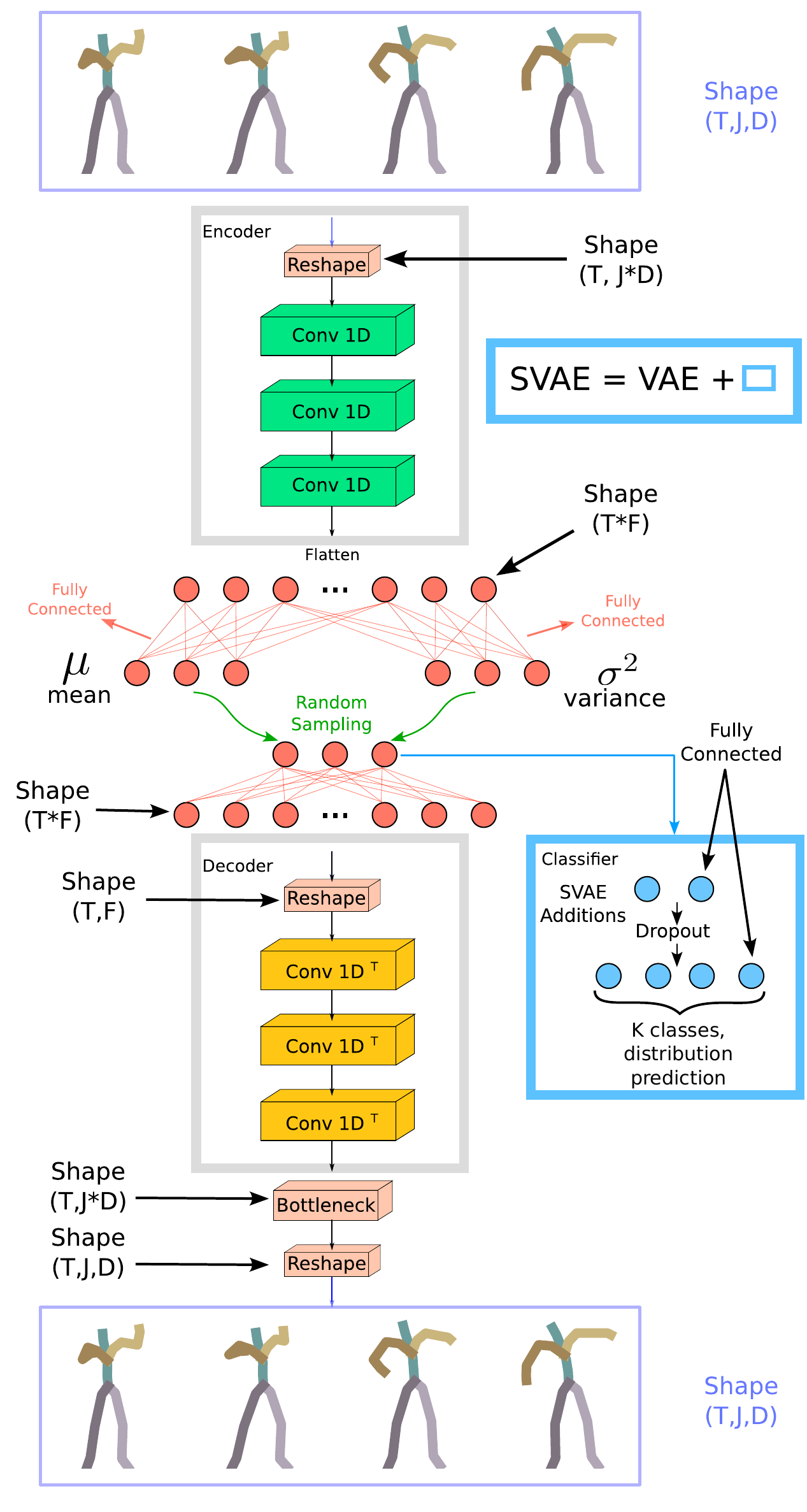}
\end{figure}

In this section, we introduce our proposed Supervised VAE (SVAE) architecture, 
which builds upon the original Variational Auto-Encoder (VAE) model by 
incorporating an associated classification task. Both models utilize a 
Fully Convolutional Network (FCN) architecture~\cite{fcn-resnet-mlp-paper} as the backbone. 
The encoder employs one-dimensional convolutions for down-sampling, 
while the decoder uses transposed one-dimensional convolutions for up-sampling.
Below, we present the original VAE and the Supervised VAE, both based on 
CNN architecture, as illustrated in Figure~\ref{fig:svae-detailed}.

\subsubsection{Original Variational Auto-Encoder (VAE)}

The VAE architecture, initially proposed in~\cite{vae-paper}, has been adapted for time series data, 
specifically targeting human motion sequences. The optimizer in this model minimizes 
two key losses: the reconstruction loss and the Kullback-Leibler (KL) divergence, as 
detailed in Chapter~\ref{chapitre_1} Section~\ref{sec:tscl}.
Notably, the reconstruction loss deviates from the standard Mean Squared Error 
loss by excluding the averaging over the time dimension, as shown below:
\begin{equation}\label{equ:svae-rec-loss}
    \mathcal{L}_{rec} (\textbf{x}, \hat{\textbf{x}}) = \dfrac{1}{J.D} \sum_{t=1}^{L}\sum_{m=1}^{J.D} (x_t^m - \hat{x}_t^m)^2,
\end{equation}
\noindent where $\textbf{x}$ and $\hat{\textbf{x}}$ are the MTS representation 
of the input skeleton sequence and the reconstructed one, both of length $L$ and $JxD$ dimensions,
and $J$ is the number of joints on the recorded skeleton each in a space of $D$ dimensions.
This modification prevents the model from converging to the average sequence during 
training, which would lead to underfitting due to reduced dimensionality in the 
optimization problem.

The KL divergence loss aligns the Gaussian distribution in the latent space to a standard
normal distribution with zero mean and unit variance, such as Eq.~\ref{equ:vae-kl} in
Chapter~\ref{chapitre_1}:

\begin{equation}\label{equ:vae-kl-human-motion}
\begin{split}
    \mathcal{L}_{KL} &= D_{KL}(q_{\theta}(\textbf{z}|\textbf{x}),\mathcal{N}(0,1))\\
    &= -\dfrac{1}{2}~\sum_{d=1}^\textbf{d}(1+\log\sigma_d^2-\mu_d^2-\sigma_d^2)
\end{split}
\end{equation}
\noindent where $\textbf{d}$ is the dimension of the latent space, specific to the architecture 
of the VAE, $\boldsymbol{\mu}$ and $\log~\boldsymbol{\sigma}^2$
are the mean and $\log$ variance vectors of the latent space features.

\subsubsection{Supervised Variational Auto-Encoder (SVAE)}

To enhance the VAE with conditioning capabilities, we introduce a classification 
task within the latent space. The SVAE model integrates an MLP based classifier~\cite{fcn-resnet-mlp-paper}
within the latent space. Both the classifier and the decoder utilize a randomly 
sampled vector from the Gaussian distribution produced by the encoder. This
additional classification task brings a Cross Entropy (CE) loss into the optimization 
process, alongside the reconstruction and KL divergence losses:
\begin{equation}\label{equ:cross-entropy-human-motion}
    \mathcal{L}_{CE}(\textbf{y},\hat{\textbf{y}}) = -\sum_{c=1}^{C}y_{c}.\log_{2}(\hat{y}_{c})
\end{equation}
\noindent where $\textbf{y}$ and $\hat{\textbf{y}}$ represent the true and predicted 
class distributions, respectively, and $C$ denotes the number of classes.

Inspired by the $\beta$-VAE~\cite{beta-vae-paper}, each loss in the SVAE model is weighted.
In $\beta$-VAE, the total loss is modulated by a hyperparameter $\beta$:
\begin{equation}\label{equ:beta-vae-total-loss}
    \mathcal{L}_{\beta-vae} = (1-\beta).\mathcal{L}_{mse} + \beta.\mathcal{L}_{KL}
\end{equation}
The $\beta$ parameter balances the emphasis between reconstruction quality and 
latent space disentanglement. Higher $\beta$ values prioritize learning 
disentangled latent features, while lower values focus on achieving 
better reconstruction quality. In our SVAE model, the total loss is calculated as:
\begin{equation}\label{equ:svae-loss-weights}
    \mathcal{L}_{total} = \lambda_{rec} \cdot \mathcal{L}_{rec} + \lambda_{KL} \cdot \mathcal{L}_{KL} + \lambda_{CE} \cdot \mathcal{L}_{CE}.
\end{equation}
For the original VAE architecture, the CE loss weight $\lambda_{CE}$
is set to zero. The optimal values for these weights were determined through
extensive experiments, detailed in the experimental section.

\subsection{Experimental Setup}

\subsubsection{Dataset}

To evaluate our proposed model, we employ the open-source HumanAct12 dataset 
as described in~\cite{action2motion-paper}. This dataset consists of $1,191$ human motion sequences, 
each depicting one of $12$ different actions. Each frame in these sequences 
represents a 3D skeleton composed of $24$ joints, for a total of $72$ dimensions.
To manage the variation in 
sequence lengths, we apply a resampling algorithm
that adjusts all sequences 
to a uniform target length of $75$, which corresponds to the average length of 
the input sequences. This resampling process is based on the Fourier Transform 
and is implemented using the SciPy Python module~\cite{scipy-paper}.
Prior to training, we normalize the skeleton sequences using $\min-\max$ normalization, 
which is applied separately to each dimension. Let $\mathcal{S}$ represent the dataset of skeleton 
sequences, organized into four dimensions: number of samples, sequence length, number 
of joints, and dimension of each joint. The normalization process is defined as follows:
\begin{equation}\label{equ:svae-normalization}
    \mathcal{S}[:,:,:,d] = \dfrac{\mathcal{S}[:,:,:,d] - \min(\mathcal{S}[:,:,:,d])}{\max(\mathcal{S}[:,:,:,d]) - \min(\mathcal{S}[:,:,:,d])},
\end{equation}
In scenarios involving a train/test split, the test set is normalized using the 
$\min$ and $\max$ values derived from each dimension of the training set. This ensures 
consistency in the data preprocessing steps and enhances the model's performance 
by standardizing the input data.

\subsubsection{Model Architecture}

For the Encoder and Decoder architecture in our model (Figure~\ref{fig:svae-detailed}), 
we have employed the following configurations:

\begin{itemize}
    \item Each convolutional layer in the network uses $128$ filters.
    \item The kernel sizes for the filters are $40$, $20$, and $10$ for the three layers
    in the Encoder, respectively, and these sizes are reversed in the Decoder.
    \item The dimension of the latent space is set to $16$.
\end{itemize}

The classifier architecture within the SVAE model comprises an FC layer 
with $8$ units, followed by a $50\%$ dropout, and then 
two fully connected layers with $C$ units each, where $C$ represents 
the number of classes. The generative models are trained for $2000$ epochs
with a batch size of $32$, using the 
Adam optimizer with a learning rate decay. We monitor the training 
loss to select the best model for evaluation.

\subsubsection{Evaluation Metrics}

For the evaluation metrics, we utilize the FID and APD fidelity and diversity 
metrics, detailed in the previous Section~\ref{sec:evaluation-wsdba}.
For feature extraction during evaluation, we adhere to the methodology in~\cite{action2motion-paper},
employing a GRU-based classifier.

\subsection{Experimental Results}

In this section, we present the experimental results on the HumanAct12 dataset. 
We start with a parameter search to determine the optimal weights for the different 
losses in SVAE while comparing its performance to the standard VAE model and the CVAE.
Next, we compare our SVAE model with state-of-the-art models. 
Finally, we conduct an experiment on label distribution balancing using data generation.

\subsubsection{Weight Losses Parameter Search}

\begin{table}
\centering
\caption{
The best loss function weights for the VAE, CVAE and SVAE models are as follows. For the VAE and CVAE, 
the classification loss weight ($W_{cls}$) is set to $0$ due to the absence of a 
classifier. The generated samples maintain the same label distribution as the 
training set but include more samples to better estimate the statistical metrics.
The best performing setup for each of the three variants is highlighted in \textbf{bold}.
}
\label{tab:svae-search}
\begin{tabular}{cccccc}
    \hline
    Model & W\_rec & W\_kl & W\_cls & FID & APD \\ 
    \hline \hline
    VAE   & 0.99900      & 1E-03     & 0.00000          & 01.38\textsuperscript{+-0.37}  & 06.42\textsuperscript{+-0.10} \\
    \textbf{VAE}   & \textbf{0.99990}     & \textbf{1E-04}    & \textbf{0.00000}          & \textbf{01.20\textsuperscript{+-0.10}}    & \textbf{06.46\textsuperscript{+-0.10}}  \\
    VAE   & 1.00000          & 1E+00         & 0.00000          & 45.03\textsuperscript{+-1.90}  & 00.07\textsuperscript{+-0.07} \\
    VAE   & 0.99999    & 1E-05     & 0.00000          & 01.23\textsuperscript{+-0.09}  & 06.45\textsuperscript{+-0.10}  \\
    VAE   & 0.99000       & 1E-02      & 0.00000          & 01.80\textsuperscript{+-0.38}   & 06.39\textsuperscript{+-0.10}  \\ \hline
    CVAE   & 0.99900      & 1E-03     & 0.00000          & 00.55\textsuperscript{+-0.13}  & 06.62\textsuperscript{+-0.11} \\
    CVAE   & 0.99990     & 1E-04    & 0.00000          & 01.35\textsuperscript{+-0.22}    & 06.39\textsuperscript{+-0.13}  \\
    CVAE   & 1.00000          & 1E+00         & 0.00000          & 05.83\textsuperscript{+-1.95}  & 06.15\textsuperscript{+-0.20} \\
    CVAE   & 0.99999    & 1E-05     & 0.00000          & 04.65\textsuperscript{+-1.85}  & 05.84\textsuperscript{+-0.23}  \\
    \textbf{CVAE}   & \textbf{0.99000}       & \textbf{1E-02}      & \textbf{0.00000}          & \textbf{00.38\textsuperscript{+-0.14}}   & \textbf{06.71\textsuperscript{+-0.11}}  \\ \hline
    SVAE  & 0.70000        & 1E-04    & 0.29000       & 00.59\textsuperscript{+-0.11}  & 06.64\textsuperscript{+-0.10}  \\
    \textbf{SVAE} & \textbf{0.49950} & \textbf{1E-03} & \textbf{0.49950} & \textbf{00.53\textsuperscript{+-0.15}} & \textbf{06.65\textsuperscript{+-0.11}} \\
    SVAE          & 0.49995         & 1E-04         & 0.49995         & 00.72\textsuperscript{+-0.23}          & 06.62\textsuperscript{+-0.11}          \\
    SVAE  & 1.00000          & 1E+00         & 1.00000          & 45.56\textsuperscript{+-1.85} & 00.05\textsuperscript{+-0.07} \\
    SVAE  & 0.29000       & 1E-04    & 0.70000        & 00.64\textsuperscript{+-0.09}  & 06.64\textsuperscript{+-0.10}  \\
    SVAE  & 0.30000        & 1E-02     & 0.70000        & 00.86\textsuperscript{+-0.30}   & 06.62\textsuperscript{+-0.12} \\
    SVAE  & 1.00000          & 1E-04    & 1.00000          & 00.79\textsuperscript{+-0.15}  & 06.59\textsuperscript{+-0.11} \\
    SVAE  & 0.70000        & 1E-04    & 0.30000        & 00.80\textsuperscript{+-0.22}   & 06.61\textsuperscript{+-0.10}  \\
    SVAE  & 0.29000       & 1E-02      & 0.70000        & 00.73\textsuperscript{+-0.27}  & 06.63\textsuperscript{+-0.13} \\
    SVAE  & 0.30000        & 1E-04    & 0.70000        & 00.68\textsuperscript{+-0.13}  & 06.63\textsuperscript{+-0.10}  \\ \hline
\end{tabular}
\end{table}

To determine the optimal set of hyperparameters for the loss function, we 
performed an ablation study using a grid search. During this process, we 
tracked the FID and APD metrics to evaluate performance for the VAE, CVAE and SVAE models.
The results 
of this ablation study are detailed in Table~\ref{tab:svae-search}, with the best-performing model 
highlighted in bold.

In this study, the SVAE model achieved optimal performance with loss function 
weights of $0.4995$ for the reconstruction loss, $0.001$ for the KL loss, and $0.4995$ for 
the CE loss. Reflecting the principles of the $\beta$-VAE model, this configuration minimizes 
the weight of the KL loss while equally weighting the reconstruction and classification 
losses. This setup ensures that the SVAE model strikes a balance between achieving 
an optimal latent space representation, maintaining high reconstruction quality, 
and effectively separating classes in the latent space.

\begin{figure}
    \centering
    \caption{
        The 2D projection of the latent space (using PCA) for the VAE, CVAE and SVAE models 
        on the training samples shows distinct separation of latent points 
        corresponding to each action label in the case of SVAE compared to no separation in the cases of 
        VAE and CVAE.
    }
    \label{fig:svae-pca}
    \includegraphics[width=0.85\linewidth]{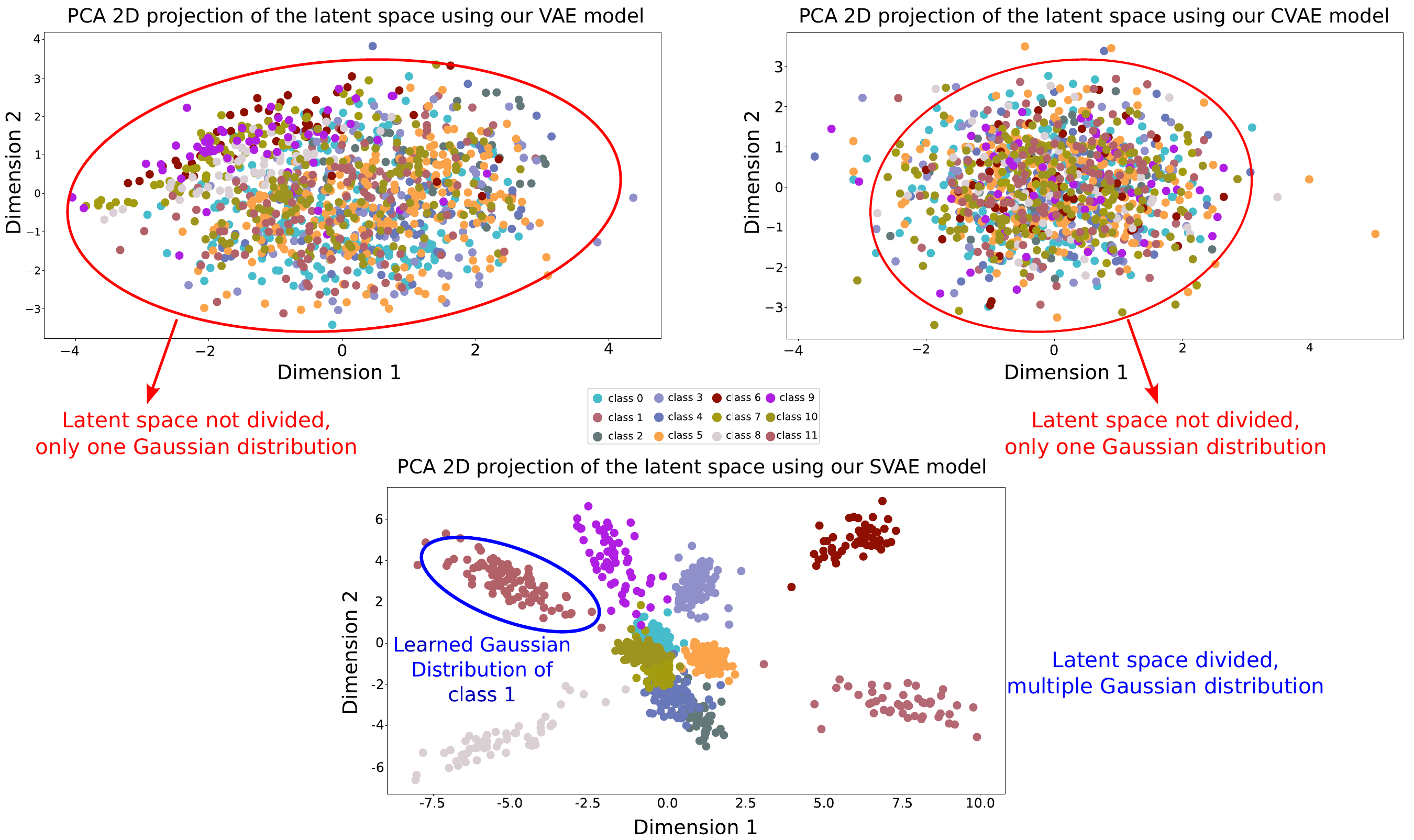}
\end{figure}

\begin{figure}
    \centering
    \caption{
        Generated samples from the proposed SVAE models are conditioned on four 
        different actions: \protect\mycolorbox{0,30,255,0.6}{Warm Up}, 
        \protect\mycolorbox{255,30,0,0.6}{Drink}, 
        \protect\mycolorbox{255,165,0,0.6}{Lift Dumbbell}
        and \protect\mycolorbox{0,128,0,0.6}{Sit}. For each action, 
        we present two distinct generated examples to showcase the diversity and 
        robustness of the SVAE model.
    }
    \label{fig:svae-generations}
    \includegraphics[width=\textwidth]{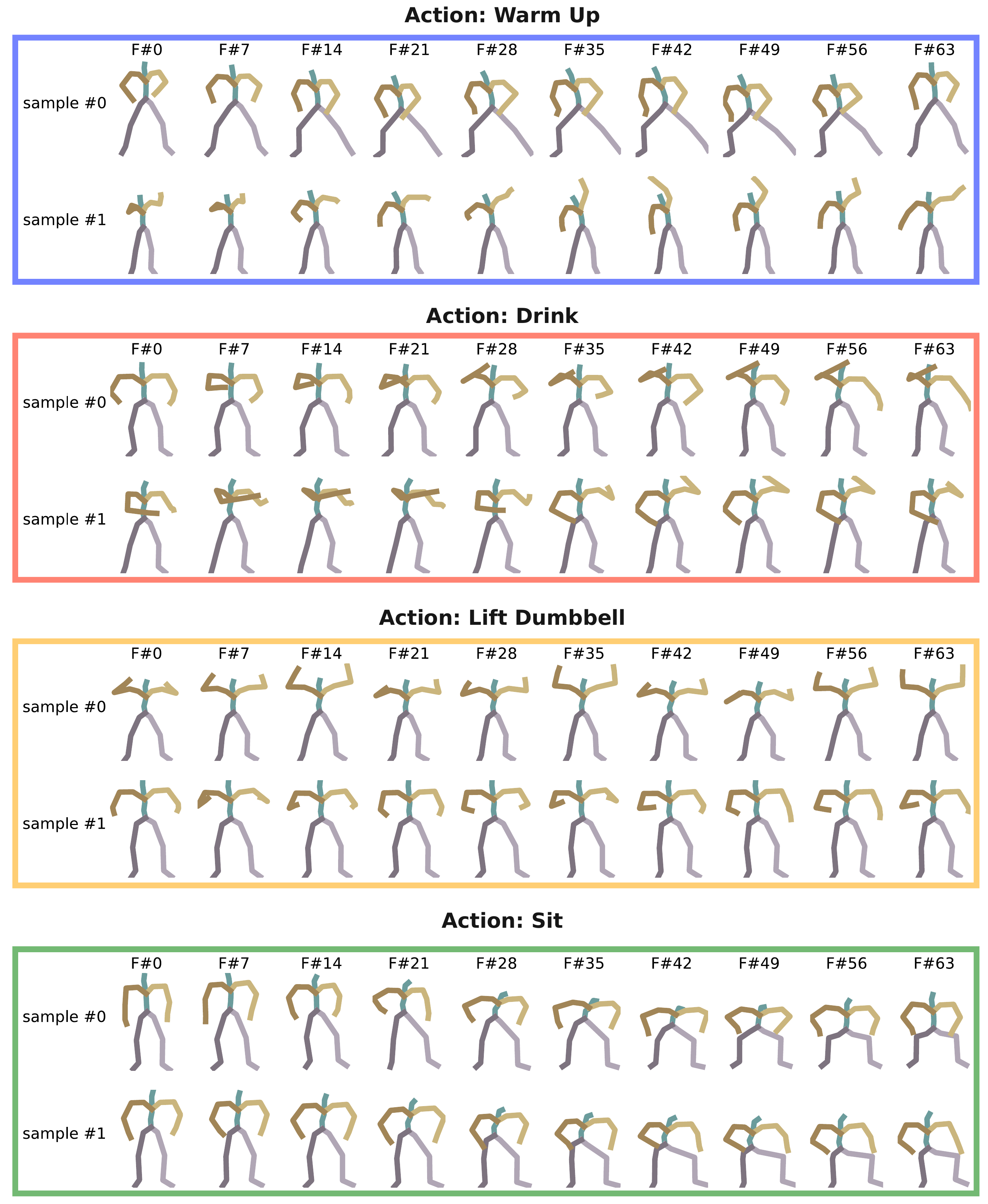}
\end{figure}

The SVAE model outperforms the traditional VAE model, though it does not surpass the best setup 
of the CVAE model. However, the SVAE model is less sensitive to changes in the loss parameters. 
This stability is due to the explicit training setup, which ensures that the action labels 
effectively distinguish between samples in the latent space.
In order to highlight such ability, a 2D visualization of the 
training samples' latent space, created using Principal Component Analysis (PCA), 
is shown in Figure~\ref{fig:svae-pca} for the three models VAE, CVAE and SVAE.
This visualization highlights the SVAE model's ability to 
clearly separate multiple classes within the latent space, providing better control 
over generation conditioned on the label, which is not provided by both the VAE and the CVAE models.
A visualization of several generated samples is shown in Figure~\ref{fig:svae-generations},
demonstrating that the quality of the generated sequences appears visually satisfactory.
However, visual inspection alone is not sufficient to fully assess the model's
performance. Therefore, quantitative evaluations are necessary to objectively measure
the fidelity and diversity of the generated samples.

\subsubsection{Comparison With State-Of-The-Art}

\begin{table}
\centering
\caption{
        We compared our proposed model, the proposed SVAE,
with other approaches from the literature using the FID and Diversity metrics. 
For each method, we reported the average $\pm$ standard deviation of the FID and 
Diversity scores. For the real samples, a random split into two sets was used 
to calculate the FID metric. This comparison allows us to assess the effectiveness 
of our SVAE model in generating diverse and high-fidelity samples relative to existing methods.
}
\label{tab:svae-sota}
\begin{tabular}{cccccc}
    \hline
    Model         &  &  & FID $\downarrow$                             &  & Diversity $\uparrow$                  \\ 
    \hline \hline
    Real          &  &  & 00.030\textsuperscript{+-0.005} &  & 06.860\textsuperscript{+-0.070}    \\ 
    \hline
    Two-stage GAN~\cite{action2motion-paper} &  &  & 10.480\textsuperscript{+-0.089}  &  & 05.960\textsuperscript{+-0.049}   \\
    Act-MoCoGAN~\cite{action2motion-paper}   &  &  & 05.610\textsuperscript{+-0.113}  &  & 06.752\textsuperscript{+-0.071}   \\
    Action2Motion~\cite{action2motion-paper} &  &  & 02.458\textsuperscript{+-0.079}  &  & 07.032\textsuperscript{+-0.038}   \\
    ACTOR~\cite{actor-paper}         &  &  & 00.120\textsuperscript{+-0.000}    &  & 06.840\textsuperscript{+-0.030}     \\
    PoseGPT~\cite{posegpt-paper}&  &  & 00.080\textsuperscript{+-*.***}                            &  & 06.850\textsuperscript{+-*.***}                             \\
    UM-CVAE~\cite{um-cvae-paper}       &  &  & 00.090\textsuperscript{+-0.000}     &  & 06.810\textsuperscript{+-0.020}       \\
    MotionDiffuse~\cite{motion-diffuse-paper} &  &  & 00.070\textsuperscript{+-0.000}    &  & 06.850\textsuperscript{+-0.020}     \\ 
    \hline
    SVAE (\textbf{ours})   &  &  & 00.560\textsuperscript{+-0.170}
      &  & 06.640\textsuperscript{+-0.100} \\ 
    \hline
\end{tabular}
\end{table}

To evaluate our proposed SVAE architecture against state-of-the-art models 
on the HumanAct12 dataset, we present the results in Table~\ref{tab:svae-sota}, focusing on 
our best SVAE model. To ensure fair metric evaluation, the generated samples 
maintain the same label distribution as the training set, for instance if the 
real data contains $10$ samples of class 1 and $5$ samples of class 2, 
then the generated data should follow the same label distribution.
This is followed by extracting the features of both real and generated samples using the GRU classifier,
i.e. extracting the output of the layer before the classification one.
The results reveal 
that our SVAE model surpasses the Action2Motion model in fidelity (FID) and 
performs closely to the leading state-of-the-art models. Moreover, the 
diversity of samples generated by the SVAE is similar to that of the real 
samples, indicating that the model effectively preserves diversity while 
maintaining high fidelity.

The results demonstrate that the SVAE model can generate high fidelity 
human motion sequences, as illustrated by the examples in Figure~\ref{fig:svae-generations}. 
This indicates that the convolution filters effectively extract temporal 
features from the input sequences. Consequently, this confirms that human 
motion skeleton sequences can be accurately represented as Multivariate Time Series (MTS).

\subsubsection{Fixing Labels Distribution With SVAE's Generations}

\begin{table}
\centering
\caption{Test accuracy values are reported for four different train/test splits: 
(first row) training on real samples; (second row) training on generated samples 
(following the same distribution as the real samples); and (third row) training 
on augmented samples (a combination of real and generated samples to achieve a
uniform label distribution).}
\label{tab:svae-accuracy}
\begin{tabular}{c|cccc}
    \hline
    Train On & Split 0 & Split 1 & Split 2 & Split 3 \\ 
    \hline \hline
    real & \textbf{87.77\textsuperscript{+-1.55}} & \textbf{76.47\textsuperscript{+-2.23}} & 87.78\textsuperscript{+-1.40} & 77.43\textsuperscript{+-2.54} \\
    generated & 72.47\textsuperscript{+-2.91} & 64.39\textsuperscript{+-2.42} & 77.37\textsuperscript{+-4.32} & 69.75\textsuperscript{+-7.31} \\
    augmented & 86.07\textsuperscript{+-2.10} & 73.51\textsuperscript{+-1.32} & \textbf{89.34\textsuperscript{+-1.77}} & \textbf{78.71\textsuperscript{+-2.53}} \\ 
    \hline
\end{tabular}
\end{table}

As previously discussed, utilizing data extension with deep generative models 
can present a paradoxical challenge. In this section, we aim to address the impact 
of the class balancing problem. By examining the effects of balancing class distributions 
in the training data, we seek to understand how this approach influences model 
performance and the quality of generated samples.
We implemented our method by balancing the label distribution in the training set,
after applying four different train/test splits with a cross subject setup.
Specifically, for each training set per split, we generated new human motion sequences using 
our SVAE model to augment each class, aiming for a uniform label distribution while 
keeping the most populated class unchanged. Additionally, we trained the model on 
generated samples that maintained the same number of samples and label distribution 
as the training set. The results, presented in Table~\ref{tab:svae-accuracy},
indicate that the performance 
is comparable to training on real samples, showing no significant differences. 
This demonstrates that the generated samples possess sufficient quality. Furthermore, 
using data generation to balance the label distribution enhances the model's performance.

\begin{figure}
    \centering
    \caption{
        The confusion matrix compares performance when training on real samples (left) 
        versus augmented samples with a fixed label distribution (right). 
        The matrix consists of $12$ rows and $12$ columns, corresponding to the 
        $12$ action labels. Each row represents the actual class of the samples 
        used for prediction, and each class is ranked in descending order based 
        on their population in the test set. The results show that when the label 
        distribution is fixed (right), the test set performance is significantly 
        affected in classes $4$, $7$, and $11$, which have higher ranks.
    }
    \label{fig:svae-confusion}
    \includegraphics[width=\textwidth]{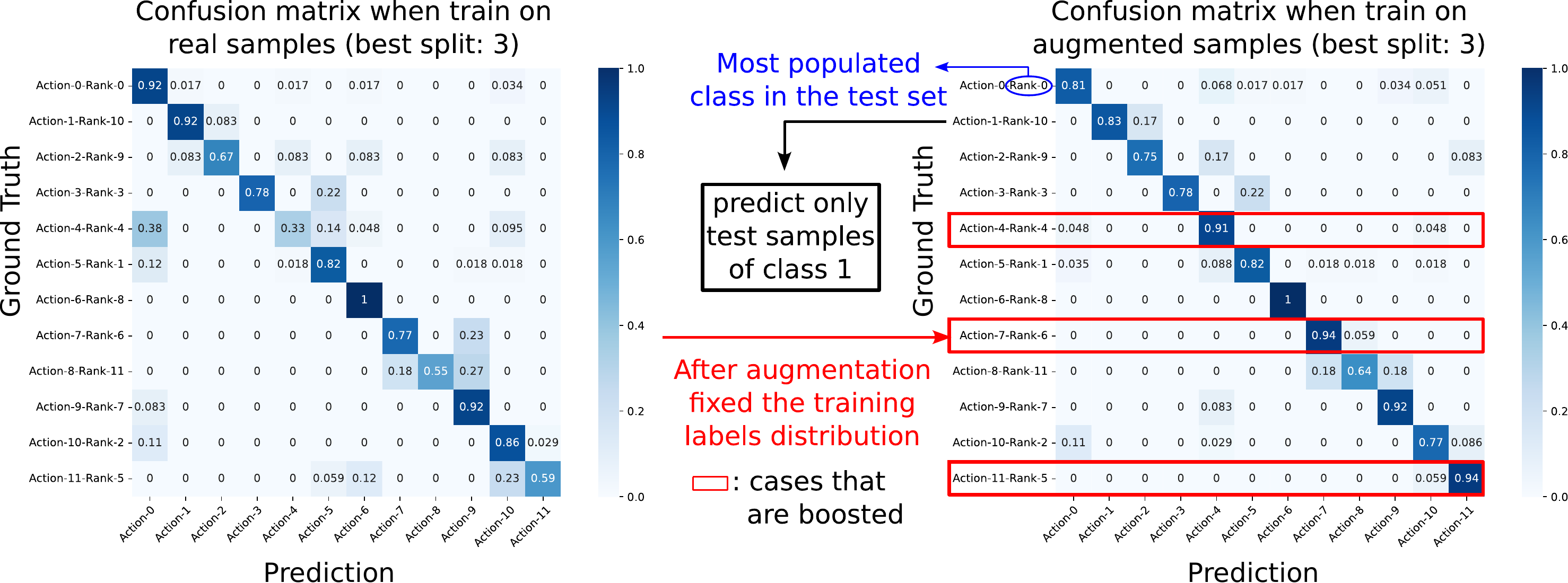}
\end{figure}

To further examine the fourth split and understand the impact on each action class, 
we present a confusion matrix in Figure~\ref{fig:svae-confusion}.
The row ticks include the rank of each 
action class, which represents the population of each class in the test set, listed 
in descending order. The confusion matrix reveals that our data generation for label 
distribution balancing 
technique effectively boosted the performance of some intermediate and low-rank 
action classes. This demonstrates that the quality of the generated samples is 
sufficient to enhance the performance of underrepresented classes.

\section{Conclusion}

In this chapter, we explored several advanced methodologies for improving 
human motion analysis represented as time series data.

Firstly, we discussed the usage of LITEMVTime, a powerful model designed for human 
rehabilitation assessments. LITEMVTime not only enhances performance but also offers 
explainability, making it highly suitable for clinical applications where understanding 
model decisions is crucial.

Next, we introduced ShapeDBA, a novel prototyping method. ShapeDBA was used in 
a weighted setup to perform data extension for human rehabilitation, effectively 
boosting regression models. This approach demonstrated significant improvements 
in handling sparse training data, enhancing the overall quality and robustness of 
the rehabilitation assessments.

Additionally, we proposed a deep generative model, the Supervised Variational 
Auto-Encoder (SVAE), for human motion generation. This CNN-based model 
competes well with state-of-the-art models. The SVAE model integrates 
classification tasks within the latent space, achieving a balance between 
reconstruction quality and class separation, ultimately generating high fidelity 
and diverse human motion sequences.

Throughout this chapter and all previous chapters, we have extensively discussed discriminative tasks 
such as classification, regression, and clustering, noting that the evaluation 
methods for these tasks are now well-established and widely adopted by the community. 
In contrast, the evaluation of generative models is more complex and diverse, 
with metrics like FID and APD being commonly used, but many new metrics being 
proposed each year. It is common in generative model studies, especially in 
applications like human motion generation, to use fewer than five datasets for 
evaluation. This limitation makes it impractical to use techniques such as the 
Critical Difference Diagram and the Multi-Comparison Matrix. 
Currently, each paper often proposes a new metric with different settings, leading 
to inconsistent and unfair comparisons. This inconsistency is due to the lack of a 
standardized framework, which the next chapter aims to address.
The next chapter 
addresses the need to unify the evaluation metrics for human motion generation.
\chapter{Evaluation Metrics For Human Motion Generation}\label{chapitre_7}

\section{Introduction}

Evaluating generative models presents unique challenges~\cite{reliable-fidelity-diversity}
that are not as prevalent in discriminative models, where comparisons to 
ground truth data are straightforward. For generative models, 
the evaluation involves measuring the validity of generated samples 
against real ones. Traditional human judgment metrics, such as 
Mean Opinion Scores (MOS)~\cite{mos-paper}, often fall short as they assume a 
uniform user perception of ideal generation, which is unrealistic. 
Therefore, quantitative evaluation is essential, focusing on two 
key dimensions: fidelity and diversity. Fidelity assesses the 
similarity between the distributions of real and generated data, 
while diversity measures the variety within the generated samples, 
ensuring they reflect the range present in real data sets.

In the previous chapter, we reviewed existing works on human motion generation, 
which typically use standard evaluation methods and sometimes introduce new metrics. 
The main challenge is defining fidelity and diversity metrics, as no single optimal 
solution exists, \textit{No Metric To Rule Them All}.
Consequently, numerous approaches and novel metrics have been developed 
to address this issue. However, evaluating these metrics alone is insufficient, 
and inconsistencies in generation setups and frameworks post-training across 
different studies make model comparisons problematic. This complexity underscores 
the need for a unified and detailed evaluation framework.

A crucial aspect of evaluating human motion data is its temporal dependency~\cite{ucr-archive}.
Temporal 
distortion, which includes time shifts, frequency changes, and warping, is vital for 
assessing multivariate time series~\cite{uea-archive}
such as human motion sequences. Existing metrics often 
overlook this, focusing instead on latent representations. To address this gap, we 
introduce a novel metric called Warping Path Diversity (WPD). WPD measures the diversity 
of temporal distortions in both real and generated data, scoring models based on their 
ability to produce varied temporal sequences, thus ensuring a more comprehensive evaluation.

Unifying the evaluation process is essential. This work consolidates evaluation metrics 
from the literature into a unified framework for fair comparisons,
providing a helpful 
resource for newcomers. Our experiments highlight the difficulty of identifying a 
universally superior model, as small changes in architecture and hyperparameters 
can significantly impact metric values. We conduct detailed experiments with three 
CVAE model variants on the same dataset, offering an in-depth analysis of each metric.

\begin{figure}
    \centering
    \caption{
        The evaluation metrics for human motion generation in this work are 
        divided into two groups: \protect\mycolorbox{0,117,185,0.64}{fidelity metrics}
        and \protect\mycolorbox{0,150,59,0.64}{diversity metrics}.
        These metrics are further categorized based on their evaluation 
        criteria, such as FID being a distribution-based metric.
    }
    \label{fig:metrics-summary}
    \includegraphics[width=0.7\textwidth]{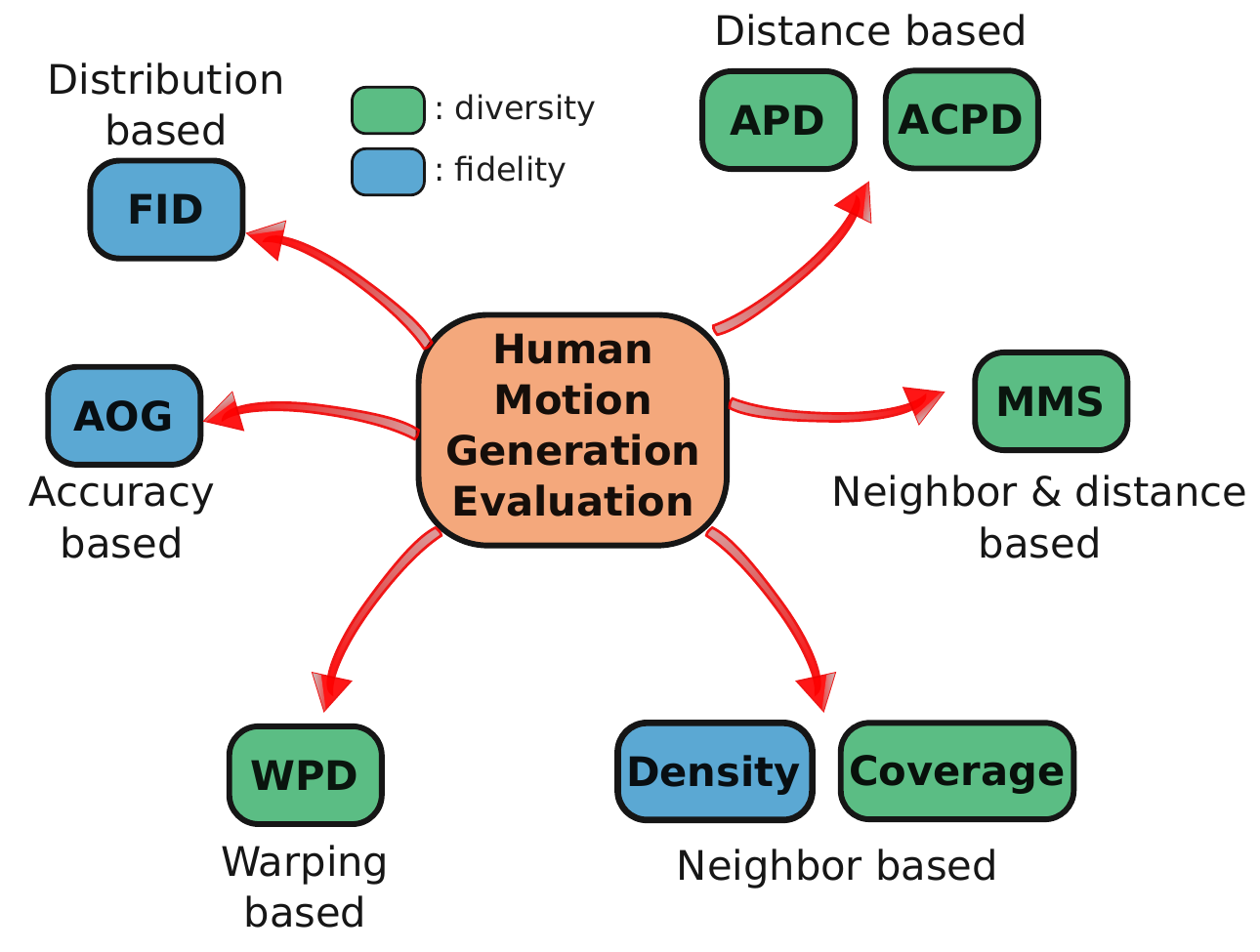}
\end{figure}

Figure~\ref{fig:metrics-summary} presents a brief summary 
of all the metrics used in this work, categorized 
into fidelity and diversity. The metrics are further organized into 
sub-categories based on their evaluation approach: accuracy-based, distribution-based, 
distance-based, neighbor-based, neighbor/distance-based, and warping-based.

In this chapter, we propose a clear, user-friendly evaluation framework for 
newcomers to the field. By establishing 
standardized practices, we aim to facilitate more consistent and meaningful 
comparisons of generative models, ultimately contributing to the advancement 
of human motion generation research.

\section{Generative Models Metrics}

The evaluation metrics for generative models are categorized into fidelity and diversity. 
Fidelity metrics evaluate how well generated samples mirror the real distribution, 
making it harder to distinguish between real and generated data, thus ensuring 
reliability. Diversity metrics assess the variation among generated samples, 
indicating the model's ability to produce a wide range of outputs. Higher diversity 
means the model isn't limited to a narrow segment of the real distribution. 

This section first defines key concepts, then thoroughly reviews the main metrics 
used for assessing both fidelity and diversity.

\subsection{Definitions}

To understand the metrics that follow, it's necessary to establish some definitions:

\mydefinition Real set of samples: a set of $N$ real samples is
referred to as $\mathcal{X}~=~\{\textbf{x}_i\}_{i=1}^{N}$ and follows a distribution \pr;
where $\textbf{x}_i$ is an MTS of length $L$ and dimensions $M=JxD$, with $J$ being the number 
of skeleton joints and $D$ the dimension of each joint.

\mydefinition Generated set of samples: a set of $G$ generated samples is referred to
as $\hat{\mathcal{X}}~=~\{\hat{\textbf{x}}_j\}_{j=1}^{G}$ and follows a distribution \pg.

\mydefinition A pre-trained deep learning model $\mathcal{G}\circ\mathcal{F}(.)$ 
is made of a feature extractor $\mathcal{F}$ and a last layer $\mathcal{G}$ 
achieving the desired task (e.g. classification).

\begin{figure}
    \centering
    \caption{
        Before calculating evaluation measures, two steps are followed. First, a 
        \protect\mycolorbox{0,194,200,0.47}{model} is trained on a 
        \protect\mycolorbox{156,120,255,0.6}{supervised task} using only
        \protect\mycolorbox{0,50,255,0.6}{real data} and not the 
        \protect\mycolorbox{255,165,0,0.6}{generated data}. Second, the 
        pre-trained encoder's \protect\mycolorbox{0,125,0,0.6}{latent representation of the
        real data}
        is extracted, as well as the \protect\mycolorbox{255,30,0,0.5}{latent 
        representation of the generated samples}.
        The metrics are then 
        computed based on this latent representation.
    }
    \label{fig:metrics-latent-space}
    \includegraphics[width=\textwidth]{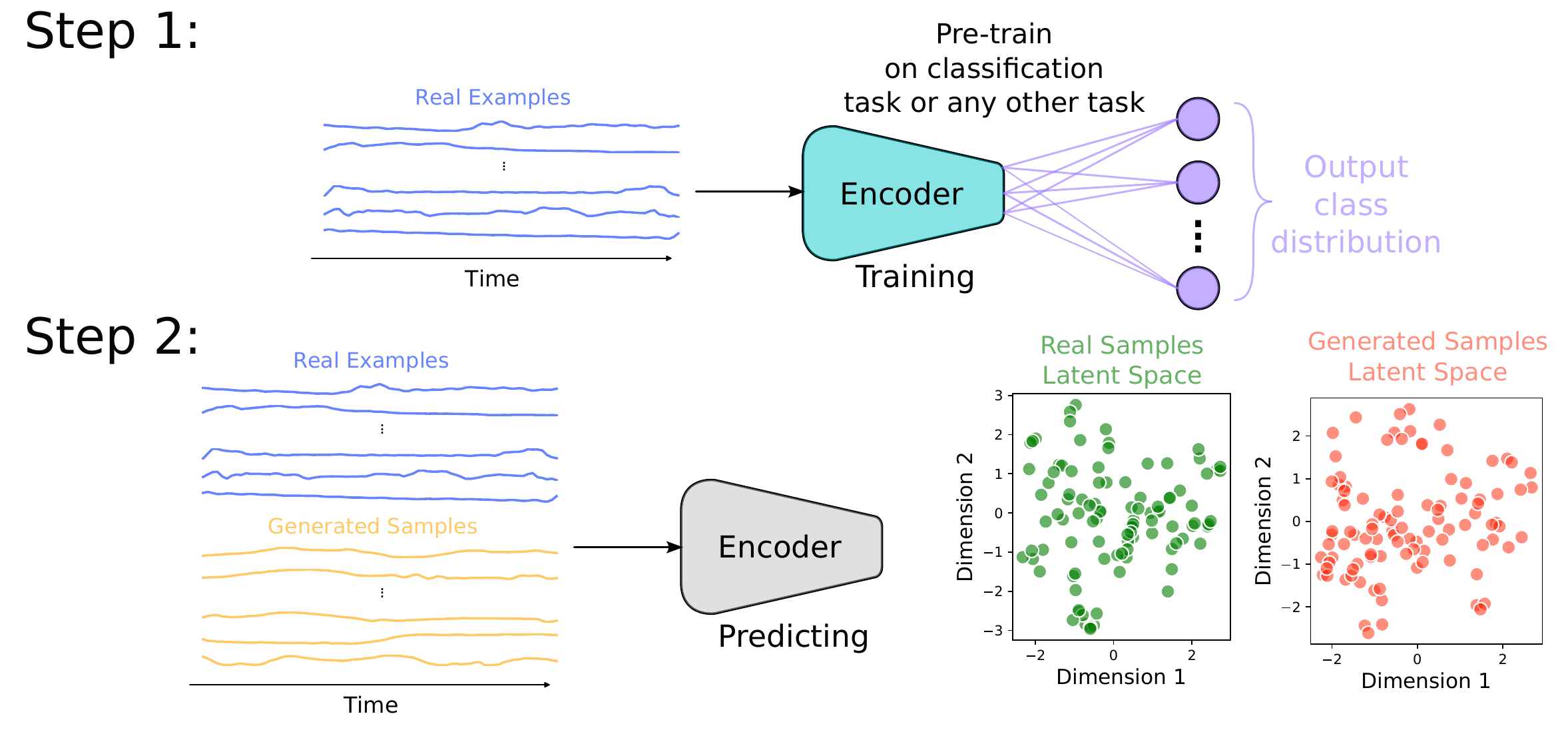}
\end{figure}

To compute most metrics, we first train a deep learning model $\mathcal{G}\circ\mathcal{F}$
on a specific 
task, typically classification, using real data. Here, $\mathcal{G}$ is a $softmax$ layer. 
The feature extractor is then used to project both real data $\mathcal{X}$ and generated 
data $\hat{\mathcal{X}}$ into a latent space, enabling metric calculations within this space. 
This process involves two steps, as illustrated in Figure~\ref{fig:metrics-latent-space}:
training the model on real data and using the 
feature extractor (excluding the final layer) to encode both real and generated 
samples into latent spaces $\textbf{V}$ and $\hat{\textbf{V}}$, respectively, such as:
\begin{equation}\label{equ:metrics-latent-space}
    \textbf{V} = \mathcal{F}(\mathcal{X}) \;\; \text{and} \;\; \hat{\textbf{V}} = \mathcal{F}(\hat{\mathcal{X}})
\end{equation}
\noindent where both $\textbf{V}$ and $\hat{\textbf{V}}$ are two-dimensional
matrices, corresponding to the number of examples in $\mathcal{X}$ and $\hat{\mathcal{X}}$
respectively, with each dimension representing features $f$.

For each metric discussed in this section, we compute two versions: one using 
generated samples (and real samples, as applicable) and one using only real 
samples. This approach provides a reference metric value for \pr. We achieve 
this by randomly splitting $\textbf{V}$ into two subsets, $\textbf{V}_1$ and 
$\textbf{V}_2$. Metrics are then calculated by treating $\textbf{V}_1$ as the 
latent space for real samples and $\textbf{V}_2$ as the latent space for generated samples.

\subsection{Fidelity Metrics}

The fidelity metrics in the literature are either
distribution based, neighbor based or accuracy based.

\subsubsection{Fréchet Inception Distance (FID)}\label{sec:metrics-fid}

In the previous chapter, we utilized the Fréchet Inception Distance (FID) due to 
its widespread use in evaluating generative models. In this section, we present 
the history and background of this metric.
Introduced by~\cite{fid-original-paper}, the Fréchet Inception Distance (FID) 
is a popular metric for evaluating generative models. It builds on the Inception Score 
(IS)~\cite{is-paper}, which assesses generated samples using
a pre-trained Inception model, indirectly 
considering the real distribution \pr. Unlike IS, FID quantifies the difference 
between the real distribution \pr~and the generated distribution \pg. It does 
this by calculating the Fréchet Distance~\cite{fd-paper} between two Gaussian distributions in the 
Inception model's latent space of both $\mathcal{X}$ and $\hat{\mathcal{X}}$.

The Fréchet Distance (FD)~\cite{fd-paper} measures the similarity between two continuous curves. 
To understand what FD
measures, a famous example goes as follows: \textit{Imagine a person and their dog, each 
wanting to traverse a different finite curved
path. The speed of the person and the dog can vary but they are not 
allowed to go backward on the path. The FD between these
two curves is the length of a leash, small enough so that both the person and the 
dog can traverse the whole finite curve}. For probability distributions, FD is 
calculated between their 
Cumulative Distribution Functions (CDFs)~\cite{fd-dist-paper,}. For multidimensional Gaussian
distributions~\cite{fid-gaus-paper} $\mathcal{P}_1 \sim \mathcal{N}(\mu_1,\Sigma_1)$ 
and $\mathcal{P}_2\sim\mathcal{N}(\mu_2,\Sigma_2)$, both of dimension $f$, the FD 
is calculated as follows:
\begin{equation}\label{equ:metrics-fid}
    FD(\mathcal{P}_1,\mathcal{P}_2)^2 = \textit{trace}(\Sigma_1+\Sigma_2-2(\Sigma_1.\Sigma_2)^{1/2}) + \sum_{i=1}^{f}(\mu_{1,i}-\mu_{2,i})^2 
\end{equation}
\noindent where the values of $FD$ (or $FID$) range from $0$ to $+\infty$.

\paragraph*{Setup for generative models}
First, we empirically estimate the mean vectors $\boldsymbol{\mu}$ and $\hat{\boldsymbol{\mu}}$
for both $\textbf{V}$ and $\hat{\textbf{V}}$, along with their covariance
matrices $\Sigma$ and $\hat{\Sigma}$.
Second, we compute the Fréchet Distance (FD) using Eq.~\ref{equ:metrics-fid}.
For consistency with the literature, we refer to this metric as FID throughout this 
work, even though the Inception network is not used for human motion.

\begin{figure}
    \centering
    \caption{
        On the left, we illustrate the energy (FID) needed to transform a 
        \protect\mycolorbox{0,100,255,0.6}{standard 
        Gaussian distribution} into another \protect\mycolorbox{255,30,0,0.6}{Gaussian
        distribution with a 
        higher mean and variance}. On the right, the plot shows that as 
        the mean ($\mu$) and variance ($\sigma^2$) of the target distribution increase, 
        the required transformation energy (FID) also increases progressively.
    }
    \label{fig:metrics-fid}
    \includegraphics[width=\textwidth]{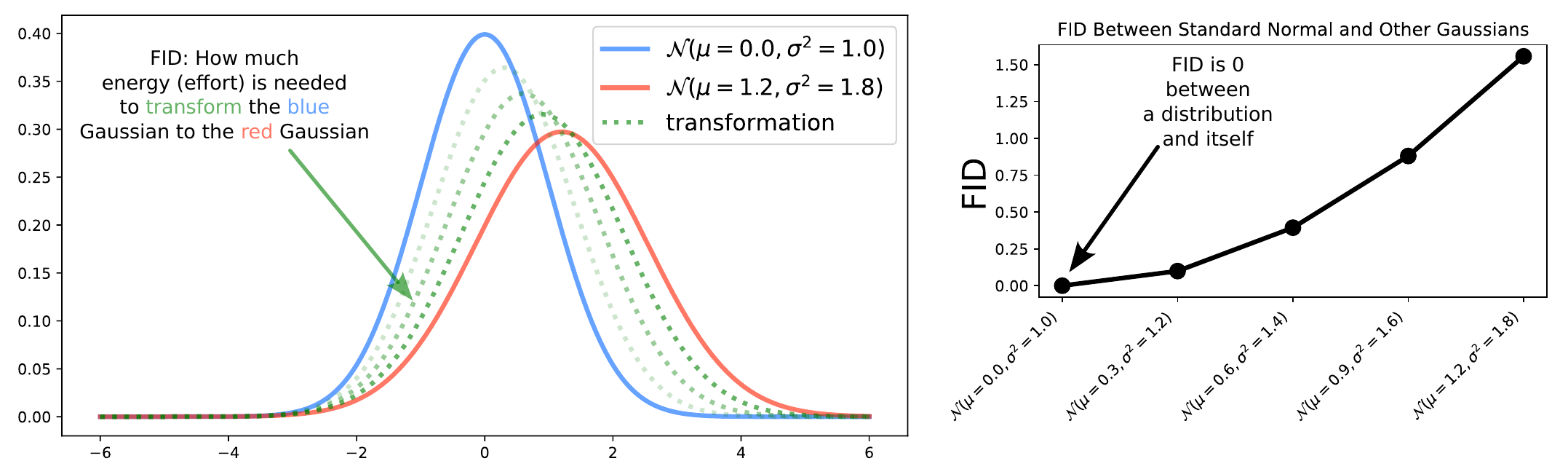}
\end{figure}

\paragraph*{Interpretation}
The FID represents the amount of energy or effort needed to transform one Gaussian 
distribution into another. Figure~\ref{fig:metrics-fid} illustrates this concept, showing the 
probability density functions of two Gaussian distributions. The energy 
required to change  $\mathcal{N}(\mu=0.0, \sigma^2=1.0)$ to $\mathcal{N}(\mu=1.2, \sigma^2=1.8)$
increases with the mean and variance differences. Since no energy is needed to 
transform a distribution into itself, the starting point on the plot 
(right side of Figure~\ref{fig:metrics-fid}) is $0.0$. This aligns with the
FID's definition as a distance metric.

Many studies claim that a lower FID indicates higher fidelity in generated 
samples. However, this can be misleading. For instance, if a generative model 
merely replicates real samples, it would achieve a perfect FID of 0, showing 
no new value. Therefore, to accurately compare two generative models using the 
FID metric, we propose following this principle:
\begin{theorem}[Fréchet Inception Distance Interpretation]\label{the:metrics-fid}
    A generative model $Gen_1$ is considered more fidelitous than another 
    model $Gen_2$ on the $FID$ metric if $FID_{gen1} < FID_{gen2}$ while 
    respecting the following constraint:
    \begin{equation*} 
        \forall~\epsilon>0, \hspace{1cm} FID_{gen} = FID_{real} + \epsilon
    \end{equation*}
\end{theorem}

\subsubsection{Accuracy On Generated (AOG)}\label{sec:metrics-aog}

Generative models can incorporate relevant characteristics of each sample for better 
control. For labeled datasets, this might include discrete labels, continuous values, 
or text descriptions. This conditioning, as mentioned in Chapter~\ref{chapitre_6}
enhances the precision of generated outputs. 
For example, a model generating human motion sequences can be conditioned to produce 
specific actions like ``running'' or ``jumping'' ensuring alignment with the desired activity.

To evaluate the conditioning capability of a generative model, we can use the score 
of a classifier pre-trained on real samples $\mathcal{X}$,
treating the generated set $\hat{\mathcal{X}}$ as unseen data.
The classifier, $\mathcal{G}\circ\mathcal{F}$,
helps measure this capability. We refer to this metric as Accuracy On Generated (AOG), adapted from the Accuracy 
metric of~\cite{action2motion-paper}, 
formulated as follows:
\begin{equation}\label{equ:metrics-aog}
    AOG(\hat{\mathcal{X}},~\hat{Y},~\mathcal{G}\circ\mathcal{F}) = \dfrac{1}{G}\sum_{i=1}^{G}\mathds{1}\{\mathcal{G}\circ\mathcal{F}(\hat{\mathcal{X}}_i) == \hat{Y}_i\}
\end{equation}
\noindent where $\hat{Y}$ represents the set of labels employed to condition 
the generation process, serving as ground truth labels that 
$\mathcal{G}\circ\mathcal{F}$ is expected to predict.
Additionally, $\mathds{1}$ denotes the indicator function defined as:
\begin{equation}
    \mathds{1}\{condition\}=
    \begin{cases}
        1 & if\text{ }condition\text{ }is\text{ }True\\
        0 & if\text{ }condition\text{ }is\text{ }False
    \end{cases}
\end{equation}
\noindent where the values of AOG in Eq.~\ref{equ:metrics-aog} range from $0$ to $1$.

\paragraph*{Setup for generative models}
The AOG metric is calculated by comparing the ground truth labels $\hat{Y}$
with the predictions made by $\mathcal{G}\circ\mathcal{F}$ using Eq.~\ref{equ:metrics-aog}.

\begin{figure}
    \centering
    \caption{
        This example demonstrates the AOG metric for two generative models.
        \protect\mycolorbox{0,175,105,0.53}{$Model1$} achieves a perfect AOG of $100\%$
        by accurately generating samples conditioned on three classes 
        \protect\mycolorbox{255,50,0,0.6}{red triangles},
        \protect\mycolorbox{0,100,255,0.6}{blue squares} and 
        \protect\mycolorbox{0,140,0,0.6}{green circles}. 
        In contrast, \protect\mycolorbox{176,175,105,0.53}{$Model2$} only achieves $50\%$,
        indicating it struggles with correct conditional generation. The AOG metric
        reflects the classification accuracy of generated samples compared to ground
        truth labels $\hat{Y}$. 
    }
    \label{fig:metrics-aog}
    \includegraphics[width=\textwidth]{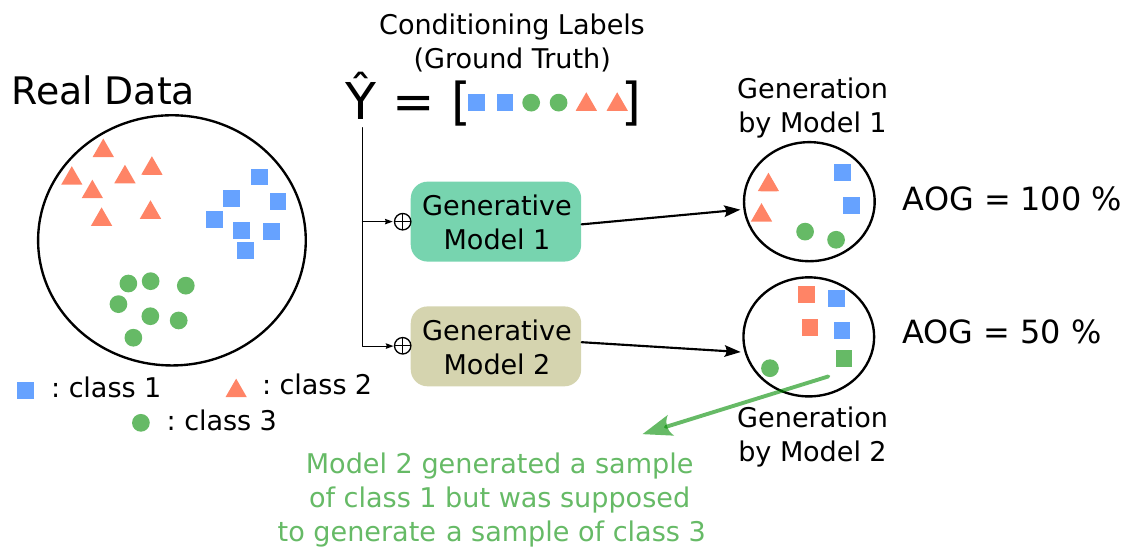}
\end{figure}

\paragraph*{Interpretation}
The AOG metric is a strong indicator of a generative model's conditioning capability, 
but it must be interpreted carefully. A very low AOG might suggest the model 
generates samples from a narrow set of labels. Conversely, a perfect AOG score 
doesn't necessarily mean high fidelity to \pr; the generated samples could contain 
residual noise that doesn't affect the classifier's performance. Therefore, a high 
AOG might give a false impression of generation quality. Figure~\ref{fig:metrics-aog}
demonstrates this nuanced behavior of the AOG metric.

\subsubsection{Density and Precision}\label{sec:metrics-density}

Using a single metric to evaluate generative models is often insufficient due to 
the complexity of assessing both fidelity and diversity.~\cite{precision-recall-paper}
highlighted that two models might share the same FID score but differ in these 
qualities. To address this,~\cite{precision-recall-paper}
introduced precision and recall metrics to 
evaluate fidelity and diversity separately.~\cite{improved-precision-recall-paper} refined 
these metrics by incorporating the $k$-nearest neighbor algorithm to better 
estimate density functions. In particular, precision represents the portion 
of samples generated by \pg~that can be sampled from \pr~as well., and 
its improved formulation (referred to as precision in the rest of this work) is as follows:
\begin{equation}\label{equ:metrics-precision}
    precision(\textbf{V},\hat{\textbf{V}},k) = \frac{1}{G}\sum_{j=1}^{G}\mathds{1}(\hat{\textbf{V}}_j\in manifold(\textbf{V}_1,\ldots,\textbf{V}_N))
\end{equation}
\noindent where $manifold(\{a_1,a_2,\ldots,a_n\}) = \bigcup_{i=1}^{n}B(a_i,NND_k(a_i))$, 
$B(c,r)$ is a sphere in $\mathds{R}^{dimension(a_i)}$ of center $c$ and radius $r$, 
and $NND_k(a_i)$ is the distance from $a_i$ to its $k_{th}$ nearest neighbor in the 
set $\{a_j\}_{j=1,j\neq i}^{N}$. The values of precision range from $0$ to $1$.

\cite{reliable-fidelity-diversity} identified key limitations in the precision and 
recall metrics for evaluating generative models. They proposed density and coverage 
as new metrics for fidelity and diversity, respectively. This section addresses fidelity, 
with diversity discussed in Section~\ref{sec:metrics-diversity}.
The precision metric has two main issues: 
(1) no closed formulation for the expected precision when real and generated samples 
follow the same distribution (i.e. \pr~and \pg~are identical), and
(2) outliers in the real data $\mathcal{X}$ can produce misleading 
precision values, suggesting good performance even when it is not accurate.

The mathematical formulation of the density metric is as follows:
\begin{equation}\label{equ:metrics-density}
    density(\textbf{V},\hat{\textbf{V}},k) = \frac{1}{k.G}\sum_{j=1}^{G}\sum_{i=1}^{N}\mathds{1}(\hat{\textbf{V}}_j\in B(\textbf{V}_i,NND_k(\textbf{V}_i)))
\end{equation}
\noindent where the values of $density$ range from $0$ to $\dfrac{N}{k}$.

\cite{reliable-fidelity-diversity} showed that when real and generated distributions (\pr~and~\pg) 
are identical, the expected value of density is $1$. This means that with enough samples 
and a high number of neighbors, the density metric converges to $1$, accurately 
reflecting the fidelity of generated samples. Additionally, density has an advantage 
over precision by detecting outliers in the real distribution.

\begin{figure}
    \centering
    \caption{
        This example demonstrates the computation of density and precision metrics on a 
        synthetic dataset. The left side shows the latent representation of: 
        \protect\mycolorbox{0,100,255,0.6}{real data},
        \protect\mycolorbox{255,30,0,0.6}{the real outlier}, 
        generated samples \protect\mycolorbox{0,0,0,0.2}{near the outlier} 
        and \protect\mycolorbox{0,127,0,0.6}{near non-outliers}.
        For each real sample we represent its neighborhood area. 
        The right side depicts the original data series. The density 
        metric, unlike precision, correctly identifies the outlier, 
        giving a score of $1.25$ instead of $1$. This illustrates how density 
        better reflects fidelity by accounting for outliers. Both metrics use $2$ neighbors.
    }
    \label{fig:metrics-density}
    \includegraphics[width=\textwidth]{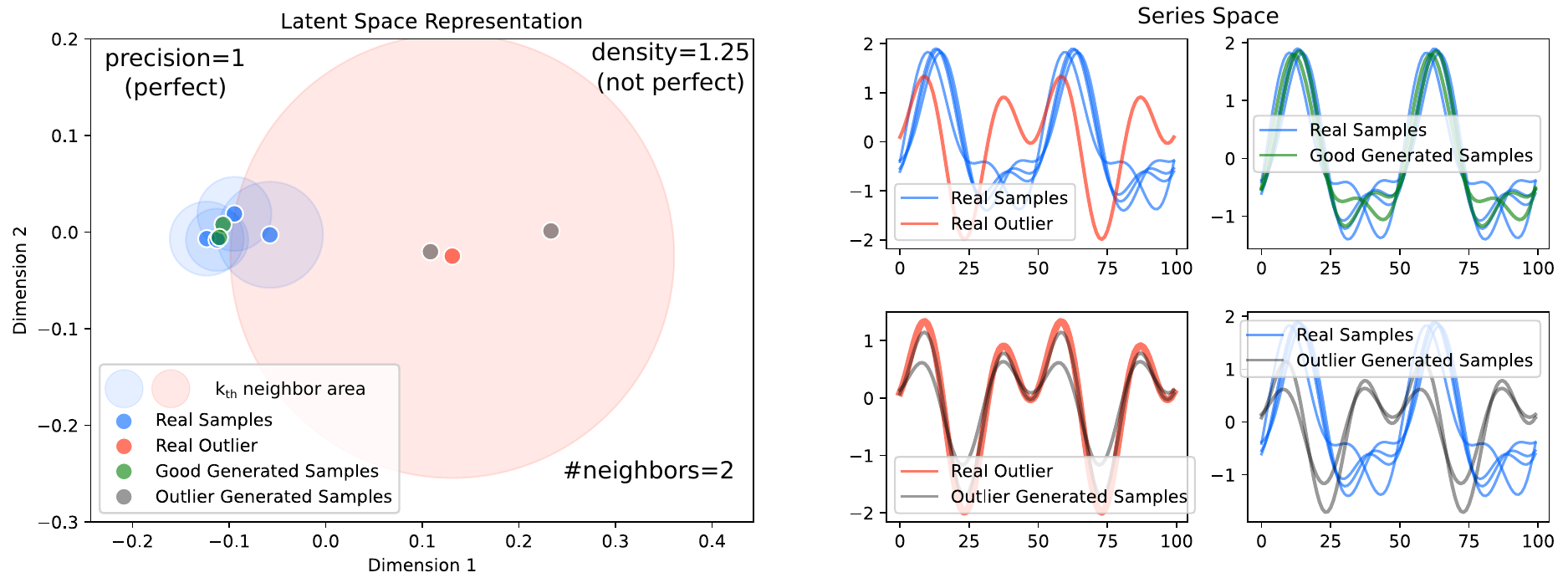}
\end{figure}

Figure~\ref{fig:metrics-density} illustrates an outlier scenario with a
synthetic dataset. On the left, 
it shows the latent space of five real samples and four generated samples, 
highlighting an outlier in red. The generated samples correctly cluster around 
the four real samples, with two near the outlier. Precision fails here, giving a 
misleading perfect score of $1$, which suggests high fidelity. However, the density 
metric reveals the issue, as it identifies the two generated samples influenced by 
the outlier, resulting in a more accurate, non-perfect score of $1.25$.

\paragraph*{Setup for generative models}
First, we determine the distance from each sample in $\textbf{V}=\mathcal{F}(\mathcal{X})$
to its $k_{th}$ nearest neighbor within $\hat{\textbf{V}}=\mathcal{F}(\hat{\mathcal{X}})$.
Next, we compute the precision and density metrics using Eqs.~\ref{equ:metrics-precision}
and~\ref{equ:metrics-density}, respectively.

\paragraph*{Interpretation}
The precision metric measures the number of generated samples that fall within at 
least one real sample's neighboring sphere. Conversely, the density metric counts 
how many neighboring spheres each generated sample occupies. While both metrics 
assess how well generated samples match real ones, density provides a more detailed 
analysis by considering each real-generated pair individually. Precision, however, 
overlooks potential biases from real outliers by focusing only on the union of neighboring 
spheres, missing the nuanced fidelity captured by density.

\subsection{Diversity Metrics}\label{sec:metrics-diversity}

Evaluating fidelity alone does not fully ensure the reliability of generated 
samples; hence, diversity measures are necessary. Next, we introduce the 
diversity metrics commonly used in the literature, including both distance-based 
and neighbor-based approaches.

\subsubsection{Average Pair Distance (APD)}\label{sec:metrics-apd}

Another common metric we used in the previous chapter is the Average Pair Distance (APD). 
Originally proposed by~\cite{apd-original-paper} for measuring distances between images, 
it was adapted by~\cite{action2motion-paper} for evaluating the diversity of human motion 
generative models. APD calculates the average Euclidean Distance, in the latent space of the pre-trained encoder,
between randomly 
selected sample pairs, repeated $R$ times over $S_{apd}$ pairs. The final APD value is the average 
result of these experiments. This metric can evaluate the diversity of any dataset, 
not just generated samples.
The APD metric calculated on the generated set of samples, for one
random selection $r~\in~\{1,2,\ldots,R\}$ is formulated as follows:
\begin{equation}\label{equ:metrics-apd}
    APD_r(\mathcal{S},\mathcal{S}^{'}) = \dfrac{1}{S_{apd}}\sum_{i=1}^{S_{apd}} \sqrt{\sum_{j=1}^{f} (\mathcal{S}_{i,j} - \mathcal{S}^{'}_{i,j})^2} 
\end{equation}
\noindent where $\mathcal{S}$ and $\mathcal{S}^{'}$ are two randomly 
selected subsets of $\hat{\textbf{V}}=\mathcal{F}(\hat{\mathcal{X}})$, 
i.e. $\mathcal{S},\mathcal{S}^{'}~\subset \hat{\textbf{V}}$.

\begin{figure}
    \centering
    \caption{
        This example demonstrates the calculation of the APD metric for both
        \protect\mycolorbox{0,100,255,0.6}{real samples} 
        and \protect\mycolorbox{255,30,0,0.6}{generated samples}. 
        For each latent representation, two sets $\mathcal{S}$ and $\mathcal{S}^{'}$
        of randomly selected values, 
        each of size $S_{apd}$, are created. The APD metric is then computed 
        between these sets. This process is repeated RR times, and the final APD 
        value is the \protect\mycolorbox{0,128,0,0.6}{average of all computed APD values}.
    }
    \label{fig:metrics-apd}
    \includegraphics[width=\textwidth]{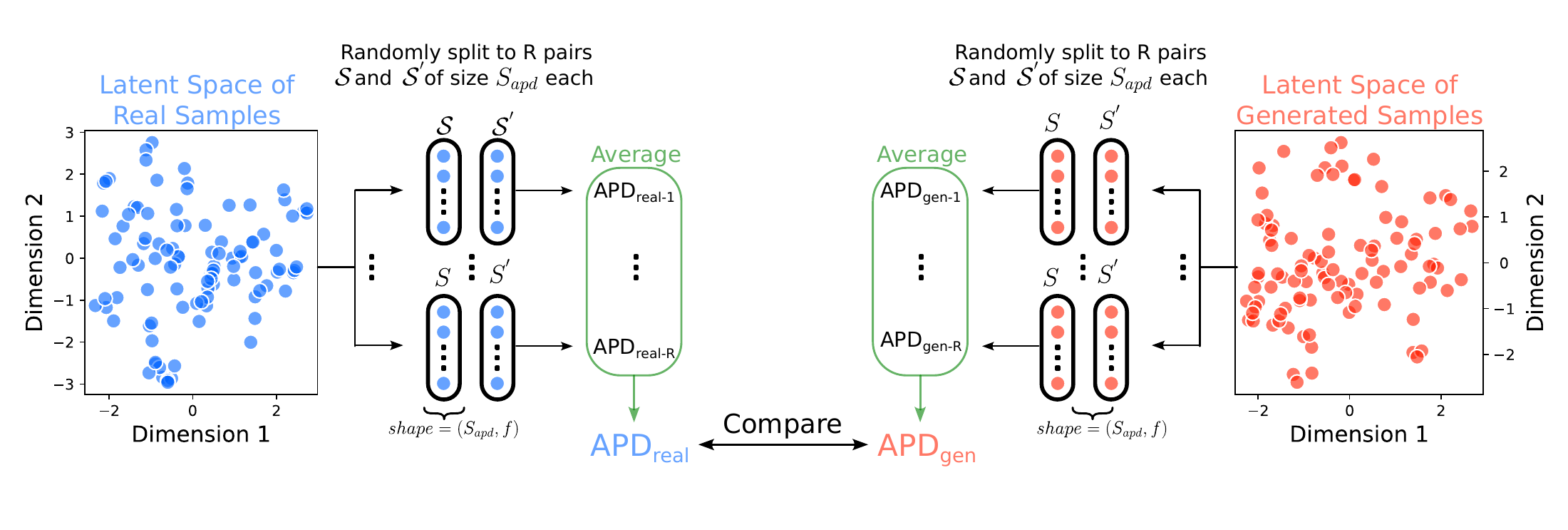}
\end{figure}

The APD metric is then calculated by averaging over the $R$ random experiments:
\begin{equation}\label{equ:metrics-apd-all}
APD(\hat{\mathcal{X}}) = \dfrac{1}{R}\sum_{r=1}^{R} APD_r(\mathcal{S}^r,\mathcal{S}^{'r})
\end{equation}

\noindent where $APD_r(\hat{\mathcal{X}})$ is calculated using Eq.~(\ref{equ:metrics-apd}).
An illustration of the APD metric is represented in Figure~\ref{fig:metrics-apd} 
highlighting the procedure to calculate APD on real and generated data separately.

\paragraph*{Setup for generative models}
We calculate the APD metric using Eqs~\ref{equ:metrics-apd} and~\ref{equ:metrics-apd-all}.
This is done for both real and generated datasets independently of each other, with only the encoder, pre-trained 
on real data, used for both computations.

\paragraph*{Interpretation}
The APD metric assesses whether a generative model can avoid mode collapse, 
where it generates the same outcome repeatedly. Ideally, the APD metric should 
be as high as possible to indicate diverse outputs. However, a potential issue 
arises if APD$_{gen}>APD_{red}$, as it suggests that the generated space is more 
diverse than the real one, which is an implausible outcome.
To address this,~\cite{action2motion-paper}
provided further interpretation, leading to the following theorem:

\begin{theorem}[Average Pair Distance Interpretation]\label{the:metrics-apd}
    A generative model $Gen_1$ is considered more diverse
    than another model $Gen_2$ if $APD_{gen1} > APD_{gen2}$,
    while respecting the following constraint:
    \begin{equation*} 
        \forall~\epsilon>0, |APD_{gen1} - APD_{real}| < \epsilon 
    \end{equation*}
\end{theorem}

In simpler terms, a generative model's APD diversity should not surpass that of the 
real distribution \pr. To illustrate, if real data has an APD$_{real}$ of $5$
and a generative model is randomly initialized without training, the APD$_{gen}$
could exceed $5$. However, this higher diversity would be due to random generation,
not a meaningful correlation with \pr. This demonstrates that exceeding too much the real 
data's diversity doesn't necessarily reflect a well-trained model.

\subsubsection{Average per Class Pair Distance (ACPD)}\label{sec:metrics-acpd}

The Average per Class Pair Distance (ACPD)~\cite{action2motion-paper}
metric, like APD, evaluates the 
diversity of generated samples but at a more detailed level. While APD 
measures diversity across the entire distribution \pr, ACPD focuses on individual 
sub-clusters within \pr. This allows for a more nuanced assessment of how well 
the model captures diversity within specific categories. ACPD computes the 
average APD for each sub-cluster, formed using the class labels, providing insights into class-specific diversity. 
The mathematical formulation of ACPD is as follows:
\begin{equation}\label{equ:metrics-acpd}
    ACPD_r(\mathcal{S},\mathcal{S}^{'}) = \dfrac{1}{C.S_{acpd}}\sum_{i=1}^{S_{acpd}}\sqrt{\sum_{j=1}^{f} (\mathcal{S}_{c,i,j} - \mathcal{S}^{'}_{c,i,j})^2} 
\end{equation}
\noindent where $C$ is the total number of classes in \pr, 
$\mathcal{S}_c,\mathcal{S}^{'}_c$ are randomly selected 
subsets from $\hat{\textbf{V}}[\hat{Y} = c]$, and $\hat{Y}$ 
are the labels used to generated $\hat{\mathcal{X}}$.

Similar to APD, due to the randomness involved in ACPD, the experiment is 
repeated $R$ times to calculate ACPD$_r$ for $r~\in~\{1,2,\ldots,R\}$.
The final ACPD value is then obtained by averaging these repeated calculations:
\begin{equation}\label{equ:metrics-acpd-all}
    ACPD(\hat{\mathcal{X}}) = \dfrac{1}{R}\sum_{r=1}^{R}ACPD_r(\mathcal{S}^r,\mathcal{S}^{'r})
\end{equation}
It is important to note that this metric is restricted only to labeled datasets where 
labels are discrete, e.g. classification.

\paragraph*{Setup for generative models}
We calculate ACPD using Eqs.~\ref{equ:metrics-acpd} and~\ref{equ:metrics-acpd-all}.
This is done for both real and generated datasets independently of each other, with only the encoder, pre-trained 
on real data, used for both computations.

\paragraph*{Interpretation}

The ACPD metric assesses the diversity of generated samples within each sub-cluster 
of \pr, ensuring that the model doesn't over-focus on a single cluster. 
This addresses the common issue of imbalanced labeled data in machine 
learning, where some classes may have more diversity than others.
Similar to the APD metric, ACPD is calculated for both real and generated samples, 
resulting in ACPD$_{real}$ and ACPD$_{gen}$.
A generative model is deemed class diverse when ACPD$_{gen}$
closely matches ACPD$_{real}$, indicating balanced generation across all categories.

\subsubsection{Coverage and Recall}\label{sec:metrics-coverage}

As mentioned in Section~\ref{sec:metrics-density},~\cite{reliable-fidelity-diversity}
proposed new metrics to replace the improved precision and recall metrics 
introduced by~\cite{precision-recall-paper,improved-precision-recall-paper}.
Their recall metric measures how well the generated distribution \pg~can
sample real examples from \pr. This is achieved by counting the number 
of real samples that appear in at least one neighborhood of a generated sample. 
The recall metric is formulated as follows:
\begin{equation}\label{equ:metrics-recall}
    recall(\textbf{V},\hat{\textbf{V}},k) = \dfrac{1}{N}\sum_{i=1}^{N}\mathds{1}(\textbf{V}_i~\in~manifold(\hat{\textbf{V}}_1,\hat{\textbf{V}}_2,\ldots,\hat{\textbf{V}}_M)) 
\end{equation}
\noindent where $manifold(.)$ and $\mathds{1}(.)$ follow the same definition 
detailed in Section~\ref{sec:metrics-density}. The recall metric is bounded between $0$ and $1$.

\cite{reliable-fidelity-diversity} identified several limitations of
the recall metric, summarized as follows:
\begin{enumerate}
    \item Defining neighborhood areas based on generated samples can lead 
    to misinterpretations, as outliers are more likely to be 
    sampled by \pg~than by~\pr.
    \item There is no closed-form solution for the expected value of the recall 
    metric when \pr~and \pg~are identical distributions, complicating the evaluation process.
\end{enumerate}

To overcome the limitations of the recall metric,~\cite{reliable-fidelity-diversity}
proposed the coverage metric. This metric focuses on neighborhood areas around 
the real samples $\mathcal{X}$. It counts how many real samples include at 
least one generated sample from $\hat{\mathcal{X}}$. The coverage metric provides 
a more accurate representation by measuring the presence of generated samples 
within the vicinity of real samples. The formulation of the coverage metric is as follows:
\begin{equation}\label{equ:metrics-coverage}
    coverage(\textbf{V},\hat{\textbf{V}},k)=\dfrac{1}{N}\sum_{i=1}^{N}\mathds{1}(\exists~j~s.t.~\hat{\textbf{V}}_j~\in~B(\textbf{V}_i,NND_k(\textbf{V}_i))) 
\end{equation}
    
\noindent where $B(.,.)$ and $NND_k(.)$ follow the same definitions detailed 
in Section~\ref{sec:metrics-density}. The coverage metric is bounded between $0$ and $1$.

\begin{figure}
    \centering
    \caption{
        The computation of coverage and recall metrics over a synthetic dataset is 
        shown on the left and right side of the figure.
        The figure's left and right sides depict the latent space with 
        \protect\mycolorbox{0,100,255,0.6}{real samples} and
        generated samples both \protect\mycolorbox{0,128,0,0.6}{reliable},
        and \protect\mycolorbox{255,30,0,0.6}{outliers}. 
        The middle shows the original series space, highlighting the differences
        between the three spaces. 
        The left plot shows neighbor areas around generated samples (recall metric), 
        while the right plot shows neighbor areas around real samples (coverage metric). 
        The coverage metric correctly identifies outliers, resulting in a non-perfect 
        measure, unlike the recall metric. Both metrics use $2$ neighbors.
    }
    \label{fig:metrics-coverage}
    \includegraphics[width=\textwidth]{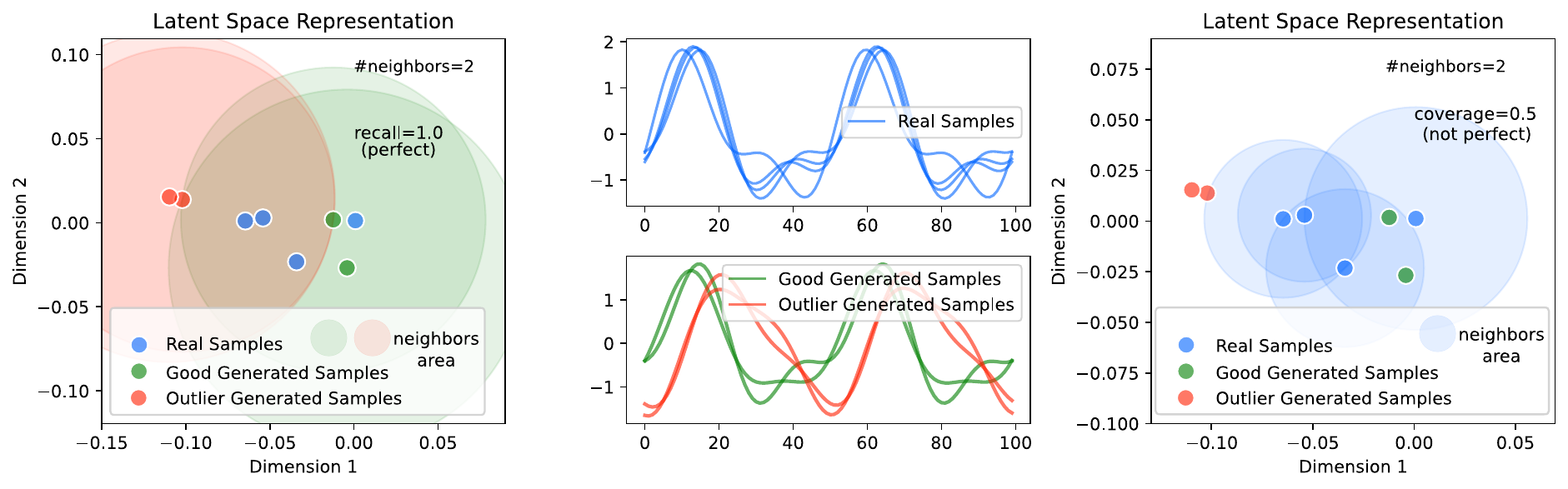}
\end{figure}

The coverage metric addresses the recall metric's limitation by avoiding the use of 
generated sample neighborhoods. Figure~\ref{fig:metrics-coverage} illustrates this 
with a synthetic example. 
The recall metric falsely shows perfect diversity due to over-estimated 
neighborhoods caused by outliers (left scatter plot). In contrast, the 
coverage metric accurately differentiates between valid and outlier samples
by focusing on the neighborhoods of real samples. This results in a more 
reliable assessment of the generative model's diversity.

The coverage metric also resolves the second limitation of recall, 
as demonstrated by~\cite{reliable-fidelity-diversity}. They showed that 
for identical distributions \pr~and~\pg, the expected value of coverage 
has a simplified closed-form solution, as such:
\begin{equation}\label{equ:metrics-expected-coverage}
    \mathds{E}[coverage] = 1-\dfrac{(N-1)\ldots(N-k)}{(G+N-1)\ldots(G+N-k)}
\end{equation}
\noindent which reduces to, in the case where both $N$ and $G$ are high enough:
\begin{equation}\label{equ:metrics-expected-coverage-high-nm}
    \mathds{E}[coverage] = 1 - \dfrac{1}{2^k} .
\end{equation}

\paragraph*{Setup for generative models}
First, we calculate the recall metric by determining the distance of each 
sample in $\hat{\textbf{V}}$ to its $k_{th}$ nearest neighbor in $\hat{\textbf{V}}$
and applying Eq.~\ref{equ:metrics-recall}. Second, we compute the coverage 
metric by finding the distance of each sample in $\textbf{V}$ to its $k_{th}$ 
nearest neighbor in $\textbf{V}$ and using Eq.~\ref{equ:metrics-coverage}.

\paragraph*{Interpretation}
The recall metric calculates the proportion of real samples that fall within 
the neighborhood of at least one generated sample. Conversely, the coverage metric 
determines the proportion of real samples that have at least one generated 
sample in their neighborhood. While both metrics assess diversity, 
the coverage metric is more reliable as it is based on real sample 
neighborhoods. With an expected value of $1-\dfrac{1}{2^k}$
for identical \pr~and~\pg, the interpretation of high diversity for the coverage 
metric depends on the chosen number of neighbors $k$.

\subsubsection{Mean Maximum Similarity (MMS)}\label{sec:metrics-mms}

Originally proposed in~\cite{mms-paper}, the Mean Maximum Similarity (MMS) 
metric evaluates the novelty of generated data. Generative models 
can sometimes produce data almost identical to the training set, 
mimicking its diversity without solving the intended task. MMS 
addresses this by quantifying the novelty of generated samples, which 
we interpret as a measure of diversity in this context. This ensures 
that the generated data is not just varied but also distinct from the training set.

\begin{figure}
    \centering
    \caption{
        This example illustrates the MMS metric computation on a synthetic dataset. 
        The left side shows the latent representation of
        \protect\mycolorbox{0,100,255,0.6}{real samples}
        and \protect\mycolorbox{0,128,0,0.6}{generated samples}. The right side displays the 
        original series space. First, each generated point's 
        nearest neighbor in the real set is identified using the 
        \protect\mycolorbox{255,30,0,0.6}{Euclidean Distance}. 
        Second, the MMS metric is obtained by averaging all these distances.
    }
    \label{fig:metrics-mms}
    \includegraphics[width=\textwidth]{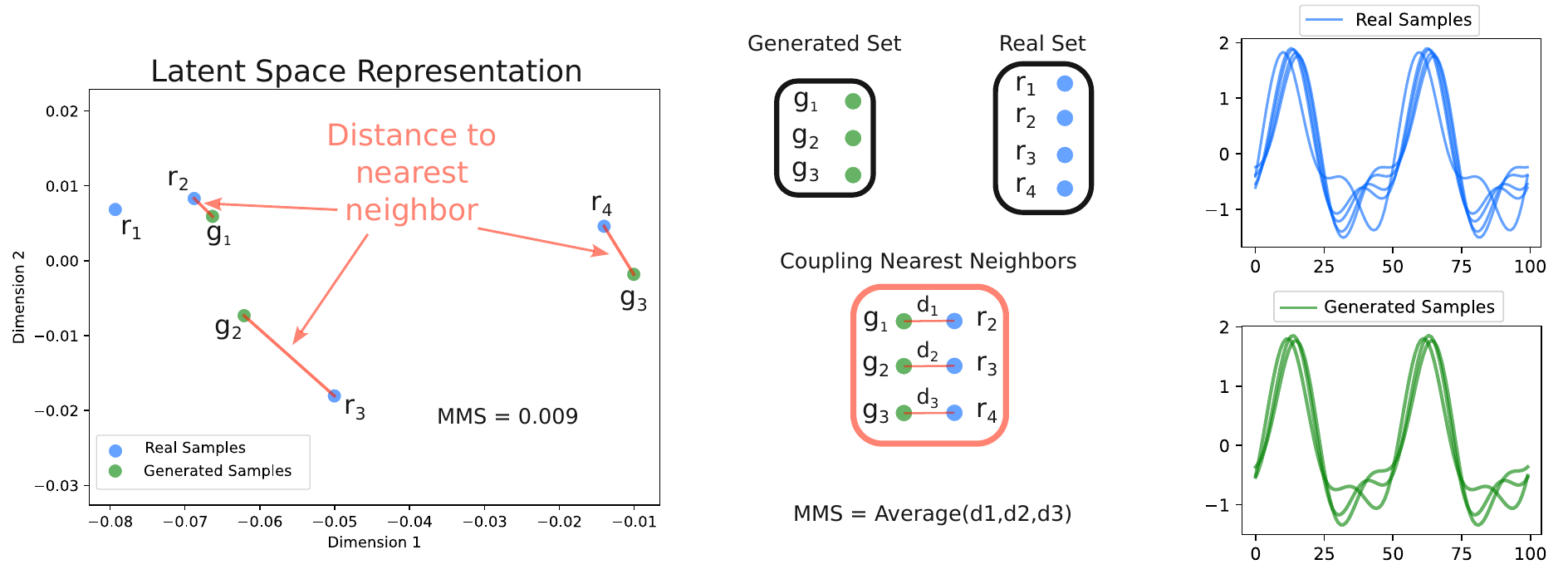}
\end{figure}

The MMS quantifies novelty/diversity by averaging the distances of each
of the generated samples to its real nearest samples.
It is given by:
\begin{equation}\label{equ:metrics-mms}
    MMS(\textbf{V},\hat{\textbf{V}}) = \dfrac{1}{G}\sum_{j=1}^{G}\sqrt{\sum_{d=1}^{f}(\hat{\textbf{V}}_{j,d} - \textbf{V}_{NN_j,d})^2} 
\end{equation}
    
\noindent where $\textbf{V}_{NN_j}$ (from the real set) is 
the nearest neighbor to $\hat{\textbf{V}}_j$ (from the generated set).
A visual representation of the $MMS$ metric is shown in Figure~\ref{fig:metrics-mms}.

\paragraph*{Setup for generative models}
First, for each sample in $\hat{\textbf{V}}$, calculate its distance to the 
nearest neighbor in $\textbf{V}$ and average these distances to obtain MMS$_{gen}$.
Second, for each sample in $\textbf{V}$, calculate the distance to the second 
nearest neighbor within $\textbf{V}$ and average these to determine MMS$_{real}$.
We adhere to the method outlined in~\cite{msm-paper} and do not use 
the $\textbf{V}_1$ and $\textbf{V}_2$ sets for the MMS metric calculation.

\paragraph*{Interpretation}
\cite{msm-paper} suggested that MMS$_{gen}$ should always be higher than
MMS$_{real}$ to signify high novelty. However, this metric has limitations, 
particularly when the model generates random samples far from the real set, 
leading to an overestimation of novelty. Thus, relying solely on MMS 
for evaluating a generative model's performance is insufficient. Additionally, 
MMS$_{real}$ is calculated within the entire real sample set, not between 
two subsets, which differs from the approach used for other metrics.

\section{Proposed Metric: Warping Path Diversity (WPD)}\label{sec:metrics-wpd}

\begin{figure}
    \centering
    \caption{
        This figure demonstrates the need for a temporal distortion 
        diversity metric. On the left side, three \protect\mycolorbox{0,100,255,0.6}{real (top)}
        and three \protect\mycolorbox{255,30,0,0.6}{generated (bottom)} 
        human motion sequences performing the "drink-with-left-hand" 
        action are presented. In the middle, the $y$-axis projection of the 
        subject's left-hand motion is displayed
        \protect\mycolorbox{30,117,179,0.6}{for the}
        \protect\mycolorbox{255,125,12,0.6}{three}
        \protect\mycolorbox{43,158,43,0.6}{samples} from both real and generated spaces. 
        The real samples show variability in the starting frame, 
        while the generated samples start consistently. On the right, 
        the latent representation of \protect\mycolorbox{0,100,255,0.6}{real}
        and \protect\mycolorbox{255,30,0,0.6}{generated} samples using a pre-trained 
        GRU classifier reveals that the model does not account for 
        temporal distortion diversity.
    }
    \label{fig:metrics-warping}
    \includegraphics[width=0.9\textwidth]{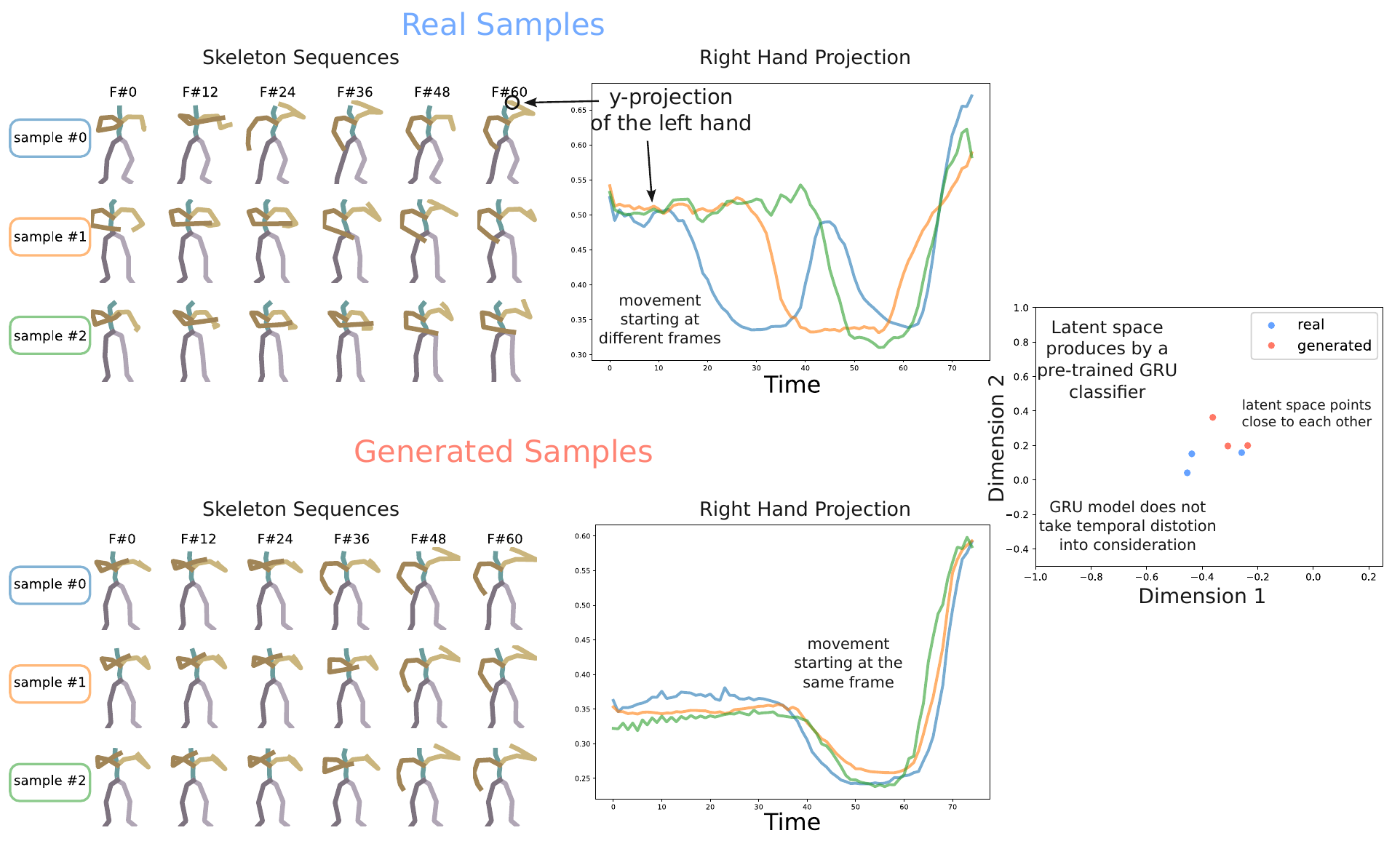}
\end{figure}

Each diversity metric in Section~\ref{sec:metrics-diversity}
relies on a pre-trained encoder $\mathcal{F}$ to extract latent 
features, assuming an input latent space. However, for temporal 
data like human motion sequences, some temporal distortions exist, such as shifting and frequency changes. 
For instance, as presented in Figure~\ref{fig:metrics-warping},
real samples from the HumanAct12 dataset show the 
action of drinking starting at different frames, while generated 
samples lack this variability. The pre-trained encoder $\mathcal{F}$ 
fails to account for these distortions, affecting metrics 
like APD, which measure Euclidean distances in latent space. 
To address this, we propose a new metric that uses Dynamic 
Time Warping (DTW)~\cite{dtw-paper} (see Chapter~\ref{chapitre_1}) to capture and quantify 
temporal distortions between sequences.

\begin{figure}
    \centering
    \caption{
        The distance matrix between \protect\mycolorbox{0,100,255,0.6}{two}
        \protect\mycolorbox{255,30,0,0.6}{time series} is shown in a 
        heat map where each point represents the squared difference 
        between corresponding time stamps. The optimal
        \protect\mycolorbox{0,255,255,0.6}{Dynamic Time Warping 
        (DTW) path}, captures the 
        temporal distortion between the series. The
        \protect\mycolorbox{144,238,144,1.0}{connections}
        between the warping path and the 
        \protect\mycolorbox{0,128,0,0.6}{diagonal} 
        indicate how much the two series deviate from having 
        no temporal distortion.
    }
    \label{fig:metrics-warping-example}
    \includegraphics[width=0.5\textwidth]{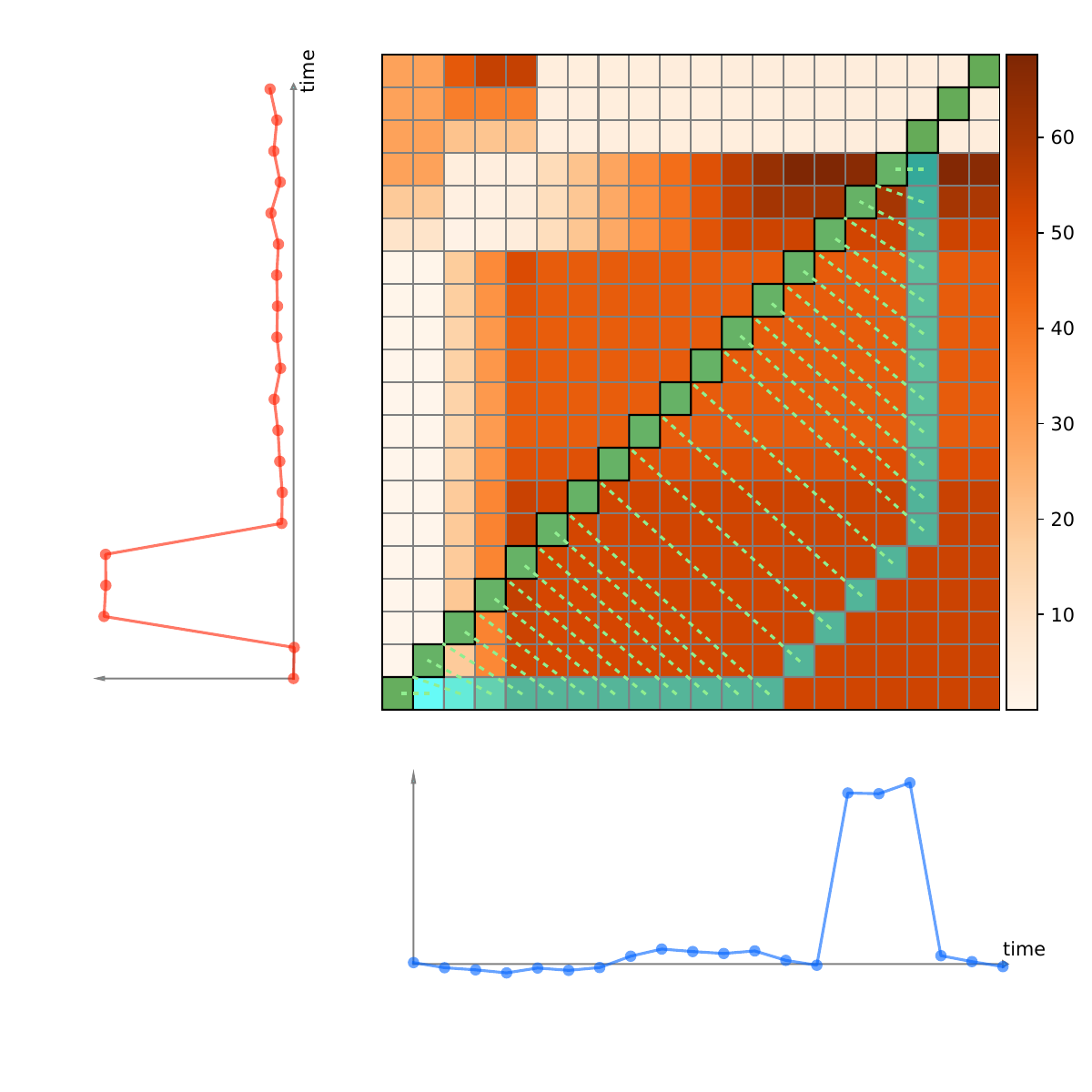}
\end{figure}

To simplify, we assume both sequences xx and yy have the same length $L$. 
For the sequences in Figure~\ref{fig:metrics-warping-example},
three scenarios can occur:

\begin{enumerate}
    \item \textbf{Worst-case scenario}: The sequences are poorly aligned, 
    with the DTW path running along the matrix edges, resulting in a 
    path length of $L_{\pi}=2L$.
    \item \textbf{Best-case scenario}: The sequences are perfectly 
    aligned along the diagonal, making DTW equivalent to Euclidean Distance.
    \item \textbf{Intermediate scenario}: Temporal distortions cause the path
    to deviate from the diagonal but remain shorter than the maximum value;
    $L_{\pi}<2L$.
\end{enumerate}

\begin{figure}
    \centering
    \caption{Mathematical basis of the WPD metric:
    For \protect\mycolorbox{255,30,0,0.6}{each point} on 
    the \protect\mycolorbox{0,128,0,0.6}{warping path}, 
    the \protect\mycolorbox{255,100,255,0.6}{corresponding triangle} is always a right 
    isosceles triangle, given that the series are of equal length.
    Hence the \protect\mycolorbox{0,100,255,0.6}{distance} 
    from the \protect\mycolorbox{255,120,0,0.6}{point}
    to the \protect\mycolorbox{220,220,220,1.0}{diagonal} can be easily calculated 
    with the Pythagorean theorem~\cite{pythagorean-theorem-paper}.}
    \label{fig:metrics-dtw-triangle}
    \includegraphics[width=0.6\textwidth]{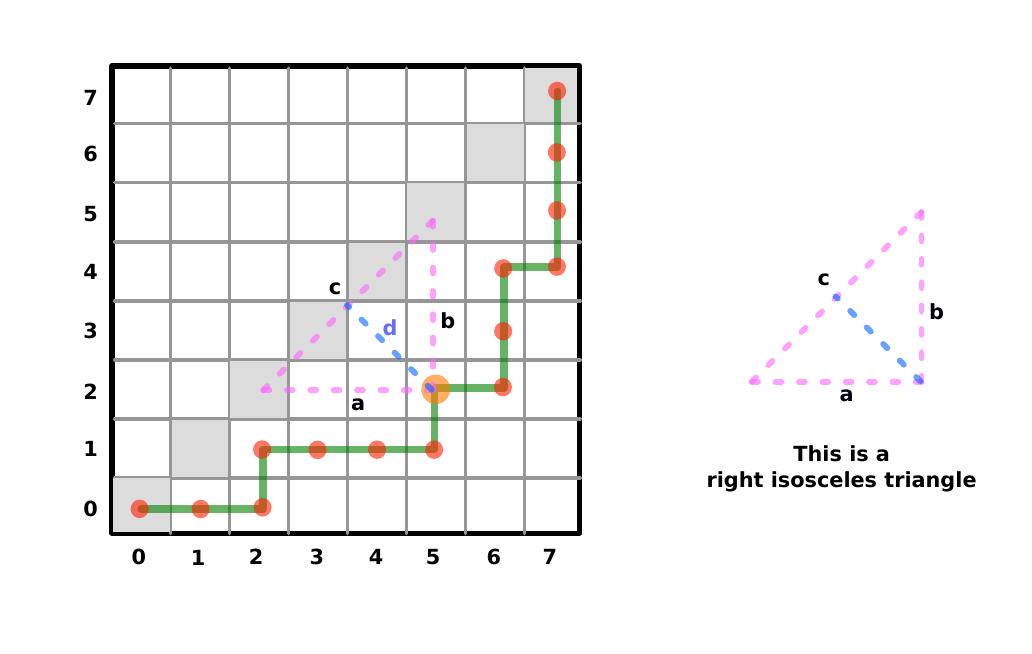}
\end{figure}

To measure the diversity of temporal distortions (warping) between two 
sequences, we propose quantifying the distance of the warping path 
from the diagonal. This involves summing the distances from each point 
on the warping path to the diagonal, as illustrated in 
Figure~\ref{fig:metrics-dtw-triangle}. Each point on the path is 
considered within an integer coordinate space with axes ranging 
from $1$ to $L$. For equal-length sequences, the triangle at each 
warping path point is a right isosceles triangle, making the hypotenuse's 
median half its length.

\begin{theorem}[Warping Path Diversity's Distance To Diaginal Computation]\label{the:metrics-distance-to-diagonal}
    Given two sequences $\textbf{x}$ and $\textbf{y}$ both of length $L$,
    with warping path $\pi$ of length $L~\leq~L_{\pi}~\leq~2L$, the distance 
    from the $t_{th}$ point of the path $\pi$,
    where $t~\in~\{1,2,\ldots,L_{\pi}\}$,
    to the diagonal (perfect alignment) is defined as follows:
    \begin{equation*}
        distance(\pi_t,diagonal) = \dfrac{\sqrt{2}}{2}|t_1-t_2|
    \end{equation*}
\end{theorem}

\textit{proof}: 
Using the annotations of $a,~b,~c$ and $d$ in
Figure~\ref{fig:metrics-dtw-triangle} as the summit of points
$\pi_t$ on the warping path $\pi$, then $d=~distance(\pi_t,diagonal)$ is
calculated as follows:
\begin{equation}\label{equ:metrics-proof-wpd}
    \begin{split}
        d &= \dfrac{1}{2}\sqrt{c^2} = \dfrac{1}{2}\sqrt{a^2+b^2} = \dfrac{1}{2}\sqrt{2 * a^2}\\
        &= \dfrac{1}{2}\sqrt{2*(t_1-t_2)^2} = \dfrac{\sqrt{2}}{2} |t_1-t_2|
    \end{split}
\end{equation}

The WPD value between $\textbf{x}$ and $\textbf{y}$ is the average distance 
of all points on the warping path to the diagonal, as follows:
\begin{equation}\label{equ:wpd-d}
    WPD_d(\textbf{x},\textbf{y}) = \dfrac{\sqrt{2}}{2.L_{\pi}}\sum_{t=1}^{L_{\pi}}|t_1-t_2|
\end{equation}

Finally, the WPD metric of a generative model is calculated, 
like for the APD metric, between random subsets of samples 
from both real and generated samples:
\begin{equation}\label{equ:wpd-r}
WPD_r(\mathcal{S},\mathcal{S}^{'}) = \dfrac{1}{S_{wpd}}\sum_{i=1}^{S_{wpd}}WPD_d(\mathcal{S}_i,\mathcal{S}^{'}_i)
\end{equation}

\noindent where $\mathcal{S}$ and $\mathcal{S}^{'}$ are two randomly 
selected subsets, of size $S_{wpd}$, from
$\hat{\textbf{V}} = \mathcal{F}(\hat{\mathcal{X}})$, 
i.e., $\mathcal{S},\mathcal{S}^{'}~\subset~\hat{\textbf{V}}$ and 
$r~\in~\{1,2,\ldots,R\}$ is the number of repetitions of this 
random experiment to avoid the bias of a random selection.
The final $WPD$ metric is calculated as:
\begin{equation}\label{equ:wpd}
WPD(\hat{\mathcal{X}}) = \dfrac{1}{R}\sum_{r=1}^{R}WPD_r(\mathcal{S}^r,\mathcal{S}^{'r}) 
\end{equation}

\noindent with $WPD$ bounded between $0$ and 
$\dfrac{\sqrt{2}}{4}(L+1)$.

The same methodology is follows to calculate WPD on the real set of samples 
$\mathcal{X}$.
The characteristics of our proposed WPD metric 
are summarized in Table~\ref{tab:metrics-summmary}
along with all the other metrics presented in this study.

\begin{table}
    \centering
    \caption{Summary of the Generative Models Metrics in this study.}
    \label{tab:metrics-summmary}
    \resizebox{\columnwidth}{!}{
    \begin{tabular}{|c|c|c|c|c|c|c|c|}
      \hline
      Metric & Category & Space & Hyperparameters & Bounds & Interpretation & Better Version & Used in Study \\
      \hline
      $FID$ & fidelity & latent & None & 
      \begin{tabular}{c}
        $0\leq FID < \infty$
      \end{tabular} & 
      \begin{tabular}{c}
        Higher but close \\ to $FID_{real}$
      \end{tabular} & None & yes \\
      \hline
      $AOG$ & fidelity/accuracy & latent & None & 
      \begin{tabular}{c}
        $0\leq AOG \leq 1$
      \end{tabular} & 
      \begin{tabular}{c}
        Close to 1 \\ (100\% accuracy)
      \end{tabular} & None & yes \\
      \hline
      $density$ & fidelity & latent & 
      \begin{tabular}{c}
        $k$: number \\ of neighbors
      \end{tabular} & 
      \begin{tabular}{c}
        $0\leq density \leq N/k$ \\ $N$ being the number \\ of real samples
      \end{tabular} & 
      \begin{tabular}{c}
        closer to $density_{real}$ \\ which is close to 1
      \end{tabular} & None & yes \\
      \hline
      $precision$ & fidelity & latent & 
      \begin{tabular}{c}
        $k$: number \\ of neighbors
      \end{tabular} & 
      \begin{tabular}{c}
        $0\leq precision \leq 1$
      \end{tabular} & 
      \begin{tabular}{c}
        closer to 1
      \end{tabular} & $density$ & no \\
      \hline
      $coverage$ & diversity & latent & 
      \begin{tabular}{c}
        $k$: number \\ of neighbors
      \end{tabular} & 
      \begin{tabular}{c}
        $0\leq coverage \leq 1$
      \end{tabular} & 
      \begin{tabular}{c}
        closer to $coverage_{real}$ \\ which is close to $1-1/2^k$
      \end{tabular} & None & yes \\
      \hline
      $recall$ & diversity & latent & 
      \begin{tabular}{c}
        $k$: number \\ of neighbors
      \end{tabular} & 
      \begin{tabular}{c}
        $0\leq recall \leq 1$
      \end{tabular} & 
      \begin{tabular}{c}
        closer to 1
      \end{tabular} & $coverage$ & no \\
      \hline
      $APD$ & diversity & latent & 
      \begin{tabular}{c}
        $S_{apd}$: size of \\ random subset \\ $R$: number of \\ random experiments
      \end{tabular} & 
      \begin{tabular}{c}
        $0\leq APD < \infty$
      \end{tabular} & 
      \begin{tabular}{c}
        Lower but close \\ to $APD_{real}$
      \end{tabular} & None & yes \\
      \hline
      $ACPD$ & diversity & latent & 
      \begin{tabular}{c}
        $S_{acpd}$: size of \\ random subset \\ $R$: number of \\ random experiments
      \end{tabular} & 
      \begin{tabular}{c}
        $0\leq ACPD < \infty$
      \end{tabular} & 
      \begin{tabular}{c}
        Lower but close \\ to $ACPD_{real}$
      \end{tabular} & None & yes \\
      \hline
      $MMS$ & diversity/novelty & latent & None & 
      \begin{tabular}{c}
        $0\leq MMS < \infty$
      \end{tabular} & 
      \begin{tabular}{c}
        Higher but close \\ to $MMS_{real}$
      \end{tabular} & None & yes \\
      \hline
      $WPD$ (\textbf{ours}) & diversity/warping & raw & 
      \begin{tabular}{c}
        $S_{wpd}$: size of \\ random subset \\ $R$: number of \\ random experiments
      \end{tabular} & 
      \begin{tabular}{c}
        $0\leq WPD \leq \dfrac{\sqrt{2}}{4}(L+1)$ \\ $L$ being the length \\ of the sequence
      \end{tabular} & 
      \begin{tabular}{c}
        Depends on the \\ application
      \end{tabular} & None & yes \\
      \hline
    \end{tabular}}
\end{table}

\section{Experimental Setup}

\begin{figure}
    \centering
    \caption{
        In our experiments, the CVAE undergoes two phases: training 
        and generation.\textbf{ During training}, the \protect\mycolorbox{0,194,0,0.47}{Encoder} 
        and \protect\mycolorbox{0,194,171,0.47}{Decoder} 
        are trained simultaneously. The \protect\mycolorbox{0,194,0,0.47}{Encoder}
        extracts features 
        from input sequences and projects them into a Gaussian latent 
        space with a learned \protect\mycolorbox{234,194,139,0.47}{mean} 
        and \protect\mycolorbox{234,112,139,0.47}{variance}, conditioned 
        on the \protect\mycolorbox{255,30,1309,0.6}{action label}. The
        \protect\mycolorbox{0,194,171,0.47}{Decoder}  then 
        reconstructs the input sequence from a
        \protect\mycolorbox{0,100,255,0.6}{sample in this space}. 
        \textbf{In the generation phase}, a \protect\mycolorbox{160,100,255,0.6}{random
        sample from a Normal 
        distribution} is fed to the \protect\mycolorbox{0,194,171,0.47}{Decoder}
        to generate a new sequence, 
        also conditioned on the desired \protect\mycolorbox{255,30,1309,0.6}{action label}.
    }
    \label{fig:metrics-cvae}
    \includegraphics[width=\textwidth]{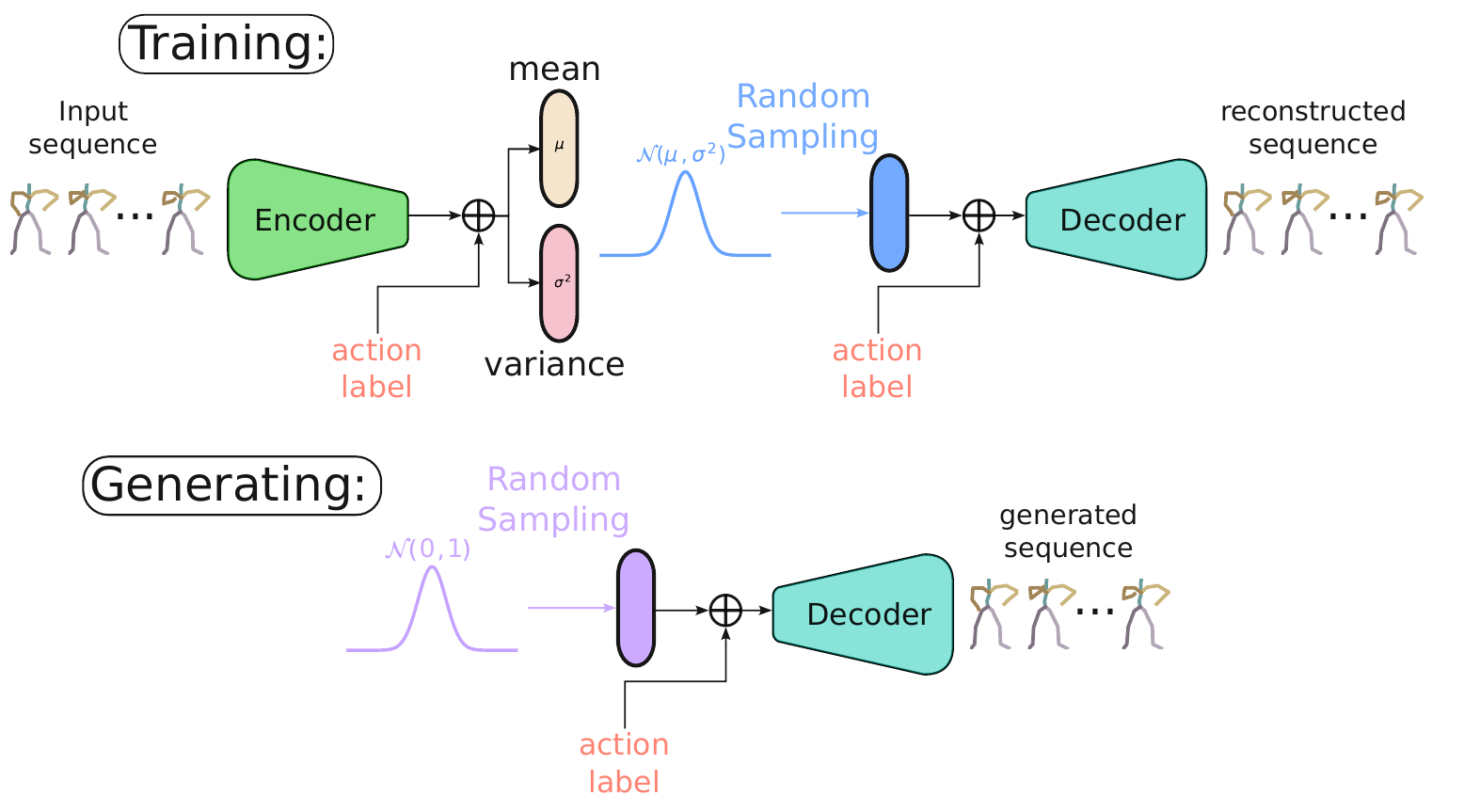}
\end{figure}

To analyze the behavior of each metric during evaluation, we 
conduct an experiment using Conditional Variational Auto-Encoders 
(CVAE) to generate human motion sequences. The conditioning feature of 
the CVAE allows precise control over the generated actions by specifying 
the action class. Figure~\ref{fig:metrics-cvae} illustrates
the training and generating 
phases of a CVAE with human motion sequences, providing a visual 
representation of our experimental setup. This approach enables a 
comprehensive assessment of how different metrics respond to the 
generated data.

\subsection{Backbone Architectures}

The CVAE model employs an encoder-decoder architecture using three 
well-known neural network backbones: Convolutional Neural Networks 
(CNNs), Recurrent Neural Networks (RNNs), and Transformer Networks. 
The CNN-based CVAE (CConvVAE) uses symmetrical convolution and 
de-convolution blocks. The RNN-based CVAE (CGRUVAE) uses stacked 
Gated Recurrent Units (GRUs) for both encoding and decoding, 
repeating the input for sequence generation. The Transformer-based 
CVAE (CTransVAE) features self-attention mechanisms in both encoder 
and decoder, with matching layers and parameters. Each architecture 
maintains symmetry between its encoder and decoder components.

\subsection{Implementation Details}

The three CVAE variants in this work are implemented using 
\textit{tensorflow}~\cite{tensorflow-paper} Python package
and trained for $2,000$ epochs with a batch size of $32$, 
utilizing a learning rate decay method. The CConvVAE employs 
three convolution and three de-convolution blocks with $128$
filters and kernel sizes of $40$, $20$, and $10$. 
The CGRUVAE has two GRU layers in both the encoder and decoder, 
with a hidden state size of $128$. The CTransVAE uses convolution 
embedding followed by two Multi-Head Attention layers in both the 
encoder and decoder, with $128$ filters and a head size of $32$. 
All three variants have a latent space dimension of $16$.
To train the generative models, we utilize a publicly available 
action recognition dataset, HumanAct12~\cite{action2motion-paper}.
Prior to
training, we normalize all sequences in the dataset using a 
$\min-\max$ scalar on each of the $x~-~y-~z$ dimensions independently.
It is important to note that for all the metrics used in this
work, no prior train-test splits are required, instead all the dataset can
be used.

\subsection{Training on Different Loss Parameters}
In this experimental work, we assess how slight changes in a model's parameters may affect
the interpretation of 
evaluation metrics by experimenting with the model's loss parameters. 
We train a CVAE model to optimize a weighted sum of two losses: 
reconstruction loss and Kullback-Leibler (KL) divergence loss.
The total loss is defined as:
\begin{equation}\label{equ:metrics-total-loss}
    \mathcal{L} = \alpha.\mathcal{L}_{rec} + \beta.\mathcal{L}_{KL}
\end{equation}
    
\noindent where $\alpha$ and $\beta$ are scalar weights between 
$0$ and $1$ for each of the reconstruction and KL loss respectively.
In the ideal case, it is preferable to maintain the following constraint:
\begin{equation}\label{equ:metrics-alpha-beta}
    \alpha + \beta = 1 
\end{equation}
As seen in chapter~\ref{chapitre_1}, $\alpha$ is set to $1-\beta$ to preserve convexity.
Instead of selecting a specific $(\alpha,\beta)$ pair, we experiment 
with various values: $(1E^{-1},9E^{-1})$, $(5E^{-1},5E^{-1})$, 
$(9E^{-1},1E^{-1})$, $(9.9E^{-1},1E^{-2})$, $(9.99E^{-1},1E^{-3})$, 
$(9.999E^{-1},1E^{-4})$. This approach highlights how different weightings 
influence model performance and metric interpretation.

\subsection{Class Imbalanced Generation Setup}

To unify the evaluation method and ensure fairness,
we propose a generation setup that addresses the class 
imbalance problem in training datasets, relevant to any supervised 
generative model, including those used for human action recognition. 
Given that all metrics compare real and generated distributions, 
it's crucial to ensure fair evaluation. To do this, we match the 
label distribution of generated samples with that of the training 
set, preventing the model from over-representing majority classes. 
When generating more samples than available in the training set, a
proportional factor is applied to maintain the original label distribution, 
ensuring balanced and unbiased sample generation.

\section{Results and Analysis}

\begin{figure}
  \centering
  \caption{
    Radar charts compare the performance of three CVAE variants 
    across eight metrics. Each chart, labeled from \textbf{a} to
    \textbf{d}, represents 
    a different $(\alpha,\beta)$ parameter set. The charts 
    feature four polygons: one for \protect\mycolorbox{255,102,138,0.25}{each}
    \protect\mycolorbox{230,171,2,0.25}{CVAE}
    \protect\mycolorbox{117,112,179,0.25}{variant} and one for 
    \protect\mycolorbox{27,158,119,0.25}{real data metrics}.
    For all metrics except FID, a higher summit 
    indicates better performance. For FID, a higher summit means 
    worse performance.
  }
  \label{fig:metrics-radar}
  \includegraphics[width=\textwidth]{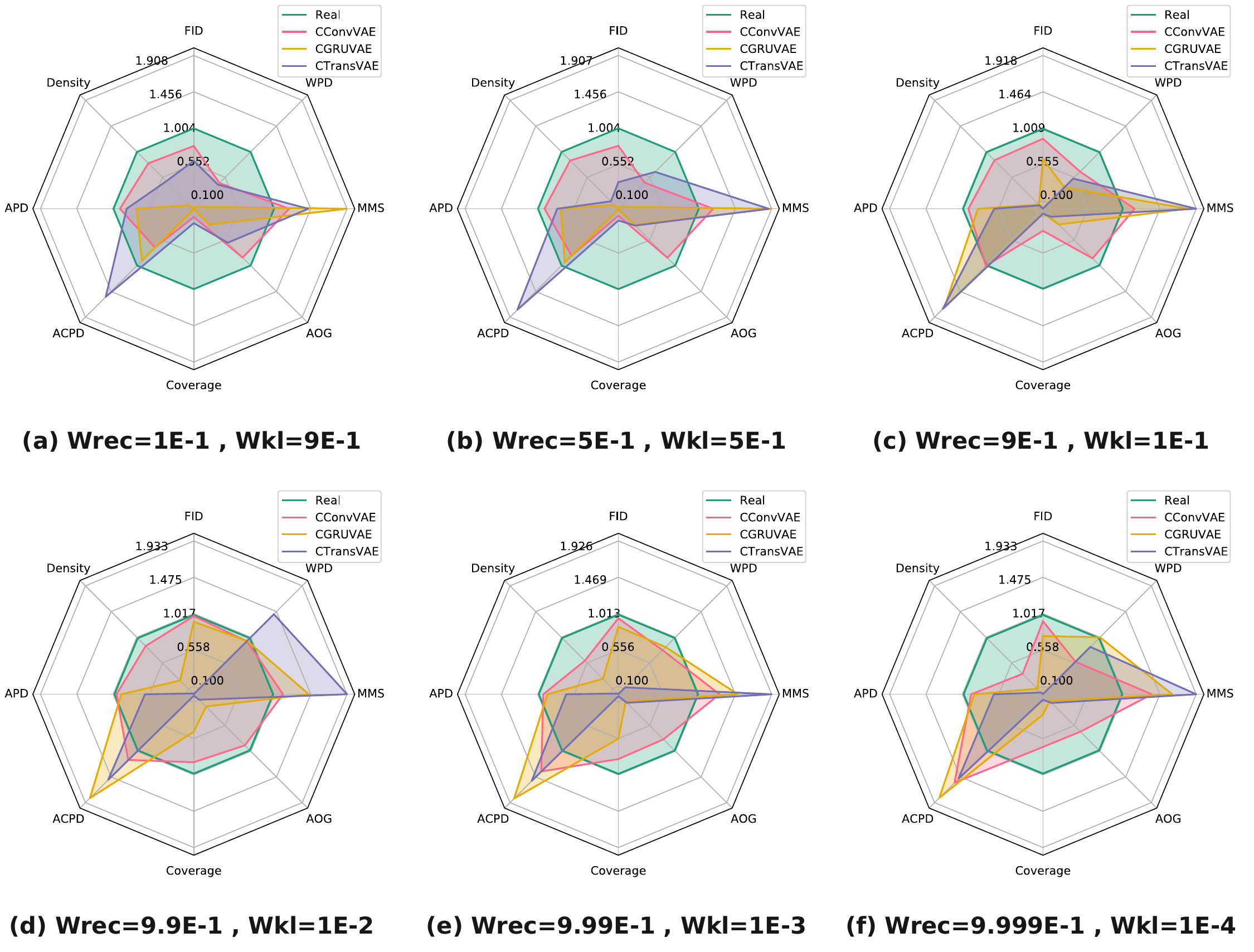}
\end{figure}

To evaluate our models, we use radar charts due to the varying ranges 
of our metrics, which offers a clearer comparison than simply stating 
numerical differences. For each $(\alpha,\beta)$ pair, a radar chart displays 
four polygons (see Figure~\ref{fig:metrics-radar}):
one for each CVAE variant and one for real samples. 
Metrics are normalized between 0 and 1 and transformed for comparison, 
with a summit lower than the real polygon indicating 
$metric_{gen} < metric_{real}$, except for FID where it 
indicates $FID_{gen} > FID_{real}$.
Optimal performance is shown by generative model 
polygons closely matching the real polygon, except 
for the MMS metric where higher generative values are better.

Figure~\ref{fig:metrics-radar} illustrates the difficulty 
in finding a generative model that excels across all metrics simultaneously. 
Changes in the backbone architecture or loss parameters can significantly 
impact metric values. Our experiments with the HumanAct12 dataset 
reveal that selecting the best model, CConvVAE, is only feasible 
for a specific set of $(\alpha,\beta)$ values, as shown in
Figure~\ref{fig:metrics-radar}-d. However, this parameter search 
is not always practical, making it challenging to identify 
the best model across all metrics. Therefore, depending on 
the application, we may need to prioritize specific metrics
or even a single metric.

We now present an analysis of the results by comparing three 
CVAE models on each metric individually. For each metric, we 
explain what it means for a model to perform best.

\begin{itemize}
  \item \textbf{FID}: In certain $(\alpha,\beta)$ configurations,
  Figure~\ref{fig:metrics-radar} demonstrates that CConvVAE 
  achieves the lowest FID value, closely approaching FID$_{real}$.
  This suggests that CConvVAE generates samples with superior 
  fidelity compared to CGRUVAE and CTransVAE. The model efficiently 
  learns a distribution \pg~that is easier to align with \pr~than 
  the distributions produced by the other models. However, 
  despite having the smallest and most accurate FID, there are 
  cases where the gap between CConvVAE and FID$_{real}$ is still 
  considerable (Figures~\ref{fig:metrics-radar}-a-b-c).
  
  \item \textbf{Density}: Just as with the FID metric, for 
  certain $(\alpha,\beta)$ pairs, CConvVAE produces a Density 
  value that is closer to Density$_{real}$ than the other CVAE 
  variants. This suggests that CConvVAE is more likely to 
  generate samples resembling \pr~than CGRUVAE and CTransVAE. 
  However, even though CConvVAE often comes closest to 
  matching Density$_{real}$, a notable gap between Density$_{gen}$ 
  and Density$_{real}$ can still be observed in some instances.

  \item \textbf{AOG}: The AOG metric is crucial for evaluating a 
  generative model's conditional effectiveness. Factors like 
  data scarcity, underfitting, or poor hyperparameters can 
  impact performance. Figure~\ref{fig:metrics-radar} shows CConvVAE often 
  leads in this metric, but not always successfully; 
  for instance, in Figure~\ref{fig:metrics-radar}-f, CConvVAE's
  AOG value diverges 
  significantly from AOG$_{real}$. Meanwhile, CTransVAE and 
  CGRUVAE consistently fail to manage sub-classes across 
  all settings, indicating persistent issues with their 
  conditional mechanisms.

  \item \textbf{APD}: Regarding the APD diversity metric, 
  CConvVAE stands out by consistently achieving values near 
  APD$_{real}$ across various hyperparameter settings, 
  particularly excelling in Figure~\ref{fig:metrics-radar}-d. 
  However, CGRUVAE also shows strong performance on the APD 
  metric in some configurations. This demonstrates that while 
  CConvVAE may significantly outperform CGRUVAE in one aspect, 
  such as FID, CGRUVAE can still excel in other metrics, 
  highlighting its overall competence.

  \item \textbf{ACPD}: The ACPD metric evaluates diversity per 
  sub-class, while the APD metric assesses overall diversity. 
  Figure~\ref{fig:metrics-radar} shows that nearly half the 
  hyperparameter settings have all three CVAE variants outperforming 
  the real data on ACPD, indicating greater sub-class diversity. 
  However, this can result from overfitting, instability, and 
  class imbalance. Excelling in one metric doesn't imply overall 
  superiority. For example, CGRUVAE outperforms CConvVAE in ACPD 
  in Figure~\ref{fig:metrics-radar}-a but has a much higher FID, 
  indicating unreliable results. This highlights the need for 
  multiple evaluation metrics.
  
  \item \textbf{Coverage}: For most hyperparameter settings, 
  CConvVAE surpasses other CVAE variants in terms of coverage, 
  indicating it generates a greater number of samples that align 
  with the real distribution \pr~compared to CGRUVAE and 
  CTransVAE. However, in Figures~\ref{fig:metrics-radar}-a-b, CTransVAE 
  outperforms CConvVAE in coverage, showing that despite 
  CConvVAE's strong APD diversity, it can perform poorly 
  in coverage under certain conditions.

  This raises the question: What is the difference between APD and 
  coverage if they both quantify diversity? Both metrics use a latent space 
  and Euclidean Distance, however APD measures the distance between 
  randomly selected pairs to measure the volume of space occupied 
  by \pr~and~\pg, independently. Conversely, coverage evaluates the nearest 
  neighbor relationships between real and generated samples to 
  quantify how much of~\pr's space is occupied by the generated samples.

  \item \textbf{MMS}: The MMS metric measures diversity by 
  evaluating the novelty of generated samples, ideally with 
  MMS$_{gen}$ values higher but close to MMS$_{real}$. This metric 
  can be more challenging to interpret. CConvVAE demonstrates 
  the most stable MMS values among the three variants, maintaining 
  a higher yet comparable level to MMS$_{real}$. In contrast, 
  CTransVAE's MMS values exceed the radar plot's limits. 
  This indicates that CConvVAE excels at producing novel human 
  motion sequences.

  \item \textbf{WPD}: The WPD metric assesses temporal diversity 
  by measuring warping and distortions between samples. 
  There are three interpretations for WPD:
  \begin{enumerate}
    \item $|WPD_{real} - WPD_{gen}| < \epsilon$ (where $\epsilon$
    is very small): This signifies a perfect replication 
    of all temporal distortions from \pr~to~\pg.
    \item $WPD_{gen} >>> WPD_{real}$: This indicates that the 
    generative model has identified and created similar but new 
    temporal distortions.
    \item $WPD_{gen} <<< WPD_{real}$: This implies the model fails 
    to replicate temporal distortions, generating consistent but 
    limited distortions.
  \end{enumerate}
  Figure~\ref{fig:metrics-radar} shows that, apart from CTransVAE 
  in Figure~\ref{fig:metrics-radar}-d, all WPD values are 
  less than WPD$_{real}$. This means most models can re-create some 
  temporal distortions, though not all. Notably, in 
  Figure~\ref{fig:metrics-radar}-d, the minimal gap between 
  WPD$_{real}$ and WPD for CConvVAE and CGRUVAE suggests 
  a near perfect replication of temporal distortions in \pr.
\end{itemize}

\section{Conclusion}

In this chapter, we provided a comprehensive review of evaluation metrics 
used to assess the reliability of generative models for human 
motion generation. Recognizing that human motion data are 
temporal and represented as multivariate time series, we 
introduced a novel metric to evaluate diversity in terms 
of temporal distortion. We proposed a unified evaluation 
framework, with eight metrics measuring fidelity and diversity,
that allows for fair comparisons between different 
models. Our experiments with three generative model variants 
on a publicly available dataset demonstrated that no single 
metric can universally determine model superiority. Instead, 
a combination of different metrics is often necessary to 
accurately evaluate model reliability.

Our findings indicate that the CConvVAE model outperforms others 
on the highest number of metrics, which can give the impression 
of being the best overall model. However, minor hyper-parameter adjustments 
can significantly impact its performance across various metrics. 
This underscores the difficulty of identifying the \textit{best model},
the challenge of finding \textit{The One Metric To Rule Them All},
and highlights the importance of tailoring model selection to 
specific applications. For instance, in gaming, where generating 
diverse actions is crucial, diversity metrics are more important 
than fidelity metrics like FID. Conversely, in medical research, 
where precise replication of movements is critical, fidelity takes 
precedence.

Additionally, we offer publicly available, user-friendly code for calculating 
all the metrics we used, applicable to any generative model with any parameterization. 
We hope this work serves as a valuable starting point for newcomers to the field 
of human motion generation and helps establish a clear framework for unified evaluation. 
However, we acknowledge that the metrics discussed are not exhaustive, as the field is 
rapidly evolving. It is also important to adapt metrics when labels in the real data 
are unavailable.

\chapter{Reproducible Research}\label{chapitre_8}

\section{Introduction}

Reproducibility is a fundamental aspect of scientific research, as it ensures that our 
work can be replicated and adapted by other researchers for their own applications. In this chapter, we 
underscore the significance of reproducibility in research and outline the measures 
taken to guarantee that the work presented in this thesis aims to adhere to high 
standards of reproducibility.

A notable feature of this thesis is that almost all of the contributions have been 
integrated into an open-source Python package called \textit{aeon}~\cite{aeon-paper},
for which I am a 
core developer. This package serves as a centralized repository for the tools and
methods developed during this research, ensuring their accessibility and usability 
for the broader scientific community. By packaging our contributions in this way, 
we not only promote reproducibility but also encourage the wider use and continuous 
improvement of our work.

A key aspect of this thesis is that all the work detailed in the previous chapters 
is supported by publicly available code. This transparency allows other researchers 
to replicate our experiments, validate our findings, and build upon our work with 
confidence. By making our code accessible, we contribute to a more open and collaborative 
scientific community.

In this chapter, we will detail the specific requirements we follow to define good 
reproducible work. These requirements include clear documentation and adherence to 
best practices in software development, to the best of our capabilities.
We believe that by following these standards 
and taking feedback from the community, we can continually improve the reproducibility 
and reliability of our research.

Finally, every program used locally to produce figures or assist in analysis has been 
made available. While these tools may not be directly associated with any particular 
paper, they are crucial to the overall research process. We will present these programs 
in this chapter, highlighting their roles and functionalities.

In summary in this chapter we:

\begin{itemize}
    \item introduce the \textit{aeon} Python package, which encapsulates the contributions of this thesis.
    \item highlight the importance of reproducibility in research, by outlining
    the requirements and practices that define a reproducible work.
    \item present the local programs used for figure generation and analysis.
\end{itemize}

By committing to these principles of reproducibility and actively seeking feedback from 
the community, we aim to enhance the transparency, reliability, and impact of our research.

\section{Time Series Analysis With \textit{aeon}}

The \textit{aeon} Python package is a versatile open-source library developed to facilitate 
various time series machine learning tasks. It provides tools for classification, clustering,
transformations, regression, forecasting, anomaly detection, similarity search and segmentation,
making it a comprehensive solution for handling time series data.

\textit{aeon} is designed with ease of use and extensibility in mind, offering clear documentation 
and an intuitive interface. As a core developer of this package, the contributions from 
this thesis have been integrated into \textit{aeon}, ensuring that the methods and tools developed 
are accessible to the broader scientific community. This section will introduce \textit{aeon}, 
showcasing its capabilities and demonstrating its application in diverse time series 
machine learning scenarios.

\subsection{Deep Learning For Time Series With \textit{Aeon}}

My involvement in the \textit{aeon} project began with the task of defining 
the deep learning framework from the ground up. This framework 
had to be developed with several key principles in mind: ease of 
contribution for new developers, clarity of code, straightforward 
functionality, and comprehensive documentation. Ensuring that new 
contributors could easily understand and extend the code was paramount. 
As a result of these efforts, all the models reviewed in the 
deep learning for TSC~\cite{dl4tsc} are now implemented in \textit{aeon} for both 
classification and regression tasks. Additionally, new models 
such as InceptionTime~\cite{inceptiontime-paper},
our own H-InceptionTime (Chapter~\ref{chapitre_3}),
and LITETime (Chapter~\ref{chapitre_4}) 
have been incorporated, further enhancing the package's 
capabilities and robustness.

The maintenance of this framework is ongoing and 
involves several critical activities:

\begin{itemize}
    \item \textbf{Bug Fixes}: Regular updates are made to 
    identify and resolve bugs promptly, ensuring the stability 
    and reliability of the framework.
    \item \textbf{Documentation Improvement}: Continuous efforts 
    are made to enhance the documentation, making it clearer 
    and more comprehensive for users and contributors.
    \item \textbf{Feature Enhancement}: New features and capabilities 
    are regularly added to the framework, expanding its functionality 
    and keeping it at the forefront of time series machine learning 
    research.
    \item \textbf{Unit Testing}: Rigorous unit testing is conducted 
    to ensure the code remains robust and compatible with updates to 
    \textit{tensorflow}~\cite{tensorflow-paper}
    and \textit{keras}~\cite{keras-website}. This testing helps 
    maintain the integrity and performance of the framework as 
    the underlying libraries evolve.
\end{itemize}

By committing to these maintenance activities, we ensure that the \textit{aeon} 
package remains a valuable and reliable tool for the scientific community.

Recently, I started working on the deep learning for time series 
clustering module in \textit{aeon}, building upon the foundational work 
established in the original review~\cite{deep-tscl-bakeoff}.
This module is still under 
development, aiming to provide robust and efficient tools for 
clustering time series data using deep learning techniques. So 
far, I have included two original models from the review into 
the module, Auto-Encoder based deep learning models with FCN and ResNet
backbone networks, laying the groundwork for further expansion. In 
addition, I have been actively involved in mentoring a Google 
Summer of Code 
internship~\footnote{\url{https://summerofcode.withgoogle.com/programs/2024/projects/Hvd0DfkD}}, 
guiding the intern to enhance and 
develop more models for this module. This collaborative effort 
aims to accelerate the development process and ensure the 
inclusion of state-of-the-art deep clustering models in \textit{aeon}.

The upcoming work in deep learning within the \textit{aeon} package includes 
developing modules for time series anomaly
detection~\cite{anomaly-detection-review},
averaging~\cite{deep-averaging-paper}, 
and domain adaptation~\cite{deep-domain-adaptation-review}.
These enhancements will broaden \textit{aeon}'s 
capabilities, enabling it to tackle a wider range of time series 
challenges and better serve the research community.

\subsection{Other Tasks With \textit{Aeon}}

My involvement with the \textit{aeon} project extends beyond deep learning. 
In the distances module, I contributed to and continue to 
maintain the ShapeDTW similarity measure~\cite{shape-dtw-distance},
which is integral to my development of the ShapeDBA (Chapter~\ref{chapitre_6})
method in the averaging module. Additionally, I significantly 
optimized the runtime complexity of the PAA~\cite{paa} and SAX~\cite{sax} 
representation codes, making them much faster and more efficient. 
Beyond code contributions, I have also been actively involved in 
improving general documentation, creating example notebooks, 
and co-authoring the open-source software paper~cite{aeon}
with the rest of the core developers.

We believe that these efforts collectively enhance reproducibility 
within the research community, making tools more accessible and 
reliable for all users.

\section{What Makes A Work Reproducible~?}\label{sec:guide-coding}

Reproducibility is crucial in scientific research, enabling others 
to verify and build upon previous work. In this section, I will 
outline the key requirements we followed to ensure our research 
is reproducible. A reproducible codebase must be well-documented, 
extendable, easily modifiable, and well-architected. These practices cover 
technical aspects like code development and data management, as well 
as broader principles like clear documentation and transparency. 
By adhering to these standards, we aim to make our research both 
rigorous and accessible.

\subsection{Code Documentation}

While providing a GitHub repository might seem sufficient for 
ensuring code reproducibility, it is crucial to provide clear 
instructions for users who wish to re-run the experiments 
associated with the paper.

\subsubsection{Dependencies}

A key component of code documentation is a complete 
list of dependencies that the paper's code 
relies on is essential. Without this list, a new user would need to 
manually inspect all the code files to determine which dependencies 
are required, and in some cases, they might also need to identify 
the specific versions, especially if the code relies on older versions 
of these dependencies.

For this reason, we ensured that every project published during 
this thesis included a complete list of dependencies, thereby 
supporting better open-source reproducible research.

\subsubsection{Adapting Configuration To The User's Side}

Often, users may need to make minor adjustments to the code in 
order to successfully re-run it. This does not diminish the code's 
reproducibility, but it does make it essential for the repository 
to include a detailed, step-by-step guide. Such a guide should 
specify which variables need to be modified, where to make these 
changes, and how to do so. These variables might include the root 
directory for datasets. Some instructions can be for downloading 
datasets from a provided link etc.

Providing this information in the code's documentation is crucial; 
without it, users may encounter errors that prevent the code 
from running properly.

\subsection{Extendibility}

Ensuring that code is easily executable by others is a crucial 
first step towards creating a reproducible repository. However, 
research often involves incremental contributions, which means 
the original work must be designed to be extendable. For example, 
consider the deep learning for TSC review by~\cite{dl4tsc},
if a researcher wanted to add new models or datasets on top of 
this work, it should be easily achievable. If the code is not 
designed to allow such modifications, it cannot be considered 
truly extendable or modifiable.

\subsection{Code Architecture}

When the code from a research paper is intended to be studied 
by students, employees, or other researchers, it is essential 
that the code be well architected. A poorly structured codebase 
makes it difficult, if not impossible, for others to understand 
and build upon the work. Clear and thoughtful architecture is 
fundamental for ensuring that the techniques, algorithms, and 
overall project can be effectively comprehended and utilized by others.

\begin{enumerate}
    \item \textbf{Variable naming}: One critical element of 
    well-architected code is the use of meaningful variable 
    names. For instance, vague or arbitrary names like 
    ``$X = Y + Z.dot(alpha\_xyz)$'' should be avoided. 
    Instead, variables should be named descriptively, 
    especially when they represent parameters or concepts 
    from the paper's algorithm. For example, if the method 
    includes a parameter for the number of filters, the code 
    should use a name like ``$n\_filters$'' rather than something 
    ambiguous like ``$f$''.

    \item \textbf{File management}:
    Proposing a repository where a single file contains over 
    $10~000$ lines of code is simply unacceptable. Code should 
    be organized in a way that allows users to easily locate 
    specific functionalities. For example, if the project includes 
    multiple classifiers and normalization functions, each 
    classifier should be placed in its own file within a 
    dedicated sub-folder, while all normalization functions 
    could be grouped together in a separate file, as these 
    are typically concise. The choice of file management 
    structure doesn't follow a rule of thumb; it should be 
    tailored to the specific needs of the project.

    \item \textbf{Code Structure}:
    While there is no one-size-fits-all approach to file management, 
    the choice of code structure often follows certain best practices. 
    For instance, if the code does not require defining objects 
    with multiple functionalities, using only functions is sufficient, 
    and introducing classes would be unnecessary. Conversely, if 
    the code benefits from encapsulating behavior within objects, 
    then using classes is more appropriate.
    Furthermore, if several classes share common code, 
    it's not good practice to copy and paste these functionalities 
    across different classes. Instead, defining a base class that 
    these classes can inherit from is a better approach. 
    This also applies when working with abstract methods, 
    class methods, and similar concepts.
\end{enumerate}

\subsection{How To Check For All These Requirements}

When serving as the main developer of a project, it's often not 
ideal to be the one assessing whether all the necessary requirements 
for reproducibility and code quality are met. This is due to inherent 
human bias, as developers tend to view their own projects as being in 
an optimal state. To mitigate this bias, a better practice is to seek 
feedback from a fellow researcher, student, PhD candidate or a 
thesis supervisor. For example, you could ask one of these peers to 
try adding new functions to your code and provide feedback on how 
easily your code can be extended.

\section{Hardware Utilization and Accessible Code Repositories}

Throughout this thesis, I utilized a diverse set of hardware to support the computational demands 
of deep learning models, including a GTX1080ti with 8GB of VRAM, an RTX3090 with 24GB of VRAM, an 
RTX4090 with 24GB of VRAM, and the Mesocentre High Performance Computing Center 
of the University of Strasbourg. Initially, setting 
up the GPUs involved manually configuring the necessary CUDA tools, which required significant time 
and effort to ensure compatibility and performance. However, as the work progressed, I transitioned 
to using Docker containers, which provided several advantages, including simplified environment 
management, enhanced portability, and the ability to encapsulate all dependencies within a container. 
This ensured that the code ran consistently across different machines and facilitated easier scaling 
and deployment.

For each published article, I ensured that the community has access to all source codes and resources 
necessary to reproduce the work, as outlined previously in Table~\ref{tab:intro-contributions}.
The provided Dockerfiles encapsulate the code and configure the environment 
for seamless GPU integration, allowing users to execute the code efficiently with just two to three commands, 
leveraging GPU acceleration without the need for complex manual configurations. Additionally, all codes were 
developed following the guidelines outlined in the previous section, ensuring consistency, reliability, and 
ease of use for others in the community.

\section{Published Work Serving For Analysis And Reproducibility}

A central goal of this thesis has been to ensure that all tools 
used in our analysis are accessible to the broader community. 
By making these resources available, we enhance the community's 
ability to understand, replicate, and build upon our work, thereby 
increasing its impact and clarity for future readers.

We accomplished this through two main strategies: first, by publishing 
public repositories that contain the code used to generate visualizations 
and support our research analysis; and second, by creating web pages 
featuring interactive tools that go beyond what can be presented in a 
paper or GitHub repository. These efforts aim to deepen the community's 
understanding and facilitate wider engagement with our research.

\subsection{Elastic Warping Visualization}

\begin{figure}
    \centering
    \caption{Warping path example between two time series from 
    the ECGFiveDays dataset.}
    \label{fig:example-warping-path-github}
    \includegraphics[width=0.6\textwidth]{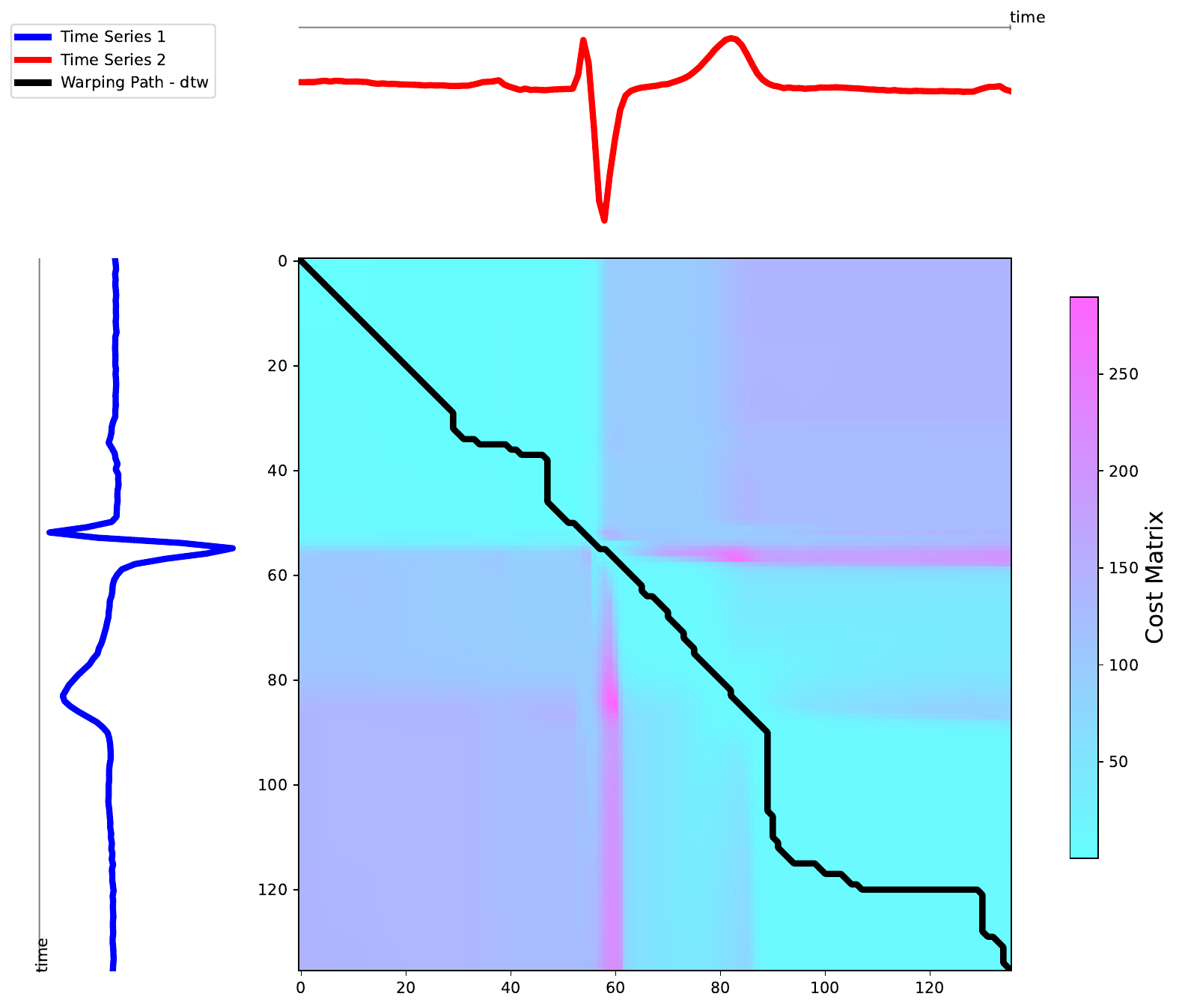}
\end{figure}

A fundamental approach to analyzing time series data involves 
assessing how sensitive the samples are to temporal distortions. 
While there are various methods to achieve this, the most effective 
way is often through direct visualization. To facilitate this, we 
developed a public GitHub repository~\cite{ismail-fawaz2024elastic-vis} 
available here:
\url{https://github.com/MSD-IRIMAS/Elastic_Warping_Vis}.
This repository allows users to visualize the warping path and 
temporal distortions between any two time series samples, with 
outputs available in either PDF format, as shown in 
Figure~\ref{fig:example-warping-path-github}, or as MP4 
format~\footnote{\url{https://github.com/MSD-IRIMAS/Elastic\_Warping\_Vis/blob/main/exps/dtw-vis/ECGFiveDays/dtw.mp4}}.

The code leverages the \textit{aeon} Python package for computing the warping 
path and distance matrix, and uses \textit{matplotlib} for 
visualization~\cite{matplotlib-paper}.
Users can apply any of the distance functions implemented in \textit{aeon}~\footnote{
\url{https://www.aeon-toolkit.org/en/stable/api_reference/distances.html}}.

The repository not only meets all the requirements discussed in
Section~\ref{sec:guide-coding} for public availability on 
GitHub but is also conveniently 
installable via PyPi~\footnote{\url{https://pypi.org/}}.

\subsection{Convolutional Filter Space Visualization}\label{sec:filters-code}

Training a CNN often requires post-training analysis of the convolutional 
filter space. For example, in Chapter~\ref{chapitre_3}
this type of analysis was conducted to ensure that the model was not relearning hand-crafted filters, as 
illustrated in Figure~\ref{fig:hcf-tsne_filters}.
After publishing this work,
we developed a GitHub repository that provides code to 
generate similar figures for any pre-trained model.
This tool~\footnote{\url{https://github.com/MSD-IRIMAS/filter1D\_visualization}}
produces visualizations in both PDF format, 
like Figure~\ref{fig:hcf-tsne_filters}
and in a web-friendly format using \textit{bokeh}~\cite{bokeh}, 
a visualization tool for web 
interfaces~\footnote{\url{https://maxime-devanne.com/pages/filter1D\_visualization/}}.
This resource will assist researchers in better analyzing 
the convolutional filter space when working with CNNs.

\subsection{Augmenting Time Series Classification Datasets With Elastic Averaging}

Since the publication of~\cite{weighted-dba-paper}, research on data augmentation for 
time series data has seen a significant increase in both quantity and diversity. 
The work in \cite{weighted-dba-paper} focuses on using weighted elastic averaging, specifically 
DBA~\cite{dba-paper}, for data augmentation to enhance 
the performance of TSC models. Although the original
implementation was written in Java, the method continues to attract attention and 
further development. In response, we proposed an open-source Python implementation 
of this approach~\cite{Ismail-Fawaz2023weighted-ba}~\footnote{\url{https://github.com/MSD-IRIMAS/Augmenting-TSC-Elastic-Averaging}}, 
utilizing the \textit{aeon}~\cite{aeon-paper} library as the backend for similarity and 
averaging computations. Our code adheres to open-source standards and offers flexibility, 
allowing users to parameterize the augmentation function with any similarity 
measure they choose.

\subsection{KAN It Work For Time Series Classification ?}

Following the publication of a new approach to neural networks 
using Kolmogorov-Arnold Networks (KANs)~\cite{kan-paper}, which are designed to 
effectively model complex relationships within tabular data by leveraging the 
Kolmogorov-Arnold representation theorem, we were interested in testing their 
applicability to time series data. Given their success with tabular data, we 
proposed a method that involves extracting Catch22~\cite{catch22} features from time 
series data before feeding them into the KAN model for classification tasks. Our goal 
was to provide an open-source repository~\cite{Ismail-Fawaz2023kan-c22-4-tsc}~\footnote{\url{https://github.com/MSD-IRIMAS/Simple-KAN-4-Time-Series}}
for this work, encouraging the time series community to engage with KAN 
models and fostering continued research and innovation in this area.

\subsection{Published Webpages With Associated Papers}

Developing webpages can be essential when visualizations, such as videos or interactive 
tools, cannot be effectively presented in a paper. For the hand-crafted convolution 
filters contribution, we created a 
webpage~\footnote{\url{https://msd-irimas.github.io/pages/HCCF-4-tsc/}}
featuring an interactive tool, utilizing the code discussed in Section~\ref{sec:filters-code}.
Similarly, for the ShapeDBA paper, we developed a 
webpage~\footnote{\url{https://msd-irimas.github.io/pages/ShapeDBA/}}
showcasing video visualizations of the ShapeDTW alignment, 
highlighting the differences between DBA and ShapeDBA in 
terms of the underlying similarity measure.

\subsection{Deep Learning For Time Series Classification: A Webpage}

Since the publication of the deep learning for TSC review~\cite{dl4tsc}, the research community 
has shown significant interest in this domain. However, with the increasing number 
of models,some of which are introduced in this thesis,revisiting the review with 
another paper wasn't ideal. Instead, we developed a 
webpage~\footnote{\url{https://msd-irimas.github.io/pages/dl4tsc/}} that features a detailed 
overview of each model from the 2019 review, along with newer models. This includes 
information on the number of parameters, FLOPS, and links to the original papers. 
The webpage also features a CD diagram~\footnote{\url{https://github.com/hfawaz/cd-diagram}} 
and an MCM from Chapter~\ref{chapitre_2} to present the results across the UCR archive~\cite{ucr-archive}.

Additionally, we created \textit{bokeh}~\cite{bokeh} based 1v1 scatter plots, allowing users to compare 
models of their choice. Furthermore, we included an interactive \textit{bokeh} plot to 
showcase the average performance of all models concerning their FLOPS and 
parameter counts across all UCR archive datasets. Users can also select specific 
datasets instead of viewing the average performance.

We believe that this webpage, with ongoing maintenance, will provide the research 
community with easy access to comprehensive details, results, and comparisons 
needed for deep learning in the TSC task.

\section{Conclusion}

In this chapter, we emphasized the critical role of reproducibility in scientific 
research, particularly within the context of time series analysis and deep learning.
Ensuring that research is transparent and replicable allows others to confidently 
build upon existing work. To this end, we have meticulously documented our code and 
methodologies, providing clear instructions providing clear instructions that facilitate 
accurate reproduction of our experiments. Additionally, we have focused 
on creating a well-structured and organized codebase, adhering to best practices in 
software engineering. This approach not only ensures the code's functionality but also 
makes it easier for other researchers to extend and adapt our work.

Moreover, we have made significant efforts to share the tools and resources 
developed during this thesis with the broader research community. By publishing 
our work on GitHub and integrating it into the \textit{aeon} Python package, we have 
made these resources widely accessible, fostering a collaborative environment 
for further innovation. The development of interactive web tools further enhances 
the accessibility and understanding of our work, allowing users to explore complex 
data and results dynamically. Through these efforts, we aim to contribute to a 
culture of reproducibility in research, ensuring that our work serves as a reliable 
foundation for future advancements in time series analysis.
%
%
\addchap{Conclusion and future works}
\label{conlusion_globale}

\section*{Overview Of Contributions}

The work presented in this thesis brings significant progress to the field of 
time series analysis, particularly in areas concerning supervised and unsupervised 
learning, benchmarking machine learning models, foundation models for time series 
classification, generative models and model complexity reduction. Through comprehensive 
experimentation and rigorous evaluation, the thesis has developed and 
demonstrated several novel methodologies for addressing some of the most 
challenging aspects of time series data, such as high-dimensionality, 
multivariate dependencies, and limited labeled data.

Chapter~\ref{chapitre_1} lays the groundwork for the research, providing an extensive 
review of the state-of-the-art techniques for time series analysis. 
It introduces both supervised and unsupervised learning methods, including Time 
Series Classification and Extrinsic Regression. In addition, the chapter discusses 
self-supervised learning techniques, paving the way for the novel approaches 
later introduced in the thesis. This chapter serves as an essential foundation, 
offering insight into existing challenges and the research landscape in deep learning for 
time series analysis.

A central contribution of this thesis is the introduction of more refined tools for
model benchmarking in time series analysis especially discriminative models. Chapter~\ref{chapitre_2}
critiques traditional methods like the Critical Difference Diagram (CDD)~\cite{demsar-cdd-paper}, 
which suffers from instabilities and overlooks the magnitude of performance differences. 
The Multiple Comparison Matrix (MCM) is introduced as an alternative, offering more 
reliable comparisons across datasets, addressing weaknesses in existing statistical tests. 
This new method improves the precision of model evaluation and provides clearer insights 
into relative model performance.

In Chapter~\ref{chapitre_3}, the search for foundational models that can generalize 
across time series data is explored. A key contribution here is the development of 
pre-trained models capable of adapting to various classification tasks. This 
includes PHIT (Pre-trained H-InceptionTime), which shows significant improvements 
in performance when compared to non-pre-trained models across a large number of 
time series datasets. By establishing a foundation for transfer learning in time series, 
this chapter paves the way for future applications of foundational models.
These models leverages a second key contribution introduced in this chapter, the hand-crafted convolution
filters that help generalization of the deep learning models by offering prior non-trainable filters
to detect specific patterns in the time series.
Extensive experiments on the UCR archive~\cite{ucr-archive} highlights how our proposed domain 
foundation models outperforms classical baseline deep learning techniques with no pre-training.

Chapter~\ref{chapitre_4} presents the LITE and LITEMV models, showing significant 
advancements in reducing the computational complexity of deep learning models 
for Time Series Classification. LITE offers a simplified architecture based on 
Inception~\cite{inceptiontime-paper}, reducing the number of parameters while 
maintaining competitive performance. Additionally, LITEMV adapts this architecture 
for multivariate time series, incorporating DepthWise Convolutions~\cite{mobilenets}
to handle data from multiple channels more efficiently. These models represent a major 
step toward making deep learning more accessible in resource-constrained environments.
Experiments on both the UCR archive (univariate)~\cite{ucr-archive} and the UEA archive 
(multivariate)~\cite{uea-archive} demonstrate that smaller models can be just as 
effective as complex ones. This is achieved with only a few boosting techniques 
to address the trade off between model complexity and performance.

Building on the challenges of acquiring labeled data, Chapter~\ref{chapitre_5} 
introduces TRILITE, a self-supervised model using triplet loss~\cite{triplet-loss-paper} 
to learn representations from unlabeled time series data. This model excels in 
situations where labeled data is scarce, offering a solution for improving classifier 
performance in both supervised and semi-supervised scenarios. The use of data augmentation 
adapted to time series enhances its ability to learn meaningful patterns. 
This work opens up new possibilities for applying self-supervised learning 
techniques to time series tasks beyond classification.
Extensive experiments on the UCR archive showcase the performance of TRILITE in two cases, 
where the dataset lacks a lot labeled samples, and when the datasets lacks a lot of samples even 
if they are labeled.

In order to highlight the practical relevance of the theoretical contributions 
made in this thesis, we explored their application in a real-world context, 
such as human motion analysis.
Hence Chapter~\ref{chapitre_6} focuses on the analysis of human movement using time 
series data captured by Kinect cameras~\cite{kinect-paper}. The model LITEMVTime, 
introduced earlier, is applied to classify the quality of human movements 
in rehabilitation contexts. Using datasets such as Kimore~\cite{kimore-paper}, the model 
provides accurate assessments of patient movements, outperforming other deep learning 
architectures.
Additionally, the chapter introduces generation pipelines for generating realistic 
human motion sequences, including a prototyping method through a novel approach we proposed 
called ShapeDBA and a CNN-based deep Supervised Variational Autoencoder (SVAE). 
Both models, capable of generating high-fidelity and diverse human motion sequences, 
are particularly beneficial for applications in the medical and entertainment domains. 
This generation capability opens the door for further exploration in areas like real-time 
motion generation in gaming and rehabilitation contexts.

In Chapter~\ref{chapitre_7}, we revisit the topic of evaluation introduced in 
Chapter~\ref{chapitre_2}, but with a specific focus on evaluating generative models.
The challenges of evaluating generative models 
for human motion are addressed. A new metric, Warping Path Diversity (WPD), 
is introduced to account for temporal distortions in generated sequences, 
complementing traditional metrics like Fréchet Inception Distance (FID)~\cite{fid-original-paper}. 
This unified evaluation framework helps in assessing both the fidelity and diversity 
of generated human movements. These metrics are essential for applications like 
video games and medical simulations, where realism and variability are crucial.

Chapter~\ref{chapitre_8} highlights the importance of reproducibility in 
machine learning research. A significant contribution here is the development 
of the open-source \emph{aeon} package~\cite{aeon-paper}, which integrates the models 
and methodologies presented throughout the thesis. This package, available to the research 
community, ensures that experiments are reproducible and that tools developed can be 
widely used and extended. By providing detailed documentation and modular code, 
this thesis promotes transparency and collaboration in time series research.

While the contributions of this research have provided strong foundations for time series 
classification and analysis, several areas remain ripe for future exploration.
In the following section we discuss some of the future perspectives for the advancements
in this field.

\section*{Discussion Of Future Works}

While the benchmarking contributions have provided a more comprehensive 
way to evaluate models, the introduction of more sophisticated statistical tools 
could refine these comparisons further. As models become increasingly complex, 
it will be essential to develop new methods for evaluating not just their 
performance, but also their scalability, robustness, and interpretability. 
Developing tools that combine the insights from MCM with other multi-objective 
optimization techniques might provide more holistic evaluation metrics.

While this thesis has laid the groundwork for foundation models in time series 
classification tasks, future work could explore extending these models to time 
series forecasting. Forecasting presents unique challenges, such as the need to
predict future data points based on past observations, which requires models to 
capture long-term dependencies effectively. Future research could investigate 
how pre-trained foundation models, like PHIT, can be adapted to handle time 
series forecasting~\cite{forecasting-foundatio}, potentially incorporating attention mechanisms to focus 
on relevant parts of the data~\cite{foundation-models-example-paper}. Another area of interest would be transfer 
learning techniques for forecasting tasks~\cite{transfer-learning-forecasting}, enabling models trained on one 
domain to generalize effectively to another, such as transitioning from medical 
time series data to financial market predictions. This extension would significantly 
broaden the applicability of foundation models in real-world scenarios.

An additional key area of future work is the development of lightweight and efficient 
models like LITE and LITEMV. These models have demonstrated great promise in low-resource 
environments, but more research is needed to optimize them for even larger and more 
complex datasets. Additionally, extending these models to other domains such as real-time 
streaming data or applications in Internet of Things (IoT)~\cite{iot-tsc} systems could unlock further potential.

Another potential direction is the extension of TRILITE and similar self-supervised 
models to other types of time series tasks beyond classification. Future research 
could explore how these models might be adapted for tasks such as time series forecasting~\cite{self-sup-forecasting}, 
anomaly detection~\cite{self-sup-anom-detection}, or even generative tasks. Moreover, there is considerable room to 
improve the data augmentation techniques used in triplet loss frameworks to further 
enhance their robustness in diverse scenarios~\cite{series2vec}.

The generative models introduced in this thesis, particularly for human motion 
generation, have shown promise in generating realistic motion sequences. However, 
future work could focus on improving the temporal diversity of these generated motions. 
Current models can generate high-quality movements but may lack variability in timing 
and execution styles. Incorporating techniques such as adversarial training or variational 
approaches could enhance the diversity of generated motions, making them more useful 
for applications like virtual reality, gaming, and medical simulations.
Furthermore, exploring multimodal data inputs, such as combining visual data with motion 
sensor data, could improve the robustness of generated sequences, allowing models 
to better mimic real-world human movements. This would broaden the scope of 
applications for human motion data generation, particularly in scenarios where 
subtle variations in motion are crucial, such as in rehabilitation exercises or 
fine motor skill training.
Regarding the Warping Path Diversity, it takes into consideration DTW based warping distortions,
however exploring the disentanglement of shape and temporal patterns~\cite{marteau2019separation}
could improve upon traditional DTW, enabling a more detailed and nuanced analysis of time series variability.

Moreover, the \emph{aeon} package's continued development~\cite{aeon-paper} should focus on incorporating 
more advanced deep learning architectures, as well as improving the ease with which 
users can contribute and extend the framework. Incorporating more interactive 
visualization tools that allow researchers to explore model outputs and hyper-parameters 
could enhance the usability of the package.

Finally, deep learning for Time Series Extrinsic Regression~\cite{tser-archive} remains an unfinished 
area of research, largely because most existing models are adaptations of architectures 
originally designed for classification tasks. While classification tasks aim to 
categorize time series data into discrete classes~\cite{bakeoff-tsc-1}, extrinsic regression deals with 
predicting continuous values, which often requires handling more nuanced relationships 
in the data. The current approach involves repurposing classification models~\cite{deep-tsc-tser}, 
such as convolutional or recurrent networks, to perform regression by simply 
changing the final layer to output continuous values. However, this overlooks 
the fundamental differences between the tasks, especially in terms of loss 
functions and evaluation metrics. No models have been specifically designed 
with the unique challenges of extrinsic regression in mind, such as effectively 
capturing long-range dependencies and handling variability in the magnitude of 
predictions. A dedicated architecture tailored for regression tasks, which could 
leverage techniques like dynamic temporal scaling or specific optimization for 
continuous outputs, would be a crucial step forward in making deep learning 
models more effective for time series extrinsic regression. This gap presents 
an opportunity for future research to innovate and address the complexities 
that are unique to regression in time series data.

These directions point to a broad set of possibilities that build upon the 
foundation laid by this thesis. By continuing to focus on model efficiency, 
self-supervised learning, benchmarking, and reproducibility, the field of 
time series analysis will be well-positioned to address both current and future challenges.

%
%
\addchap{Financing}
\label{financements}

The DELEGATION project, "DEep LEarning for Generating humAn moTION," is 
focused on developing a sophisticated deep learning framework for generating 
expressive, skeleton-based human motions. This system addresses complex 
challenges in motion analysis, such as noise and variability in human movement 
data, and aims to create realistic, controllable motion sequences. Its applications 
span areas like physical rehabilitation, where accurate motion representation is essential.

This thesis is funded under the DELEGATION project, which is supported by the 
Agence Nationale de la Recherche (ANR) under grant number ANR-21-CE23-0014. 
The financial support from ANR facilitates in-depth research into cutting-edge 
techniques in human motion generation and analysis, allowing for significant 
contributions to this emerging field.

In addition to its innovative goals, the DELEGATION project involves a strong 
collaboration between several key partners. The project is coordinated by 
Dr. Maxime Devanne from the Institut de Recherche en Informatique, 
Mathématiques, Automatique et Signal (IRIMAS) at Université de Haute-Alsace in Mulhouse, France, 
alongside Prof. Germain Forestier and Dr. Jonathan Weber. Other partners 
include the Media Integration and Communication Center (MICC) at University of Florence in Italy, 
led by Dr. Stefano Berretti as the local PI, the CHRU of Brest, France, with Prof. Olivier 
Remy-Neris as the local PI, and the Centre de Réadaptation de Mulhouse
(CRM) in Mulhouse, France, with Fabienne Ernst Kuteifan as the local PI.

A portion of the DELEGATION grant was dedicated to facilitating an academic 
visit to Dr. Stefano Berretti at the Media Integration and Communication Center (MICC) 
in Florence, Italy. I had the opportunity to spend two weeks there in June 2024, 
where I gained valuable expertise and discussed with other PhD students involved in similar 
projects. The visit was highly beneficial, fostering new ideas and approaches that 
enriched my research, while strengthening international collaboration 
within the project's framework.


\printbibliography[heading=bibintoc]

@article{elastic-ensemble,
  title     = {Time series classification with ensembles of elastic distance measures},
  author    = {Lines, Jason and Bagnall, Anthony},
  journal   = {Data Mining and Knowledge Discovery},
  volume    = {29},
  pages     = {565--592},
  year      = {2015},
  publisher = {Springer}
}

@article{ucr-archive,
  title     = {The UCR time series archive},
  author    = {Dau, Hoang Anh and Bagnall, Anthony and Kamgar, Kaveh and Yeh, Chin-Chia Michael and Zhu, Yan and Gharghabi, Shaghayegh and Ratanamahatana, Chotirat Ann and Keogh, Eamonn},
  journal   = {IEEE/CAA Journal of Automatica Sinica},
  volume    = {6},
  number    = {6},
  pages     = {1293--1305},
  year      = {2019},
  publisher = {IEEE}
}

@article{uea-archive,
  title   = {The UEA multivariate time series classification archive, 2018},
  author  = {Bagnall, Anthony and Dau, Hoang Anh and Lines, Jason and Flynn, Michael and Large, James and Bostrom, Aaron and Southam, Paul and Keogh, Eamonn},
  journal = {arXiv preprint arXiv:1811.00075},
  year    = {2018}
}

@article{bertalanivc2022resource,
  title     = {Resource-aware time series imaging classification for wireless link layer anomalies},
  author    = {Bertalani{\v{c}}, Bla{\v{z}} and Me{\v{z}}a, Marko and Fortuna, Carolina},
  journal   = {IEEE Transactions on Neural Networks and Learning Systems},
  year      = {2022},
  publisher = {IEEE}
}

@article{bakeoff-tsc-1,
  title     = {The great time series classification bake off: a review and experimental evaluation of recent algorithmic advances},
  author    = {Bagnall, Anthony and Lines, Jason and Bostrom, Aaron and Large, James and Keogh, Eamonn},
  journal   = {Data mining and knowledge discovery},
  volume    = {31},
  pages     = {606--660},
  year      = {2017},
  publisher = {Springer}
}

@article{bakeoff-tsc-2,
  title     = {Bake off redux: a review and experimental evaluation of recent time series classification algorithms},
  author    = {Middlehurst, Matthew and Sch{\"a}fer, Patrick and Bagnall, Anthony},
  journal   = {Data Mining and Knowledge Discovery},
  pages     = {1--74},
  year      = {2024},
  publisher = {Springer}
}

@article{dl4tsc,
  title     = {Deep learning for time series classification: a review},
  author    = {Ismail Fawaz, Hassan and Forestier, Germain and Weber, Jonathan and Idoumghar, Lhassane and Muller, Pierre-Alain},
  journal   = {Data mining and knowledge discovery},
  volume    = {33},
  number    = {4},
  pages     = {917--963},
  year      = {2019},
  publisher = {Springer}
}

@article{msm-distance,
  title     = {The move-split-merge metric for time series},
  author    = {Stefan, Alexandra and Athitsos, Vassilis and Das, Gautam},
  journal   = {IEEE transactions on Knowledge and Data Engineering},
  volume    = {25},
  number    = {6},
  pages     = {1425--1438},
  year      = {2012},
  publisher = {IEEE}
}

@inproceedings{soft-dtw-distance,
  title        = {Soft-dtw: a differentiable loss function for time-series},
  author       = {Cuturi, Marco and Blondel, Mathieu},
  booktitle    = {International conference on machine learning},
  pages        = {894--903},
  year         = {2017},
  organization = {PMLR}
}

@article{shape-dtw-distance,
  title     = {shapedtw: Shape dynamic time warping},
  author    = {Zhao, Jiaping and Itti, Laurent},
  journal   = {Pattern Recognition},
  volume    = {74},
  pages     = {171--184},
  year      = {2018},
  publisher = {Elsevier}
}

@article{proximity-forest,
  title     = {Proximity forest: an effective and scalable distance-based classifier for time series},
  author    = {Lucas, Benjamin and Shifaz, Ahmed and Pelletier, Charlotte and O’Neill, Lachlan and Zaidi, Nayyar and Goethals, Bart and Petitjean, Fran{\c{c}}ois and Webb, Geoffrey I},
  journal   = {Data Mining and Knowledge Discovery},
  volume    = {33},
  number    = {3},
  pages     = {607--635},
  year      = {2019},
  publisher = {Springer}
}

@article{random-forest,
  title     = {Random forests},
  author    = {Breiman, Leo},
  journal   = {Machine learning},
  volume    = {45},
  pages     = {5--32},
  year      = {2001},
  publisher = {Springer}
}

@article{proximity-forest-2,
  title   = {Proximity Forest 2.0: A new effective and scalable similarity-based classifier for time series},
  author  = {Herrmann, Matthieu and Tan, Chang Wei and Salehi, Mahsa and Webb, Geoffrey I},
  journal = {arXiv preprint arXiv:2304.05800},
  year    = {2023}
}

@article{adtw-distance,
  title     = {Amercing: an intuitive and effective constraint for dynamic time warping},
  author    = {Herrmann, Matthieu and Webb, Geoffrey I},
  journal   = {Pattern Recognition},
  volume    = {137},
  pages     = {109333},
  year      = {2023},
  publisher = {Elsevier}
}

@article{hoerl1970ridge,
  title     = {Ridge regression: Biased estimation for nonorthogonal problems},
  author    = {Hoerl, Arthur E and Kennard, Robert W},
  journal   = {Technometrics},
  volume    = {12},
  number    = {1},
  pages     = {55--67},
  year      = {1970},
  publisher = {Taylor \& Francis}
}

@article{hctsa,
  title     = {hctsa: A computational framework for automated time-series phenotyping using massive feature extraction},
  author    = {Fulcher, Ben D and Jones, Nick S},
  journal   = {Cell systems},
  volume    = {5},
  number    = {5},
  pages     = {527--531},
  year      = {2017},
  publisher = {Elsevier}
}

@article{catch22,
  title     = {catch22: CAnonical Time-series CHaracteristics: Selected through highly comparative time-series analysis},
  author    = {Lubba, Carl H and Sethi, Sarab S and Knaute, Philip and Schultz, Simon R and Fulcher, Ben D and Jones, Nick S},
  journal   = {Data Mining and Knowledge Discovery},
  volume    = {33},
  number    = {6},
  pages     = {1821--1852},
  year      = {2019},
  publisher = {Springer}
}

@article{tsfresh,
  title     = {Time series feature extraction on basis of scalable hypothesis tests (tsfresh--a python package)},
  author    = {Christ, Maximilian and Braun, Nils and Neuffer, Julius and Kempa-Liehr, Andreas W},
  journal   = {Neurocomputing},
  volume    = {307},
  pages     = {72--77},
  year      = {2018},
  publisher = {Elsevier}
}

@inproceedings{adaboost,
  title        = {Experiments with a new boosting algorithm},
  author       = {Freund, Yoav and Schapire, Robert E and others},
  booktitle    = {icml},
  volume       = {96},
  pages        = {148--156},
  year         = {1996},
  organization = {Citeseer}
}

@article{fresh,
  title   = {Distributed and parallel time series feature extraction for industrial big data applications},
  author  = {Christ, Maximilian and Kempa-Liehr, Andreas W and Feindt, Michael},
  journal = {arXiv preprint arXiv:1610.07717},
  year    = {2016}
}

@inproceedings{fresh-prince,
  title        = {The freshprince: A simple transformation based pipeline time series classifier},
  author       = {Middlehurst, Matthew and Bagnall, Anthony},
  booktitle    = {International Conference on Pattern Recognition and Artificial Intelligence},
  pages        = {150--161},
  year         = {2022},
  organization = {Springer}
}

@article{rotation-forest,
  title     = {Rotation forest: A new classifier ensemble method},
  author    = {Rodriguez, Juan Jos{\'e} and Kuncheva, Ludmila I and Alonso, Carlos J},
  journal   = {IEEE transactions on pattern analysis and machine intelligence},
  volume    = {28},
  number    = {10},
  pages     = {1619--1630},
  year      = {2006},
  publisher = {IEEE}
}

@article{lecun2015deep,
  title     = {Deep learning},
  author    = {LeCun, Yann and Bengio, Yoshua and Hinton, Geoffrey},
  journal   = {nature},
  volume    = {521},
  number    = {7553},
  pages     = {436--444},
  year      = {2015},
  publisher = {Nature Publishing Group UK London}
}

@article{rumelhart1986learning,
  title     = {Learning representations by back-propagating errors},
  author    = {Rumelhart, David E and Hinton, Geoffrey E and Williams, Ronald J},
  journal   = {nature},
  volume    = {323},
  number    = {6088},
  pages     = {533--536},
  year      = {1986},
  publisher = {Nature Publishing Group UK London}
}

@article{rocket,
  title     = {ROCKET: exceptionally fast and accurate time series classification using random convolutional kernels},
  author    = {Dempster, Angus and Petitjean, Fran{\c{c}}ois and Webb, Geoffrey I},
  journal   = {Data Mining and Knowledge Discovery},
  volume    = {34},
  number    = {5},
  pages     = {1454--1495},
  year      = {2020},
  publisher = {Springer}
}

@inproceedings{mini-rocket,
  title     = {Minirocket: A very fast (almost) deterministic transform for time series classification},
  author    = {Dempster, Angus and Schmidt, Daniel F and Webb, Geoffrey I},
  booktitle = {Proceedings of the 27th ACM SIGKDD conference on knowledge discovery \& data mining},
  pages     = {248--257},
  year      = {2021}
}

@article{multi-rocket,
  title     = {MultiRocket: multiple pooling operators and transformations for fast and effective time series classification},
  author    = {Tan, Chang Wei and Dempster, Angus and Bergmeir, Christoph and Webb, Geoffrey I},
  journal   = {Data Mining and Knowledge Discovery},
  volume    = {36},
  number    = {5},
  pages     = {1623--1646},
  year      = {2022},
  publisher = {Springer}
}

@article{hydra,
  title     = {Hydra: Competing convolutional kernels for fast and accurate time series classification},
  author    = {Dempster, Angus and Schmidt, Daniel F and Webb, Geoffrey I},
  journal   = {Data Mining and Knowledge Discovery},
  volume    = {37},
  number    = {5},
  pages     = {1779--1805},
  year      = {2023},
  publisher = {Springer}
}

@article{shapelets,
  title     = {Time series shapelets: a novel technique that allows accurate, interpretable and fast classification},
  author    = {Ye, Lexiang and Keogh, Eamonn},
  journal   = {Data mining and knowledge discovery},
  volume    = {22},
  pages     = {149--182},
  year      = {2011},
  publisher = {Springer}
}

@article{stc,
  title     = {Classification of time series by shapelet transformation},
  author    = {Hills, Jon and Lines, Jason and Baranauskas, Edgaras and Mapp, James and Bagnall, Anthony},
  journal   = {Data mining and knowledge discovery},
  volume    = {28},
  pages     = {851--881},
  year      = {2014},
  publisher = {Springer}
}

@inproceedings{rdst,
  title        = {Random dilated shapelet transform: A new approach for time series shapelets},
  author       = {Guillaume, Antoine and Vrain, Christel and Elloumi, Wael},
  booktitle    = {International Conference on Pattern Recognition and Artificial Intelligence},
  pages        = {653--664},
  year         = {2022},
  organization = {Springer}
}

@article{sax,
  title     = {Experiencing SAX: a novel symbolic representation of time series},
  author    = {Lin, Jessica and Keogh, Eamonn and Wei, Li and Lonardi, Stefano},
  journal   = {Data Mining and knowledge discovery},
  volume    = {15},
  pages     = {107--144},
  year      = {2007},
  publisher = {Springer}
}

@article{paa,
  title     = {Dimensionality reduction for fast similarity search in large time series databases},
  author    = {Keogh, Eamonn and Chakrabarti, Kaushik and Pazzani, Michael and Mehrotra, Sharad},
  journal   = {Knowledge and information Systems},
  volume    = {3},
  pages     = {263--286},
  year      = {2001},
  publisher = {Springer}
}

@inproceedings{sfa,
  title     = {SFA: a symbolic fourier approximation and index for similarity search in high dimensional datasets},
  author    = {Sch{\"a}fer, Patrick and H{\"o}gqvist, Mikael},
  booktitle = {Proceedings of the 15th international conference on extending database technology},
  pages     = {516--527},
  year      = {2012}
}

@inproceedings{sax-vsm,
  title        = {Sax-vsm: Interpretable time series classification using sax and vector space model},
  author       = {Senin, Pavel and Malinchik, Sergey},
  booktitle    = {2013 IEEE 13th international conference on data mining},
  pages        = {1175--1180},
  year         = {2013},
  organization = {IEEE}
}

@article{boss,
  title     = {The BOSS is concerned with time series classification in the presence of noise},
  author    = {Sch{\"a}fer, Patrick},
  journal   = {Data Mining and Knowledge Discovery},
  volume    = {29},
  pages     = {1505--1530},
  year      = {2015},
  publisher = {Springer}
}

@inproceedings{weasel,
  title     = {Fast and accurate time series classification with weasel},
  author    = {Sch{\"a}fer, Patrick and Leser, Ulf},
  booktitle = {Proceedings of the 2017 ACM on Conference on Information and Knowledge Management},
  pages     = {637--646},
  year      = {2017}
}

@article{weasel2,
  title     = {WEASEL 2.0: a random dilated dictionary transform for fast, accurate and memory constrained time series classification},
  author    = {Sch{\"a}fer, Patrick and Leser, Ulf},
  journal   = {Machine Learning},
  volume    = {112},
  number    = {12},
  pages     = {4763--4788},
  year      = {2023},
  publisher = {Springer}
}

@article{tsf,
  title     = {A time series forest for classification and feature extraction},
  author    = {Deng, Houtao and Runger, George and Tuv, Eugene and Vladimir, Martyanov},
  journal   = {Information Sciences},
  volume    = {239},
  pages     = {142--153},
  year      = {2013},
  publisher = {Elsevier}
}

@inproceedings{cif,
  title        = {The canonical interval forest (CIF) classifier for time series classification},
  author       = {Middlehurst, Matthew and Large, James and Bagnall, Anthony},
  booktitle    = {2020 IEEE international conference on big data (big data)},
  pages        = {188--195},
  year         = {2020},
  organization = {IEEE}
}

@article{hive-cote2.0,
  title     = {HIVE-COTE 2.0: a new meta ensemble for time series classification},
  author    = {Middlehurst, Matthew and Large, James and Flynn, Michael and Lines, Jason and Bostrom, Aaron and Bagnall, Anthony},
  journal   = {Machine Learning},
  volume    = {110},
  number    = {11},
  pages     = {3211--3243},
  year      = {2021},
  publisher = {Springer}
}

@article{quant,
  title   = {QUANT: A Minimalist Interval Method for Time Series Classification},
  author  = {Dempster, Angus and Schmidt, Daniel F and Webb, Geoffrey I},
  journal = {arXiv preprint arXiv:2308.00928},
  year    = {2023}
}

@article{extremely-randomized-trees,
  title     = {Extremely randomized trees},
  author    = {Geurts, Pierre and Ernst, Damien and Wehenkel, Louis},
  journal   = {Machine learning},
  volume    = {63},
  pages     = {3--42},
  year      = {2006},
  publisher = {Springer}
}

@article{cote,
  title     = {Time-series classification with COTE: the collective of transformation-based ensembles},
  author    = {Bagnall, Anthony and Lines, Jason and Hills, Jon and Bostrom, Aaron},
  journal   = {IEEE Transactions on Knowledge and Data Engineering},
  volume    = {27},
  number    = {9},
  pages     = {2522--2535},
  year      = {2015},
  publisher = {IEEE}
}

@article{hive-cote,
  title     = {Time series classification with HIVE-COTE: The hierarchical vote collective of transformation-based ensembles},
  author    = {Lines, Jason and Taylor, Sarah and Bagnall, Anthony},
  journal   = {ACM Transactions on Knowledge Discovery from Data (TKDD)},
  volume    = {12},
  number    = {5},
  pages     = {1--35},
  year      = {2018},
  publisher = {ACM New York, NY, USA}
}

@article{cawpe,
  title     = {A probabilistic classifier ensemble weighting scheme based on cross-validated accuracy estimates},
  author    = {Large, James and Lines, Jason and Bagnall, Anthony},
  journal   = {Data mining and knowledge discovery},
  volume    = {33},
  number    = {6},
  pages     = {1674--1709},
  year      = {2019},
  publisher = {Springer}
}

@inproceedings{hive-cote1,
  title        = {On the usage and performance of the hierarchical vote collective of transformation-based ensembles version 1.0 (hive-cote v1. 0)},
  author       = {Bagnall, Anthony and Flynn, Michael and Large, James and Lines, Jason and Middlehurst, Matthew},
  booktitle    = {Advanced Analytics and Learning on Temporal Data: 5th ECML PKDD Workshop, AALTD 2020, Ghent, Belgium, September 18, 2020, Revised Selected Papers 6},
  pages        = {3--18},
  year         = {2020},
  organization = {Springer}
}

@article{back-prop,
  title     = {Learning representations by back-propagating errors},
  author    = {Rumelhart, David E and Hinton, Geoffrey E and Williams, Ronald J},
  journal   = {nature},
  volume    = {323},
  number    = {6088},
  pages     = {533--536},
  year      = {1986},
  publisher = {Nature Publishing Group UK London}
}

@inproceedings{resnet-images,
  title     = {Deep residual learning for image recognition},
  author    = {He, Kaiming and Zhang, Xiangyu and Ren, Shaoqing and Sun, Jian},
  booktitle = {Proceedings of the IEEE conference on computer vision and pattern recognition},
  pages     = {770--778},
  year      = {2016}
}

@article{mobilenets,
  title   = {Mobilenets: Efficient convolutional neural networks for mobile vision applications},
  author  = {Howard, Andrew G and Zhu, Menglong and Chen, Bo and Kalenichenko, Dmitry and Wang, Weijun and Weyand, Tobias and Andreetto, Marco and Adam, Hartwig},
  journal = {arXiv preprint arXiv:1704.04861},
  year    = {2017}
}

@article{attention-all-you-need,
  title   = {Attention is all you need},
  author  = {Vaswani, Ashish and Shazeer, Noam and Parmar, Niki and Uszkoreit, Jakob and Jones, Llion and Gomez, Aidan N and Kaiser, {\L}ukasz and Polosukhin, Illia},
  journal = {Advances in neural information processing systems},
  volume  = {30},
  year    = {2017}
}

@article{attention-mechanism,
  title   = {Neural machine translation by jointly learning to align and translate},
  author  = {Bahdanau, Dzmitry and Cho, Kyunghyun and Bengio, Yoshua},
  journal = {arXiv preprint arXiv:1409.0473},
  year    = {2014}
}

@article{rnn-paper,
  title     = {Finding structure in time},
  author    = {Elman, Jeffrey L},
  journal   = {Cognitive science},
  volume    = {14},
  number    = {2},
  pages     = {179--211},
  year      = {1990},
  publisher = {Wiley Online Library}
}

@article{lstm-paper,
  title     = {Long short-term memory},
  author    = {Hochreiter, Sepp and Schmidhuber, J{\"u}rgen},
  journal   = {Neural computation},
  volume    = {9},
  number    = {8},
  pages     = {1735--1780},
  year      = {1997},
  publisher = {MIT press}
}

@article{gru-paper,
  title   = {Learning phrase representations using RNN encoder-decoder for statistical machine translation},
  author  = {Cho, Kyunghyun and Van Merri{\"e}nboer, Bart and Gulcehre, Caglar and Bahdanau, Dzmitry and Bougares, Fethi and Schwenk, Holger and Bengio, Yoshua},
  journal = {arXiv preprint arXiv:1406.1078},
  year    = {2014}
}

@inproceedings{graves2013speech,
  title        = {Speech recognition with deep recurrent neural networks},
  author       = {Graves, Alex and Mohamed, Abdel-rahman and Hinton, Geoffrey},
  booktitle    = {2013 IEEE international conference on acoustics, speech and signal processing},
  pages        = {6645--6649},
  year         = {2013},
  organization = {Ieee}
}

@article{seq-to-seq,
  title   = {Sequence to sequence learning with neural networks},
  author  = {Sutskever, Ilya and Vinyals, Oriol and Le, Quoc V},
  journal = {Advances in neural information processing systems},
  volume  = {27},
  year    = {2014}
}

@article{alexnet-paper,
  title   = {Imagenet classification with deep convolutional neural networks},
  author  = {Krizhevsky, Alex and Sutskever, Ilya and Hinton, Geoffrey E},
  journal = {Advances in neural information processing systems},
  volume  = {25},
  year    = {2012}
}

@inproceedings{imagenet-paper,
  title        = {Imagenet: A large-scale hierarchical image database},
  author       = {Deng, Jia and Dong, Wei and Socher, Richard and Li, Li-Jia and Li, Kai and Fei-Fei, Li},
  booktitle    = {2009 IEEE conference on computer vision and pattern recognition},
  pages        = {248--255},
  year         = {2009},
  organization = {Ieee}
}

@article{time-cnn,
  title     = {Convolutional neural networks for time series classification},
  author    = {Zhao, Bendong and Lu, Huanzhang and Chen, Shangfeng and Liu, Junliang and Wu, Dongya},
  journal   = {Journal of Systems Engineering and Electronics},
  volume    = {28},
  number    = {1},
  pages     = {162--169},
  year      = {2017},
  publisher = {BIAI}
}

@article{mlp-original-paper,
  title   = {An introduction to computational geometry},
  author  = {Minsky, Marvin and Papert, Seymour},
  journal = {Cambridge tiass., HIT},
  volume  = {479},
  number  = {480},
  pages   = {104},
  year    = {1969}
}

@inproceedings{fcn-resnet-mlp-paper,
  title        = {Time series classification from scratch with deep neural networks: A strong baseline},
  author       = {Wang, Zhiguang and Yan, Weizhong and Oates, Tim},
  booktitle    = {2017 International joint conference on neural networks (IJCNN)},
  pages        = {1578--1585},
  year         = {2017},
  organization = {IEEE}
}

@inproceedings{nne-paper,
  title        = {Deep neural network ensembles for time series classification},
  author       = {Fawaz, Hassan Ismail and Forestier, Germain and Weber, Jonathan and Idoumghar, Lhassane and Muller, Pierre-Alain},
  booktitle    = {2019 International Joint Conference on Neural Networks (IJCNN)},
  pages        = {1--6},
  year         = {2019},
  organization = {IEEE}
}

@inproceedings{encoder-paper,
  title     = {Towards a Universal Neural Network Encoder for Time Series.},
  author    = {Serra, Joan and Pascual, Santiago and Karatzoglou, Alexandros},
  booktitle = {CCIA},
  pages     = {120--129},
  year      = {2018}
}

@article{inceptiontime-paper,
  title     = {Inceptiontime: Finding alexnet for time series classification},
  author    = {Ismail Fawaz, Hassan and Lucas, Benjamin and Forestier, Germain and Pelletier, Charlotte and Schmidt, Daniel F and Weber, Jonathan and Webb, Geoffrey I and Idoumghar, Lhassane and Muller, Pierre-Alain and Petitjean, Fran{\c{c}}ois},
  journal   = {Data Mining and Knowledge Discovery},
  volume    = {34},
  number    = {6},
  pages     = {1936--1962},
  year      = {2020},
  publisher = {Springer}
}

@inproceedings{inceptionv4-paper,
  title     = {Inception-v4, inception-resnet and the impact of residual connections on learning},
  author    = {Szegedy, Christian and Ioffe, Sergey and Vanhoucke, Vincent and Alemi, Alexander},
  booktitle = {Proceedings of the AAAI conference on artificial intelligence},
  volume    = {31},
  number    = {1},
  year      = {2017}
}

@inproceedings{disjoint-cnn-paper,
  title        = {Disjoint-cnn for multivariate time series classification},
  author       = {Foumani, Seyed Navid Mohammadi and Tan, Chang Wei and Salehi, Mahsa},
  booktitle    = {2021 International Conference on Data Mining Workshops (ICDMW)},
  pages        = {760--769},
  year         = {2021},
  organization = {IEEE}
}

@article{convtran-paper,
  title     = {Improving position encoding of transformers for multivariate time series classification},
  author    = {Foumani, Navid Mohammadi and Tan, Chang Wei and Webb, Geoffrey I and Salehi, Mahsa},
  journal   = {Data Mining and Knowledge Discovery},
  volume    = {38},
  number    = {1},
  pages     = {22--48},
  year      = {2024},
  publisher = {Springer}
}

@article{relative-pe-paper1,
  title   = {Self-attention with relative position representations},
  author  = {Shaw, Peter and Uszkoreit, Jakob and Vaswani, Ashish},
  journal = {arXiv preprint arXiv:1803.02155},
  year    = {2018}
}

@article{tser-live-fuel-moisture,
  title     = {Live fuel moisture content estimation from MODIS: A deep learning approach},
  author    = {Zhu, Liujun and Webb, Geoffrey I and Yebra, Marta and Scortechini, Gianluca and Miller, Lynn and Petitjean, Fran{\c{c}}ois},
  journal   = {ISPRS Journal of Photogrammetry and Remote Sensing},
  volume    = {179},
  pages     = {81--91},
  year      = {2021},
  publisher = {Elsevier}
}

@article{heart-rate-estimattion,
  title     = {Toward a robust estimation of respiratory rate from pulse oximeters},
  author    = {Pimentel, Marco AF and Johnson, Alistair EW and Charlton, Peter H and Birrenkott, Drew and Watkinson, Peter J and Tarassenko, Lionel and Clifton, David A},
  journal   = {IEEE Transactions on Biomedical Engineering},
  volume    = {64},
  number    = {8},
  pages     = {1914--1923},
  year      = {2016},
  publisher = {IEEE}
}

@article{tser-archive,
  title   = {Monash University, UEA, UCR time series extrinsic regression archive},
  author  = {Tan, Chang Wei and Bergmeir, Christoph and Petitjean, Francois and Webb, Geoffrey I},
  journal = {arXiv preprint arXiv:2006.10996},
  year    = {2020}
}

@article{bakeoff-tser,
  title     = {Unsupervised feature based algorithms for time series extrinsic regression},
  author    = {Guijo-Rubio, David and Middlehurst, Matthew and Arcencio, Guilherme and Silva, Diego Furtado and Bagnall, Anthony},
  journal   = {Data Mining and Knowledge Discovery},
  pages     = {1--45},
  year      = {2024},
  publisher = {Springer}
}

@article{cls-and-res-tree,
  title     = {Classification and regression trees},
  author    = {Loh, Wei-Yin},
  journal   = {Wiley interdisciplinary reviews: data mining and knowledge discovery},
  volume    = {1},
  number    = {1},
  pages     = {14--23},
  year      = {2011},
  publisher = {Wiley Online Library}
}

@inproceedings{eamonn-prototyping,
  title     = {An enhanced representation of time series which allows fast and accurate classification, clustering and relevance feedback.},
  author    = {Keogh, Eamonn J and Pazzani, Michael J},
  booktitle = {Kdd},
  volume    = {98},
  pages     = {239--243},
  year      = {1998}
}

@article{dba-paper,
  title     = {A global averaging method for dynamic time warping, with applications to clustering},
  author    = {Petitjean, Fran{\c{c}}ois and Ketterlin, Alain and Gan{\c{c}}arski, Pierre},
  journal   = {Pattern recognition},
  volume    = {44},
  number    = {3},
  pages     = {678--693},
  year      = {2011},
  publisher = {Elsevier}
}

@inproceedings{mba-paper,
  title     = {Barycentre averaging for the move-split-merge time series distance measure},
  author    = {Holder, Christopher and Guijo-Rubio, David and Bagnall, Anthony},
  booktitle = {15th International Joint Conference on Knowledge Discovery, Knowledge Engineering and Knowledge Management},
  year      = {2023}
}

@article{elastic-clustering-review,
  title     = {A review and evaluation of elastic distance functions for time series clustering},
  author    = {Holder, Christopher and Middlehurst, Matthew and Bagnall, Anthony},
  journal   = {Knowledge and Information Systems},
  volume    = {66},
  number    = {2},
  pages     = {765--809},
  year      = {2024},
  publisher = {Springer}
}

@article{msm-paper,
  title     = {The move-split-merge metric for time series},
  author    = {Stefan, Alexandra and Athitsos, Vassilis and Das, Gautam},
  journal   = {IEEE transactions on Knowledge and Data Engineering},
  volume    = {25},
  number    = {6},
  pages     = {1425--1438},
  year      = {2012},
  publisher = {IEEE}
}

@inproceedings{neares-centroid-dba-paper,
  title        = {Dynamic time warping averaging of time series allows faster and more accurate classification},
  author       = {Petitjean, Fran{\c{c}}ois and Forestier, Germain and Webb, Geoffrey I and Nicholson, Ann E and Chen, Yanping and Keogh, Eamonn},
  booktitle    = {2014 IEEE international conference on data mining},
  pages        = {470--479},
  year         = {2014},
  organization = {IEEE}
}

@inproceedings{kmeans-paper,
  title        = {Some methods for classification and analysis of multivariate observations},
  author       = {MacQueen, James and others},
  booktitle    = {Proceedings of the fifth Berkeley symposium on mathematical statistics and probability},
  volume       = {1},
  number       = {14},
  pages        = {281--297},
  year         = {1967},
  organization = {Oakland, CA, USA}
}

@inproceedings{kshape-paper,
  title     = {k-shape: Efficient and accurate clustering of time series},
  author    = {Paparrizos, John and Gravano, Luis},
  booktitle = {Proceedings of the 2015 ACM SIGMOD international conference on management of data},
  pages     = {1855--1870},
  year      = {2015}
}

@book{golub2013matrix,
  title     = {Matrix computations},
  author    = {Golub, Gene H and Van Loan, Charles F},
  year      = {2013},
  publisher = {JHU press}
}

@article{deep-tscl-bakeoff,
  title     = {End-to-end deep representation learning for time series clustering: a comparative study},
  author    = {Lafabregue, Baptiste and Weber, Jonathan and Gan{\c{c}}arski, Pierre and Forestier, Germain},
  journal   = {Data mining and knowledge discovery},
  volume    = {36},
  number    = {1},
  pages     = {29--81},
  year      = {2022},
  publisher = {Springer}
}

@article{triplet-loss-paper,
  title   = {Unsupervised scalable representation learning for multivariate time series},
  author  = {Franceschi, Jean-Yves and Dieuleveut, Aymeric and Jaggi, Martin},
  journal = {Advances in neural information processing systems},
  volume  = {32},
  year    = {2019}
}

@inproceedings{multi-rec-paper,
  title     = {Deep clustering via joint convolutional autoencoder embedding and relative entropy minimization},
  author    = {Ghasedi Dizaji, Kamran and Herandi, Amirhossein and Deng, Cheng and Cai, Weidong and Huang, Heng},
  booktitle = {Proceedings of the IEEE international conference on computer vision},
  pages     = {5736--5745},
  year      = {2017}
}

@article{drnn-paper,
  title   = {Learning representations for time series clustering},
  author  = {Ma, Qianli and Zheng, Jiawei and Li, Sen and Cottrell, Gary W},
  journal = {Advances in neural information processing systems},
  volume  = {32},
  year    = {2019}
}

@inproceedings{clustering-loss-paper,
  title        = {Unsupervised deep embedding for clustering analysis},
  author       = {Xie, Junyuan and Girshick, Ross and Farhadi, Ali},
  booktitle    = {International conference on machine learning},
  pages        = {478--487},
  year         = {2016},
  organization = {PMLR}
}

@article{kramer1991nonlinear,
  title     = {Nonlinear principal component analysis using autoassociative neural networks},
  author    = {Kramer, Mark A},
  journal   = {AIChE journal},
  volume    = {37},
  number    = {2},
  pages     = {233--243},
  year      = {1991},
  publisher = {Wiley Online Library}
}

@article{vae-paper,
  title   = {Auto-encoding variational bayes},
  author  = {Kingma, Diederik P and Welling, Max},
  journal = {arXiv preprint arXiv:1312.6114},
  year    = {2013}
}

@article{siamese-networks,
  title   = {Signature verification using a" siamese" time delay neural network},
  author  = {Bromley, Jane and Guyon, Isabelle and LeCun, Yann and S{\"a}ckinger, Eduard and Shah, Roopak},
  journal = {Advances in neural information processing systems},
  volume  = {6},
  year    = {1993}
}

@inproceedings{facenet-paper,
  title     = {Facenet: A unified embedding for face recognition and clustering},
  author    = {Schroff, Florian and Kalenichenko, Dmitry and Philbin, James},
  booktitle = {Proceedings of the IEEE conference on computer vision and pattern recognition},
  pages     = {815--823},
  year      = {2015}
}

@article{mixing-up-paper,
  title     = {Mixing up contrastive learning: Self-supervised representation learning for time series},
  author    = {Wickstr{\o}m, Kristoffer and Kampffmeyer, Michael and Mikalsen, Karl {\O}yvind and Jenssen, Robert},
  journal   = {Pattern Recognition Letters},
  volume    = {155},
  pages     = {54--61},
  year      = {2022},
  publisher = {Elsevier}
}

@article{time-clr-paper,
  title     = {Timeclr: A self-supervised contrastive learning framework for univariate time series representation},
  author    = {Yang, Xinyu and Zhang, Zhenguo and Cui, Rongyi},
  journal   = {Knowledge-Based Systems},
  volume    = {245},
  pages     = {108606},
  year      = {2022},
  publisher = {Elsevier}
}

@inproceedings{simCLR-paper,
  title        = {A simple framework for contrastive learning of visual representations},
  author       = {Chen, Ting and Kornblith, Simon and Norouzi, Mohammad and Hinton, Geoffrey},
  booktitle    = {International conference on machine learning},
  pages        = {1597--1607},
  year         = {2020},
  organization = {PMLR}
}

@article{demsar-cdd-paper,
  title     = {Statistical comparisons of classifiers over multiple data sets},
  author    = {Dem{\v{s}}ar, Janez},
  journal   = {The Journal of Machine learning research},
  volume    = {7},
  pages     = {1--30},
  year      = {2006},
  publisher = {JMLR. org}
}

@article{mcm-paper,
  title   = {An approach to multiple comparison benchmark evaluations that is stable under manipulation of the comparate set},
  author  = {Ismail-Fawaz, Ali and Dempster, Angus and Tan, Chang Wei and Herrmann, Matthieu and Miller, Lynn and Schmidt, Daniel F and Berretti, Stefano and Weber, Jonathan and Devanne, Maxime and Forestier, Germain and others},
  journal = {arXiv preprint arXiv:2305.11921},
  year    = {2023}
}

@article{demsar-extension-paper,
  title   = {An Extension on" Statistical Comparisons of Classifiers over Multiple Data Sets" for all Pairwise Comparisons.},
  author  = {Garcia, Salvador and Herrera, Francisco},
  journal = {Journal of machine learning research},
  volume  = {9},
  number  = {12},
  year    = {2008}
}

@article{bayseian-benavoli-paper,
  title   = {Time for a change: a tutorial for comparing multiple classifiers through Bayesian analysis},
  author  = {Benavoli, Alessio and Corani, Giorgio and Dem{\v{s}}ar, Janez and Zaffalon, Marco},
  journal = {Journal of Machine Learning Research},
  volume  = {18},
  number  = {77},
  pages   = {1--36},
  year    = {2017}
}

@article{pvalues-pitfal-paper,
  title     = {Using p-values for the comparison of classifiers: pitfalls and alternatives},
  author    = {Berrar, Daniel},
  journal   = {Data Mining and Knowledge Discovery},
  volume    = {36},
  number    = {3},
  pages     = {1102--1139},
  year      = {2022},
  publisher = {Springer}
}

@article{cdd-benavoli-paper,
  title     = {Should we really use post-hoc tests based on mean-ranks?},
  author    = {Benavoli, Alessio and Corani, Giorgio and Mangili, Francesca},
  journal   = {The Journal of Machine Learning Research},
  volume    = {17},
  number    = {1},
  pages     = {152--161},
  year      = {2016},
  publisher = {JMLR. org}
}

@incollection{wilcoxon-paper,
  title     = {Individual comparisons by ranking methods},
  author    = {Wilcoxon, Frank},
  booktitle = {Breakthroughs in statistics: Methodology and distribution},
  pages     = {196--202},
  year      = {1992},
  publisher = {Springer}
}

@book{nemenyi-paper,
  title     = {Distribution-free multiple comparisons.},
  author    = {Nemenyi, Peter Bjorn},
  year      = {1963},
  publisher = {Princeton University}
}

@article{friedman-paper,
  title     = {A comparison of alternative tests of significance for the problem of m rankings},
  author    = {Friedman, Milton},
  journal   = {The annals of mathematical statistics},
  volume    = {11},
  number    = {1},
  pages     = {86--92},
  year      = {1940},
  publisher = {JSTOR}
}

@article{holm-correction-paper,
  title     = {A simple sequentially rejective multiple test procedure},
  author    = {Holm, Sture},
  journal   = {Scandinavian journal of statistics},
  pages     = {65--70},
  year      = {1979},
  publisher = {JSTOR}
}

@book{lecoutre2014significance,
  title     = {The significance test controversy revisited},
  author    = {Lecoutre, Bruno and Poitevineau, Jacques and Lecoutre, Bruno and Poitevineau, Jacques},
  year      = {2014},
  publisher = {Springer}
}

@inproceedings{hand-crafted-filters-paper,
  title        = {Deep learning for time series classification using new hand-crafted convolution filters},
  author       = {Ismail-Fawaz, Ali and Devanne, Maxime and Weber, Jonathan and Forestier, Germain},
  booktitle    = {2022 IEEE International Conference on Big Data (Big Data)},
  pages        = {972--981},
  year         = {2022},
  organization = {IEEE}
}

@inproceedings{pretext-task-paper,
  title        = {Finding foundation models for time series classification with a pretext task},
  author       = {Ismail-Fawaz, Ali and Devanne, Maxime and Berretti, Stefano and Weber, Jonathan and Forestier, Germain},
  booktitle    = {Pacific-Asia Conference on Knowledge Discovery and Data Mining},
  pages        = {123--135},
  year         = {2024},
  organization = {Springer}
}

@article{sobel-paper1,
  title   = {Custom extended sobel filters},
  author  = {Bogdan, Victor and Bonchi{\c{s}}, Cosmin and Orhei, Ciprian},
  journal = {arXiv preprint arXiv:1910.00138},
  year    = {2019}
}

@inproceedings{sobel-paper2,
  title        = {An improved Sobel edge detection},
  author       = {Gao, Wenshuo and Zhang, Xiaoguang and Yang, Lei and Liu, Huizhong},
  booktitle    = {2010 3rd International conference on computer science and information technology},
  volume       = {5},
  pages        = {67--71},
  year         = {2010},
  organization = {IEEE}
}

@article{t-sne-paper,
  title   = {Visualizing data using t-SNE.},
  author  = {Van der Maaten, Laurens and Hinton, Geoffrey},
  journal = {Journal of machine learning research},
  volume  = {9},
  number  = {11},
  year    = {2008}
}

@inproceedings{transfer-learning-paper,
  title        = {Transfer learning for time series classification},
  author       = {Fawaz, Hassan Ismail and Forestier, Germain and Weber, Jonathan and Idoumghar, Lhassane and Muller, Pierre-Alain},
  booktitle    = {2018 IEEE international conference on big data (Big Data)},
  pages        = {1367--1376},
  year         = {2018},
  organization = {IEEE}
}

@misc{tensorflow-paper,
  title     = {TensorFlow: Large-scale machine learning on heterogeneous systems},
  author    = {Abadi, Mart{\'\i}n and Agarwal, Ashish and Barham, Paul and Brevdo, Eugene and Chen, Zhifeng and Citro, Craig and Corrado, Greg S and Davis, Andy and Dean, Jeffrey and Devin, Matthieu and others},
  year      = {2015},
  publisher = {Mountain View, CA: Tensorflow}
}

@book{keras-book,
  title     = {Deep learning with Python},
  author    = {Chollet, Francois},
  year      = {2021},
  publisher = {Simon and Schuster}
}

@inproceedings{lite-paper,
  title        = {Lite: Light inception with boosting techniques for time series classification},
  author       = {Ismail-Fawaz, Ali and Devanne, Maxime and Berretti, Stefano and Weber, Jonathan and Forestier, Germain},
  booktitle    = {2023 IEEE 10th International Conference on Data Science and Advanced Analytics (DSAA)},
  pages        = {1--10},
  year         = {2023},
  organization = {IEEE}
}

@software{codecarbon,
  author    = {Benoit Courty and
               Victor Schmidt and
               Sasha Luccioni and
               Goyal-Kamal and
               MarionCoutarel and
               Boris Feld and
               Jérémy Lecourt and
               LiamConnell and
               Amine Saboni and
               Inimaz and
               supatomic and
               Mathilde Léval and
               Luis Blanche and
               Alexis Cruveiller and
               ouminasara and
               Franklin Zhao and
               Aditya Joshi and
               Alexis Bogroff and
               Hugues de Lavoreille and
               Niko Laskaris and
               Edoardo Abati and
               Douglas Blank and
               Ziyao Wang and
               Armin Catovic and
               Marc Alencon and
               Michał Stęchły and
               Christian Bauer and
               Lucas-Otavio and
               JPW and
               MinervaBooks},
  title     = {mlco2/codecarbon: v2.4.1},
  month     = may,
  year      = 2024,
  publisher = {Zenodo},
  version   = {v2.4.1},
  doi       = {10.5281/zenodo.11171501},
  url       = {https://doi.org/10.5281/zenodo.11171501}
}

@inproceedings{action2motion-paper,
  title     = {Action2motion: Conditioned generation of 3d human motions},
  author    = {Guo, Chuan and Zuo, Xinxin and Wang, Sen and Zou, Shihao and Sun, Qingyao and Deng, Annan and Gong, Minglun and Cheng, Li},
  booktitle = {Proceedings of the 28th ACM International Conference on Multimedia},
  pages     = {2021--2029},
  year      = {2020}
}

@article{kimore-paper,
  title     = {The kimore dataset: Kinematic assessment of movement and clinical scores for remote monitoring of physical rehabilitation},
  author    = {Capecci, Marianna and Ceravolo, Maria Gabriella and Ferracuti, Francesco and Iarlori, Sabrina and Monteriu, Andrea and Romeo, Luca and Verdini, Federica},
  journal   = {IEEE Transactions on Neural Systems and Rehabilitation Engineering},
  volume    = {27},
  number    = {7},
  pages     = {1436--1448},
  year      = {2019},
  publisher = {IEEE}
}

@inproceedings{actor-paper,
  title     = {Action-conditioned 3d human motion synthesis with transformer vae},
  author    = {Petrovich, Mathis and Black, Michael J and Varol, G{\"u}l},
  booktitle = {Proceedings of the IEEE/CVF International Conference on Computer Vision},
  pages     = {10985--10995},
  year      = {2021}
}

@inproceedings{hm-prediction-paper,
  title     = {On human motion prediction using recurrent neural networks},
  author    = {Martinez, Julieta and Black, Michael J and Romero, Javier},
  booktitle = {Proceedings of the IEEE conference on computer vision and pattern recognition},
  pages     = {2891--2900},
  year      = {2017}
}

@article{kinect-survey-paper,
  title     = {A survey of applications and human motion recognition with microsoft kinect},
  author    = {Lun, Roanna and Zhao, Wenbing},
  journal   = {International Journal of Pattern Recognition and Artificial Intelligence},
  volume    = {29},
  number    = {05},
  pages     = {1555008},
  year      = {2015},
  publisher = {World Scientific}
}

@inproceedings{kinect-paper,
  title     = {Estimating human motion from multiple kinect sensors},
  author    = {Asteriadis, Stylianos and Chatzitofis, Anargyros and Zarpalas, Dimitrios and Alexiadis, Dimitrios S and Daras, Petros},
  booktitle = {Int. Conf. on Computer Vision/Computer Graphics Collaboration Techniques and Applications},
  pages     = {1--6},
  year      = {2013}
}

@article{mocap-paper,
  title     = {Practical motion capture in everyday surroundings},
  author    = {Vlasic, Daniel and Adelsberger, Rolf and Vannucci, Giovanni and Barnwell, John and Gross, Markus and Matusik, Wojciech and Popovi{\'c}, Jovan},
  journal   = {ACM transactions on graphics (TOG)},
  volume    = {26},
  number    = {3},
  pages     = {35--es},
  year      = {2007},
  publisher = {Acm New York, NY, USA}
}

@inproceedings{sensors-activity-recog,
  title     = {Human activity recognition using wearable sensors by deep convolutional neural networks},
  author    = {Jiang, Wenchao and Yin, Zhaozheng},
  booktitle = {Proceedings of the 23rd ACM international conference on Multimedia},
  pages     = {1307--1310},
  year      = {2015}
}

@article{gait-analysis-wearable,
  title     = {Gait analysis methods: An overview of wearable and non-wearable systems, highlighting clinical applications},
  author    = {Muro-De-La-Herran, Alvaro and Garcia-Zapirain, Begonya and Mendez-Zorrilla, Amaia},
  journal   = {Sensors},
  volume    = {14},
  number    = {2},
  pages     = {3362--3394},
  year      = {2014},
  publisher = {Molecular Diversity Preservation International (MDPI)}
}

@article{activity-recog-wearable,
  title     = {Deep learning algorithms for human activity recognition using mobile and wearable sensor networks: State of the art and research challenges},
  author    = {Nweke, Henry Friday and Teh, Ying Wah and Al-Garadi, Mohammed Ali and Alo, Uzoma Rita},
  journal   = {Expert Systems with Applications},
  volume    = {105},
  pages     = {233--261},
  year      = {2018},
  publisher = {Elsevier}
}

@inproceedings{lime-paper,
  title     = {" Why should i trust you?" Explaining the predictions of any classifier},
  author    = {Ribeiro, Marco Tulio and Singh, Sameer and Guestrin, Carlos},
  booktitle = {Proceedings of the 22nd ACM SIGKDD international conference on knowledge discovery and data mining},
  pages     = {1135--1144},
  year      = {2016}
}

@article{scipy-paper,
  title     = {SciPy 1.0: fundamental algorithms for scientific computing in Python},
  author    = {Virtanen, Pauli and Gommers, Ralf and Oliphant, Travis E and Haberland, Matt and Reddy, Tyler and Cournapeau, David and Burovski, Evgeni and Peterson, Pearu and Weckesser, Warren and Bright, Jonathan and others},
  journal   = {Nature methods},
  volume    = {17},
  number    = {3},
  pages     = {261--272},
  year      = {2020},
  publisher = {Nature Publishing Group}
}

@article{explainability-tsc,
  title     = {Explainable AI for time series classification: a review, taxonomy and research directions},
  author    = {Theissler, Andreas and Spinnato, Francesco and Schlegel, Udo and Guidotti, Riccardo},
  journal   = {Ieee Access},
  volume    = {10},
  pages     = {100700--100724},
  year      = {2022},
  publisher = {IEEE}
}

@inproceedings{original-cam-paper,
  title     = {Learning deep features for discriminative localization},
  author    = {Zhou, Bolei and Khosla, Aditya and Lapedriza, Agata and Oliva, Aude and Torralba, Antonio},
  booktitle = {Proceedings of the IEEE conference on computer vision and pattern recognition},
  pages     = {2921--2929},
  year      = {2016}
}

@inproceedings{weighted-dba-paper,
  title        = {Generating synthetic time series to augment sparse datasets},
  author       = {Forestier, Germain and Petitjean, Fran{\c{c}}ois and Dau, Hoang Anh and Webb, Geoffrey I and Keogh, Eamonn},
  booktitle    = {2017 IEEE international conference on data mining (ICDM)},
  pages        = {865--870},
  year         = {2017},
  organization = {IEEE}
}

@inproceedings{shape-dba-paper,
  author    = {Ismail-Fawaz, Ali and Ismail Fawaz, Hassan and Petitjean, François and Devanne, Maxime and Weber, Jonathan and Berretti, Stefano and Webb, Geoffrey I. and Forestier, Germain},
  title     = {ShapeDBA: Generating Effective Time Series Prototypes using ShapeDTW Barycenter Averaging},
  booktitle = {ECML/PKDD Workshop on Advanced Analytics and Learning on Temporal Data},
  city      = {Turin},
  country   = {Italy},
  pages     = {1--8},
  url       = {https://doi.org/10.1007/978-3-031-49896-1_9},
  year      = {2023}
}

@inproceedings{prototyping-explainability,
  title        = {Explaining deep classification of time-series data with learned prototypes},
  author       = {Gee, Alan H and Garcia-Olano, Diego and Ghosh, Joydeep and Paydarfar, David},
  booktitle    = {CEUR workshop proceedings},
  volume       = {2429},
  pages        = {15},
  year         = {2019},
  organization = {NIH Public Access}
}

@article{ari-paper,
  title     = {Comparing partitions},
  author    = {Hubert, Lawrence and Arabie, Phipps},
  journal   = {Journal of classification},
  volume    = {2},
  pages     = {193--218},
  year      = {1985},
  publisher = {Springer}
}

@article{human-motion-data-aug-noise,
  title     = {Enhancing Human Action Recognition with 3D Skeleton Data: A Comprehensive Study of Deep Learning and Data Augmentation},
  author    = {Xin, Chu and Kim, Seokhwan and Cho, Yongjoo and Park, Kyoung Shin},
  journal   = {Electronics},
  volume    = {13},
  number    = {4},
  pages     = {747},
  year      = {2024},
  publisher = {MDPI}
}

@inproceedings{reliable-fidelity-diversity,
  title        = {Reliable fidelity and diversity metrics for generative models},
  author       = {Naeem, Muhammad Ferjad and Oh, Seong Joon and Uh, Youngjung and Choi, Yunjey and Yoo, Jaejun},
  booktitle    = {International Conference on Machine Learning},
  pages        = {7176--7185},
  year         = {2020},
  organization = {PMLR}
}

@article{regression-metrics-paper,
  title     = {A Hidden Semi-Markov Model based approach for rehabilitation exercise assessment},
  author    = {Capecci, Marianna and Ceravolo, Maria Gabriella and Ferracuti, Francesco and Iarlori, Sabrina and Kyrki, Ville and Monteriu, Andrea and Romeo, Luca and Verdini, Federica},
  journal   = {Journal of biomedical informatics},
  volume    = {78},
  pages     = {1--11},
  year      = {2018},
  publisher = {Elsevier}
}

@inproceedings{gta-dataset-paper,
  title        = {Long-term human motion prediction with scene context},
  author       = {Cao, Zhe and Gao, Hang and Mangalam, Karttikeya and Cai, Qi-Zhi and Vo, Minh and Malik, Jitendra},
  booktitle    = {Computer Vision--ECCV 2020: 16th European Conference, Glasgow, UK, August 23--28, 2020, Proceedings, Part I 16},
  pages        = {387--404},
  year         = {2020},
  organization = {Springer}
}

@article{generating-cut-scenes-paper,
  title     = {Cine-AI: Generating video game cutscenes in the style of human directors},
  author    = {Evin, Inan and H{\"a}m{\"a}l{\"a}inen, Perttu and Guckelsberger, Christian},
  journal   = {Proceedings of the ACM on Human-Computer Interaction},
  volume    = {6},
  number    = {CHI PLAY},
  pages     = {1--23},
  year      = {2022},
  publisher = {ACM New York, NY, USA}
}

@article{rehab-generation-paper,
  title     = {Generative adversarial networks for generation and classification of physical rehabilitation movement episodes},
  author    = {Li, Longze and Vakanski, Aleksandar},
  journal   = {International journal of machine learning and computing},
  volume    = {8},
  number    = {5},
  pages     = {428},
  year      = {2018},
  publisher = {NIH Public Access}
}

@inproceedings{motiongpt-fine-tune-paper,
  title     = {Motiongpt: Finetuned llms are general-purpose motion generators},
  author    = {Zhang, Yaqi and Huang, Di and Liu, Bin and Tang, Shixiang and Lu, Yan and Chen, Lu and Bai, Lei and Chu, Qi and Yu, Nenghai and Ouyang, Wanli},
  booktitle = {Proceedings of the AAAI Conference on Artificial Intelligence},
  volume    = {38},
  number    = {7},
  pages     = {7368--7376},
  year      = {2024}
}

@article{gan-paper,
  title   = {Generative adversarial nets},
  author  = {Goodfellow, Ian and Pouget-Abadie, Jean and Mirza, Mehdi and Xu, Bing and Warde-Farley, David and Ozair, Sherjil and Courville, Aaron and Bengio, Yoshua},
  journal = {Advances in neural information processing systems},
  volume  = {27},
  year    = {2014}
}

@inproceedings{two-stage-gan-paper,
  title     = {Deep video generation, prediction and completion of human action sequences},
  author    = {Cai, Haoye and Bai, Chunyan and Tai, Yu-Wing and Tang, Chi-Keung},
  booktitle = {Proceedings of the European conference on computer vision (ECCV)},
  pages     = {366--382},
  year      = {2018}
}

@inproceedings{mocogan-paper,
  title     = {Mocogan: Decomposing motion and content for video generation},
  author    = {Tulyakov, Sergey and Liu, Ming-Yu and Yang, Xiaodong and Kautz, Jan},
  booktitle = {Proceedings of the IEEE conference on computer vision and pattern recognition},
  pages     = {1526--1535},
  year      = {2018}
}

@inproceedings{PHSPD-dataset-paper,
  title        = {3D human shape reconstruction from a polarization image},
  author       = {Zou, Shihao and Zuo, Xinxin and Qian, Yiming and Wang, Sen and Xu, Chi and Gong, Minglun and Cheng, Li},
  booktitle    = {Computer Vision--ECCV 2020: 16th European Conference, Glasgow, UK, August 23--28, 2020, Proceedings, Part XIV 16},
  pages        = {351--368},
  year         = {2020},
  organization = {Springer}
}

@inproceedings{LCP-VAE-paper,
  title     = {Contextually plausible and diverse 3d human motion prediction},
  author    = {Aliakbarian, Sadegh and Saleh, Fatemeh and Petersson, Lars and Gould, Stephen and Salzmann, Mathieu},
  booktitle = {Proceedings of the IEEE/CVF International Conference on Computer Vision},
  pages     = {11333--11342},
  year      = {2021}
}

@article{vision-transformer-paper,
  title   = {An image is worth 16x16 words: Transformers for image recognition at scale},
  author  = {Dosovitskiy, Alexey and Beyer, Lucas and Kolesnikov, Alexander and Weissenborn, Dirk and Zhai, Xiaohua and Unterthiner, Thomas and Dehghani, Mostafa and Minderer, Matthias and Heigold, Georg and Gelly, Sylvain and others},
  journal = {arXiv preprint arXiv:2010.11929},
  year    = {2020}
}

@inproceedings{posegpt-paper,
  title        = {Posegpt: Quantization-based 3d human motion generation and forecasting},
  author       = {Lucas, Thomas and Baradel, Fabien and Weinzaepfel, Philippe and Rogez, Gr{\'e}gory},
  booktitle    = {European Conference on Computer Vision},
  pages        = {417--435},
  year         = {2022},
  organization = {Springer}
}

@article{vq-vae-paper,
  title   = {Neural discrete representation learning},
  author  = {Van Den Oord, Aaron and Vinyals, Oriol and others},
  journal = {Advances in neural information processing systems},
  volume  = {30},
  year    = {2017}
}

@article{gpt-1-paper,
  title     = {Improving language understanding by generative pre-training},
  author    = {Radford, Alec and Narasimhan, Karthik and Salimans, Tim and Sutskever, Ilya and others},
  year      = {2018},
  publisher = {San Francisco, CA, USA}
}

@inproceedings{um-cvae-paper,
  title        = {Learning uncoupled-modulation cvae for 3d action-conditioned human motion synthesis},
  author       = {Zhong, Chongyang and Hu, Lei and Zhang, Zihao and Xia, Shihong},
  booktitle    = {European Conference on Computer Vision},
  pages        = {716--732},
  year         = {2022},
  organization = {Springer}
}

@article{ddpm-paper,
  title   = {Denoising diffusion probabilistic models},
  author  = {Ho, Jonathan and Jain, Ajay and Abbeel, Pieter},
  journal = {Advances in neural information processing systems},
  volume  = {33},
  pages   = {6840--6851},
  year    = {2020}
}

@inproceedings{u-net-paper,
  title        = {U-net: Convolutional networks for biomedical image segmentation},
  author       = {Ronneberger, Olaf and Fischer, Philipp and Brox, Thomas},
  booktitle    = {Medical image computing and computer-assisted intervention--MICCAI 2015: 18th international conference, Munich, Germany, October 5-9, 2015, proceedings, part III 18},
  pages        = {234--241},
  year         = {2015},
  organization = {Springer}
}

@article{motion-diffuse-paper,
  title     = {Motiondiffuse: Text-driven human motion generation with diffusion model},
  author    = {Zhang, Mingyuan and Cai, Zhongang and Pan, Liang and Hong, Fangzhou and Guo, Xinying and Yang, Lei and Liu, Ziwei},
  journal   = {IEEE Transactions on Pattern Analysis and Machine Intelligence},
  year      = {2024},
  publisher = {IEEE}
}

@inproceedings{svae-paper,
  author    = {Ismail-Fawaz, Ali and Devanne, Maxime and Berretti, Stefano and Weber, Jonathan and Forestier, Germain},
  title     = {A Supervised Variational Auto-Encoder for Human Motion Generation using Convolutional Neural Networks},
  booktitle = {4th International Conference on Pattern Recognition and Artificial Intelligence (ICPRAI)},
  city      = {Jeju Island},
  country   = {Korea},
  year      = {2024}
}

@article{beta-vae-paper,
  title   = {beta-vae: Learning basic visual concepts with a constrained variational framework.},
  author  = {Higgins, Irina and Matthey, Loic and Pal, Arka and Burgess, Christopher P and Glorot, Xavier and Botvinick, Matthew M and Mohamed, Shakir and Lerchner, Alexander},
  journal = {ICLR (Poster)},
  volume  = {3},
  year    = {2017}
}

@article{mos-paper,
  title     = {Mean opinion score (MOS) revisited: methods and applications,  limitations and alternatives},
  volume    = {22},
  issn      = {1432-1882},
  url       = {http://dx.doi.org/10.1007/s00530-014-0446-1},
  doi       = {10.1007/s00530-014-0446-1},
  number    = {2},
  journal   = {Multimedia Systems},
  publisher = {Springer Science and Business Media LLC},
  author    = {Streijl,  Robert C. and Winkler,  Stefan and Hands,  David S.},
  year      = {2014},
  pages     = {213-227}
}

@article{fid-original-paper,
  title   = {Gans trained by a two time-scale update rule converge to a local nash equilibrium},
  author  = {Heusel, Martin and Ramsauer, Hubert and Unterthiner, Thomas and Nessler, Bernhard and Hochreiter, Sepp},
  journal = {Advances in neural information processing systems},
  volume  = {30},
  year    = {2017}
}

@article{is-paper,
  title   = {Improved techniques for training gans},
  author  = {Salimans, Tim and Goodfellow, Ian and Zaremba, Wojciech and Cheung, Vicki and Radford, Alec and Chen, Xi},
  journal = {Advances in neural information processing systems},
  volume  = {29},
  year    = {2016}
}

@article{fd-paper,
  title  = {Sur quelques points du calcul fonctionnel},
  author = {Fr{\'e}chet, Maurice},
  year   = {1906}
}

@inproceedings{fd-dist-paper,
  title     = {Sur la distance de deux lois de probabilit{\'e}},
  author    = {Fr{\'e}chet, Maurice},
  booktitle = {Annales de l'ISUP},
  volume    = {6},
  number    = {3},
  pages     = {183--198},
  year      = {1957}
}

@article{fid-gaus-paper,
  title     = {The Fr{\'e}chet distance between multivariate normal distributions},
  author    = {Dowson, DC and Landau, BV666017},
  journal   = {Journal of multivariate analysis},
  volume    = {12},
  number    = {3},
  pages     = {450--455},
  year      = {1982},
  publisher = {Elsevier}
}

@article{precision-recall-paper,
  title   = {Assessing generative models via precision and recall},
  author  = {Sajjadi, Mehdi SM and Bachem, Olivier and Lucic, Mario and Bousquet, Olivier and Gelly, Sylvain},
  journal = {Advances in neural information processing systems},
  volume  = {31},
  year    = {2018}
}

@article{improved-precision-recall-paper,
  title   = {Improved precision and recall metric for assessing generative models},
  author  = {Kynk{\"a}{\"a}nniemi, Tuomas and Karras, Tero and Laine, Samuli and Lehtinen, Jaakko and Aila, Timo},
  journal = {Advances in Neural Information Processing Systems},
  volume  = {32},
  year    = {2019}
}

@inproceedings{apd-original-paper,
  title     = {The unreasonable effectiveness of deep features as a perceptual metric},
  author    = {Zhang, Richard and Isola, Phillip and Efros, Alexei A and Shechtman, Eli and Wang, Oliver},
  booktitle = {Proceedings of the IEEE conference on computer vision and pattern recognition},
  pages     = {586--595},
  year      = {2018}
}

@inproceedings{mms-paper,
  title        = {Implicit neural representations for variable length human motion generation},
  author       = {Cervantes, Pablo and Sekikawa, Yusuke and Sato, Ikuro and Shinoda, Koichi},
  booktitle    = {European Conference on Computer Vision},
  pages        = {356--372},
  year         = {2022},
  organization = {Springer}
}

@article{dtw-paper,
  title     = {Dynamic time warping},
  author    = {M{\"u}ller, Meinard},
  journal   = {Information retrieval for music and motion},
  pages     = {69--84},
  year      = {2007},
  publisher = {Springer}
}

@article{aeon-paper,
  title   = {aeon: a Python toolkit for learning from time series},
  author  = {Middlehurst, Matthew and Ismail-Fawaz, Ali and Guillaume, Antoine and Holder, Christopher and Rubio, David Guijo and Bulatova, Guzal and Tsaprounis, Leonidas and Mentel, Lukasz and Walter, Martin and Sch{\"a}fer, Patrick and others},
  journal = {arXiv preprint arXiv:2406.14231},
  year    = {2024}
}

@misc{keras-website,
  title        = {Keras},
  author       = {Chollet, Fran\c{c}ois and others},
  year         = {2015},
  howpublished = {\url{https://keras.io}}
}

@inproceedings{deep-averaging-paper,
  title        = {Time series averaging using multi-tasking autoencoder},
  author       = {Terefe, Tsegamlak and Devanne, Maxime and Weber, Jonathan and Hailemariam, Dereje and Forestier, Germain},
  booktitle    = {2020 IEEE 32nd International Conference on Tools with Artificial Intelligence (ICTAI)},
  pages        = {1065--1072},
  year         = {2020},
  organization = {IEEE}
}

@article{anomaly-detection-review,
  title     = {Anomaly detection in time series: a comprehensive evaluation},
  author    = {Schmidl, Sebastian and Wenig, Phillip and Papenbrock, Thorsten},
  journal   = {Proceedings of the VLDB Endowment},
  volume    = {15},
  number    = {9},
  pages     = {1779--1797},
  year      = {2022},
  publisher = {VLDB Endowment}
}

@article{deep-domain-adaptation-review,
  title   = {Deep Unsupervised Domain Adaptation for Time Series Classification: a Benchmark},
  author  = {Fawaz, Hassan Ismail and Del Grosso, Ganesh and Kerdoncuff, Tanguy and Boisbunon, Aurelie and Saffar, Illyyne},
  journal = {arXiv preprint arXiv:2312.09857},
  year    = {2023}
}

@misc{ismail-fawaz2024elastic-vis,
  author       = {Ismail-Fawaz, Ali and Devanne, Maxime and Berretti, Stefano and Weber, Jonathan and Forestier, Germain},
  title        = {Elastic Warping Visualization For Time Series},
  year         = {2024},
  publisher    = {Github},
  journal      = {GitHub repository},
  howpublished = {\url{https://github.com/yourusername/dtw-visualization}}
}

@article{matplotlib-paper,
  title     = {Matplotlib: A 2D graphics environment},
  author    = {Hunter, John D},
  journal   = {Computing in science \& engineering},
  volume    = {9},
  number    = {03},
  pages     = {90--95},
  year      = {2007},
  publisher = {IEEE Computer Society}
}

@misc{bokeh,
  title        = {Bokeh: Python library for interactive visualization},
  author       = {{Bokeh Development Team}},
  year         = {2018},
  publisher    = {Github},
  howpublished = {\url{https://bokeh.pydata.org/en/latest/}}
}

@inproceedings{ecg-example-paper,
  title        = {A generative modeling approach to limited channel ECG classification},
  author       = {Rajan, Deepta and Thiagarajan, Jayaraman J},
  booktitle    = {2018 40th Annual International Conference of the IEEE Engineering in Medicine and Biology Society (EMBC)},
  pages        = {2571--2574},
  year         = {2018},
  organization = {IEEE}
}

@article{eeg-example-paper,
  title     = {Use of features from RR-time series and EEG signals for automated classification of sleep stages in deep neural network framework},
  author    = {Tripathy, RK and Acharya, U Rajendra},
  journal   = {Biocybernetics and Biomedical Engineering},
  volume    = {38},
  number    = {4},
  pages     = {890--902},
  year      = {2018},
  publisher = {Elsevier}
}

@article{human-motion-example-paper,
  title     = {3-d human action recognition by shape analysis of motion trajectories on riemannian manifold},
  author    = {Devanne, Maxime and Wannous, Hazem and Berretti, Stefano and Pala, Pietro and Daoudi, Mohamed and Del Bimbo, Alberto},
  journal   = {IEEE transactions on cybernetics},
  volume    = {45},
  number    = {7},
  pages     = {1340--1352},
  year      = {2014},
  publisher = {IEEE}
}

@article{stock-exchange-example-paper,
  title     = {Time series trend detection and forecasting using complex network topology analysis},
  author    = {Anghinoni, Leandro and Zhao, Liang and Ji, Donghong and Pan, Heng},
  journal   = {Neural Networks},
  volume    = {117},
  pages     = {295--306},
  year      = {2019},
  publisher = {Elsevier}
}

@article{telecom-example-paper,
  title     = {Forecasting in telecommunications and ICT—A review},
  author    = {Meade, Nigel and Islam, Towhidul},
  journal   = {International Journal of Forecasting},
  volume    = {31},
  number    = {4},
  pages     = {1105--1126},
  year      = {2015},
  publisher = {Elsevier}
}

@inproceedings{image-countour-shapelets-paper,
  title     = {Time series shapelets: a new primitive for data mining},
  author    = {Ye, Lexiang and Keogh, Eamonn},
  booktitle = {Proceedings of the 15th ACM SIGKDD international conference on Knowledge discovery and data mining},
  pages     = {947--956},
  year      = {2009}
}

@article{data-mining-challenges-paper,
  title     = {10 challenging problems in data mining research},
  author    = {Yang, Qiang and Wu, Xindong},
  journal   = {International Journal of Information Technology \& Decision Making},
  volume    = {5},
  number    = {04},
  pages     = {597--604},
  year      = {2006},
  publisher = {World Scientific}
}

@article{monash-forecasting,
  title   = {Monash time series forecasting archive},
  author  = {Godahewa, Rakshitha and Bergmeir, Christoph and Webb, Geoffrey I and Hyndman, Rob J and Montero-Manso, Pablo},
  journal = {arXiv preprint arXiv:2105.06643},
  year    = {2021}
}

@article{weather-forecasting,
  title     = {Transductive LSTM for time-series prediction: An application to weather forecasting},
  author    = {Karevan, Zahra and Suykens, Johan AK},
  journal   = {Neural Networks},
  volume    = {125},
  pages     = {1--9},
  year      = {2020},
  publisher = {Elsevier}
}

@article{deep-live-fuel,
  title     = {Multi-modal temporal CNNs for live fuel moisture content estimation},
  author    = {Miller, Lynn and Zhu, Liujun and Yebra, Marta and R{\"u}diger, Christoph and Webb, Geoffrey I},
  journal   = {Environmental Modelling \& Software},
  volume    = {156},
  pages     = {105467},
  year      = {2022},
  publisher = {Elsevier}
}

@inproceedings{network-anomaly-detection,
  title        = {Network-based intrusion detection with support vector machines},
  author       = {Kim, Dong Seong and Park, Jong Sou},
  booktitle    = {Information Networking: International Conference, ICOIN 2003, Cheju Island, Korea, February 12-14, 2003. Revised Selected Papers},
  pages        = {747--756},
  year         = {2003},
  organization = {Springer}
}

@inproceedings{healthcare-anomaly-detection,
  title        = {Learning representations from healthcare time series data for unsupervised anomaly detection},
  author       = {Pereira, Joao and Silveira, Margarida},
  booktitle    = {2019 IEEE international conference on big data and smart computing (BigComp)},
  pages        = {1--7},
  year         = {2019},
  organization = {IEEE}
}

@article{timeseries-clustering-application,
  title     = {Forecasting sales through time series clustering},
  author    = {Sanwlani, Monica and Vijayalakshmi, M},
  journal   = {International Journal of Data Mining \& Knowledge Management Process},
  volume    = {3},
  number    = {1},
  pages     = {39},
  year      = {2013},
  publisher = {Academy \& Industry Research Collaboration Center (AIRCC)}
}

@inproceedings{wind-clustering,
  title        = {Wind power forecasting using time series cluster analysis},
  author       = {Pravilovic, Sonja and Appice, Annalisa and Lanza, Antonietta and Malerba, Donato},
  booktitle    = {Discovery Science: 17th International Conference, DS 2014, Bled, Slovenia, October 8-10, 2014. Proceedings 17},
  pages        = {276--287},
  year         = {2014},
  organization = {Springer}
}

@article{deep-tsc-tser,
  title     = {Deep learning for time series classification and extrinsic regression: A current survey},
  author    = {Mohammadi Foumani, Navid and Miller, Lynn and Tan, Chang Wei and Webb, Geoffrey I and Forestier, Germain and Salehi, Mahsa},
  journal   = {ACM Computing Surveys},
  volume    = {56},
  number    = {9},
  pages     = {1--45},
  year      = {2024},
  publisher = {ACM New York, NY}
}

@article{foundation-models-example-paper,
  title   = {Moment: A family of open time-series foundation models},
  author  = {Goswami, Mononito and Szafer, Konrad and Choudhry, Arjun and Cai, Yifu and Li, Shuo and Dubrawski, Artur},
  journal = {arXiv preprint arXiv:2402.03885},
  year    = {2024}
}

@inproceedings{weighted-shape-dba-paper,
  author    = {Ismail-Fawaz, Ali and Devanne, Maxime and Berretti, Stefano and Weber, Jonathan and Forestier, Germain},
  title     = {Weighted Average of Human Motion Sequences for Improving Rehabilitation Assessment},
  booktitle = {ECML/PKDD Workshop on Advanced Analytics and Learning on Temporal Data},
  city      = {Vilnius},
  country   = {Lithuania},
  year      = {2024}
}

@article{metrics-paper,
  title   = {Establishing a Unified Evaluation Framework for Human Motion Generation: A Comparative Analysis of Metrics},
  author  = {Ismail-Fawaz, Ali and Devanne, Maxime and Berretti, Stefano and Weber, Jonathan and Forestier, Germain},
  journal = {arXiv preprint arXiv:2405.07680},
  year    = {2024}
}

@article{lite-extension-paper,
  title={Look Into the LITE in Deep Learning for Time Series Classification},
  author={Ismail-Fawaz, Ali and Devanne, Maxime and Berretti, Stefano and Weber, Jonathan and Forestier, Germain},
  journal={arXiv preprint arXiv:2409.02869},
  year={2024}
}

@inproceedings{trilite-paper,
  title        = {Enhancing time series classification with self-supervised learning},
  author       = {Ismail-Fawaz, Ali and Devanne, Maxime and Weber, Jonathan and Forestier, Germain},
  booktitle    = {International Conference on Agents and Artificial Intelligence (ICAART)},
  pages        = {40--47},
  year         = {2023},
  organization = {SCITEPRESS-Science and Technology Publications}
}

@misc{Ismail-Fawaz2023weighted-ba,
  title        = {Weighted Elastic Barycetner Averaging to Augment Time Series Data},
  author       = {Ismail-Fawaz, Ali and Devanne, Maxime and Berretti, Stefano and Weber, Jonathan and Forestier, Germain},
  year         = {2024},
  publisher    = {Github},
  journal      = {GitHub repository},
  howpublished = {\url{https://github.com/MSD-IRIMAS/Augmenting-TSC-Elastic-Averaging}}
}

@misc{Ismail-Fawaz2023kan-c22-4-tsc,
  title        = {Feature-Based Time Series Classification with Kolmogorov-Arnold Networks},
  author       = {Ismail-Fawaz, Ali and Devanne, Maxime and Berretti, Stefano and Weber, Jonathan and Forestier, Germain},
  year         = {2024},
  publisher    = {Github},
  journal      = {GitHub repository},
  howpublished = {\url{https://github.com/MSD-IRIMAS/Simple-KAN-4-Time-Series}}
}

@article{kan-paper,
  title   = {Kan: Kolmogorov-arnold networks},
  author  = {Liu, Ziming and Wang, Yixuan and Vaidya, Sachin and Ruehle, Fabian and Halverson, James and Solja{\v{c}}i{\'c}, Marin and Hou, Thomas Y and Tegmark, Max},
  journal = {arXiv preprint arXiv:2404.19756},
  year    = {2024}
}

@article{ts-chief-paper,
  title     = {TS-CHIEF: a scalable and accurate forest algorithm for time series classification},
  author    = {Shifaz, Ahmed and Pelletier, Charlotte and Petitjean, Fran{\c{c}}ois and Webb, Geoffrey I},
  journal   = {Data Mining and Knowledge Discovery},
  volume    = {34},
  number    = {3},
  pages     = {742--775},
  year      = {2020},
  publisher = {Springer}
}

@inproceedings{rise,
  title        = {The contract random interval spectral ensemble (c-RISE): The effect of contracting a classifier on accuracy},
  author       = {Flynn, Michael and Large, James and Bagnall, Tony},
  booktitle    = {Hybrid Artificial Intelligent Systems: 14th International Conference, HAIS 2019, Le{\'o}n, Spain, September 4--6, 2019, Proceedings 14},
  pages        = {381--392},
  year         = {2019},
  organization = {Springer}
}

@book{pythagorean-theorem-paper,
  title     = {The Pythagorean theorem: a 4,000-year history},
  author    = {Maor, Eli},
  volume    = {65},
  year      = {2019},
  publisher = {Princeton University Press}
}

@article{lb-keogh-paper,
  title={Exact indexing of dynamic time warping},
  author={Keogh, Eamonn and Ratanamahatana, Chotirat Ann},
  journal={Knowledge and information systems},
  volume={7},
  pages={358--386},
  year={2005},
  publisher={Springer}
}

@article{lb-webb-paper,
  title={Tight lower bounds for dynamic time warping},
  author={Webb, Geoffrey I and Petitjean, Fran{\c{c}}ois},
  journal={Pattern Recognition},
  volume={115},
  pages={107895},
  year={2021},
  publisher={Elsevier}
}

@book{svm-paper,
  title={The nature of statistical learning theory},
  author={Vapnik, Vladimir},
  year={2013},
  publisher={Springer science \& business media}
}

@article{time-series-snippets,
  title={Introducing time series snippets: a new primitive for summarizing long time series},
  author={Imani, Shima and Madrid, Frank and Ding, Wei and Crouter, Scott E and Keogh, Eamonn},
  journal={Data Mining and Knowledge Discovery},
  volume={34},
  pages={1713--1743},
  year={2020},
  publisher={Springer}
}

@article{transfer-learning-forecasting,
  title={The impact of data set similarity and diversity on transfer learning success in time series forecasting},
  author={Ehrig, Claudia and Cleophas, Catherine and Forestier, Germain},
  journal={arXiv preprint arXiv:2404.06198},
  year={2024}
}

@article{iot-tsc,
  title={Robust IoT time series classification with data compression and deep learning},
  author={Azar, Joseph and Makhoul, Abdallah and Couturier, Rapha{\"e}l and Demerjian, Jacques},
  journal={Neurocomputing},
  volume={398},
  pages={222--234},
  year={2020},
  publisher={Elsevier}
}

@article{self-sup-anom-detection,
  title={CARLA: Self-supervised contrastive representation learning for time series anomaly detection},
  author={Darban, Zahra Zamanzadeh and Webb, Geoffrey I and Pan, Shirui and Aggarwal, Charu C and Salehi, Mahsa},
  journal={Pattern Recognition},
  volume={157},
  pages={110874},
  year={2025},
  publisher={Elsevier}
}

@inproceedings{self-sup-forecasting,
  title={Self-Supervised Contrastive Learning for Long-term Forecasting},
  author={Park, Junwoo and Gwak, Daehoon and Choo, Jaegul and Choi, Edward},
  booktitle={The Twelfth International Conference on Learning Representations},
  year={2024}
}

@article{series2vec,
  title={Series2vec: similarity-based self-supervised representation learning for time series classification},
  author={Foumani, Navid Mohammadi and Tan, Chang Wei and Webb, Geoffrey I and Rezatofighi, Hamid and Salehi, Mahsa},
  journal={Data Mining and Knowledge Discovery},
  pages={1--25},
  year={2024},
  publisher={Springer}
}

@article{marteau2014recursive,
  title={On recursive edit distance kernels with application to time series classification},
  author={Marteau, Pierre-Fran{\c{c}}ois and Gibet, Sylvie},
  journal={IEEE transactions on neural networks and learning systems},
  volume={26},
  number={6},
  pages={1121--1133},
  year={2014},
  publisher={IEEE}
}

@inproceedings{cuturi2007kernel,
  title={A kernel for time series based on global alignments},
  author={Cuturi, Marco and Vert, Jean-Philippe and Birkenes, Oystein and Matsui, Tomoko},
  booktitle={2007 IEEE International Conference on Acoustics, Speech and Signal Processing-ICASSP'07},
  volume={2},
  pages={II--413},
  year={2007},
  organization={IEEE}
}

@article{marteau2019times,
  title={Times series averaging and denoising from a probabilistic perspective on time-elastic kernels},
  author={Marteau, Pierre-Francois},
  journal={International Journal of Applied Mathematics and Computer Science},
  volume={29},
  number={2},
  year={2019}
}

@article{marteau2019separation,
  title={On the separation of shape and temporal patterns in time series-Application to signature authentication},
  author={Marteau, Pierre-Fran{\c{c}}ois},
  journal={arXiv preprint arXiv:1911.09360},
  year={2019}
}



\appendix 



\end{document}